\documentclass{article}

\usepackage{arxiv}

\usepackage[utf8]{inputenc} 
\usepackage[T1]{fontenc}    
\usepackage{hyperref}       
\usepackage{url}            
\usepackage{booktabs}       
\usepackage{amsfonts}       
\usepackage{nicefrac}       
\usepackage{microtype}      
\usepackage{lipsum}		
\usepackage{graphicx}
\usepackage{natbib}
\usepackage{doi}

\usepackage{subcaption}
\usepackage{amsmath}
\usepackage{xcolor}
\usepackage{color}
\usepackage{verbatim}
\usepackage{mathabx}
\usepackage{tikz}
\usepackage{multirow}
\usepackage{floatrow} 

\usepackage{pifont}

\usepackage[noend]{algpseudocode}
\usepackage{algorithm,algorithmicx}  

\usepackage[toc,page]{appendix}

\algdef{SE}[SUBALG]{Indent}{EndIndent}{}{\algorithmicend\ }%
\algtext*{Indent}
\algtext*{EndIndent}

\begin{document}

\title{Tight basis cycle representatives for persistent homology of large data sets}

\author{Manu Aggarwal\thanks{Corresponding author}\\
       \texttt{manu.aggarwal@nih.gov}  \\
       Laboratory of Biological Modeling, NIDDK\\
       National Insitutes of Health\\
       31 Center Dr, Bethesda, MD 20892
       \AND
       Vipul Periwal\\
       \texttt{vipulp@niddk.nih.gov} \\
       Laboratory of Biological Modeling, NIDDK\\
       National Insitutes of Health\\
       31 Center Dr, Bethesda, MD 20892
       }

\maketitle

\begin{abstract}

Persistent homology (PH) is a popular tool for topological data analysis that has found
  applications across diverse areas of research. It provides a rigorous method to compute robust
  topological features in discrete experimental observations that often contain various sources of
  uncertainties. Although powerful in theory, PH suffers from high computation cost that
  precludes its application to large data sets. Additionally, most analyses using PH are limited to
  computing the existence of nontrivial features. Precise localization of these features is not generally attempted because, by definition, localized representations are not unique and because of even higher computation cost. For scientific applications, such a precise location is a sine qua non for determining functional significance. Here, we provide a strategy and algorithms to compute tight representative
  boundaries around nontrivial robust features in large data sets. To showcase the efficiency of our
  algorithms and the precision of computed boundaries, we analyze three data sets from
  different scientific fields.
  In the human genome, we found an unexpected effect on loops through chromosome 13 and the sex chromosomes, upon impairment of chromatin loop formation. In a distribution of galaxies in the universe, we found statistically significant voids. In protein homologs with significantly different topology, we found voids attributable to ligand-interaction, mutation, and differences between species.

\end{abstract}

\section{Introduction}

Quantitative observations of physical systems from galaxies to proteins often lead to data sets that are discrete and have uncertainties~\citep{klir2006}. The identification of robust features in such data sets is the impetus for the development of topological data analysis (TDA). Persistent homology (PH) is, perhaps, the most mathematically rigorous method developed for TDA. PH applies techniques developed in algebraic topology to find robust lacunae in discrete data sets. Most implementations of PH end their analysis at the stage where the existence of nontrivial homology cycles has been demonstrated. However, scientific interest in nontrivial cycles often requires finding specific locations for homology cycles. For example, while the existence of voids in the distribution of galaxies is interesting, cosmologists need to know the locations and sizes of these voids to compare them to theoretical models of galaxy formation. As another example, the looping of chromatin strands of the genome is not random. It regulates gene expression by bringing regulators close to regulated genes that otherwise are far along the linear
strands~\citep{kadauke2009chromatin,rowley2018organizational}. Hence, determining the precise genomic location
of loops is important. PH has found useful
applications in areas as diverse as neuroscience~\citep{bendich2016persistent}, computational
biology~\citep{mcguirl2020topological}, natural language processing~\citep{zhu2013persistent}, the
spread of contagions~\citep{taylor2015topological}, cancer~\citep{nicolau2011topology,
lawson2019persistent}, material science~\citep{kramar2013persistence}, computer
graphics~\citep{bruel2020topology}, among many others. However, its application is still limited
because of the challenges of high computation cost and identifying the precise locations of features.

\begin{figure}[tbhp!]
      \centering
      \begin{subfigure}{\textwidth} \centering
        \includegraphics[width=0.95\linewidth]{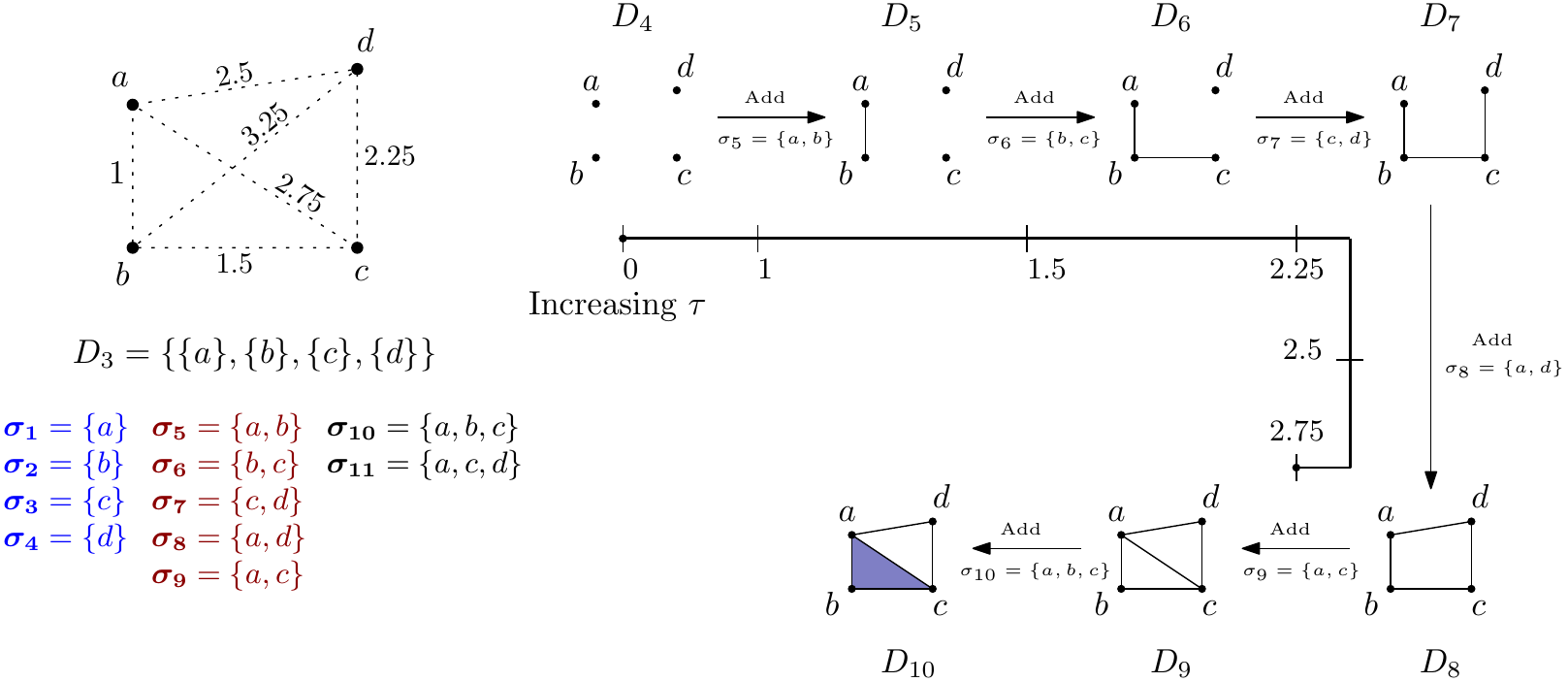}
        \caption{}
        \label{fig:PH_filtration}
      \end{subfigure}
      \centering
      \begin{subfigure}{0.33\textwidth} \centering
        \includegraphics[width=0.95\linewidth]{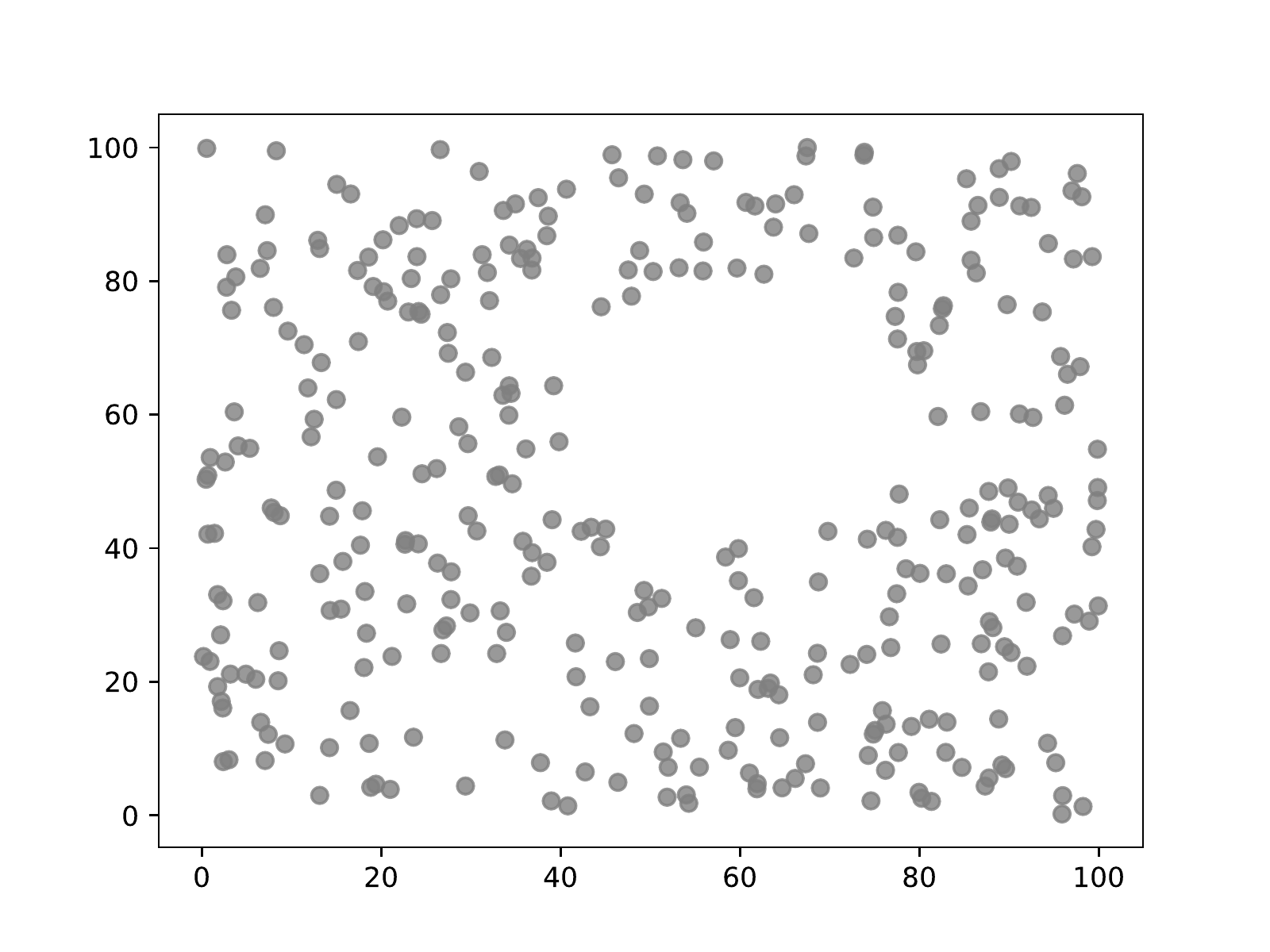}
        \caption{}
        \label{fig:example_2d_scatter}
      \end{subfigure}
      \centering
      \begin{subfigure}{0.33\textwidth} \centering
        \includegraphics[width=0.95\linewidth]{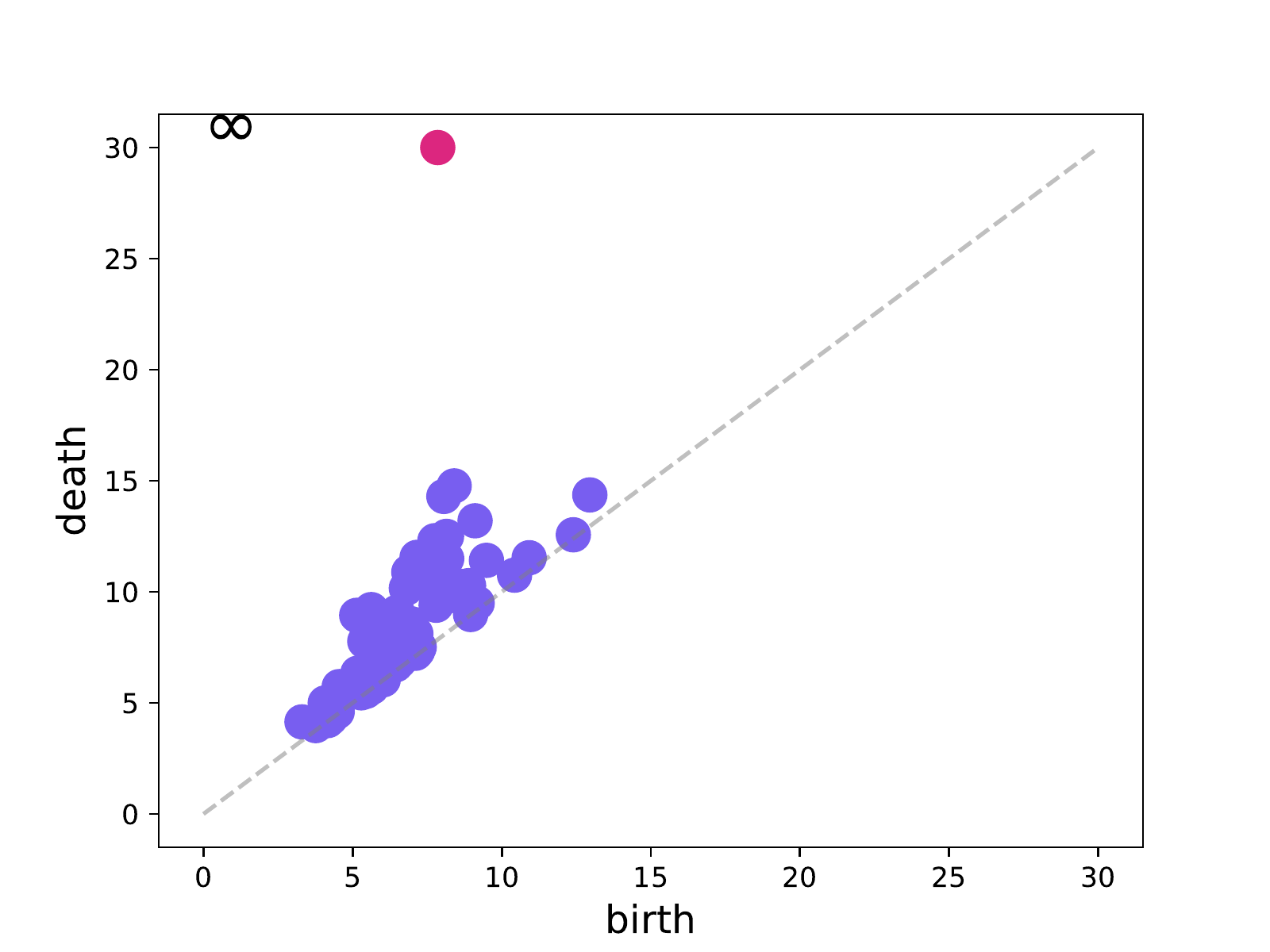}
        \caption{}
        \label{fig:example_2d_PD}
      \end{subfigure}
      \centering
      \begin{subfigure}{0.33\textwidth} \centering
        \includegraphics[width=0.95\linewidth]{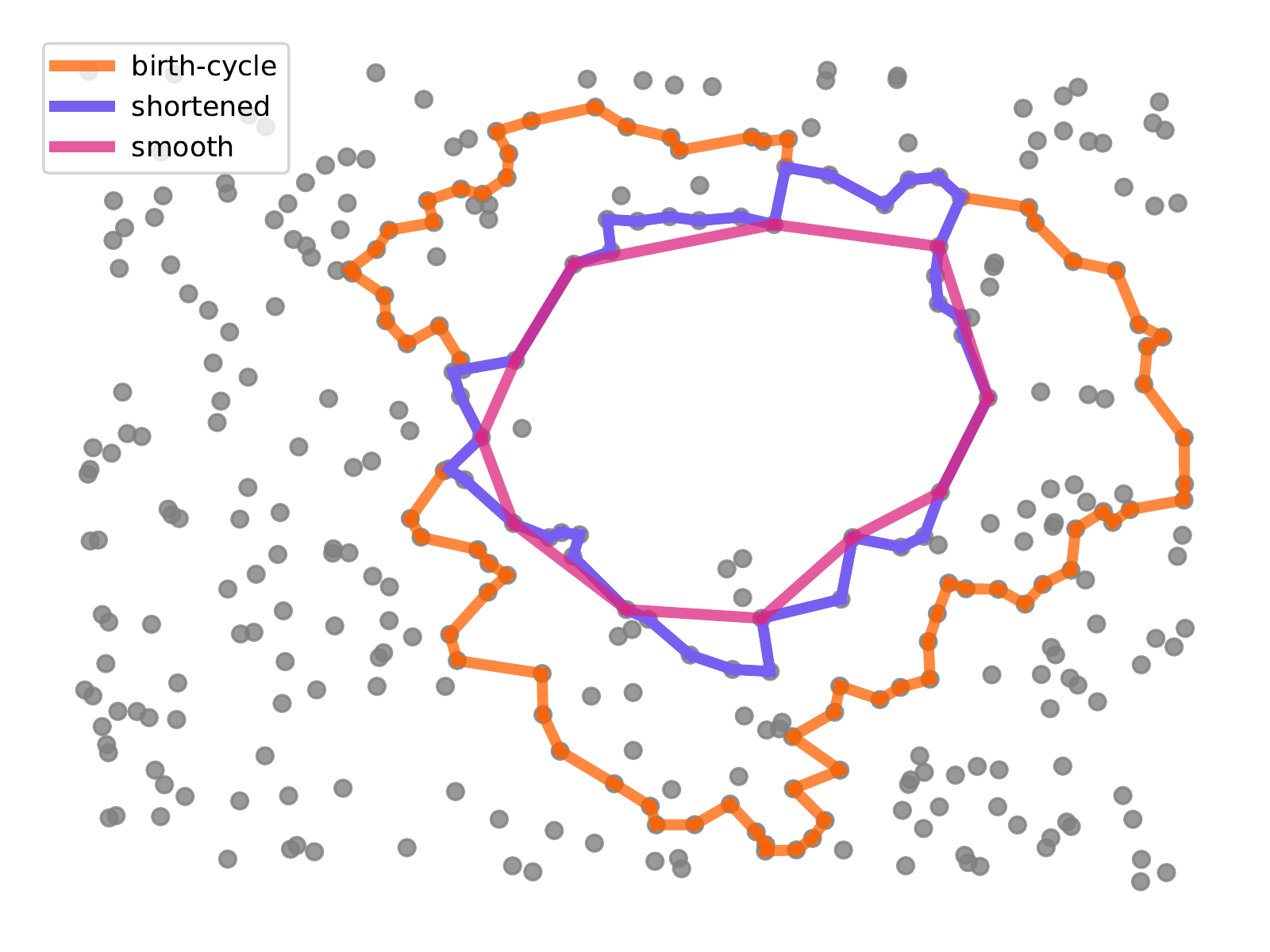}
        \caption{}
        \label{fig:example_2d_cycles}
      \end{subfigure}
      \centering
      \begin{subfigure}{0.28\textwidth} \centering
        \includegraphics[width=0.95\linewidth]{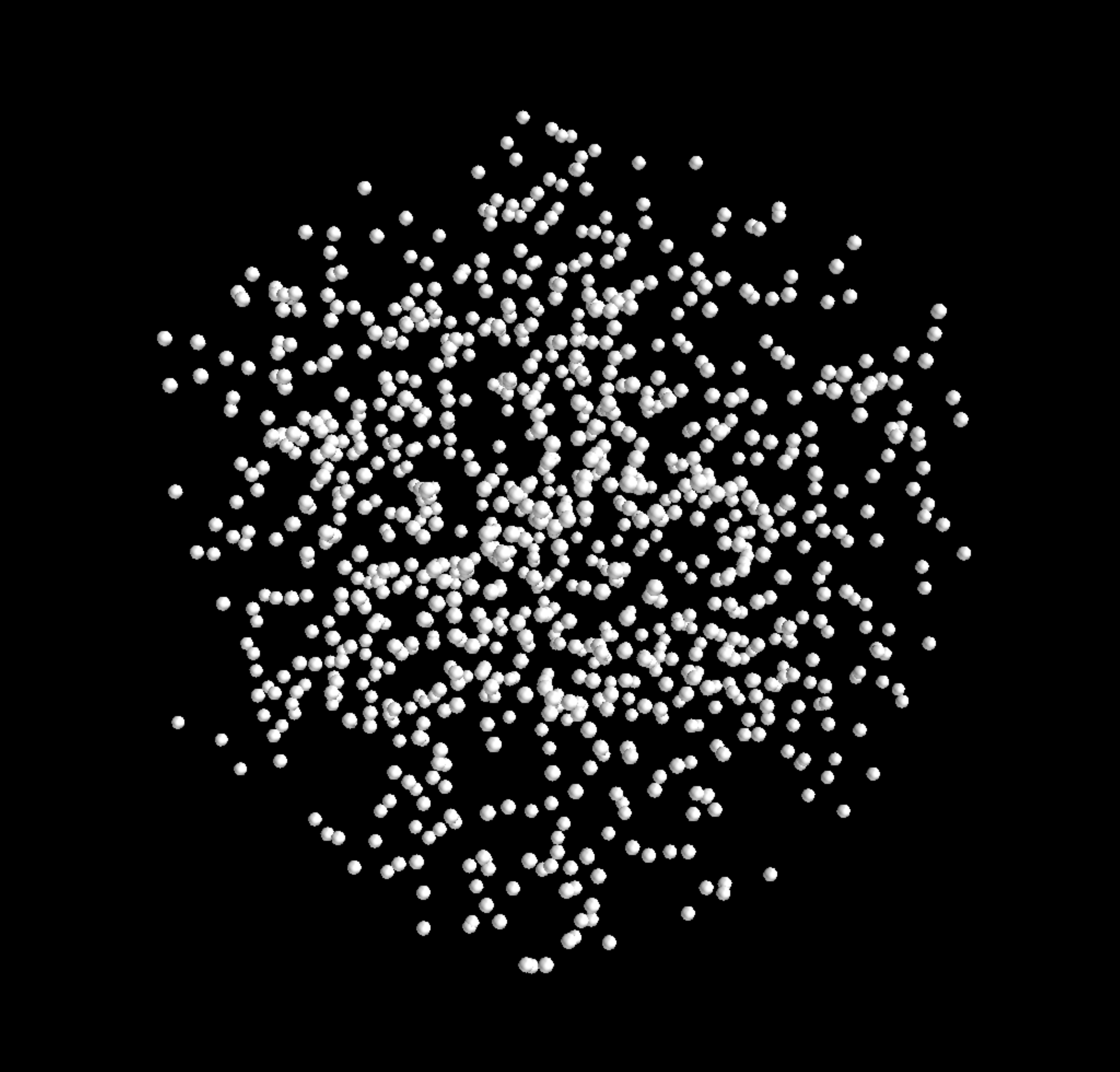}
        \caption{}
        \label{fig:example_3d_scatter}
      \end{subfigure}
      \centering
      \begin{subfigure}{0.43\textwidth} \centering
        \includegraphics[width=0.95\linewidth]{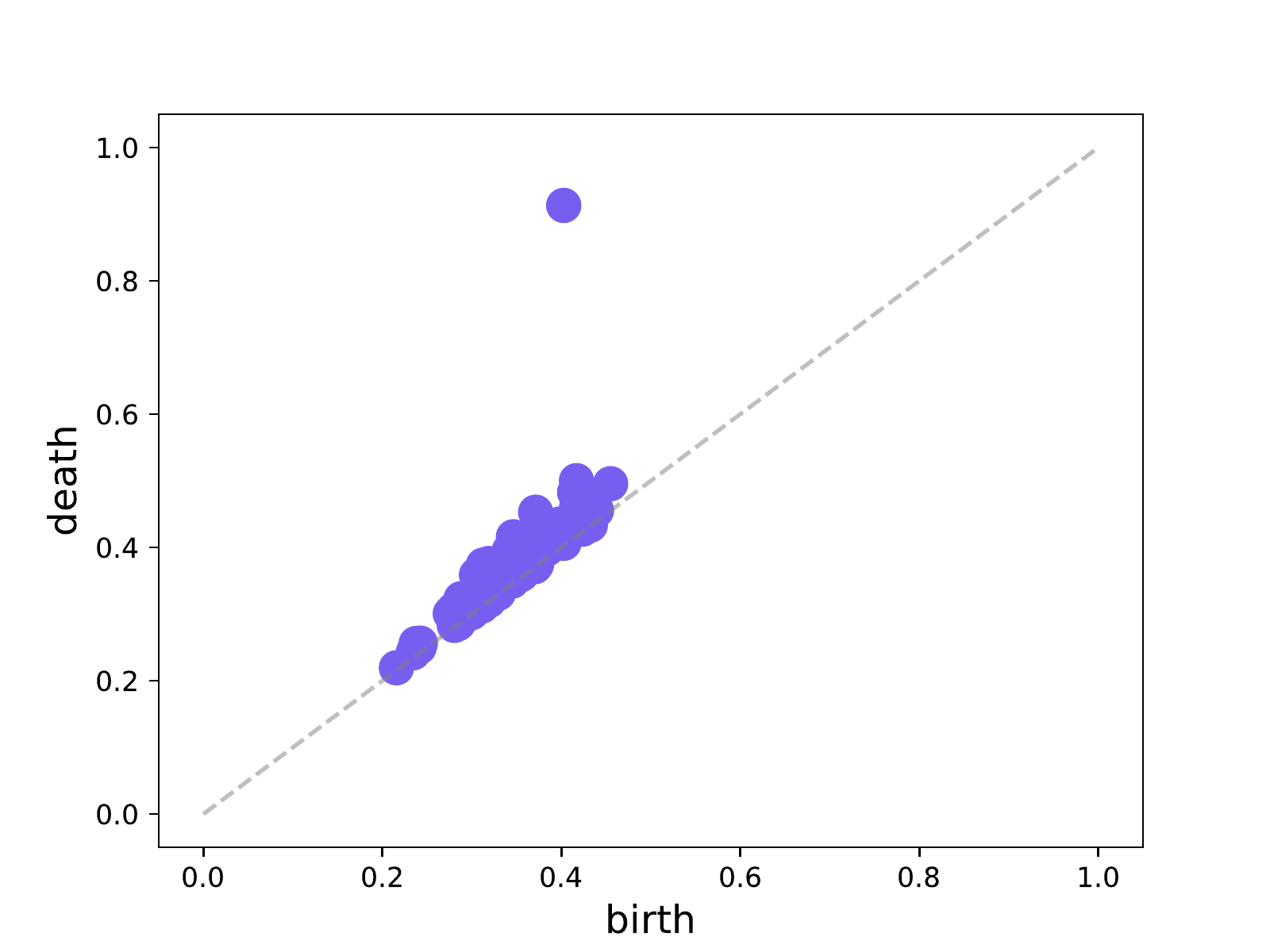}
        \caption{}
        \label{fig:example_3d_PD}
      \end{subfigure}
      \centering
      \begin{subfigure}{0.28\textwidth} \centering
        \includegraphics[width=0.95\linewidth]{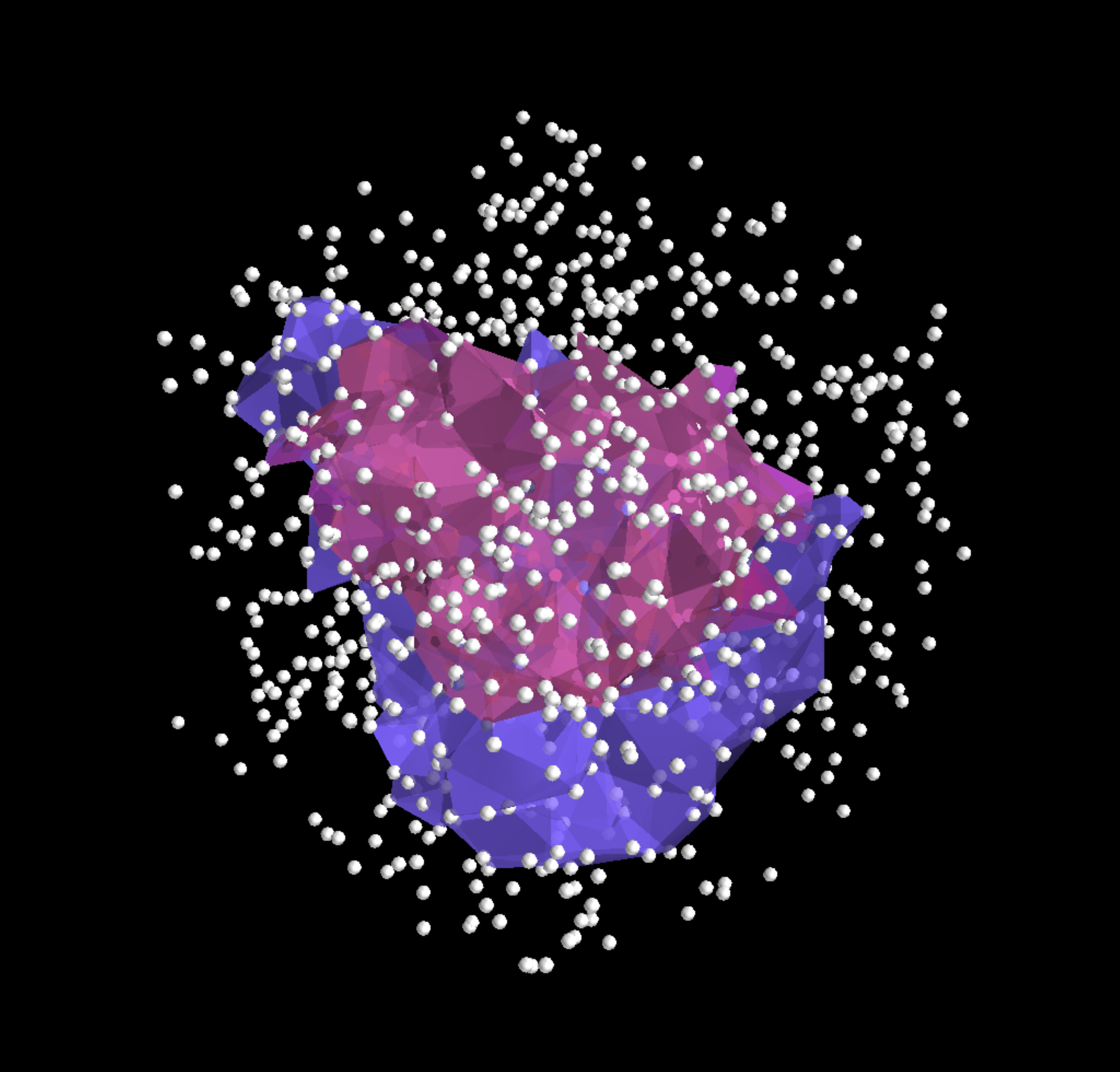}
        \caption{}
        \label{fig:example_3d_birth_smooth}
      \end{subfigure}

      \caption{PH finds robust holes in discrete data sets and our algorithms improve geometric
      precision of representative boundaries. (a) A data-set with 4 points and annotated pairwise
      distances. Simplicial complex ($D_i$) at a spatial scale $\tau$ is defined as a set of
      simplices ($\sigma_i$) with diameter at most $\tau.$ At $\tau=2.5,$ \#holes in simplicial
      complex increases from $0$ to $1$ when $\sigma_8$ is added ($D_8$). At $\tau=2.75$ \#holes
      increase $1$ to $2$ when $\sigma_9$ is added ($D_9$). One hole gets filled in at $\tau=2.75,$
      when $\sigma_{10} = \{a,b,c\}$ is added ($D_{10}$), but one remains alive. (b) A noisy
      discrete data set in $\mathbb{R}^2$ with one significant hole. (c) Birth and death of holes is
      plotted as persistence diagram. It shows that one hole stands out with relatively high
      persistence. (d) Our algorithms improve geometric precision of representative boundaries in
      multiple steps. Birth-cycle is from recursive algorithm. Shortened is from greedy shortening
      algorithm. Smooth cycle is from smoothing algorithm. (e) A point-cloud in $\mathbb{R}^3.$ It
      is difficult to identify any significant hole by visual inspection. (f) PD shows that one
      feature has relatively high persistence. (g) Our algorithms again improve geometric precision.
      Blue boundary is birth-cycle from recursive algorithm. Red boundary is smooth cycle after
      greedy shortening and smoothing algorithms.}

      \label{fig:first_cycle_examples}

\end{figure}

Colloquially, we can think of the features that PH finds as `holes' in the cloud of points that constitute a data set, regions of lower density enclosed in regions of higher density. As one can surmise from the appearance of the term `density', the features that PH finds are dependent on the scale of distances between neighboring data points. The use of an objective mathematically sound method to compute the existence of holes is called for because the human eye is adept at pattern detection even when there is no pattern. For example, we can readily pick out a chain in a uniform distribution of constituents in a three-dimensional
space~\citep{soneira1978computer}. On the other hand, we may be unable to find topologically significant
holes in a 3D point-cloud by visual inspection (Figure~\ref{fig:example_3d_scatter}). 


We briefly introduce terminology and background to explain the challenges in applying PH. 
A $n$-simplex is a set of $(n+1)$ points (Supplementary
figure~\ref{fig:supp_simplices}). 
A simplicial complex
at a spatial scale of $\tau$ is defined as the set of all possible simplices with maximal 
pairwise distance at most $\tau$ between their points.  Hence, we get a sequence
of complexes as $\tau$ increases (Figure~\ref{fig:PH_filtration}). Loops in a simplicial complex
are computed as basis elements of its homology group of dimension one (H$_1$). Voids are computed
as basis elements of its homology group of dimension two (H$_2$). PH tracks changes in the number of
basis elements of homology groups across the sequences of complexes. The $\tau$ values of the births and deaths of
topological features or holes are recorded and plotted as a persistence diagram (PD) (Figure~\ref{fig:example_2d_PD}). We refer to these $\tau$ values as births and deaths for brevity in this paper. Every hole is classified by the range of values of $\tau$ across
which it persists, called its \textit{persistence} or \textit{barcode} length and computed as
$\text{death} - \text{birth}.$ 
The key point is that the features with
relatively high persistence are robust to larger experimental variability.

This multiscale construction and processing of high-dimensional simplices (0-, 1-, 2-, and
3-simplices in this work) incurs a high computation cost. Test data sets used to benchmark PD
computation are commonly limited to a few thousand points. Some of the different methods to compute
PD are matrix reduction~\citep{edelsbrunner2008persistent}, Morse
theory~\citep{mischaikow2013morse}, and matroid theory~\citep{henselman2016matroid}. Ripser is one
of the most efficient software packages computing PD~\citep{bauer2021ripser}. 
However, even Ripser was unable to process human genome data sets because they contain millions of
discrete points. In previous work, we developed Dory~\citep{aggarwal2021dory} and used it to compute the PD of human genome
data sets within five minutes using only 6 GB of memory. We also showed that its efficiency is
not limited to these specific data sets since it used the least memory and run-time in almost all
test data sets. 
A different approach to
tackle computational feasibility of PD is to approximate a topological structure with a large number
of simplices by one with fewer simplices~\citep{dey2019simba}. However, theorems justifying such a coarsening so as to not lose information about possibly functionally important features are hard to come by and adding sampling stochasticity to an approach that is trying to overcome data uncertainty is counter-intuitive.
Most applications of PH are limited to computing PD 
and thereby demonstrating the existence and persistence of holes, but not their locations, and therefore do not allow an assessment of their functional significance. Why is the location information elided? By definition, representative boundaries around topological features
are not uniquely defined~\citep{carlsson2009topology} so the resulting boundaries can be
geometrically imprecise (Figure~\ref{fig:example_2d_cycles}). Furthermore, computing candidate representative boundaries is expensive in both memory usage and run time. Hence, the development of
efficient and scalable algorithms to compute PD and precise representative boundaries of persistent
features are active areas of research.

One approach to compute a canonical set of representative boundaries has been to define minimal
boundaries as solutions to an optimization problem. For example, a possible set of boundaries is one that
minimizes the aggregated weight of boundaries at different spatial scales~\citep{dey2011optimal}. 
Boundaries that are solutions
to such optimization problems are called \textit{optimal homologous cycles}. Strategies for their
efficient computation were introduced in~\cite{dey2018efficient}. However, computing optimal
homologous cycles still is of order $O\left (n^{11} \right)$ for a data set with $n$
points~\citep{guerra2021homological}. Hence, they are usually computed for small data sets. A
comparison of different possible optimization functions and implementations is discussed
in~\cite{li2021minimal}. Another issue is that they are not computed for voids. This is because it has
been shown that an optimization strategy for H$_2$ representative boundaries is infeasible
(NP-hard)~\citep{chen2011hardness}.

\begin{figure}[tbhp!]
      \centering
      \begin{subfigure}{0.55\textwidth} \centering
        \includegraphics[width=0.9\linewidth]{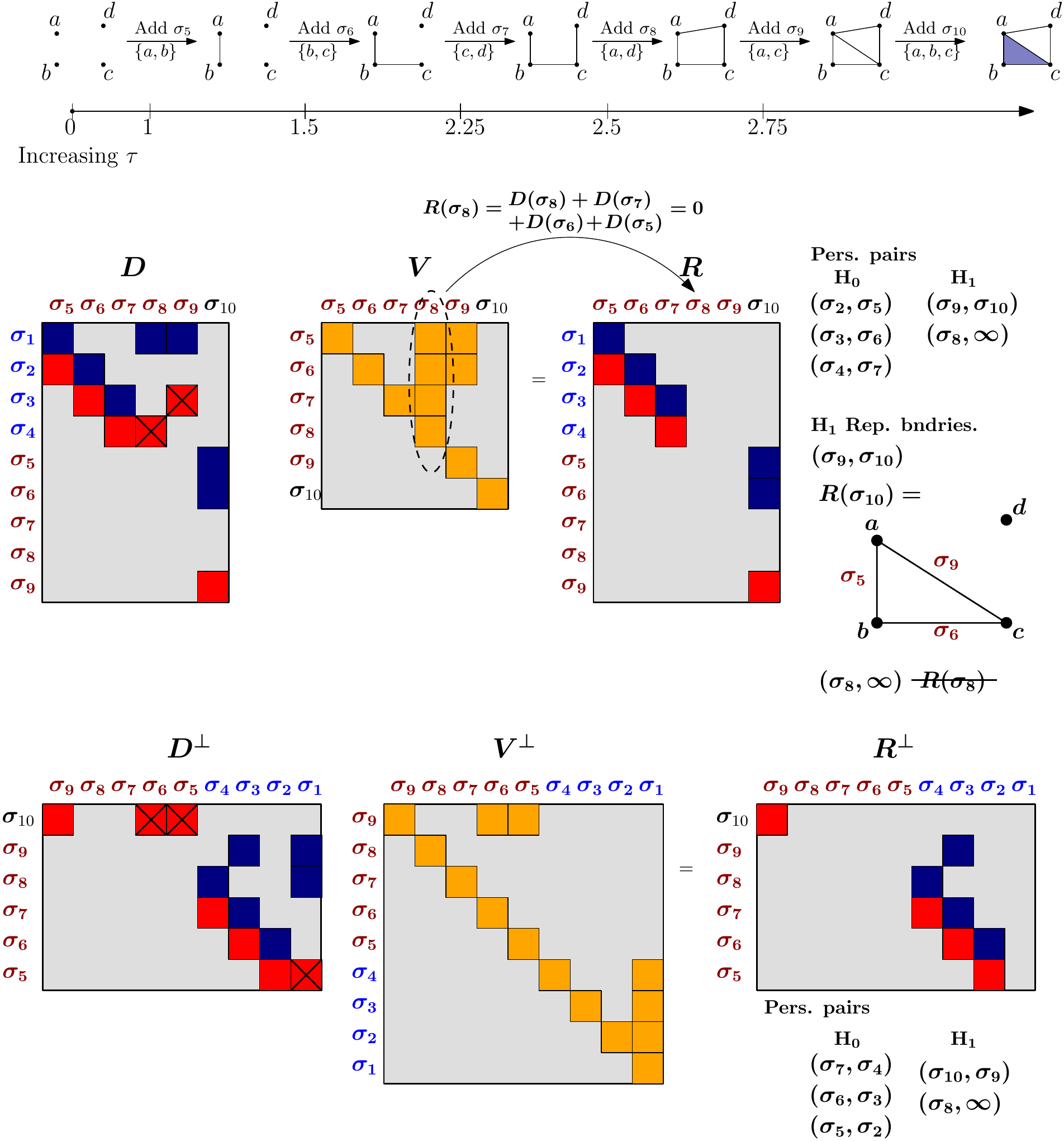}
        \caption{}
        \label{fig:PH_matrix_reduction}
      \end{subfigure}
      \centering
      \begin{subfigure}{.44\textwidth} \centering
        \includegraphics[width=0.9\linewidth]{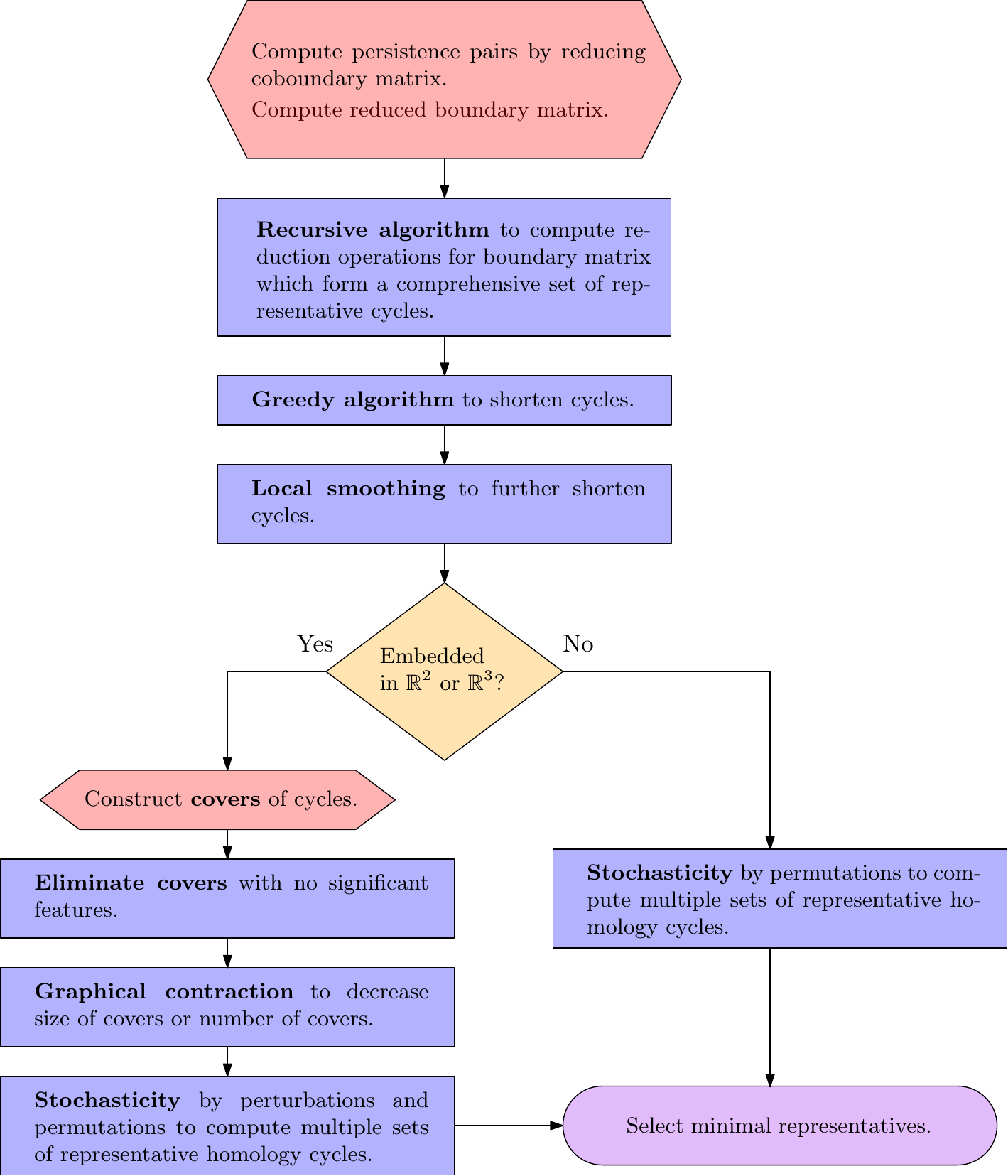}
        \caption{}
        \label{fig:flowchart}
      \end{subfigure}

  \caption{Matrix reduction algorithm and our strategy to compute representative boundaries in large
  data sets with improved precision. (a) Simplices are indexed by the order in which they are added
  to the simplicial complex. Column and row $i$ of boundary matrix $D$ correspond to simplex
  $\sigma_i.$ Columns for 0-simplices are not shown because they are always empty. Column
  $D(\sigma_i)$ has $1$ at boundary-simplices of simplex $\sigma_i,$ and $0$ otherwise. Lowest
  non-zero entry in each column is called a \textit{low}, shown as red box. Columns are reduced (mod
  2 sum) from left to right till each row has at most one red box, called a pivot. Pivot at $(i, j)$
  is a persistence pair $(\sigma_i, \sigma_j).$ If $R(\sigma_i) = \mathbf{0}$ and $\sigma_i$ is not
  in any persistence pair, then $(\sigma_i, \infty)$ is a feature that did not die. In the example,
  there are two H$_1$ persistence pairs---$(\sigma_9, \sigma_{10})$ and $(\sigma_{8}, \infty),$ or
  $(2.75, 2.75)$ and $(2.5, \infty).$   Columns of $R$ give a set of representative homology
  boundary. There is only one H$_1$ representative boundary, $R(\sigma_{10}),$ but there are two
  H$_1$ features. We propose to use columns of $V$ as a comprehensive set of representative
  boundaries, that we call birth-cycles. Coboundary matrix $D^\bot$ is off-diagonal transpose of
  $D.$ It is reduced similarly and same persistence pairs are obtained from $R^\bot.$ (b) Flowchart
  of our algorithms and strategy.}

   \label{fig:PH_example}
\end{figure}

Our contribution in this work is a set of algorithms that can compute minimal representative
boundaries around \textit{significant} loops and voids in large data sets. Figure~\ref{fig:flowchart} shows an outline of our strategy. We classify a hole as significant if it
has persistence larger than a threshold $\epsilon$ (a higher value means robust to larger variability
in the data set) and is born at a spatial scale of at most $\tau_u$ (a lower value suggests holes with
denser boundary). First, we developed a recursive algorithm that can compute a comprehensive set of
representative boundaries around all nontrivial holes (persistence > 0). This is not always
possible in extant algorithms because of high computation cost. We call these boundaries 
birth-cycles. Next, we developed a greedy algorithm that decreases lengths of birth-cycles to give a
set of shortened representative boundaries. Third, we implemented a local smoothing of the
boundaries. Figures~\ref{fig:example_2d_cycles} and~\ref{fig:example_3d_birth_smooth} show examples
of computed shortened and smoothed H$_1$ and H$_2$ representatives. Note how they tighten around
the respective holes when they are shortened. Fourth, the algorithm identifies sub-regions of
a spatial embedding of the data set that contain significant features so we can use a divide and conquer strategy. We show in examples
that this significantly decreases the  maximum size of data sets that need to be
processed for subsequent analysis. We also developed a graphical contraction that can possibly decrease
the number of sub-regions. Finally, we introduce stochasticity in two different ways that can help to overcome the greedy algorithm's local minimum problem. 
The computational efficiency of our algorithms is evidenced by the ability to process a human genome data set with three million points within a few minutes and
using around 3 GB of memory. We illustrate the geometrical precision of computed representative
boundaries for voids using case studies of the distribution of galaxies in universe and protein crystal structures.
Hence, this work enables scientific investigation of the functional significance of topologically robust
features in large noisy data sets from experimental observations.

\section{Results}

\subsection{Genome-wide high resolution loops in human genome computed within a minute}

We analyzed the human genome at high resolution to show scalability and application of our algorithms to
large data sets. This data set has approximately three million points. The 23 chromosomes in human cells collectively contain approximately 6.7 billion base
pairs (bp). The 2 m length of the genome is folded at multiple scales to fit inside a cell of around
10 $\mu$m diameter. Loops from this folding play specific roles in gene regulation. They
enable long range interaction between gene regulatory regions that lie far along a linear
chromosome or even on different chromosomes. A correlation between chromatin loops and aggregation of
the cohesin protein at loop anchors has been demonstrated~\citep{rao20143d}. It has been hypothesized
that cohesin plays a role in chromatin loop formation, confirmed by treating cells
with a molecule, auxin, that disrupts cohesin function. Addition of auxin led to a loss of
chromatin loops~\citep{rao2017cohesin}. We observed a congruent trend for H$_1$ loops when we computed PH for data sets with and without auxin treatment.

We estimated genome-wide pairwise distance matrices at $1$ kb resolution using data from Hi-C experiments for control and auxin-treated human cells~\cite{rao2017cohesin} (Section~\ref{methods:HiC_analysis}). $1$ kb resolution means that all linear chromosomes are partitioned into bins of a thousand
contiguous base pairs. A distance matrix $\widehat{D} = \widehat{d}_{ij}$ then estimates the
spatial distance between bin $i$ and bin $j$ in the folded genome (see
Section~\ref{methods:HiC_analysis} for details). This resulted in matrices of size
$3088281 \times 3088281$ for genome-wide analysis of the human genome. We denote the matrices for control and
auxin-treated by $\widehat{D}^c$ and $\widehat{D}^a,$ respectively.

\begin{figure}[h]
      \centering
      \begin{subfigure}{.48\textwidth} \centering
        \includegraphics[width=0.9\linewidth]{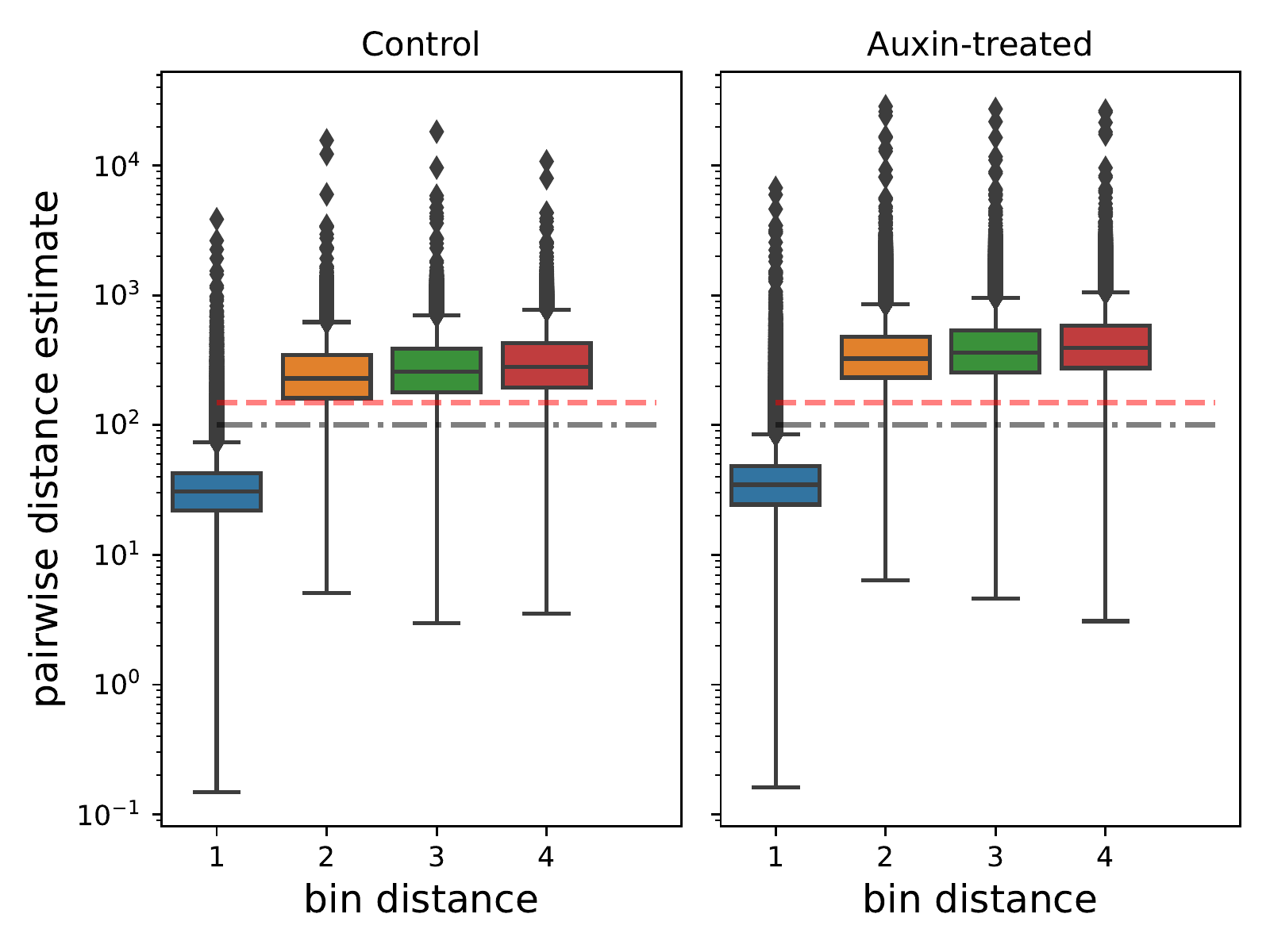}
        \caption{}
        \label{fig:HiC_edge_distribution}
    \end{subfigure}
      \centering
      \begin{subfigure}{.48\textwidth} \centering
        \includegraphics[width=0.9\linewidth]{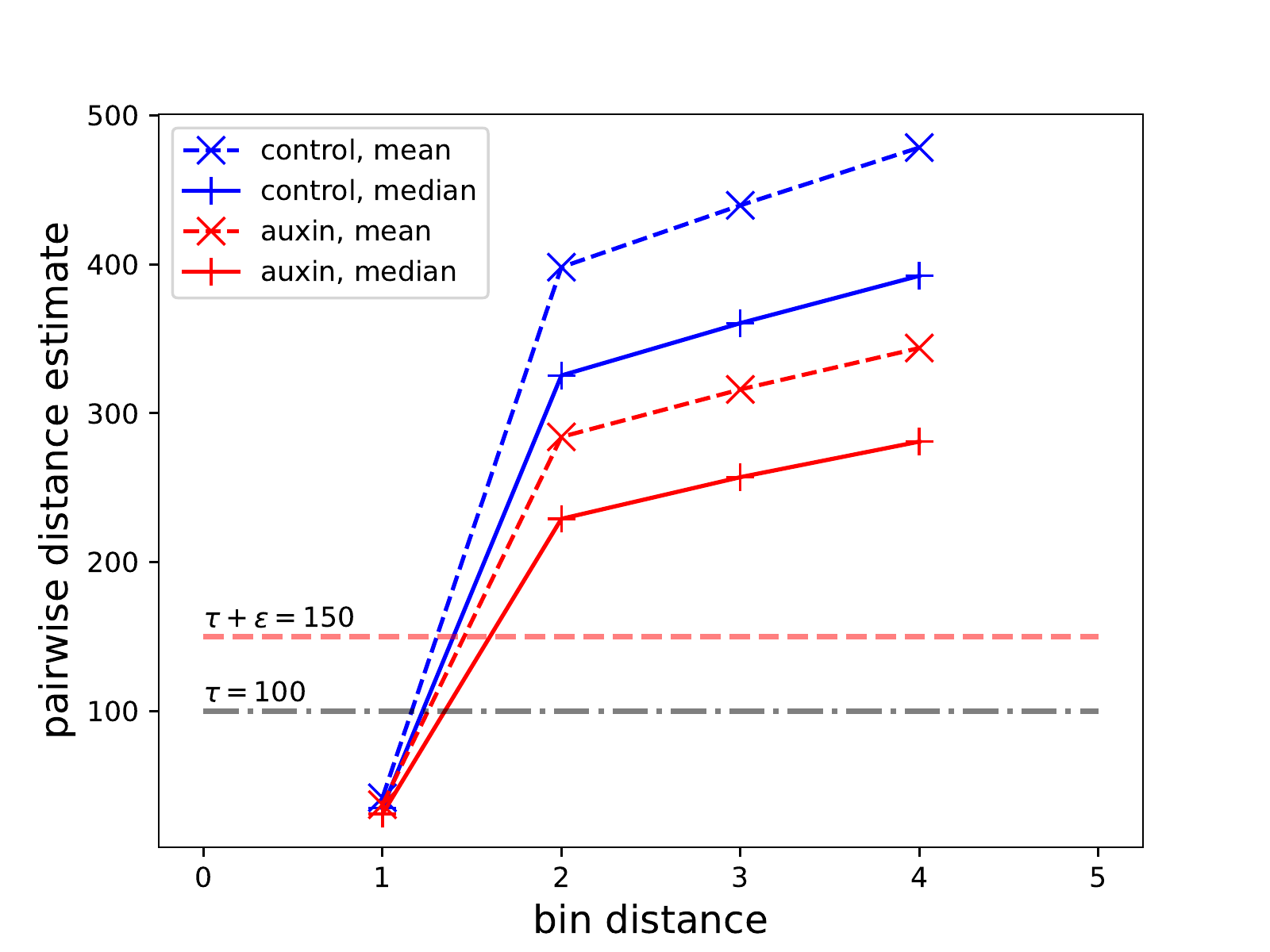}
        \caption{}
        \label{fig:HiC_edge_distribution_stats}
    \end{subfigure}

  \caption{Spatial pairwise estimates from Hi-C matrices reasonably estimate distance between
  adjacent bins on a chromosome.  (a) Distribution of pairwise estimates for bin distance of 1 is
  significantly lower as compared to those of higher bin distances. (b) Both means and medians of
  the distributions increase with increase in bin distances. }

  \label{fig:HiC_edge_bindistances}
\end{figure}

Figure~\ref{fig:HiC_edge_distribution} shows distributions of our estimates, $\widehat{d}^c_{ij}$
and $\widehat{d}^a_{ij},$ for bin distances $|i - j| = 1,2,3,$ and $4.$
Figure~\ref{fig:HiC_edge_distribution_stats} shows that their means and medians increase with bin
distances. We expect that in most cases the bins that are farther apart along the linear chromosome will
also be farther in the folded chromosome. Therefore, this is an expected property of our pairwise
estimation and provides a sanity check. Figure~\ref{fig:HiC_edge_distribution} shows that the
distributions for bin distances greater than $2$ are not as significantly different as compared to
that for bin distance of one. This suggested that the estimates might not be accurate for bin
distances larger than one. Hence, we defined a feature to be significant if it is born at a spatial
scale less than or equal to $\tau_u=100$ and has persistence at least $\epsilon=50.$ These choices
are based on the fact that $100$ is above the $75\%$-ile of estimates for bin distance of $1$ and
$150$ is below the $25\%$-ile of estimates for bin distance of $2$ (dashed lines in
Fig~\ref{fig:HiC_edge_bindistances}). Computing PH up to $\tau=\tau_u+\epsilon=150$ resulted in
3,206,797 and 2,701,615 valid entries in $\widehat{D}^c$ and $\widehat{D}^a,$ respectively, out of
the ${3088281\choose2} \approx 4.7 \times 10^{12}$ possible pairs.

Our algorithms computed PD and representative boundaries in around one minute using $3.1$ GB of
memory for $\widehat{D}^c$ and $35$ seconds with $3$ GB for $\widehat{D}^a$ (see
Table~\ref{tab:HiC_computation}). The low memory usage is 
because all data structures throughout our algorithms and strategy are designed to be on the order
of the maximal number of valid edges ($n_e$). This is specifically beneficial in this case because
$n_e \ll O(n^2).$ In general, we expect our memory usage to be lower than that of extant algorithms
by multiple orders of magnitude for topological structures where the number of simplices is much less than
the maximum number of simplices possible. Such topological structures are termed 
\textit{sparse}. It is difficult to theoretically generalize efficiency in run-time of the matrix
reduction algorithm \textit{a priori} since it depends on the number of reduction operations which
are decided by the structure of the coboundary and boundary matrices. The coboundary (and boundary)
matrices in turn depend upon the topology of the data set. Even for the same data set, different
coboundary and boundary matrices might be constructed that require different number of reduction
operations. This is possible because of the arbitrary choice of ordering of simplices that are added
to the topological structure at the same spatial scale.

\begin{table}
\begin{center}
  {
  \begin{tabular}{|c|c|c|c|c|c|c|c|c|} 
    \hline
     & $n$ & $\tau_u$ & $\epsilon$ & $n_e$ & \#H$_{1}$ & \#cycles &
     $\text{T}_\text{tot}$ (s) & $\text{M}_\text{tot}$ (GB) \\
    \hline
    Control & 3088281 & 100 & 50 & 3206797 &  375281 & 43855 & 67 & 3.1 \\
    Auxin-treated & 3088281 & 100 & 50 & 2701615 & 148628 & 21363 & 35 & 3 \\
    \hline
  \end{tabular}
}
\end{center}

  \caption{Our algorithms process large data sets within a minute. $n$ is the number of points,
  $\tau=\tau_u + \epsilon$ is the threshold for PH computation, $\#\text{H}_1$ is the number of
  nontrivial H$_1$ features, $\#$cycles is the number of cycles born at spatial scale less than or
  equal to $\tau_u,$ $\text{T}_\text{tot}$ is time in sec to compute PD, birth-cycles, and shorten
  and smooth birth-cycles, $\text{M}_\text{tot}$ is peak memory usage in GB.}

\label{tab:HiC_computation}
\end{table}

\begin{figure}[h]

      \centering
      \begin{subfigure}{.48\textwidth} \centering
        \includegraphics[width=0.9\linewidth]{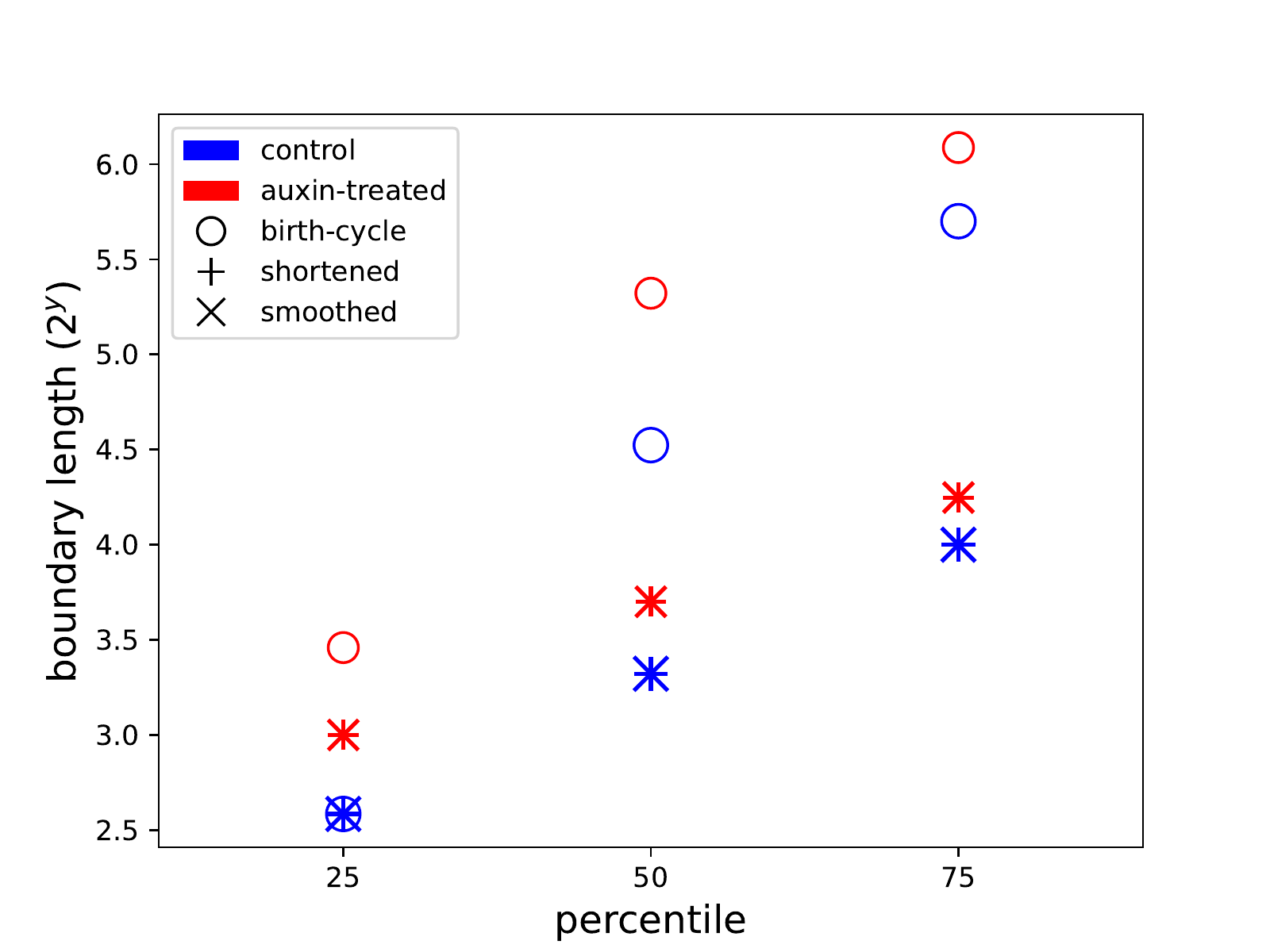}
        \caption{}
        \label{fig:HiC_percentiles_H1}
    \end{subfigure}
    \centering
      \begin{subfigure}{.48\textwidth} \centering
        \includegraphics[width=0.9\linewidth]{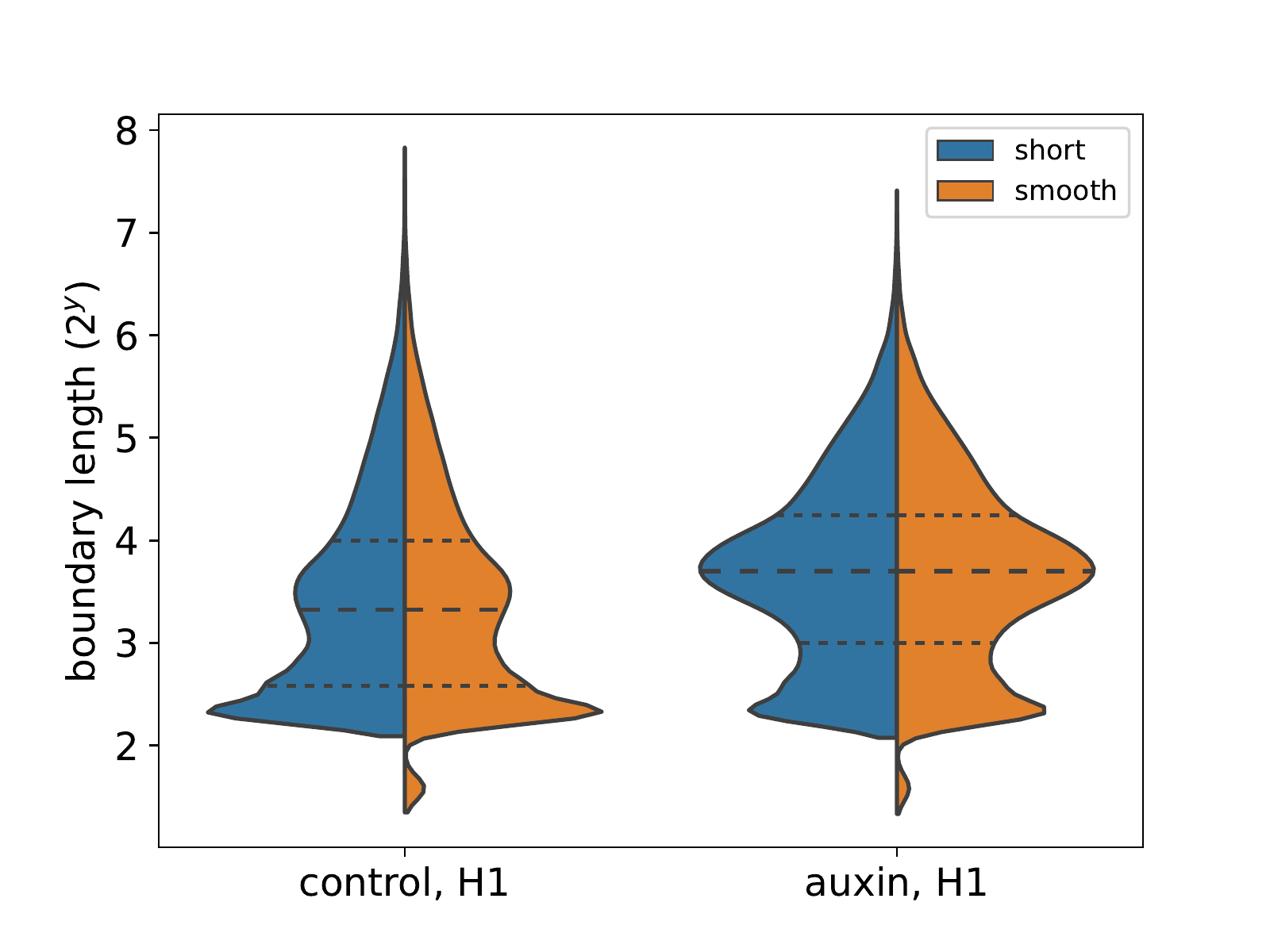}
        \caption{}
        \label{fig:HiC_short_to_smooth_H1_main}
    \end{subfigure}

  \caption{Lengths of cycles is decreased at every stage of the algorithm. (a) Interquartile lengths
  of the set of representative boundaries are decreased by multiple log scales by the greedy
  shortening algorithm. (b) Local smoothing reduces some boundaries to degenerate cycles of length
  two as is shown by the lower most humps in the distribution of smooth cycles.}

  \label{fig:HiC_compare_percentiles_lens}
\end{figure}

Figure~\ref{fig:HiC_compare_percentiles_lens} shows that the median and $75\%$-ile boundary lengths
decreased across multiple log scales (base 2) after being shortened by the greedy algorithm.
Figure~\ref{fig:HiC_compare_percentiles_lens} shows that smoothing of the shortened cycles reduced
some to degenerate cycles of length $2$ (lowermost humps in the distribution of the smooth cycles
lacking in the distribution of lengths of shortened cycles). Specifically, $902$ cycles in control
and $209$ in auxin-treated were reduced to degenerate cycles. These were ignored in subsequent
analysis.

\begin{figure}[h]
      \centering
      \begin{subfigure}{.33\textwidth} \centering
        \includegraphics[width=0.9\linewidth]{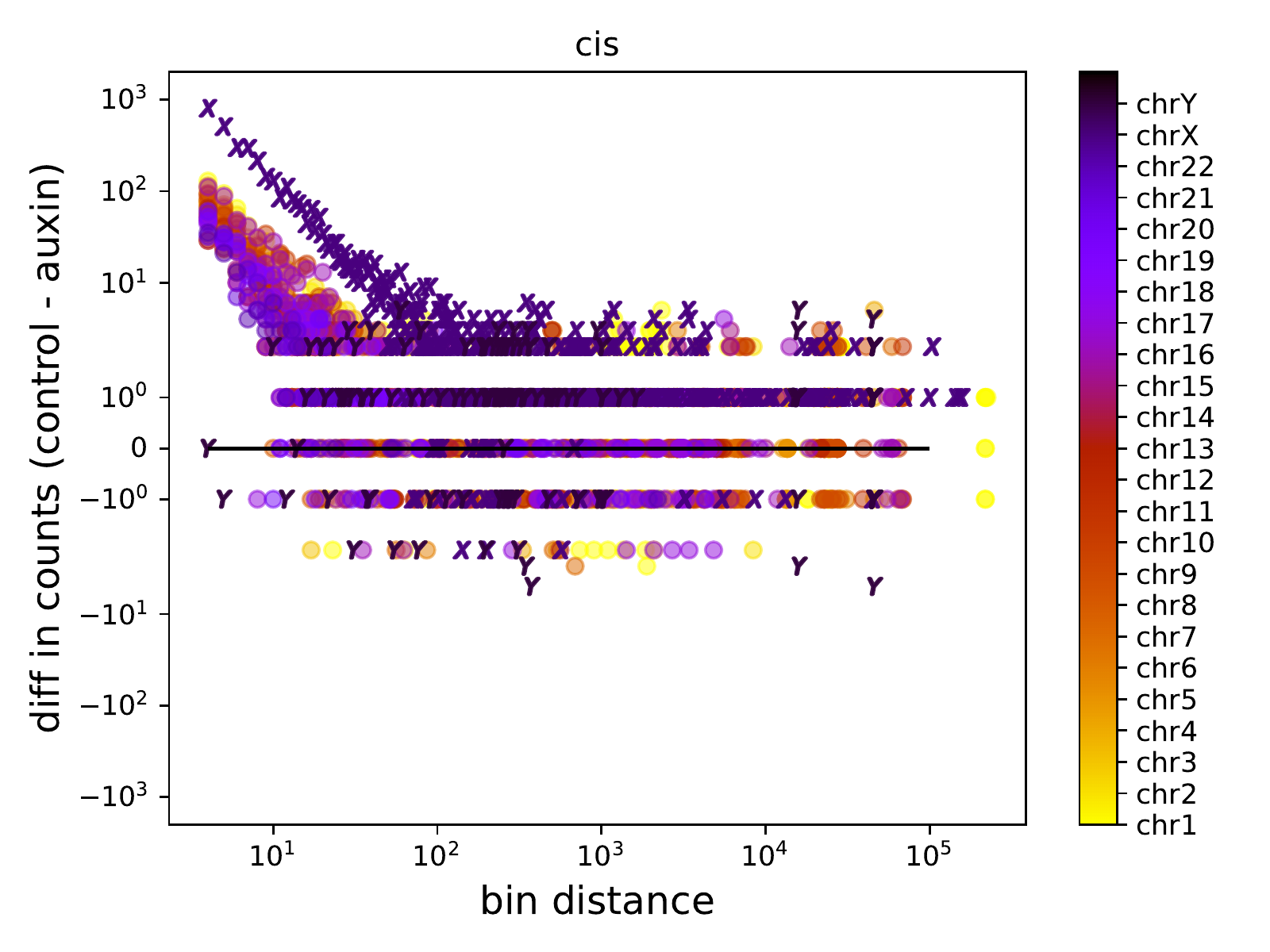}
        \caption{}
        \label{fig:HiC_cis_diff_bindist}
    \end{subfigure}
      \centering
      \begin{subfigure}{0.33\textwidth} \centering
        \includegraphics[width=0.9\linewidth]{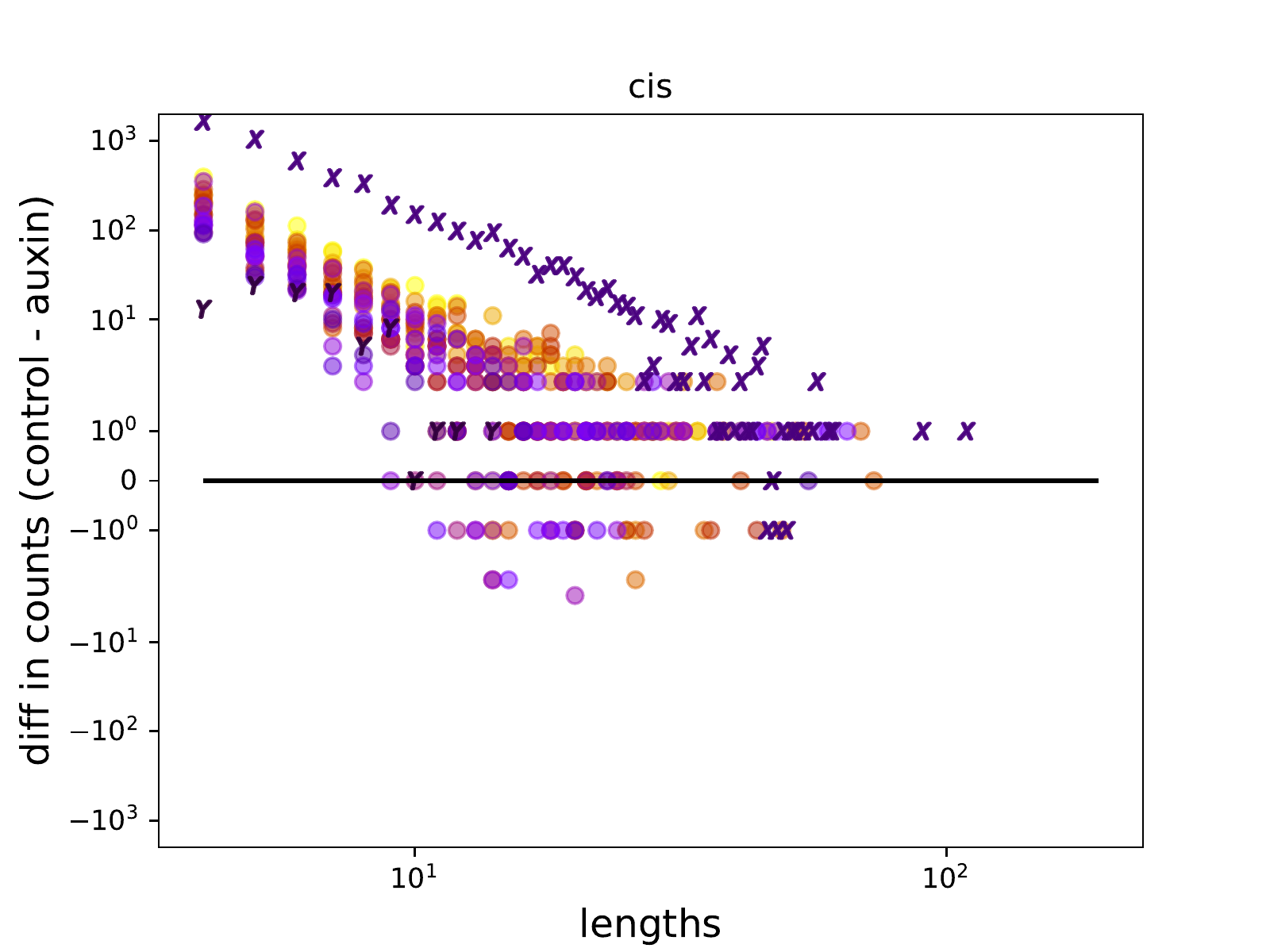}
        \caption{}
        \label{fig:HiC_cis_diff_lens}
    \end{subfigure}
      \centering
      \begin{subfigure}{0.33\textwidth} \centering
        \includegraphics[width=0.9\linewidth]{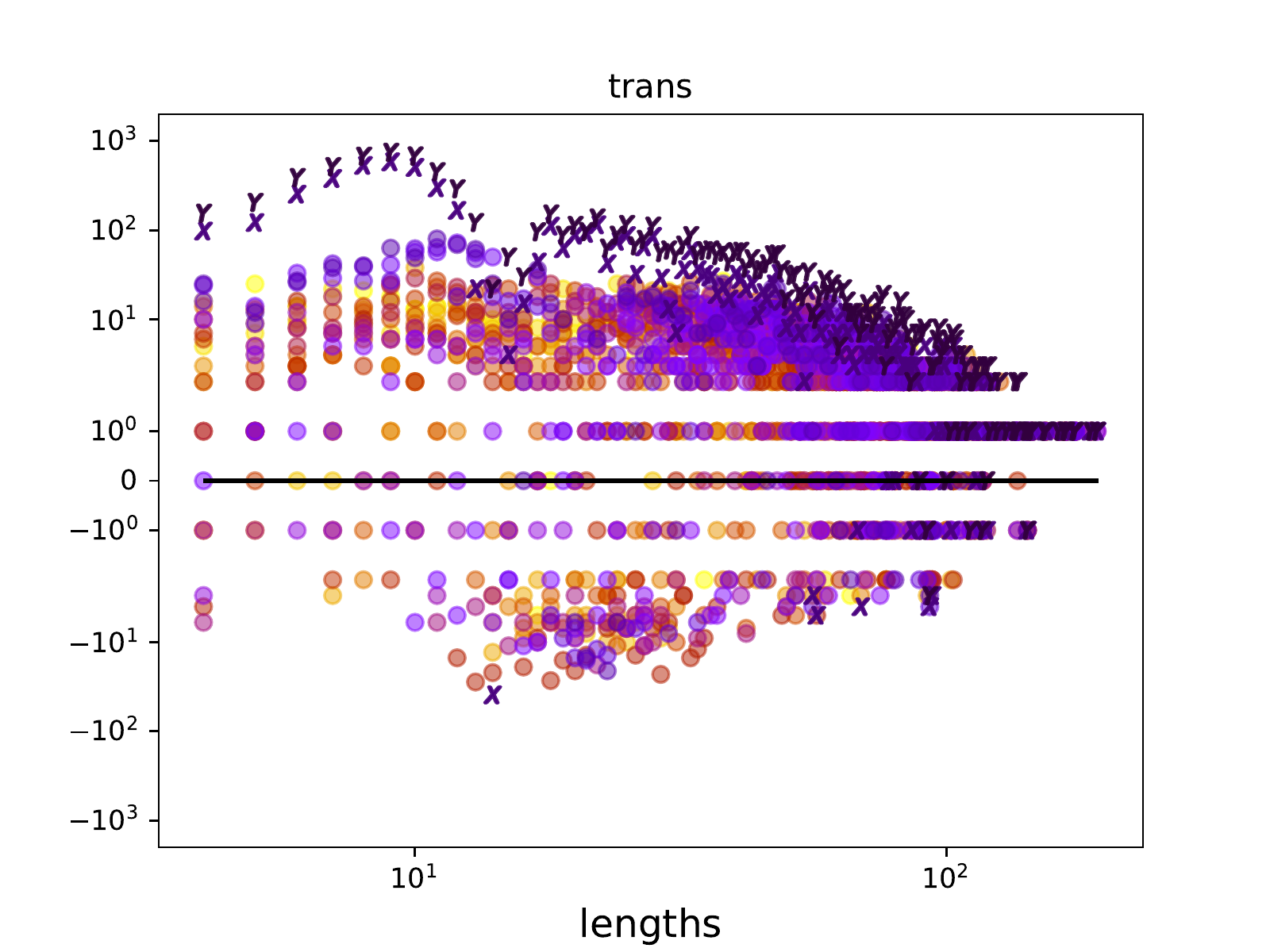}
        \caption{}
        \label{fig:HiC_trans_diff_lens}
    \end{subfigure}
      \centering
      \begin{subfigure}{.48\textwidth} \centering
        \includegraphics[width=0.9\linewidth]{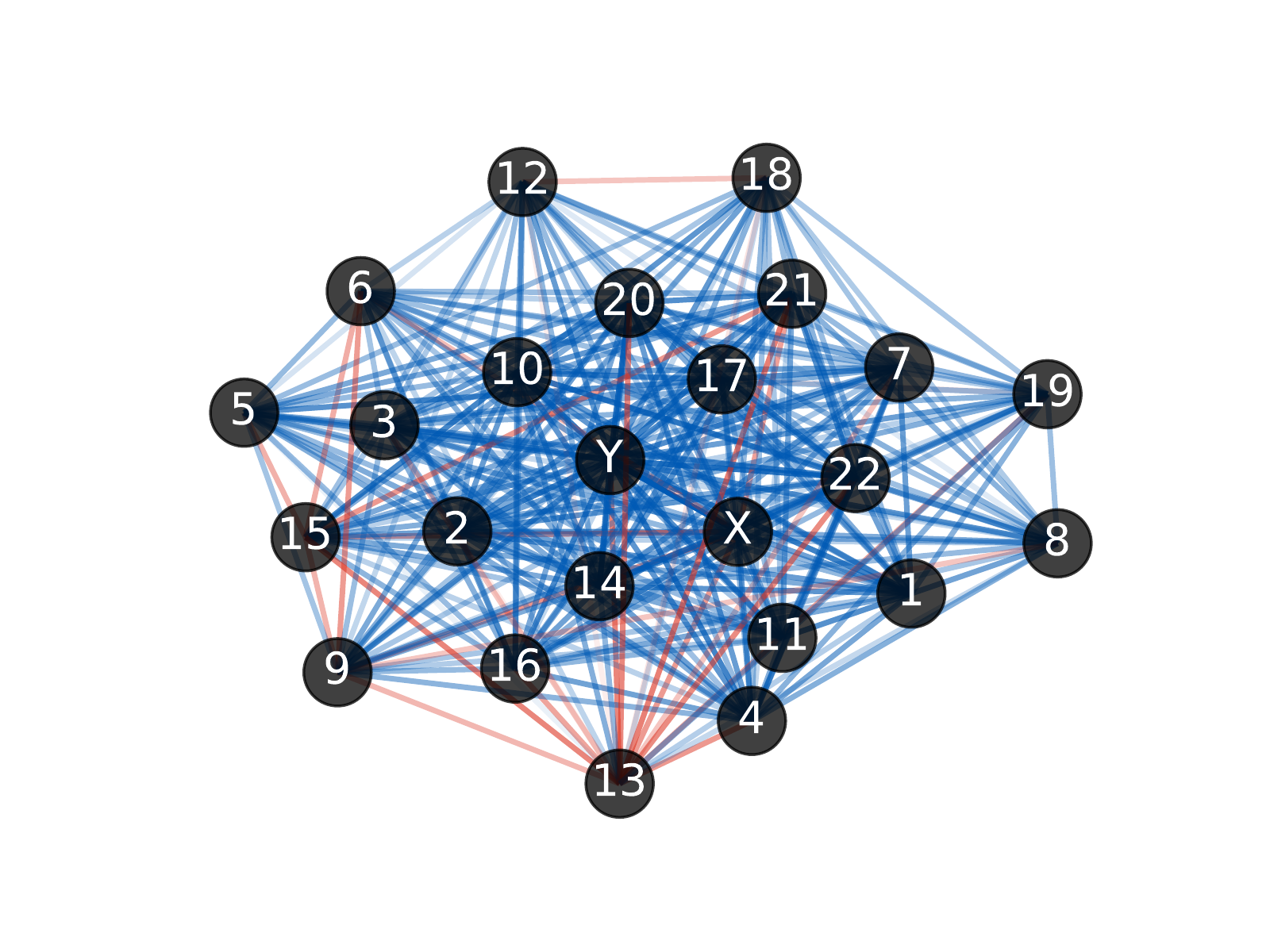}
        \caption{}
        \label{fig:HiC_compare_diff_trans_XY}
    \end{subfigure}
      \centering
      \begin{subfigure}{.48\textwidth} \centering
        \includegraphics[width=0.9\linewidth]{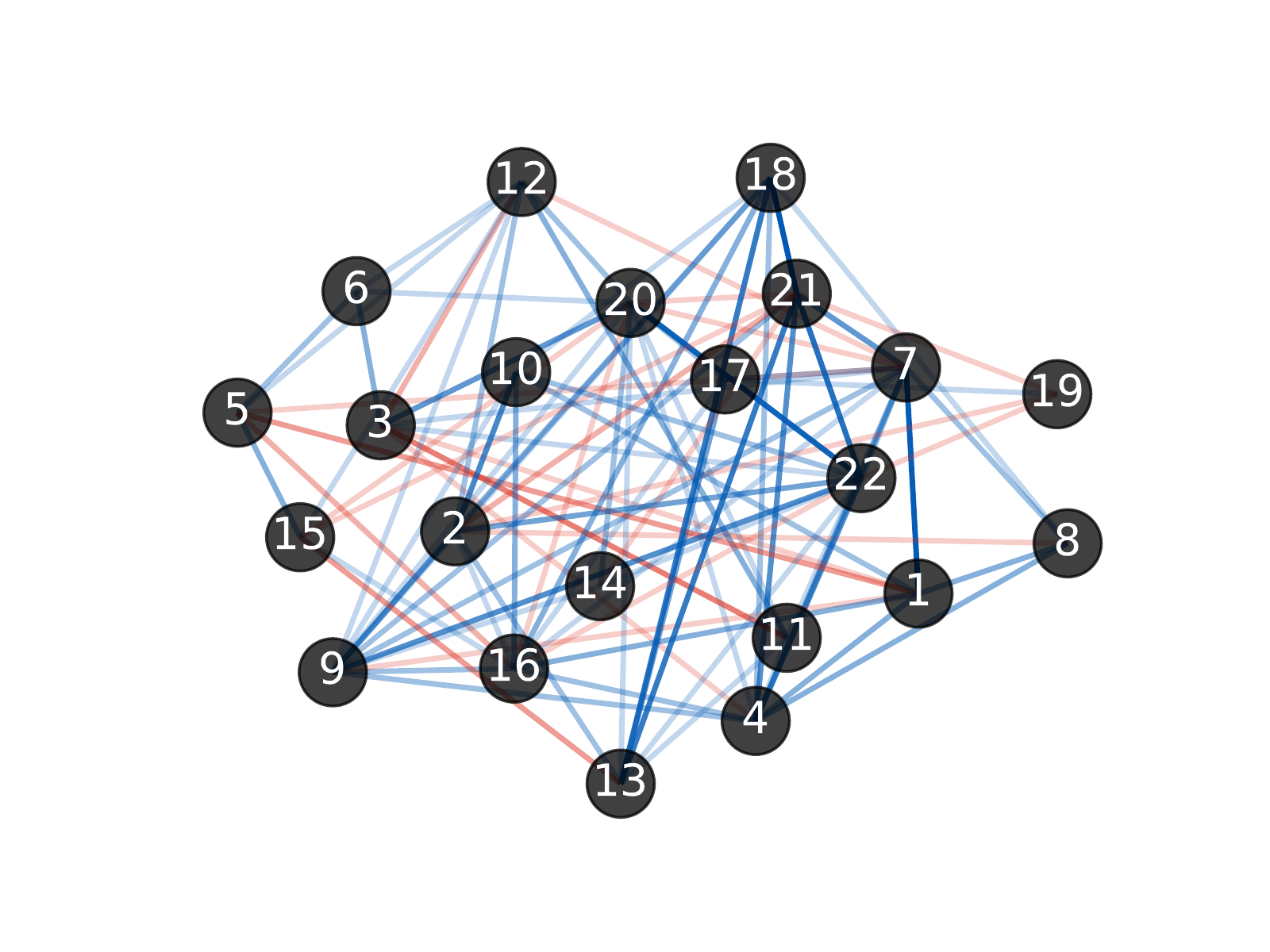}
        \caption{}
        \label{fig:HiC_compare_diff_trans_notXY}
    \end{subfigure}

  \caption{Both cis- and trans-cycles decrease in number upon treatment with auxin. (a) and (b)
  Number of cis-cycles is decreased upon addition of auxin, irrespective of cycle length and the
  maximal bin distance between contiguous bins. (c) Number of trans-cycles is also decreased, but
  there are relatively more instances of more trans-cycles in auxin-treated. (d) Opacity of edge
  $(c_i, c_j)$ indicates difference in number of trans cycles between control and auxin-treated that
  go through chromosomes $c_i$ and $c_j.$ Blue indicates more in control and red means more in
  auxin-treated. Chromosome 13 stands out with many red edges. (e) Same as previous but without the
  sex chromosomes. There are very few red edges at chromosome 13, indicating that auxin-treated has
  a larger number of trans-cycles that go through chromosome 13 and sex chromosomes.}

  \label{fig:HiC_control_auxin_diff}
\end{figure}

Next, we determine which of the shortened and smoothed boundaries cannot be around a significant feature. For every cycle $C,$ we estimate the theoretically maximum persistence that a feature, with $C$ as its representative boundary, can have. The birth is estimated as the longest edge in the boundary, because at that spatial scale the cycle is born. Death is estimated as the maximum of all possible pairwise distances between points of the cycle. This is because, at that spatial scale all simplices on the points of the cycle will be added to the simplicial complex. Hence, the theoretically maximum persistence is (estimated death) - (estimated birth). If this value is less than $\epsilon,$ then $C$ is ignored from subsequent analysis. There were $271$ such cycles in control and $22$ in auxin-treated. This left us with $42695$ cycles in
control and $21137$ in auxin-treated.
Hence, auxin treatment resulted in a loss of around $50\%$ of possibly significant H$_1$ loops. For further analysis, we considered cis- and trans-cycles separately. If a cycle is defined by edges with all bins from the same chromosome, we call it a cis-chromosome cycle or a cis-cycle. 
If a cycle is defined by edges with at least two bins on different chromosomes, we call it a trans-chromosome cycle or a trans-cycle. Analysis of these sets of minimal cycles showed that the control data set has a higher number of cis-cycles in all chromosomes, irrespective of their
length and maximal bin distance between contiguous bins in the cycle (see
Figures~\ref{fig:HiC_cis_diff_lens} and ~\ref{fig:HiC_cis_diff_bindist}). The 
differences in the number of cycles at bin-distances of up to $10^5$ in
Figure~\ref{fig:HiC_cis_diff_bindist} indicates that we found cis-cycles that have contiguous bins
which are at some distance along the linear chromosome. This provides evidence for long-range interactions.
The number of trans-cycles also decreased upon treatment with auxin. However, there are relatively
more trans-cycles in auxin-treated data (more negative points in
Figure~\ref{fig:HiC_trans_diff_lens}). For every pair of chromosomes, we counted the number of
trans-cycles that go through the pair. We plot the difference in the counts between control and auxin treatment
as a graph in Figure~\ref{fig:HiC_compare_diff_trans_XY}. A red edge between a pair of
chromosomes $(c_i, c_j)$ implies that there are more trans-cycles going through $c_i$ and $c_j$ in auxin-treated than in control, and blue edges indicate the opposite. The predominance of blue edges indicates that for almost all chromosome pairs there are more
trans-cycles in control as compared to auxin-treated. An exception that stands out is chromosome 13 which has more red edges. We repeated this analysis without the sex chromosomes to analyze
trans-cycles that go through only the autosomes. Figure~\ref{fig:HiC_compare_diff_trans_notXY} shows that such
trans-cycles through chromosome 13 are mostly blue. Hence, the auxin-treated data set has more trans-cycles that go through chromosome 13 and the sex chromosomes. This might yield insights into functional relations between chromosome 13 and the sex chromosomes. For example, ambiguous genitalia is known to be associated with sex chromosome disorders. However, at least one case has been observed of ambiguous genitalia with the autosomal chromosome anomaly of ring chromosome 13~\citep{ozsu2014ring}.

\subsection{Rare voids in a distribution of more than 100,000 galaxies}

\begin{figure}[h]
 \centering
   \begin{subfigure}{.32\textwidth}
   \centering
    \includegraphics[width=\linewidth]{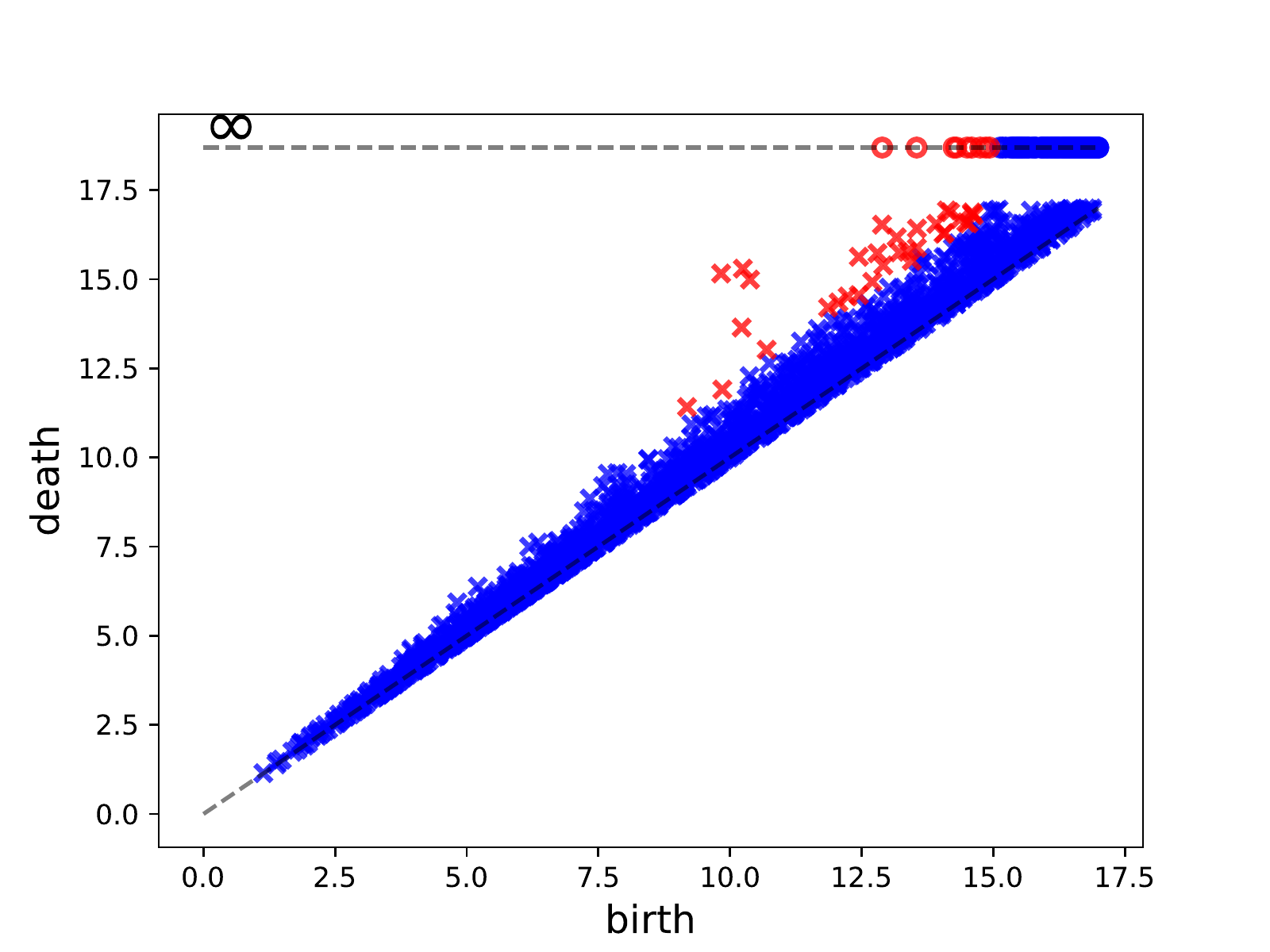}
     \caption{}
   \label{fig:universe_H2}
   \end{subfigure}
 \centering
   \begin{subfigure}{.32\textwidth}
   \centering
    \includegraphics[width=\linewidth]{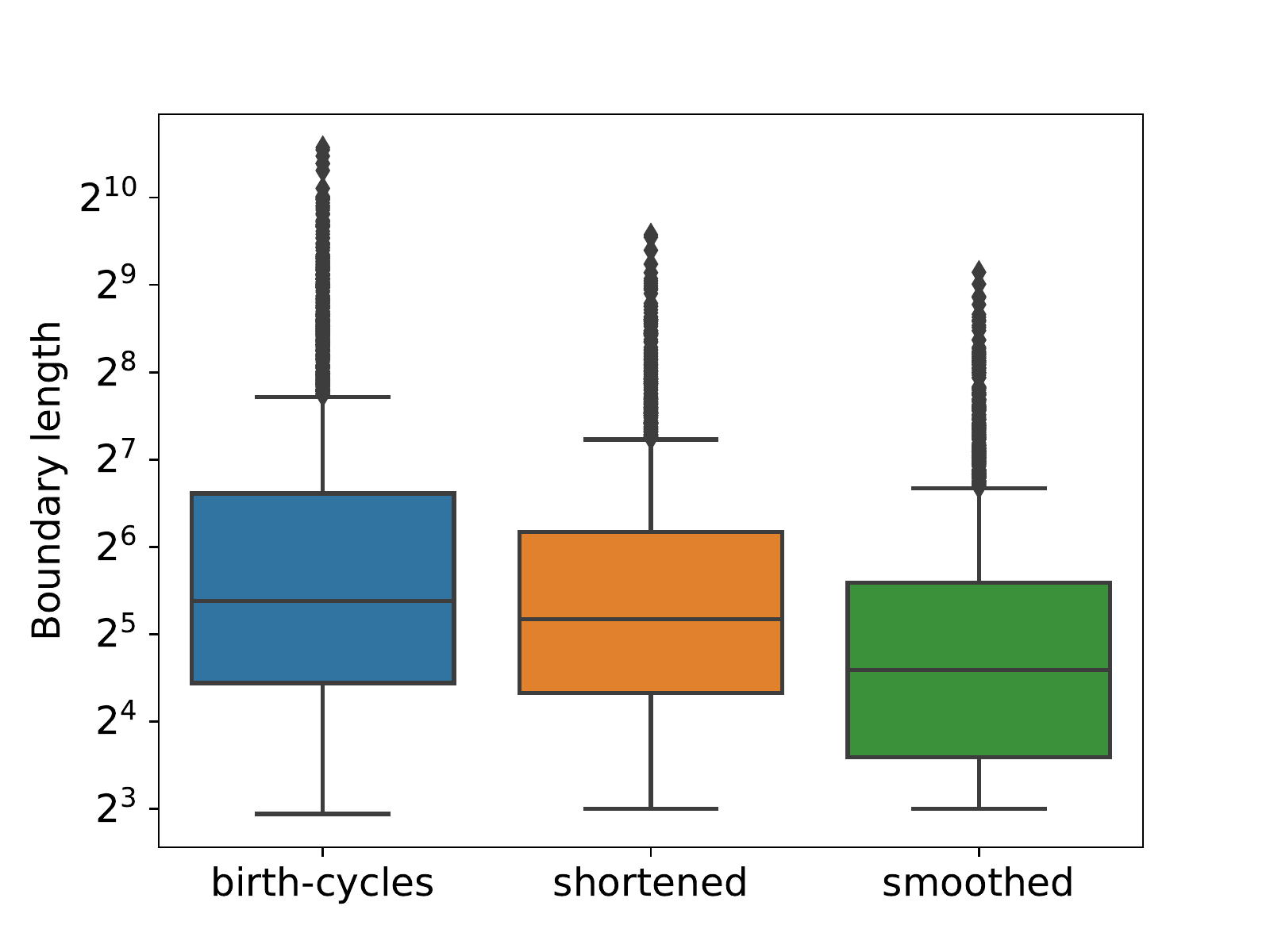}
     \caption{}
   \label{fig:universe_compare_boundary_lens}
   \end{subfigure}
 \centering
   \begin{subfigure}{.32\textwidth}
   \centering
    \includegraphics[width=\linewidth]{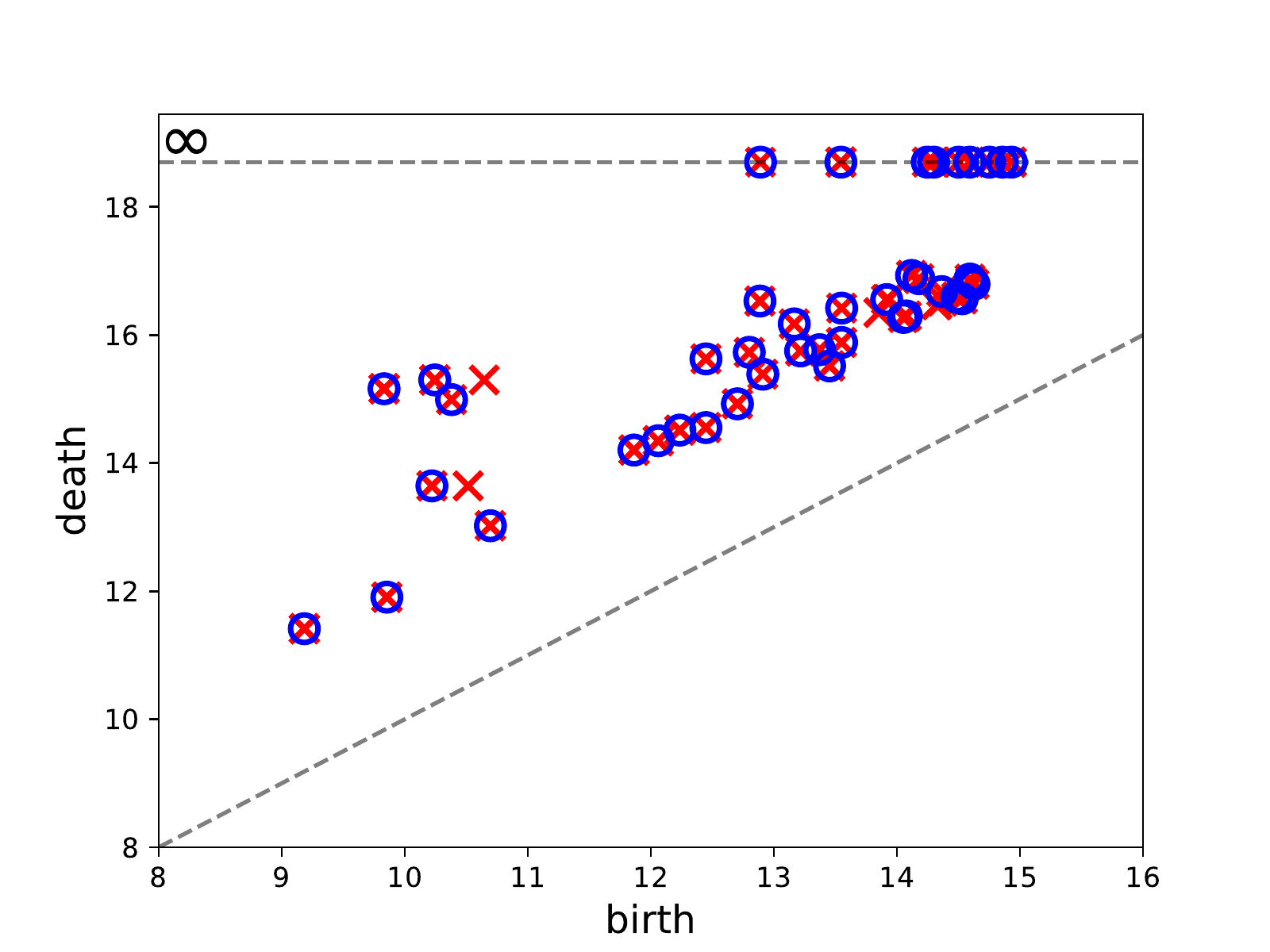}
     \caption{}
   \label{fig:universe_compare_PD_birth}
   \end{subfigure}
 \centering
   \begin{subfigure}{.48\textwidth}
   \centering
    \includegraphics[width=\linewidth]{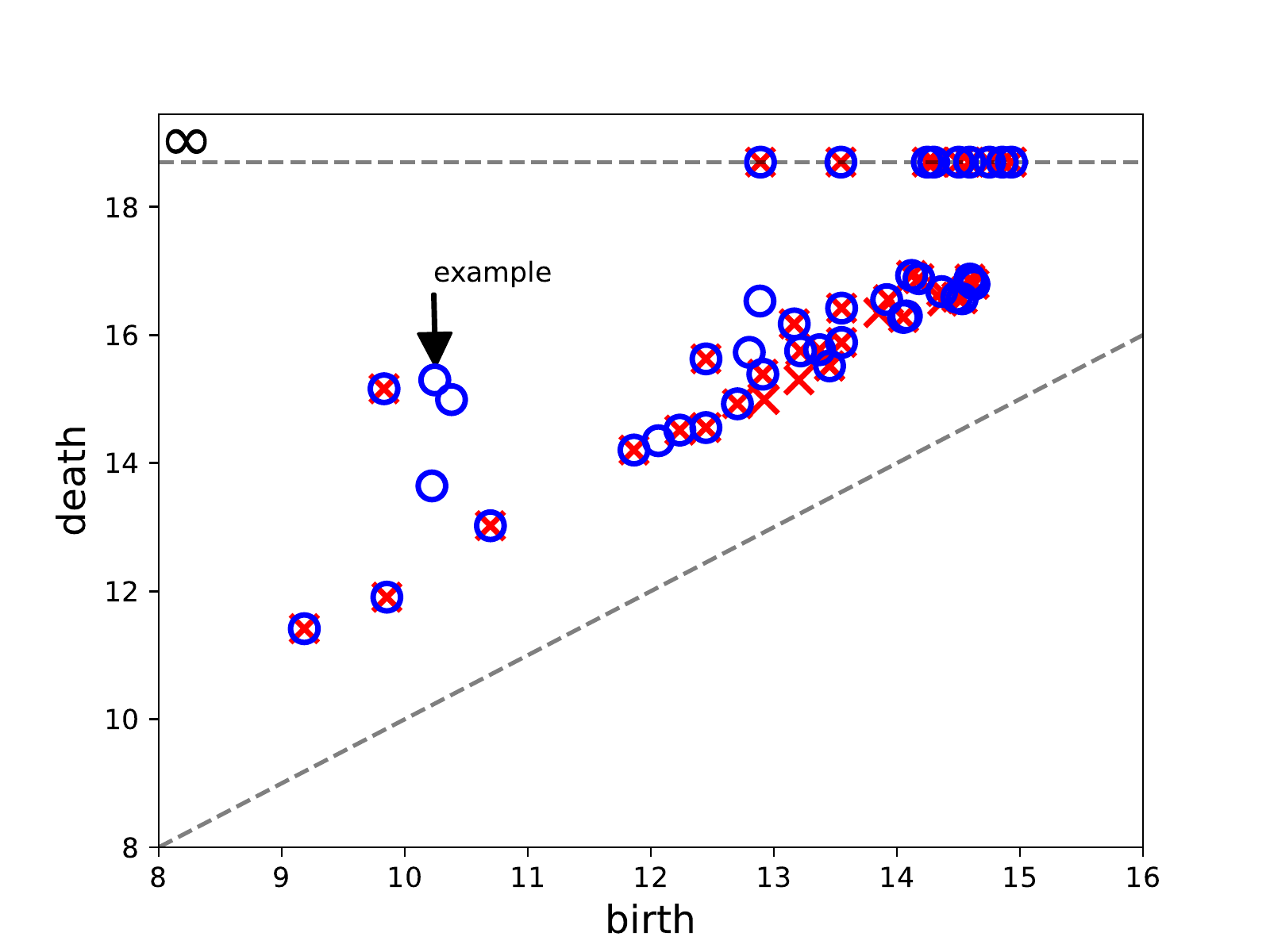}
     \caption{}
   \label{fig:universe_compare_PD_short}
   \end{subfigure}
 \centering
   \begin{subfigure}{.48\textwidth}
   \centering
    \includegraphics[width=\linewidth]{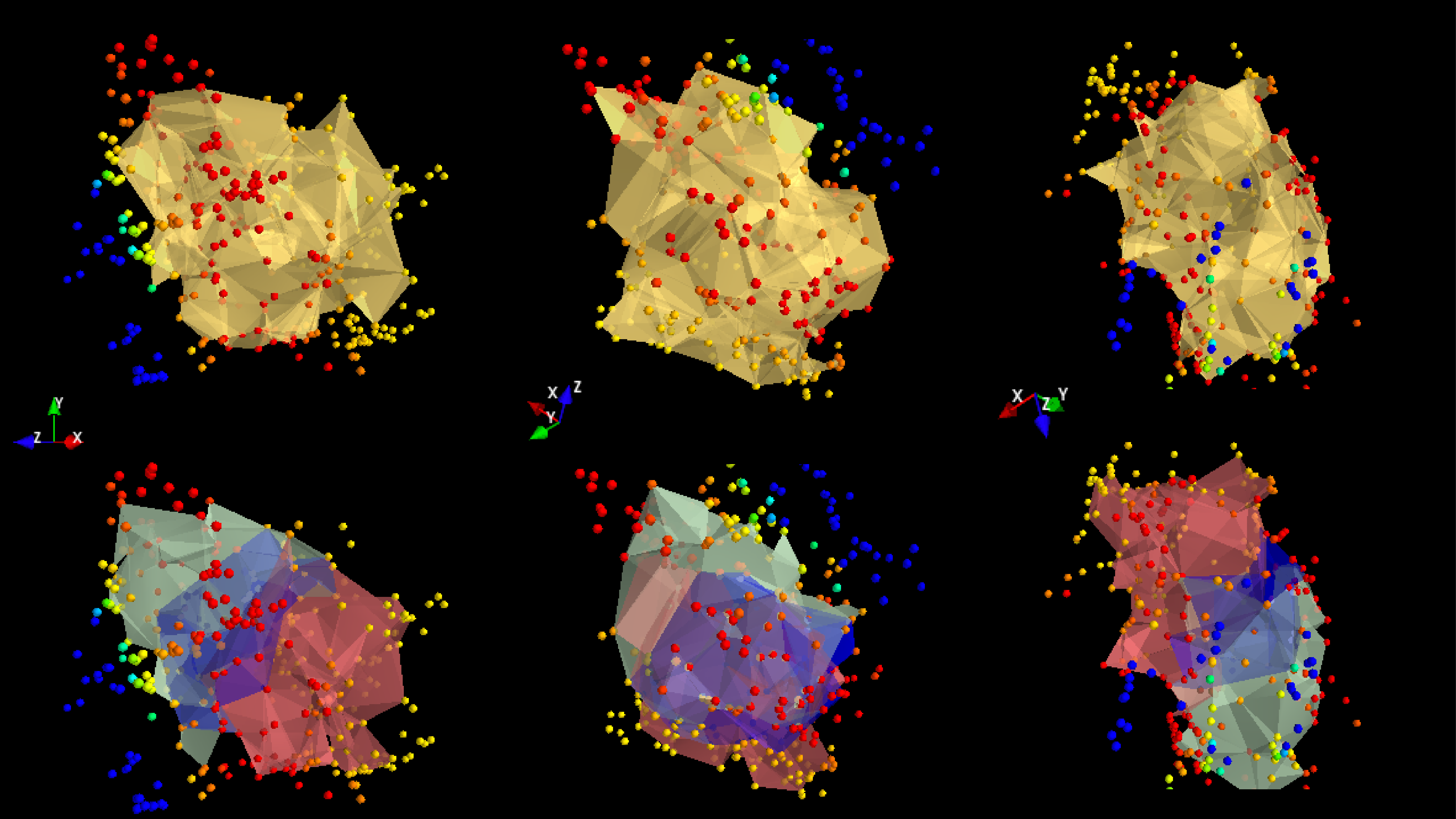}
     \caption{}
   \label{fig:universe_case_example}
   \end{subfigure}

  \caption{Results of PH computation, shortening, and smoothing for spatial embedding of around
  $108,000$ galaxies (a) H$_2$ PD with significant features ($\tau_u = 15, \epsilon = 2$) marked by
  red. (b) Lengths of birth-cycles are decreased by both shortening and smoothing. (c) Open-face
  blue circles are significant features in the H$_2$ PD of full data set. Red crosses are
  significant features in covers of birth-cycles. The set of birth-cycles wrap around all features
  classified as significant in the PD of the full data set. (d) Open-face blue circles are
  significant features in the H$_2$ PD of full data set. Red crosses are significant features in
  covers shortened cycles. Shortened cycles do not wrap around all features classified as
  significant in the PD of full data set. This is a feature of the shortening algorithm, as we show
  by comparing homology cycles with shortened cycles for the annotated example. (e) Top row shows
  birth-cycle, for the annotated example, in three different orientations. Bottom row shows the
  shorter cycles found by the greedy algorithm, in same orientations. These shorter boundaries have
  birth $< \tau_u,$ but higher than that of the birth-cycle. The persistence of holes that they wrap
  around is less than $\epsilon,$ and they are not classified as significant.}

  \label{fig:universe_fig1}
\end{figure}

We benchmarked computation of minimal boundaries around voids in $\mathbb{R}^3$ by analyzing a
distribution of 108,030 galaxies in the universe. We also showed that the computed boundaries
identified distinct voids in the universe. The locations of galaxies in $\mathbb{R}^3$
was estimated from a redshift data set (\texttt{http://www-wfau.roe.ac.uk/6dFGS/}) using Hubble's law (see
Section~\ref{method:universe} for details). Significant features were defined by $\tau_u=15$ and
$\epsilon=2.$ The number of nontrivial H$_2$ features was $2351,$ out of which $42$ were classified
as significant (shown in red in the H$_2$ PD shown in Figure~\ref{fig:universe_H2}). Computation took
$42$ mins on a 2.4 Ghz Intel Core i9 to compute PD, compute a set of birth-cycles, shorten them, and
finally smooth them. Figure~\ref{fig:universe_compare_boundary_lens} shows that both shortening
and smoothing algorithms decreased the lengths of boundaries.

 Next, we show that the computed birth-cycles wrap around all features classified as significant in
 the data set. We define the cover of a set of embedded points as the smallest hyper-rectangle in Cartesian coordinates that contains all those points on it or inside it (see~\ref{method:strategy_algorithm} for a rigorous definition). We estimated significant features that a birth-cycle might contain by computing PD of
 its cover. 
In Figure~\ref{fig:universe_compare_PD_birth}, we indicate the significant features in the full data set
 using blue \textbf{o}'s and the estimated significant features in the birth-cycle using red \textbf{x}'s. For
 every \textbf{o} there exists an exactly matching \textbf{x}. Hence, birth cycles wrap around all
 features classified as significant in the PD of the full data set.

Do the shortened boundaries also wrap around all features that are classified as significant in the
PD of the full data set? We estimated significant features for the set of shortened cycles to
compare with the significant features in the full data set.
Figure~\ref{fig:universe_compare_PD_short} shows that not all significant features in the full data
set (\textbf{o}) seem to have a matching significant feature in shortened cycles (\textbf{x}).

We explain why this happens and how it is a feature of our algorithm by comparing birth-cycles with
shortened cycles for an annotated example:  Figure~\ref{fig:universe_case_example}
shows six different views (two rows, three columns) of the embedded points in the spatial region of
the example. The columns show three different orientations. The top row shows the birth-cycle. The
bottom row shows three shortened cycles in that region as computed by the greedy algorithm. This
region has a significant feature found by the matrix
reduction algorithm with a basis cycle that is born at $10.24$ and that dies at $15.3,$ denoted as $(10.24, 15.3),$ in the full data set. However,
the persistence pairs with maximum persistence in covers of shortened cycles are $(13.2, 15.3),
(14.32, 15.34),$ and $(13.96, 14.2).$ The persistence of all these is less than $\epsilon,$ but
constituents in these boundaries can interact because their births are less than $\tau_u.$ Hence,
the shorter cycles found by the greedy algorithm show that the annotated feature is not significant.

\begin{figure}[h]
 \centering
   \begin{subfigure}{.48\textwidth}
   \centering
    \includegraphics[width=\linewidth]{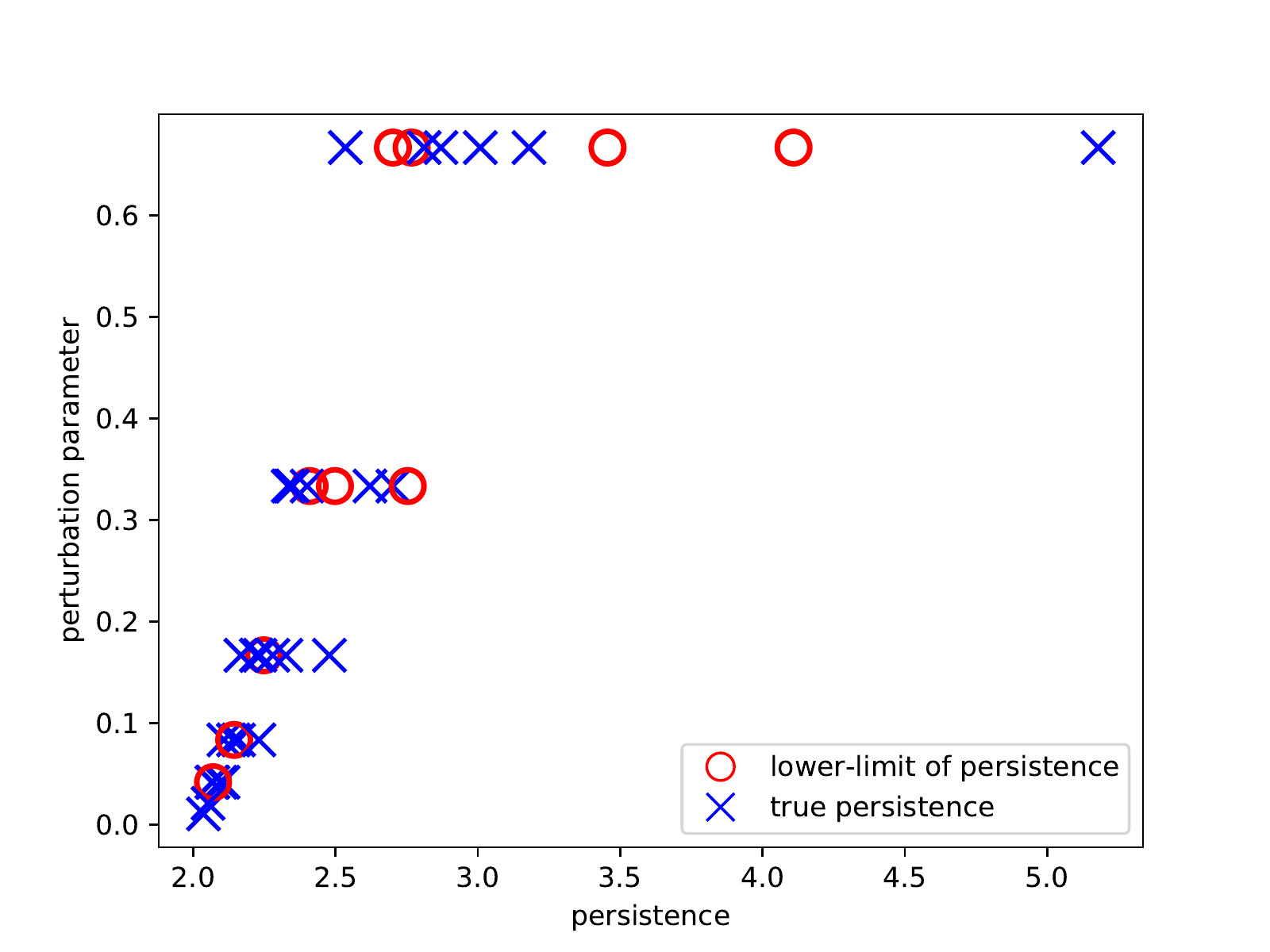}
     \caption{}
   \label{fig:universe_perturbation_parameter}
   \end{subfigure}
 \centering
   \begin{subfigure}{.48\textwidth}
   \centering
    \includegraphics[width=\linewidth]{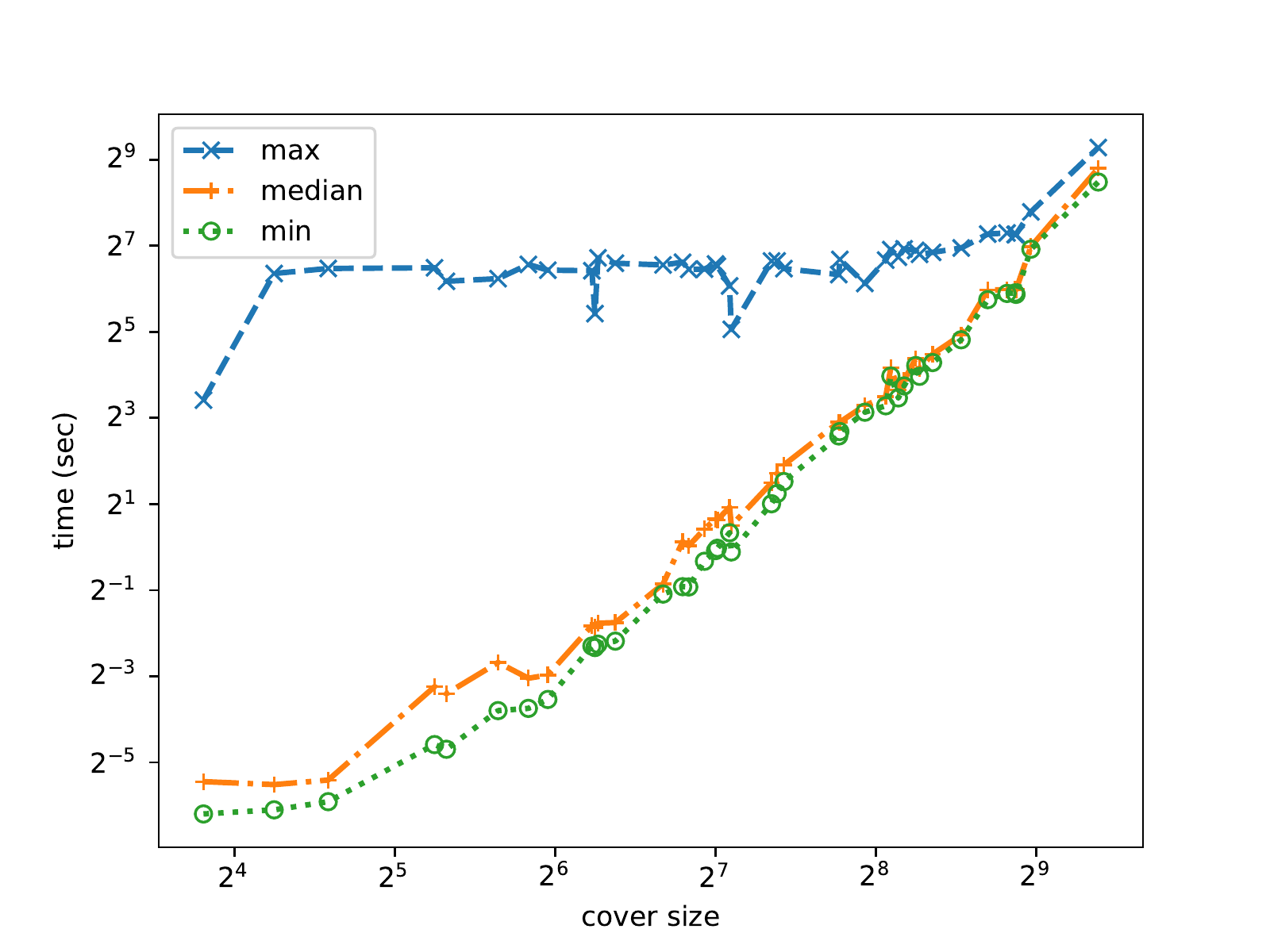}
     \caption{}
   \label{fig:universe_computetimes_stochastic}
   \end{subfigure}
 \centering
   \begin{subfigure}{.48\textwidth}
   \centering
    \includegraphics[width=\linewidth]{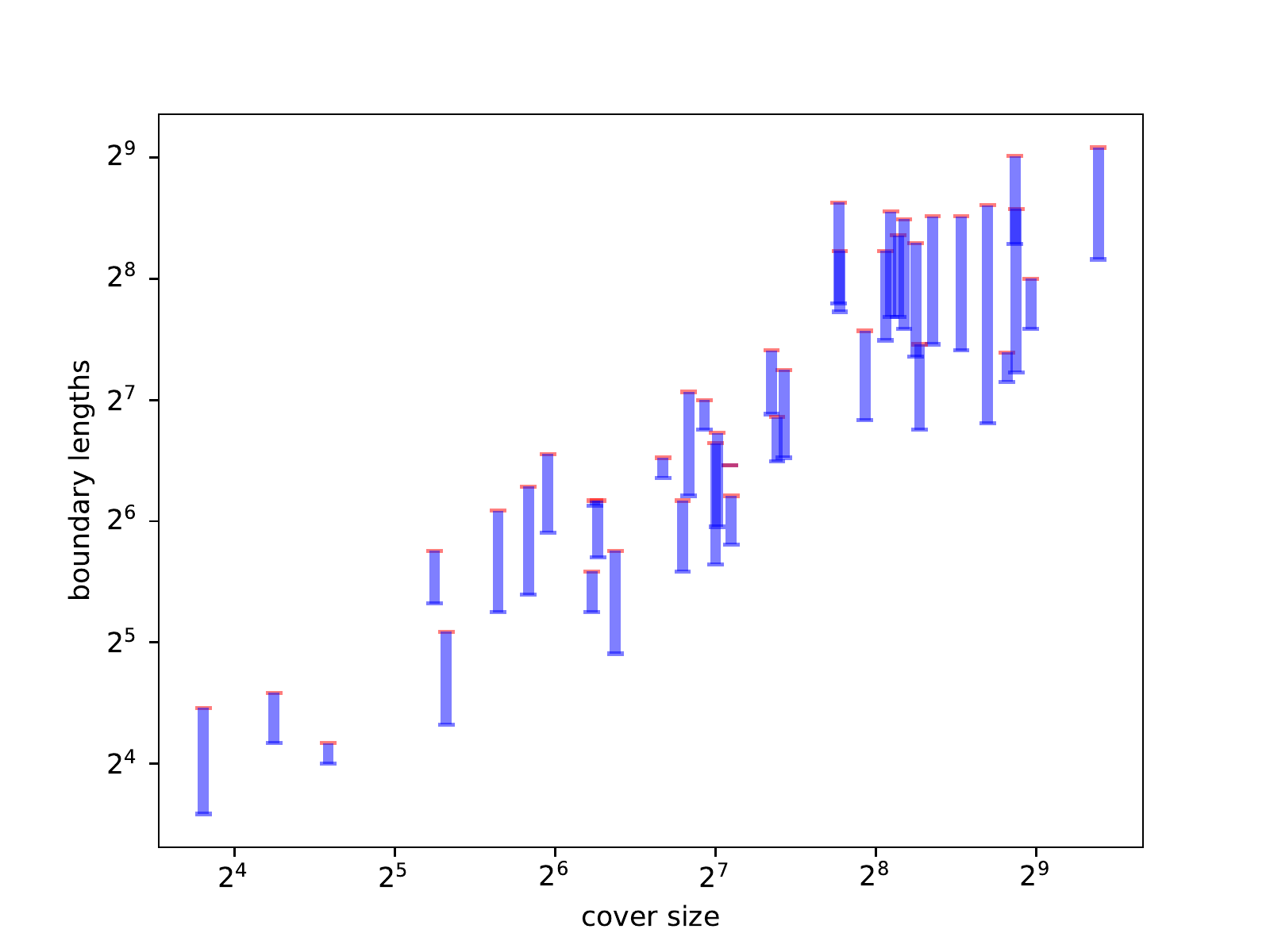}
     \caption{}
   \label{fig:universe_stochastic_lens}
   \end{subfigure}
 \centering
   \begin{subfigure}{.48\textwidth}
   \centering
    \includegraphics[width=\linewidth]{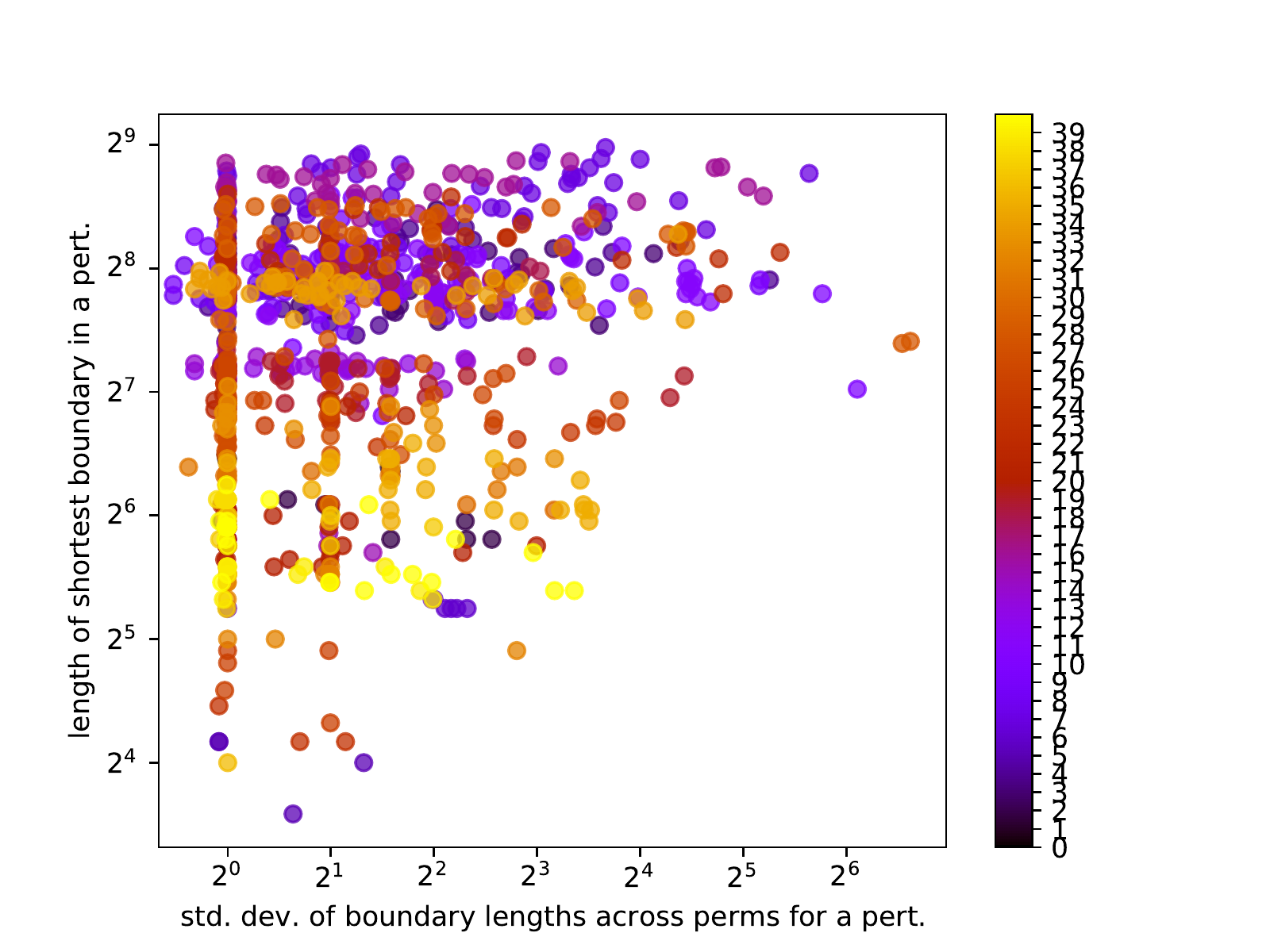}
     \caption{}
   \label{fig:universe_stochastic_lens_variation}
   \end{subfigure}

  \caption{Statistics on perturbations and permutations (a) Covers with significant feature of
  higher persistence have a higher perturbation parameter. (b) Minimum, median, and maximum of
  distribution of computation times for all permutations of all perturbations of a cover. Median and
  minimum increase with increase in number of points in the cover. A large disparity is seen for the
  maximum time taken for covers of small size, showing that it is difficult to estimate run-time
  \textit{a priori} with high precision. (c) Lengths of representative homology boundaries vary
  across all permutations of all perturbations for every cover. (d) Both perturbations and
  permutations contribute to variation in lengths of the representative boundaries.}

  \label{fig:universe_stochastic } 
  \end{figure}

Covers of $41$ of the smoothed boundaries had a non-zero number of significant features. The largest
cover had $0.6\%$ of the number of points in the original data set. Hence, the computational cost of any
subsequent analysis was considerably reduced. Graphical contraction reduced the number of possible
covers with significant features from $41$ to $40,$ 
but it did not reduce the number of points in
any cover. All $40$ covers had exactly one significant feature. We constructed $\text{n}_\text{pert} =
50$ perturbations for every cover. Figure~\ref{fig:universe_perturbation_parameter} shows that
features with larger persistence have a larger perturbation parameter
(see~\ref{method:strategy_algorithm} for definition of perturbation parameter). This is expected
since larger persistence implies robustness to larger perturbations. We constructed
$\text{n}_\text{perm} = 50$ permutations for every perturbation by permuting indices of edges of same length in the topological structure. 
Hence, every random permutation is a re-indexing of edges in the simplicial complex. However, it is possible that all edges are indexed similarly in two permutations. Chances of this are higher when there are very few edges of same lengths in the complex. Therefore, we only consider unique sets of labels out of the $50$ random permutations (see Section~\ref{method:compute_boundaries} for more details on perturbations and permutations).

It was feasible to compute PH up to the maximum possible threshold (see
Figure~\ref{fig:universe_computetimes_stochastic}). Consequently, we computed representative
homology boundaries for all unique permutations of perturbations of every cover.
Figure~\ref{fig:universe_stochastic_lens} shows that the lengths of the computed homology boundaries
varied considerably across all permuted sets of all perturbations of a cover.
Figure~\ref{fig:universe_stochastic_lens_variation} shows that both permutations and perturbations
contributed to variation in lengths.

\begin{figure}[tbhp!]
 \centering
   \begin{subfigure}{.48\textwidth}
   \centering
    \includegraphics[width=\linewidth]{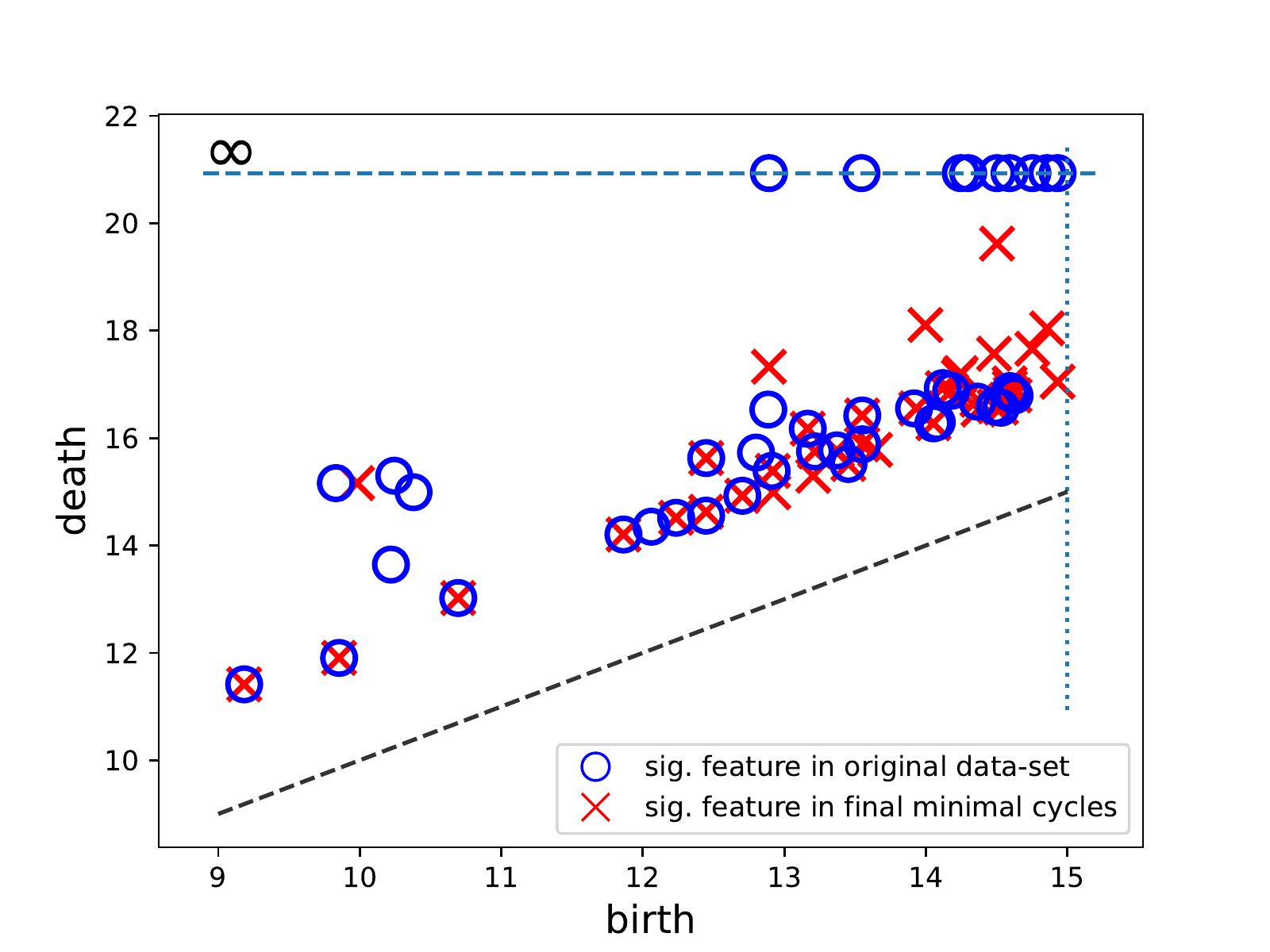}
     \caption{}
   \label{fig:universe_final_PD}
   \end{subfigure}
 \centering
   \begin{subfigure}{.48\textwidth}
   \centering
    \includegraphics[width=\linewidth]{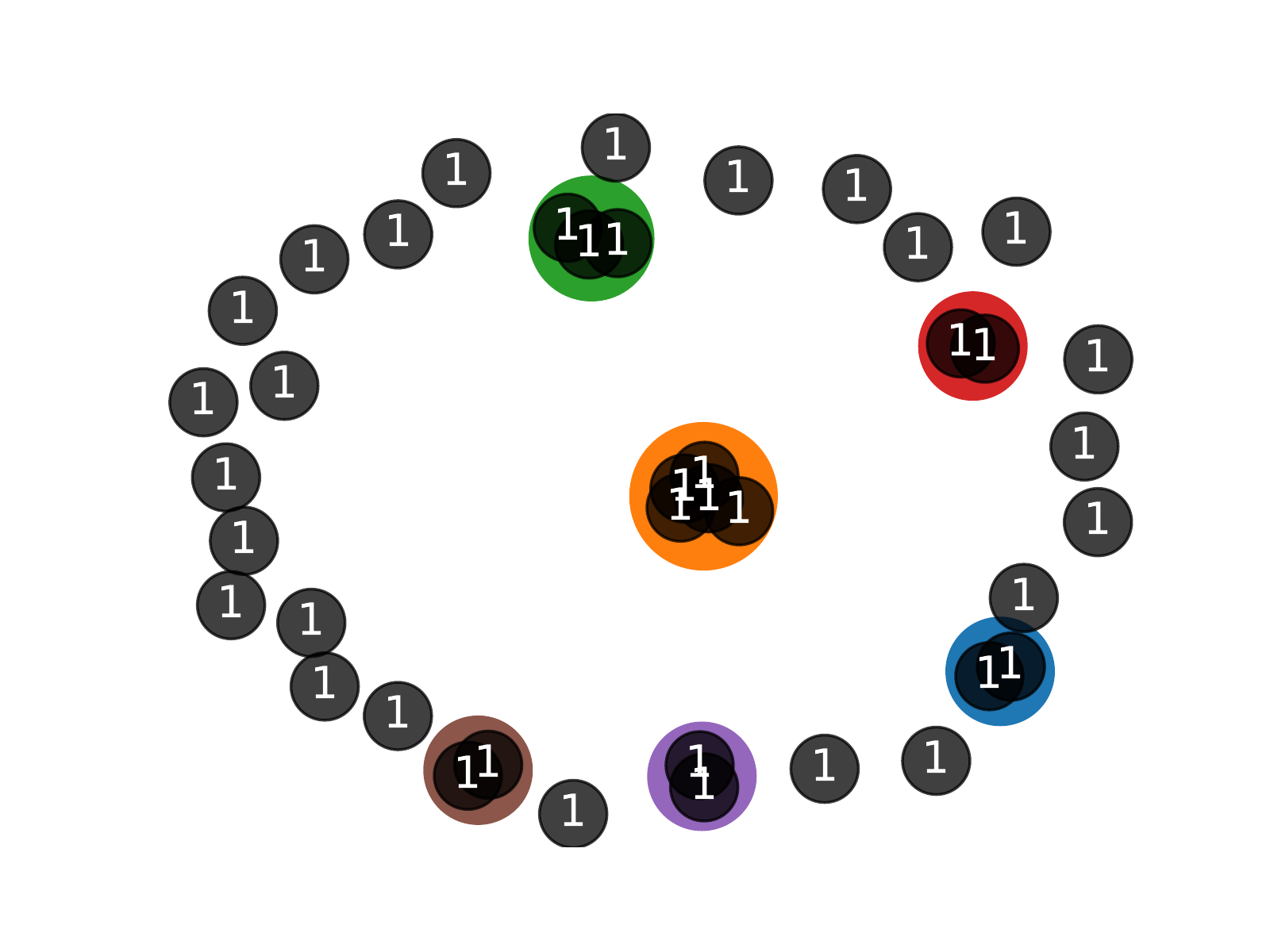}
     \caption{}
     \label{fig:universe_covers_intersection}
   \end{subfigure}
 \centering
   \begin{subfigure}{.48\textwidth}
   \centering
    \includegraphics[width=\linewidth]{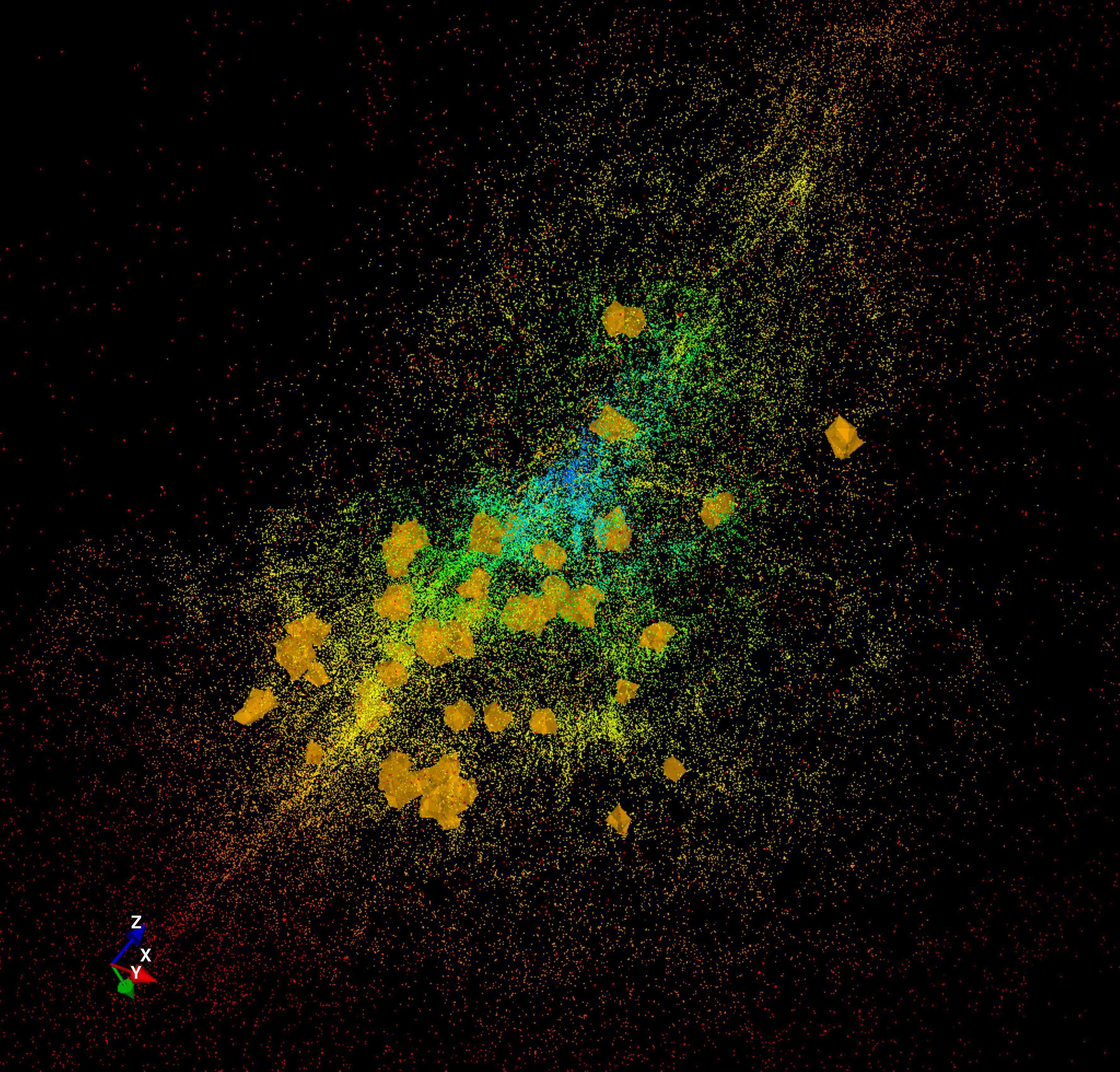}
     \caption{}
   \label{fig:universe_all_voids_view1}
   \end{subfigure}
 \centering
   \begin{subfigure}{.48\textwidth}
   \centering
    \includegraphics[width=\linewidth]{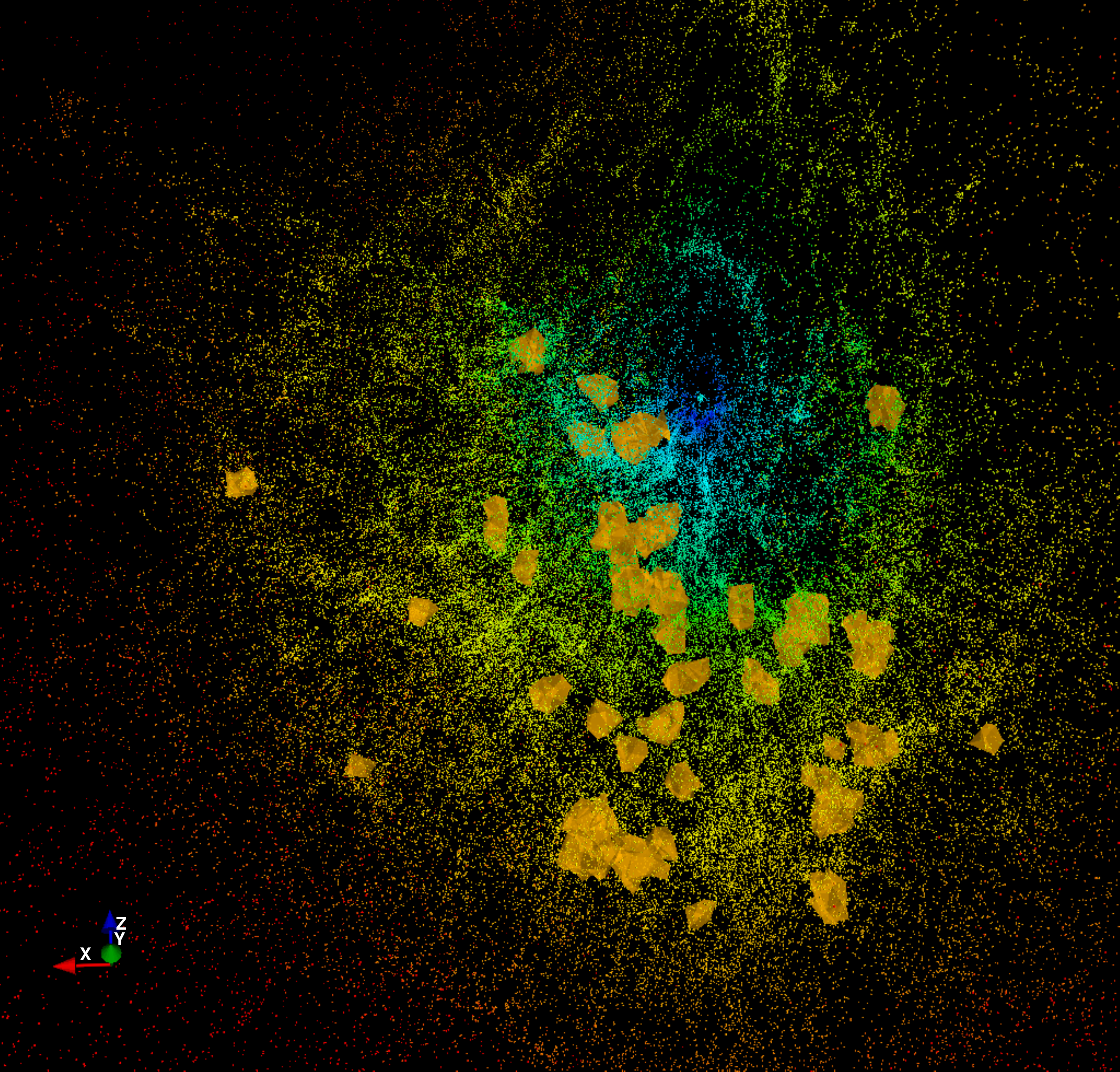}
     \caption{}
   \label{fig:universe_all_voids_view2}
   \end{subfigure}
 \centering
   \begin{subfigure}{.48\textwidth}
   \centering
    \includegraphics[width=\linewidth]{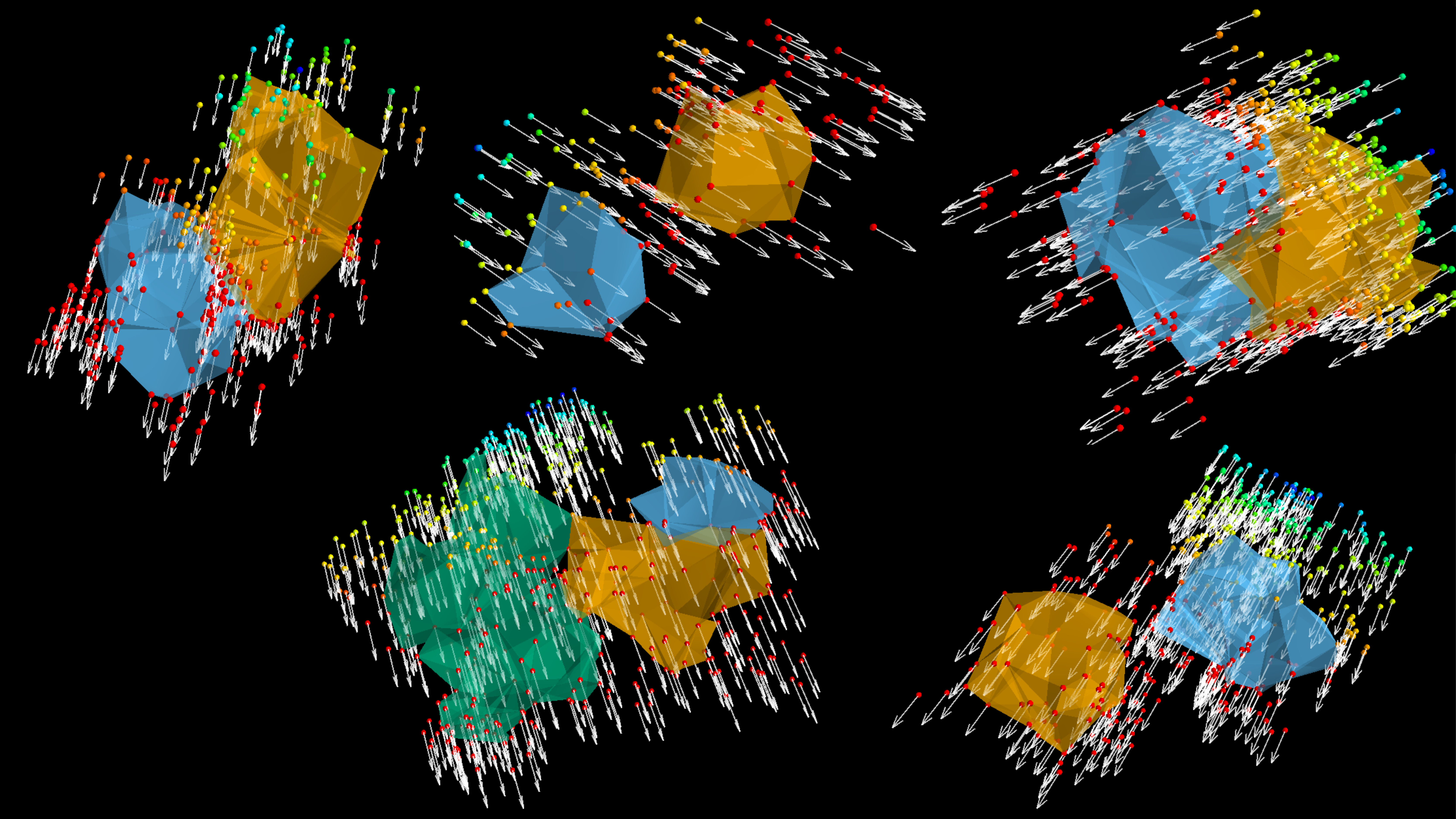}
     \caption{}
   \label{fig:universe_final_multiple_1}
   \end{subfigure}
 \centering
   \begin{subfigure}{.48\textwidth}
   \centering
    \includegraphics[width=\linewidth]{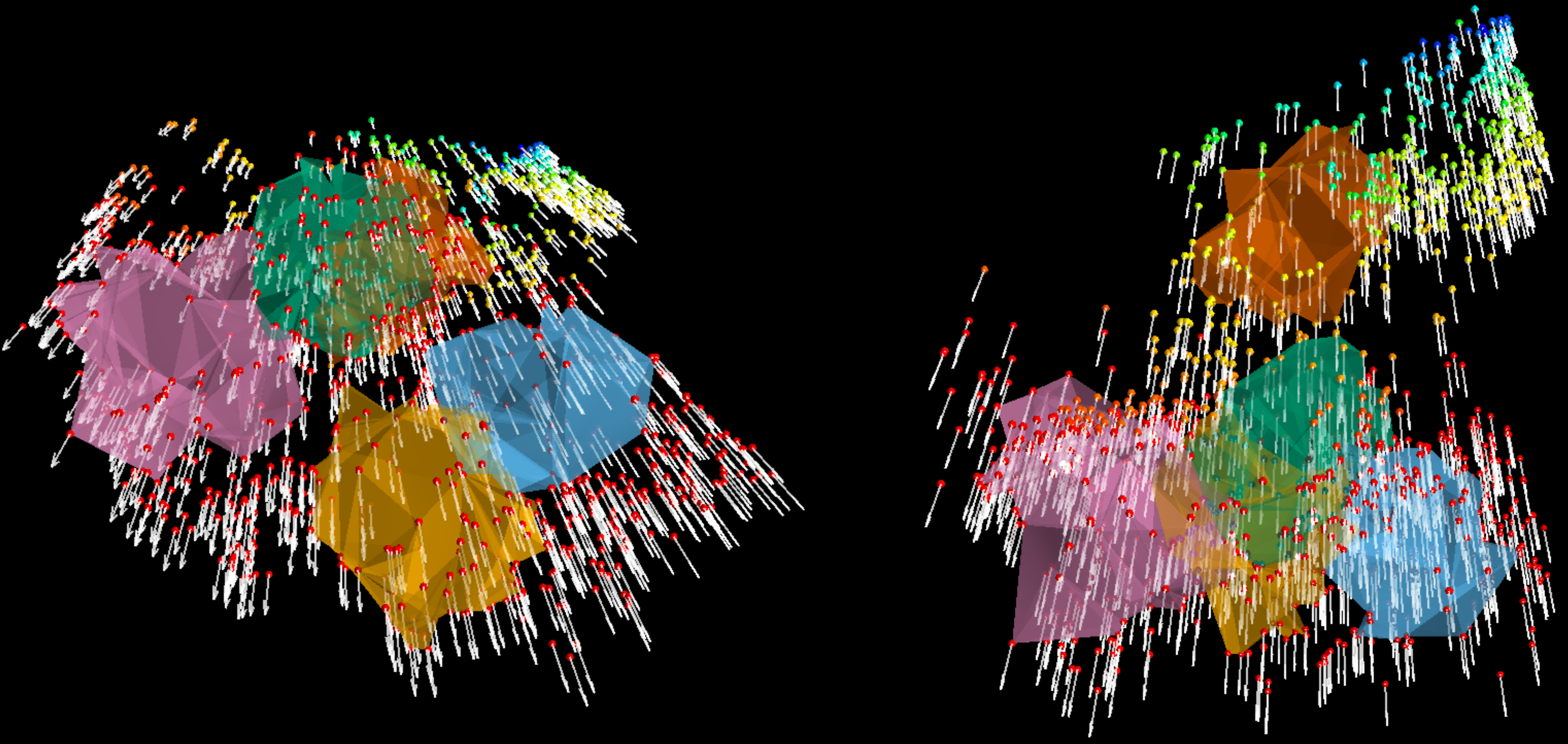}
     \caption{}
   \label{fig:universe_final_multiple_0}
   \end{subfigure}

  \caption{The final computed boundaries identified locations of $40$ distinct voids in the spatial
  embedding. (a) Persistence pairs of significant features in covers of the computed set of minimal
  representatives and of significant features in the full data set. (b) A graph with nodes as covers
  of minimal boundaries. Labels show that each cover has one significant feature. An edge denoted
  non-empty intersection between two covers. $24$ out of $30$ components are singletons. They
  identify distinct regions in the universe, each with one significant void. (b) and (d) Two
  orientations of all computed minimal boundaries around significant H$_2$ voids in the spatial
  embedding of galaxies. (e) $5$ of the components with intersecting covers to visually confirm that
  all minimal representatives are around distinct voids.(f) Last remaining component that has $5$
  intersecting covers and up to $5$ significant voids. Two different orientations show that the $5$
  minimal cycles are around $5$ distinct voids.}

  \label{fig:universe_final}
\end{figure}

We used the computed multiple sets of homology boundaries for all permutations, to define
boundary-simplices that are in minimal representatives (Section~\ref{method:strategy_algorithm}
for details). Do these minimal boundaries identify unique singular voids? They wrapped around only
one significant feature since each cover had one significant feature. Next, we confirmed that they
wrapped around unique voids. We computed their covers and checked whether they have non-empty
set-intersection. Figure~\ref{fig:universe_covers_intersection} shows a graph with nodes as covers
and an edge defining a non-empty intersection. Each node is labeled by the number of significant
features in it. The presence of $24$ singletons in this graph shows that $24$ covers do not
intersect with any other cover. This assures that the corresponding minimal representatives are
around distinct singular voids. There are $6$ components (marked with bigger colored circles) which
indicate non-empty intersection of covers. We visually confirmed that the minimal boundaries in these
components are also around distinct voids (Figures~\ref{fig:universe_final_multiple_1}
and~\ref{fig:universe_final_multiple_0}). Hence, we computed $40$ minimal representatives that are
around distinct significant voids in the distribution of galaxies in the universe.
Figure~\ref{fig:universe_supp_all_voids} shows each of the voids.
Figures~\ref{fig:universe_all_voids_view1} and~\ref{fig:universe_all_voids_view2} show two views of all voids in the embedding of the full data set.
Figure~\ref{fig:universe_final_PD} shows the $40$ significant persistence-pairs in covers of minimal
representatives and the $42$ classified as significant in the full data set. The decrease in the number
of significant features by two is attributed to the aforementioned possibility of a decrease in
persistence as boundaries are shortened. Note that we were able to compute the deaths of all features in
these regions because of the small cover sizes. This was not possible for the full data set because of the large number of galaxies. 

%

\begin{figure}[tbhp!]
 \centering
   \begin{subfigure}{\textwidth}
   \centering
    \includegraphics[width=\linewidth]{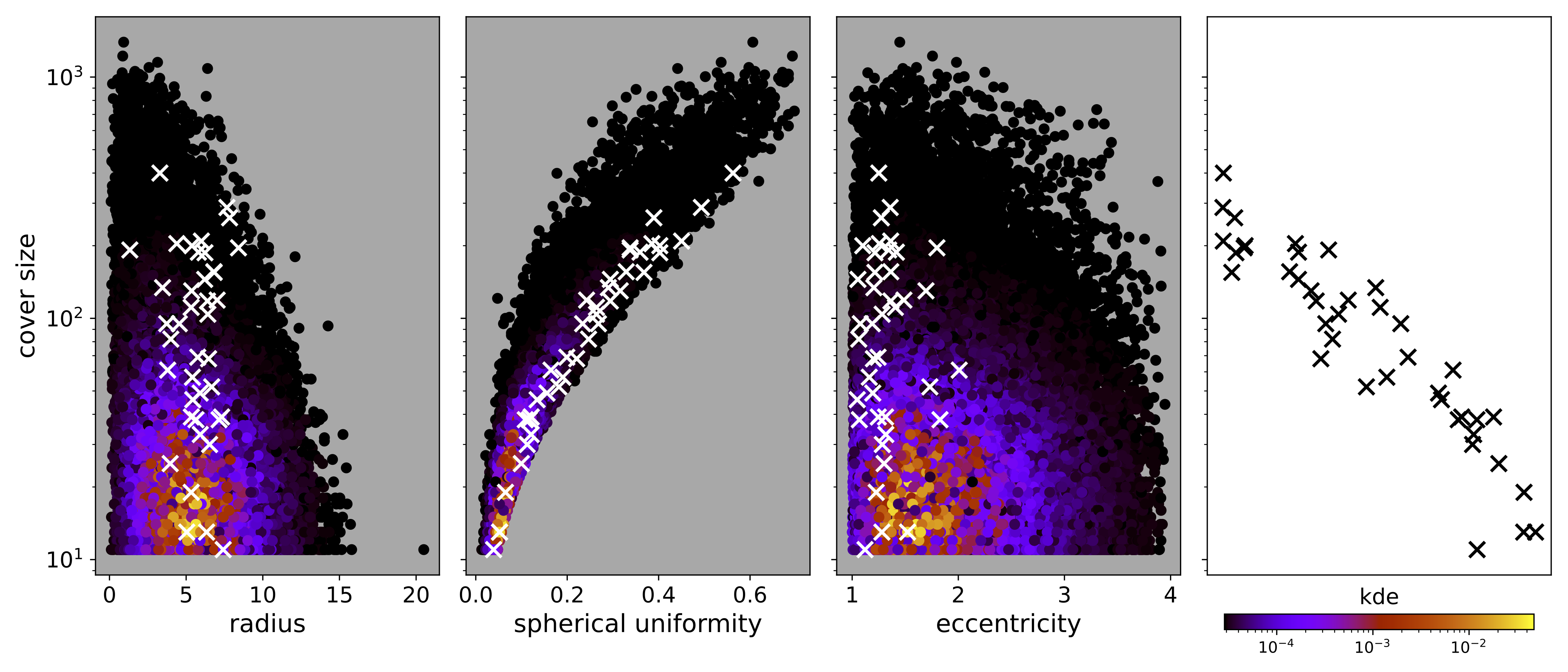}
     \caption{}
   \label{fig:universe_spatial_sampling}
   \end{subfigure}
 \centering
   \begin{subfigure}{\textwidth}
   \centering
    \includegraphics[width=\linewidth]{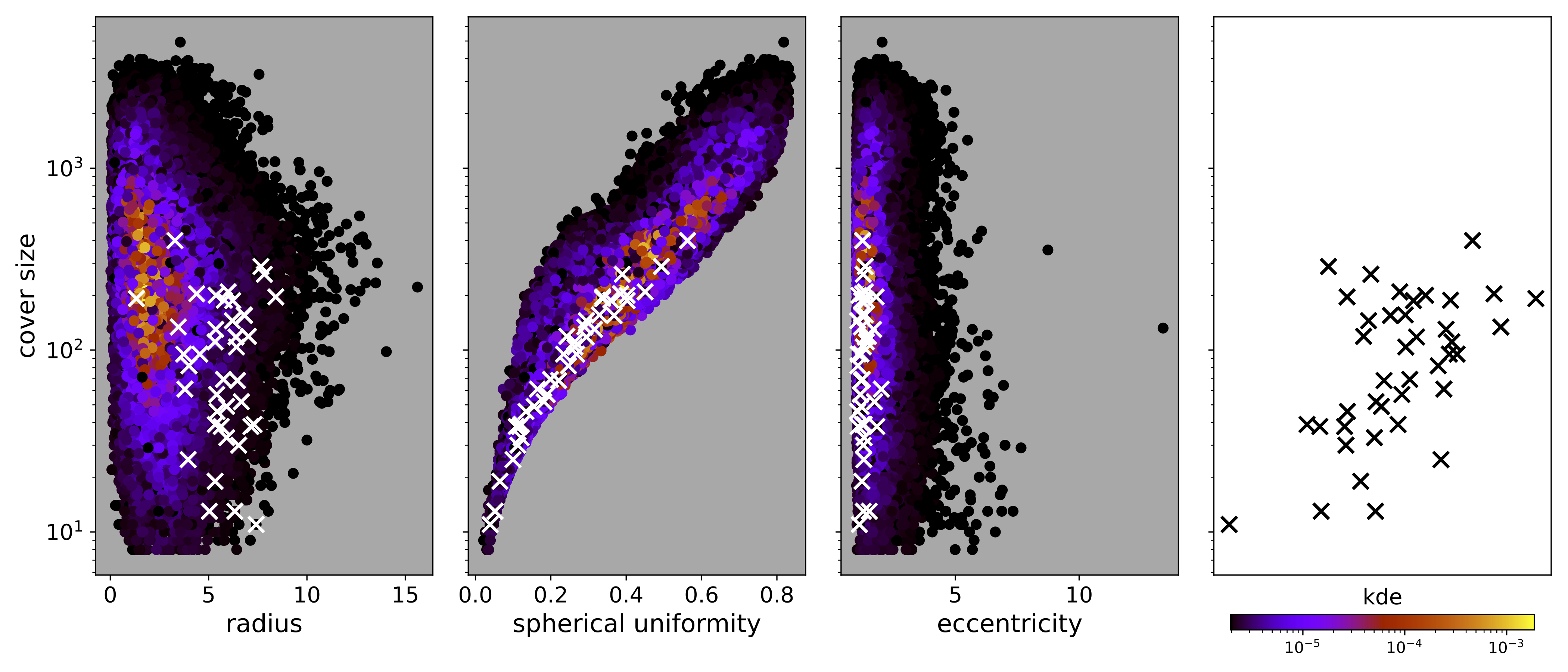}
     \caption{}
   \label{fig:universe_graphical_sampling}
   \end{subfigure}
 \centering
   \begin{subfigure}{.48\textwidth}
   \centering
    \includegraphics[width=\linewidth]{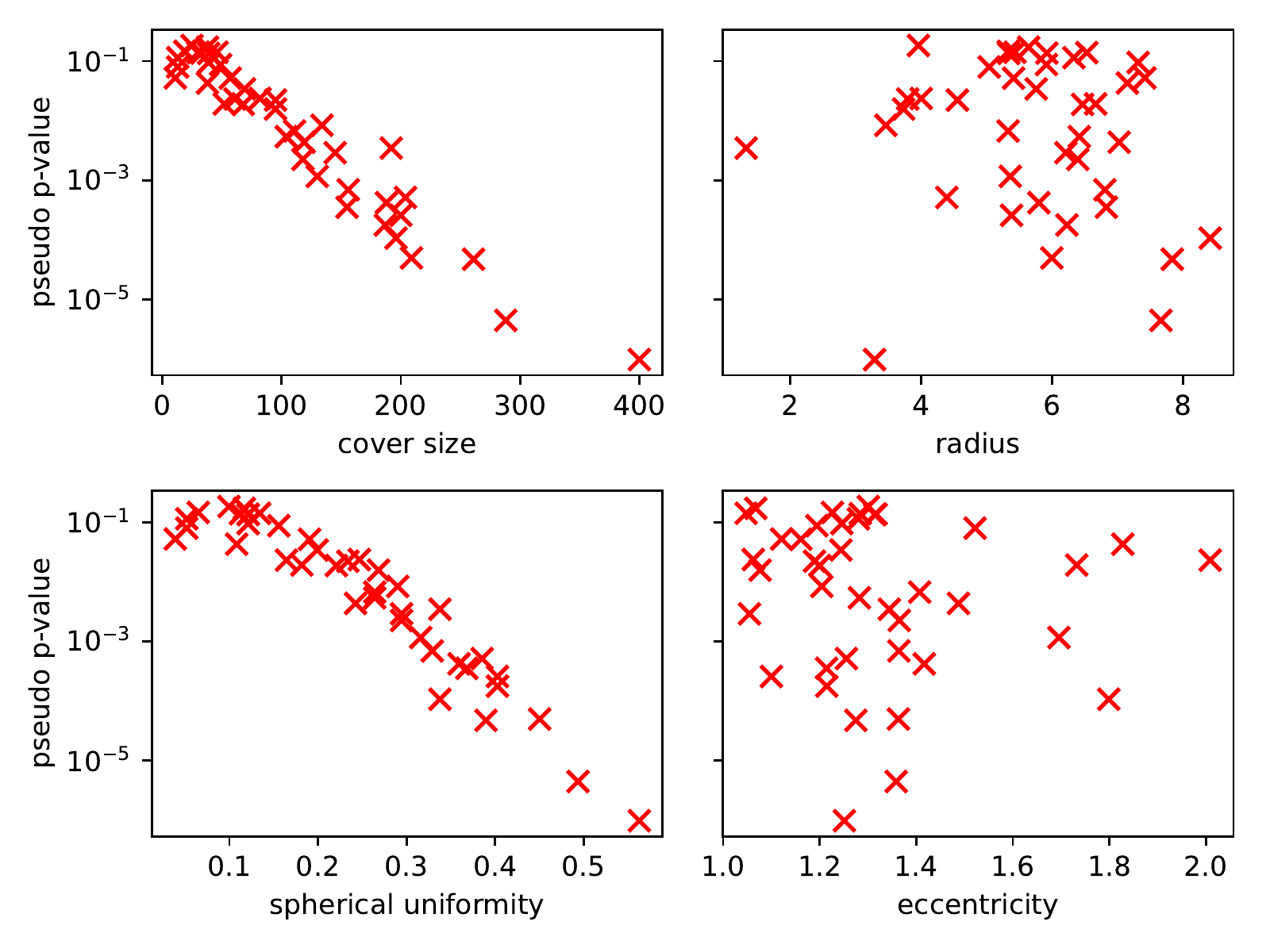}
     \caption{}
   \label{fig:universe_spatial_pval}
   \end{subfigure}
\centering
   \begin{subfigure}{.48\textwidth}
   \centering
    \includegraphics[width=\linewidth]{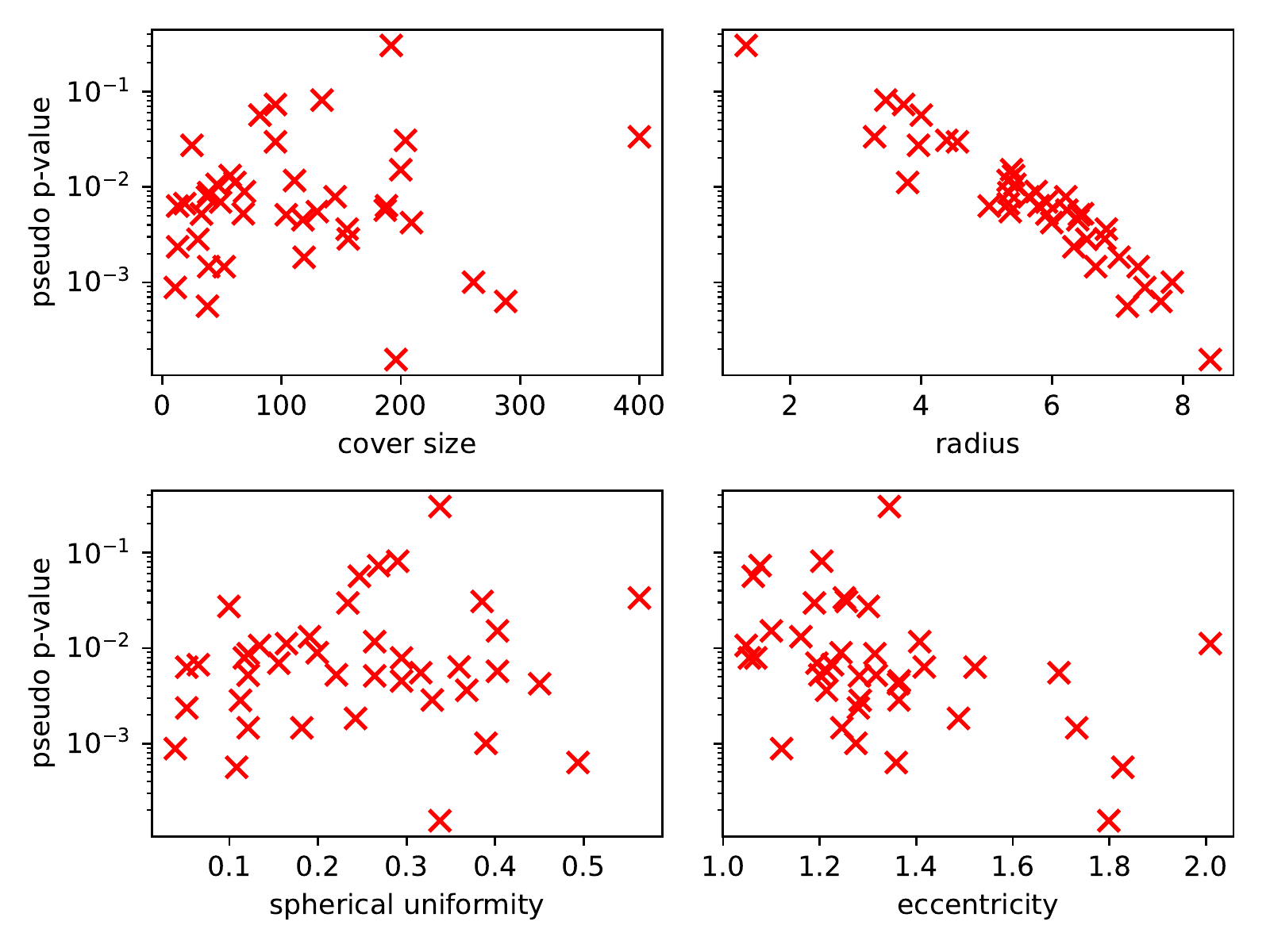}
     \caption{}
   \label{fig:universe_graphical_pval}
   \end{subfigure}

  \caption{Voids determined by minimal representatives have set of features that are rare to find by
  random sampling. $100,000$ samples each of (a) Spatial sampling and (b) Graphical sampling show that
  the feature sets of most voids have very low kde ($0.001$), especially for voids that have large
  covers. Statistical significance of features sets of the computed voids in (c) Spatial sampling and (d) Graphical sampling.}

  \label{fig:universe_random_sampling}
\end{figure}

Next, we show that it is rare to find similar voids at random. We conducted random searches in two
different ways, spatial and graphical (Section~\ref{method:universe} for details). In spatial
sampling, we sampled hyper-rectangles in the embedding. In graphical sampling, we sampled connected
subgraphs of a graph with nodes as galaxies and edges defined for pairs of galaxies 
at most $\tau_u$ distance apart. We defined four features for comparison between random samples and the sets of
minimal representatives---size of their cover, radial distance of galaxy closest to the center of
the cover (units in Mpc), spherical uniformity of the points in the cover, and eccentricity of the cover.
Figure~\ref{fig:universe_random_sampling} shows projections of the four-dimensional feature-set with circle markers for the random samples and the crosses for the voids computed by our strategy. The color of the random samples shows the Gaussian kernel density estimation (kde). We defined and computed a pseudo p-value (see Methods~\ref{method:universe_p_values}) for every computed void that is indicative of statistical significance of its feature set. Figure~\ref{fig:universe_spatial_pval} shows that feature sets with larger cover size are statistically significant (p-value $\leq 0.01$) in spatial sampling. Figure~\ref{fig:universe_spatial_pval} shows that feature sets with larger radius are statistically significant in graphical sampling. Table~\ref{tab:universe_voids} shows features sets of all $40$ voids and the corresponding pseudo p-values. There are $12$ voids with feature sets that are statistically significant in both spatial and graphical sampling.

\begin{table}
\begin{center}
{
\begin{tabular}{|c|c|c|c|c|c|}
\hline
cover size & radius (Mpc) & sph. uni. & eccen. & spatial & graphical\\
\hline
69 & 5.76 & 0.2 & 1.24 & 1.47 & \textbf{2.05}\\
49 & 5.91 & 0.16 & 1.19 & 1.06 & \textbf{2.16}\\
\textbf{187} & \textbf{6.23} & \textbf{0.4} & \textbf{1.21} & \textbf{3.75} & \textbf{2.24}\\
\textbf{145} & \textbf{6.21} & \textbf{0.29} & \textbf{1.05} & \textbf{2.54} & \textbf{2.1}\\
11 & 7.42 & 0.04 & 1.12 & 1.28 & \textbf{3.05}\\
30 & 6.53 & 0.11 & 1.28 & 0.86 & \textbf{2.55}\\
400 & 3.29 & 0.56 & 1.25 & \textbf{6.01} & 1.47\\
192 & 1.33 & 0.34 & 1.34 & \textbf{2.46} & 0.52\\
\textbf{196} & \textbf{8.41} & \textbf{0.34} & \textbf{1.8} & \textbf{3.97} & \textbf{3.81}\\
\textbf{130} & \textbf{5.36} & \textbf{0.32} & \textbf{1.7} & \textbf{2.94} & \textbf{2.26}\\
200 & 5.38 & 0.4 & 1.1 & \textbf{3.59} & 1.82\\
\textbf{261} & \textbf{7.83} & \textbf{0.39} & \textbf{1.27} & \textbf{4.33} & \textbf{3.0}\\
\textbf{119} & \textbf{7.02} & \textbf{0.24} & \textbf{1.49} & \textbf{2.36} & \textbf{2.74}\\
118 & 6.39 & 0.29 & 1.36 & \textbf{2.65} & 2.34\\
39 & 7.31 & 0.12 & 1.25 & 1.02 & \textbf{2.84}\\
\textbf{288} & \textbf{7.66} & \textbf{0.49} & \textbf{1.36} & \textbf{5.35} & \textbf{3.2}\\
82 & 4.0 & 0.25 & 1.06 & 1.63 & 1.25\\
\textbf{155} & \textbf{6.83} & \textbf{0.37} & \textbf{1.21} & \textbf{3.45} & \textbf{2.44}\\
134 & 3.46 & 0.29 & 1.2 & \textbf{2.08} & 1.09\\
61 & 3.79 & 0.16 & 2.01 & 1.64 & 1.95\\
95 & 3.73 & 0.27 & 1.08 & 1.8 & 1.14\\
\textbf{209} & \textbf{5.99} & \textbf{0.45} & \textbf{1.36} & \textbf{4.31} & \textbf{2.37}\\
13 & 6.33 & 0.05 & 1.28 & 0.94 & \textbf{2.63}\\
111 & 5.33 & 0.26 & 1.41 & \textbf{2.17} & 1.93\\
19 & 5.33 & 0.06 & 1.23 & 0.83 & \textbf{2.17}\\
\textbf{104} & \textbf{6.42} & \textbf{0.26} & \textbf{1.28} & \textbf{2.27} & \textbf{2.29}\\
68 & 6.46 & 0.22 & 1.2 & 1.73 & \textbf{2.28}\\
\textbf{156} & \textbf{6.81} & \textbf{0.33} & \textbf{1.36} & \textbf{3.16} & \textbf{2.54}\\
\textbf{188} & \textbf{5.8} & \textbf{0.36} & \textbf{1.42} & \textbf{3.38} & \textbf{2.2}\\
52 & 6.67 & 0.18 & 1.73 & 1.72 & \textbf{2.84}\\
38 & 7.15 & 0.11 & 1.83 & 1.37 & \textbf{3.25}\\
38 & 5.64 & 0.12 & 1.07 & 0.77 & \textbf{2.1}\\
46 & 5.43 & 0.13 & 1.05 & 0.85 & 1.97\\
204 & 4.39 & 0.39 & 1.26 & \textbf{3.29} & 1.51\\
95 & 4.55 & 0.23 & 1.19 & 1.65 & 1.53\\
13 & 5.04 & 0.05 & 1.52 & 1.1 & \textbf{2.2}\\
25 & 3.96 & 0.1 & 1.3 & 0.74 & 1.56\\
57 & 5.41 & 0.19 & 1.16 & 1.29 & 1.88\\
33 & 5.92 & 0.12 & 1.32 & 0.86 & \textbf{2.28}\\
39 & 5.34 & 0.12 & 1.31 & 0.87 & \textbf{2.06}\\
\hline
\end{tabular}
}
\end{center}
\caption{Features of voids computed by PH. Last two columns are $-\text{log}_{10}$ of the p-value, and those $\geq 2$ (or p-value $\leq 0.01$) are highlighted. Entire rows of voids with statistical significance in both spatial and graphical sampling are highlighted.}

\label{tab:universe_voids}
\end{table}

\subsection{Protein homologs with significantly different topology}

\begin{figure}[tbhp!]
 \centering
   \begin{subfigure}{.33\textwidth}
   \centering
    \includegraphics[width=\linewidth]{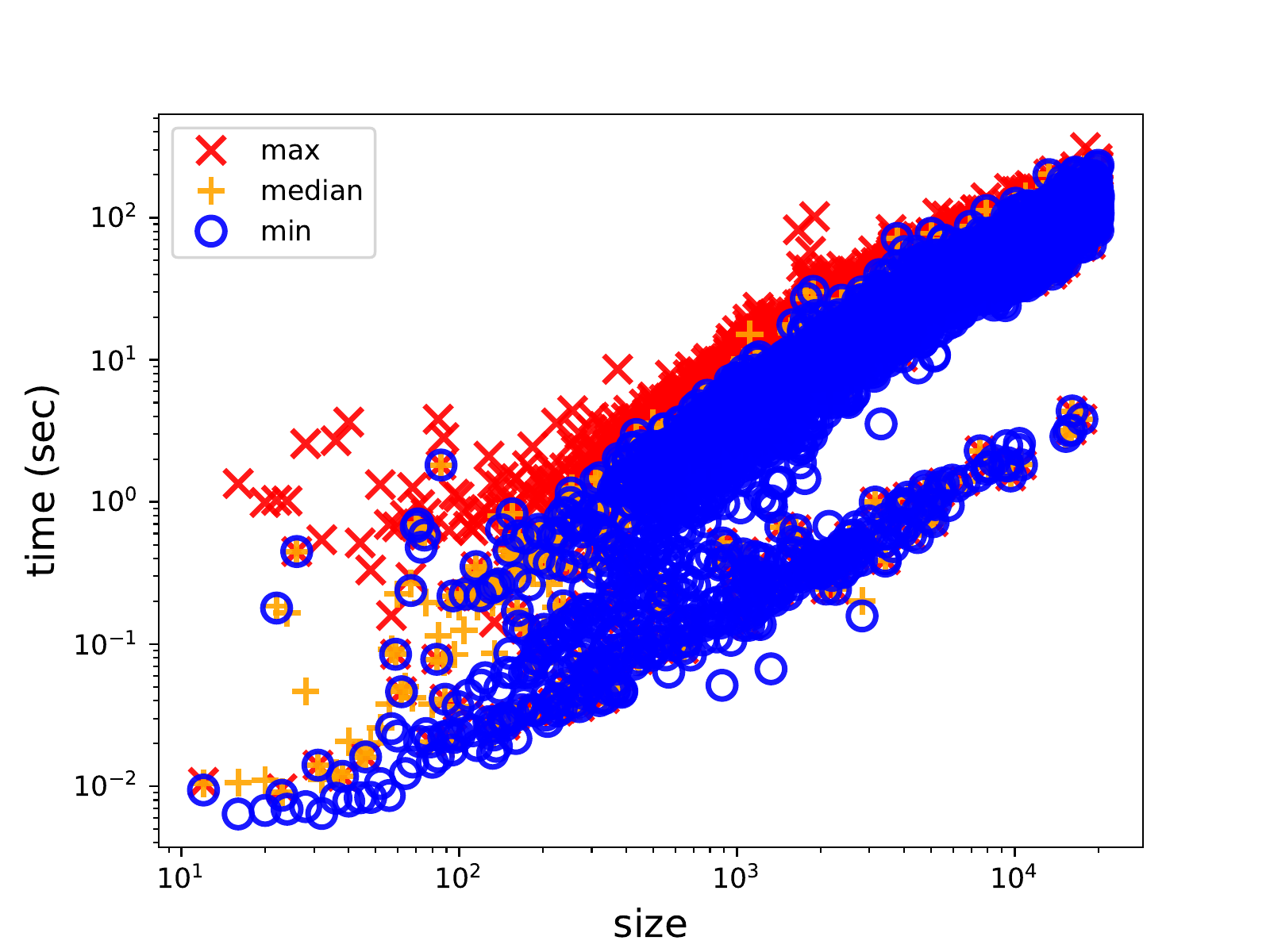}
     \caption{}
   \label{fig:protein_PD_computation}
   \end{subfigure}
 \centering
   \begin{subfigure}{.33\textwidth}
   \centering
    \includegraphics[width=\linewidth]{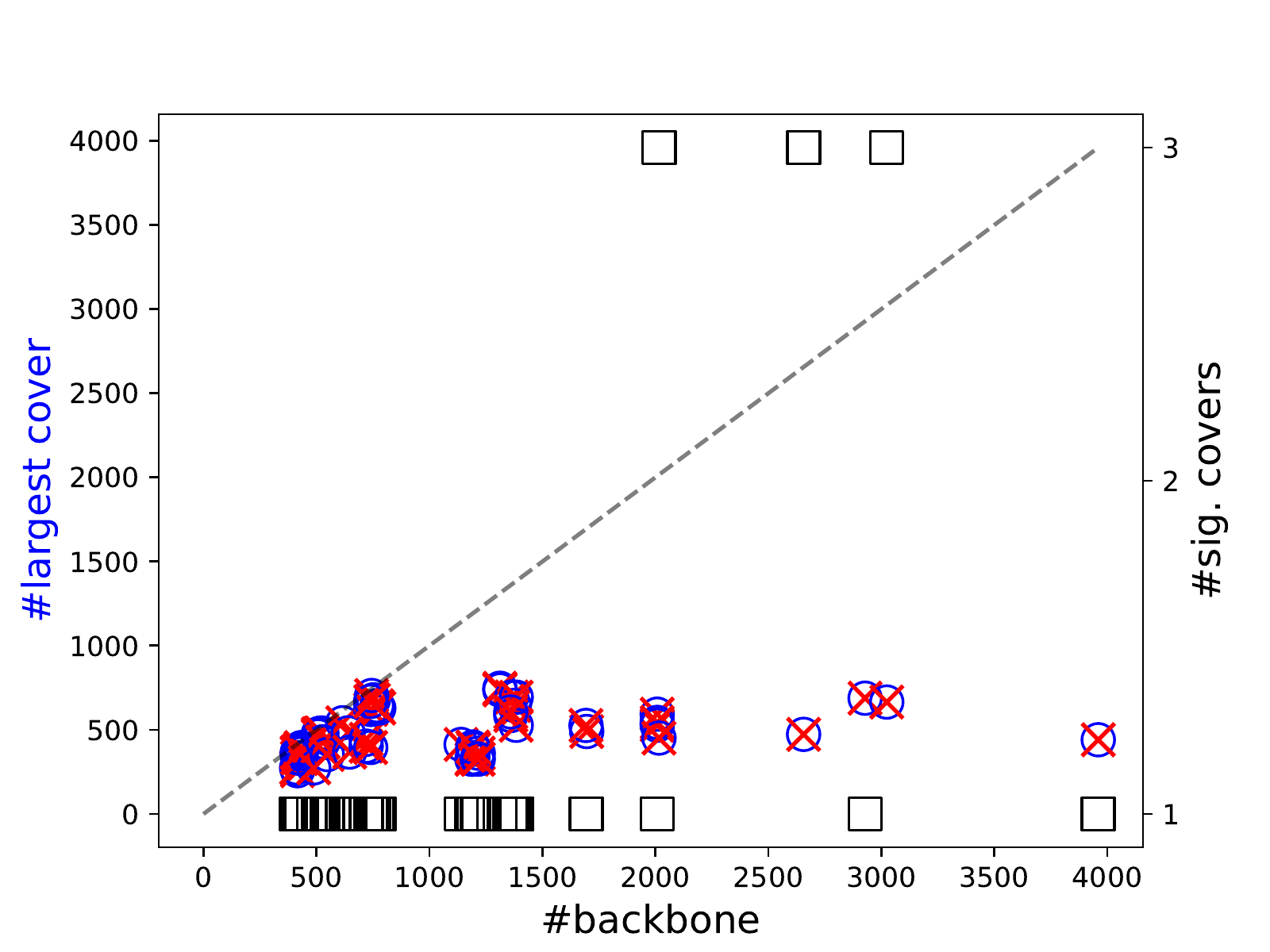}
     \caption{}
   \label{fig:protein_full_vs_largest_cover}
   \end{subfigure}
 \centering
   \begin{subfigure}{.33\textwidth}
   \centering
    \includegraphics[width=\linewidth]{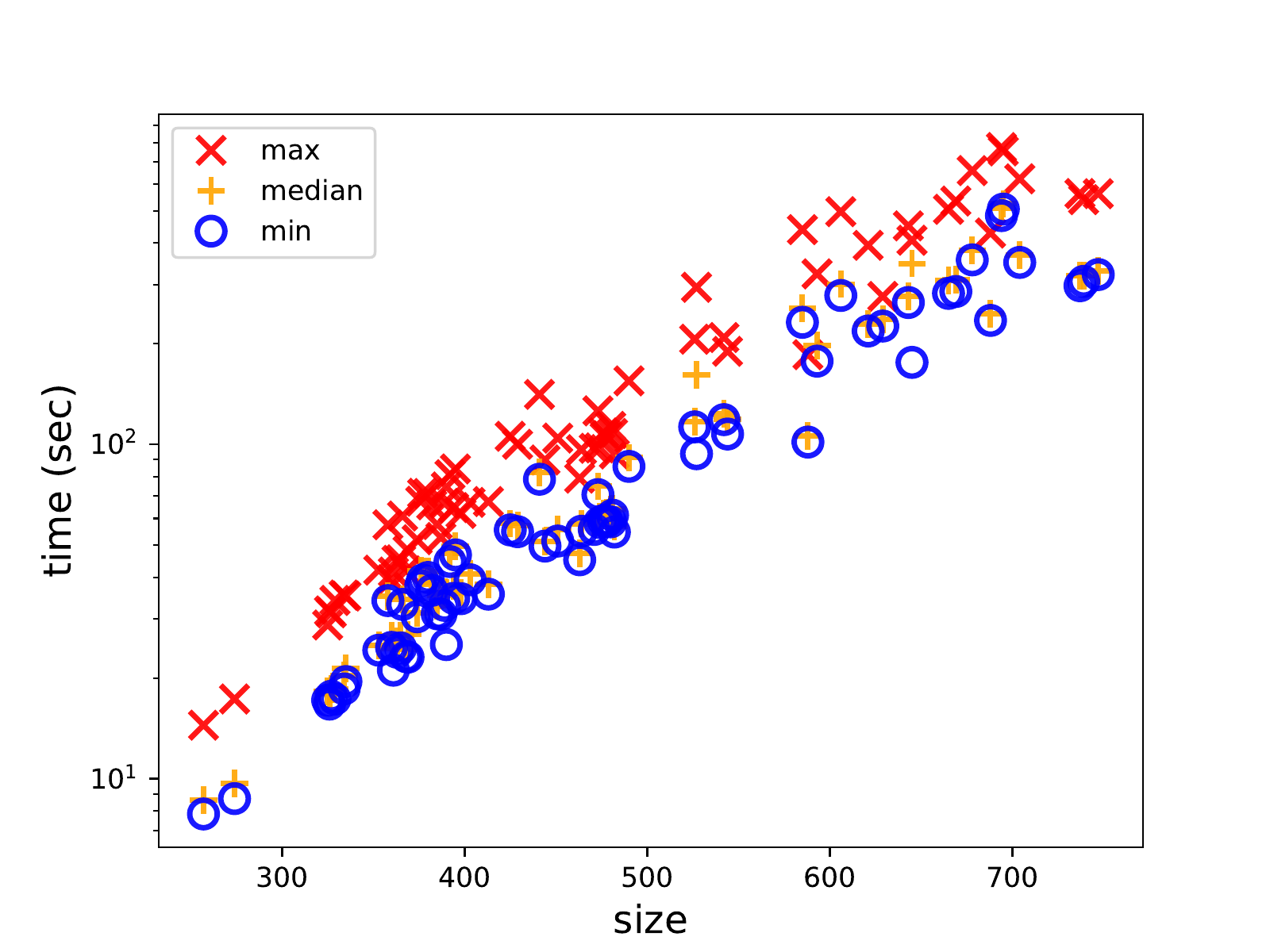}
     \caption{}
   \label{fig:protein_computation_rep_hom_77}
   \end{subfigure}
 \centering
   \begin{subfigure}{0.48\textwidth}
   \centering
    \includegraphics[width=\linewidth]{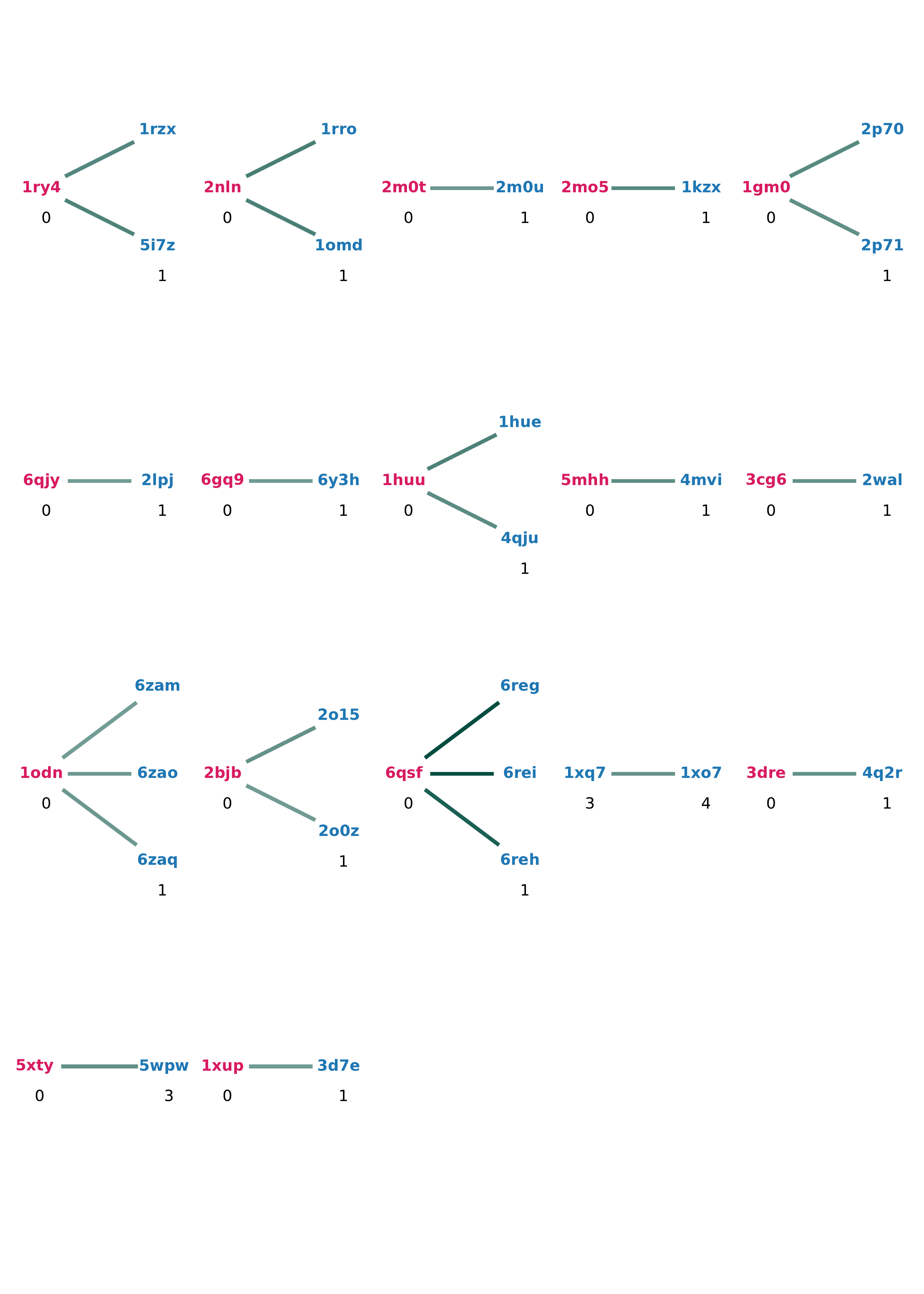}
     \caption{}
   \label{fig:protein_all_homolog_graph_1}
   \end{subfigure}
 \centering
   \begin{subfigure}{0.48\textwidth}
   \centering
    \includegraphics[width=\linewidth]{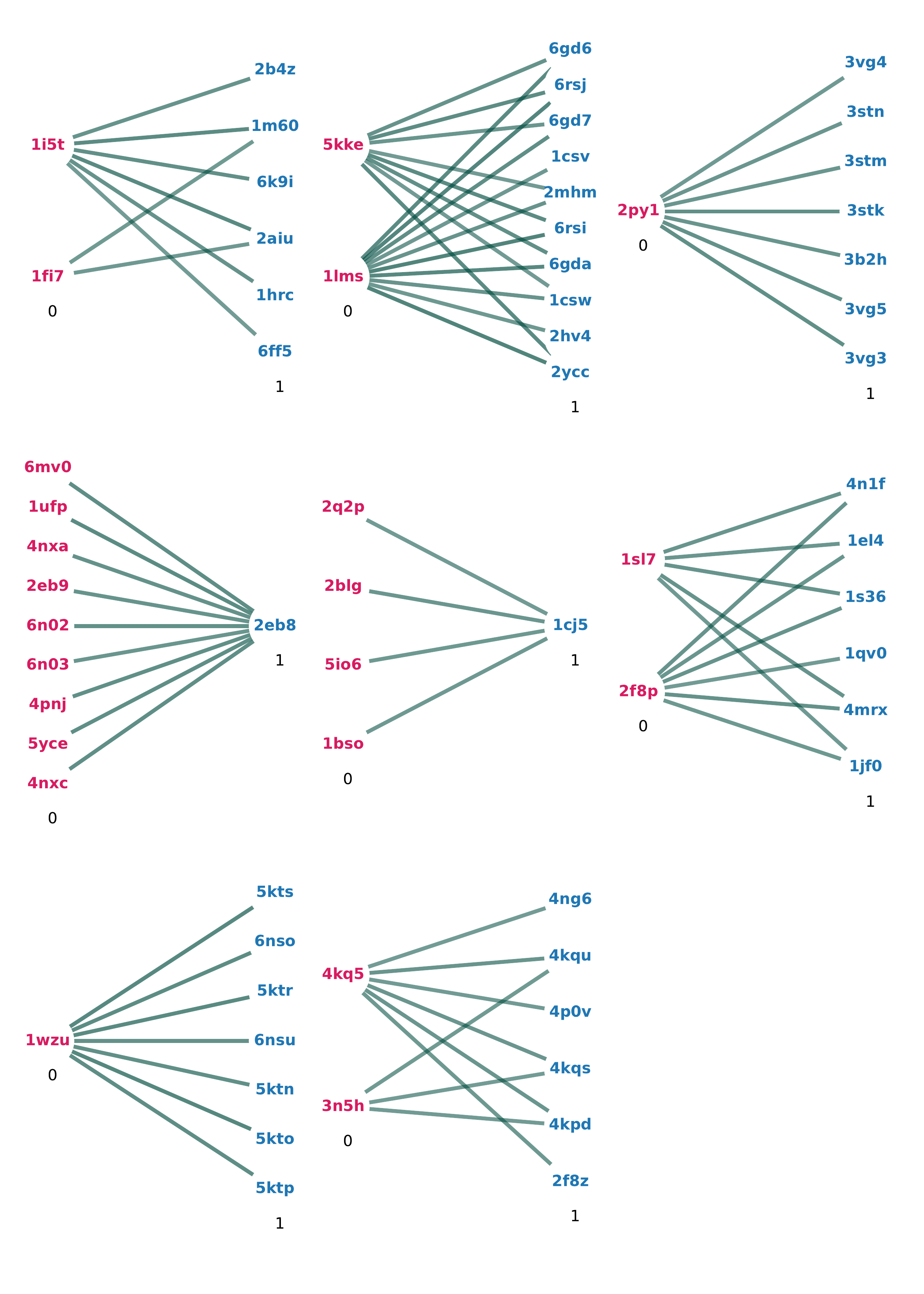}
     \caption{}
   \label{fig:protein_all_homolog_graph_2}
   \end{subfigure}

  \caption{Homologs with significant topological differences (a) PD computation of 174,574 PDB
  entries up to and including threshold $\tau = \tau_u + \epsilon = 10 + 3.5$ (b) $71$ proteins in
  $25$ homolog graphs that have significant features. There are $77$ covers with significant
  features after graphical contraction. The largest of the covers around smooth cycles is less than
  $1000$ for every protein. (c) Run-times for computing representative homology boundaries in
  permutations of perturbations of $77$ covers. (d) and (e) $25$ homolog graphs (different \#sig.
  features and L$_0$ norm of at least $3$). Black labels are the number of significant H$_2$ features in every PDB in that column.}

  \label{fig:protein_computation}
\end{figure}

Can we find protein homologs with differences in the number of significant H$_2$ features in their
backbone? We classified voids in protein structures as significant if 
they are born at a spatial scale of at most $10$ and if their persistence is at least $3.5,$ i.e.\ $\tau_u=10$ and $\epsilon = 3.5.$ These choices are
based on the fact that the average bond lengths in protein molecules ranges from around $1.5$ {\AA}
to $3$ {\AA}. We considered all publicly available Protein Data Bank (PDB) entries with at least $10$ and at most
$20000$ atoms in their backbone (N, C, C$\alpha$, and O atoms). PH of the resulting $174,574$ valid
PDB entries was computed for the backbone, up to and including threshold of $\tau = \tau_u +
\epsilon = 13.5,$ on x$2650$ processors. Figure~\ref{fig:protein_PD_computation} shows the
computation times against the number of atoms in the backbones. Out of these, $12198$ PDB entries
had at least one significant H$_2$ feature. For each of these, we queried for PDB entries that
matched at least $75\%$ of length with a score of at least $75\%$ and the length of backbones
differed by at most $25\%.$ 
It is generally presumed that sequences with high similarity did not arise independently and share a common ancestor~\citep{pearson2013introduction}. Therefore, we call the matching entries homologs. We determined $874$ unique
sets of homologs. A
higher value of L$_0$ norm in barcode lengths indicates that the topological differences are more significant (see
Section~\ref{method:proteins} for definition). We iterated over pairs of PDB structures in each set of homologs to determine those with a
different number of significant features and with L$_0$ norm in barcode lengths at least $3.$ We constructed a new graph with edges as all pairs
that satisfied these criteria. Figures~\ref{fig:protein_all_homolog_graph_1} and~\ref{fig:protein_all_homolog_graph_2} show the $25$ components of this graph. Overall, there are $110$ nodes or PDB entries. Out of these, $71$ had at least one
significant void. The largest of these proteins had around $4000$ atoms in its backbone.  First, we
computed the shortened and smoothed birth-cycles. Figure~\ref{fig:protein_full_vs_largest_cover}
shows that the largest set in covers of smoothed cycles has less than $1000$ points (open \textbf{o} in
Figure~\ref{fig:protein_full_vs_largest_cover}). Hence, again, the computational cost of any subsequent
analysis was significantly reduced as compared to the largest backbone. Graphical contraction resulted in $77$ covers that
possibly contain significant voids. The  number of significant voids are shown by open squares in
Figure~\ref{fig:protein_full_vs_largest_cover}. The size of the largest of these covers in each PDB
entry is marked by red \textbf{x}. The overlapping \textbf{x}'s and open \textbf{o}'s show that
graphical contraction did not result in further decrease in size of the covers for subsequent
analysis.

We implemented the stochastic analysis with $\text{n}_\text{pert}$ and $\text{n}_\text{perm}$ chosen
as $15.$ We were able to compute PH and representative homology boundaries.
Figure~\ref{fig:protein_computation_rep_hom_77} shows the variation in computation times. The
maximum time taken was around $15$ mins. Note that, the computation time is not strictly increasing
with increase in the number of points in the data set. We constructed the set of minimal
representatives for significant voids in every protein, as detailed in our strategy. All of the $25$
homolog graphs along with 3D figures showing protein crystal structure and voids are in
Supplementary figures~\ref{fig:pdb_1} to~\ref{fig:pdb_25}. All boundaries are around distinct
singular voids.

\begin{figure}[h]

 \centering
   \begin{subfigure}{.48\textwidth}
   \centering
    \includegraphics[width=\linewidth]{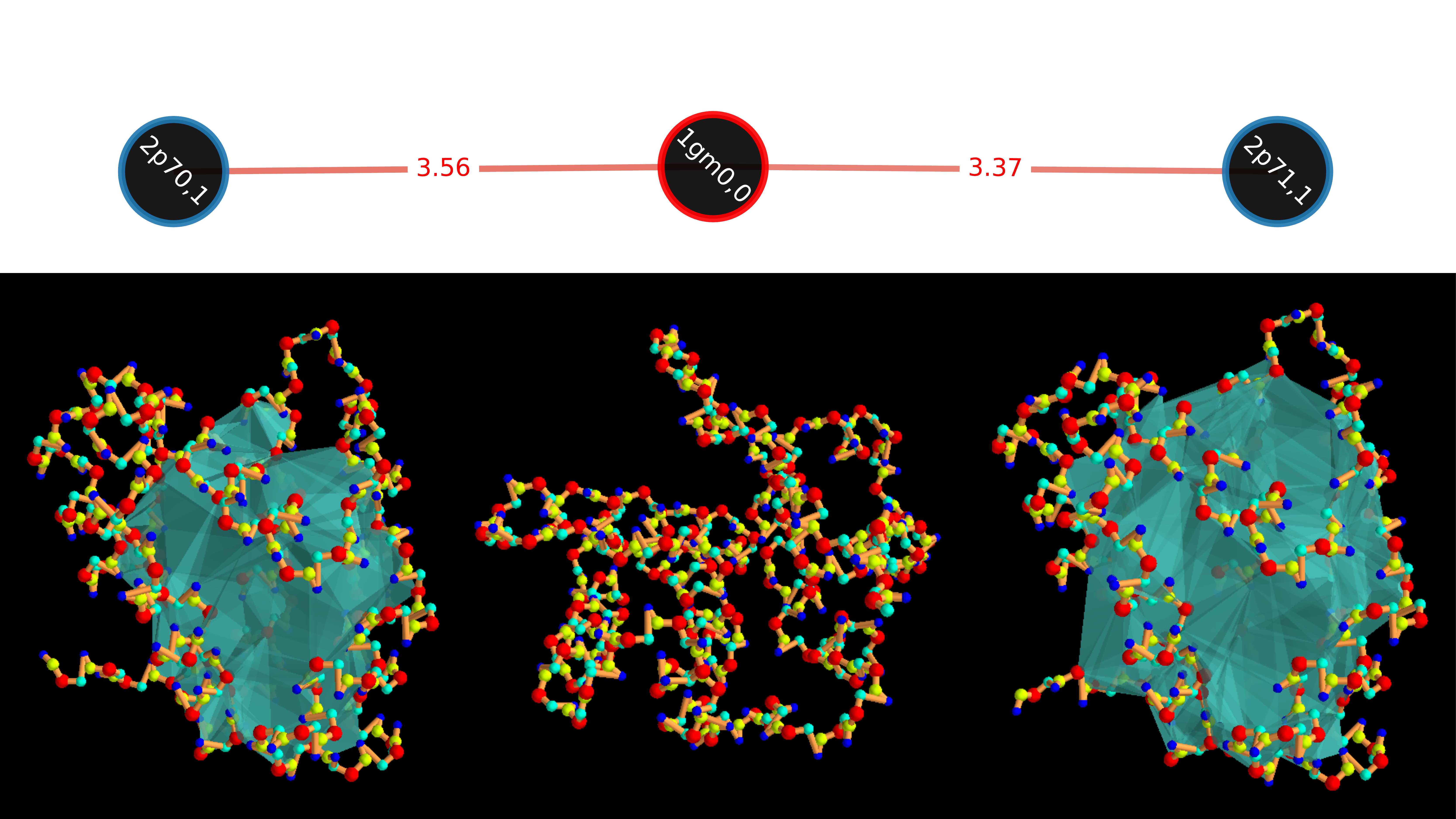}
     \caption{}
   \label{fig:homolog_example_3d_1}
   \end{subfigure}
 \centering
   \begin{subfigure}{.48\textwidth}
   \centering
    \includegraphics[width=\linewidth]{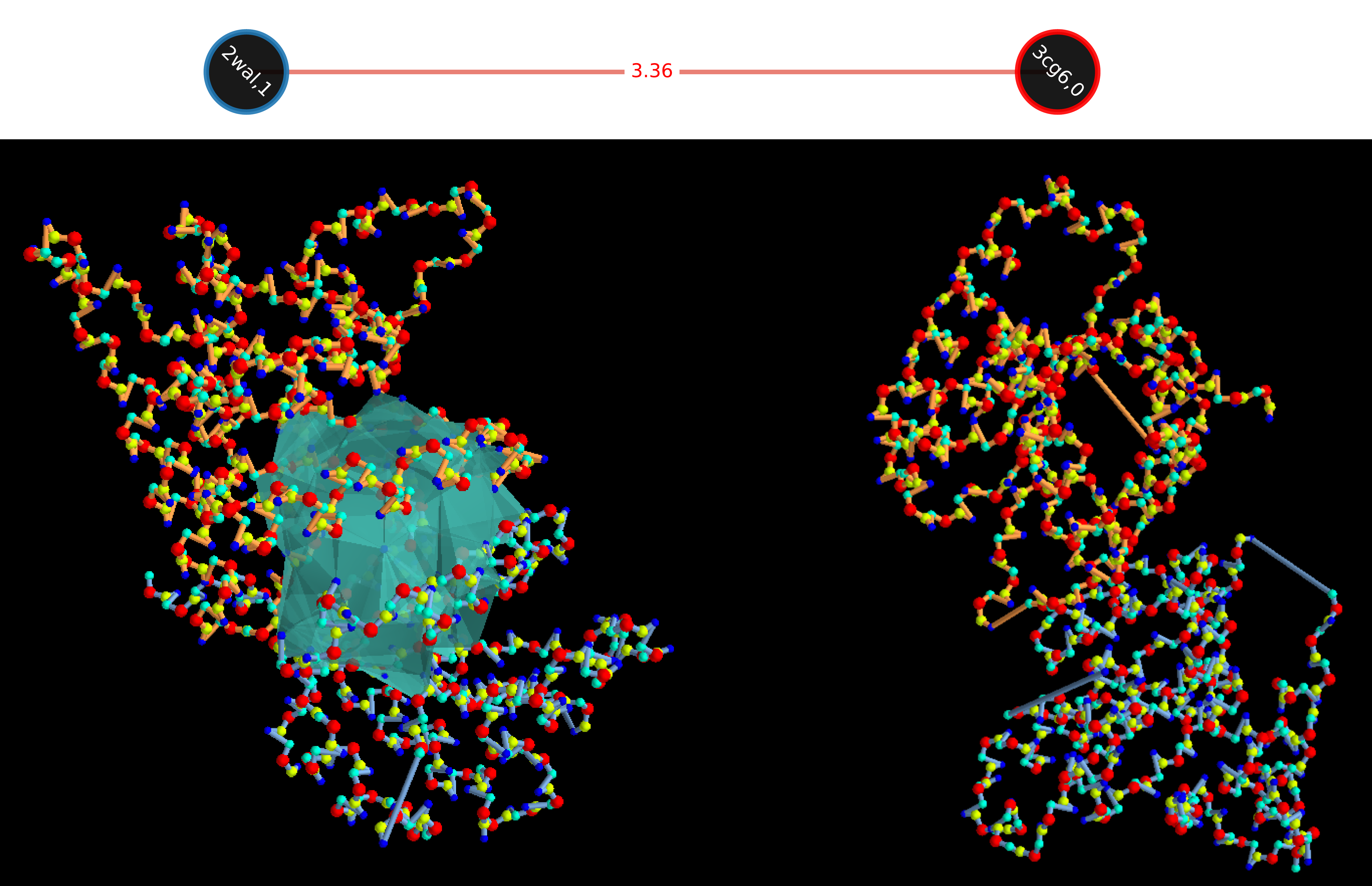}
     \caption{}
   \label{fig:homolog_example_3d_2}
   \end{subfigure}
 \centering
   \begin{subfigure}{.48\textwidth}
   \centering
    \includegraphics[width=\linewidth]{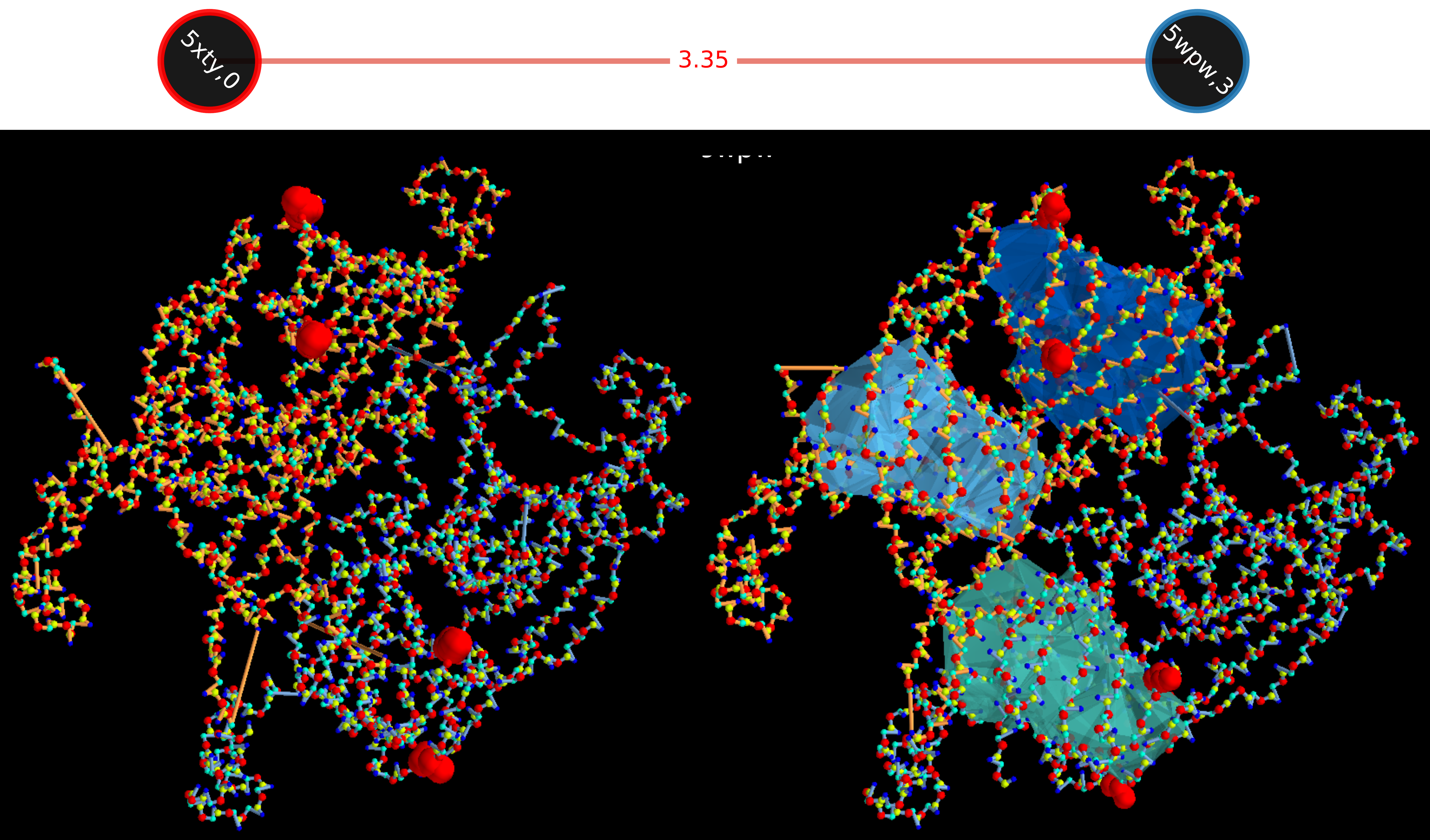}
     \caption{}
   \label{fig:homolog_example_3d_3}
   \end{subfigure}

  \caption{Three selected examples from the 25 homolog graphs. (a) Ligand-binding: 1gm0 is a form of
  pheromone-binding protein from \textit{Bombyx mori} that has no significant voids. Its bound
  configurations---with bell pepper odorant (2p70) and with iodohexadecane (2p71)---have significant
  voids at the binding sites. (b) Different species: Dimeric configuration of GADD45g in humans has
  a void, but not in mice. (c) Mutation: Two mutations (large red markers) in each of the two chains
  of protein Cocosin, introduce three significant voids.}

  \label{fig:pdb_3d_examples}
\end{figure}

We showcase three examples here. First is an example of ligand-binding
(Figure~\ref{fig:homolog_example_3d_2}). The pheromone-binding protein of \textit{Bombyx mori} has no
significant voids in its unbounded form. However, it has significant voids when bound with bell
pepper odorant (PBD 2p70) and also when bound with iodohexadecane (PBD 2p71). Second is an example
of different topology between different species. The growth arrest and DNA damage (GADD45) genes is
a highly conserved family of proteins (GADD45a, GADD45b, and GADD45g) that respond to stress on
mammalian cells and have a crucial involvement in DNA repair. We found that a dimeric conformation of
GADD45g from humans (PBD 2wal) has a significant void, but the dimeric configuration reported for
mouse (PBD 3cg6) does not have any voids (Figure~\ref{fig:homolog_example_3d_2}). This
structural difference might lead to differences in gene expression. For example, a closely related
gene in the same family, GADD45a, is up-regulated in humans but down-regulated in mice upon
irradiation~\citep{ghandhi2019discordant}. Third is an example of mutation leading to a different
topology. Cocosin, a protein in coconut fruit, is a possible food allergen. Its reported
crystal structure (PDB 5xty) has no significant voids. However, it has $3$ significant voids when
mutated in two residues in each of its two chains (PDB 5wpw). Figure~\ref{fig:homolog_example_3d_3}
shows the 3D crystal structures with the voids. The mutated regions are shown with large red markers.

\section{Discussion}\label{sec:discussion}

PH is an immensely popular method of TDA because it is based on clear mathematical foundations and it is computable with, essentially, basic linear algebra. The existence of nontrivial robust holes in noisy experimental
observations, as computed by PH, has found applications across diverse areas of research. However,
computing the locations of holes has not received much attention because of the non-uniqueness of their boundaries and high
computational cost. This information, unfortunately, is crucial for understanding the functional significance of nontrivial features. This is especially important in scientific contexts, as we have demonstrated in three disparate areas. In
this work, we provided a set of algorithms and strategy to compute locations of nontrivial holes
with high precision. Moreover, we were able to process large data sets with millions of points.

Our method is based on the established matrix reduction algorithm that reduces boundary or
coboundary matrices to compute PH. The following is a summary of our contributions and the main ideas of our
strategy: We use duality between cohomology and homology to reduce the computation cost of reducing
the boundary matrix. We show that the reduction operations of the boundary matrix form a
comprehensive set of representative boundaries in every scenario, that we call birth-cycles. We
developed a recursive algorithm that computes reduction operations for the boundary of any given simplex
on the fly. This eliminated memory overhead since the entire matrix of reduction operations is never
stored in memory. It was additionally optimized in both memory usage and run-time by selective
storage of reduction operations of some simplices. It may be interesting to explore whether these
simplices are related to critical simplices in Morse theory. The next major algorithm is the greedy
shortening algorithm that updates the set of representative boundaries with shorter ones. This
algorithm is optimized by division into different cases. Greedy shortening decreased lengths of
representatives significantly in our case studies, shortening them by multiple log-scales in some
data sets. We also locally smooth these boundaries. If an embedding of the data set is available, we
use the shortened and smoothed boundaries to find subsets of the complete data set that contain
significant features.  In our case studies, the number of points in these subsets was much less than
the number of points in the full data set. This reduced the computational cost of subsequent analysis.
We add stochasticity in two different ways to possibly find shorter boundaries in the embedding of
these subsets. From the multiple computed sets of boundaries, we construct a set of minimal representatives. It is important to note here that our introductions of stochasticity were analytical tools, in the same way that randomization can improve the performance of sorting algorithms, and in no way was the data reduced or sampled.

We suggest some alternative algorithm design choices that could be explored and lead to improvements on our methodology. For
example, we have defined the cover of an embedded set of points as the smallest hyper-rectangle that
contains the points on it or inside it. A tighter cover of the points in the embedding might be to
fit an oriented bounding box by computing eigenvectors of the covariance matrix of embedded points. However, it will
incur a higher computational cost in comparison to the current choice of cover. As another example,
the perturbation parameter of a cover is based on the number of significant features being the same
across all of its perturbations. This is a weak check to ensure that the topology across
perturbations does not change significantly. A stronger check could be to use the L$_0$ norm between two
PDs that we defined for our case study of proteins. However, the threshold of L$_0$ norm at which
two PDs will be judged as being significantly different, will have to be defined as a
hyperparameter. We generally avoided increasing the number of hyperparameters. An even stronger check
could be to compute metrics like the bottleneck and Wasserstein distances between two
PDs. However, in addition to requiring a threshold as a hyperparameter that defines significantly
different PDs, these are computationally expensive.

We list some technical limitations and possible resolutions as avenues for future work. All
algorithms have been designed with maximal size of data structures as a linear
factor of $O(n_e),$ where $n_e$ is the number of edges in the final simplicial complex. However, the
size of the reduction operations of the coboundary matrix and the size of the reduced boundary
matrix depend upon the reduction operations and cannot be predetermined. They are stored as
one-dimensional sparse matrices. Their maximum permissible size in our current implementation is the
limit of $\texttt{unsigned int}$, $l = $ 4,294,967,295. This limit was reached when we attempted to
compute PD of the HIV capsid (PDB 3j3q). One possible resolution is to increase the maximum permissible size of
our data structures from $l$ to $l^2$ by using two-dimensional sparse matrices, such that an overflow
continues data entry to a new row. Another limitation is the dependence of our definition of covers
on availability of a spatial embedding. It is possible to define covers using only the pairwise
distances. A comparison of computational benchmarks and results for different definitions of covers
may be useful. A third limitation is the assumption that it is feasible to compute PH up to the
maximum possible threshold for covers of shortened and smoothed boundaries. If that is not feasible, then we
cannot compute representative homology boundaries for the cover(s). Instead, we can compute birth-cycles by
computing PH up to and including threshold of $\tau=\tau_u + \epsilon$ for
all permutations of all perturbations. A strategy will have to be developed that will construct a
set of minimal representatives from the multiple sets of shortened and smoothed birth-cycles.

To illustrate its applications, we provided three cases studies from real data sets across diverse
scientific fields. First, we computed loops in genome wide human genome at high resolution of $1$
kb. We found that auxin affects trans-cycles that go through chromosome 13 and the sex chromosomes differently, as compared to all other chromosomes. Second, we computed voids in a distribution of more than 108,030 galaxies in the
universe. We showed that the computed voids have properties that are statistically rare, and therefore unlikely to be a product of chance.
Third, we computed voids in proteins homologs with different number of significant H$_2$ features. We highlighted three examples that showed the difference in significant voids related to ligand
interaction, mutation, and difference in species. We visually confirmed that the representatives
that we computed for voids in the universe and in protein structures wrap tightly around singular
unique voids. In conclusion, we believe that this work enables research into the functional
significance of robust features in a plethora of scientific data sets.

\section{Methods}

\subsection{Computing boundaries}\label{method:compute_boundaries}

We describe our algorithms and strategy to compute minimal boundaries around all significant holes
(loops and voids) in a discrete data set, $\mathcal{U},$ with all pairwise distances available.

\subsubsection{Background and terminology}\label{method:PH_terminology}

\begin{figure}[h]
  \centering
  \includegraphics[scale=0.75]{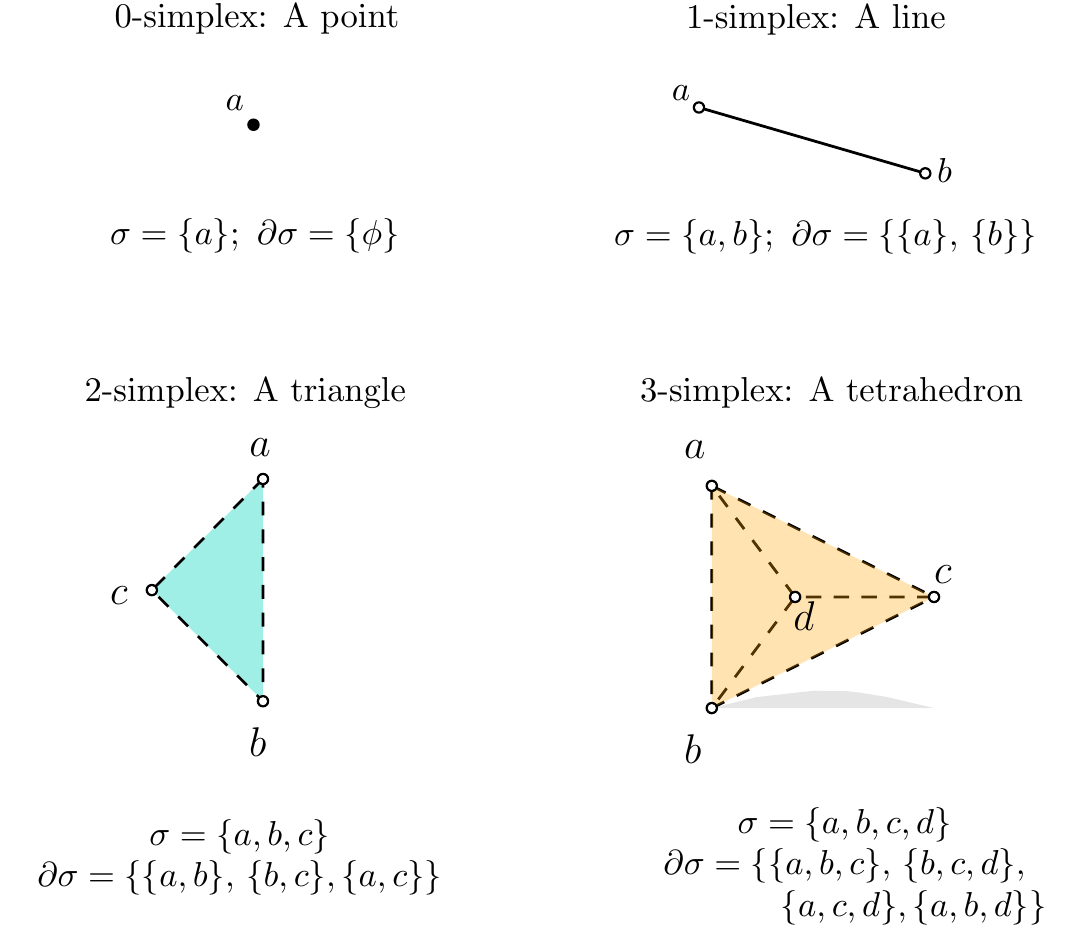}
  \caption{Simplices}
  \label{fig:supp_simplices}
\end{figure}

A $n$-simplex is a set of $(n+1)$-points (supplementary Figure~\ref{fig:supp_simplices}). We
require 0, 1, 2, and 3-simplices to compute PH up to and including H$_2.$ These simplices are also
called vertices, edges ($e$), triangles ($t$), and tetrahedrons ($h$), respectively. We define
\textit{diameter} of a simplex as the maximal pairwise distance between its points. A simplicial 
complex at a spatial scale of $\tau$ is defined as the collection of all simplices with diameter at most $\tau.$
Topologically distinct loops and voids in a simplicial complex are computed as basis elements of its
homology groups H$_1$ and H$_2,$ respectively~\citep{hatcher2001algebraic}. Persistent homology
computes changes in the number of basis elements as the spatial scale of observation, $\tau,$
increases~\citep{edelsbrunner2008persistent}. These changes are represented as \textit{birth} and
\textit{death} of basis elements of the homology groups. A (birth, death) pair is also called a
persistence pair. Nontrivial persistence pairs are plotted with birth along the $x$-axis and death along
the $y$-axis. These plots are called persistence diagrams (PD). A feature with higher
\textit{persistence} $= \text{death} - \text{birth},$ will be robust to larger variability in the
data set.

The matrix reduction algorithm~\citep{edelsbrunner2008persistent} is an established method to compute PD. 
It also yields a set of representative boundaries. We briefly describe the method and refer to
Figures~\ref{fig:PH_filtration} and~\ref{fig:PH_matrix_reduction} as an example. Simplices are
denoted by $\sigma_i,$ where $1 \leq i \leq N$ is the index of the simplex. They are indexed in the
order in which they are added to the simplicial complex. Hence, they are indexed in the order of increasing
diameters. Those with the same diameter are assigned unique indices arbitrarily. The \textit{boundary
matrix} $D$ is constructed for a simplicial complex as follows. Row and column $i$ of $D$
correspond to simplex $\sigma_i$ in the simplicial complex. Column $i$ of $D$ has $1$ at the
boundary-simplices of $\sigma_i$ (blue boxes in~\ref{fig:PH_matrix_reduction}), and is $0$ (gray
boxes) otherwise. Boundary-simplices of a $n$-simplex are all $(n-1)$-simplices that can be
constructed using its points. We denote the column corresponding to a simplex $\sigma_i$ in a matrix $M$
by $M(\sigma_i).$ Every non-zero column of $D$ has a lowest non-zero element (shown by red boxes),
called a \textit{low}. The matrix reduction algorithm then dictates to \textit{reduce} columns of $D,$
such that each row of $D$ has at most one low (red box). Each reduction operation is a sum of columns
mod $p,$ where $p$ is a prime number. In this work, $p=2.$ The reduction of $D$ is specifically from
left to right as follows. A column of $j$ of $D$ is reduced only when all columns $i<j$ have been reduced.
Also, column $j$ is reduced only with a column $i<j.$ This is written as a matrix multiplication,
$DV=R,$ where $V$ records the reduction operations for columns of $D$ that result in the reduced
matrix $R.$ Note that $V$ is always an upper triangular matrix because reductions operations are
from left to right. There are two kinds of persistence pairs that are determined from $R.$ First, a
low (red box) in $R$ at $(i, j)$ implies that a feature was born when $\sigma_i$ was added to the
simplicial complex and it died when $\sigma_j$ was added. This persistence pair is denoted by 
$(\sigma_i, \sigma_j).$ It is also called a pivot element of $R.$ Also, $R(\sigma_j)$ is a 
representative homology cycle for this feature. Second, if $R(\sigma_i) = \mathbf{0}$ and $\sigma_i$ 
is not in any low of $R$ ($\sigma_8$ in the example), then the persistence pair $(\sigma_i, \infty)$ is 
a feature that was born when $\sigma_i$ was added, but it does not die. Such features might exist 
when PH is computed up to a spatial scale that is less than the maximum of all pairwise distances. 
There is no representative boundary in $R$ for such a feature. Another way to compute  exactly the
same PD is by reducing the coboundary matrix, $D^\bot.$ It is the off-diagonal transpose of $D.$ 
Hence, column and row $i$ of $D^\bot$ correspond to simplex $\sigma_{N-i+1}.$ Similar to the reduction of 
boundary matrix, the coboundary matrix $D^\bot$ is also reduced from left to right and written as 
$D^\bot V^\bot = R^\bot.$ Persistence pairs from $R^\bot$ and $R$ have
a bijective mapping~\citep{de2011dualities}. If $(\sigma_j, \sigma_i)$ is a pivot of $R^\bot,$ 
then $(\sigma_i, \sigma_j)$ is a pivot of $R.$ The pair $(\sigma_i, \sigma_j)$ is a persistence pair
with birth when $\sigma_i$ is added to the complex and death when $\sigma_j$ is added to the
complex. If $R^\bot(\sigma_i) = \mathbf{0}$ and there is no pivot in row for $\sigma_i,$ then
$R(\sigma_i) = \mathbf{0}$ and there will be no pivot in the row for $\sigma_i$ in $R.$ Hence, the
pair $(\sigma_i, \infty)$ is a feature that does not die. In the example there are two H$_1$ features, 
$(\sigma_9, \sigma_{10}) \equiv (2.75, 2.75)$ and $(\sigma_8, \infty) \equiv (2.5, \infty).$ Non-zero 
columns of $R$ form a set of representative boundaries. In the example, $\{R(\sigma_{10}) = \{\sigma_5,
\sigma_6, \sigma_9\}\}$ is the only H$_1$ representative homology boundary. However,
there are two H$_1$ features. Hence, columns of $R$ do not form a comprehensive set of boundaries
when there are features that do not die.

\subsubsection{Our strategy and algorithms}\label{method:strategy_algorithm}

The aim is to compute the tightest dense boundaries that surround robust topological features. We quantify these
conditions by two user-defined thresholds. First is a lower limit on persistence, denoted by
$\epsilon.$ It defines the desired robustness. Second is an upper threshold for birth of features,
denoted by $\tau_u.$ A lower value of $\tau_u$ is expected to restrict the results to denser
boundaries. This is because any point in an H$_1$ representative that is born at a spatial scale
$\leq \tau_u,$ will have at least two neighbors within spatial distance of $\tau_u.$ Similarly, a
point in an H$_2$ representative with birth less than $\tau_u,$ will have at least three neighbors 
within $\tau_u.$ Hence, we define a hole with $\text{birth} \leq \tau_u$ and $\text{persistence} \geq \epsilon$ 
as significant. Now, persistence of a feature $(b, \infty)$ is at least $\tau - b,$ where $\tau$ is the threshold
for PH computation. The threshold for PH computation then has to be at least $\tau = \tau_u + \epsilon.$
Otherwise, a feature born at $b \in (\tau_u - \epsilon, \tau_u]$ that does not die might be
erroneously considered as insignificant if its persistence is estimated as $\tau - b.$

We first compute the persistence pairs by reducing the coboundary matrix $D^\bot$ to compute $R^\bot.$
From the bijective mapping between pivots of $R^\bot$ and $R,$ we know that $R(\sigma_i) \neq 0$ if
$\sigma_i$ is in some pivot of $R^\bot.$ Hence, we reduce all such $D(\sigma_i).$ As a result, we
determine exactly which columns of $D$ need reduction. This significantly reduces the run-time of
computing $R.$ For an even lower run-time, we implement reduction of columns of $D$ using
paired-indexing and serial-parallel algorithm that we developed for coboundary matrix
reduction in our previous work (supplementary~\ref{supp:final_hom_computation}). The default batch-size for
parallel reduction of boundaries is chosen to be 1000.

\begin{figure}[tbhp]
  \centering
\begin{subfigure}{.48\textwidth}
  \centering
  \includegraphics[width=0.9\linewidth]{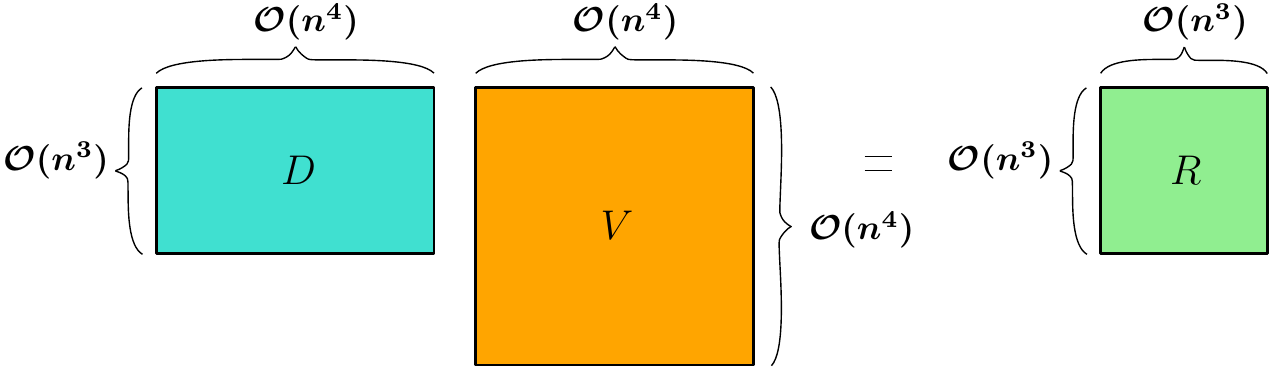}  
  \caption{}
  \label{fig:hom_matrix_size}
\end{subfigure}
  \centering
\begin{subfigure}{.48\textwidth}
  \centering
  \includegraphics[width=0.9\linewidth]{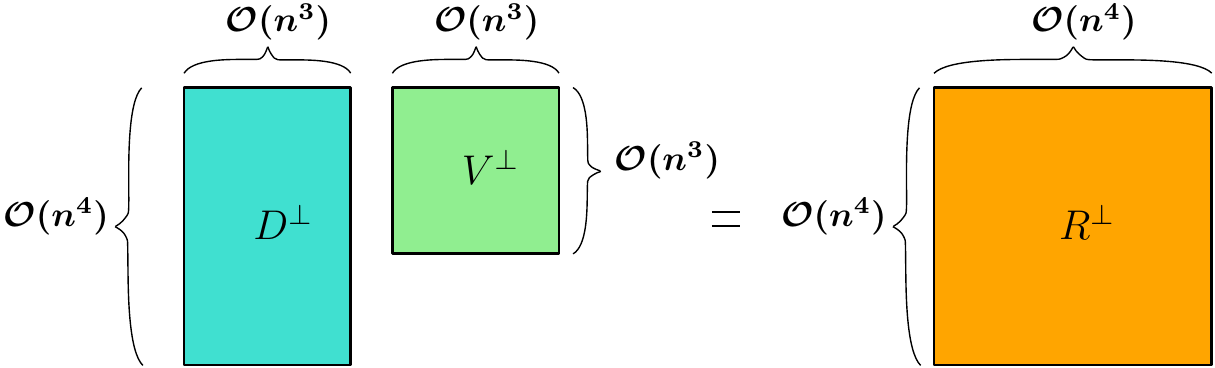}  
  \caption{}
  \label{fig:cohom_matrix_size}
\end{subfigure}

  \caption{Matrix representation and their bounds in PH computation, up to and including H$_2,$ for
  a data set with $n$ points. (a) Homology computation by reducing boundaries. (b) Cohomology
  computation by reducing coboundaries.}

\label{fig:matrix_sizes}
\end{figure}

Next, we compute a comprehensive set of representative boundaries. Columns of $R$ will not yield a
comprehensive set of boundaries if there are features that do not die. Therefore, we define a comprehensive 
set of representative boundaries by columns $V(\sigma_i)$ for all simplices $\sigma_i$ such that a feature
is born when they are added to the complex. We call these \textit{birth-cycles}. In the example in
Figure~\ref{fig:PH_matrix_reduction}, $\{V(\sigma_8), V(\sigma_9)\} = \{\{\sigma_5, \sigma_6,
\sigma_7, \sigma_8\}, \{\sigma_5, \sigma_6, \sigma_9\}\}$ is the set of birth-cycles. The matrix $V$
is a sparse upper triangular matrix. However, even then it is not feasible to store it in computer
memory since the size of a column can go up to $O(n^4),$ for a data set of $n$ points in the worst case 
for H$_2$ computation (see Figure~\ref{fig:matrix_sizes}). We developed a recursive algorithm that uses $R$ to
compute a column of $V$ on the fly without having to store any other column of $V$ in computer memory (see
Supplementary~\ref{supp:birth_cyc} for details). The computed column is directly written to a file.
This eliminates memory overhead for computing birth-cycles but can have a high run-time based on the
number of recursions and their depths. We observed in our test data sets that a small number of
columns of $V$ were used more frequently in recursions than average. The algorithm selectively
stores such columns in computer memory when they are first computed. This decreases the run-time of
any further recursions that will require these columns. Hence, the recursive algorithm is efficient in both
memory and run-time.

After computing the birth-cycles, we shorten them in multiple stages as follows.

\begin{itemize}

  \item[I] Greedy shortening: We initialize the set of representative boundaries, $\mathcal{C} =
    \{C_i\},$ as the set of all columns $V(\sigma_i)$ such that a nontrivial feature is born when $\sigma_i$ is added to the simplicial complex. The length of a cycle $C$ is the number of simplices in it---\#edges in H$_1$ representatives and \#triangles in H$_2$ representatives. We denote it by $\left | C \right |.$ The basic idea of the shortening algorithm is defined by the following steps:

    \begin{itemize}

      \item[a.] Compute $d^* = \text{max}_{C_i, C_j \in \mathcal{C}}\{\left | C_i \oplus C_j \right
        | - \left | C_i \right |\}.$ If $d^*=0,$ then exit the algorithm.

      \item[b.] Replace $C_i$ by $C_i \oplus C_j$ if $\left | C_i \oplus C_j \right | - \left | C_i
        \right | = d^*.$ If for a $C_i$ there are multiple such $C_j,$ we arbitrarily pick one to shorten $C_i.$

      \item[c.] Go to step a.

    \end{itemize}

    A straightforward implementation of computing $d^*$ is to check $\left| C_i \oplus C_j \right |$
    for all possible pairs. However, it might not be feasible in run-time if the number of cycles is
    large. For example, it took around $12$ hours to shorten a set of around $100,000$ H$_1$ representatives using
    this approach. We optimized the greedy algorithm into different cases (Supplementary~\ref{supp:shorten_cycles} for details) such that it took under 20 minutes to shorten 
    the $100,000$ H$_1$ representatives. We implement this optimization for H$_1$ representatives but not for H$_2.$
    This is because a similar optimization for H$_2$ representatives will theoretically require a data 
    structure of size $O(n^3)$ in the worst case. However, our aim is to ensure that all data structures are 
    $O(n^2).$ The straightforward implementation for H$_2$ representatives was computationally feasible in all
    of our examples because of the small number of H$_2$ features.

  \item[II] Check for connectedness: $C_i \oplus C_j$ can possibly result in
    multiple disconnected cycles. We say that two H$_1$ representatives (H$_2$ representatives) are 
    disconnected if they have no common $1$-simplex ($2$-simplex). At the conclusion of the greedy algorithm, we
    check for connectedness of every cycle (Supplementary~\ref{supp:connectedness}),
    and we record disconnected ones as separate cycles in $\mathcal{C}.$

  \item[III] Local smoothing: We smooth the resulting shortened cycles by reducing them with trivial topological
    features. Hence, we reduce H$_1$ representatives with triangles and H$_2$ representatives with tetrahedrons. 
    We develop algorithms in which we smooth a cycle by iterating over only those trivial features
    that share a boundary with it (Supplementary~\ref{supp:smoothing} for details).

\end{itemize}

We now have $\mathcal{C} = \{C_i\}$ as a set of shortened and smoothed boundaries. If a spatial
embedding of the data set is available, we add stochasticity to possibly get shorter representative
boundaries.

\begin{itemize}

  \item[IV] We presume that a spatial embedding $\mathcal{E}$ of the data set $\mathcal{U}$ is available that 
  maps all of its points bijectively to points in either $\mathbb{R}^2$ or $\mathbb{R}^3$ (Cartesian 
  coordinates), $\mathcal{E}:U \to \mathbb{R}^2 \text{ or } \mathbb{R}^3.$ WLOG we will consider embedding to be 
  in $\mathbb{R}^3.$

    \begin{itemize}

          \item[a.] Defining and computing covers of cycles: We denote a local cover of a cycle $C$
            in the embedding of the full data set as follows. Let there be $k$ points in a cycle
            $C$, denoted by $\{c_1, ..., c_k\} \subset U.$ Its embedding is $\mathcal{E}(C) =
            \{\mathbf{p}_1, ..., \mathbf{p}_k\},$ where $ \mathbf{p}_j= (p^1_j, p^2_j, p^3_j), 1
            \leq j \leq k.$ Then, the dimensions of the smallest hyper-rectangle (in Cartesian
            coordinates) that contains the cycle is $\prod_{d = 1}^3\left
            [\text{min}\{p^d_j\}_{j=1}^k, \text{max}\{p^d_j\}_{j=1}^k\right ].$ We define a cover of
            $C$ as the set of all points of the data set that are embedded inside or on this
            hyper-rectangle. In other words, $\widebar{C} = \{c \in U \mid \mathcal{E}(c) \in \prod_{d =
            1}^3\left [\text{min}\{p^d_j\}_{j=1}^k, \text{max}\{p^d_j\}_{j=1}^k \right ]\}$ is the
            cover of $C.$

          \item[b.] Eliminating cycles that cannot be around significant holes: We determine whether
            a cycle $C_i \in \mathcal{C}$ cannot wrap around a significant topological feature.
            The cover of every cycle $C_i$ is computed, denoted by $\widebar{C}_i.$ We compute PH up
            to and including $\tau$ for the embedded points $\mathcal{E}(\widebar{C}_i).$ The number
            of significant features is denoted by $n(\widebar{C}_i).$ If $n(\widebar{C}_i) = 0,$
            then the embedding of cover $\widebar{C}_i$ does not contain any significant feature.
            Hence, cycle $C_i$ cannot wrap around any significant feature in $\mathcal{E}(U).$ We
            ignore such cycles in subsequent analysis.

          \item[c.] Graphical contraction of covers: At this point, $\mathcal{\widebar{C}} =
            \{\widebar{C}_i\}$ is a collection of covers that possibly contain at least one
            significant feature in $\mathcal{E}(\widebar{C}_i).$ We define two rules to update this
            collection such that there is a possible decrease in the number of covers and/or
            decrease in the number of points in some covers. First, if $\widebar{C}_j \subset
            \widebar{C}_i$ and $n\left(\widebar{C}_i\right ) = n\left(\widebar{C}_j\right),$ then
            $\widebar{C}_i$ can be removed from $\mathcal{\widebar{C}}.$ Second, if
            $n\left(\widebar{C}_i \cap \widebar{C}_j\right) = n\left(\widebar{C}_i\right)$ and/or
            $n\left(\widebar{C}_j\right),$ then $\widebar{C}_i$ and/or $\widebar{C}_j$ is/are
            replaced by $\widebar{C}_i \cap \widebar{C}_j.$ We implement these two rules by
            representing the elements of $\mathcal{\widebar{C}}$ as nodes of a graph and conducting
            a graphical analysis (Supplementary~\ref{supp:graph_contract} for the algorithm). 

          \item[d.] Stochasticity to find multiple sets of representative boundaries: We add
            stochasticity in two different ways in each $\widebar{C}_i.$ In Supplementary~\ref{supp:pert_perm}, 
            we give a rationale for why these might make it possible to discover shorter representative cycles. 
            We detail these two methods as follows.

            \begin{itemize}

              \item[i.] Spatial perturbation of points in the embedding: For every cover
                $\widebar{C}_i,$ we construct a user-defined number of perturbations, denoted by
                $\text{n}_\text{pert}$. Every point $p_j$ in $\mathcal{E}(\widebar{C}_i)$ is
                perturbed randomly in a ball centered at the point. The radius of this ball
                is defined as the smaller of (distance of nearest neighbor of $p_j$)/3 and a maximum
                perturbation $\Delta_i$ that is allowed for all points in
                $\mathcal{E}(\widebar{C}_i).$ Such an upper threshold on the magnitude of a perturbation
                is required so that the topology does not change significantly as compared to the
                unperturbed embedding. We call $\Delta_i$ the perturbation parameter for
                $\widebar{C}_i.$ It is computed as $\Delta_i = \dfrac{\epsilon/3}{2^m},$ where $m$
                is the smallest positive integer such that the number of significant features in all
                 the $\text{n}_\text{pert}$ perturbations is the same as that in
                $\mathcal{E}(\widebar{C}_i)$ (PH computed up to $\tau = \tau_u + \epsilon$).

              \item[ii.] Permutations for every perturbation: We permute the indices of edges that
                have the same diameter. This results in permutation of columns of the coboundary and
                the boundary matrices. This stochasticity is merely a change in labels and therefore introduces nothing 
                topologically new into the computation, but helps to overcome the greedy algorithm’s local minimum problem. 
                We construct $\text{n}_\text{perm}$ number of permutations, a user-defined hyperparameter. It is possible 
                that some re-indexed sets result in the exact same indexing of edges. There is a higher chance of this if the
                number of maximum possible unique indexing is not very large as compared to $\text{n}_\text{perm}.$
                Therefore, we discard a permutation if its indexing of edges already exists in the set of permutations. 

              \item[iii.] Processing permutations of perturbations to construct the final set of
                minimal representatives: We compute representative homology boundaries for the valid permutations
                of all perturbations of all covers. We then determine minimal representative
                boundaries from the multiple computed sets. One way can be to compare lengths of the
                multiple computed boundaries around a
                topological feature. To do so, first a matching of features across different PDs is
                required. There are methods to match persistence pairs between two PDs, for example,
                minimizing bottleneck distance and minimizing Wasserstein distance, to name a few. 
                However, no metric can ensure that two features
                being matched across PDs will correspond to a significant hole located in a
                similar region in the embedding. PDs across permutations of the same perturbation will have the
                same set of persistence pairs, but even then there can be ambiguity if there exist two significant features with 
                the same birth and death. 

                Therefore, we do not match persistence pairs and define the set of
                minimal representative boundaries as follows. We first disregard all representative homology boundaries
                that do not contain significant features. Then, we pick all sets that minimize the longer of
                the remaining boundaries as follows. We sort boundaries in each set by decreasing
                order of length. The number of boundaries is equalized across all sets by inserting
                boundaries of length $0.$ Then the list of sets of representative boundaries is
                sorted in increasing order of the length of longest boundary as first priority,
                second longest length as second priority, and so on. We select the sets of representative boundaries 
                with the lowest order in the resulting sorted list. If only one set has the minimum order, then that
                set of representative boundaries is defined as the minimal set and we are done. If
                there are multiple sets with minimum order, then we define the union of all simplices in
                the representatives in these sets as the set of minimal representatives.

            \end{itemize}
    
      \end{itemize}

  \item[IV'] If a spatial embedding is not available, then we can implement multiple permutations
    and select a set of minimal boundaries as defined above.

\end{itemize}

\subsection{Hi-C analysis}\label{methods:HiC_analysis}

We first summarize the methodology of a Hi-C experiment to understand the data obtained. Nuclear DNA
from millions of cells is shredded. Pairs of spatially close shredded fragments have a high chance
to ligate due to their inherent stickiness. Paired fragments are sequenced and counted.
The counts are aggregated at a specific resolution, say $r$ kb, and reported as follows. The linear
chromosomes are partitioned into fixed length segments of $r$ base pairs, called \textit{bins}. The Hi-C
contact matrix, $M=[m_{ij}],$ is constructed where $m_{ij}$ reports the aggregated counts, or
contact frequency, of paired fragments where one fragment is in bin $i$ and the other is in bin $j.$
The contact matrix is balanced to account for experimental biases. There are multiple methods to
balance or normalize Hi-C contact matrices~\citep{lyu2020comparison}. We implemented iterative
cross-entropy (ICE) normalization using Cooler~\citep{abdennur2020cooler}. Fragments that are
spatially close in the folded genome, are expected to have higher counts or contact frequencies. Hence, we estimated the
pairwise-distance matrix by taking the multiplicative inverse of non-zero contact frequencies, i.e.\
$\widehat{D} = [\widehat{d}_{ij} = 1/m_{ij}].$ The pairwise distance for zero contact frequencies
was estimated as infinity. In other words, an edge between bin $i$ and bin $j$ is not added to the
simplicial complex if $m_{ij}=0,$ regardless of the threshold of PH computation. 
ICE normalization resulted in $m_{ij}=\text{NaN}$ for a few pairs. The corresponding edges were not added to the simplicial complex.

\subsection{Redshift data set}\label{method:universe}

The redshift data set was obtained from \texttt{http://www-wfau.roe.ac.uk/6dFGS/}. It comprises of
110,256 unique and reliable redshifts and holds 136,304 spectra. We only considered entries
classified as galaxies by the SuperCOSMOS star/galaxy classifier provided in the data set. This resulted
in a data set of 108,030 galaxies. We computed an embedding of this data set in $\mathbb{R}^3$ using
recession velocity $(cz \text{ km s$^{-1}$}),$ galactic latitude $(b),$ and galactic
longitude $(l)$ as follows. Hubble's law was used to compute the proper distance, $D =
\frac{cz}{H_0}$ Mpc where $H_0 = 72.1$ km s$^{-1}$ Mpc$^{-1}$ is the Hubble's constant. We computed
spherical coordinates $(\rho, \theta, \phi)$ using $(D, b, l)$ as follows---$\rho = D,$ $l =
\theta,$ and $\phi = (\pi/2 - b).$ Spherical coordinates were transformed to Cartesian coordinates,
$(x, y, z),$ which defines the embedding of this data set in $\mathbb{R}^3.$

To test that the voids computed have a rare set of features, we implemented two random searches, spatial and
graphical as follows.

\subsubsection{Spatial sampling}\label{method:spatial_sampling}

\begin{itemize}

  \item[1.] For every computed minimal representative boundary $C_i$, $1 \leq i \leq 40,$ we
    computed its cover $\widebar{C}_i = [x^-_i, x^+_i]\times[y^-_i, y^+_i]\times[z^-_i, z^+_i].$ Its
    dimensions are then $\Delta x_i = x^+_i - x^-_i, \Delta y_i = y^+_i - y^-_i,$ and $\Delta z_i =
    z^+_i - z^-_i.$ The minimum number of galaxies in or on the covers is denoted by $n_\text{min}.$

  \item[2.] We defined the minimum and maximum dimensions over all covers by $\Delta_\text{min} x
    = \text{argmin}_i \{\Delta x_i\}, \Delta_\text{max} x = \text{argmax}_i \{\Delta x_i\}, $ and
    similarly in $y$ and $z$ dimensions.

  \item[2.] We created a grid in 3D Cartesian coordinates with each voxel of dimensions $\Delta x =
    \Delta_{\text{min}} x, \Delta y = \Delta_{\text{min}} y, $ and $\Delta z = \Delta_{\text{min}}
    z.$

  \item[3.] We picked a point $(x,y,z)$ randomly inside or on a non-empty voxel.

  \item[4.] We randomly picked a set of dimensions, $\Delta x, \Delta y, \Delta z,$ from
    $[\Delta_\text{min} x, \Delta_\text{max} x], [\Delta_\text{min} y, \Delta_\text{max} y],
    [\Delta_\text{min} z, \Delta_\text{max} z],$ respectively. A hyper-rectangle of picked dimensions with
    $(x,y,z)$ as center is defined as a sampled cover. If the number of galaxies in or on this cover
    is at least $n_\text{min},$ then we recorded this as a sample, otherwise it was discarded.

  \item[5.] The locations of all galaxies inside
    and on 
    a valid sampled cover from previous step, were transformed to spherical coordinates.
    The center of the cover was translated to the origin. Then we
    computed four features---size of the cover as the number of galaxies in and on it, spherical
    uniformity (see~\ref{method:spherical_uniformity} for details on our definition and estimation),
    radius as the radial distance of the closest galaxy from the center, and eccentricity as the ratio
    of the smallest dimension of the cover to the longest.

\end{itemize}

\subsubsection{Graphical sampling}\label{method:graphical_sampling}

    \begin{itemize}

      \item[1.] We constructed a graph, $G = (V, E)$ on the embedding. The set of nodes $V$ was
        defined as all galaxies and an edge between two nodes was defined iff spatial distance
        between them is at most $\tau_u$ (picked as $15$ in this case study).

      \item[2.] The number of 0-simplices in the computed set of minimal representatives 
        vary in the interval $[8, 149].$ We randomly selected an integer, $n_i,$ in this range, that defines the
        number of nodes of our graphical sample. We then selected a random connected subgraph $G$
        with number of nodes equal to $n_i$ and with minimum degree $3.$ This was done for a fair comparison
        with the computed representative H$_2$ homology boundaries because every 0-simplex in them
        has at least three neighbors as they are sums of tetrahedrons. For a valid subgraph, we
        compute its cover and features of the cover that were also defined in spatial sampling.

    \end{itemize}

\subsubsection{Spherical uniformity}\label{method:spherical_uniformity}

We defined spherical uniformity of a set of points in $\mathbb{R}^3$ by estimating how evenly they
are distributed in the space of polar angle ($\theta$) and azimuthal angle ($\phi$). A grid of pixel
size $\pi/10$ rads $\times$ $\pi/10$ rads was constructed on this space. We defined spherical
uniformity, $u,$ as the ratio of non-empty pixels to the total number of pixels.

\subsubsection{Pseudo p-values}\label{method:universe_p_values}

\begin{figure}[h]

 \centering
   \begin{subfigure}{.48\textwidth}
   \centering
    \includegraphics[width=\linewidth]{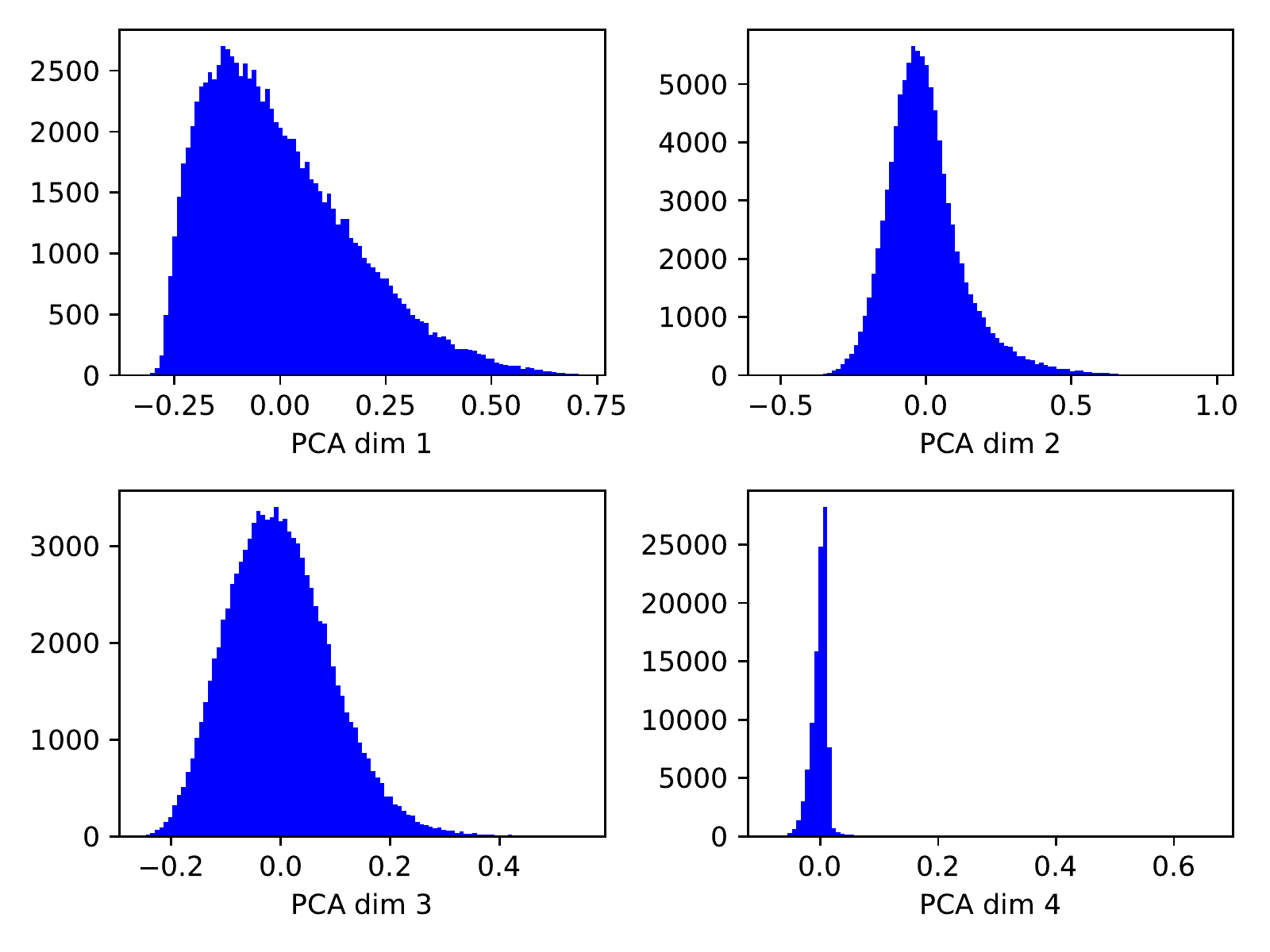}
     \caption{}
   \label{fig:universe_spatial_PCA_hist}
   \end{subfigure}
 \centering
   \begin{subfigure}{.48\textwidth}
   \centering
    \includegraphics[width=\linewidth]{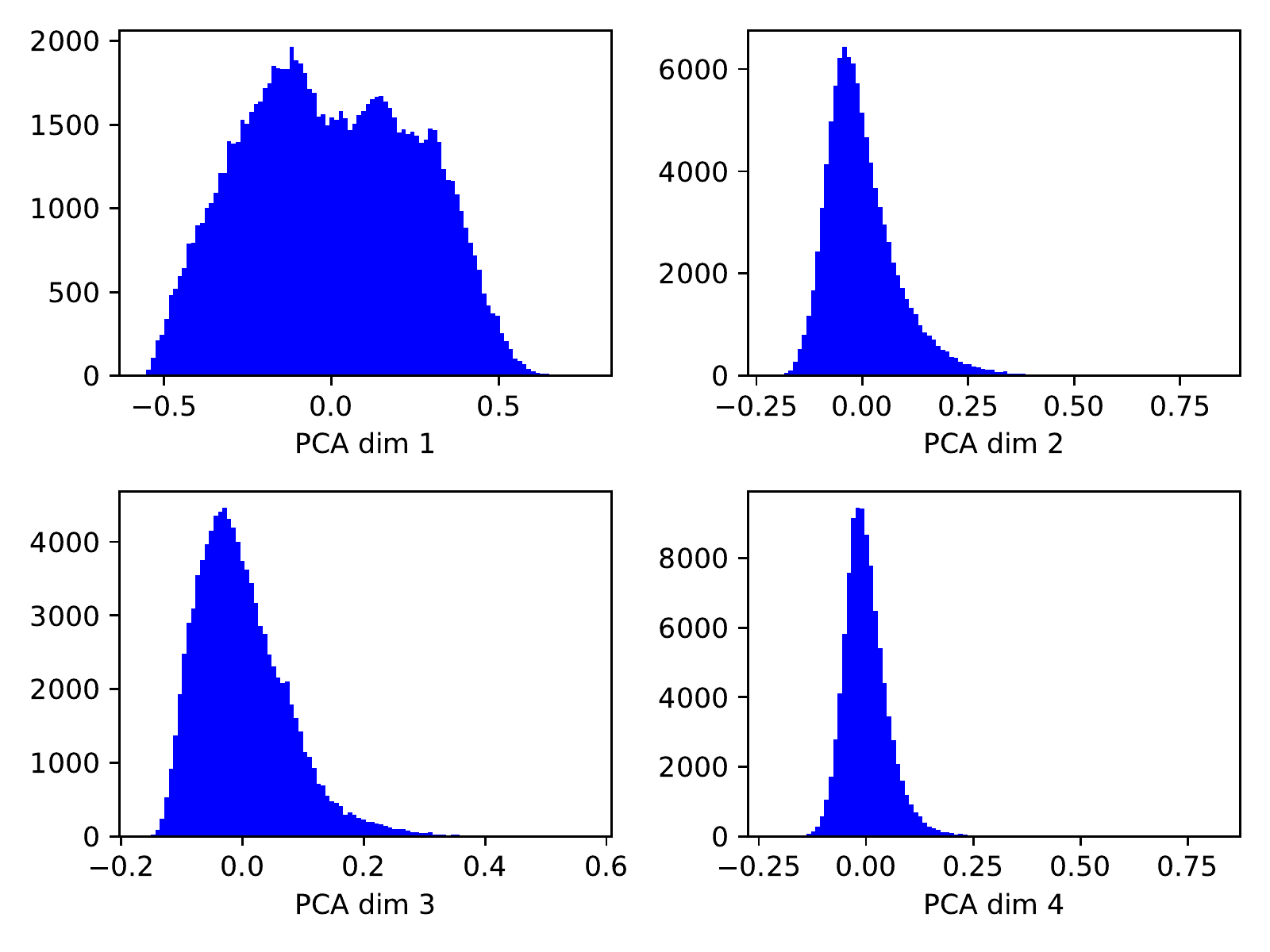}
     \caption{}
   \label{fig:universe_grahical_PCA_hist}
   \end{subfigure}

  \caption{Histograms of the PCA coordinates of random samples in (a) Spatial sampling and (b) Graphical sampling. }

  \label{fig:universe_PCA_histograms}
\end{figure}

To estimate rarity of the set of the four features of the voids computed by our strategy, we computed a pseudo p-value as follows. Let $V=\{(v^1_i, v^2_i, v^3_i, v^4_i)\}$ denote the set of features of the covers of voids computed by PH, $1 \leq i \leq 40.$ There are two sets of $100,000$ random covers, one each from spatial and graphical sampling. WLOG let $S=\{(f^1_i, f^2_i, f^3_i, f^4_i)\}$ denote a set of the four features of $N$ random samples, $1 \leq i \leq N.$ We re-scale the data to $[0,1].$ Suppose $\text{min}_j\{f^d_j\} = f^d_{-}$ and $\text{max}_j\{f^d_j\}=f^d_{+},$ then $g^d_{i} = \dfrac{f^d_i - f^d_{-}}{f^d_{+} - f^d_{-}}.$ We performed PCA on this set of points, separately for re-scaled spatial and graphical samples. Let the transformed coordinates of samples along the PCA dimensions be $S_* = \{(h^1_i, h^2_i, h^3_i, h^4_i)\}.$ Figure~\ref{fig:universe_PCA_histograms} shows the histograms of the coordinates in four PCA dimensions for both. The variance ratios of the four PCA components for spatial sampling are $[0.52, 0.31, 0.15, 0.007]$ and for graphical sampling are $[0.8,  0.093, 0.072, 0.034].$ Note that the last PCA component of spatial sampling accounts for only 0.7\% of variance in the embedded points. We re-scaled $V, u^d_i = \dfrac{v^d_i - f^d_{-}}{f^d_{+} - f^d_{-}},$ and computed coordinates, $(w^1_i, w^2_i, w^3_i, w^4_i),$ of the rescaled features sets along the PCA dimensions. Then, for a void in $V$ with re-scaled and transformed coordinates $(w^1, w^2, w^3, w^4)$ along the PCA dimensions, we computed a pseudo p-value as follows. Let $c^d$ be the number of samples with PCA coordinate value greater than $w^d$ in PCA dimension $d.$ Then, the p-value for the void was defined as $\prod_{d=1}^{4} (c^d/n),$ where $n$ is the number of total samples.

\subsection{PDB analysis to find homologs with different H$_2$ topology}\label{method:proteins}

Spatial coordinates of backbone of PDB entries were extracted using the Prody
package~\citep{zhang2021prody}. A PDB entry $q_i$ was marked as a homolog of $p_i$ if it satisfied
following criteria---at least $75\%$ of $p_i$ matched with $q_i$ with a score of at least $75\%$ and
lengths of their backbones differed by at most $25\%.$ A script was written to automate this search
using the Python package pyPDB~\citep{gilpin2016pypdb}. PDB entries that had a residue entry not among
the $20$ amino acids, were ignored. 

We defined an L$_0$ distance between two H$_2$ PD as follows. For each PD, we constructed a list of
significant barcode lengths, sorted in decreasing order. The number of entries in the two lists were
equalized by padding with $0.$ The maximum of the element-wise difference between these two lists was defined as 
the L$_0$ distance between the two PDs.

\section{Supplementary information}


\begin{table}
\begin{center}
  {
  \begin{tabular}{|c|c|} 
    \hline
    Symbol/Term & Description \\
    \hline
    $n$-simplex & A set of $n+1$ points\\
    vertex & A set of a single point, 0-simplex\\
    edge & A set of two points, 1-simplex\\
    triangle & A set of three points, 2-simplex \\
    tetrahedron & A set of four points, 3-simplex\\
    boundary ($\partial \sigma$) & All $(n-1)$-simplices that are in the $n$-simplex $\sigma$\\
    coboundary ($\delta \sigma$) & All $n$-simplices boundaries of which contain the $(n-1)$-simplex $\sigma$\\
    diameter of a simplex & Longest edge in it\\
    H$_d$ & Dimension $d$ homology group\\
    H$^*_d$ & Dimension $d$ cohomology group\\
    $D$ & Boundary matrix\\
    $V$ & Matrix that records operations to reduce boundaries\\
    $R$ & Matrix with reduced boundaries\\
    simplicial complex & A set of simplices\\
    filtration ($F$) & Sequence in which simplices are added to the complex\\
    $F_d$ & Sequence of $d$-simplices in the filtration \\
    $D^\bot$ & Coboundary matrix\\
    $V^\bot$ & Matrix that records operations to reduce coboundaries\\
    $R^\bot$ & Matrix with reduced coboundaries\\
    $p^\bot$ & List of persistence pairs from cohomology computation\\
    \hline
  \end{tabular}
}
\end{center}

  \caption{Some basic terminology, symbols, and descriptions.}

\label{tab:terminology}
\end{table}

\subsection{Computing homology}\label{suppl:compute_hom}

We first compute the persistence pairs by reducing the coboundary matrix and store them as $p^\bot.$ From duality between homology
and cohomology, it follows that we need to reduce a column $D(\sigma_i)$ only if $(\sigma_i, \sigma_j)$ is a persistence pair in 
$p^\bot.$ Additionally, we provide a way to efficiently determine trivial persistence pairs that, similar to apparent 
pairs~\citep{bauer2021ripser}, do not require any reduction. Hence, we reduce column $D(\sigma_i)$ only if 
$(\sigma_i, \sigma_j)$ is a persistence pair in $p^\bot$ and is not a trivial persistence pair.

\subsubsection{Trivial persistence pairs}\label{subsec:triv_pers}

Suppose $t$ is the smallest triangle in the coboundary of an edge $e.$ Additionally, if $e$ is the
diameter of $t,$ then it will be the lowest edge in $D(t).$  Further, since $t$ is the smallest
triangle in the coboundary of $e,$ there is no triangle smaller than $t$ that has $e$ in its
boundary. Hence, all entries in row $e$ and to the left of column $t$ will be 0. As a result, the
lowest entry in column $t$ is the first non-zero entry in row $e,$ making it a pivot entry (first
low in the row from left). Hence, column $t$ will not require any reduction operations, and we say
that $(e, t)$ is a trivial persistence pair in H$_1.$

Now, we will show that if $(e, t)$ is a trivial persistence pair in H$_1,$ then $(t, e)$ is a
persistence pair in H$^*_1$ such that coboundary of $e$ does not require any reduction operations.
Since $t$ is the lowest triangle in the coboundary of $e,$ all entries below row $t$ in column $e$
of the coboundary matrix will be 0. Also, since $e$ is the diameter of $t,$ all entries in row $t$ to
the left of column $e$ will be 0, making $(t, e)$ a pivot entry. As a result, column $e$ in the
coboundary matrix will not require any reduction operations, and we say that $(t, e)$ is a trivial
persistence pair in H$^*_1.$  In Dory, we compute cohomology and store the persistence pairs that are not trivial in
$p^\bot.$ 
Therefore, to compute H$_1,$ we
need to iterate only over the triangles that are in some persistence pair in $p^\bot$ because these
triangles are also the ones that are not in a trivial persistence pair of H$_1.$

Since we do not store trivial persistence pairs, we have to check for them at every reduction step
as well. Suppose $e$ is the lowest edge in the partially reduced boundary of a triangle. Then, if $(e,
t')$ is a trivial persistence pair, the next reduction has to be with the boundary of $t'.$ So, our
aim is to determine whether there is a trivial persistence pair $(e, t')$ for a given edge $e.$ To
do so efficiently, we use paired-indexing that we introduced in~\citet{aggarwal2021dory}.
Paired-indexing encodes a simplex such that it stores information about its diameter. A triangle $t$
is stored as $\langle k_p, k_s \rangle$ where $k_p$ is its diameter and $k_s$ is the third vertex in
the triangle that is not in the edge corresponding to its diameter. Then, for an edge $e,$ if $t' =
\langle e, k_s \rangle$ is the lowest triangle in the coboundary of $e,$ then $(e, t')$ is a trivial
persistent pair. These checks are computationally feasible because the number of reduction steps is
lowered on account of processing only the triangles that are in some persistence pair in $p^\bot.$

Similarly, to compute H$_2$ we iterate over the tetrahedrons that are in some persistence pair in
$p^\bot.$ As before, we have to check for trivial persistence pairs during reduction because we do
not store them in computer memory. Suppose $t$ is the lowest triangle in partially reduced boundary
of a tetrahedron. Then, if $(t, h')$ is a trivial persistence pair, the next reduction has to be
with the boundary of $h'.$ If $h'$ is the smallest tetrahedron in the coboundary of $t$ and the maximum
triangle in boundary of $h'$ is also $t,$ then $(t, h')$ is a trivial persistent pair, and the next
reduction has to be with the boundary of $h'.$

\subsubsection{Final algorithm for homology computation}\label{supp:final_hom_computation}

\begin{figure}[h]
  \centering
  \includegraphics[scale=0.85]{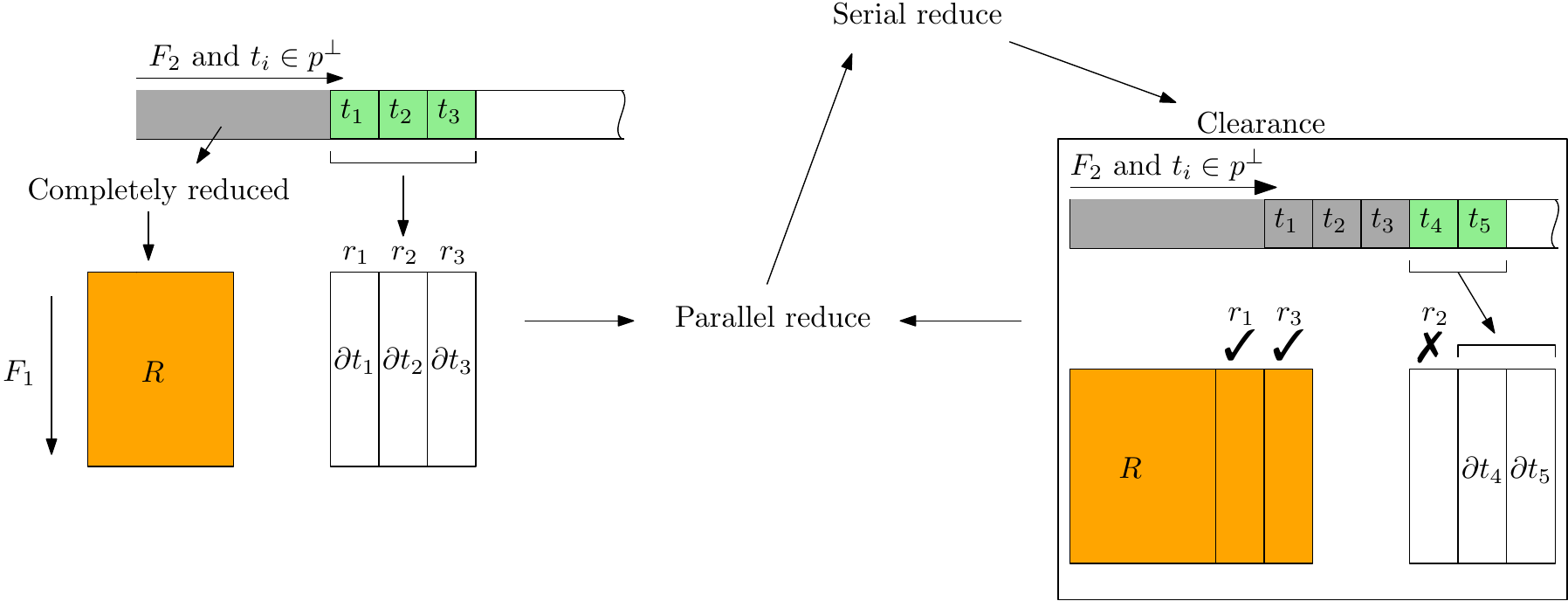}

  \caption{Serial-parallel reduction to reduce boundaries of triangles. Only those triangles that
  are in some persistence pair in $p^\bot$ need to be considered for reduction.}

  \label{fig:serial_parallel_hom}
\end{figure}

We reduce the boundaries of simplices in batches using the serial-parallel reduction that we
introduced in~\citet{aggarwal2021dory} (Figure~\ref{fig:serial_parallel_hom}). The default
batch-size for parallel reduction of boundaries is chosen to be 1000. The flowcharts for serial and
parallel are shown in appendix~\ref{app:serial_parallel}.

\subsection{Recursive algorithm to compute columns of $V$}\label{supp:birth_cyc}

We develop a recursive algorithm that computes $V(\sigma)$ without having to store $V.$ We also show
that by selectively storing a few columns of $V,$ we can significantly reduce the computation time
at a very low cost of additional memory.

\subsubsection{Computing birth-cycles for H$_1$}

 We show how we use $R$ to reduce the boundary of any edge $e_o$ in the simplicial complex on the fly.
 We consider two cases, $R(e_o)$ is non-empty (or non-zero) and $R(e_o)$ is empty (or zero). We
 first go over the former case since the latter will require the former. The function \texttt{FindV}
 in algorithm~\ref{alg:recursive_V} computes $V(e_o)$ when $R(e_o)$ is non-empty. We begin by
 defining $r_{e_o} = \partial e$ and $V(e_o) = [\,].$ We reduce $r_{e_o}$ with $R$ by updating
 $r_{e_o} \gets r_{e_o} \oplus R(e')$ if low$(R(e'))=$ low$(r_{e_o})$ and append $e_o$ to $V(e_o).$
 However, we have to append to $V(e_o)$ the reduction operations corresponding to $R(e'),$ that is,
 $V(e').$ This sets up a recursion to compute $V(e').$ It remains to define the termination of
 recursion. We show that $\partial e_o$ will always reduce to zero. Initially, $r_{e_o} = \partial
 e_o.$ Since $R(e_o)$ is the complete reduction of $\partial e_o$ that exists in $R,$ we know that
 reduction of $\partial e_o$ with $R$ will eventually make it equal to $R(e_o).$ Hence, $r_{e_o} =
 R(e_o)$ at some point in the reduction, and the next reduction step will be summing $r_{e_o}$ and
 $R(e_o)$ because they have the same low. This will result in zero.

\begin{algorithm}
  \begin{algorithmic}[1]

    \State \textbf{Input} $R, e_o$
    \State \textbf{Output} $V(e_o)$
    \State $V(e_o) \gets [\,]$

    \State \texttt{FindV}$(e_o, V(e_o))$

  \item[]
    \State \textbf{function} \texttt{FindV} $(e, V(e_o))$
  \Indent

     \State Append $e$ to $V(e_o)$ 

     \State $r_e \gets \partial e$

     \While{$r_e$ not empty}
        \If{low$(R(e')) = $ low$(r_e)$}
            \State $r_e \gets r_e \oplus R(e')$ 
        \EndIf
        \If{$r_e$ is NOT empty}
            \State \texttt{FindV}$(e', V(e_o))$
        \EndIf
     \EndWhile

  \EndIndent

  \end{algorithmic}
  \caption{Compute $V(e_o)$ for edge $e_o$ with $R(e_o)$ non-empty}
  \label{alg:recursive_V}
\end{algorithm}

We now tackle the latter case of $R(e_o)$ empty. Initially, $r_{e_o} = \partial e_o.$ In contrast to
the former case, $V(e_o)$ is initialized with $[e_o]$ and is not empty. We reduce $r_{e_o}$ with $R$
by updating $r_{e_o} \gets r_{e_o} \oplus R(e')$ if low$(R(e'))=$ low$(r_{e_o}).$ Since $R(e')$ will
be a non-empty column in $R,$ we use \texttt{FindV} (algorithm~\ref{alg:recursive_V}) to compute
reduction operations required to reduce $\partial e'$ to $R(e')$ and append them to $V(e_o).$ The
reduction ends when $r_{e_o}$ inevitably reduces to zero since $R(e_o)$ is empty. We do not keep
track of the coefficients of edges in $V(e_o)$ during reduction to reduce computation time.  Hence,
we remove edges with zero coefficient (modulo 2) from $V(e_o)$ at the conclusion of recursion since
their boundaries will sum to 0, at the end of the reduction loop.

\begin{algorithm}
  \begin{algorithmic}[1]

    \State \textbf{Input} $R, e_o$
    \State \textbf{Output} $V(e_o)$
    \State $V(e_o) = [e_o]$

    \State $r_{e_o}\gets \partial e_o$

    \While{$r_{e_o}$ is not empty}
        \If{low$(R(e')) = $ low$(r_{e_o})$}
            \State $r_{e_o} \gets r_{e_o} \oplus R(e')$ 

            \State \texttt{FindV}$(e', V(e_o))$

        \EndIf

    \EndWhile

    \State Remove edges from $V(e_o)$ with zero coefficient

  \end{algorithmic}
  \caption{Compute $V(e_o)$ for edge $e_o$ with $R(e_o)$ empty}
  \label{alg:recursive_V_empty}
\end{algorithm}

\subsubsection{Computing birth-cycles for H$_2$}

The algorithm to compute birth-cycles for H$_2$ is written similarly by recursively reducing the
boundaries of triangles at which topological features are born. However, we do not store trivial
persistence pairs of H$_1$ in $R,$ and compute these on the fly. Also, if $(e', t')$ is a trivial
persistence pair, we know that $R(t') = \partial t'.$ Hence, we do not compute $V(t')$ in such cases
(algorithm~\ref{alg:recursive_V_dim1}).

\begin{algorithm}
  \begin{algorithmic}[1]

    \State \textbf{Input} $R, t_o$
    \State \textbf{Output} $V(t_o)$
    \State $V(t_o) = [t_o]$

    \State $r_{t_o}\gets \partial t_o$

    \While{$r_{t_o}$ is not empty}
        \If{$(\text{low}(r_{t_o}), t')$ is a triv. pers. pair}

            \State $r_{t_o} \gets r_{t_o} \oplus \partial t'$ 

            \State Append $t'$ to $V(t_o)$ 

        \ElsIf{low$(R(t')) = $ low$(r_{t_o})$}

            \State $r_{t_o} \gets r_{t_o} \oplus R(t')$ 

            \State \texttt{FindV}$(t', V(t_o))$

        \EndIf

    \EndWhile

    \State Remove triangles from $V(t_o)$ with zero coefficient

  \item[]

    \State \textbf{function} \texttt{FindV} $(t, V(t_o))$
  \Indent

     \State Append $t$ to $V(t_o)$ 
     \State $r_t \gets \partial t$
     \While{$r_t$ not empty}

        \If{$(\text{low}(r_{t}), t')$ is triv. pers. pair}

            \State $r_{t} \gets r_{t} \oplus \partial t'$ 

        \ElsIf{low$(R(t')) = $ low$(r_t)$}
            \State $r_t \gets r_t \oplus R(t')$ 
        \EndIf

        \If{$r_t$ is NOT empty}
            \State \texttt{FindV}$(t', V(t_o))$
        \EndIf
     \EndWhile

  \EndIndent

  \end{algorithmic}
  \caption{Compute $V(t_o)$ for triangle $t_o$ with $R(t_o)$ empty}
  \label{alg:recursive_V_dim1}
\end{algorithm}

\subsubsection{Selective storage for computational optimization}

$V$ does not need to be stored in memory in the recursive algorithm. We write each
birth-cycle to a file when it is computed. The only additional memory taken is the maximum possible
size of $V(\sigma).$ However, recursive reduction can be expensive depending upon the number of
reductions and the recursion-depth of every reduction. To alleviate this problem, we note that it is
possible for a column $R(\sigma_i)$ to be used multiple times in the recursive computation of
birth-cycles. In this case, storing its reduction operations, $V(\sigma_i),$ in computer memory can
decrease the computational overhead. It is also possible that the time taken to compute
$V(\sigma_i)$ is low and the multiple computations have a negligible computational overhead. Hence,
we developed selective storage of columns of $V$ based on two criteria. For notational
convenience, we drop the subscript $i.$ First, we consider the number of reductions, inclusive of the
reductions carried out recursively, required to compute $V(\sigma).$ We denote this parameter by
$n_r(\sigma)$ (\#reductions). Second, we maintain a counter for the number of times $V(\sigma)$ is
requested as the birth-cycles are being computed. We denote this parameter by $n_u(\sigma)$
(\#usage). If $n_r(\sigma) > n_r^*$ and when $n_u(\sigma)$ surpasses a user-defined threshold,
denoted by $n_u^*,$ then we store $V(\sigma)$ in the memory. Note that an optimal value of $n_u^*$
that can provide the best trade-off between memory usage and run-time can be determined only after
computing all of the birth-cycles recursively and analyzing usage of $V(\sigma).$ We chose $n_u^* =
2$ as its default value, and it can be increased if storage of fewer $V(\sigma)$ is desired and the
resulting increase in computation time is insignificant.

The benefits of the above strategy will be most significant if $n_r$ is low for most simplices and
the ones with large $n_r$ also have a large $n_u.$ The former is desirable because it will reduce
the memory taken by selective storage, and the latter means the reduction in computation time by
using selective storage will be significant.

\subsection{Greedy algorithm to compute a set of shorter cycles}\label{supp:shorten_cycles}

A birth-cycle corresponding to a topological feature that is born in homology group H$_d$ is a set
of $d$-simplices. We will denote the set of all birth-cycles computed for H$_d$ by $\mathcal{B}_d.$
For notational convenience, we drop the subscript $d$ since the analysis follows similarly in any
dimension. We denote a birth-cycle by $X_i \in \mathcal{B}$ where $X_i$ is a set of simplices. We
define the length of $X_i$ as the number of simplices in $X_i,$ denoted by $\lvert X_i \rvert.$
Since the set of birth-cycles forms a basis for the corresponding homology group, we propose a
Gram-Schmidt like method to reduce the lengths of cycles.  We substitute a cycle $X_i$ by $X_i
\oplus X_j = (X_i \cup X_j) \setminus (X_i \cap X_j)$ if such a substitution reduces its length by
$r_m,$ the maximum possible reduction in length when all pairs of cycles are considered. The
iterations are continued till the sum of no two cycles results in a cycle of shorter length. We know
that this condition will be satisfied eventually because every iteration results in a reduction in
length of at least one cycle and the lengths are bounded below by 0. Note that $X_j$ is possibly one
of multiple cycles that can be added to $X_i$ to reduce its length by $r_m.$ A choice of a different
cycle might result in a different set of cycles at the end of the algorithm that may be shorter.
Hence, it is not guaranteed that a single run of this algorithm to termination will result in cycles
of minimal lengths, and the only condition satisfied is that sum of no two cycles will give a cycle
of shorter length. We also note that $X_i \oplus X_j$ might result in disconnected cycles.  

A brute-force implementation that scans all possible pairs is computationally infeasible. $N$ cycles will require $O(N^2)$ comparisons in every
iteration and every comparison involves determining $(X_i \cup X_j) \setminus (X_i \cap X_j),$ which
can be costly for long cycles. Hence, such a brute-force approach is not feasible to process a large
number of cycles. 

We make the following observations.

\begin{enumerate}

  \item \label{O1} If $\lvert X_i \rvert - \lvert X_i \oplus X_j \vert > r_i,$ then both $\lvert
    X_i \rvert$ and $\lvert X_j \rvert$ have to be greater than $r_i.$

  \item \label{O2} If $\lvert X_i \rvert \geq \lvert X_j \rvert,$ then $\lvert X_i \rvert -
    \lvert X_i \oplus X_j\rvert \geq \lvert X_j \rvert - \lvert X_i \oplus X_j\rvert .$

\end{enumerate}

To utilize these observations, we index the birth cycles in decreasing order of their lengths.
Hence, if $i < j,$ then $\lvert X_i \vert > \lvert X_j \rvert.$ Indices of cycles of the same length are
defined arbitrarily by the sorting algorithm. Since the relative lengths of cycles might change in
an iteration, a re-indexing of cycles by sorting them in decreasing order of lengths is done at
every iteration. Then, from observation~\ref{O2} it follows that for a given a cycle $X_i,$ we only
need to consider substitution of $X_i$ by $X_i \oplus X_j$ if $j > i.$ Hence, we decrease the number
of comparisons required in every iteration. 

We begin by initializing two data structures. First, for every cycle $X_i$ we store $f(X_i) =
(X^i_*, r_i),$ where $X^i_*$ is the cycle such that updating $X_i$ by $X_i \oplus X^i_*$ decreases
its length the most, and the decrease in length is stored as $r_i.$ Second, for every simplex
$\sigma,$ that is in at least one cycle, we maintain a list of all cycles that contain it. We denote
it by $g(\sigma) = [X_{i_1}, X_{i_2}, ...].$

To initialize $f(X_i),$ we first set $r_i \gets 0, X_i = \text{NULL},$ and consider the sum of $X_i$
with all cycles $X_j$ with $j > i.$ The following ideas reduce the number of comparisons necessary.

\begin{enumerate}

  \item \label{init1} If $\lvert X_i \rvert - \lvert X_i \oplus X_j \vert > r_i,$ we update $X^i_*
    \gets X_j$ and $r_i \gets \lvert X_i \rvert - \lvert X_i \oplus X_j \vert.$ We increment $j$ by
    1 and repeat till either $j = N$ or $X_j < r_i.$ The latter condition follows from
    observation~\ref{O1} because, if $X_j < r_i,$ then summing $X_i$ with any cycle $X_k$ for $k
    \geq j$ cannot result in a reduction in length greater than $r_i.$ Since the cycles are indexed
    by decreasing lengths, we can quit when $X_j < r_i.$ This potentially reduces the number of
    comparisons to be done. Note that if $r_i = 0$ upon termination, then there is no cycle that can
    be added to $X_i$ to reduce its length.

  \item \label{init2} To compute $\lvert X_i \oplus X_j \rvert,$ we iterate over elements in $X_i$
    and $X_j$ and count the elements that are in either but not in both. To make this operation of
    order $O(\lvert X_i \rvert + \lvert X_j \rvert),$ we store every cycle as a list of simplices
    ordered by their index. We quit when the count is more than $\lvert X_i \rvert - r_i$ because
    that implies that the reduction in length will be less than $r_i.$ In such a case, we do not
    iterate over all elements in $X_i$ and $X_j,$ reducing the computation time.

\end{enumerate}

We compute the initial $f(X_i)$ for all birth-cycles in parallel using \texttt{OpenMP} for-loop with
\texttt{static} scheduling and a \texttt{chunksize} of 1000.

We start the iterative algorithm after $f(X_i)$ and $g(\sigma)$ have been initialized. We denote the
maximum $r_i$ in $f(X_i)$ by $r_m.$ If $r_m = 0,$ the algorithm terminates. Otherwise, all $X_i$
with $r_i = r_m$ are substituted with $X_i \oplus X^i_*.$ For every substitution, $g(\sigma)$ has to
be updated as follows---$X_i$ is to be added to $g(\sigma)$ for all $\sigma \in X_*^i \setminus (X_i
\cap X_*^i)$ and it has to be removed from $g(\sigma)$ for all $\sigma \in X_i \cap X_*^i.$ This
information is tracked during the summation of cycles as a list $u(\sigma) = [(X_{i_1}, f_1), ...],$
where $f_1 = 0$ if $X_{i_1}$ has to be removed from $g(\sigma)$ and, otherwise, $f_1 = 1.$ Then,
$g(\sigma)$ is updated in parallel.

After these substitutions, we update the indices of cycles by re-sorting them in
decreasing order of lengths. 
We also have to
determine whether $f(X_i)$ has changed. To reduce the number of comparisons, we consider four
different cases for every cycle $X_i$ as follows.

\begin{enumerate}

  \item \label{case1} $X_i$ was updated: If $X_i$ was updated in this iteration, then we reset $r_i
    \gets 0$ and check the sum of $X_i$ with 
    each cycle $X_j$ where $j > i.$

  \item \label{case2} $X_i$ was not updated, $r_i = 0$: This means that prior to the updates, there
    was no cycle that could be added to $X_i$ to result in a cycle of shorter length. Hence, after
    the updates, we need to check the sum of $X_i$ with only the cycles that were updated.

  \item \label{case3} $X_i$ was not updated, $r_i \neq 0,$ $X_*^i$ was updated: Since $X_*^i$ has
    been updated, we check the sum of $X_i$ with 
    each cycle $X_j$ where $j > i.$

  \item \label{case4} $X_i$ was not updated, $r_i \neq 0,$ $X_*^i$ was not updated: In this case we
    only need to check the sum of $X_i$ with the updated cycles to determine if a reduction in length
    more than $r_i$ is possible.

\end{enumerate}

The above four cases can be categorized into two scenarios---checking the sum of $X_i$ with all cycles
and checking the sum of $X_i$ with only the cycles that were updated in the iteration. These scenarios
are implemented using different strategies.

\textit{Strategy 1 to check the sum of $X_i$ with all cycles (cases~\ref{case1} and~\ref{case3})}: For
every simplex $\sigma$ in $X_i,$ we iterate over $g(\sigma).$ Then, if $X_j \in g(\sigma),$ we know
that $\sigma \in X_i \cap X_j.$  Hence, we can compute $\lvert X_i \cap X_j \rvert$ on the fly by
initializing it as 0 and increment it by 1 whenever $X_j$ is in $g(\sigma)$ for $\sigma \in X_i.$
Since $\lvert X_i \oplus X_j \rvert = \lvert X_i \rvert + \lvert X_j \rvert - 2\lvert X_i \cap X_j
\rvert,$ we can compute the reduction in length on the fly when $X_i$ is summed with $X_j$ as well.
If this reduction is greater than $r_i,$ we update $r_i \gets \lvert X_i \oplus X_j \rvert$ and
$X_*^i \gets X_j.$

\textit{Strategy 2 to check the sum of $X_i$ with updated cycles (cases~\ref{case2} and~\ref{case4})}:
Suppose the $U = [X_{i_1}, X_{i_2}, ...]$ is the list of updated cycles ordered in decreasing order
of their lengths. Then, we sum $X_i$ with cycles in $U$ as they are ordered in $U.$ If we get a
reduction in length of $X_i$ that is greater than $r_i,$ we update $r_i$ and $X_*^i$ accordingly.
For efficiency, we implement the two ideas,~\ref{init1} and~\ref{init2}, from initialization.
Hence, we may need not compare $X_i$ with all cycles in $U$ and also may not need to iterate over
all simplices in the two cycles that are being summed.

Strategy 1 and strategy 2 are implemented as embarrassingly parallel over all cycles and updated
cycles, respectively. The four cases (case~\ref{case1} to~\ref{case4}) are implemented separately in
parallel for better load balance. We use \texttt{OpenMP} for-loop with \texttt{static} scheduling
and a \texttt{chunksize} of 50.

Why do we use two different strategies? The strength of strategy 1 lies in the fact that iterating
over $g(\sigma)$ compares only those cycles that have at least one simplex in common. However, it
also means that we iterate over every simplex in every cycle $X_i$ that is to be processed. In our
experiments with test data sets, we observed that in the first few iterations almost all of the
cycles are in cases~\ref{case2} and~\ref{case4} and only a few cycles updated. Hence, strategy 1 takes long because it iterates over the entire length of every cycle. As an
alternative, we implement strategy 2 that iterates over pairs of cycles and checks their sums. It is
significantly faster than strategy 1 because the ideas~\ref{init1} and ~\ref{init2} reduce the
number of pairs that are considered. Now, as iterations of the algorithm reduce the lengths of the
cycles, there is a decrease in both the number of simplices in the cycles and the number of cycles
that have common simplices. As a result, combined with strategy 1, the algorithm scales efficiently
for a large number of cycles. The efficiency is higher if the iterations of the algorithm are
accompanied with an increase in the number of cycles in cases~\ref{case1} and~\ref{case3}.

This algorithm can be implemented for both H$_1$ and H$_2$ birth-cycles. However, since the memory
requirement for $g(\sigma)$ can be $O(n^3)$ in the worst case, we implement it for only H$_1.$ To
shorten H$_2$ birth-cycles, we simply iterate over all pairs of nontrivial cycles in every
iteration. This was computationally feasible for all of the data sets in this work because the
number of nontrivial H$_2$ features was very low.

\subsection{Connectedness}\label{supp:connectedness}

For every shortened H$_1$ representative, we compute its cycle basis using the networkx Python package and record 
different basis elements as separate cycles. For every shortened H$_2$ representative, we compute a connectivity graph, $G=(V,E),$ with the set of nodes $V$ as all triangles in the representative and an edge between nodes is defined if the corresponding triangles have two points in common. We record disconnected components of $G$ as separate H$_2$ representatives.

\subsection{Smoothing}\label{supp:smoothing}

An H$_1$ representative can be written as a sequence of 0-simplices, $[v_0, ..., v_n, v_0].$ To
smooth it, we simply remove $v_i$ if the pairwise distance between $v_{i-i}$ and $v_{i+1}$ is at
most $\tau_u.$ 

An H$_2$ representative boundary is a set of triangular faces, $B = \{t_i\},$ and each triangle,
$t_i,$ is a set of its three vertices. We define a graph $G$ with set of nodes $V = \{t_i\}$ and set
of edges $E = \{(t_i, t_j) \text{ where } \left |t_i \cap t_j \right | = 2\},$ i.e.\ an edge between
two nodes of $G$ denotes that the corresponding triangular faces share two vertices. For each
triangle $t_i,$ we define its diameter as the length of the longest edge in it, denoted by $d_i.$

Suppose a tetrahedron $h = \{t_1, t_2, t_3, t_4\}$ exists in the embedding, with birth $\leq
\tau_u,$ such that exactly three of its four faces are in $B.$ Then, we remove those three faces
from $B,$ add the fourth face of $h$ to $B,$ and add edges between the fourth face and neighbors of the
three removed faces. If there are multiple such possibilities, then the algorithm greedily picks a
tetrahedron with smallest diameter. We update $G$ and repeat till no update to $G$ is possible.
At the end of the iterations, we remove any disconnected tetrahedrons in $B.$ It is not efficient to
loop through all valid (with birth $\leq \tau_u$) tetrahedrons in the embedding of the entire data
set. We developed an algorithm that, instead, is of the order of number of triangular faces in $B$
(Algorithm~\ref{alg:smoothing_2_cycle}).

\begin{algorithm}[!th]
  \begin{algorithmic}[1]

    \State \textbf{Input:} Boundary $B=\{t_i\}$ (set of triangular faces), birth threshold $\tau_u$ 
    \State Make graph $G\gets(B, E),$ where $(t_i, t_j) \in E$ if they share two points. 

    \State update $\gets 1$

    \While{update}
      
        \State update $\gets 0$

        \State $d_* \gets \infty$

        \State $\mathcal{C}\gets$ cliques in $G$ with at least 3 nodes/triangular faces

        \For{$C$ in $\mathcal{C}$}

            \For{$(t_1, t_2, t_3)$ in combinations of 3 from $C$}

                \State all points $P = t_1 \cup t_2 \cup t_3$

                \State $t_4 \gets P\setminus \{t_1\}$

                \State $t_4 \gets t_4 \cup (P\setminus \{t_2\})$

                \State $t_4 \gets t_4 \cup (P\setminus \{t_3\})$

                \If{$t_4 \in C$}

                  \State Skip because tetrahedron $\{t_1, t_2, t_3, t_4\} \in B$

                \Else
                  
                  \State $d \gets$ diameter of $h=\{t_1, t_2, t_3, t_4\}$

                  \If{$d < d_*$ AND $d \leq \tau_u$}
                  
                    \State $d_* \gets d$

                    \State Mark $\{t_1^*, t_2^*, t_3^*\} \gets \{t_1, t_2, t_3\}$ to be removed and
                    $t_4^* \gets t_4$ to be added to $G$

                  \EndIf

                \EndIf

            \EndFor

        \EndFor

        \If{$d_*$ is not $\infty$}

          \State Add $t_4^*$ to $V$

          \State $N \gets$ union of neighbors of $\{t_1^*, t_2^*, t_3^*\}$

          \For{$t \in N$}
            
              \If{$t_4^*$ and $t$ have two points in common}

                \State Add edge $(t_4^*, t)$ to $E$

              \EndIf

          \EndFor

          \State Remove $\{t_1^*, t_2^*, t_3^*\}$ from $B$

          \State update $\gets 1$

        \EndIf

    \EndWhile

    \State Remove disconnected tetrahedrons from $B$ by removing its components of length $\leq 4$

    \State \textbf{Output:} Nodes in $B$  

  \end{algorithmic}
  \caption{Smoothing a 2-cycle}
  \label{alg:smoothing_2_cycle}
\end{algorithm}

\subsection{Covers and graphical contraction}\label{supp:graph_contract}

Given an upper threshold for PH $\tau_u,$ threshold for significance $\epsilon,$ and
$\mathcal{\widebar{C}} = \{\widebar{C}_i\}$ with $n(\widebar{C}_i)>0,$ we implement the following
conditions.

\begin{itemize}

  \item[1.] Subset check: If $\widebar{C}_j \subset \widebar{C}_i$ and $n\left(\widebar{C}_i\right )
    = n\left(\widebar{C}_j\right),$ then remove $\widebar{C}_i$ from $\mathcal{\widebar{C}}.$

  \item[2.] Intersection check: If $\widebar{C}_i \cup \widebar{C}_j \notin \{\phi, \widebar{C}_i,
    \widebar{C}_j\}$ and $n\left(\widebar{C}_i \cap \widebar{C}_j\right) =
    n\left(\widebar{C}_i\right)$ and/or $n\left(\widebar{C}_j\right),$ then replace $\widebar{C}_i$
    and/or $\widebar{C}_j$ by $\widebar{C}_i \cap \widebar{C}_j.$ Note that we can impose a stronger
    condition that $\left | \widebar{C}_i \cup \widebar{C}_j \right | > 3,$ because otherwise the
    intersection cannot have a significant feature.

\end{itemize}

We implement this by constructing a graph with nodes as the elements of $\mathcal{\widebar{C}}$
(Algorithm~\ref{alg:graphical_contraction}).

\begin{algorithm}[!th]
  \begin{algorithmic}[1]

    \State \textbf{Input:} $\tau_u, \epsilon, \mathcal{\widebar{C}} = \{\widebar{C}_i\}$ where
    $n(\widebar{C}_i) > 0$

    \State Set of vertices $V \gets \mathcal{\widebar{C}}$

    \State Remove $C_i$ from $V$ if there exists $n(\widebar{C}_i) = n(\widebar{C}_j)$ AND
    $\widebar{C}_i \subset \widebar{C}_j$ \Comment{Subset check}

    \State Define edge set $E \gets \{(\widebar{C}_i, \widebar{C}_j)\}$ if $\widebar{C}_i \cap
    \widebar{C}_j \notin \{\phi, \widebar{C}_i, \widebar{C}_j\}$

    \State Intersection graph $G \gets (V, E)$

    \For{component $c$ in components of $G$}

      \State $G_c \gets$ subgraph $G$ defined by nodes in $c$

      \State update $\gets 1$

      \While{update}
          
          \State update $\gets 0$

          \For{$(c_i, c_j)$ in pairs of nodes of $G_c$} \Comment{Intersection check}

            \State $c_k \gets c_i \cap c_j $
            \State \textbf{Continue/Skip} if $c_k \in \{c_i, c_j\}$ AND $\left | c_k \right | < 4$

            \State Compute $n(c_k)$ by computing PD up to $\tau_u + \epsilon$

            \If{$n(c_k) = n(c_i)$}

              \State Remove $c_i$ from $G_c$

            \EndIf

            \If{$n(c_k) = n(c_j)$}

              \State Remove $c_j$ from $G_c$

            \EndIf

            \If{$c_i$ and/or $c_j$ removed}\Comment{Add $c_k$ to $G_c$}

                \For{$c_i$ in $G_c$}

                    \State If $c_k \subset c_i$ AND $n(c_k) = n(c_i),$ remove $c_i$ \Comment{Subset
                    check}

                \EndFor

                \State Update edge set $E$ to add neighbors of $c_k$

            \EndIf

            \State update $\gets 1$

            \State \textbf{break}

          \EndFor

      \EndWhile

    \EndFor

    \State $G \gets$ composition of all $G_c$

    \State \textbf{Output:} List of nodes of $G$
      
  \end{algorithmic} \caption{Graphical contraction} \label{alg:graphical_contraction}
\end{algorithm}

\subsection{Perturbations and permutations}\label{supp:pert_perm}

The rationale for perturbations and permutations possibly yielding shorter representatives is as follows.

\begin{itemize}

\item[a.] Perturbations: Consider a persistence pair $(b, d).$ The algorithm will find a representative boundary
  around this feature, say $C^1,$ that is born at $b,$ i.e.\ the length of longest edge in $C^1$ is
  $b.$ Suppose that there exists a cycle $C^2$ in the embedding that is born at $b+\Delta$ $(\Delta >
  0)$ and is closer to the hole as compared to $C^1.$ We are interested in discovering $C^2$ if
  $\Delta$ is sufficiently small $(b + \Delta < \tau_u).$ By perturbing the data set, we allow the
  possibility that the longest edge in $C^2$ is smaller than the longest edge in $C^1$ in the
  perturbed data set, and hence, the former can be discovered by the algorithm as a representative
  boundary.

\item[b.] Permutations: The indices of the edges that are added to the simplicial complex at the same spatial scale, are permuted.
  The resulting change in the order of columns of boundary matrices, can give a different set of
  representative boundaries. Note that, unlike perturbations which are essentially relying on the robustness of meaningful topological features
  , a permutation
  preserves the PD.

\end{itemize}

\section{Universe}\label{supp:universe}

\begin{figure}[!tbhp]
      \centering
      \begin{subfigure}{.18\textwidth} \centering
        \includegraphics[width=0.9\linewidth]{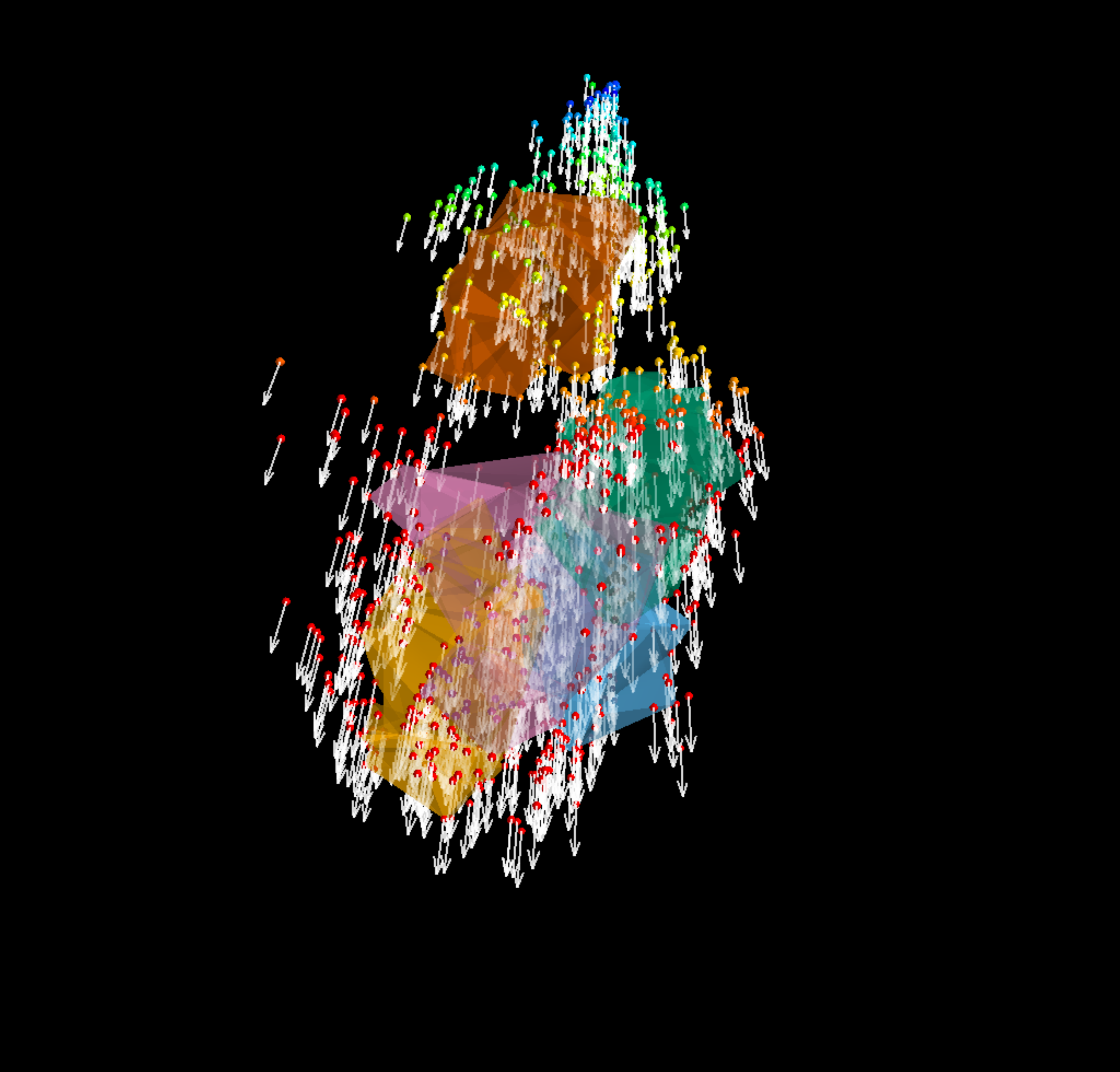}
        \captionsetup{labelformat=empty}
        \caption{}
        \label{supp_fig:universe_1}
    \end{subfigure}
      \centering
      \begin{subfigure}{.18\textwidth} \centering
        \includegraphics[width=0.9\linewidth]{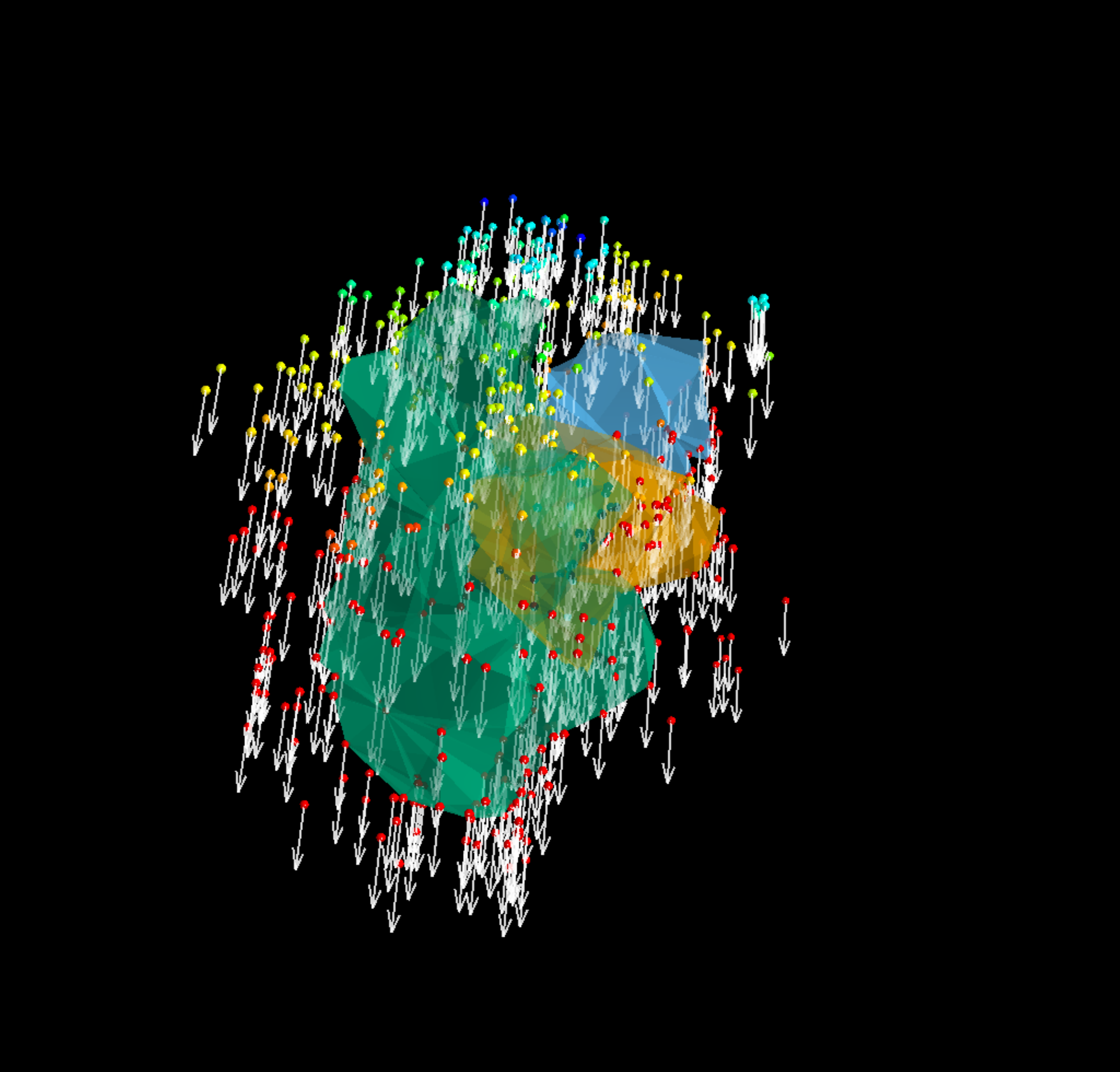}
        \captionsetup{labelformat=empty}
        \caption{}
        \label{supp_fig:universe_1}
    \end{subfigure}
      \centering
      \begin{subfigure}{.18\textwidth} \centering
        \includegraphics[width=0.9\linewidth]{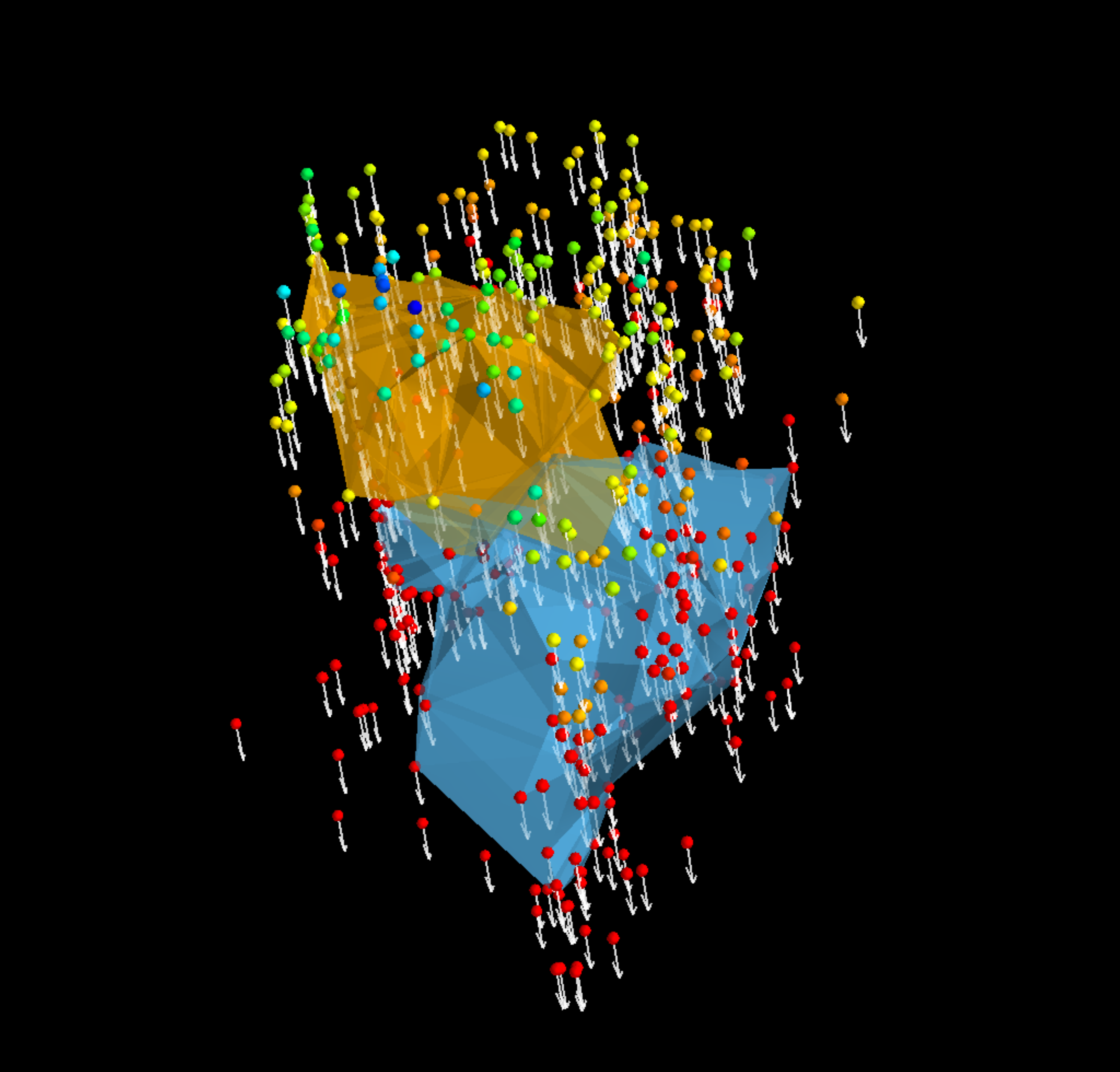}
        \captionsetup{labelformat=empty}
        \caption{}
        \label{supp_fig:universe_1}
    \end{subfigure}
      \centering
      \begin{subfigure}{.18\textwidth} \centering
        \includegraphics[width=0.9\linewidth]{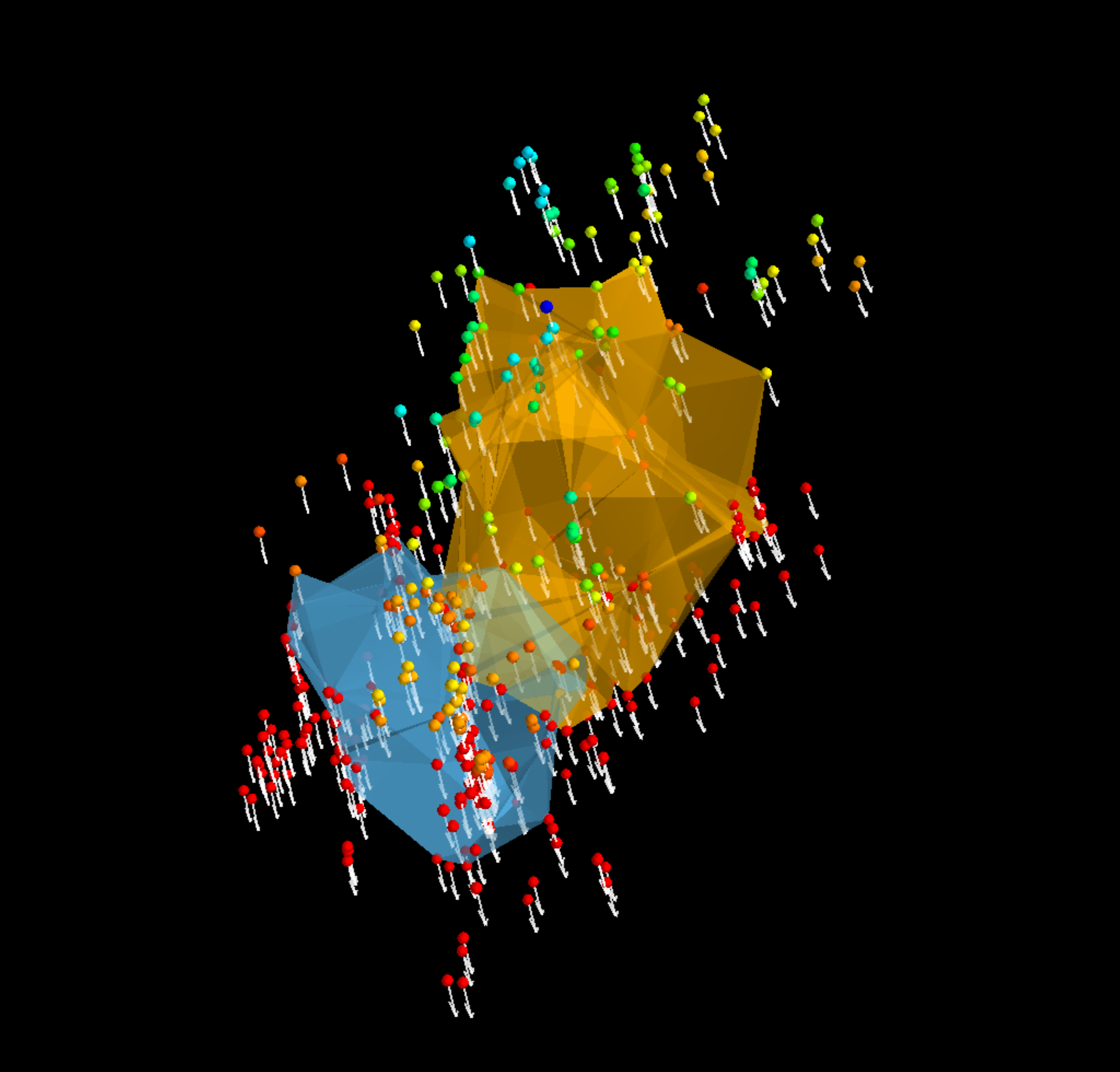}
        \captionsetup{labelformat=empty}
        \caption{}
        \label{supp_fig:universe_1}
    \end{subfigure}
      \centering
      \begin{subfigure}{.18\textwidth} \centering
        \includegraphics[width=0.9\linewidth]{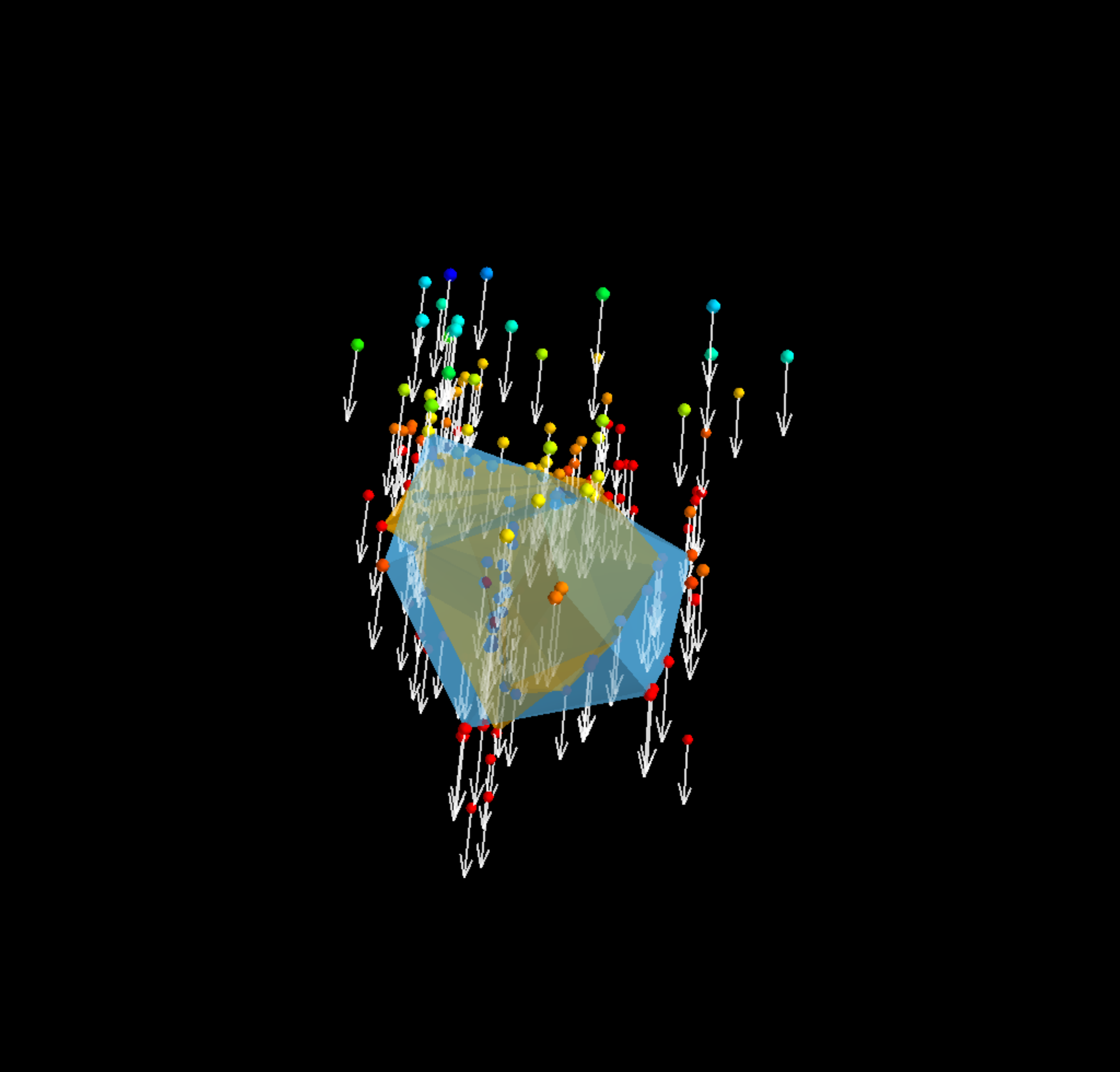}
        \captionsetup{labelformat=empty}
        \caption{}
        \label{supp_fig:universe_1}
    \end{subfigure}
      \centering
      \begin{subfigure}{.18\textwidth} \centering
        \includegraphics[width=0.9\linewidth]{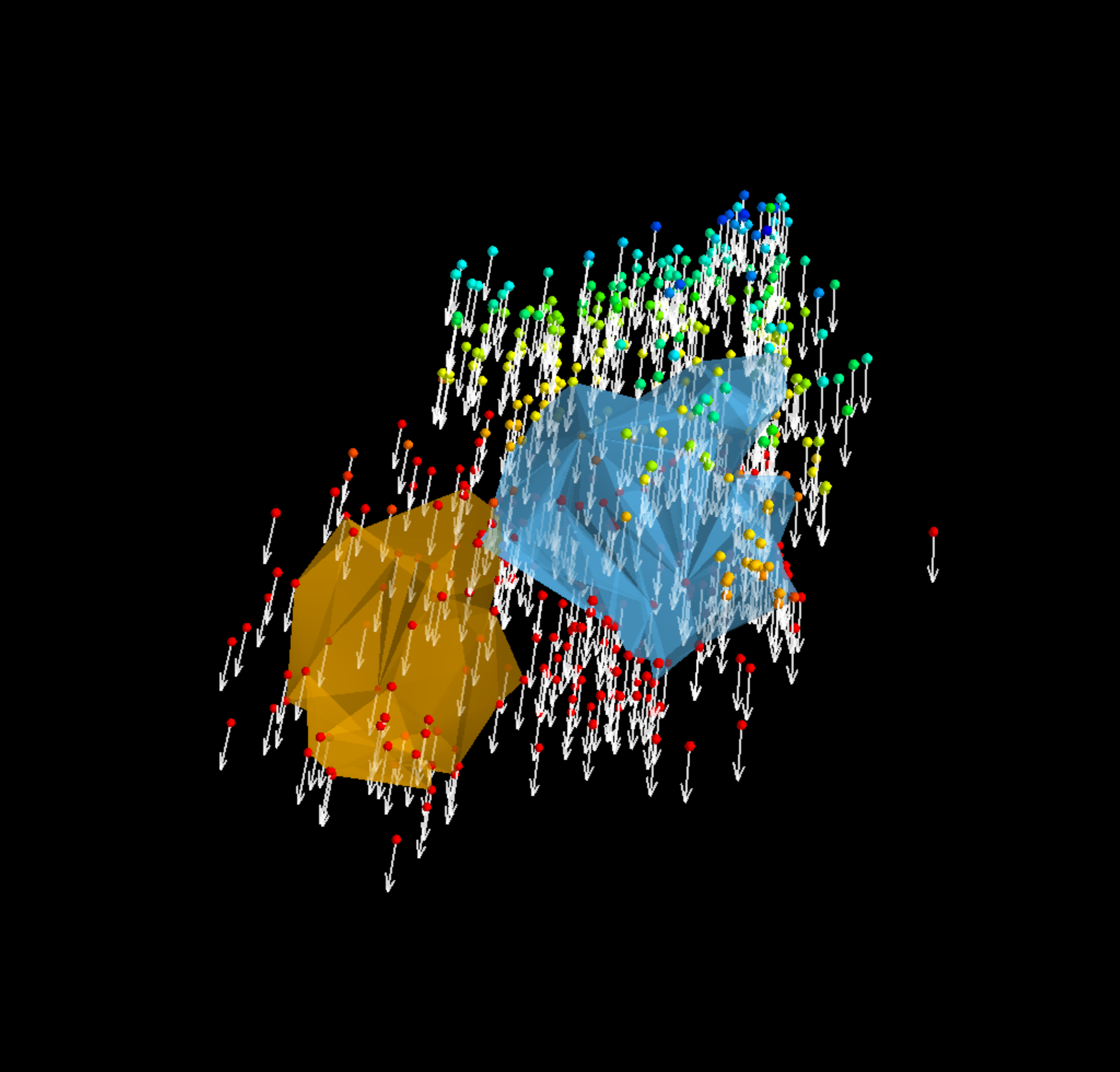}
        \captionsetup{labelformat=empty}
        \caption{}
        \label{supp_fig:universe_1}
    \end{subfigure}
      \centering
      \begin{subfigure}{.18\textwidth} \centering
        \includegraphics[width=0.9\linewidth]{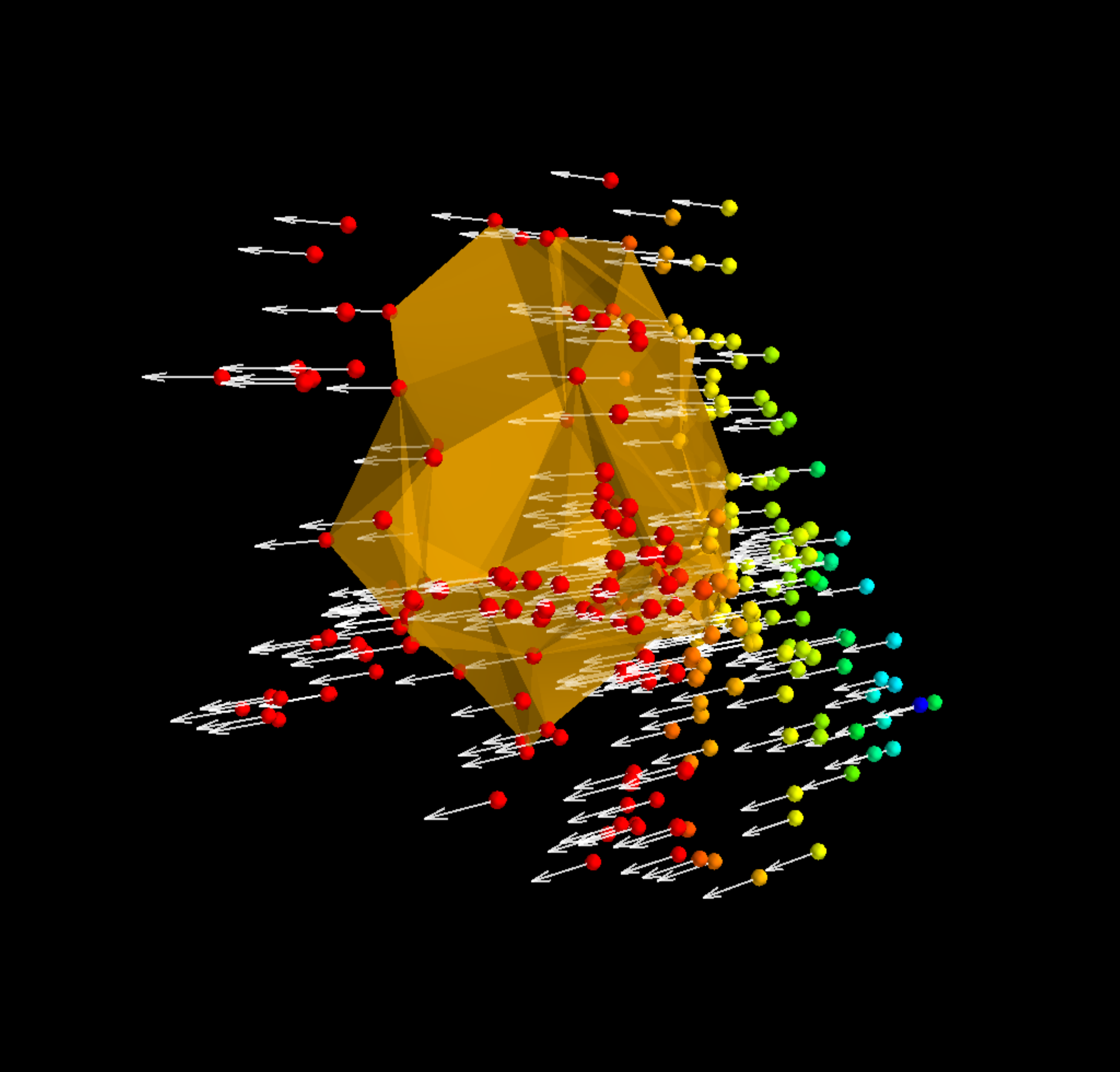}
        \captionsetup{labelformat=empty}
        \caption{}
        \label{supp_fig:universe_1}
    \end{subfigure}
      \centering
      \begin{subfigure}{.18\textwidth} \centering
        \includegraphics[width=0.9\linewidth]{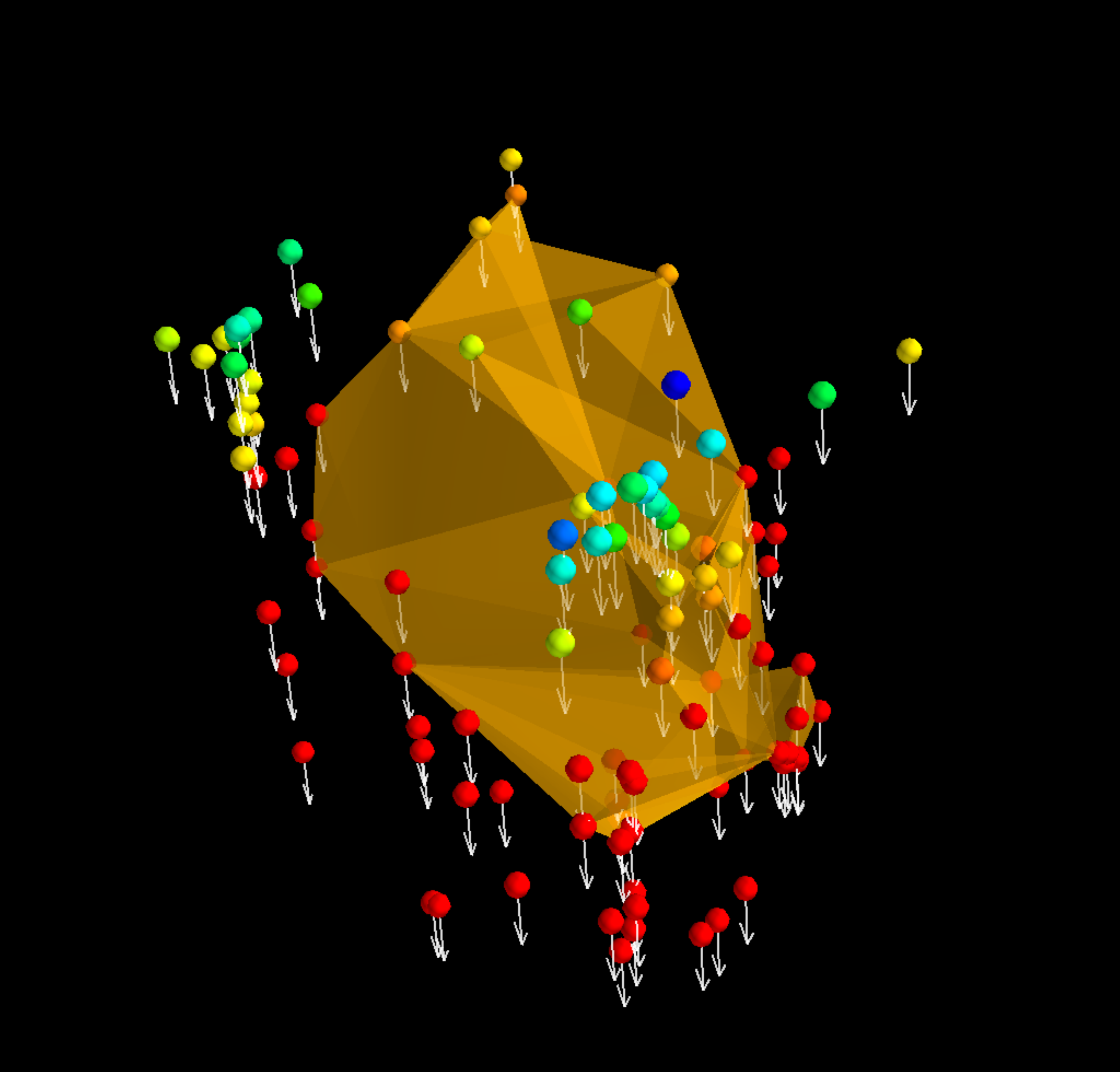}
        \captionsetup{labelformat=empty}
        \caption{}
        \label{supp_fig:universe_1}
    \end{subfigure}
      \centering
      \begin{subfigure}{.18\textwidth} \centering
        \includegraphics[width=0.9\linewidth]{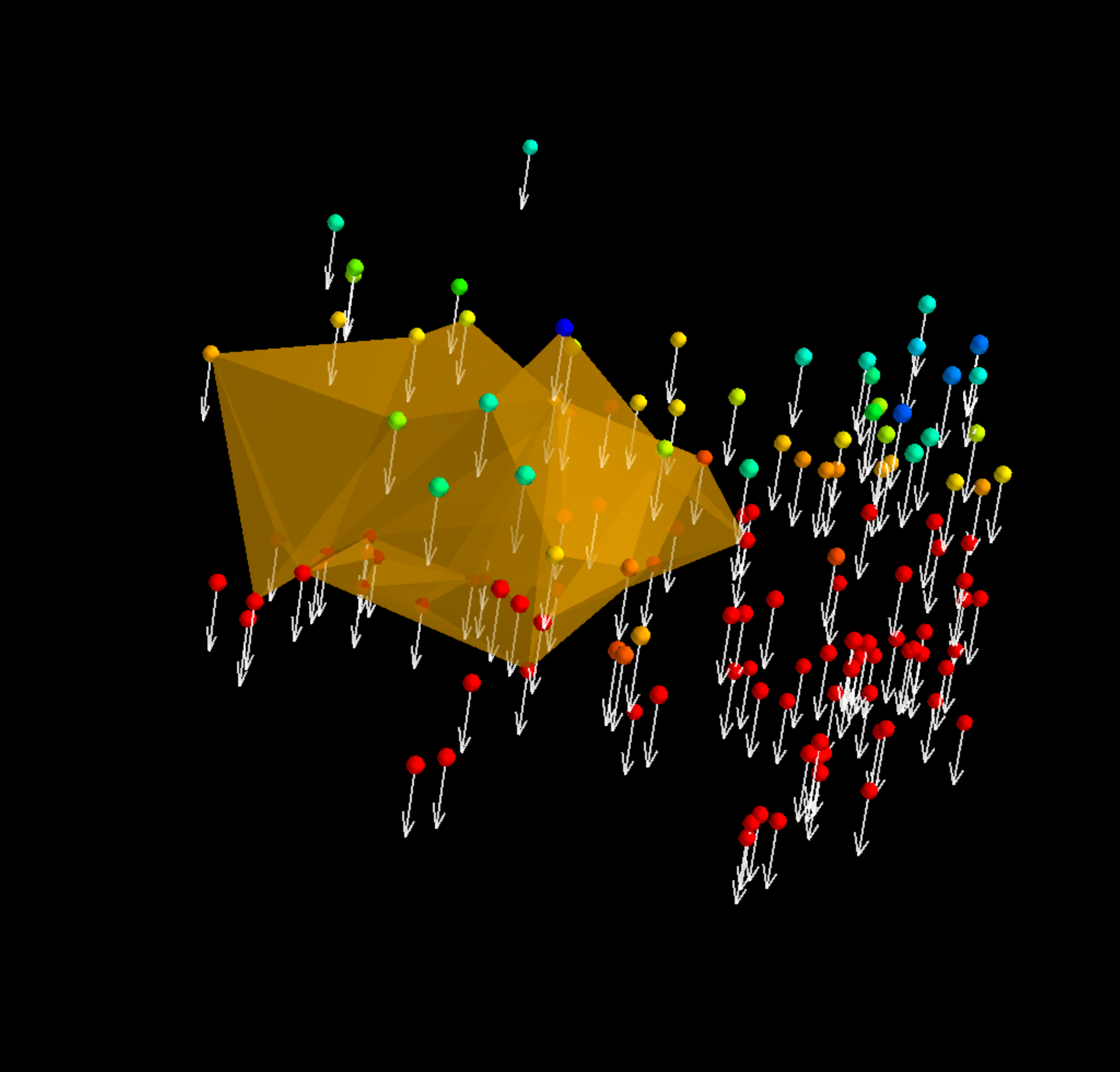}
        \captionsetup{labelformat=empty}
        \caption{}
        \label{supp_fig:universe_1}
    \end{subfigure}
      \centering
      \begin{subfigure}{.18\textwidth} \centering
        \includegraphics[width=0.9\linewidth]{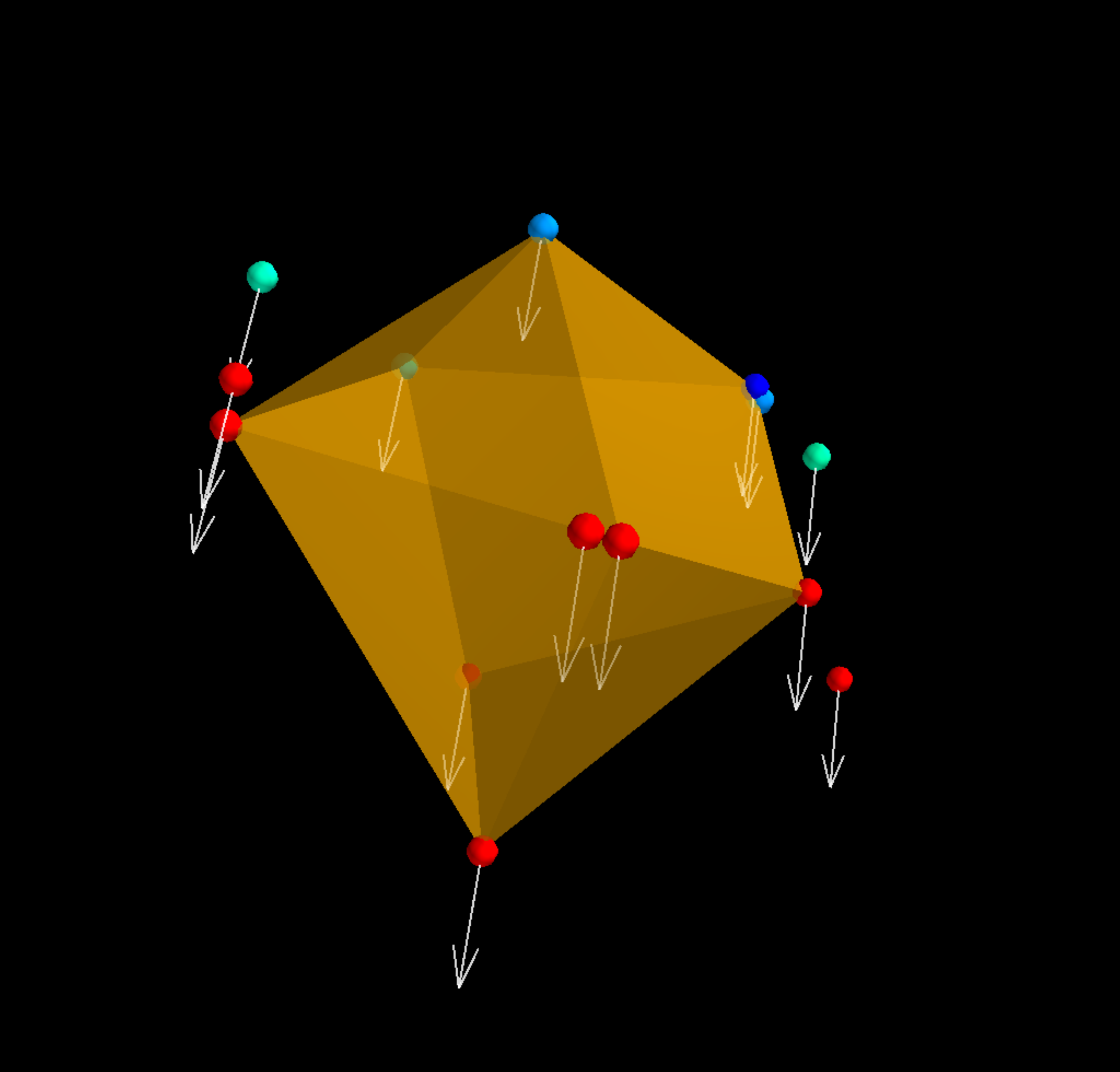}
        \captionsetup{labelformat=empty}
        \caption{}
        \label{supp_fig:universe_1}
    \end{subfigure}
      \centering
      \begin{subfigure}{.18\textwidth} \centering
        \includegraphics[width=0.9\linewidth]{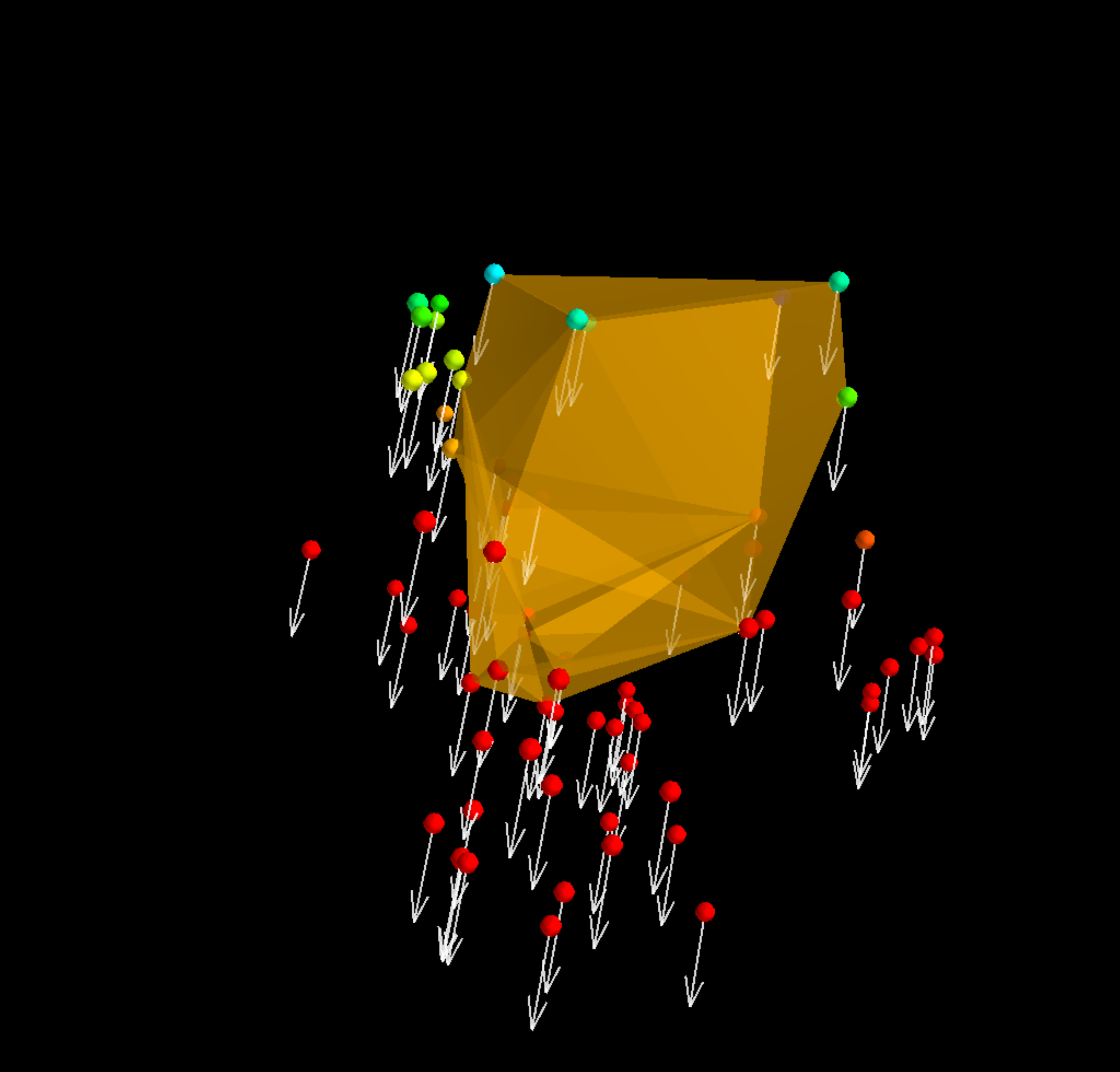}
        \captionsetup{labelformat=empty}
        \caption{}
        \label{supp_fig:universe_1}
    \end{subfigure}
      \centering
      \begin{subfigure}{.18\textwidth} \centering
        \includegraphics[width=0.9\linewidth]{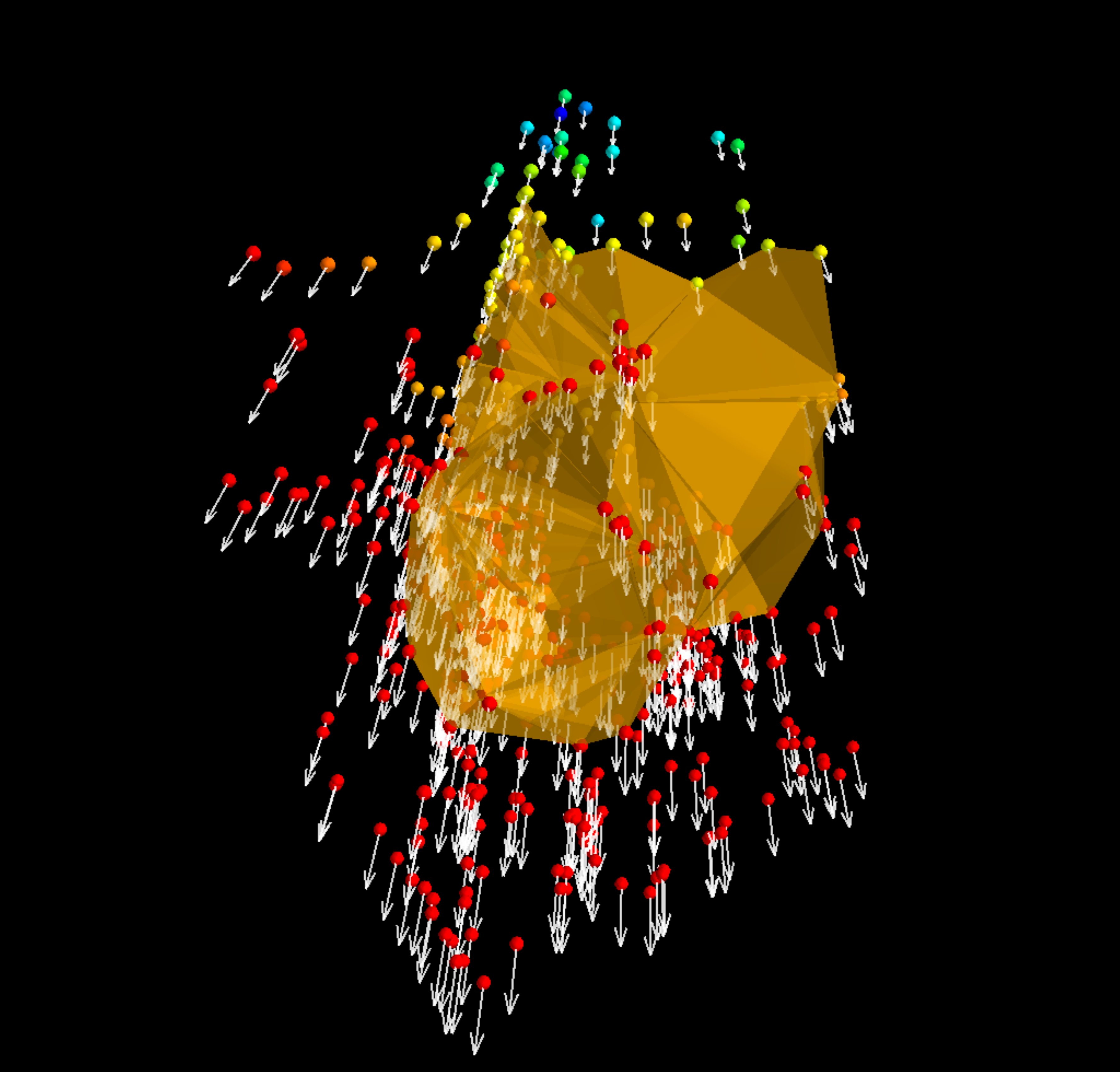}
        \captionsetup{labelformat=empty}
        \caption{}
        \label{supp_fig:universe_1}
    \end{subfigure}
      \centering
      \begin{subfigure}{.18\textwidth} \centering
        \includegraphics[width=0.9\linewidth]{figures/universe_final3d_auto_view_6_1.pdf}
        \captionsetup{labelformat=empty}
        \caption{}
        \label{supp_fig:universe_1}
    \end{subfigure}
      \centering
      \begin{subfigure}{.18\textwidth} \centering
        \includegraphics[width=0.9\linewidth]{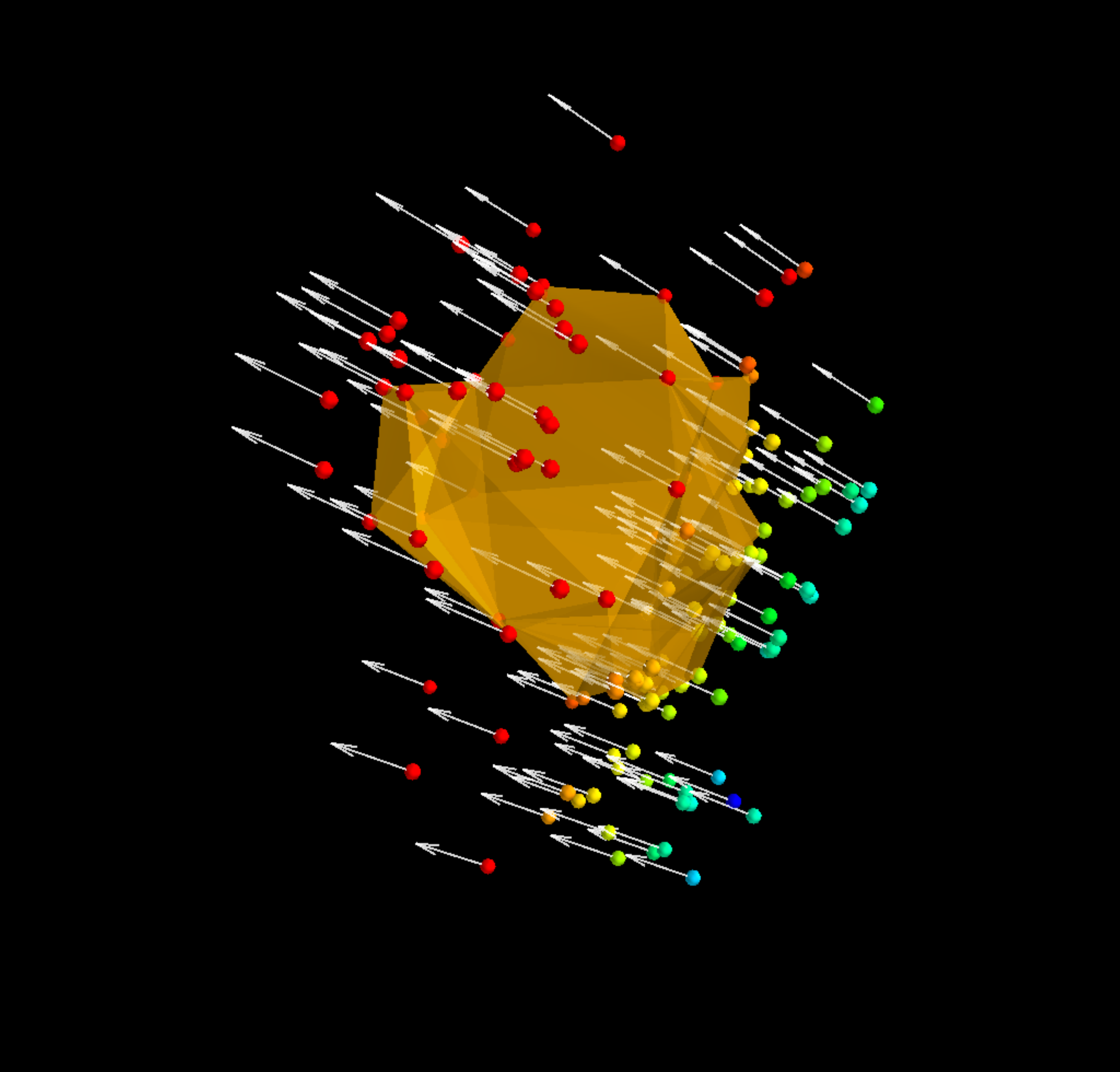}
        \captionsetup{labelformat=empty}
        \caption{}
        \label{supp_fig:universe_1}
    \end{subfigure}
      \centering
      \begin{subfigure}{.18\textwidth} \centering
        \includegraphics[width=0.9\linewidth]{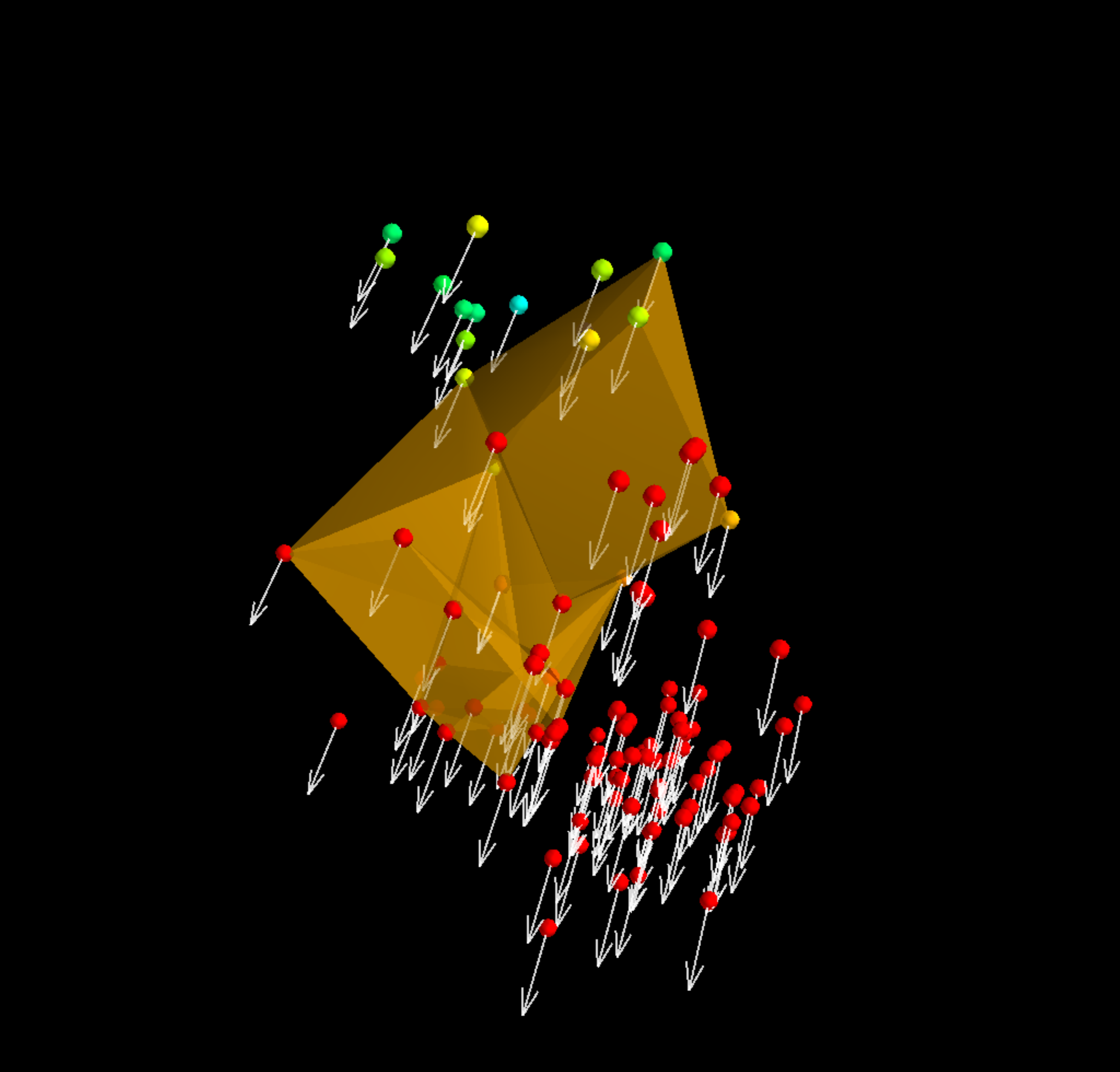}
        \captionsetup{labelformat=empty}
        \caption{}
        \label{supp_fig:universe_1}
    \end{subfigure}
      \centering
      \begin{subfigure}{.18\textwidth} \centering
        \includegraphics[width=0.9\linewidth]{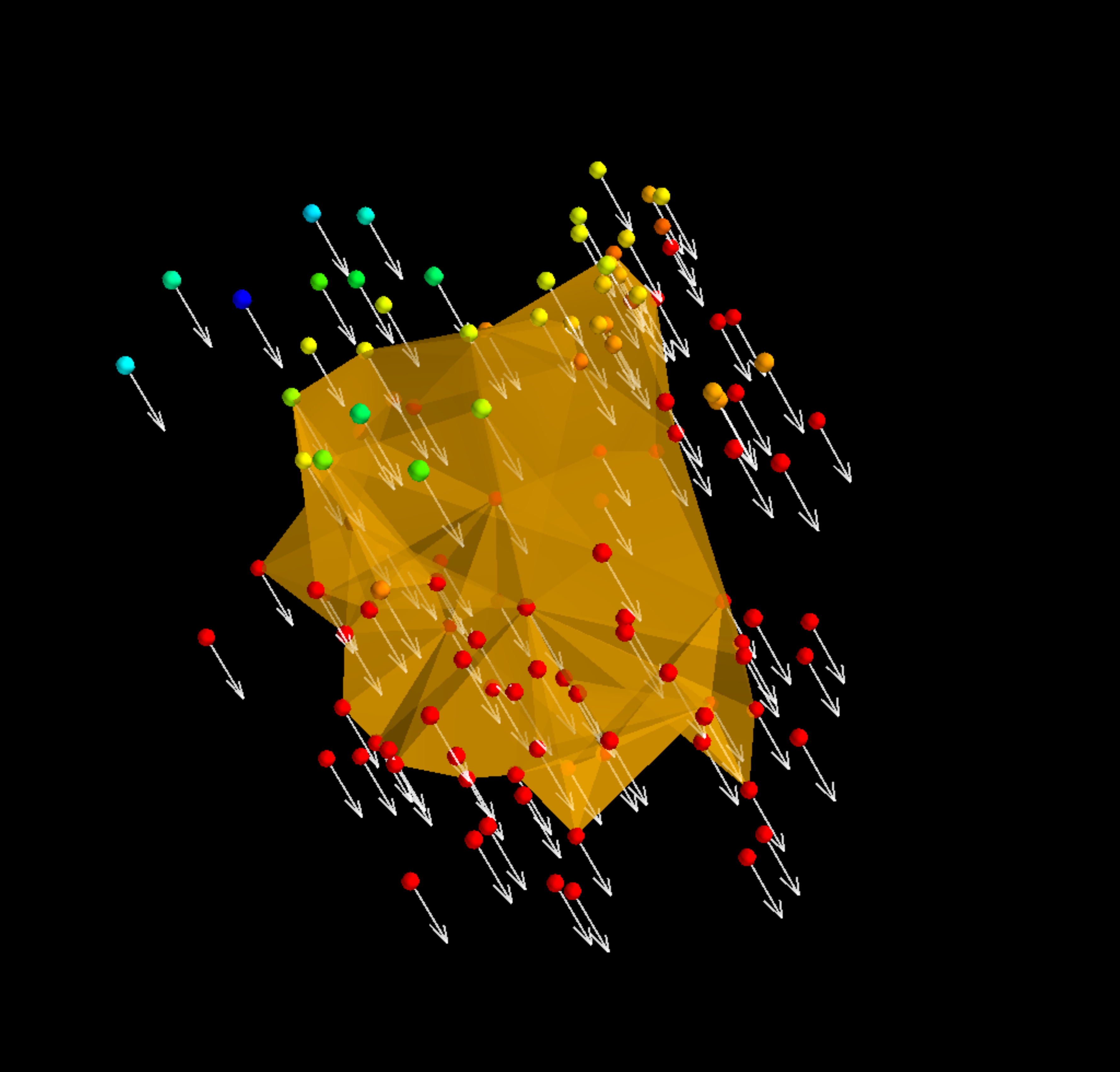}
        \captionsetup{labelformat=empty}
        \caption{}
        \label{supp_fig:universe_1}
    \end{subfigure}
      \centering
      \begin{subfigure}{.18\textwidth} \centering
        \includegraphics[width=0.9\linewidth]{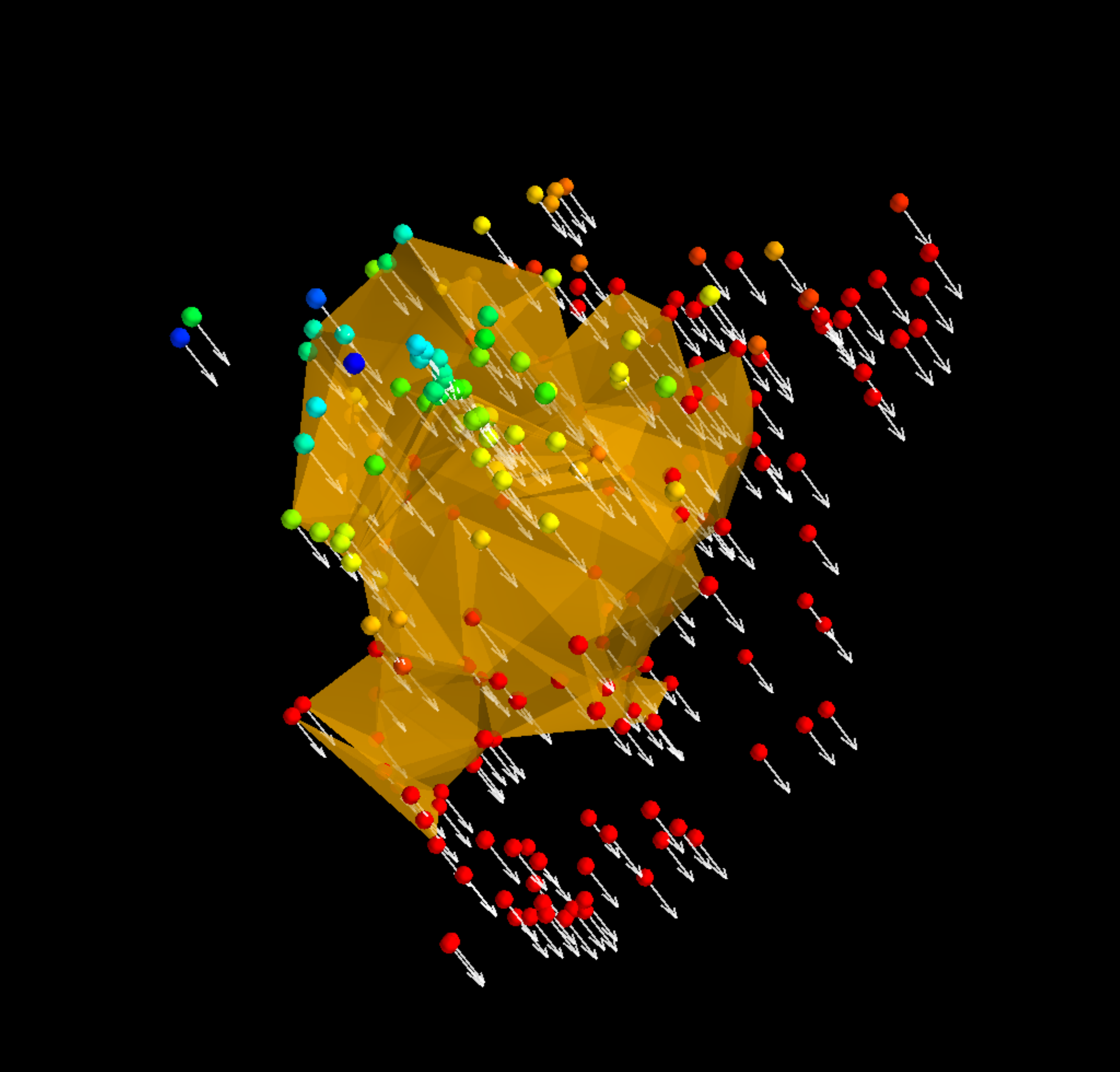}
        \captionsetup{labelformat=empty}
        \caption{}
        \label{supp_fig:universe_1}
    \end{subfigure}
      \centering
      \begin{subfigure}{.18\textwidth} \centering
        \includegraphics[width=0.9\linewidth]{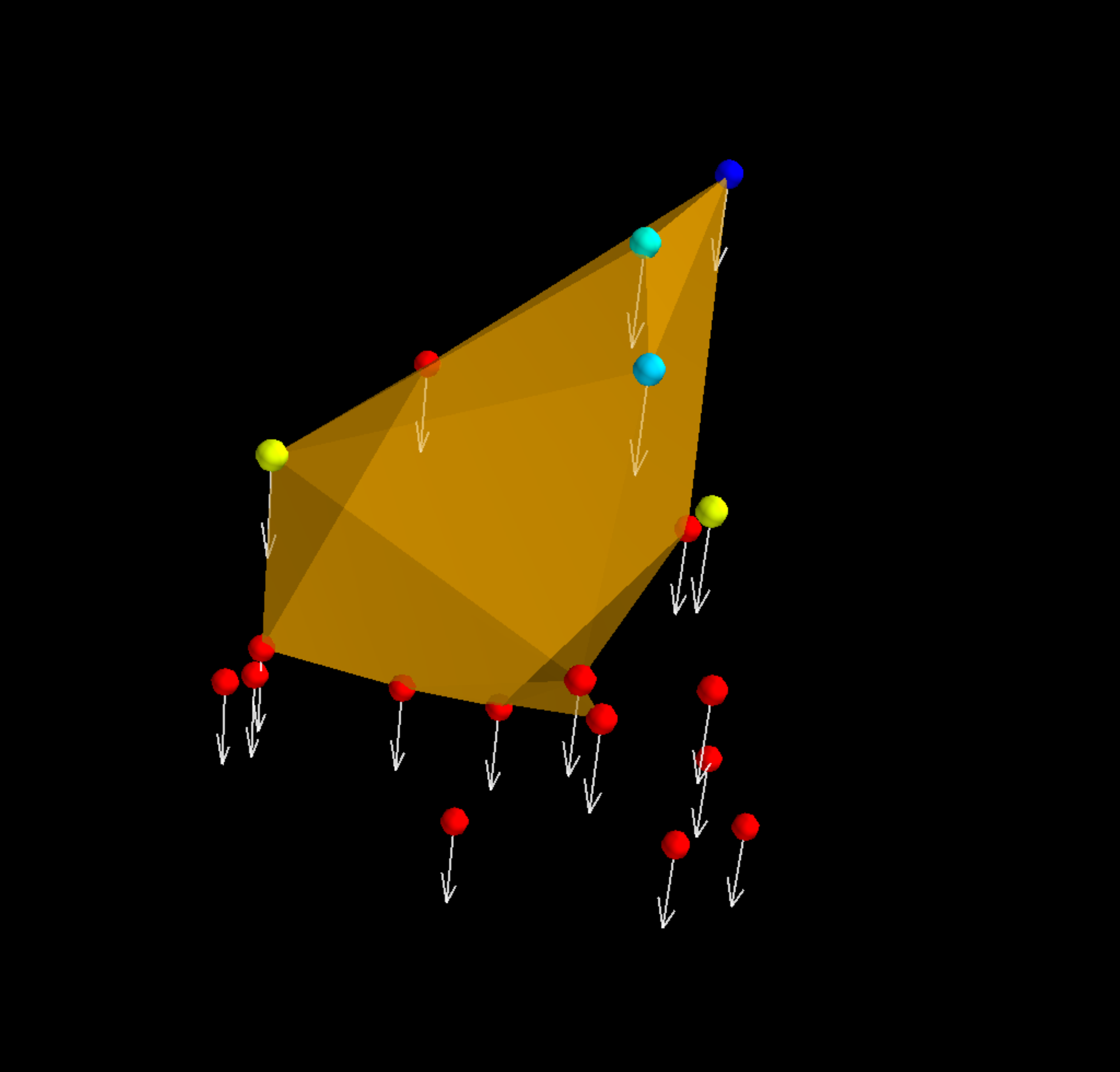}
        \captionsetup{labelformat=empty}
        \caption{}
        \label{supp_fig:universe_1}
    \end{subfigure}
      \centering
      \begin{subfigure}{.18\textwidth} \centering
        \includegraphics[width=0.9\linewidth]{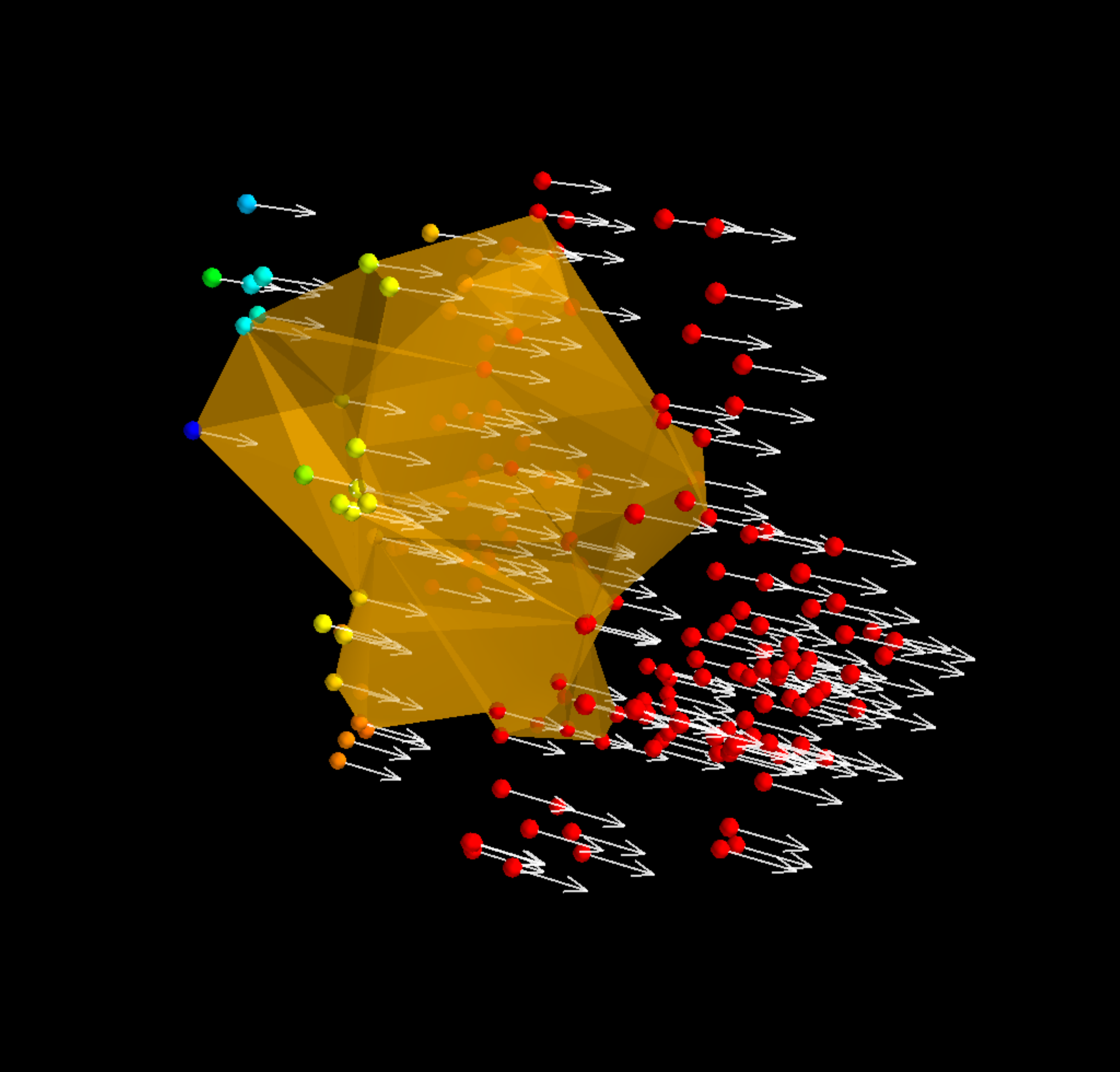}
        \captionsetup{labelformat=empty}
        \caption{}
        \label{supp_fig:universe_1}
    \end{subfigure}
      \centering
      \begin{subfigure}{.18\textwidth} \centering
        \includegraphics[width=0.9\linewidth]{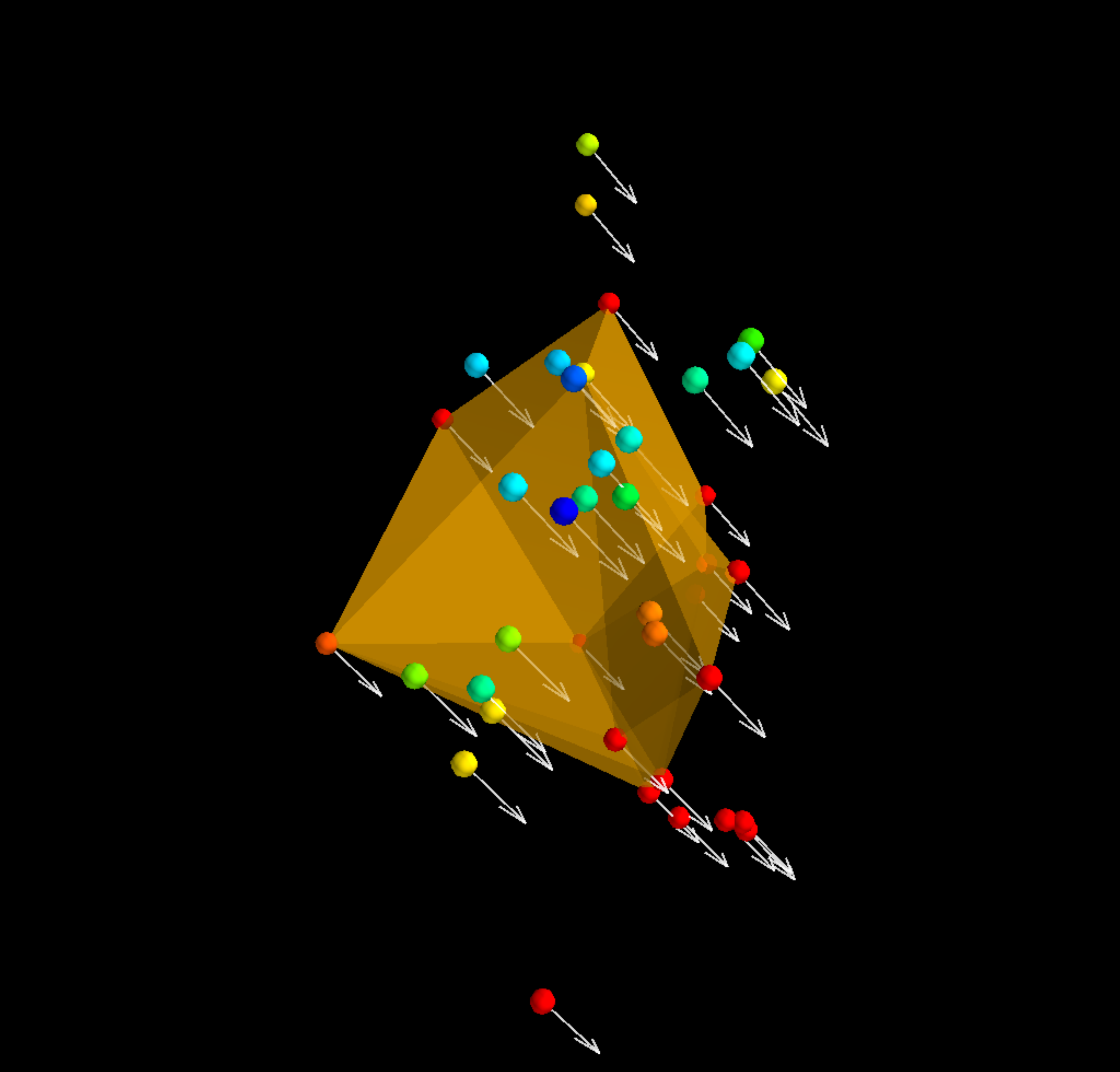}
        \captionsetup{labelformat=empty}
        \caption{}
        \label{supp_fig:universe_1}
    \end{subfigure}
      \centering
      \begin{subfigure}{.18\textwidth} \centering
        \includegraphics[width=0.9\linewidth]{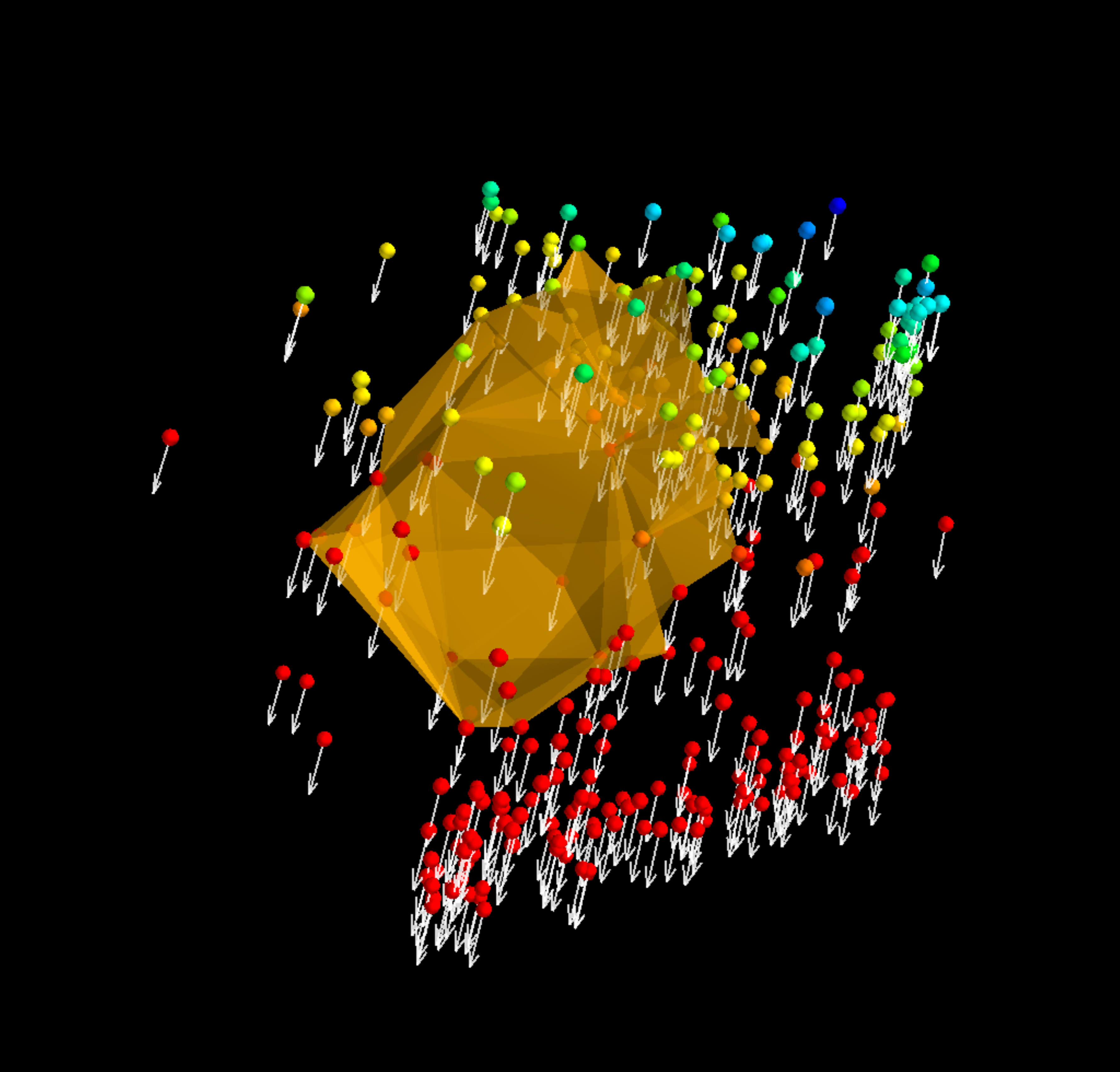}
        \captionsetup{labelformat=empty}
        \caption{}
        \label{supp_fig:universe_1}
    \end{subfigure}
      \centering
      \begin{subfigure}{.18\textwidth} \centering
        \includegraphics[width=0.9\linewidth]{figures/universe_final3d_auto_view_17_1.pdf}
        \captionsetup{labelformat=empty}
        \caption{}
        \label{supp_fig:universe_1}
    \end{subfigure}
      \centering
      \begin{subfigure}{.18\textwidth} \centering
        \includegraphics[width=0.9\linewidth]{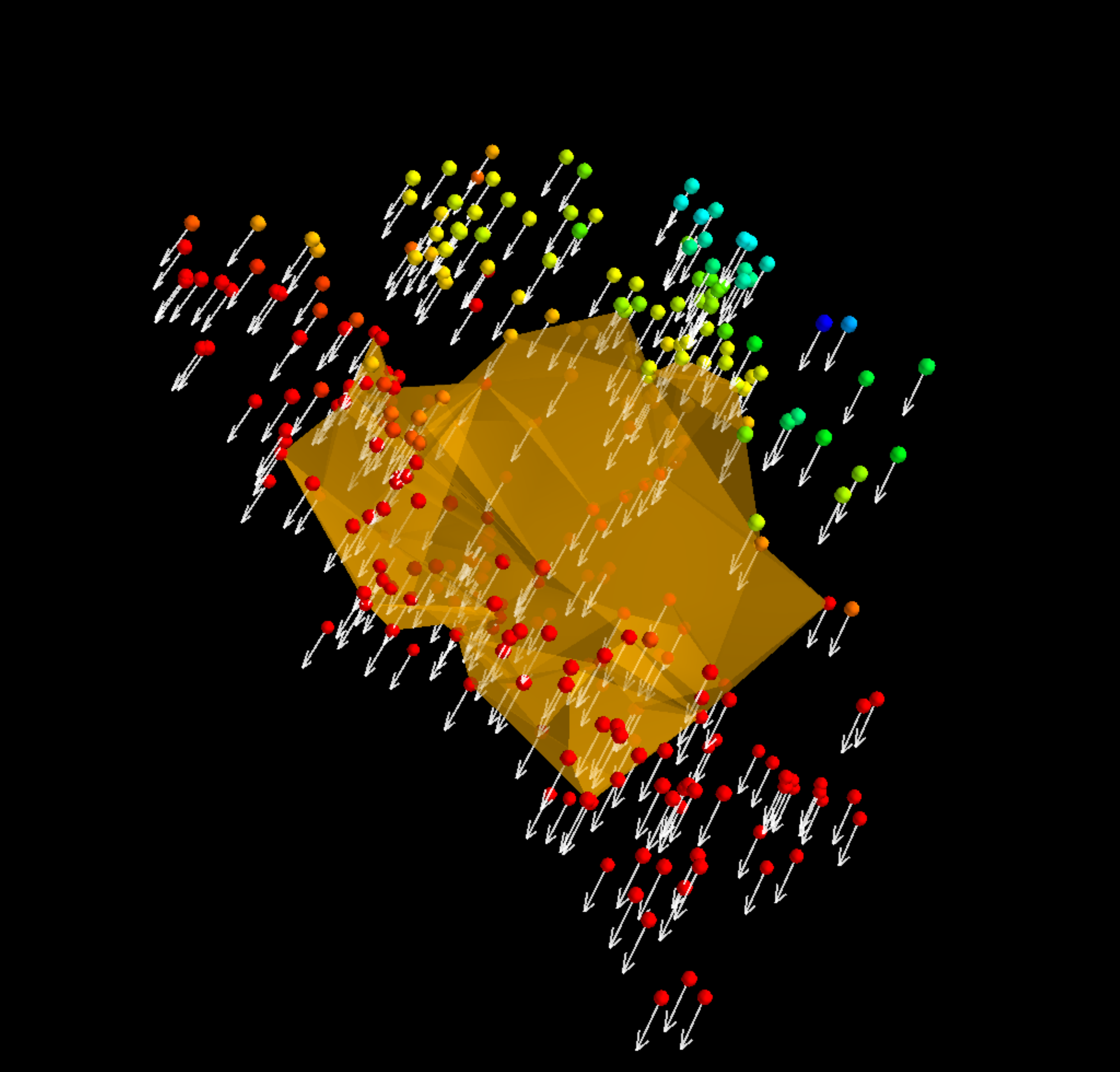}
        \captionsetup{labelformat=empty}
        \caption{}
        \label{supp_fig:universe_1}
    \end{subfigure}
      \centering
      \begin{subfigure}{.18\textwidth} \centering
        \includegraphics[width=0.9\linewidth]{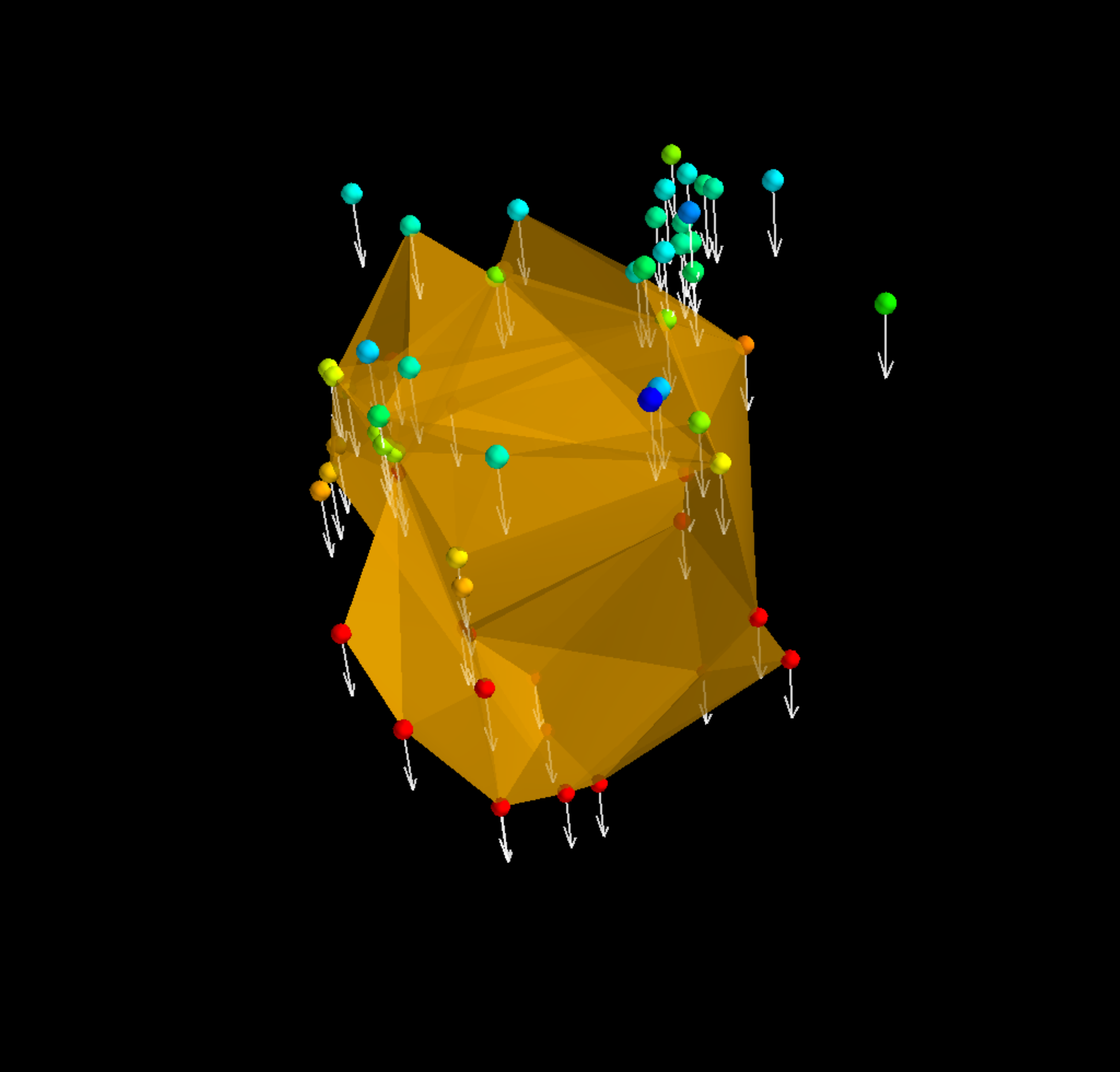}
        \captionsetup{labelformat=empty}
        \caption{}
        \label{supp_fig:universe_1}
    \end{subfigure}
      \centering
      \begin{subfigure}{.18\textwidth} \centering
        \includegraphics[width=0.9\linewidth]{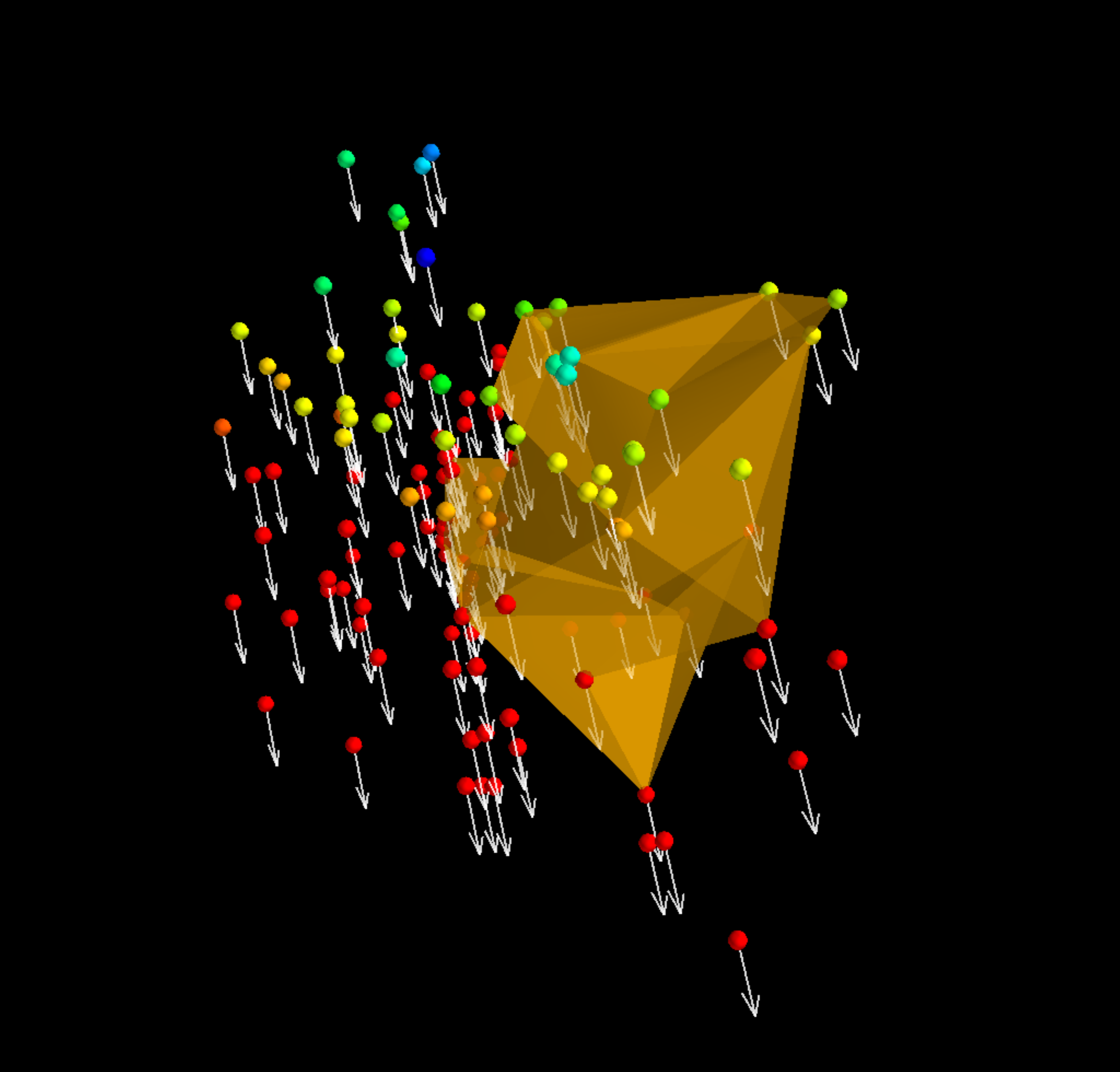}
        \captionsetup{labelformat=empty}
        \caption{}
        \label{supp_fig:universe_1}
    \end{subfigure}
      \centering
      \begin{subfigure}{.18\textwidth} \centering
        \includegraphics[width=0.9\linewidth]{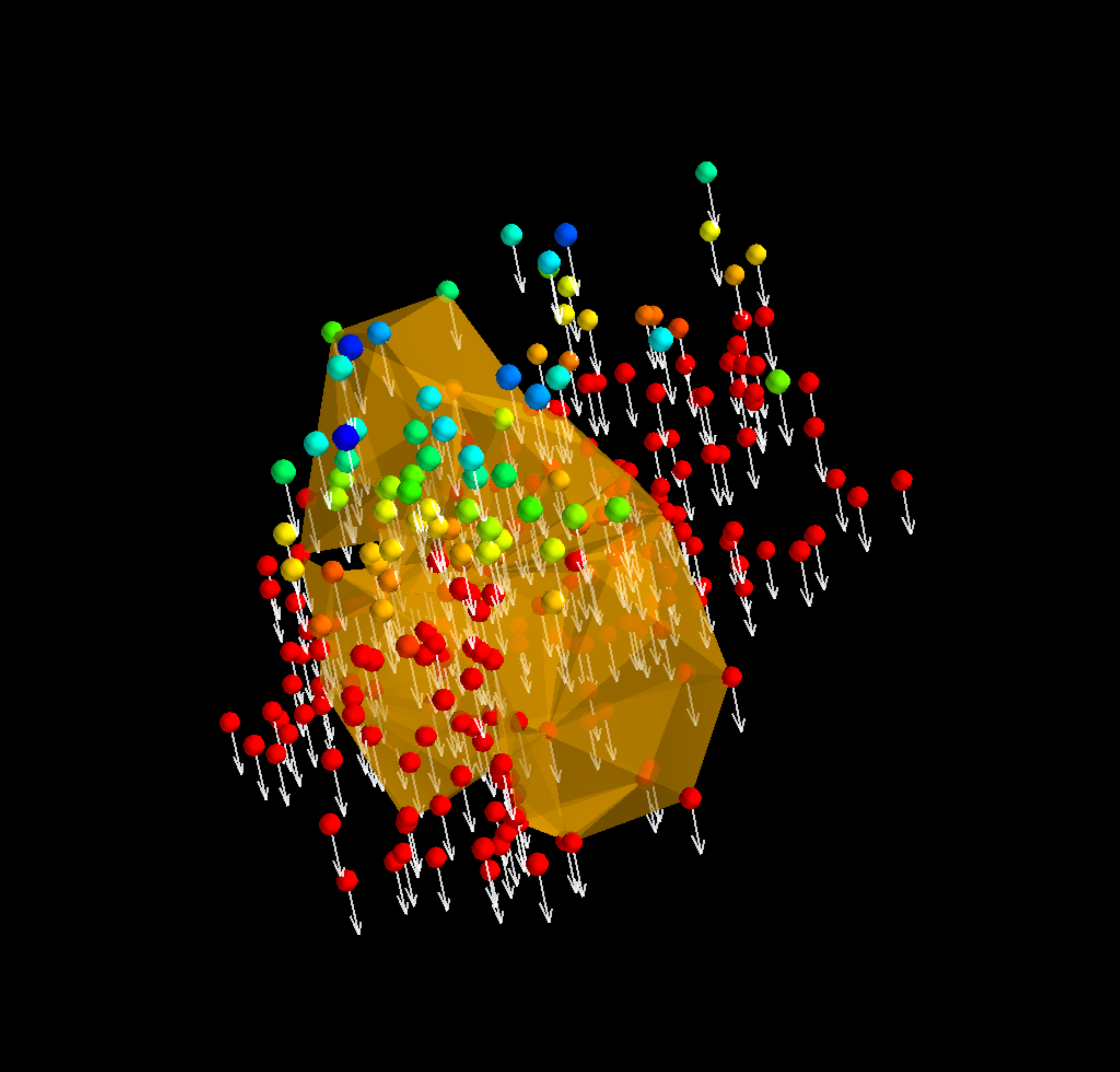}
        \captionsetup{labelformat=empty}
        \caption{}
        \label{supp_fig:universe_1}
    \end{subfigure}
      \centering
      \begin{subfigure}{.18\textwidth} \centering
        \includegraphics[width=0.9\linewidth]{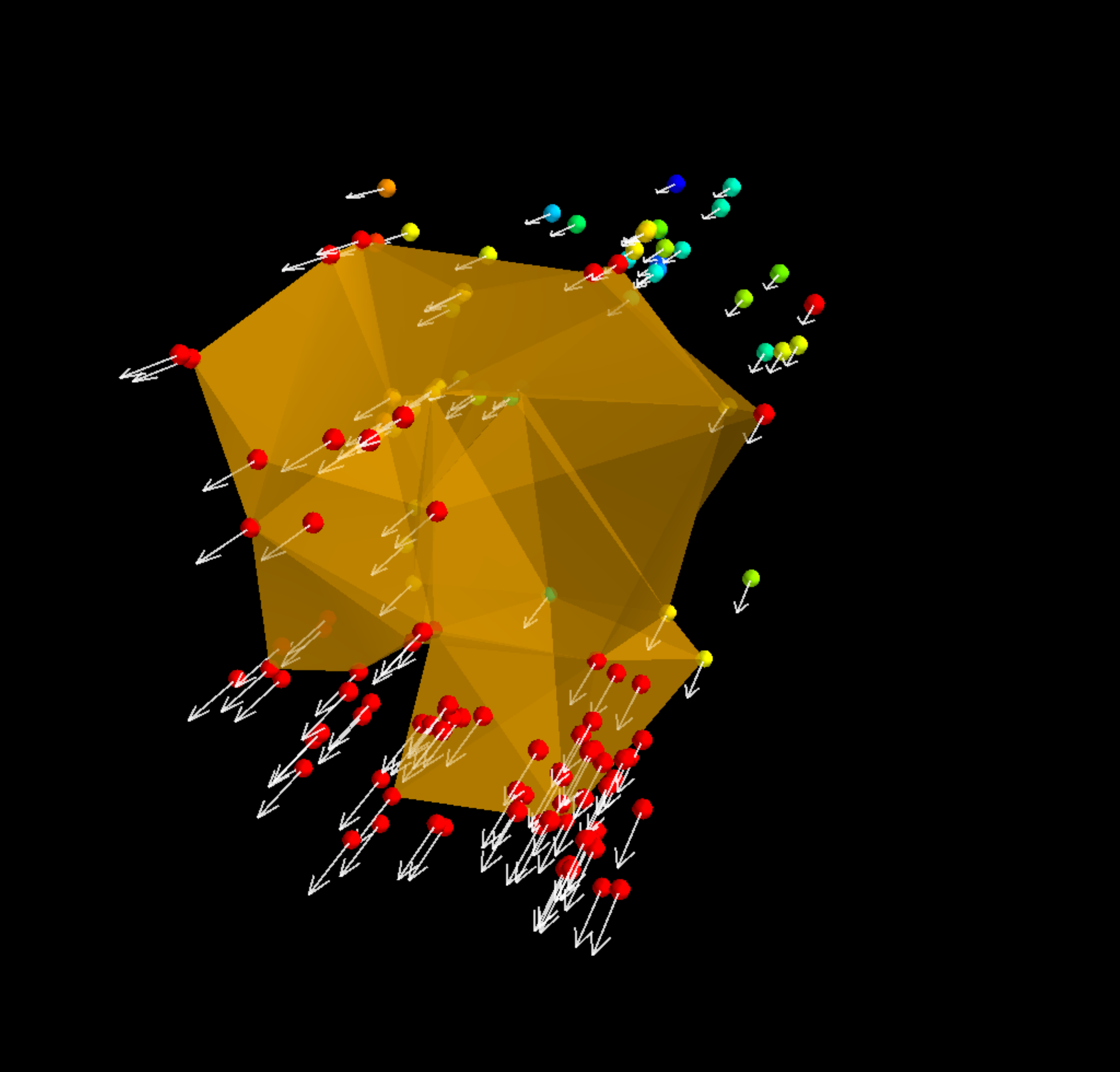}
        \captionsetup{labelformat=empty}
        \caption{}
        \label{supp_fig:universe_1}
    \end{subfigure}
      \centering
      \begin{subfigure}{.18\textwidth} \centering
        \includegraphics[width=0.9\linewidth]{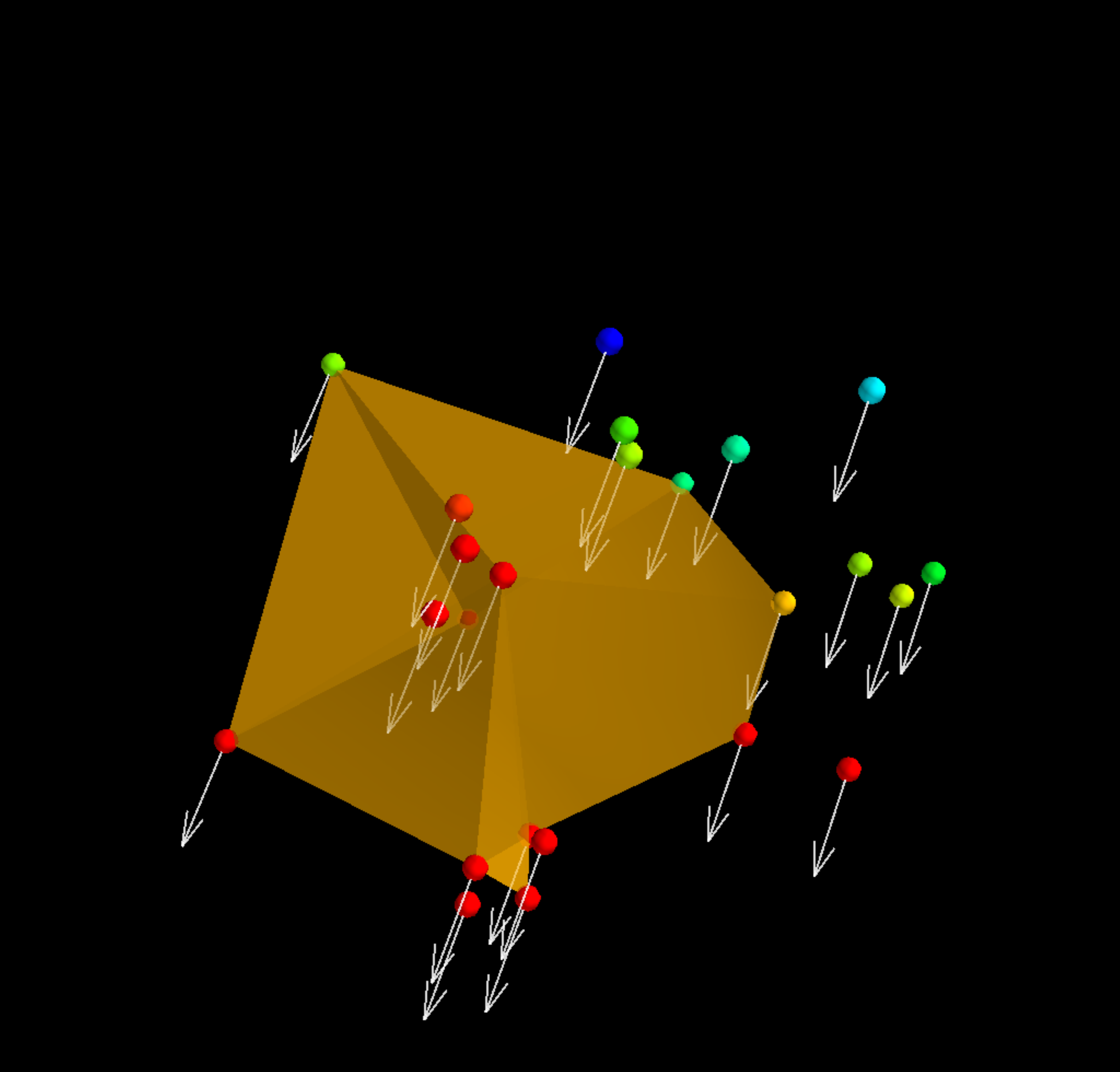}
        \captionsetup{labelformat=empty}
        \caption{}
        \label{supp_fig:universe_1}
    \end{subfigure}
      \centering
      \begin{subfigure}{.18\textwidth} \centering
        \includegraphics[width=0.9\linewidth]{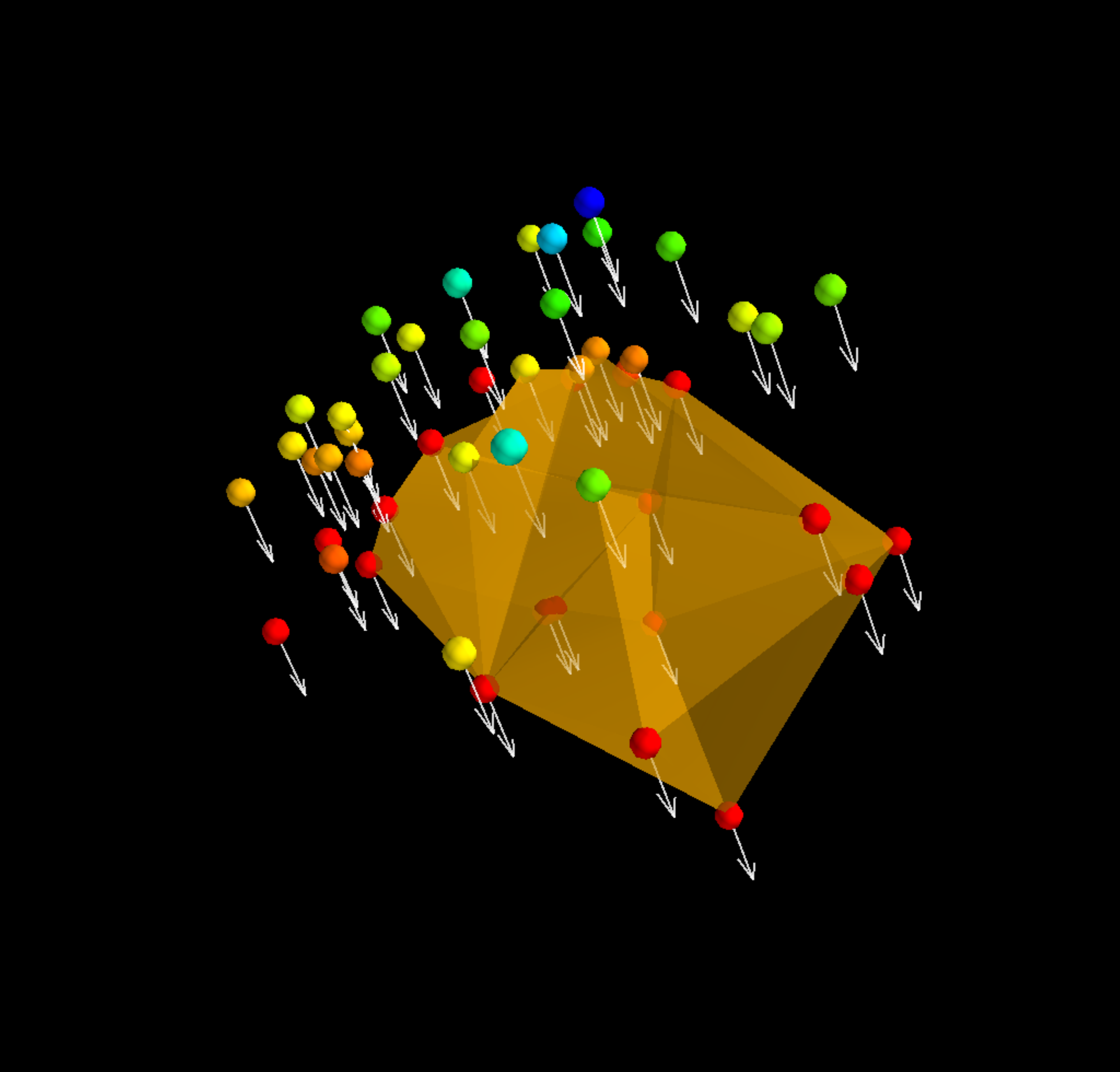}
        \captionsetup{labelformat=empty}
        \caption{}
        \label{supp_fig:universe_1}
    \end{subfigure}
      \centering
      \begin{subfigure}{.18\textwidth} \centering
        \includegraphics[width=0.9\linewidth]{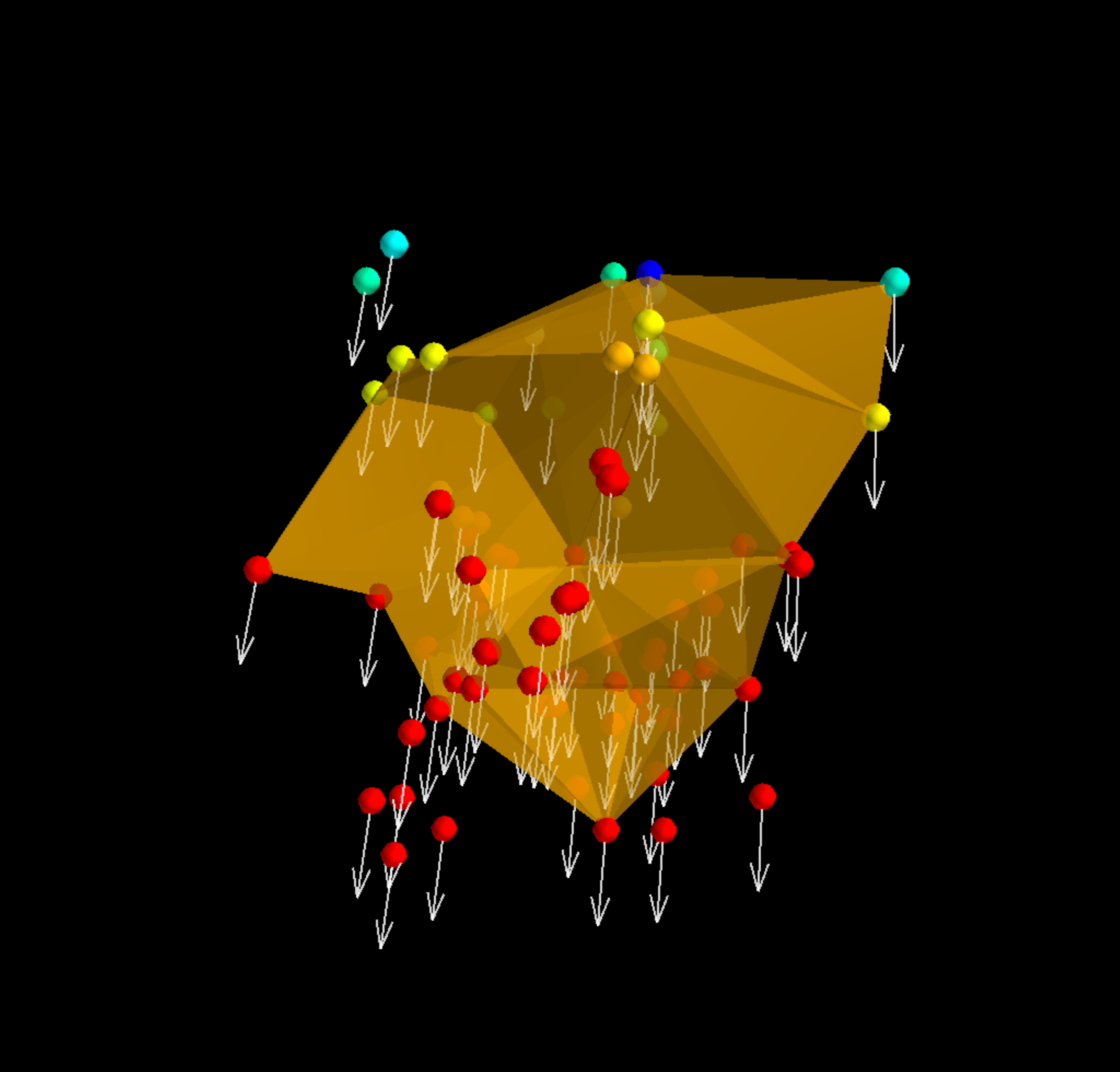}
        \captionsetup{labelformat=empty}
        \caption{}
        \label{supp_fig:universe_1}
    \end{subfigure}
      \centering
      \begin{subfigure}{.18\textwidth} \centering
        \includegraphics[width=0.9\linewidth]{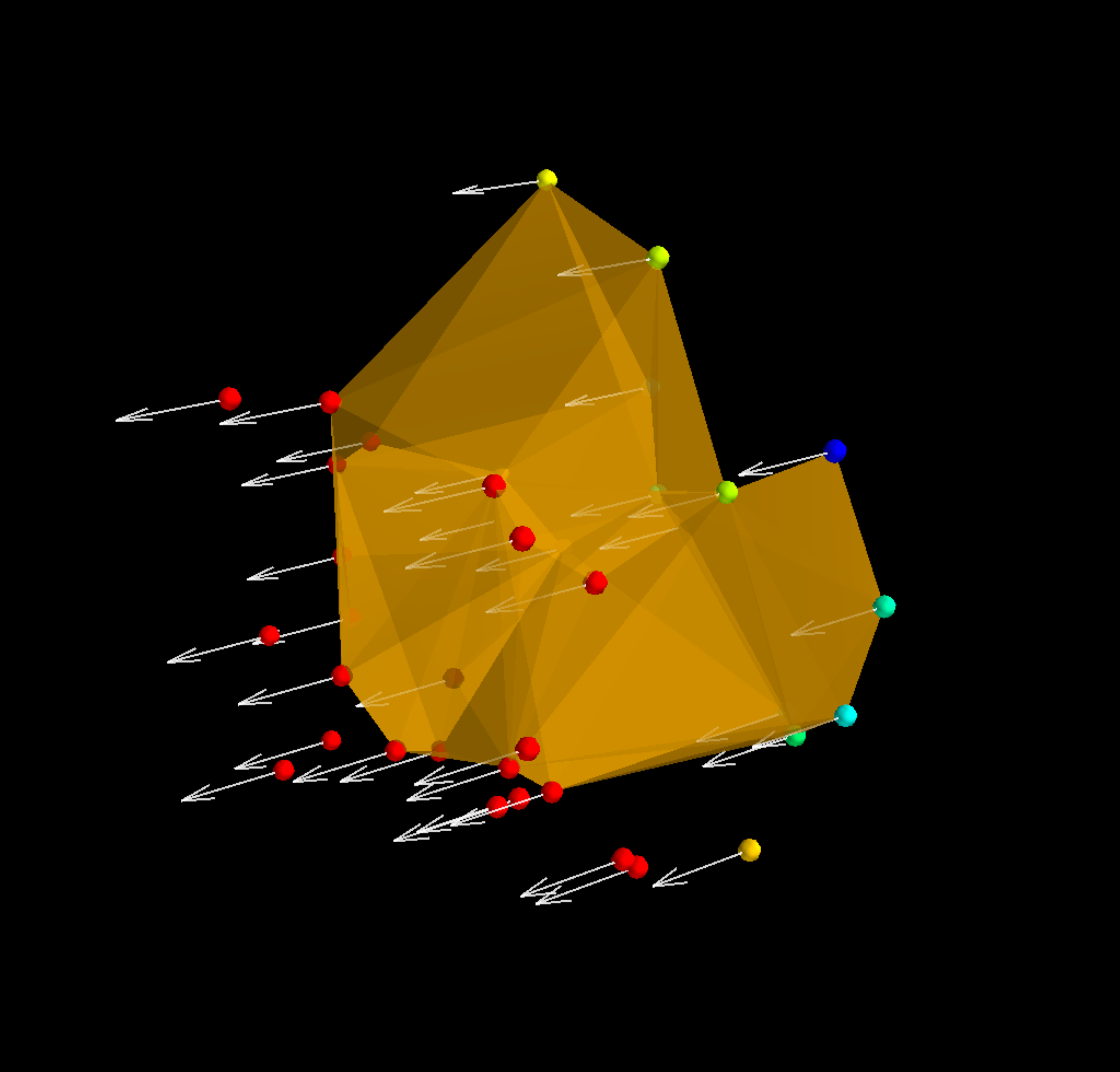}
        \captionsetup{labelformat=empty}
        \caption{}
        \label{supp_fig:universe_1}
    \end{subfigure}
      \centering
      \begin{subfigure}{.18\textwidth} \centering
        \includegraphics[width=0.9\linewidth]{figures/universe_final3d_auto_view_28_1.pdf}
        \captionsetup{labelformat=empty}
        \caption{}
        \label{supp_fig:universe_1}
    \end{subfigure}

      \caption{}
      \label{fig:universe_supp_all_voids}
    \end{figure}

Figure~\ref{fig:universe_supp_all_voids} shows the computed set of minimal representative boundaries
around significant voids in the universe.

\section{PDB analysis}\label{supp:pdb_homologs}

    \begin{figure}[!tbhp]
      \centering
      \begin{subfigure}{.48\textwidth} \centering
        \includegraphics[width=0.9\linewidth]{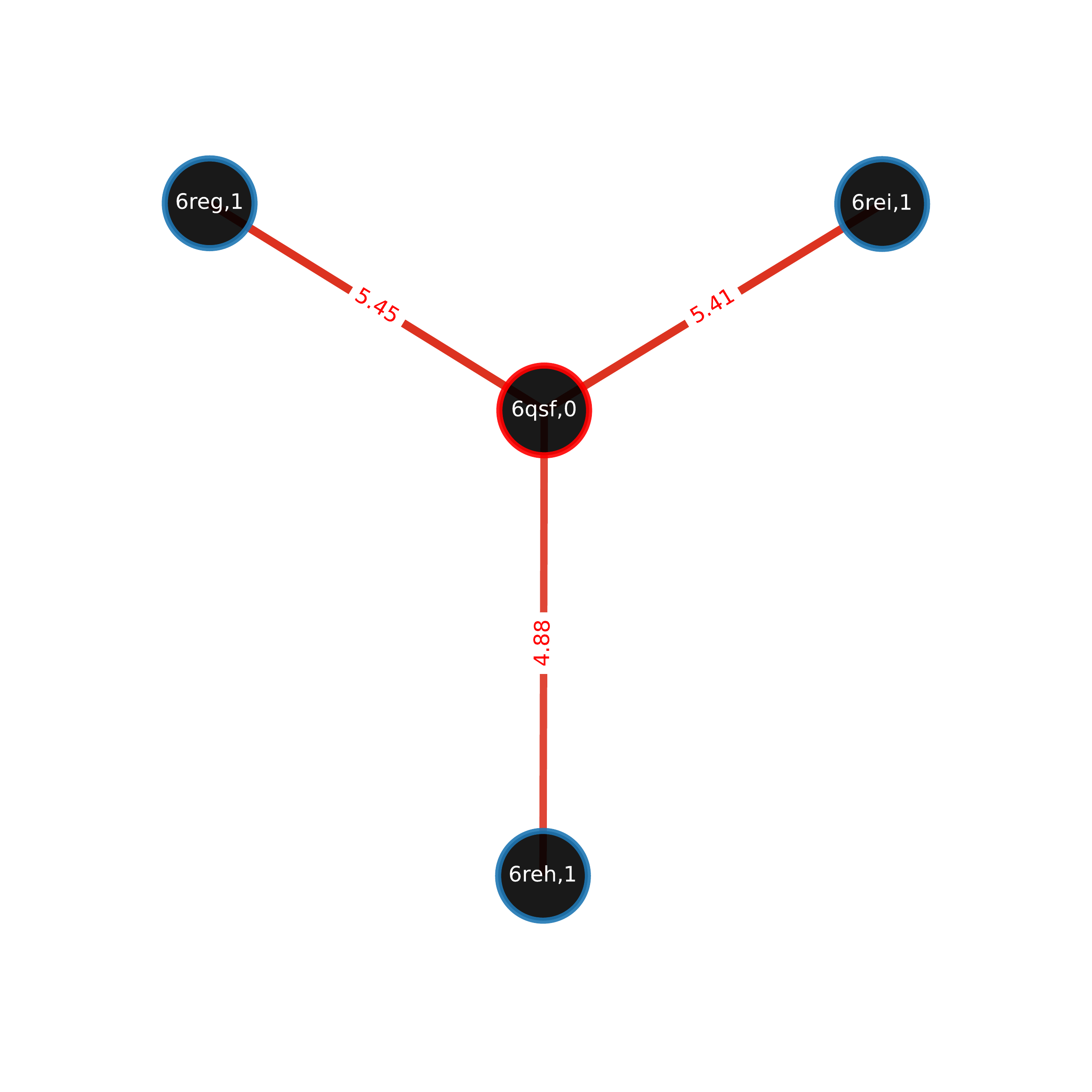}
        \captionsetup{labelformat=empty}
        \caption{}
        \label{supp_fig:pdb_hom_graph_1}
    \end{subfigure}
      \centering
      \begin{subfigure}{.48\textwidth} \centering
        \includegraphics[width=0.9\linewidth]{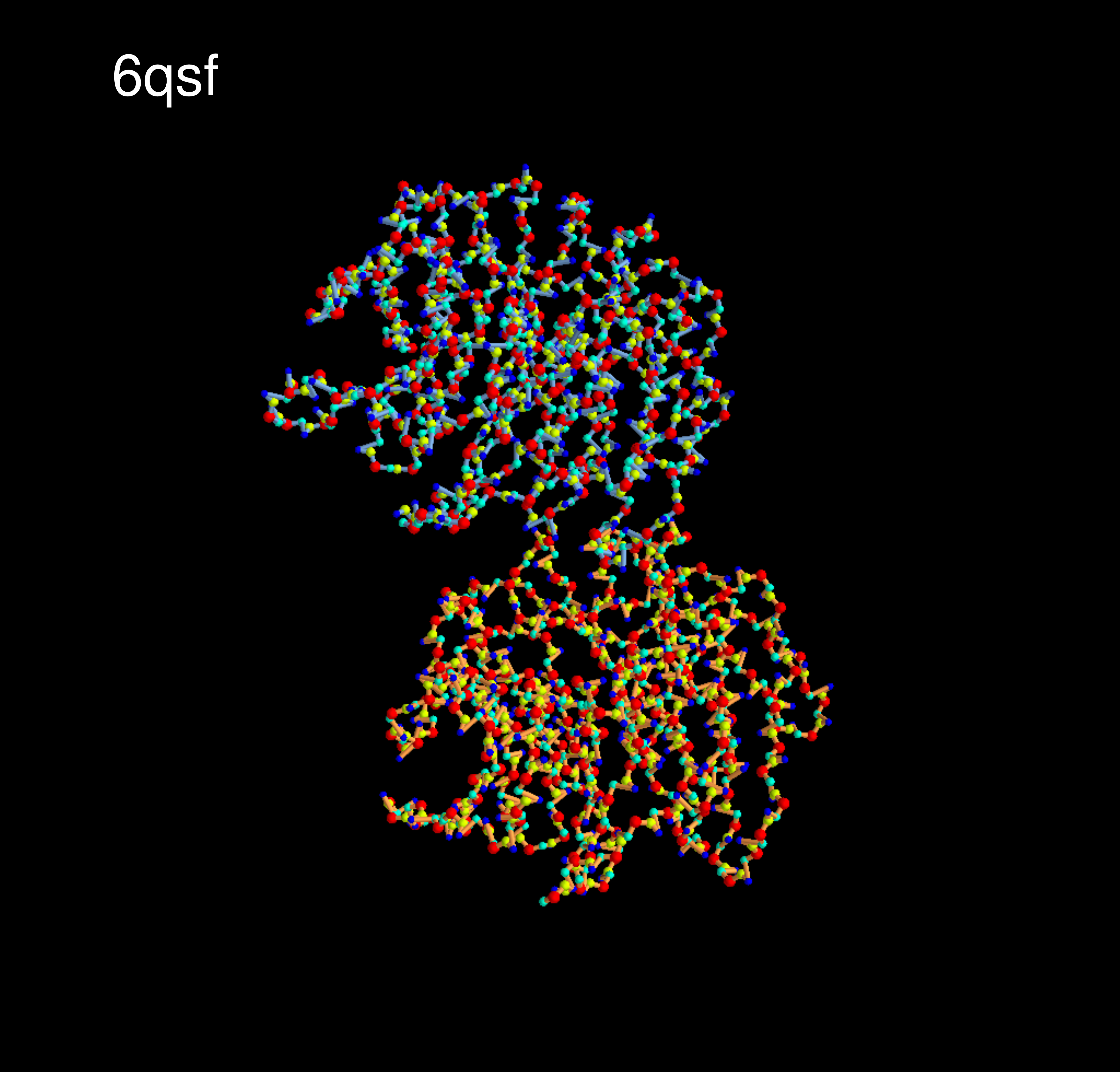}
        \captionsetup{labelformat=empty}
        \caption{}
        \label{supp_fig:pdb_6qsf}
    \end{subfigure}
      \centering
      \begin{subfigure}{.48\textwidth} \centering
        \includegraphics[width=0.9\linewidth]{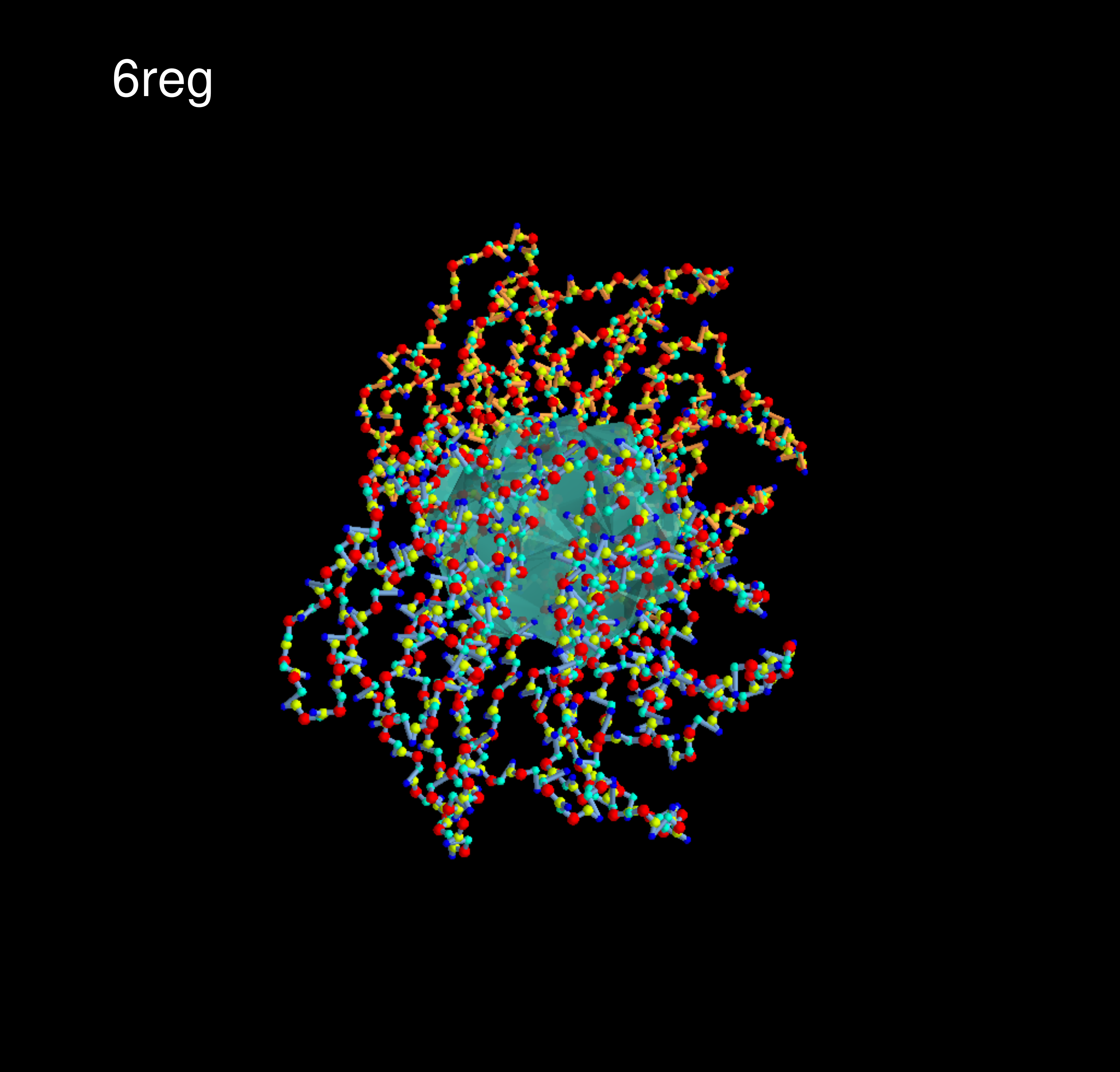}
        \captionsetup{labelformat=empty}
        \caption{}
        \label{supp_fig:pdb_6reg}
    \end{subfigure}
      \centering
      \begin{subfigure}{.48\textwidth} \centering
        \includegraphics[width=0.9\linewidth]{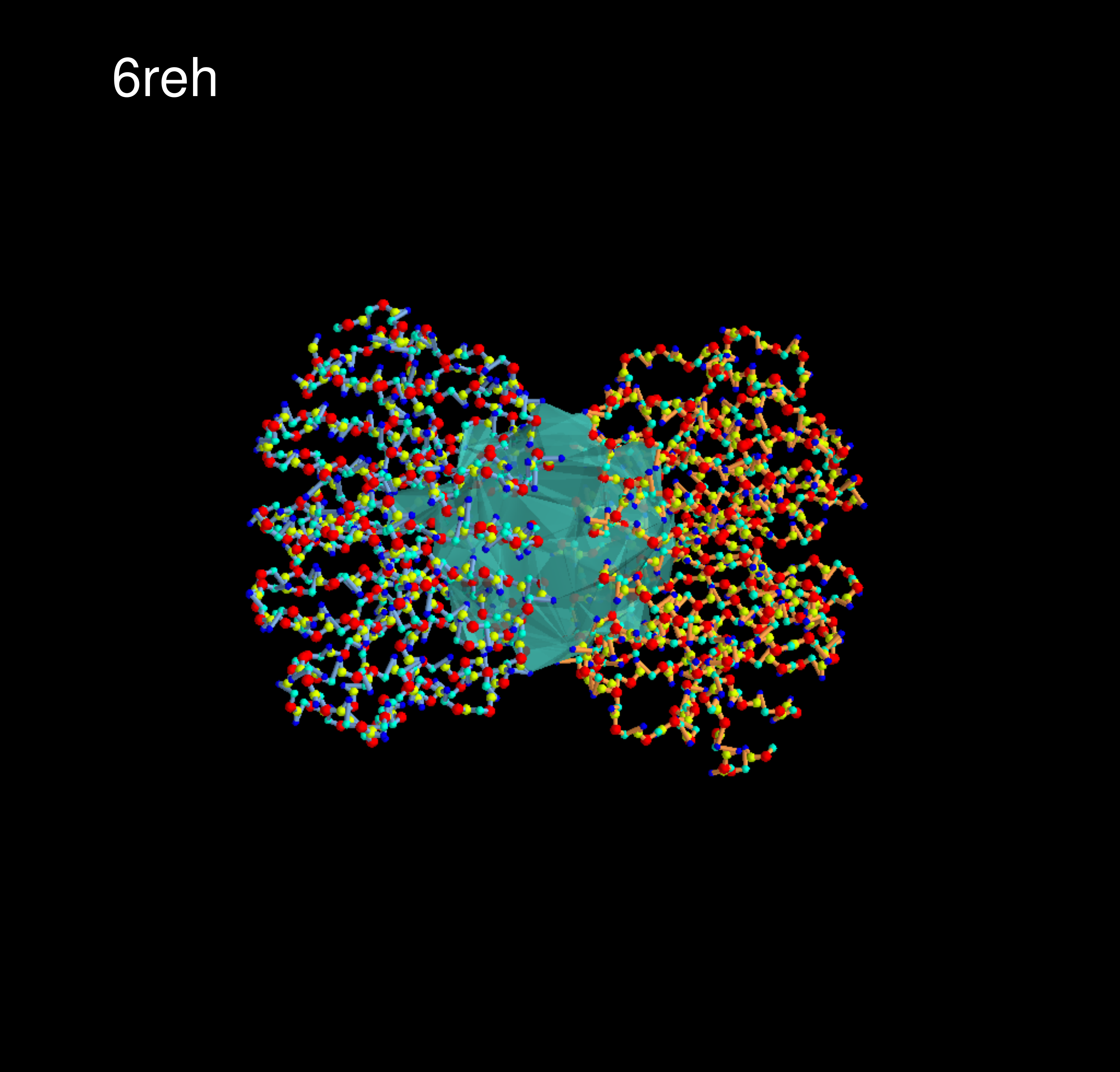}
        \captionsetup{labelformat=empty}
        \caption{}
        \label{supp_fig:pdb_6reh}
    \end{subfigure}
      \centering
      \begin{subfigure}{.48\textwidth} \centering
        \includegraphics[width=0.9\linewidth]{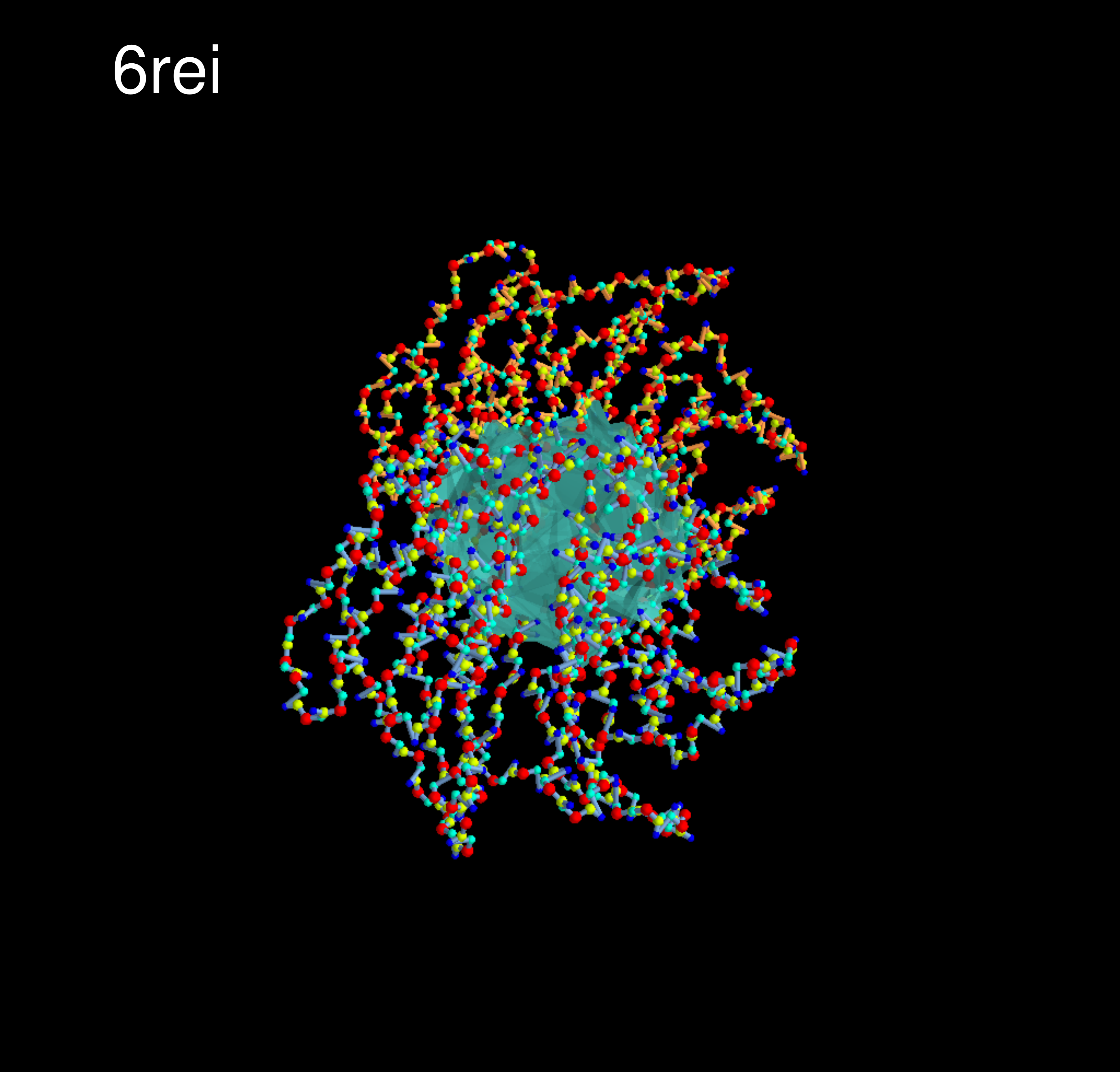}
        \captionsetup{labelformat=empty}
        \caption{}
        \label{supp_fig:pdb_6rei}
    \end{subfigure}

      \caption{}
      \label{fig:pdb_1}
    \end{figure}

    \begin{figure}[!tbhp]
      \centering
      \begin{subfigure}{.48\textwidth} \centering
        \includegraphics[width=0.9\linewidth]{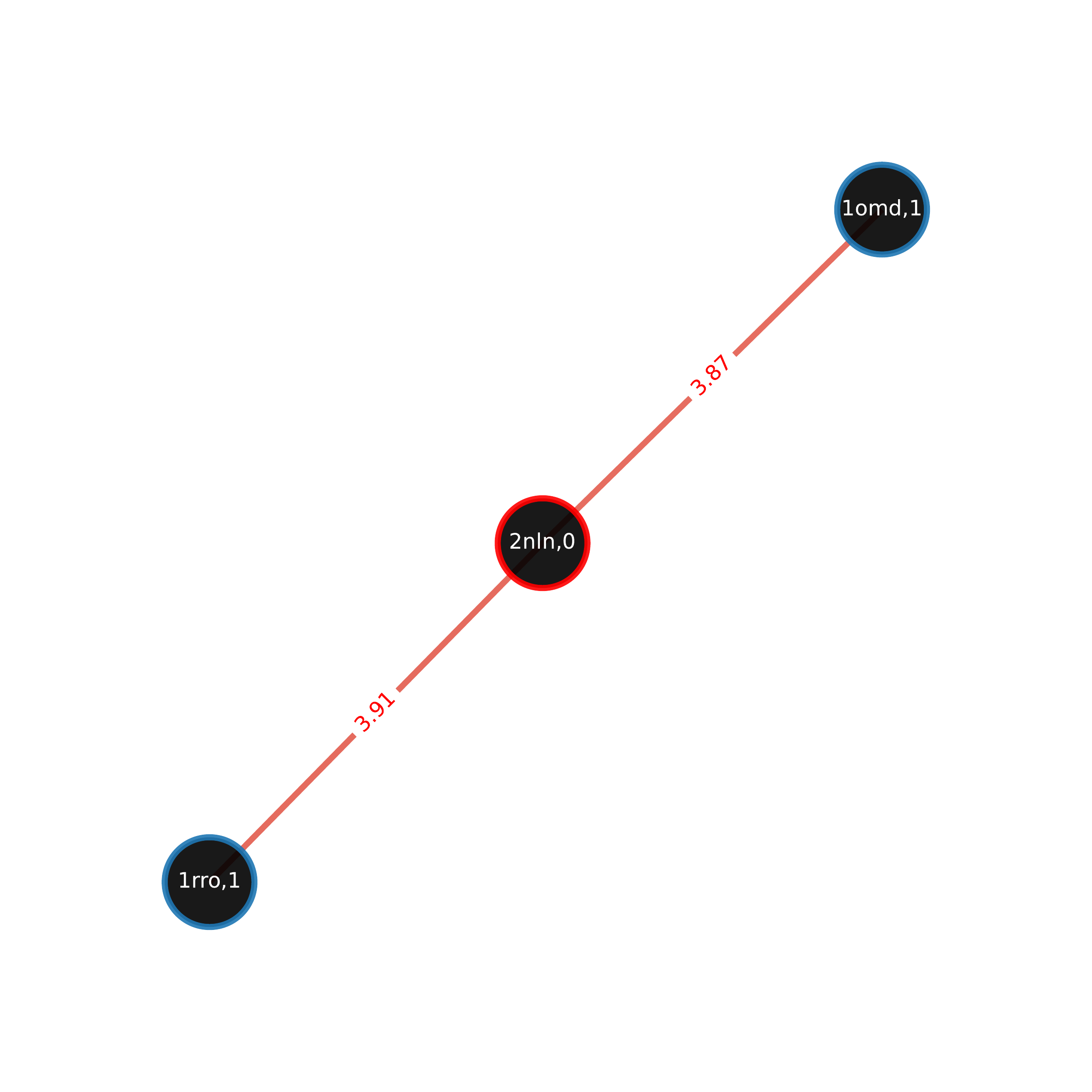}
        \captionsetup{labelformat=empty}
        \caption{}
        \label{supp_fig:pdb_hom_graph_2}
    \end{subfigure}
      \centering
      \begin{subfigure}{.48\textwidth} \centering
        \includegraphics[width=0.9\linewidth]{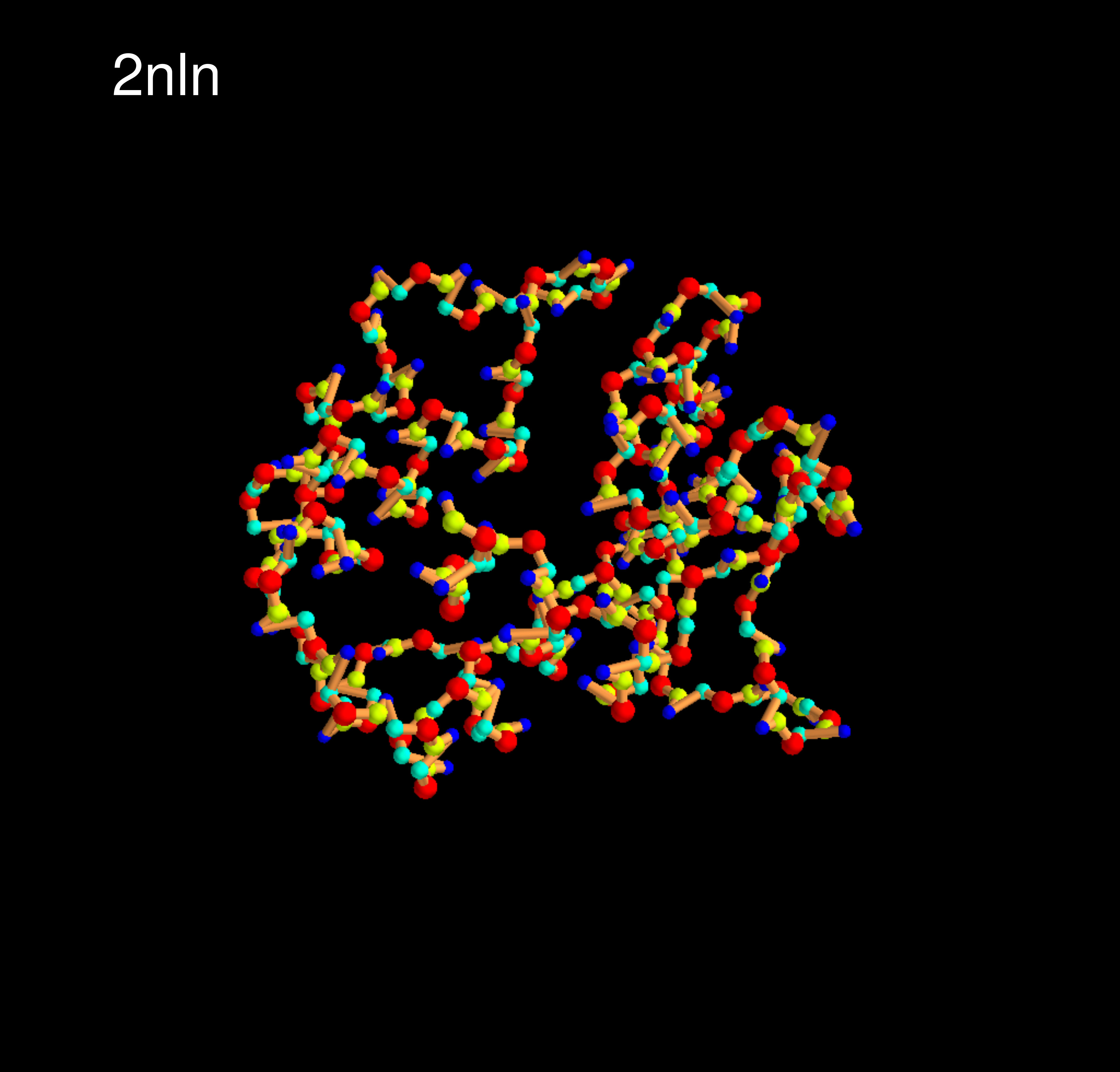}
        \captionsetup{labelformat=empty}
        \caption{}
        \label{supp_fig:pdb_2nln}
    \end{subfigure}
      \centering
      \begin{subfigure}{.48\textwidth} \centering
        \includegraphics[width=0.9\linewidth]{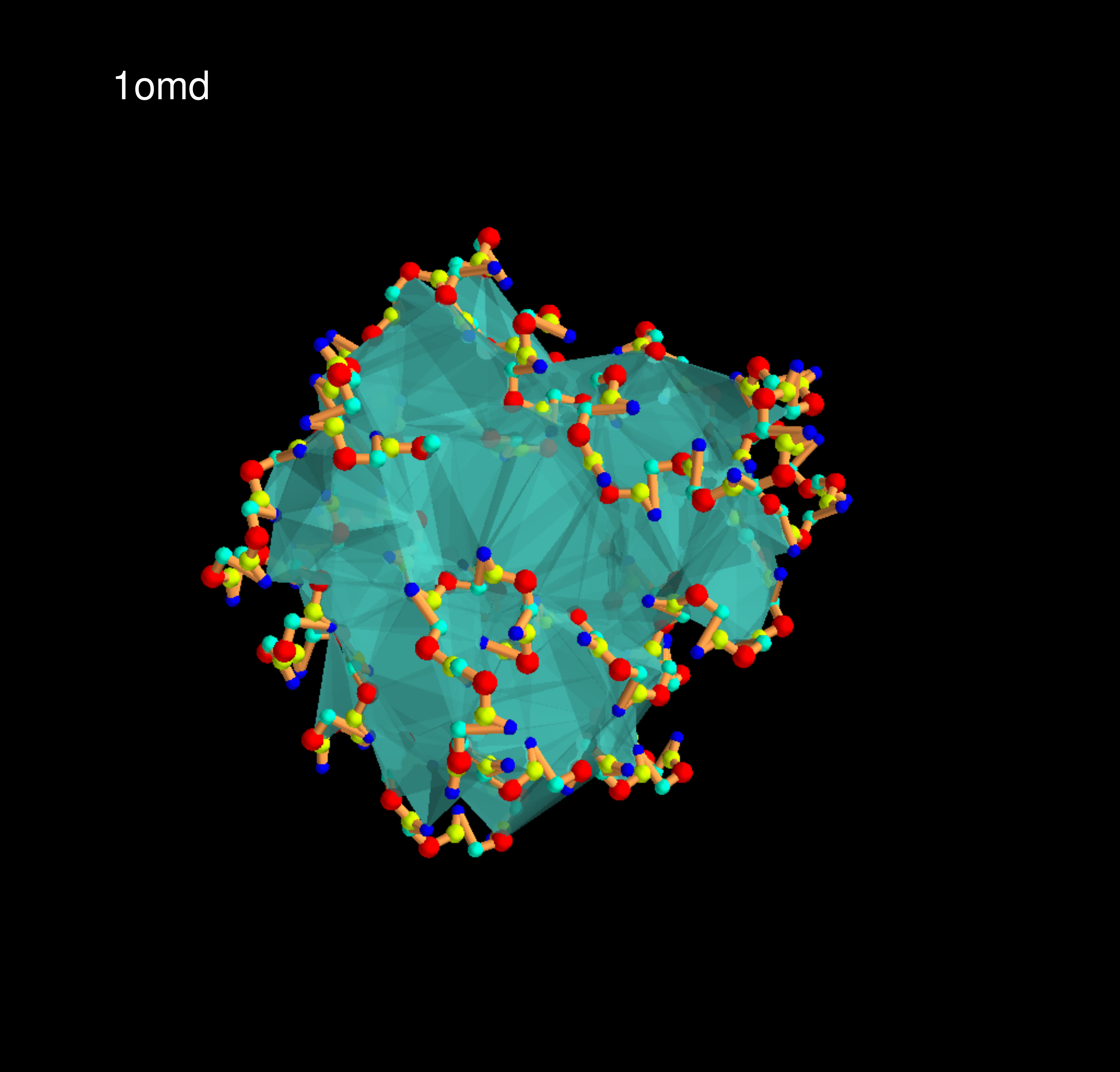}
        \captionsetup{labelformat=empty}
        \caption{}
        \label{supp_fig:pdb_1omd}
    \end{subfigure}
      \centering
      \begin{subfigure}{.48\textwidth} \centering
        \includegraphics[width=0.9\linewidth]{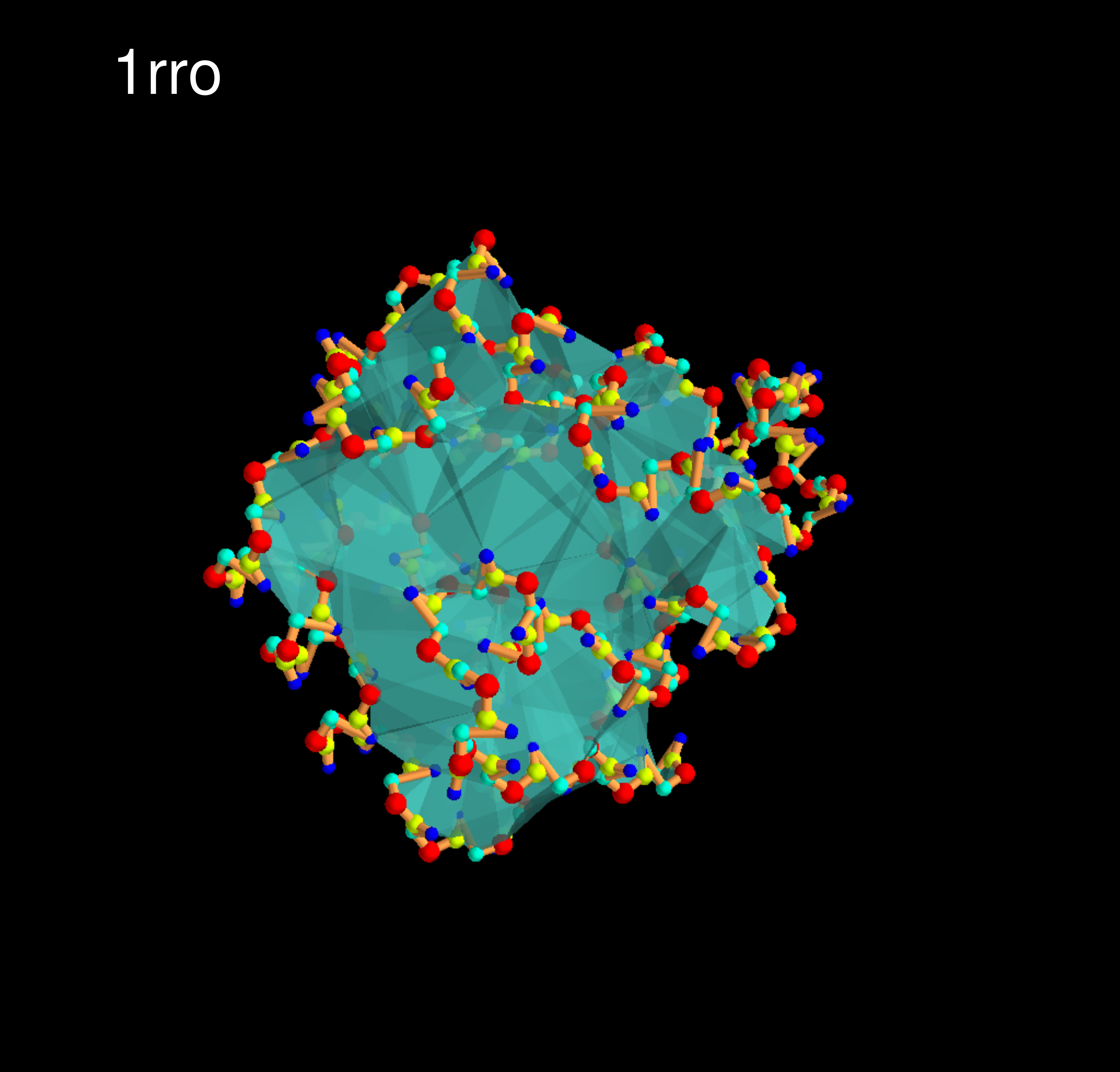}
        \captionsetup{labelformat=empty}
        \caption{}
        \label{supp_fig:pdb_1rro}
    \end{subfigure}

      \caption{}
      \label{fig:pdb_2}
    \end{figure}

    \begin{figure}[!tbhp]
      \centering
      \begin{subfigure}{.48\textwidth} \centering
        \includegraphics[width=0.9\linewidth]{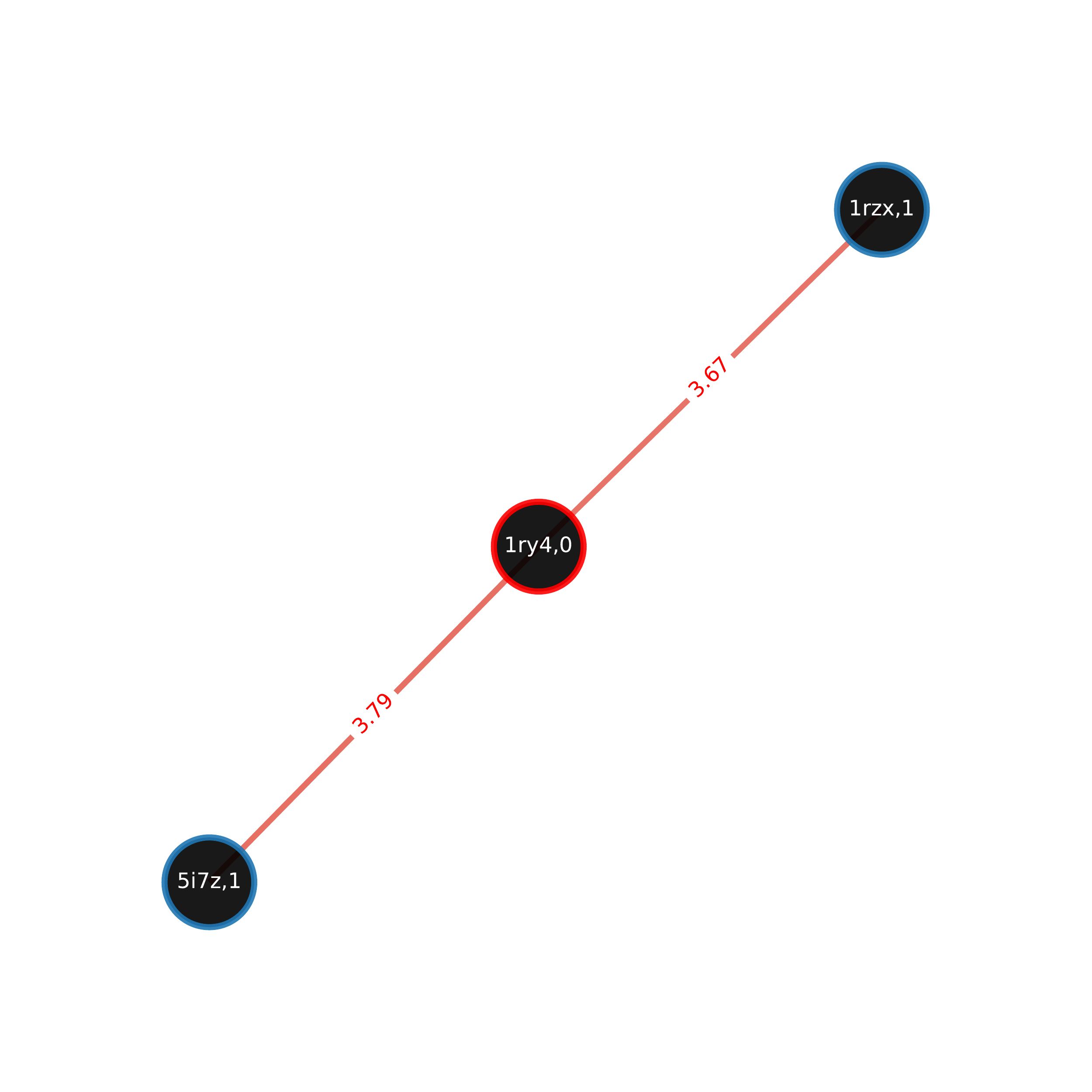}
        \captionsetup{labelformat=empty}
        \caption{}
        \label{supp_fig:pdb_hom_graph_3}
    \end{subfigure}
      \centering
      \begin{subfigure}{.48\textwidth} \centering
        \includegraphics[width=0.9\linewidth]{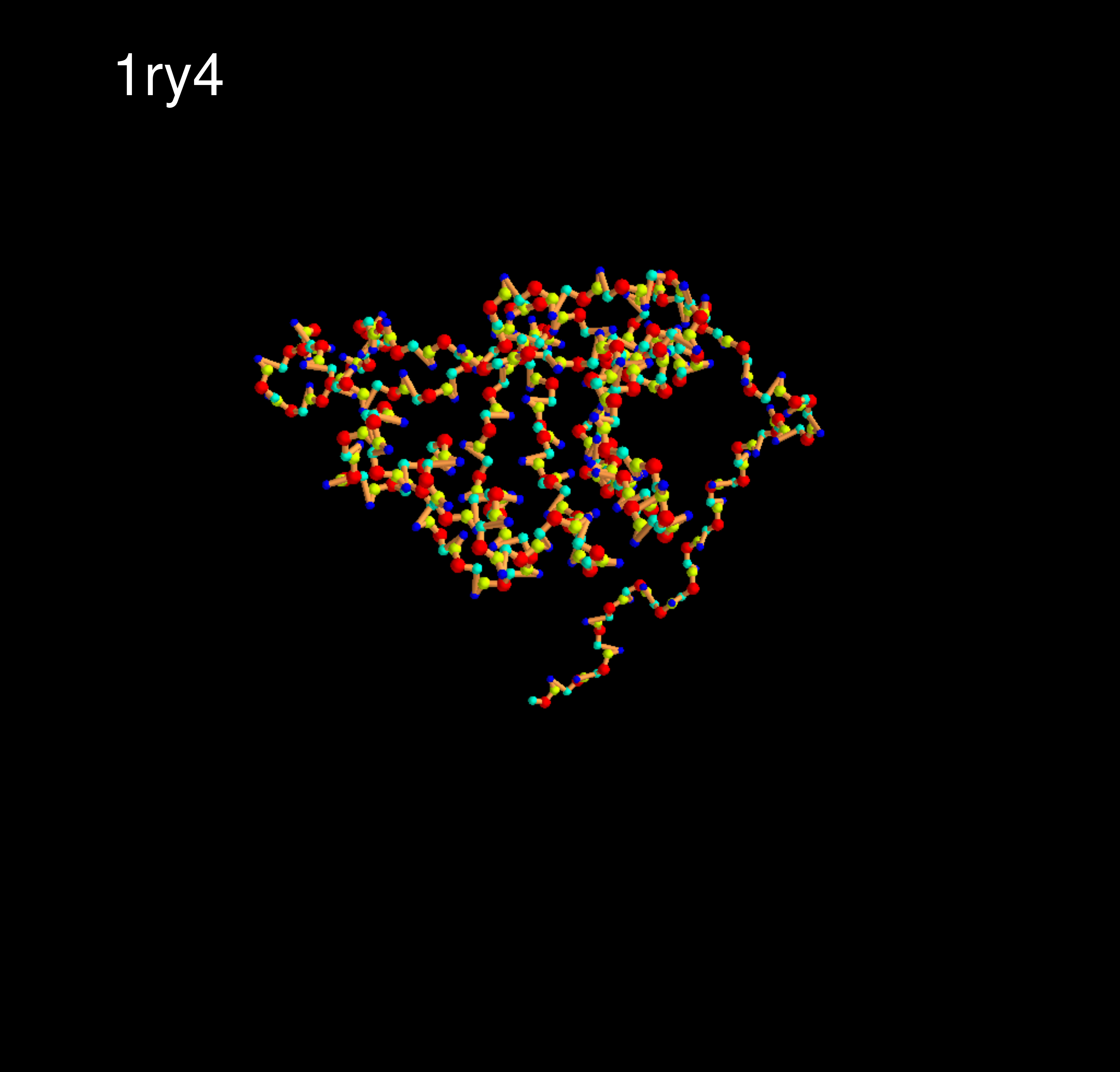}
        \captionsetup{labelformat=empty}
        \caption{}
        \label{supp_fig:pdb_1ry4}
    \end{subfigure}
      \centering
      \begin{subfigure}{.48\textwidth} \centering
        \includegraphics[width=0.9\linewidth]{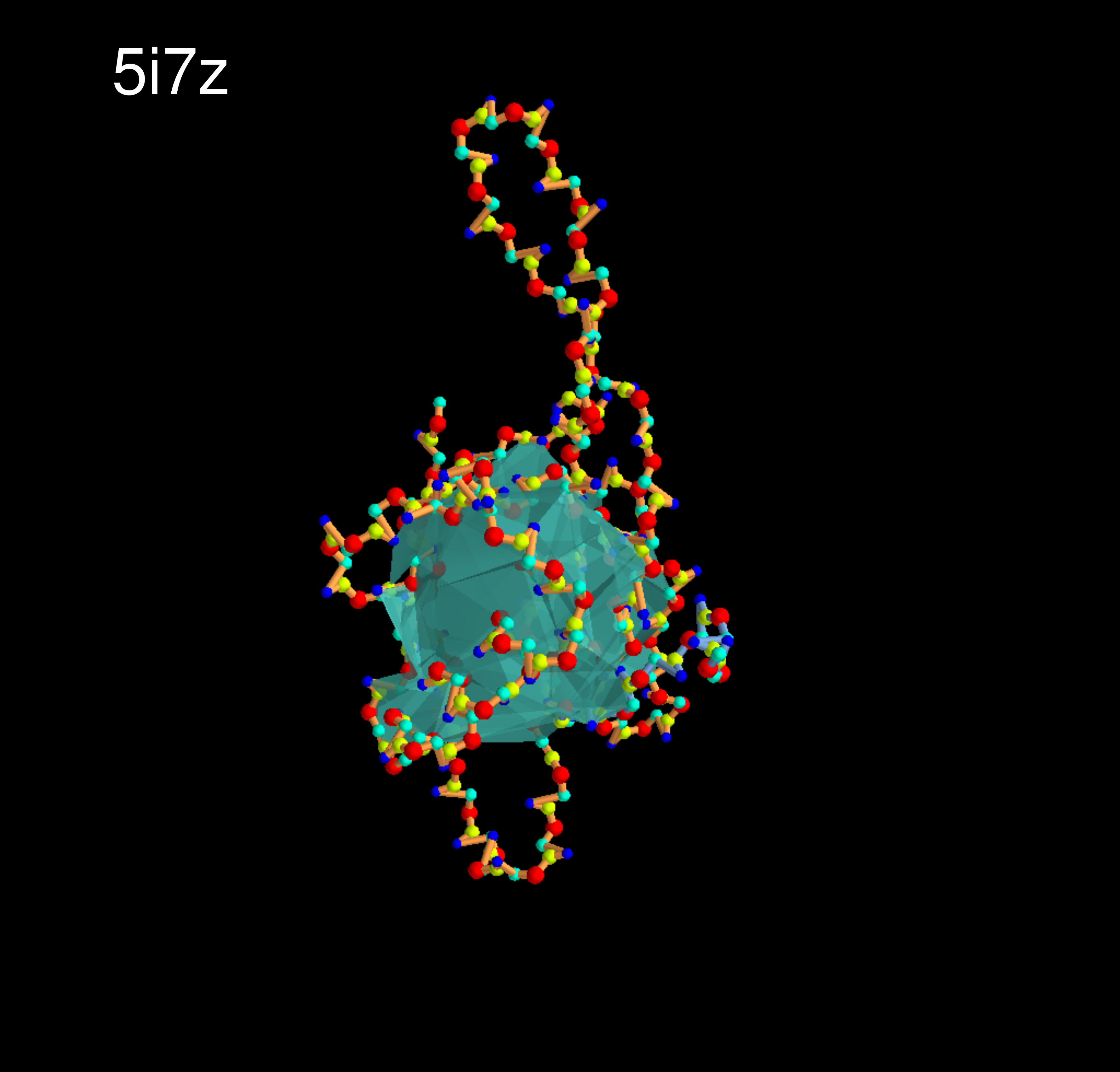}
        \captionsetup{labelformat=empty}
        \caption{}
        \label{supp_fig:pdb_5i7z}
    \end{subfigure}
      \centering
      \begin{subfigure}{.48\textwidth} \centering
        \includegraphics[width=0.9\linewidth]{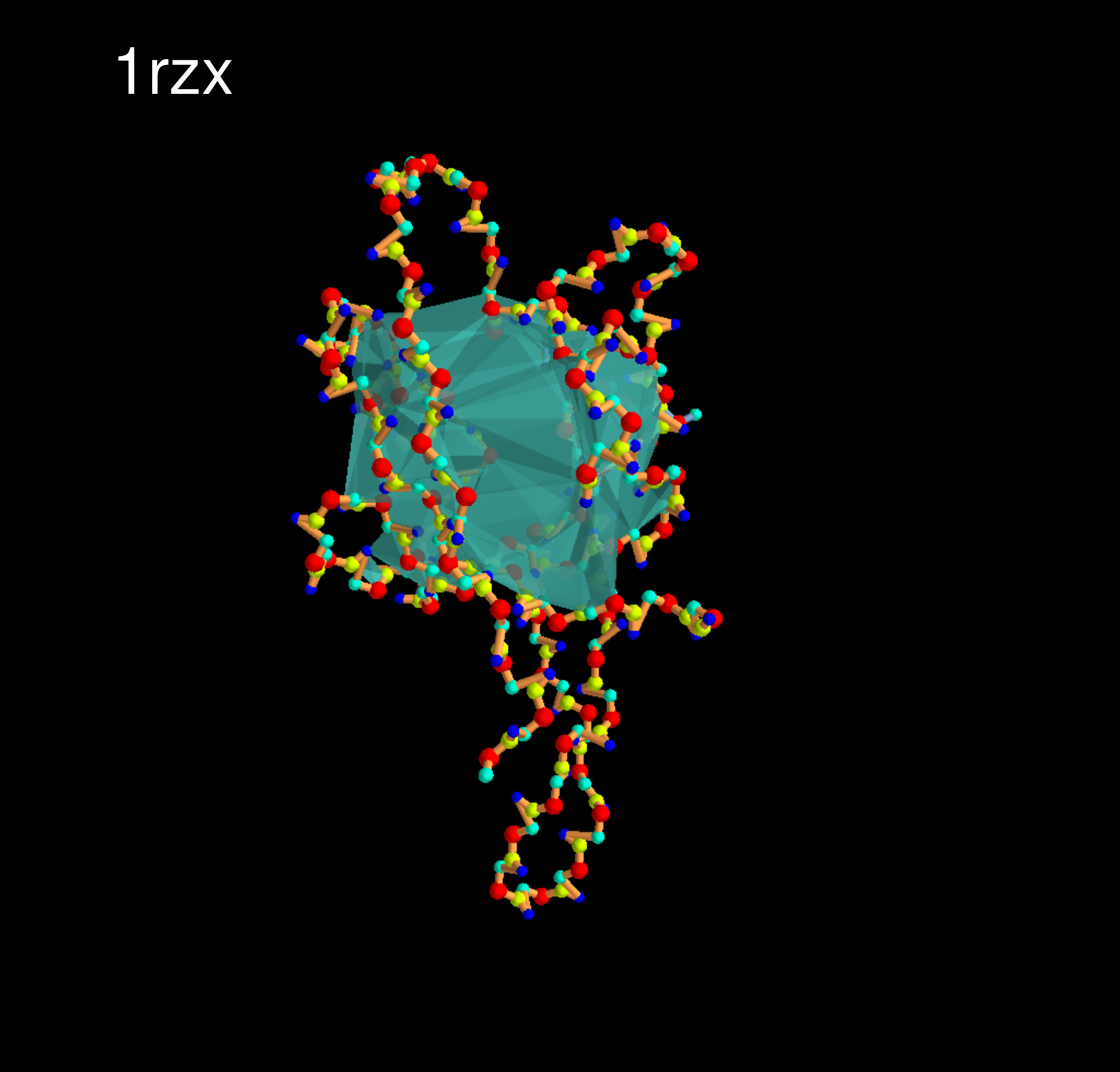}
        \captionsetup{labelformat=empty}
        \caption{}
        \label{supp_fig:pdb_1rzx}
    \end{subfigure}
      \caption{}
      \label{fig:pdb_3}
    \end{figure}

    \begin{figure}[!tbhp]
      \centering
      \begin{subfigure}{.48\textwidth} \centering
        \includegraphics[width=0.9\linewidth]{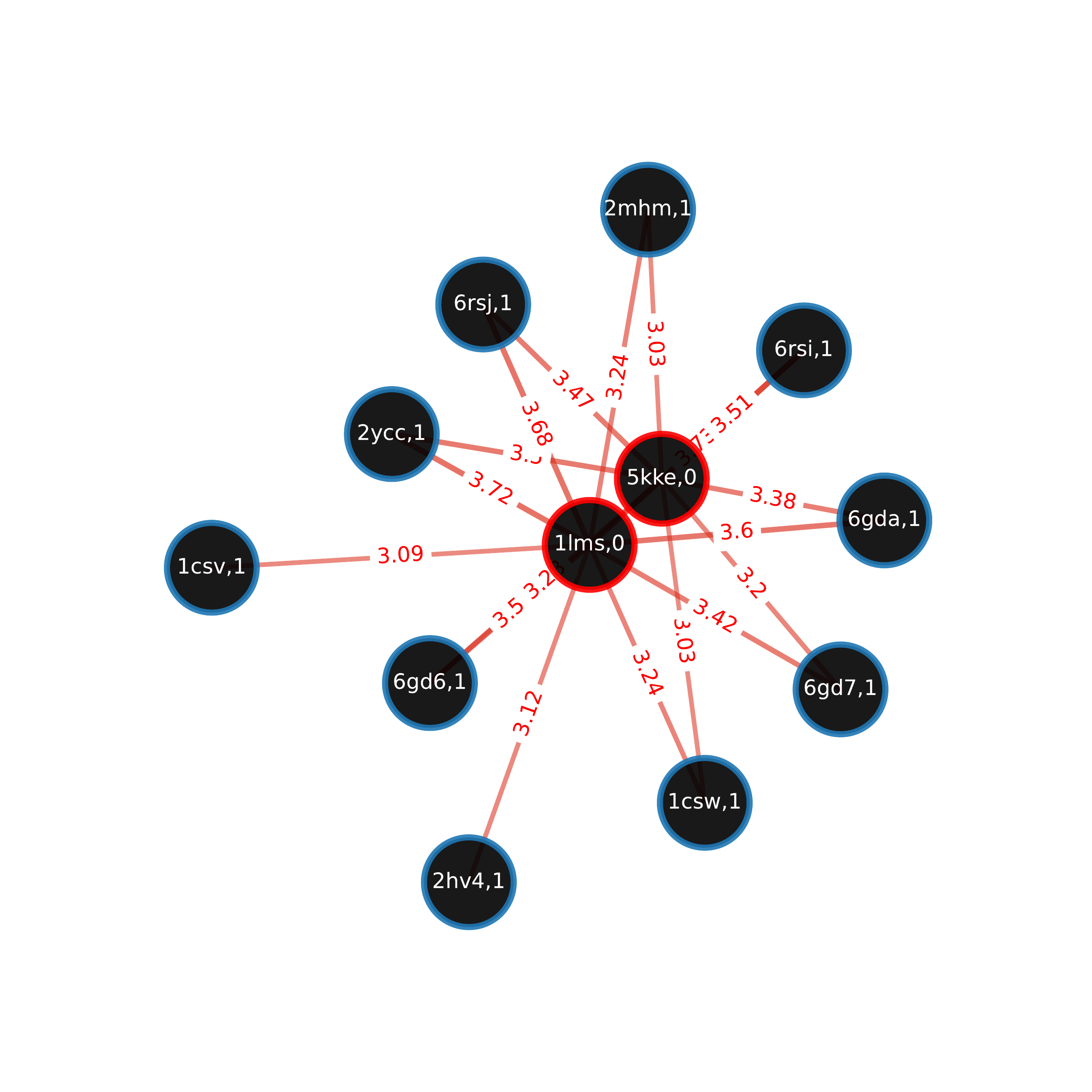}
        \captionsetup{labelformat=empty}
        \caption{}
        \label{supp_fig:pdb_hom_graph_4}
    \end{subfigure}
      \centering
      \begin{subfigure}{.24\textwidth} \centering
        \includegraphics[width=0.9\linewidth]{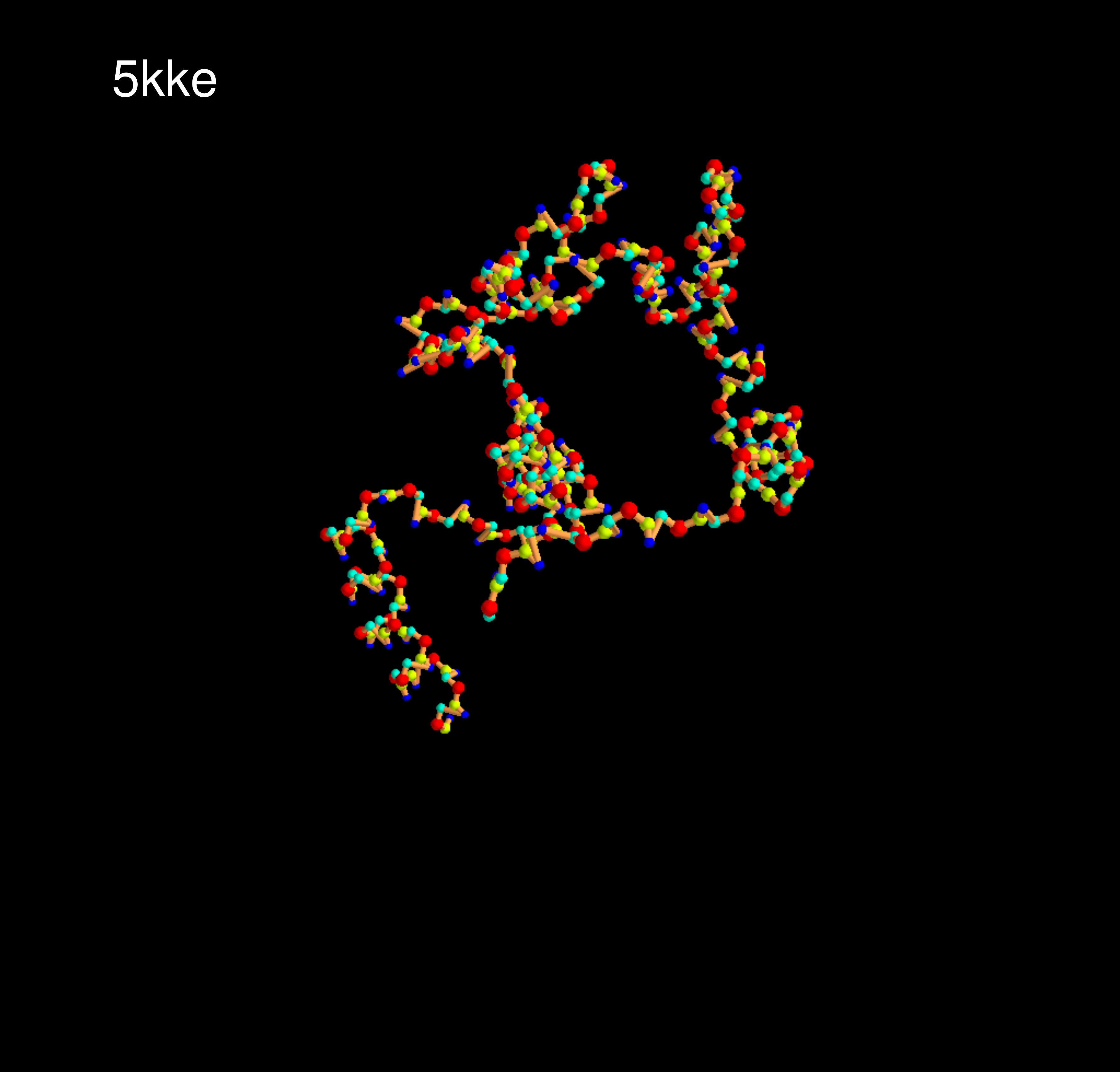}
        \captionsetup{labelformat=empty}
        \caption{}
        \label{supp_fig:pdb_5kke}
    \end{subfigure}
      \centering
      \begin{subfigure}{.24\textwidth} \centering
        \includegraphics[width=0.9\linewidth]{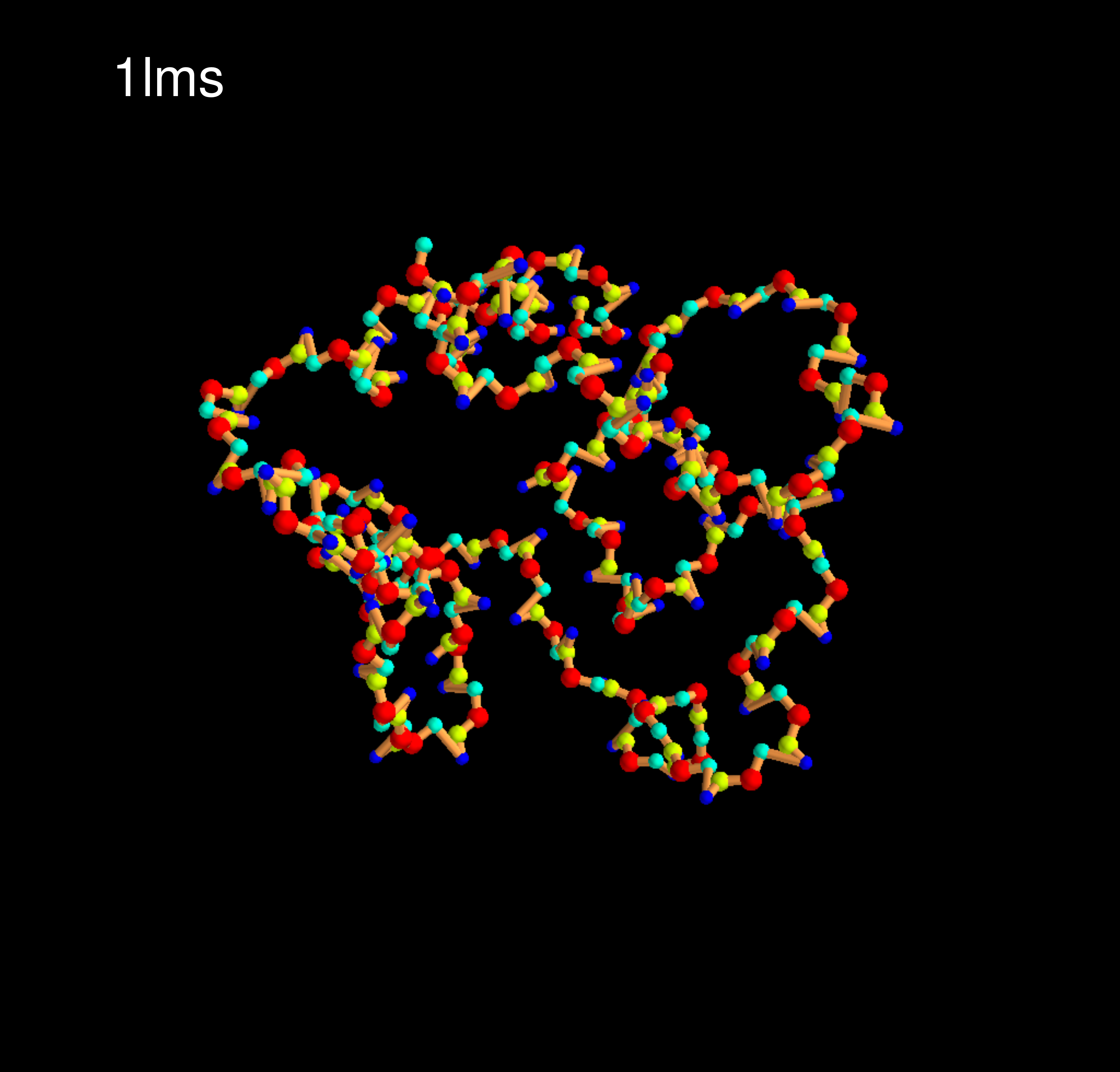}
        \captionsetup{labelformat=empty}
        \caption{}
        \label{supp_fig:pdb_1lms}
    \end{subfigure}
      \centering
      \begin{subfigure}{.24\textwidth} \centering
        \includegraphics[width=0.9\linewidth]{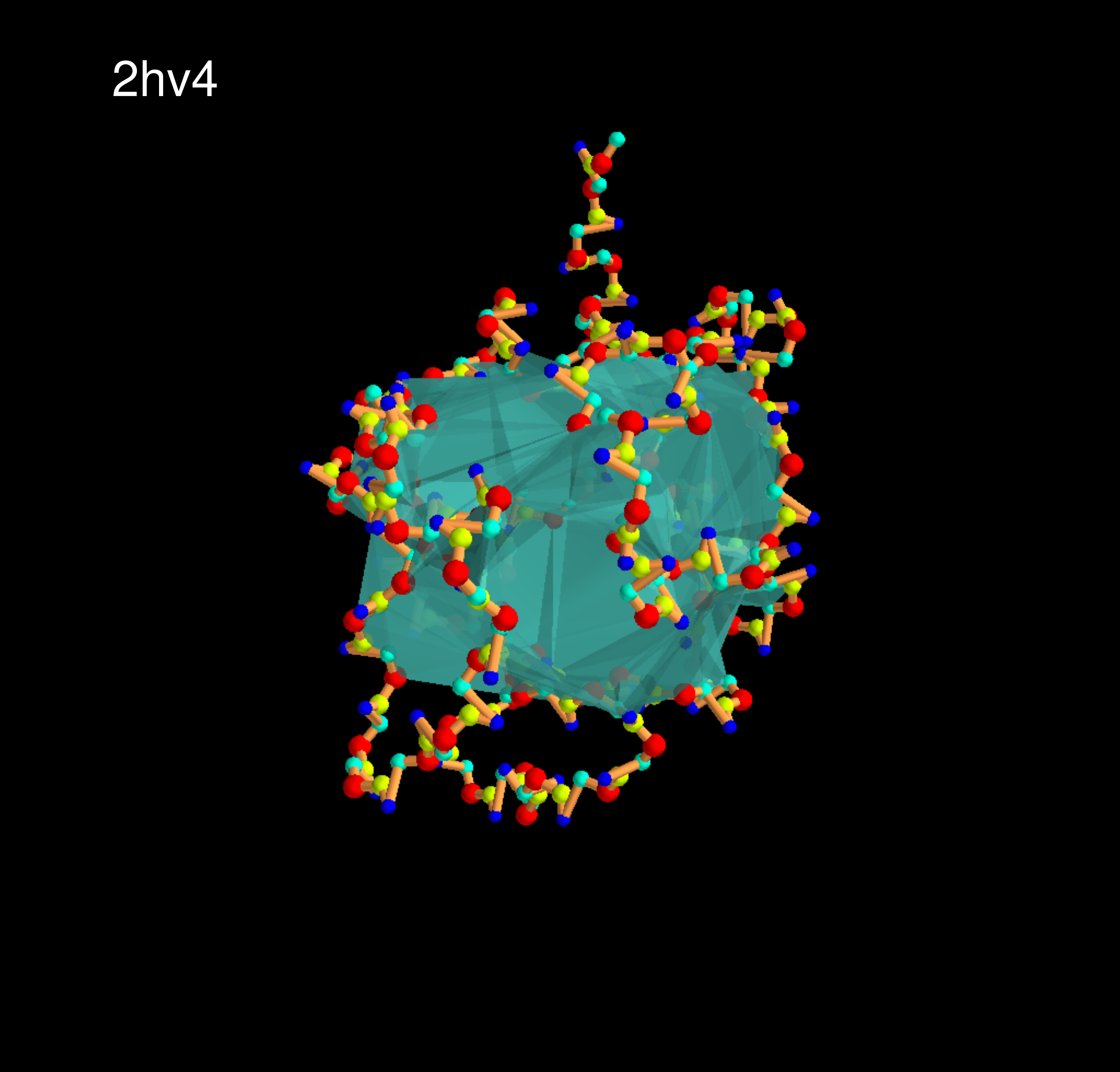}
        \captionsetup{labelformat=empty}
        \caption{}
        \label{supp_fig:pdb_2hv4}
    \end{subfigure}
      \centering
      \begin{subfigure}{.24\textwidth} \centering
        \includegraphics[width=0.9\linewidth]{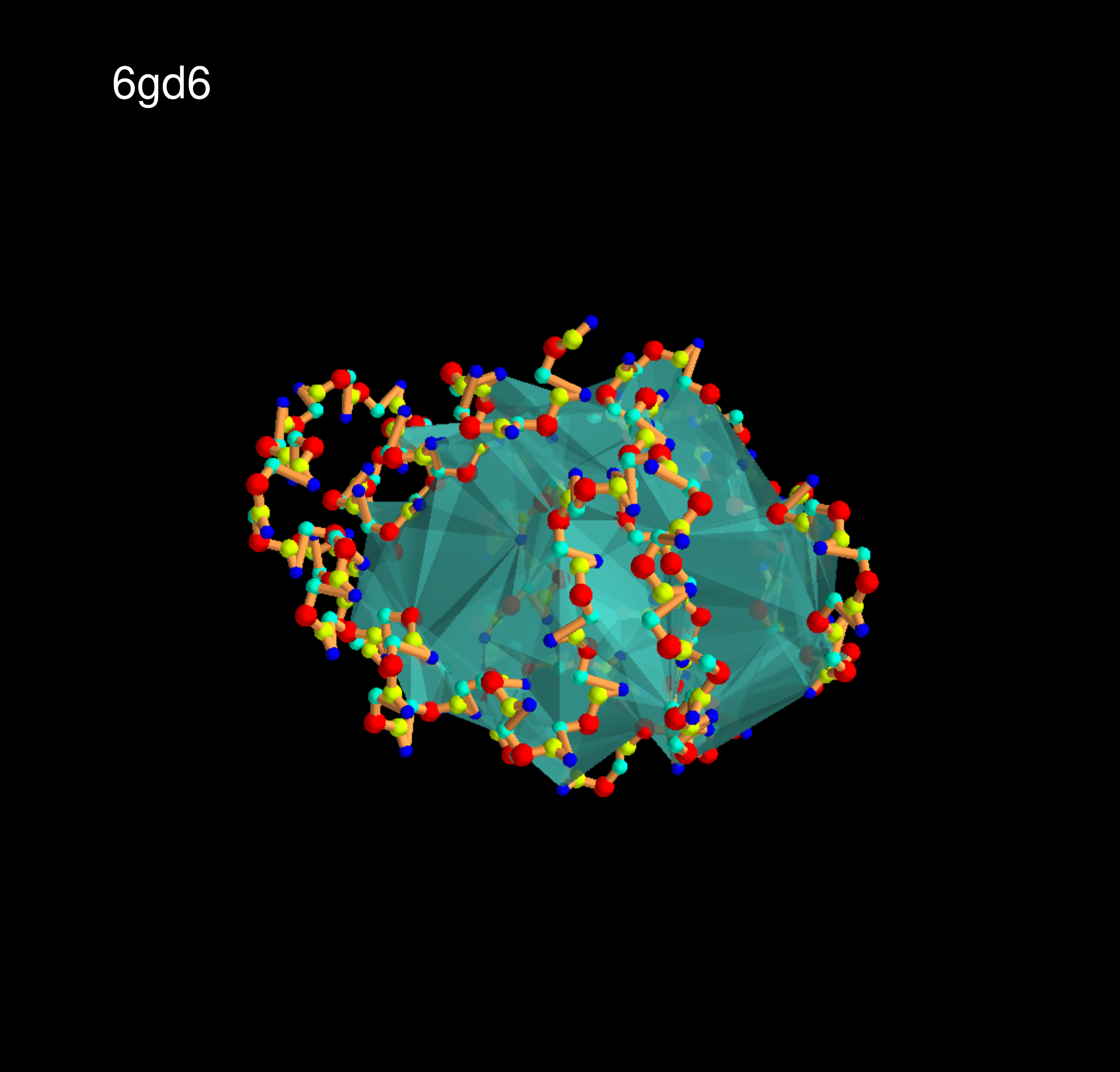}
        \captionsetup{labelformat=empty}
        \caption{}
        \label{supp_fig:pdb_6gd6}
    \end{subfigure}
      \centering
      \begin{subfigure}{.24\textwidth} \centering
        \includegraphics[width=0.9\linewidth]{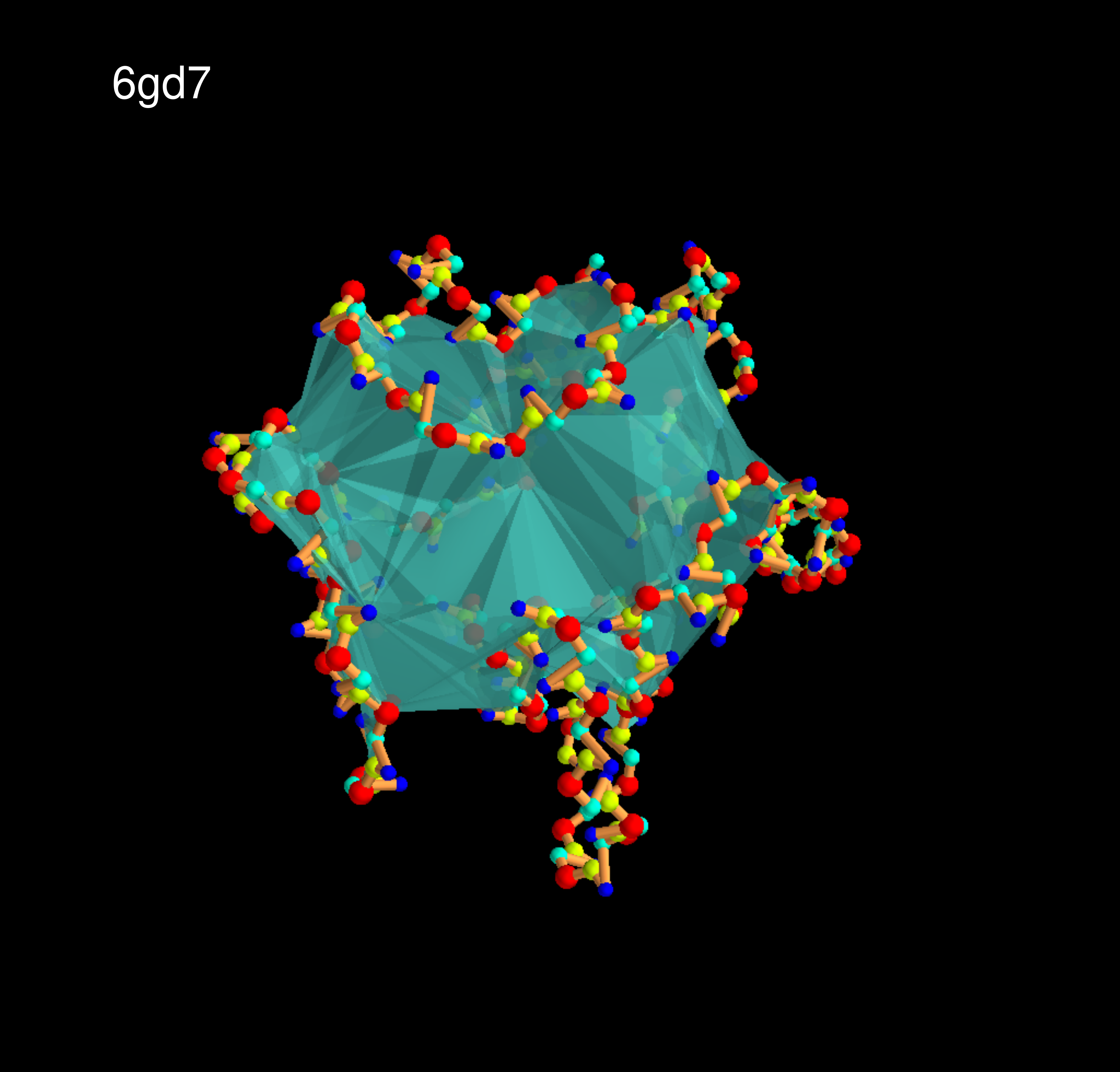}
        \captionsetup{labelformat=empty}
        \caption{}
        \label{supp_fig:pdb_6gd7}
    \end{subfigure}
      \centering
      \begin{subfigure}{.24\textwidth} \centering
        \includegraphics[width=0.9\linewidth]{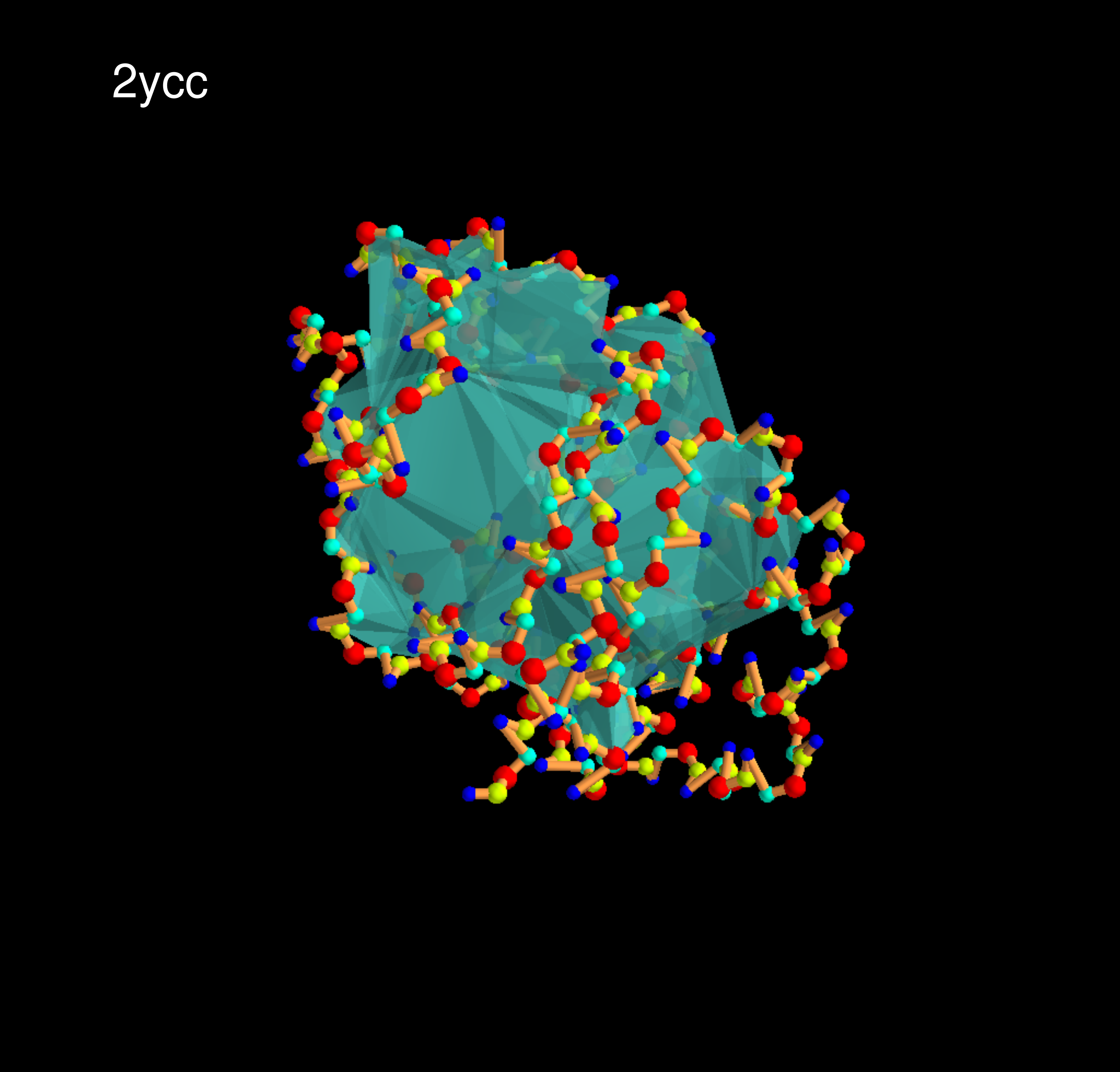}
        \captionsetup{labelformat=empty}
        \caption{}
        \label{supp_fig:pdb_2ycc}
    \end{subfigure}
      \centering
      \begin{subfigure}{.24\textwidth} \centering
        \includegraphics[width=0.9\linewidth]{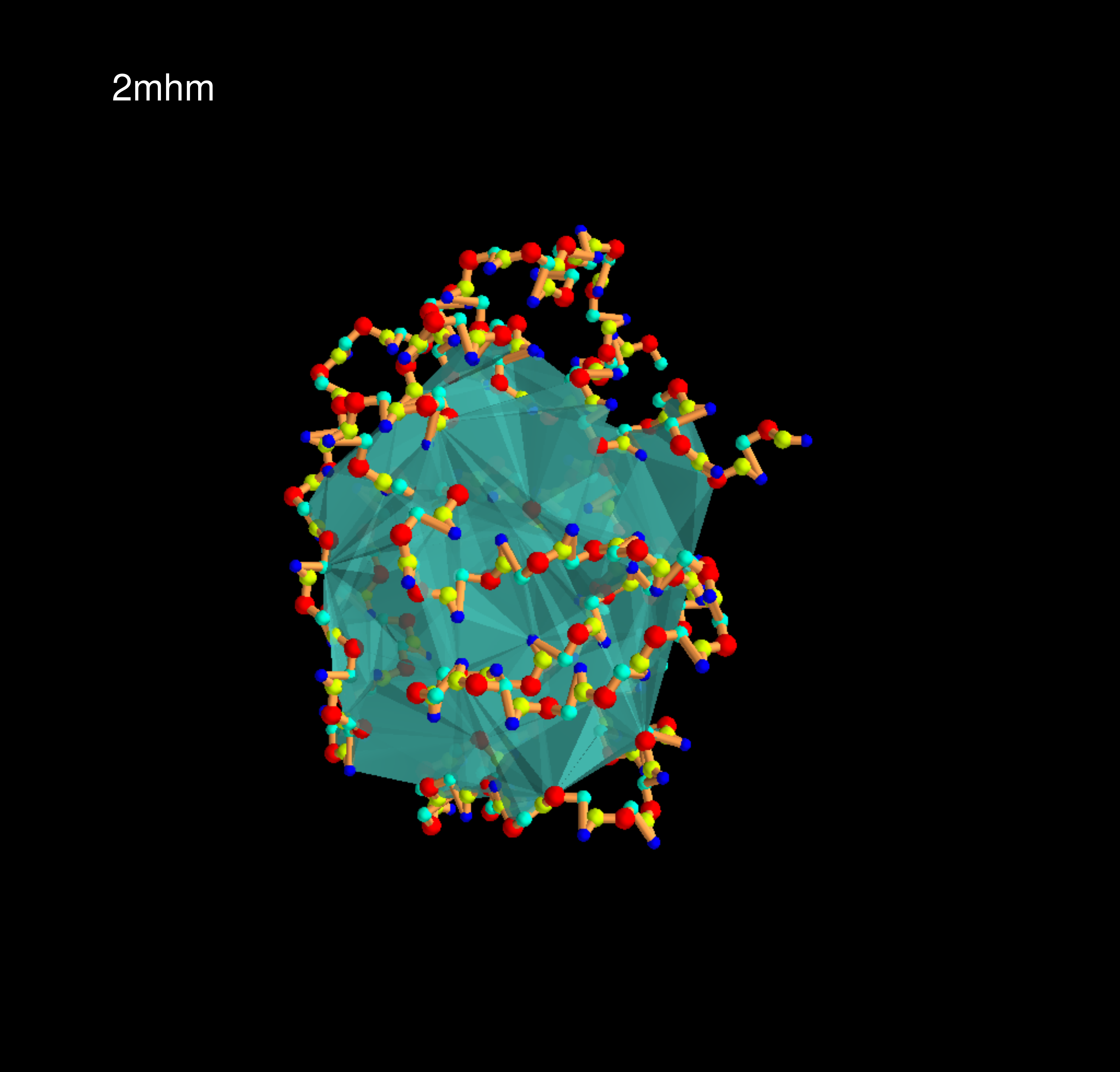}
        \captionsetup{labelformat=empty}
        \caption{}
        \label{supp_fig:pdb_2mhm}
    \end{subfigure}
      \centering
      \begin{subfigure}{.24\textwidth} \centering
        \includegraphics[width=0.9\linewidth]{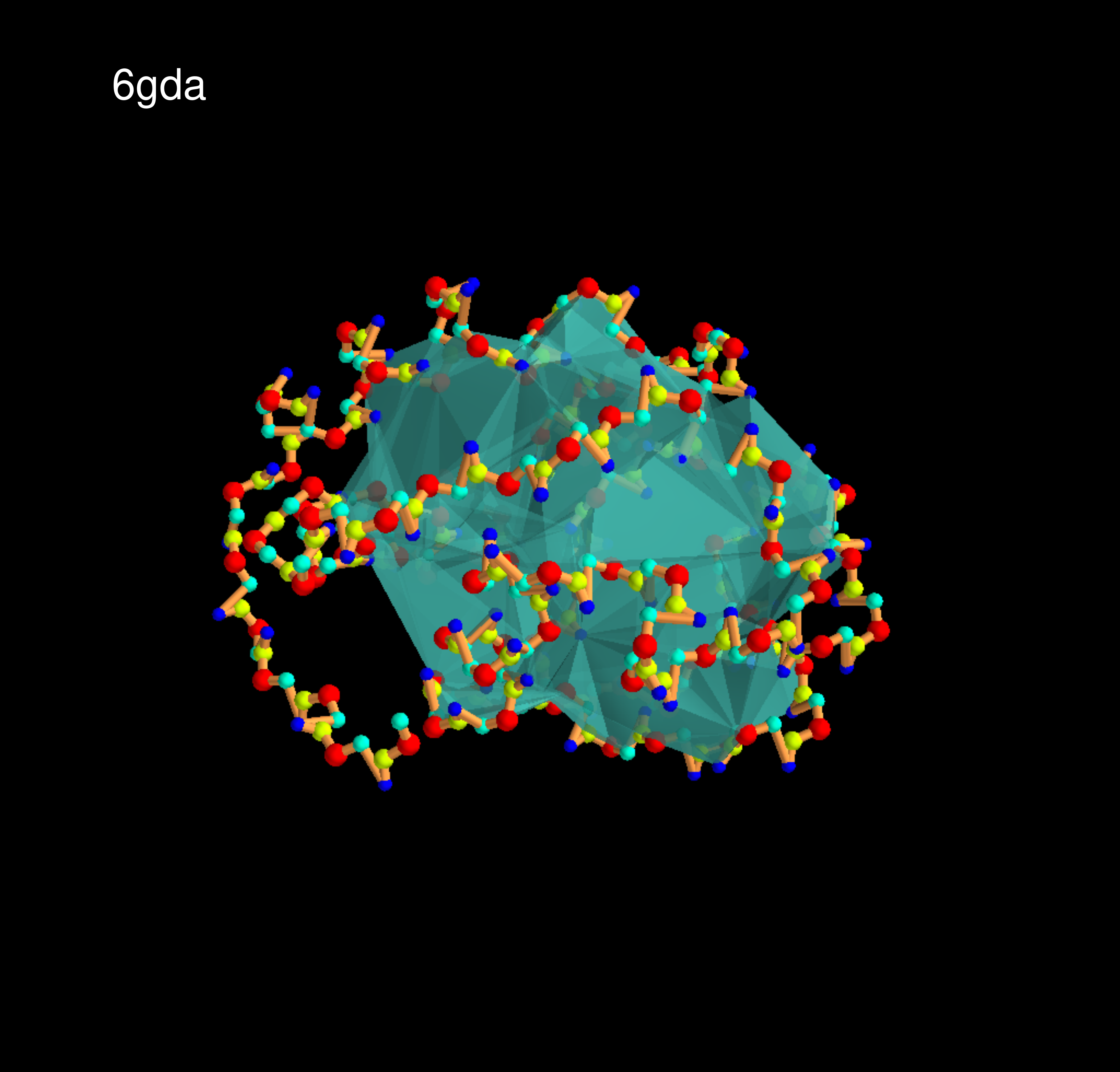}
        \captionsetup{labelformat=empty}
        \caption{}
        \label{supp_fig:pdb_6gda}
    \end{subfigure}
      \centering
      \begin{subfigure}{.24\textwidth} \centering
        \includegraphics[width=0.9\linewidth]{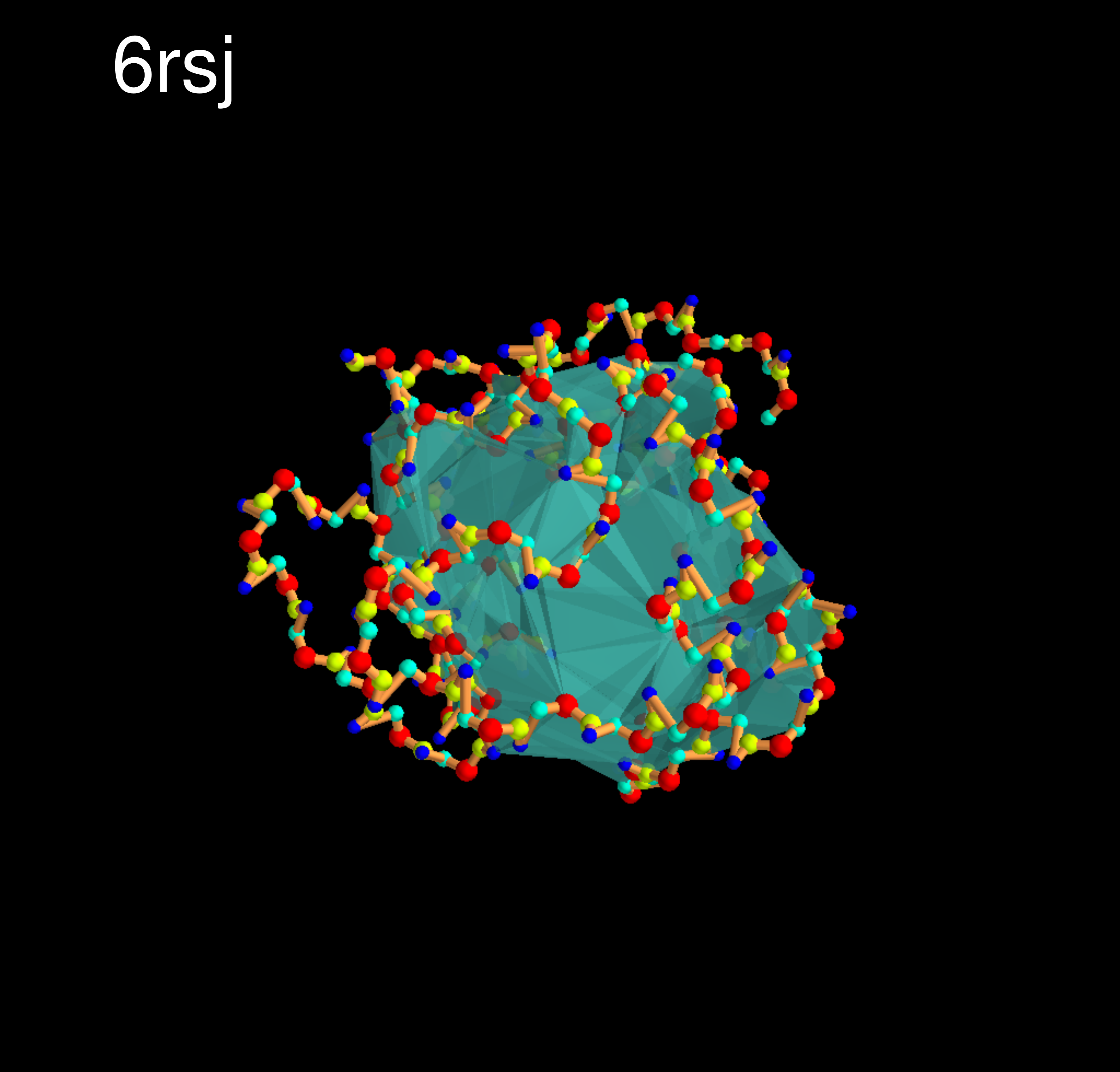}
        \captionsetup{labelformat=empty}
        \caption{}
        \label{supp_fig:pdb_6rsj}
    \end{subfigure}
      \centering
      \begin{subfigure}{.24\textwidth} \centering
        \includegraphics[width=0.9\linewidth]{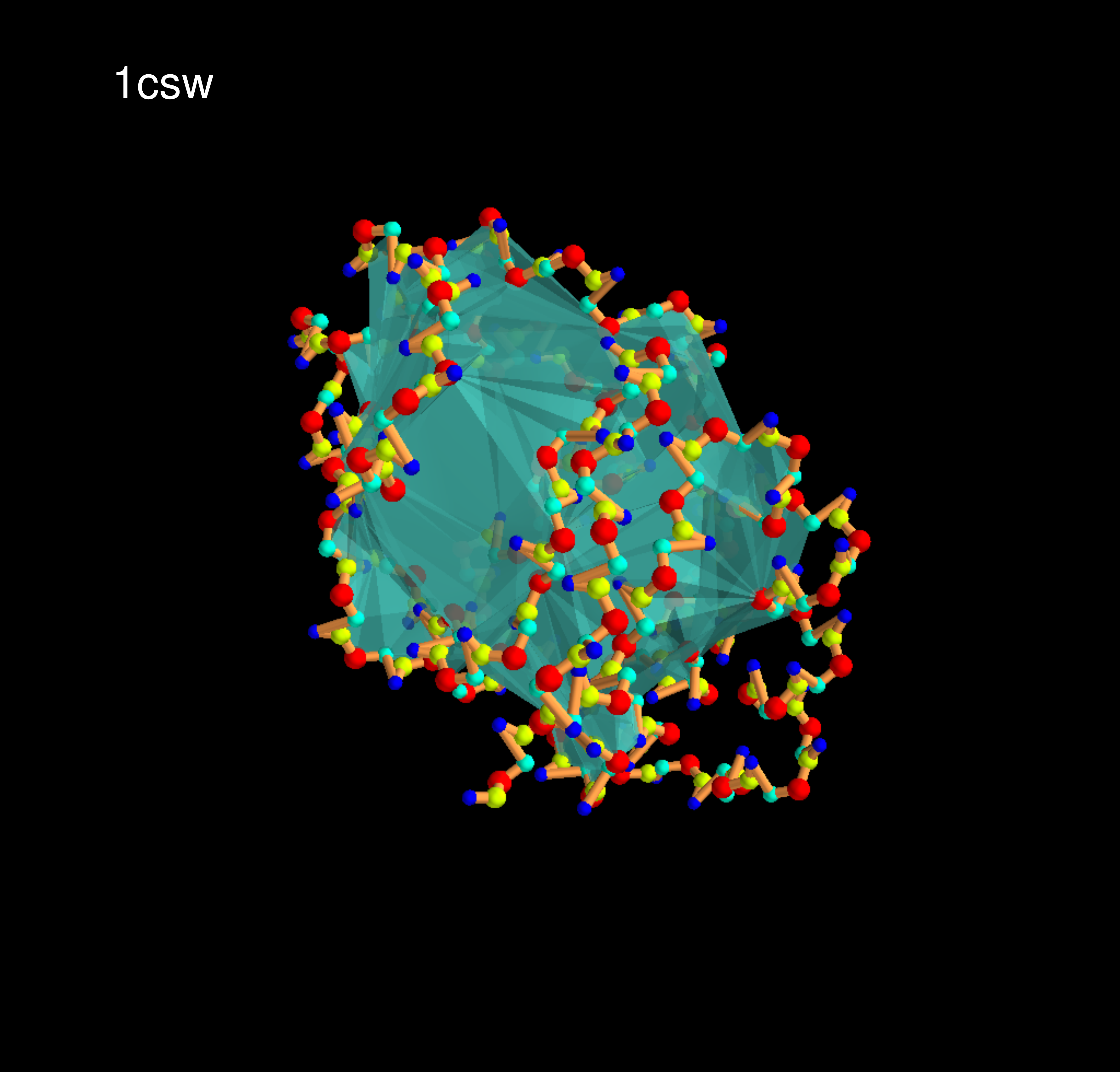}
        \captionsetup{labelformat=empty}
        \caption{}
        \label{supp_fig:pdb_1csw}
    \end{subfigure}
      \centering
      \begin{subfigure}{.24\textwidth} \centering
        \includegraphics[width=0.9\linewidth]{figures/pdb_2hv4.pdf}
        \captionsetup{labelformat=empty}
        \caption{}
        \label{supp_fig:pdb_2hv4}
    \end{subfigure}
      \centering
      \begin{subfigure}{.24\textwidth} \centering
        \includegraphics[width=0.9\linewidth]{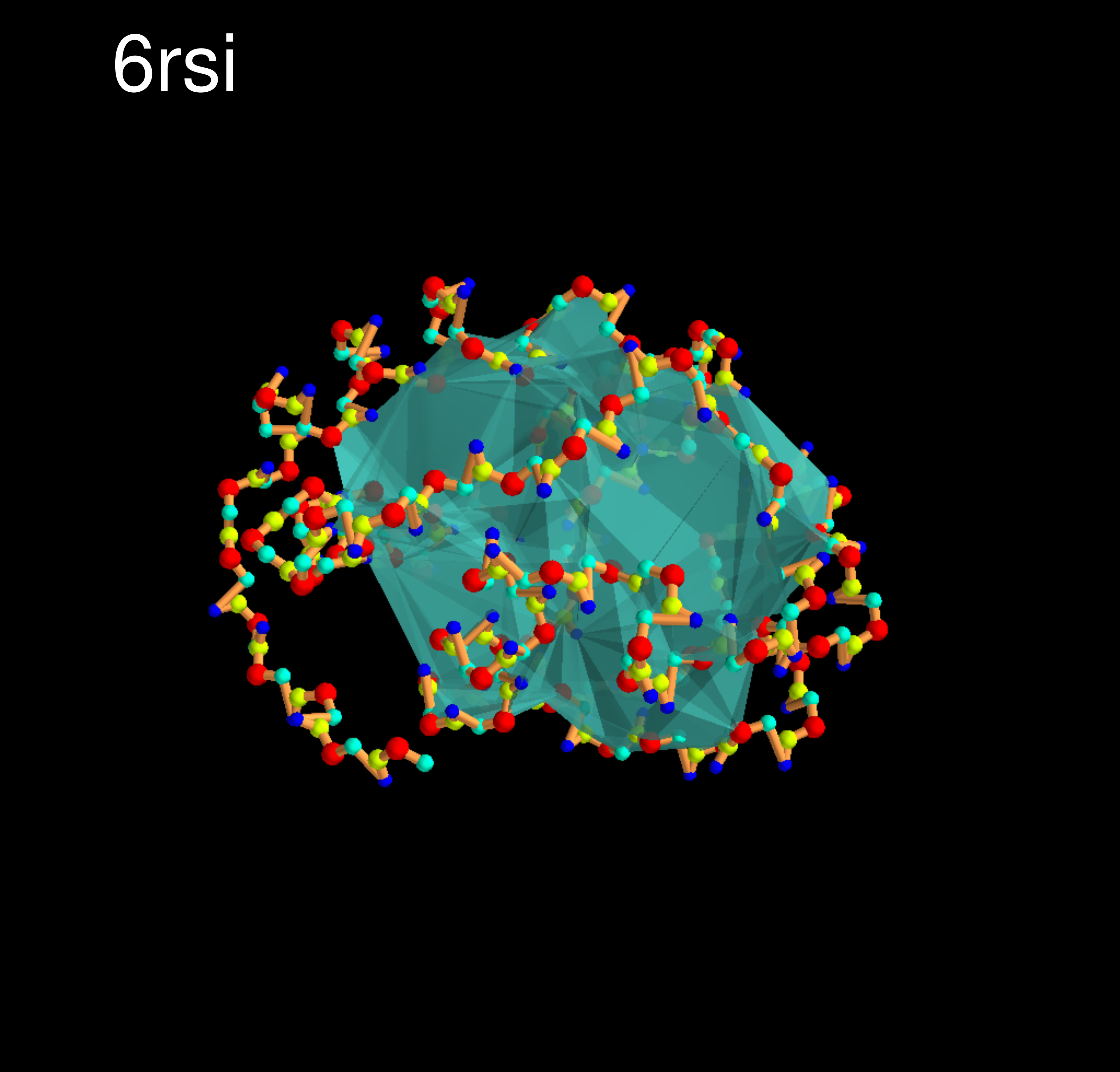}
        \captionsetup{labelformat=empty}
        \caption{}
        \label{supp_fig:pdb_6rsi}
    \end{subfigure}
      \centering
      \begin{subfigure}{.24\textwidth} \centering
        \includegraphics[width=0.9\linewidth]{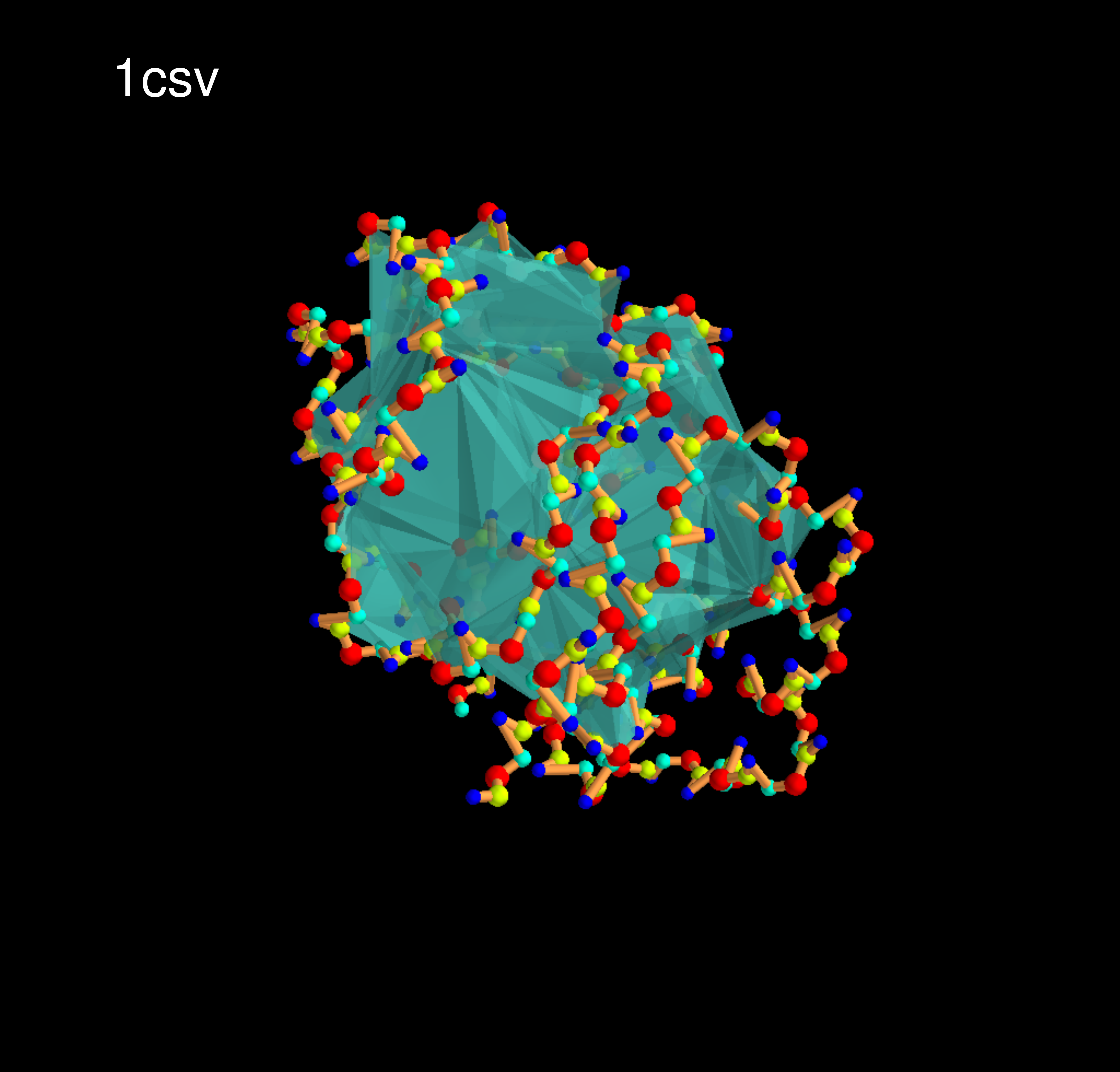}
        \captionsetup{labelformat=empty}
        \caption{}
        \label{supp_fig:pdb_1csv}
    \end{subfigure}

      \caption{}
      \label{fig:pdb_4}
    \end{figure}

    \begin{figure}[!tbhp]
      \centering
      \begin{subfigure}{.48\textwidth} \centering
        \includegraphics[width=0.9\linewidth]{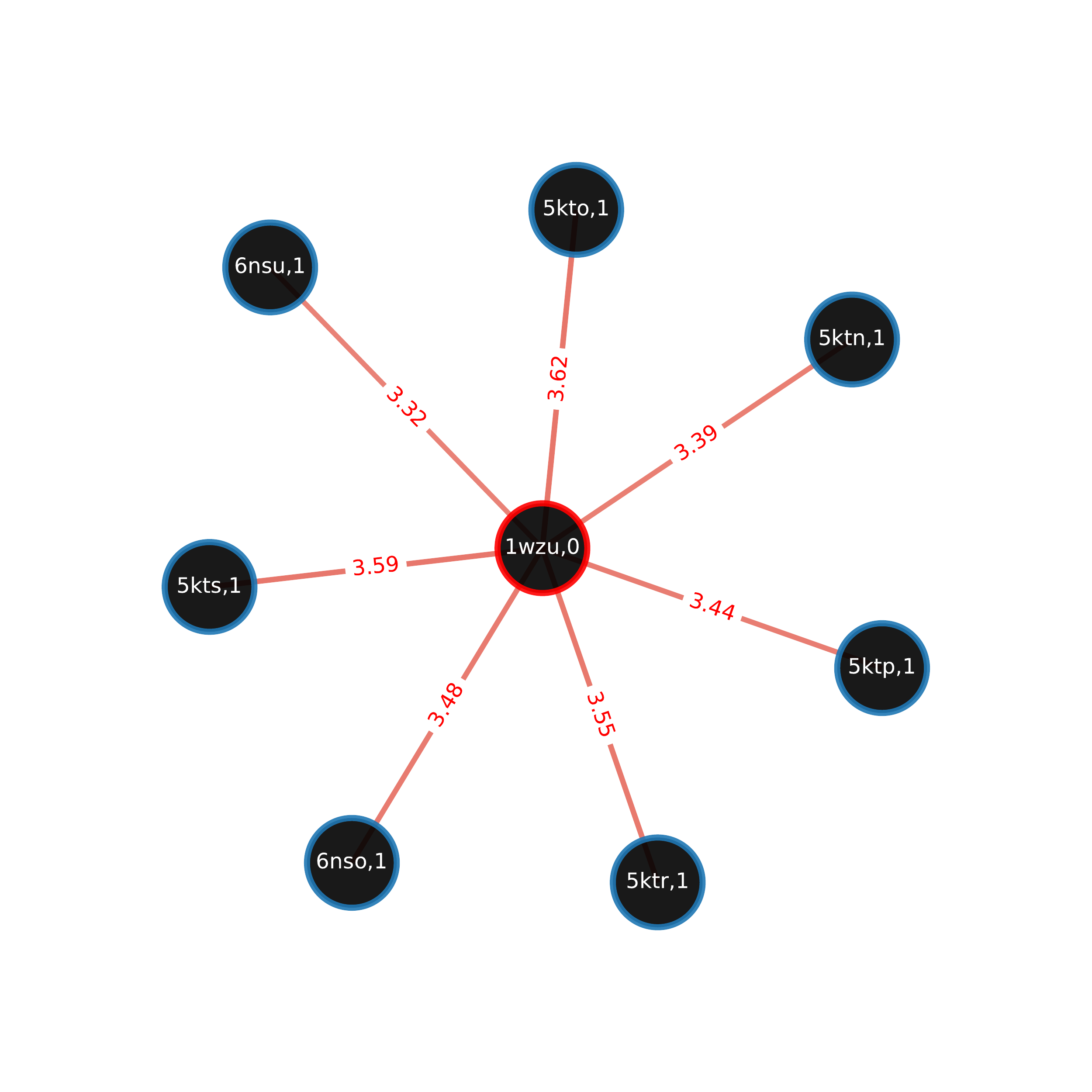}
        \captionsetup{labelformat=empty}
        \caption{}
        \label{supp_fig:pdb_hom_graph_5}
    \end{subfigure}
      \centering
      \begin{subfigure}{.24\textwidth} \centering
        \includegraphics[width=0.9\linewidth]{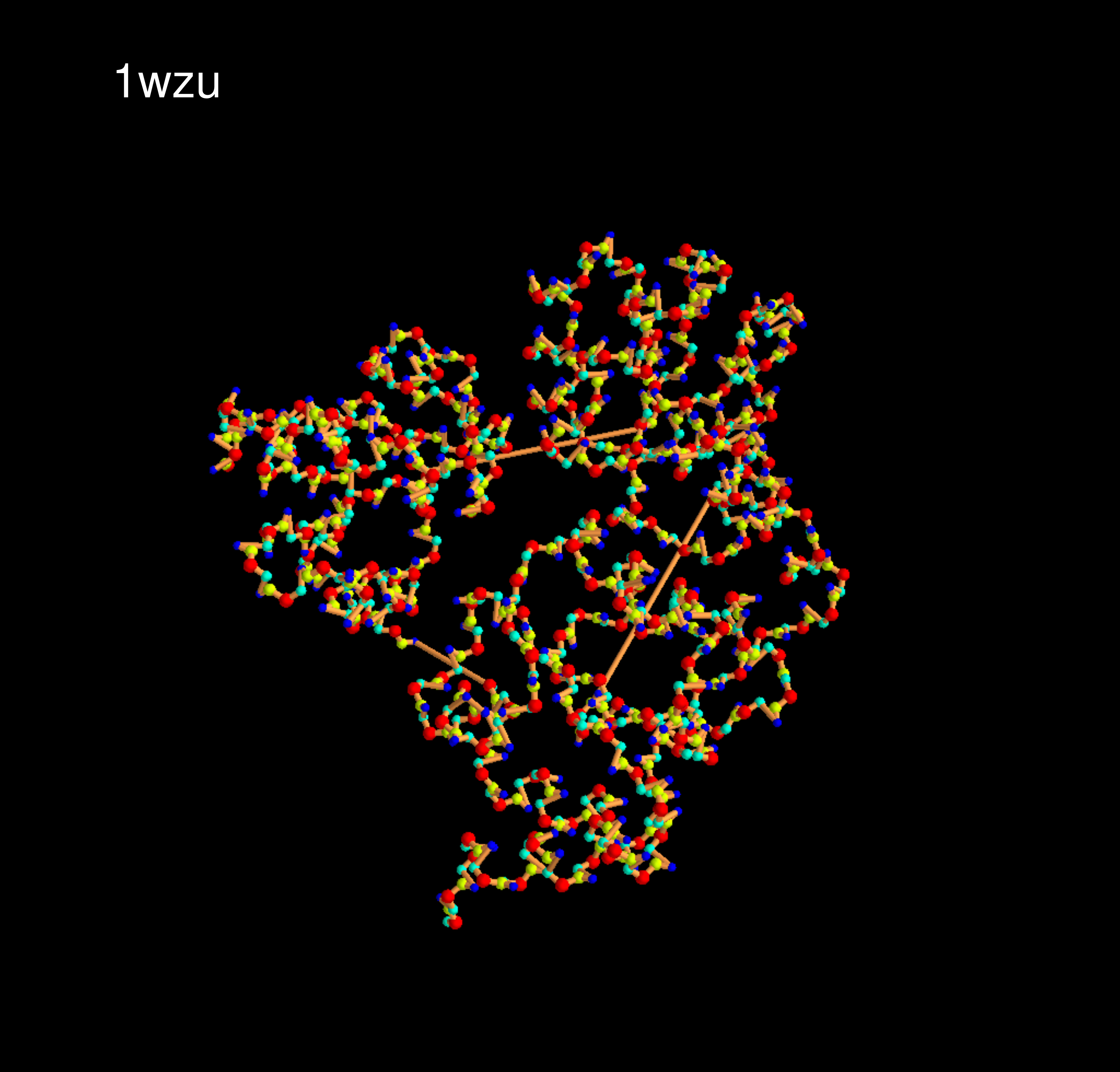}
        \captionsetup{labelformat=empty}
        \caption{}
        \label{supp_fig:pdb_1wzu}
    \end{subfigure}
      \centering
      \begin{subfigure}{.24\textwidth} \centering
        \includegraphics[width=0.9\linewidth]{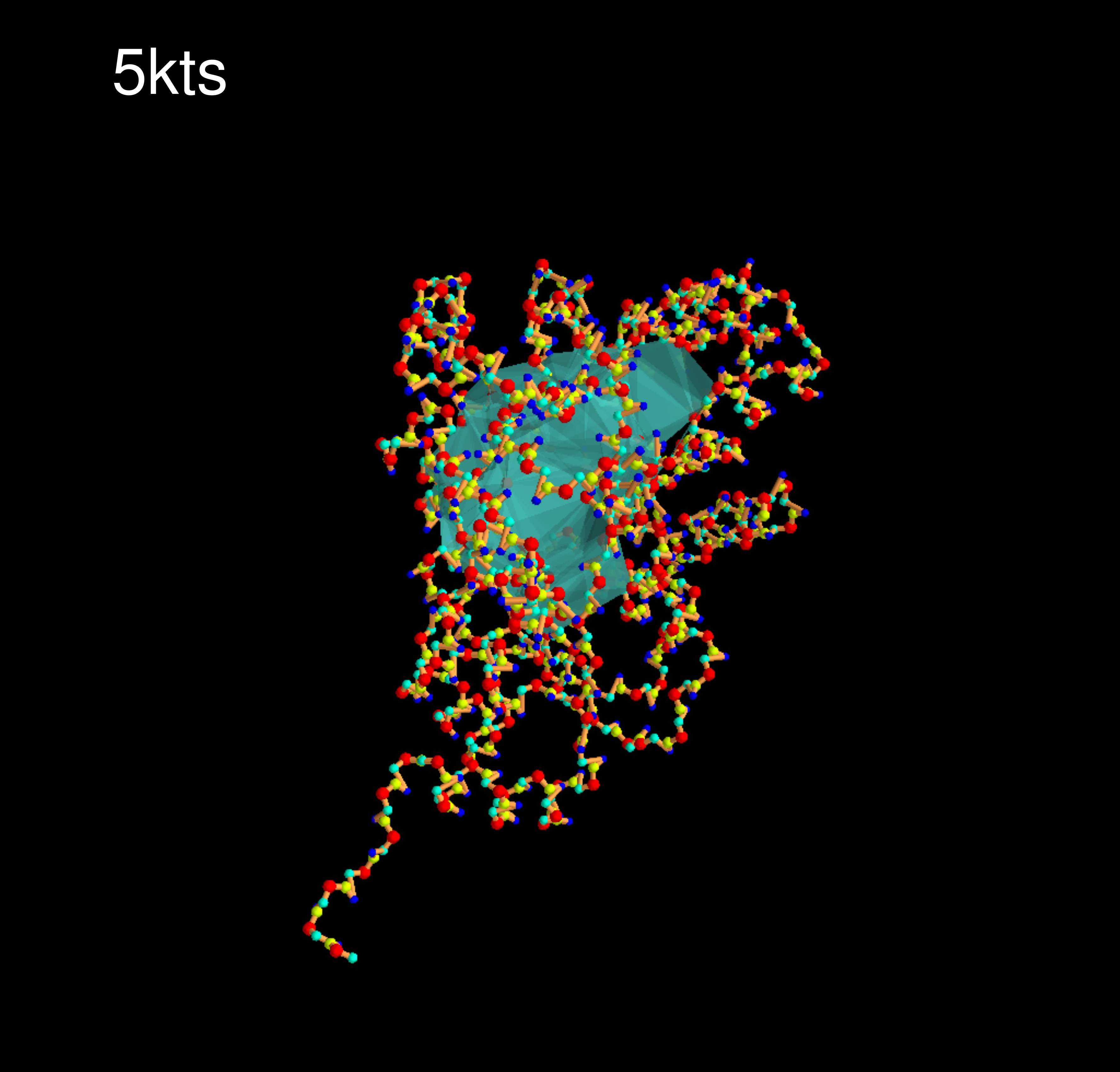}
        \captionsetup{labelformat=empty}
        \caption{}
        \label{supp_fig:pdb_5kts}
    \end{subfigure}
      \centering
      \begin{subfigure}{.24\textwidth} \centering
        \includegraphics[width=0.9\linewidth]{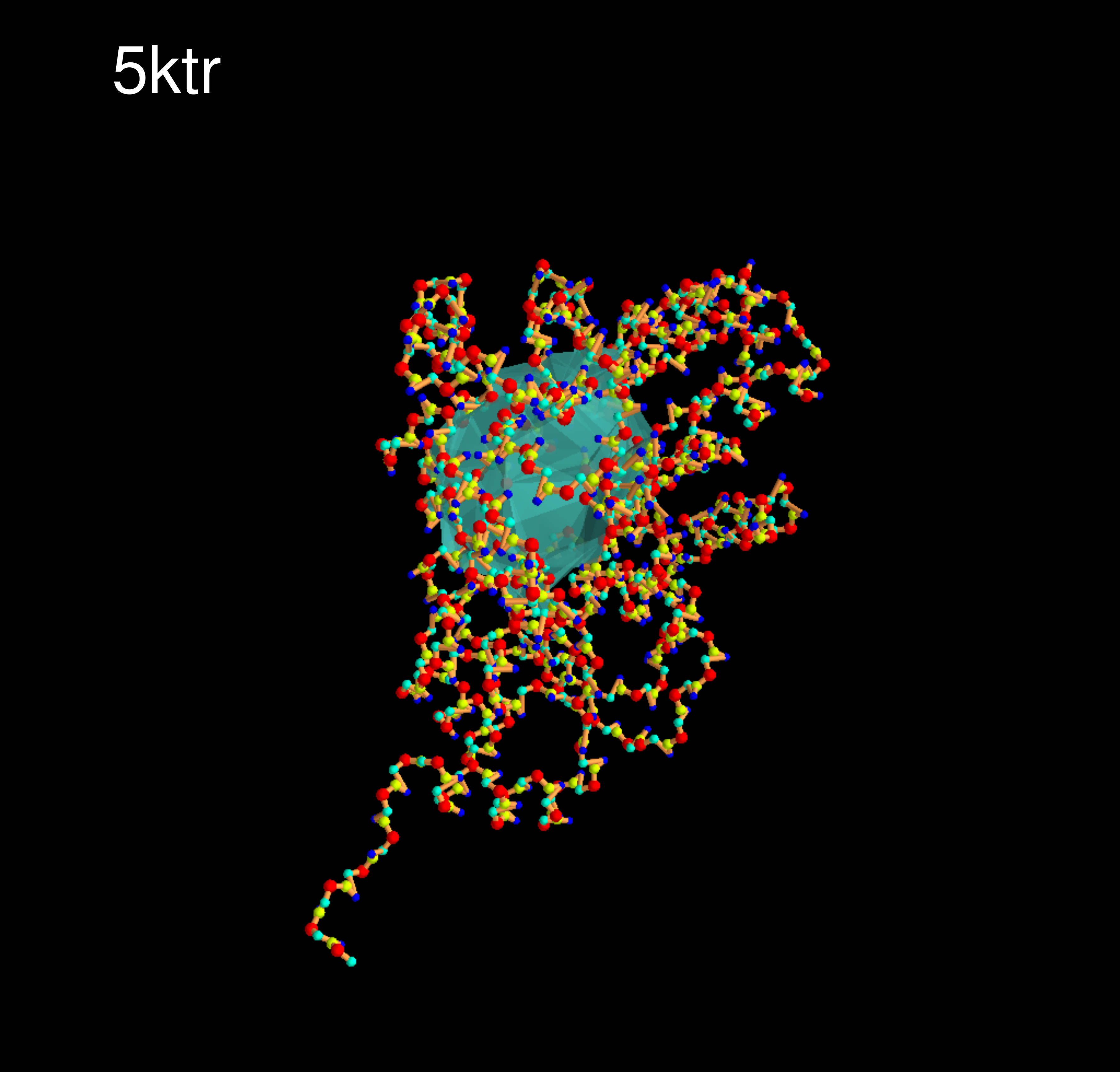}
        \captionsetup{labelformat=empty}
        \caption{}
        \label{supp_fig:pdb_5ktr}
    \end{subfigure}
      \centering
      \begin{subfigure}{.24\textwidth} \centering
        \includegraphics[width=0.9\linewidth]{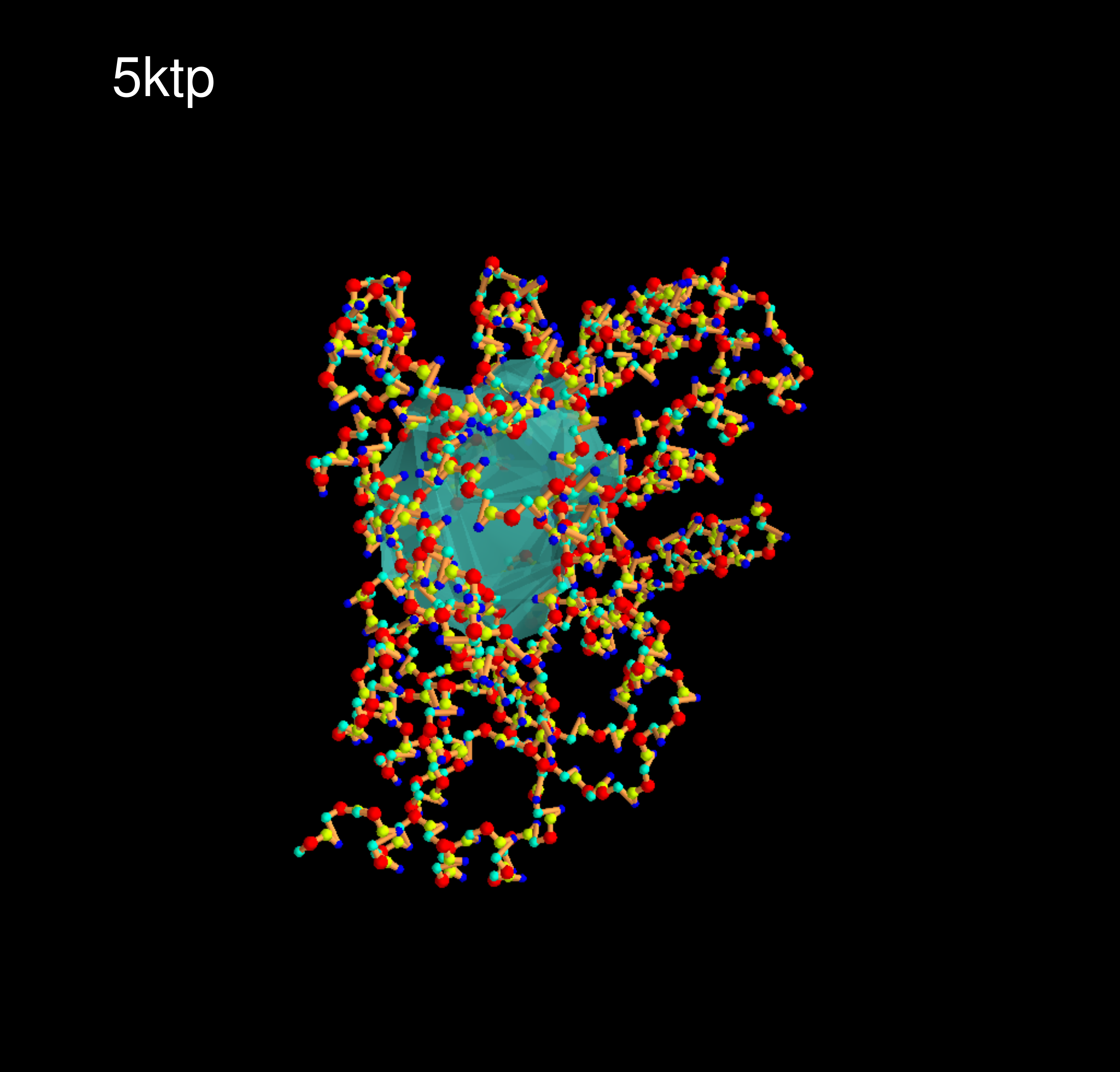}
        \captionsetup{labelformat=empty}
        \caption{}
        \label{supp_fig:pdb_5ktp}
    \end{subfigure}
      \centering
      \begin{subfigure}{.24\textwidth} \centering
        \includegraphics[width=0.9\linewidth]{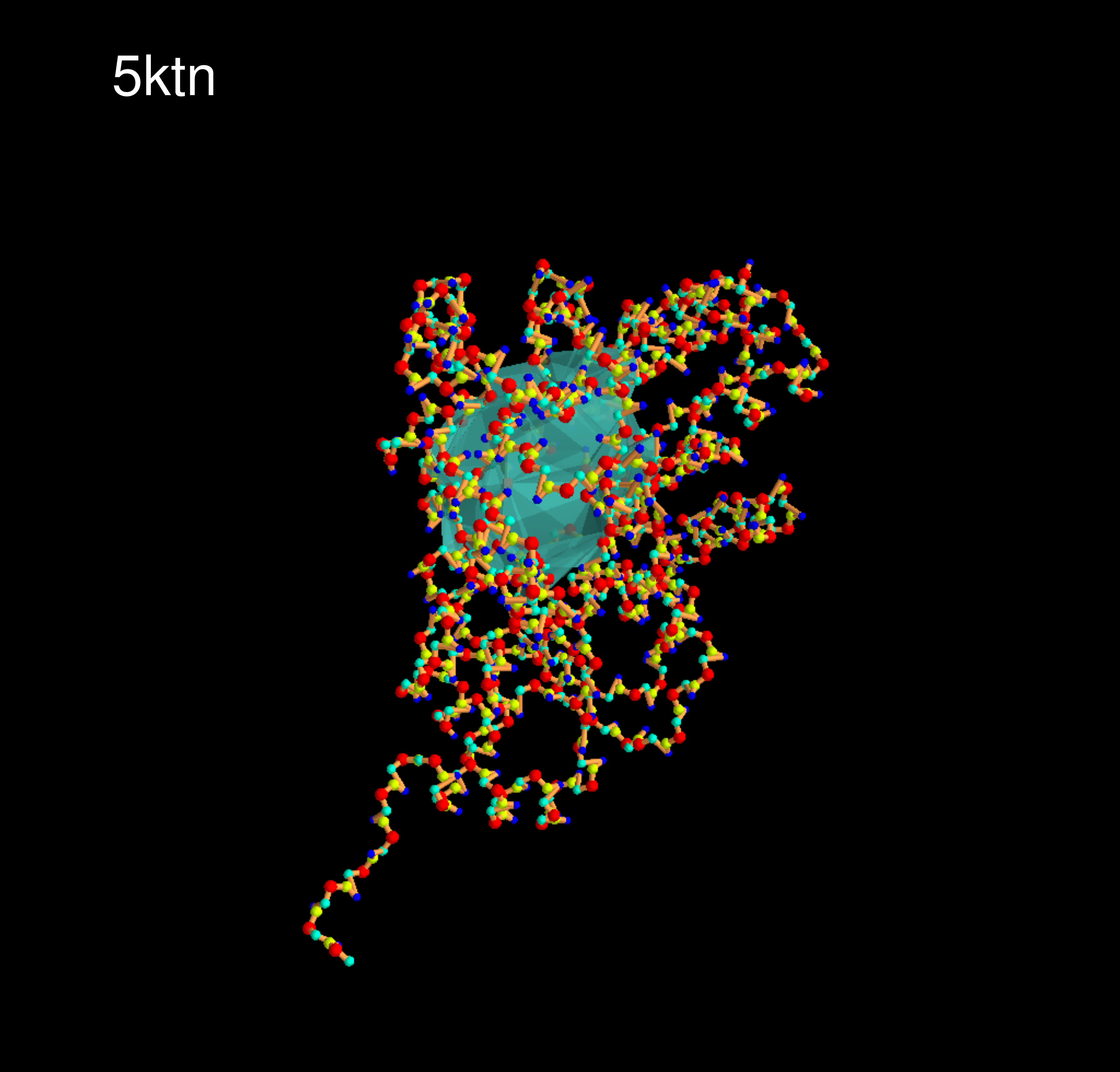}
        \captionsetup{labelformat=empty}
        \caption{}
        \label{supp_fig:pdb_5ktn}
    \end{subfigure}
      \centering
      \begin{subfigure}{.24\textwidth} \centering
        \includegraphics[width=0.9\linewidth]{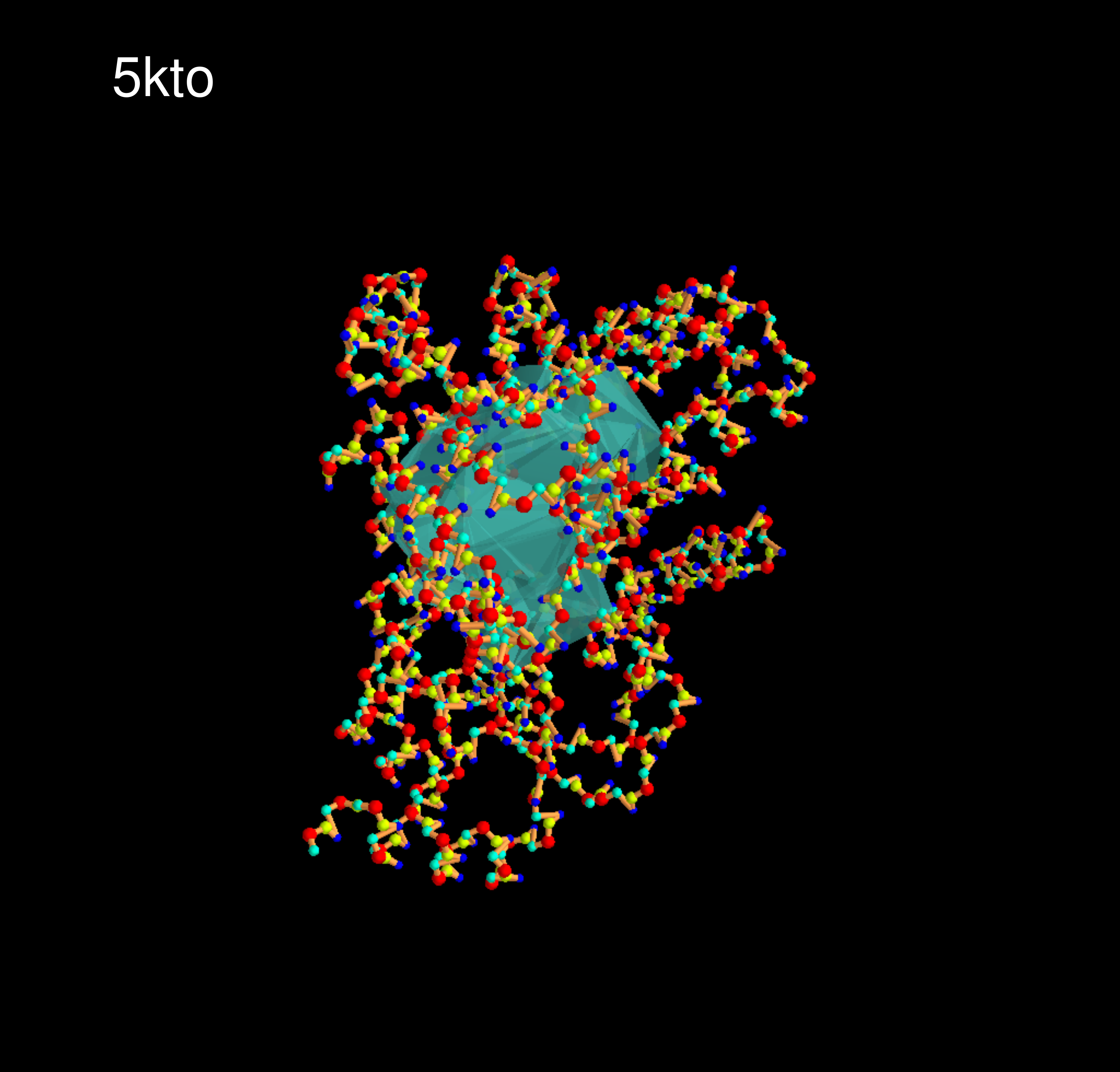}
        \captionsetup{labelformat=empty}
        \caption{}
        \label{supp_fig:pdb_5kto}
    \end{subfigure}
      \centering
      \begin{subfigure}{.24\textwidth} \centering
        \includegraphics[width=0.9\linewidth]{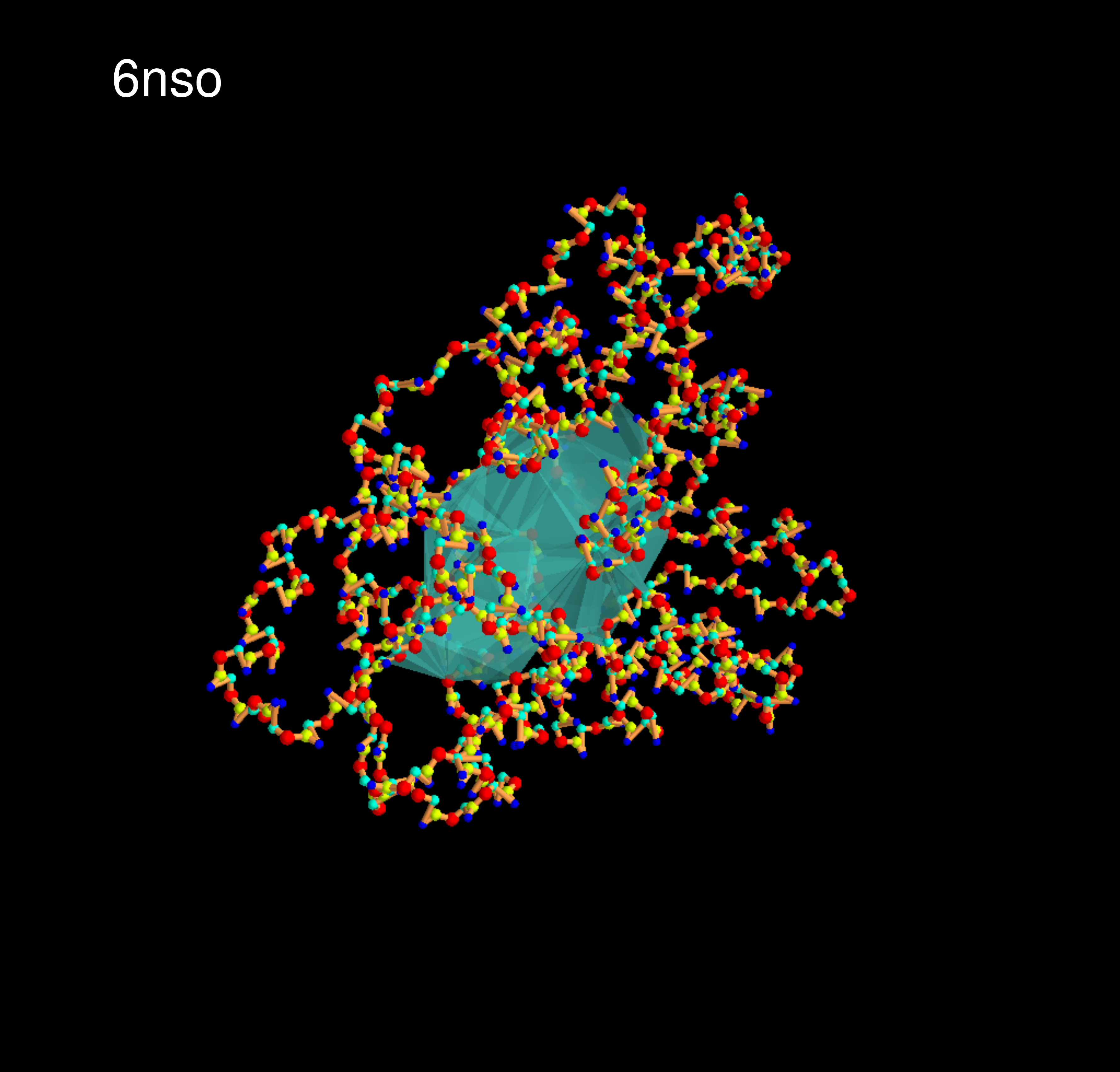}
        \captionsetup{labelformat=empty}
        \caption{}
        \label{supp_fig:pdb_6nso}
    \end{subfigure}
      \centering
      \begin{subfigure}{.24\textwidth} \centering
        \includegraphics[width=0.9\linewidth]{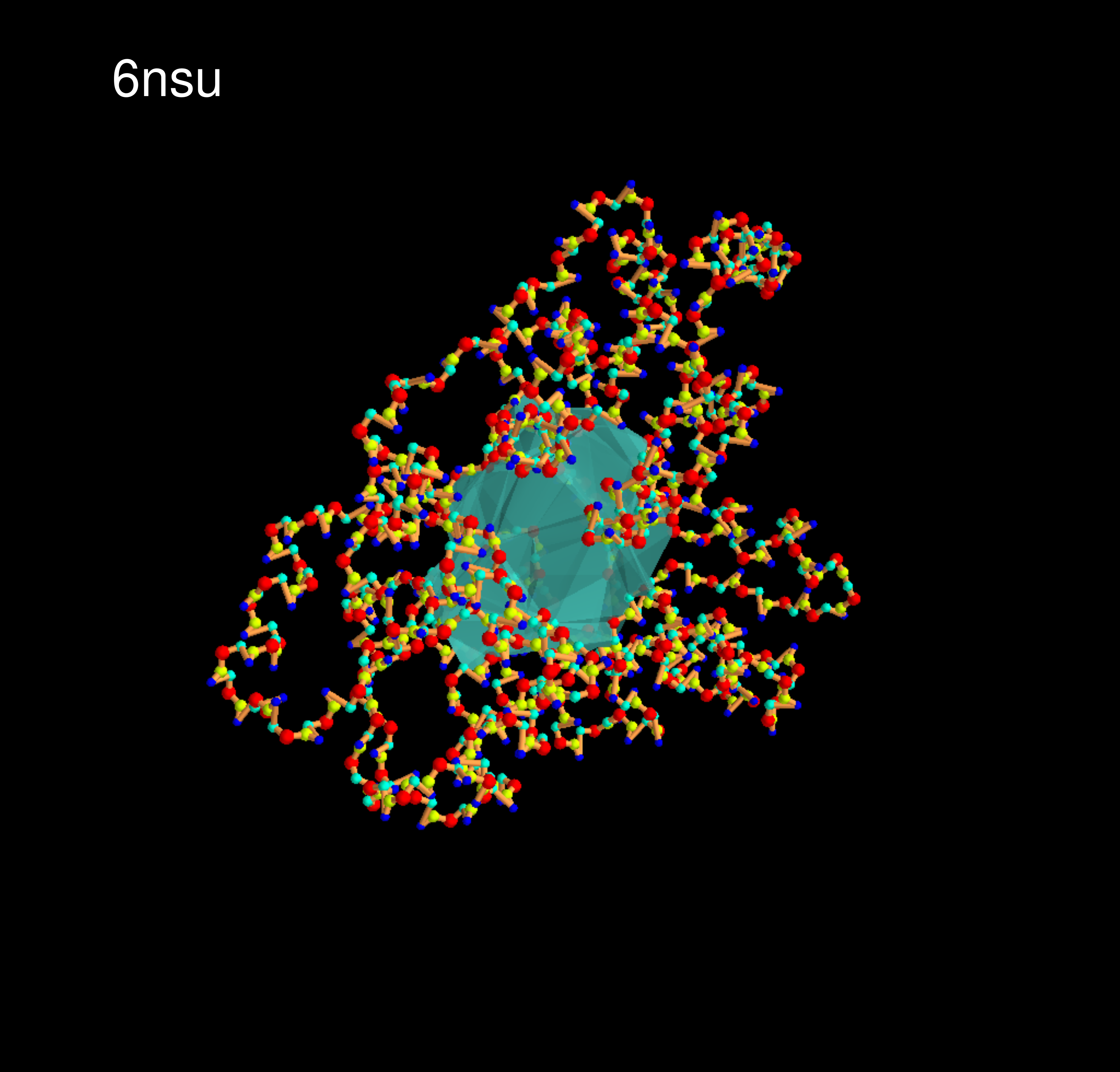}
        \captionsetup{labelformat=empty}
        \caption{}
        \label{supp_fig:pdb_6nsu}
    \end{subfigure}
      \caption{}
      \label{fig:pdb_5}
    \end{figure}

    \begin{figure}[!tbhp]
      \centering
      \begin{subfigure}{.48\textwidth} \centering
        \includegraphics[width=0.9\linewidth]{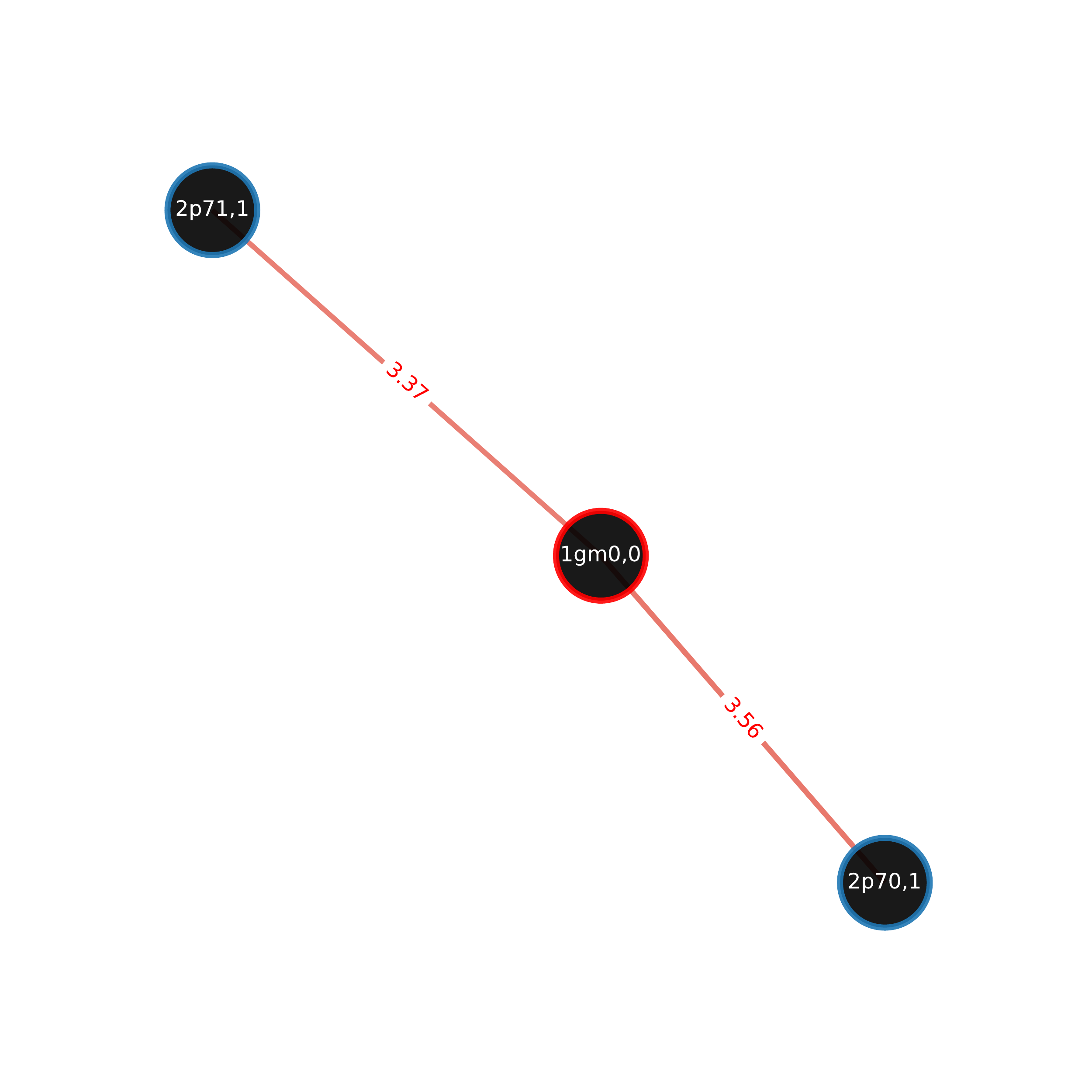}
        \captionsetup{labelformat=empty}
        \caption{}
        \label{supp_fig:pdb_hom_graph_6}
    \end{subfigure}
      \centering
      \begin{subfigure}{.24\textwidth} \centering
        \includegraphics[width=0.9\linewidth]{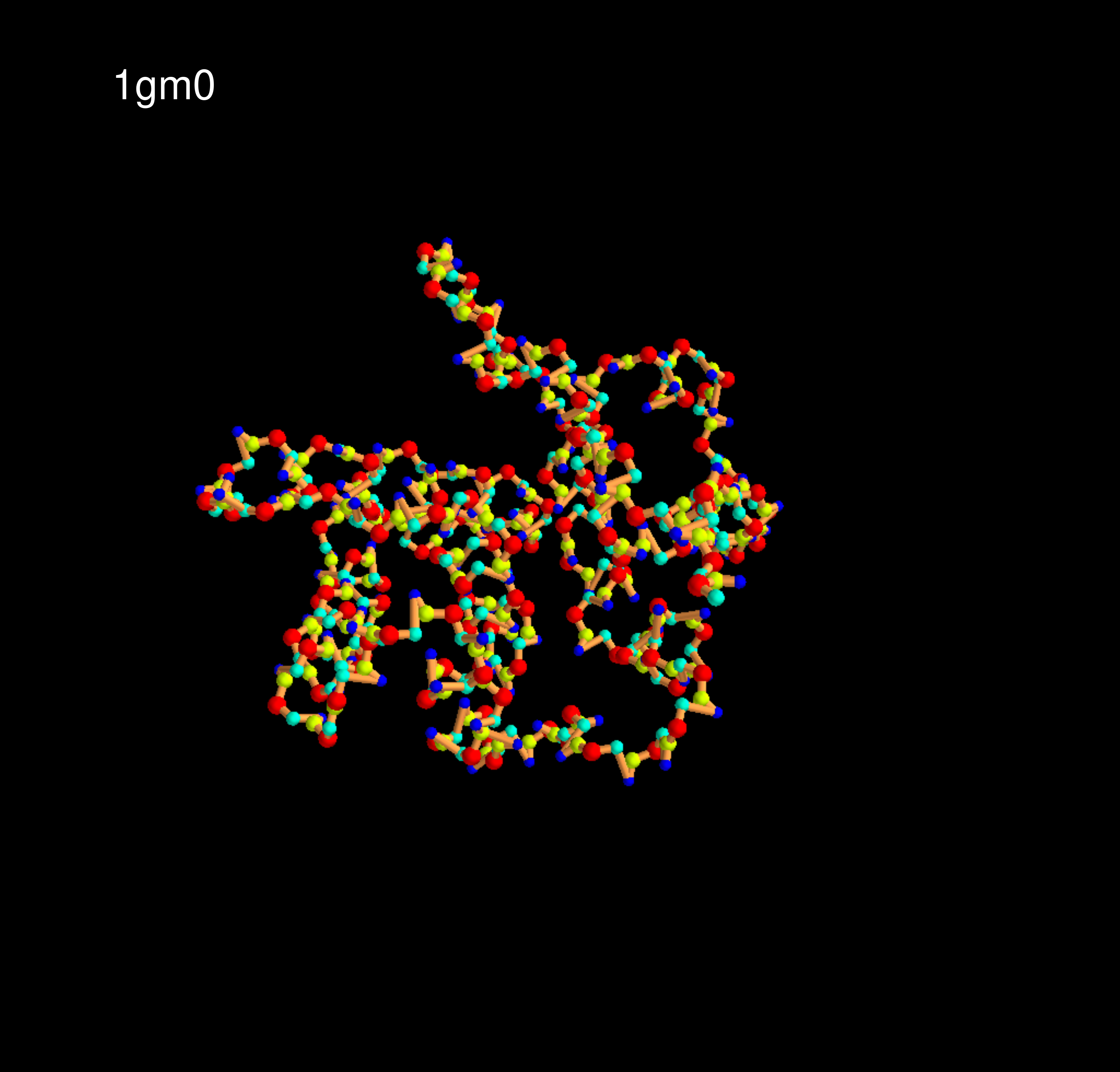}
        \captionsetup{labelformat=empty}
        \caption{}
        \label{supp_fig:pdb_1gm0}
    \end{subfigure}
      \centering
      \begin{subfigure}{.24\textwidth} \centering
        \includegraphics[width=0.9\linewidth]{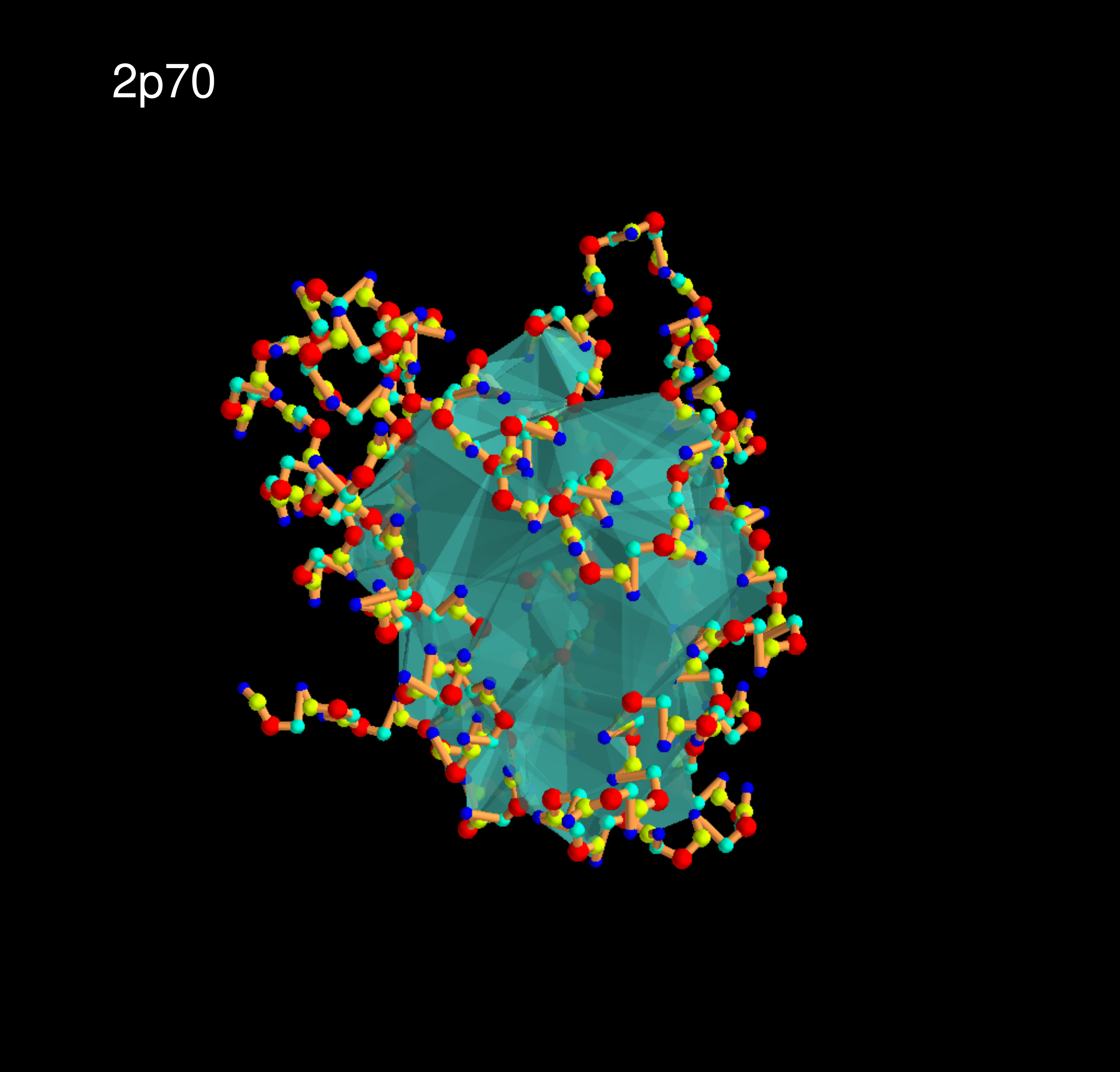}
        \captionsetup{labelformat=empty}
        \caption{}
        \label{supp_fig:pdb_2p70}
    \end{subfigure}
      \centering
      \begin{subfigure}{.24\textwidth} \centering
        \includegraphics[width=0.9\linewidth]{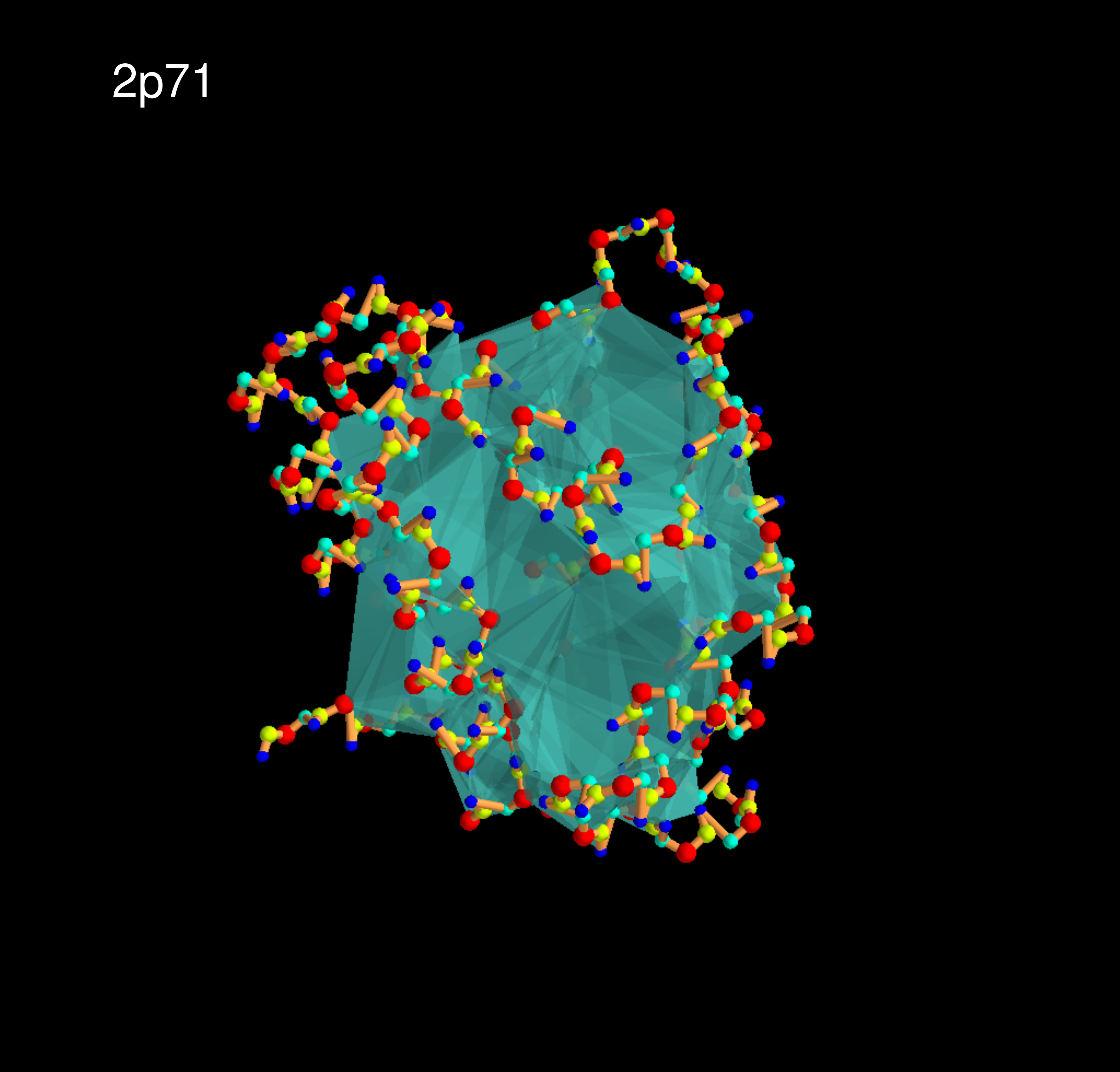}
        \captionsetup{labelformat=empty}
        \caption{}
        \label{supp_fig:pdb_2p71}
    \end{subfigure}
      \caption{}
      \label{fig:pdb_6}
    \end{figure}

    \begin{figure}[!tbhp]
      \centering
      \begin{subfigure}{.48\textwidth} \centering
        \includegraphics[width=0.9\linewidth]{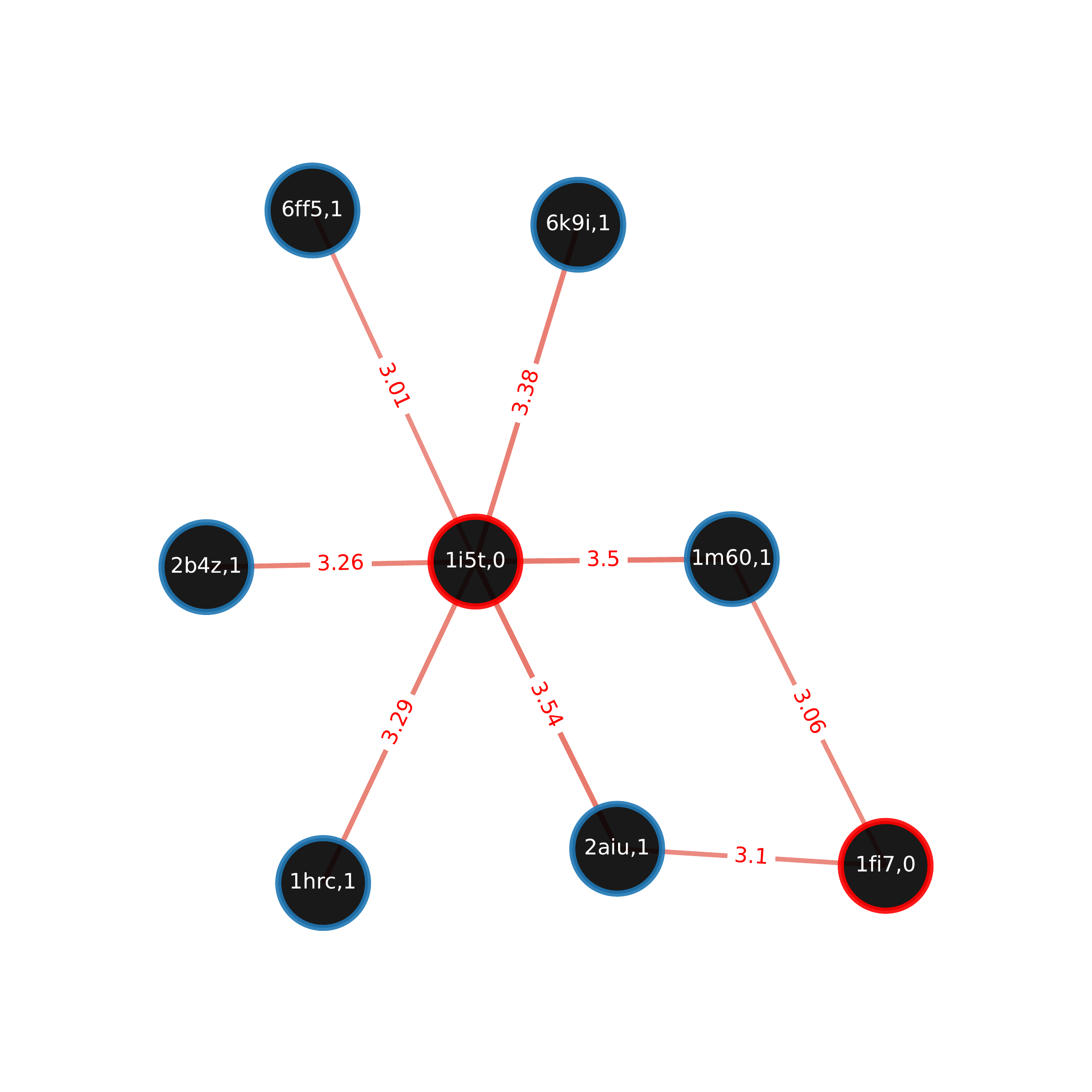}
        \captionsetup{labelformat=empty}
        \caption{}
        \label{supp_fig:pdb_hom_graph_7}
    \end{subfigure}
      \centering
      \begin{subfigure}{.24\textwidth} \centering
        \includegraphics[width=0.9\linewidth]{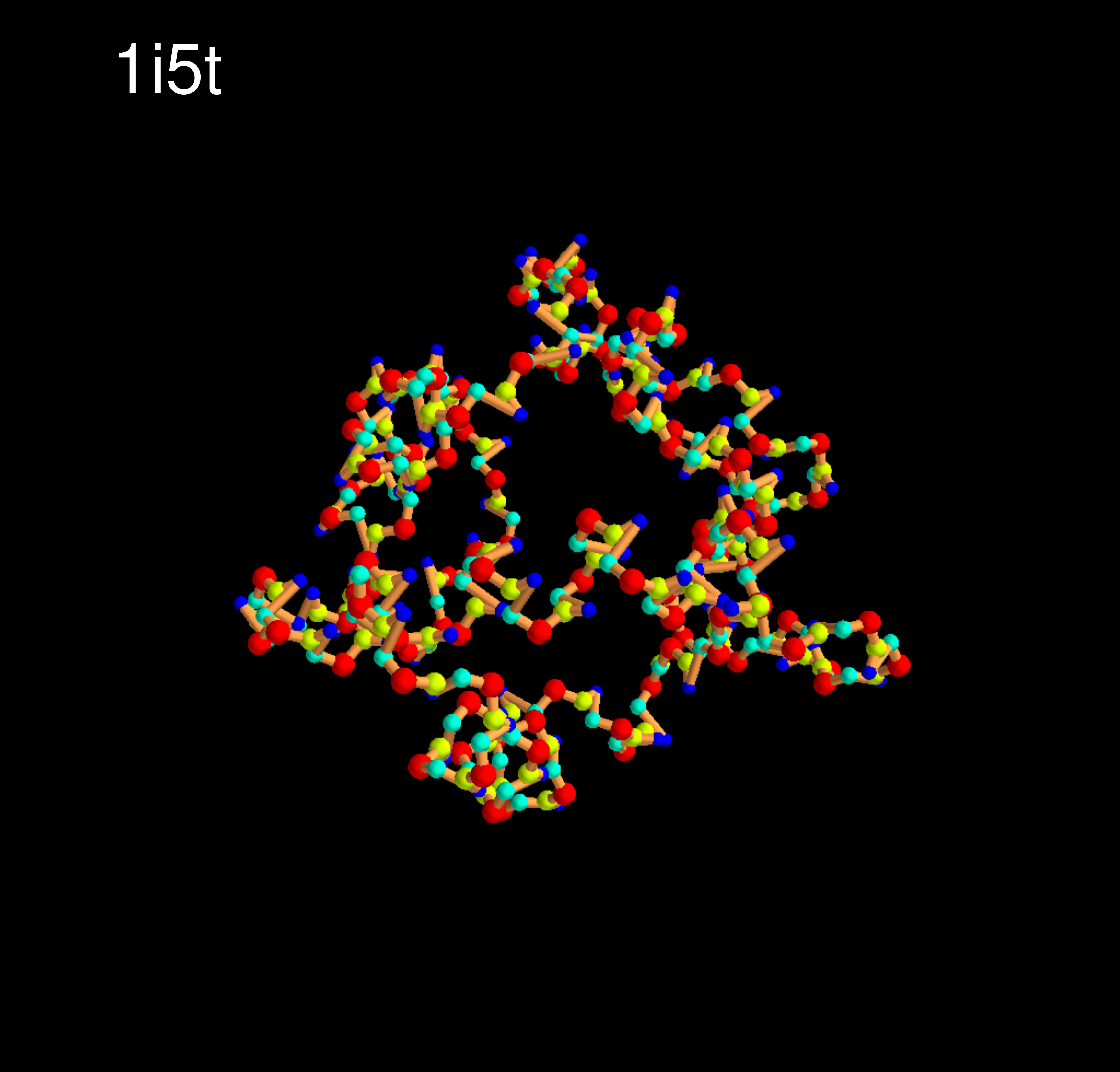}
        \captionsetup{labelformat=empty}
        \caption{}
        \label{supp_fig:pdb_1i5t}
    \end{subfigure}
      \centering
      \begin{subfigure}{.24\textwidth} \centering
        \includegraphics[width=0.9\linewidth]{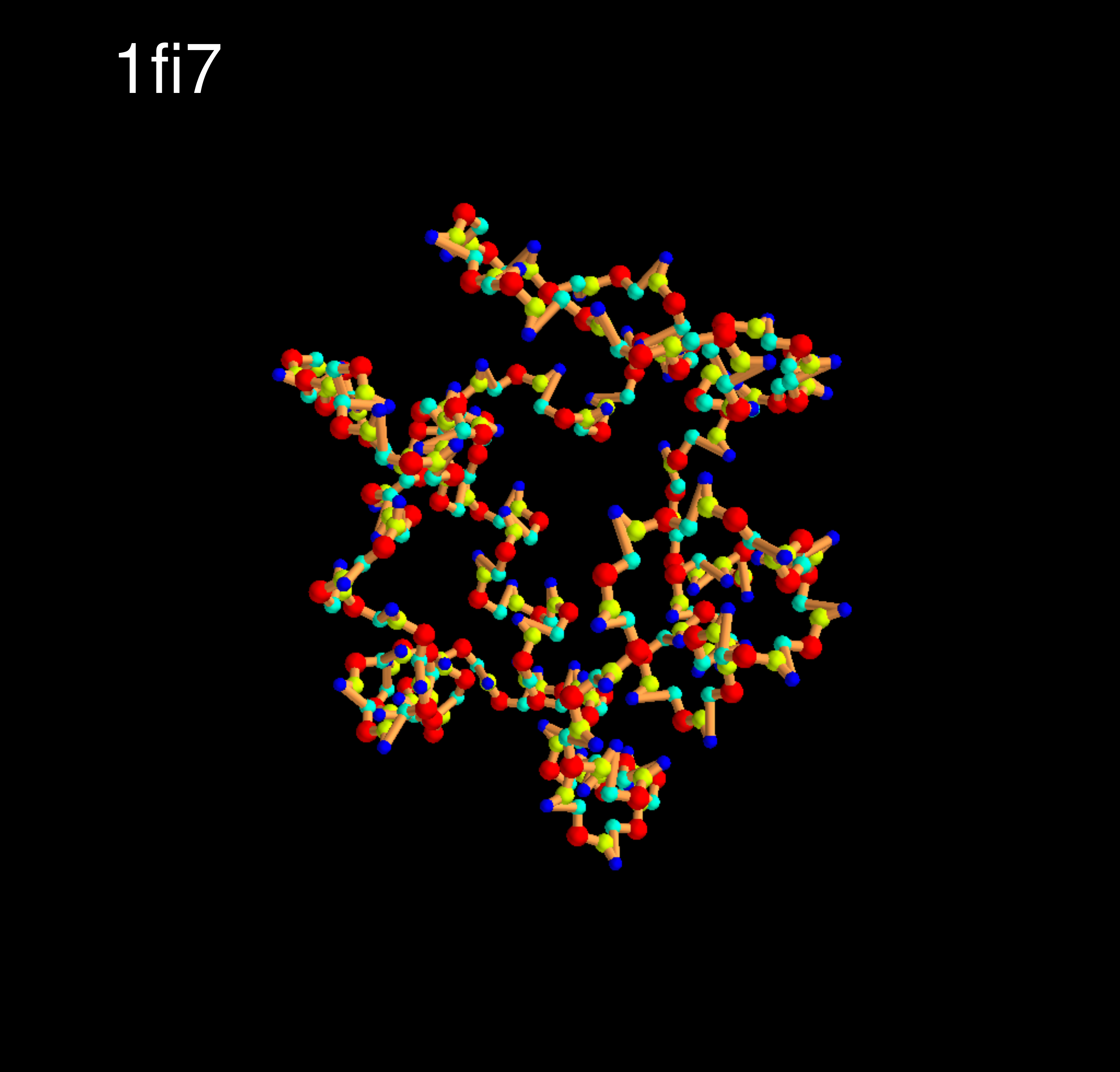}
        \captionsetup{labelformat=empty}
        \caption{}
        \label{supp_fig:pdb_1fi7}
    \end{subfigure}
      \centering
      \begin{subfigure}{.24\textwidth} \centering
        \includegraphics[width=0.9\linewidth]{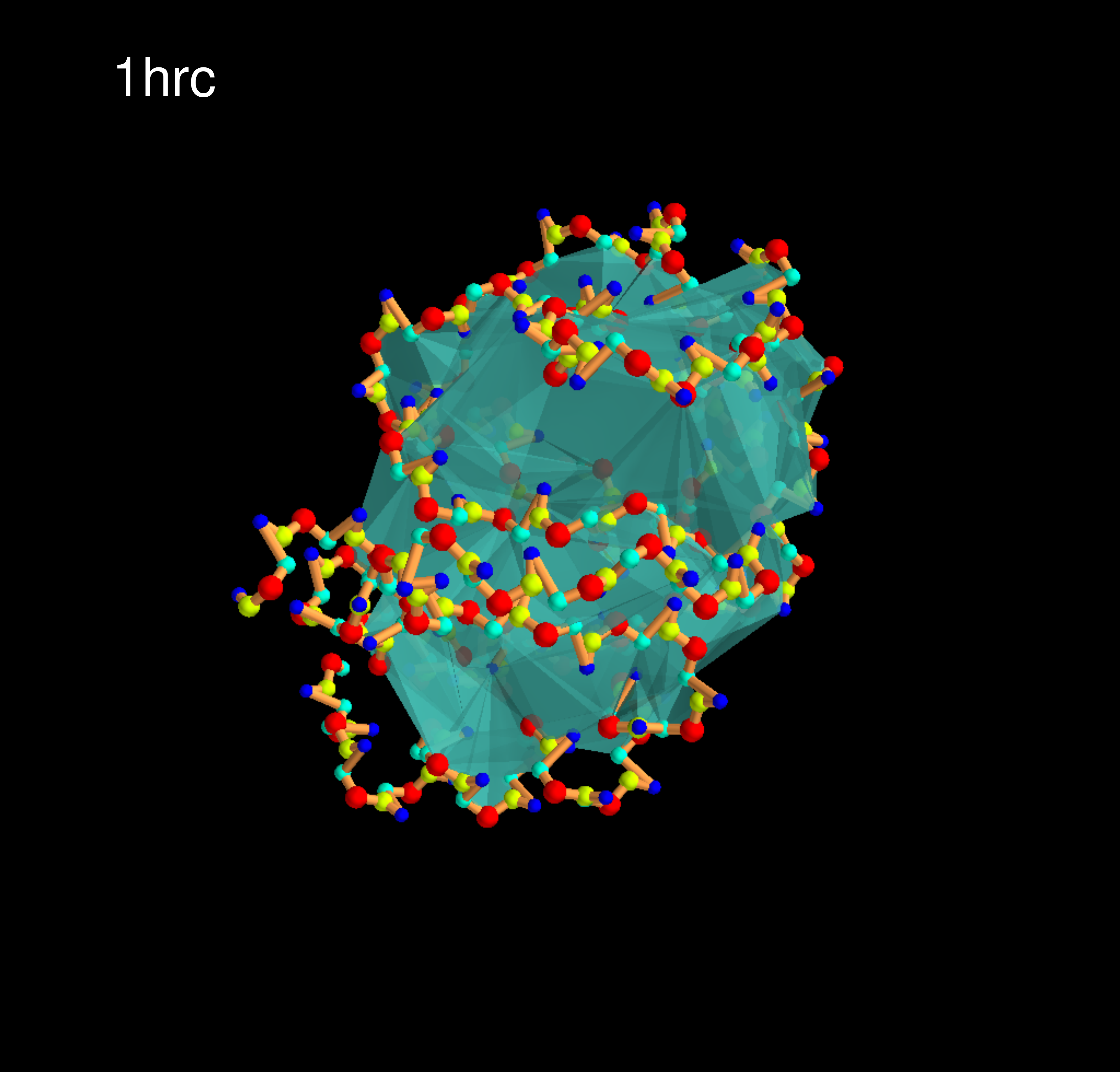}
        \captionsetup{labelformat=empty}
        \caption{}
        \label{supp_fig:pdb_1hrc}
    \end{subfigure}
      \centering
      \begin{subfigure}{.24\textwidth} \centering
        \includegraphics[width=0.9\linewidth]{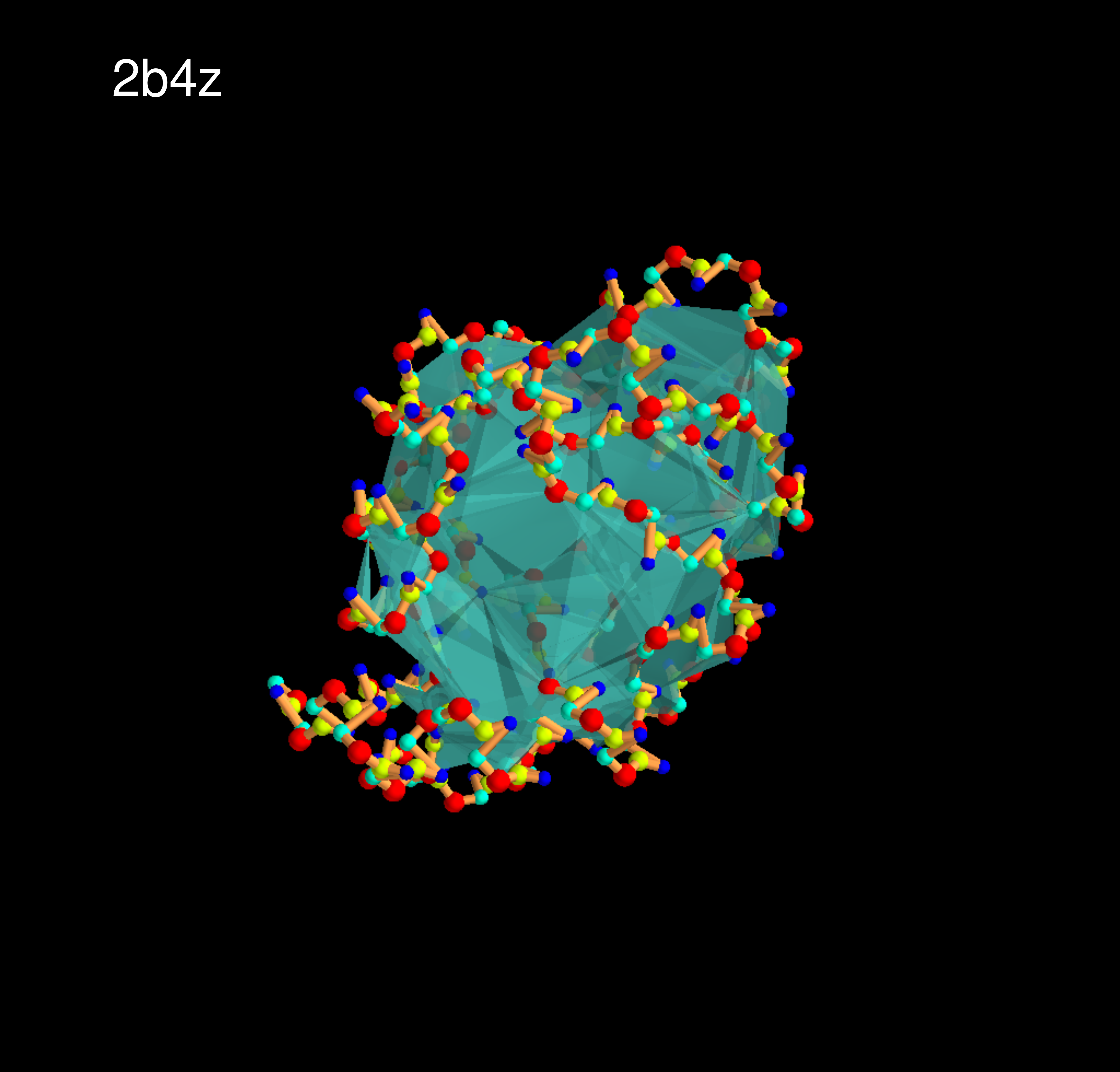}
        \captionsetup{labelformat=empty}
        \caption{}
        \label{supp_fig:pdb_2b4z}
    \end{subfigure}
      \centering
      \begin{subfigure}{.24\textwidth} \centering
        \includegraphics[width=0.9\linewidth]{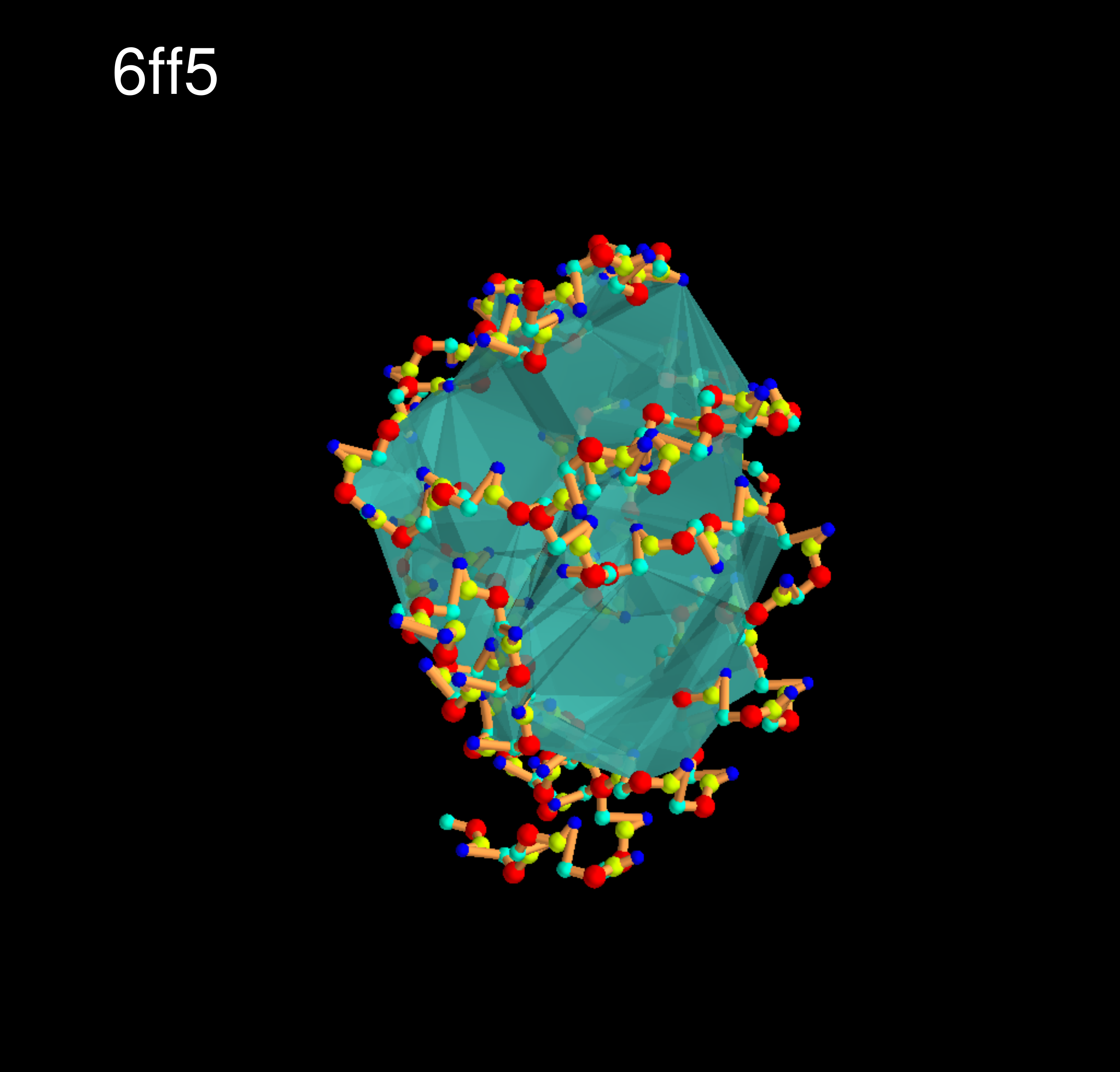}
        \captionsetup{labelformat=empty}
        \caption{}
        \label{supp_fig:pdb_6ff5}
    \end{subfigure}
      \centering
      \begin{subfigure}{.24\textwidth} \centering
        \includegraphics[width=0.9\linewidth]{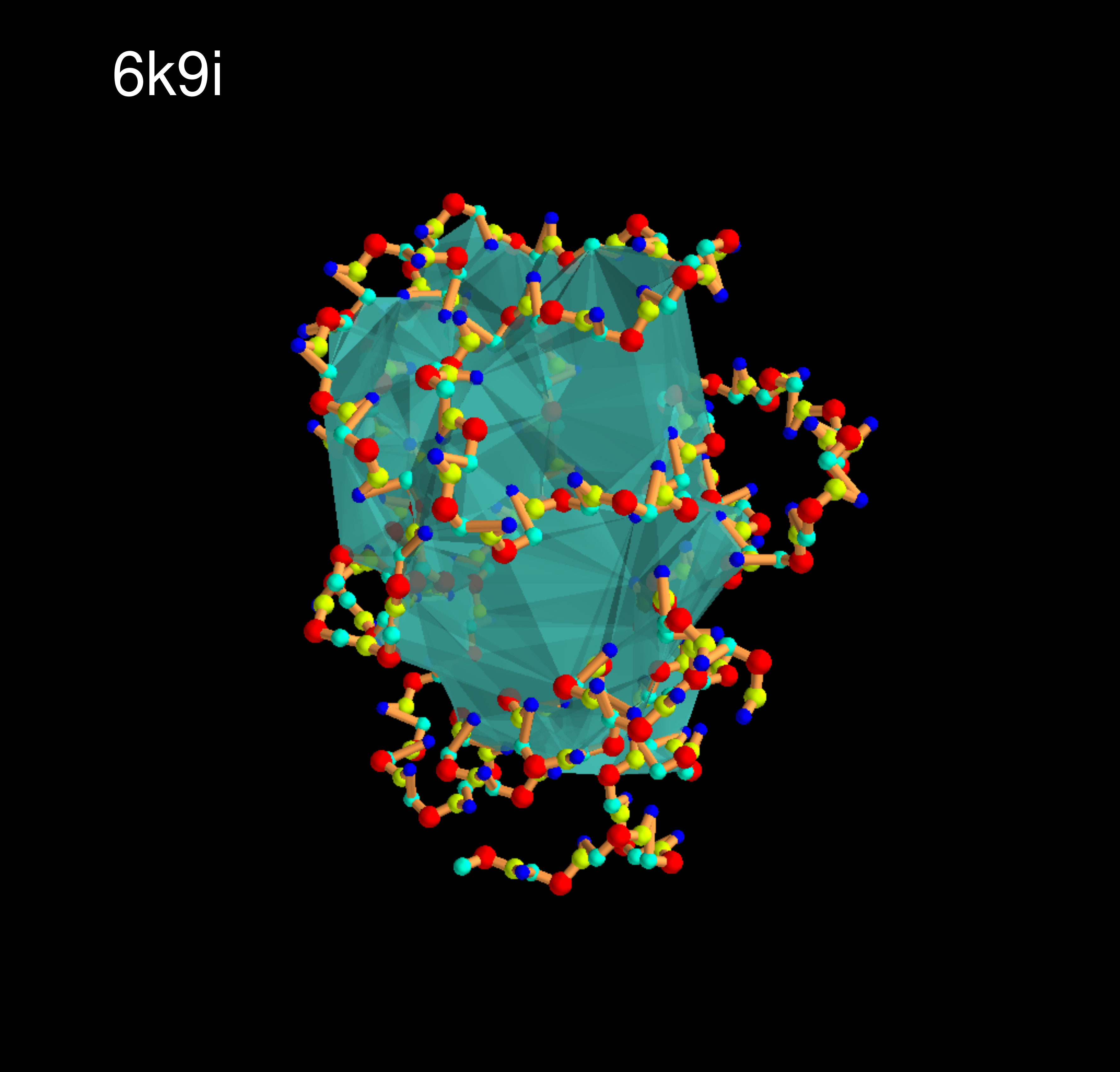}
        \captionsetup{labelformat=empty}
        \caption{}
        \label{supp_fig:pdb_6k9i}
    \end{subfigure}
      \centering
      \begin{subfigure}{.24\textwidth} \centering
        \includegraphics[width=0.9\linewidth]{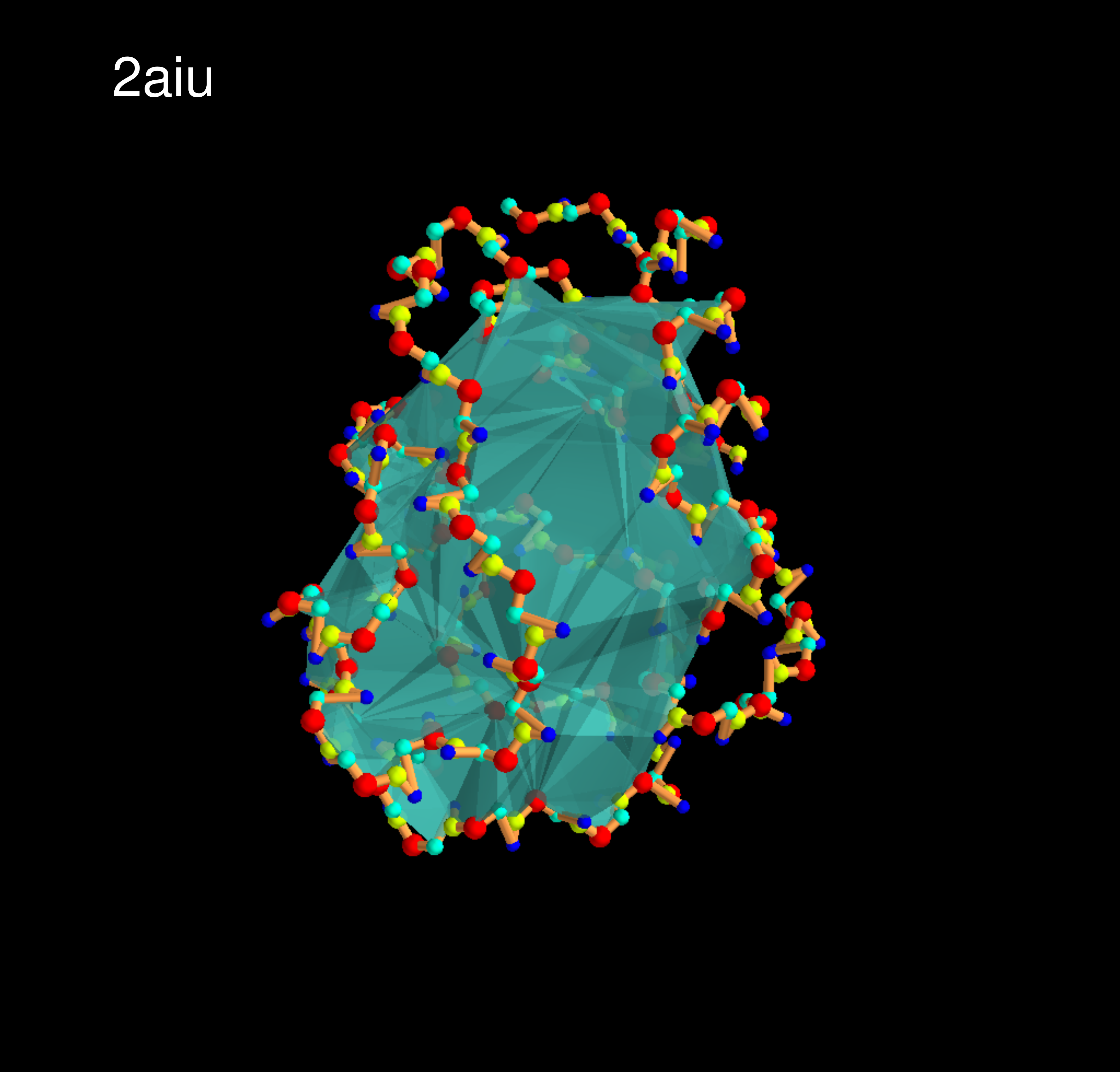}
        \captionsetup{labelformat=empty}
        \caption{}
        \label{supp_fig:pdb_2aiu}
    \end{subfigure}
      \centering
      \begin{subfigure}{.24\textwidth} \centering
        \includegraphics[width=0.9\linewidth]{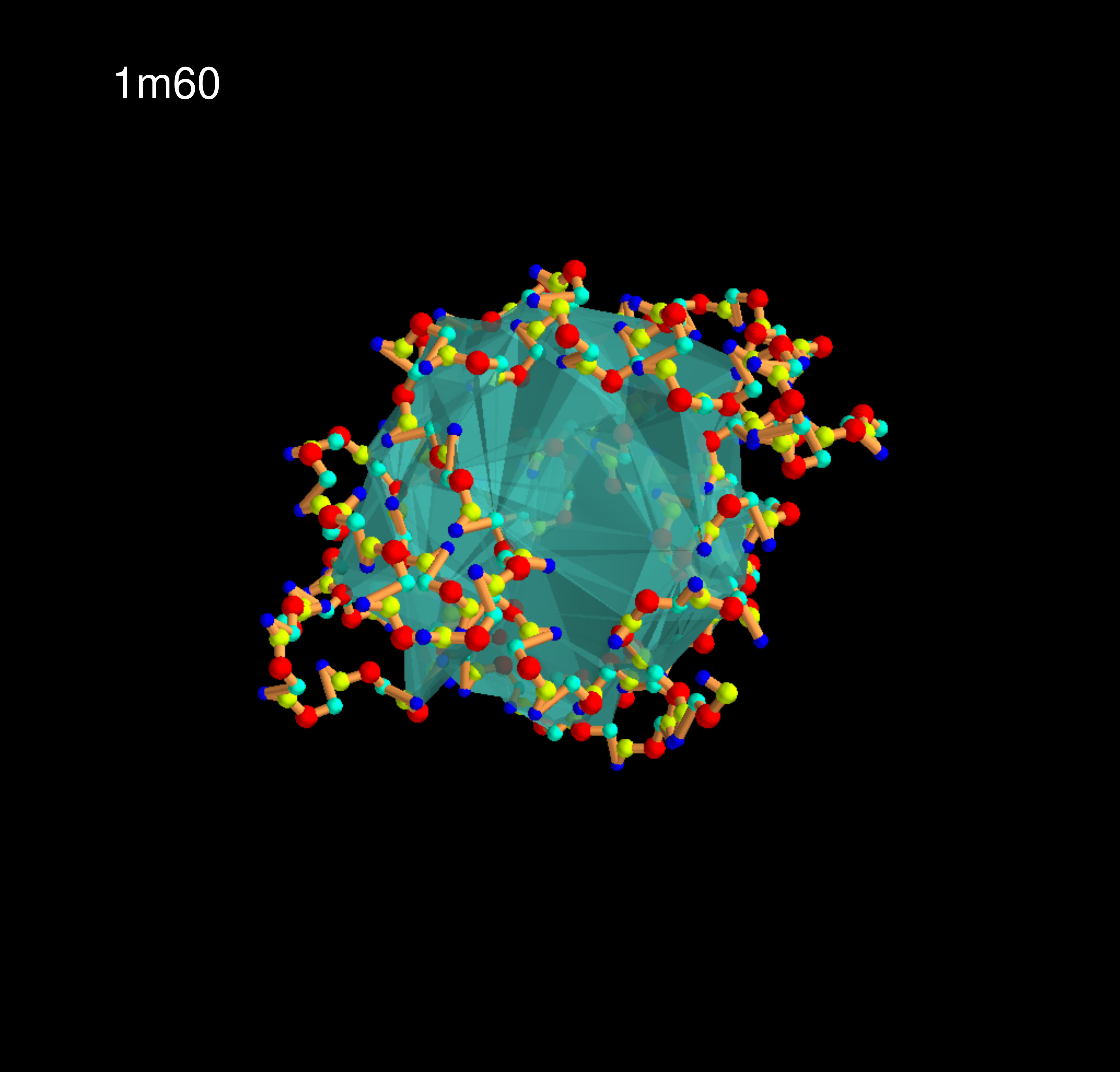}
        \captionsetup{labelformat=empty}
        \caption{}
        \label{supp_fig:pdb_1m60}
    \end{subfigure}
      \caption{}
      \label{fig:pdb_7}
    \end{figure}

    \begin{figure}[!tbhp]
      \centering
      \begin{subfigure}{.48\textwidth} \centering
        \includegraphics[width=0.9\linewidth]{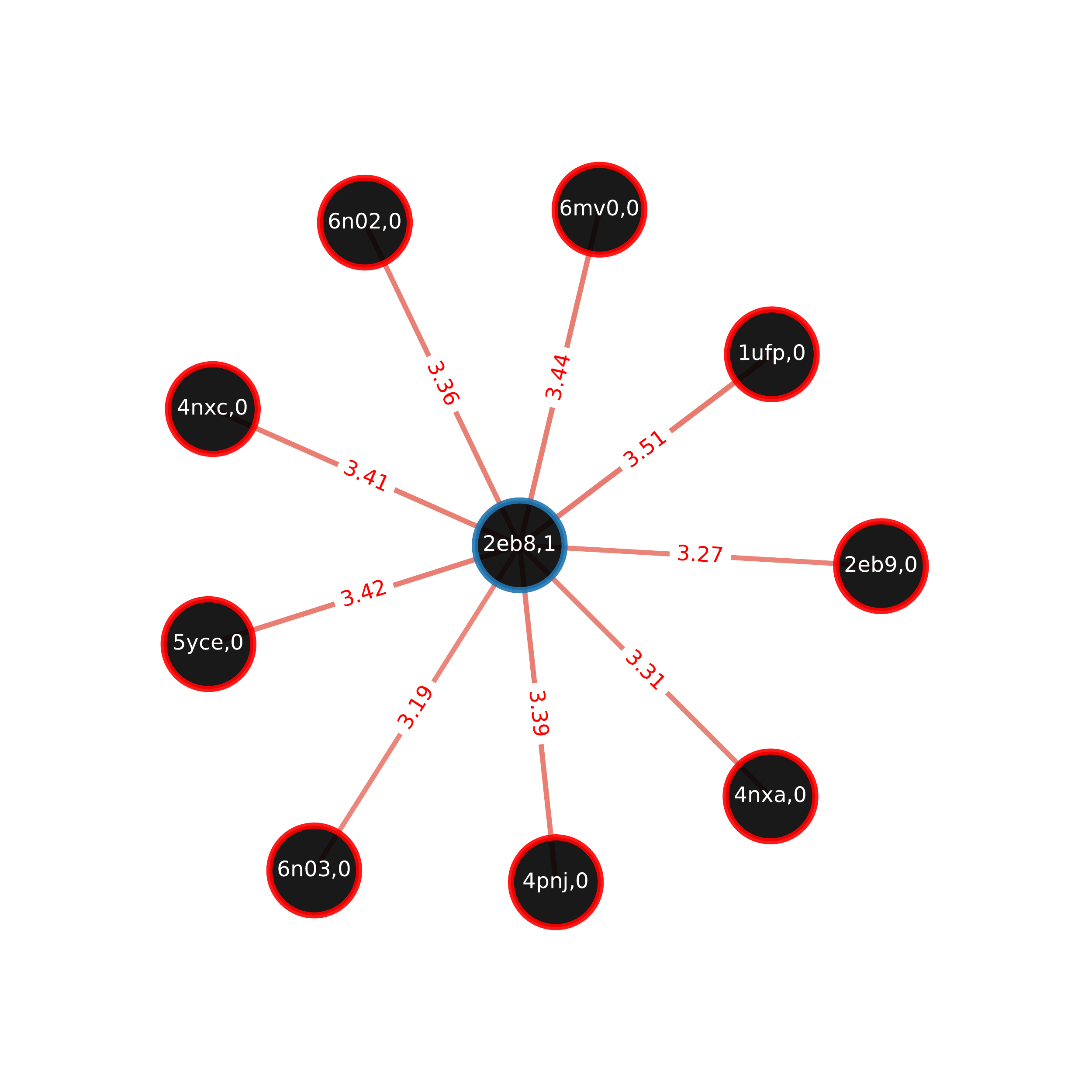}
        \captionsetup{labelformat=empty}
        \caption{}
        \label{supp_fig:pdb_hom_graph_8}
    \end{subfigure}
      \centering
      \begin{subfigure}{.24\textwidth} \centering
        \includegraphics[width=0.9\linewidth]{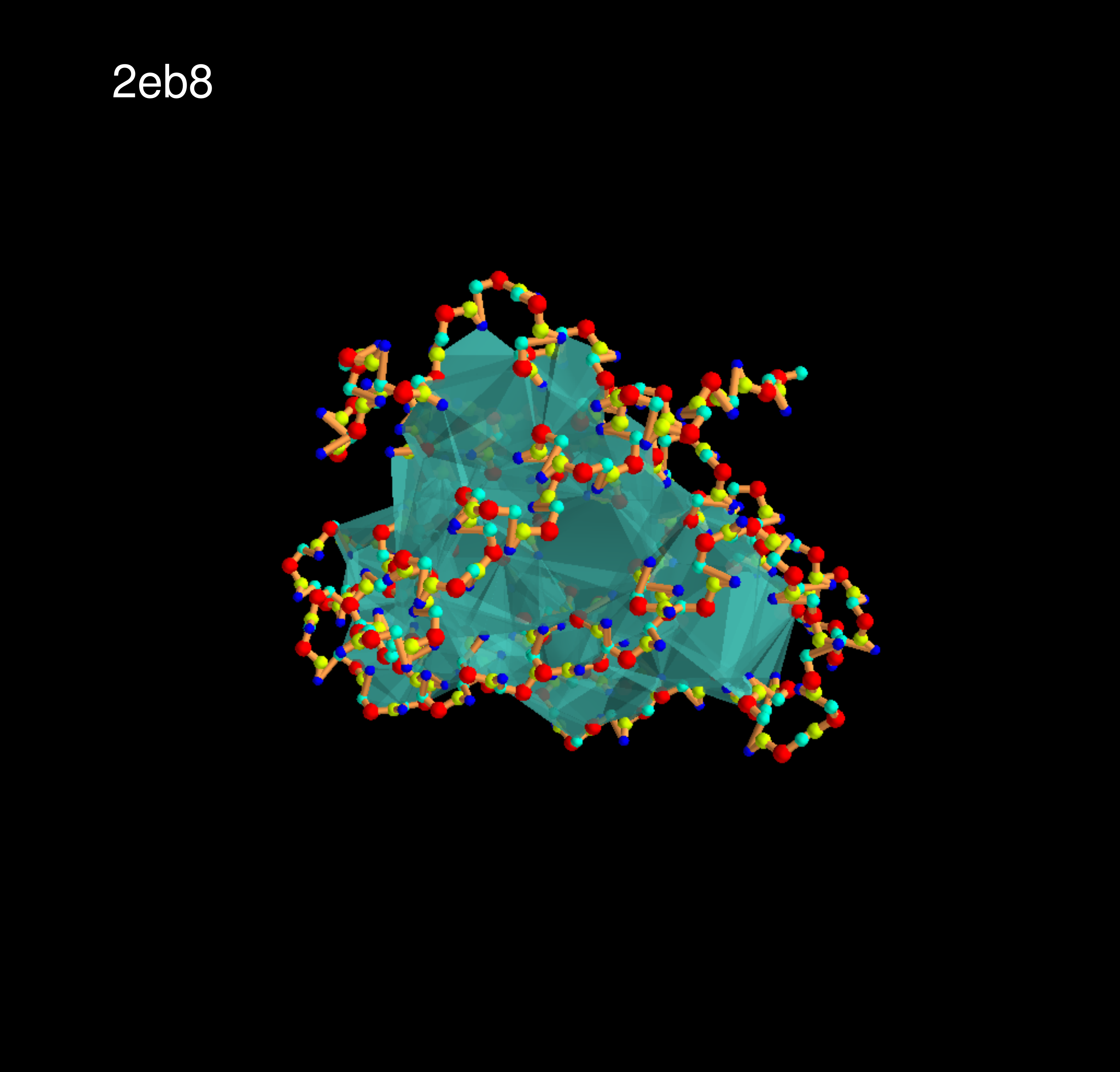}
        \captionsetup{labelformat=empty}
        \caption{}
        \label{supp_fig:pdb_2eb8}
    \end{subfigure}
      \centering
      \begin{subfigure}{.24\textwidth} \centering
        \includegraphics[width=0.9\linewidth]{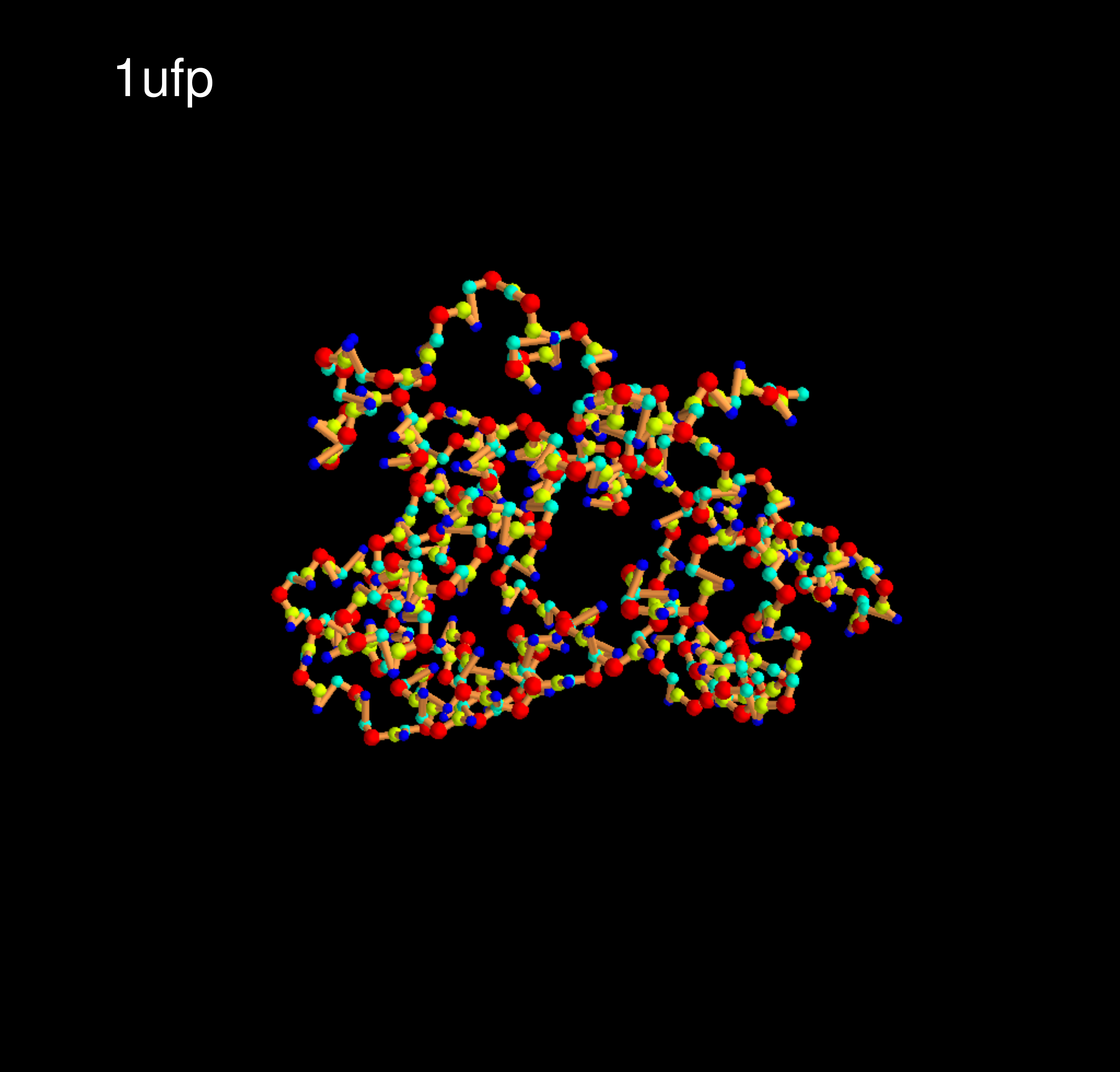}
        \captionsetup{labelformat=empty}
        \caption{}
        \label{supp_fig:pdb_1ufp}
    \end{subfigure}
      \centering
      \begin{subfigure}{.24\textwidth} \centering
        \includegraphics[width=0.9\linewidth]{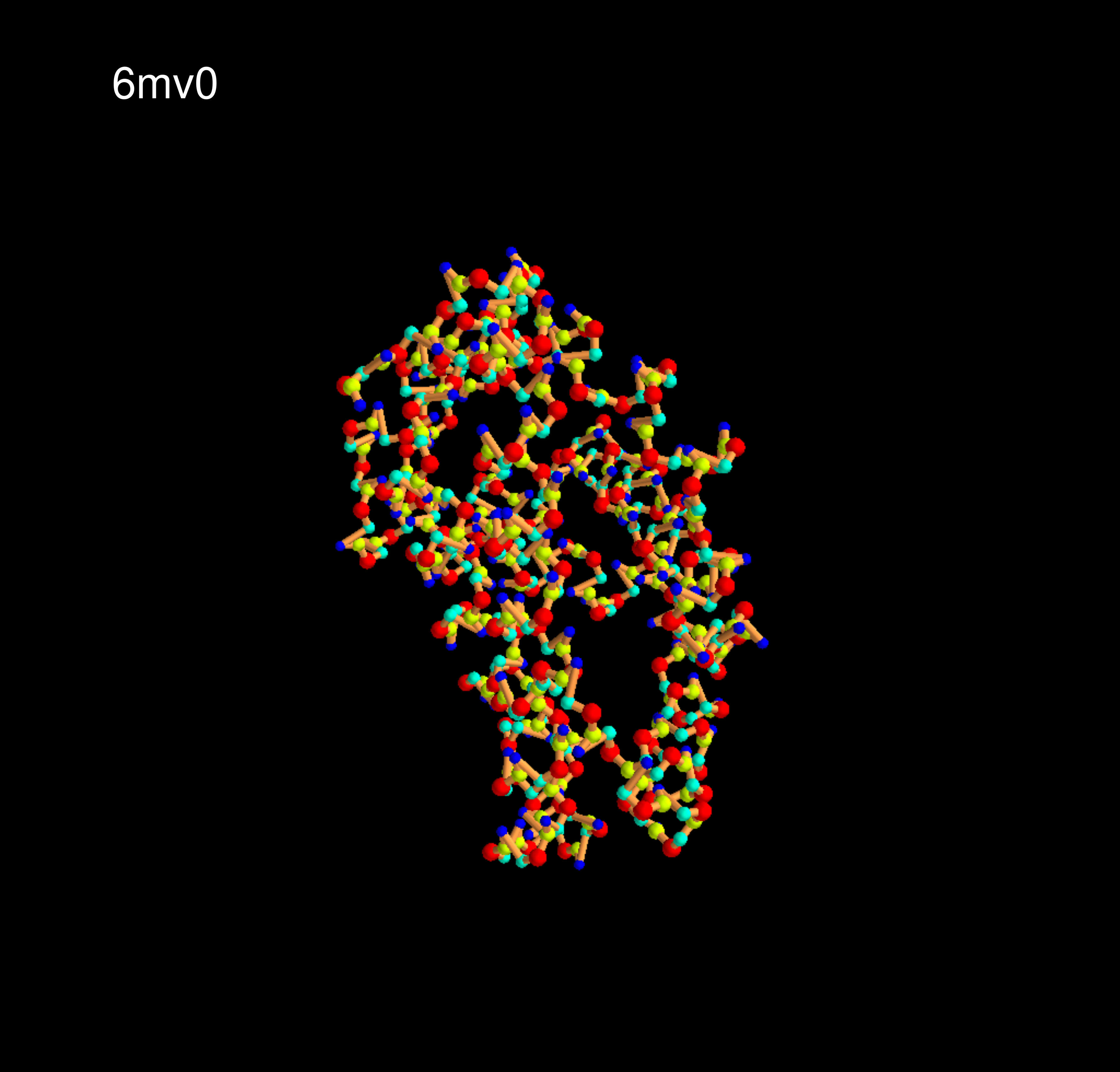}
        \captionsetup{labelformat=empty}
        \caption{}
        \label{supp_fig:pdb_6mv0}
    \end{subfigure}
      \centering
      \begin{subfigure}{.24\textwidth} \centering
        \includegraphics[width=0.9\linewidth]{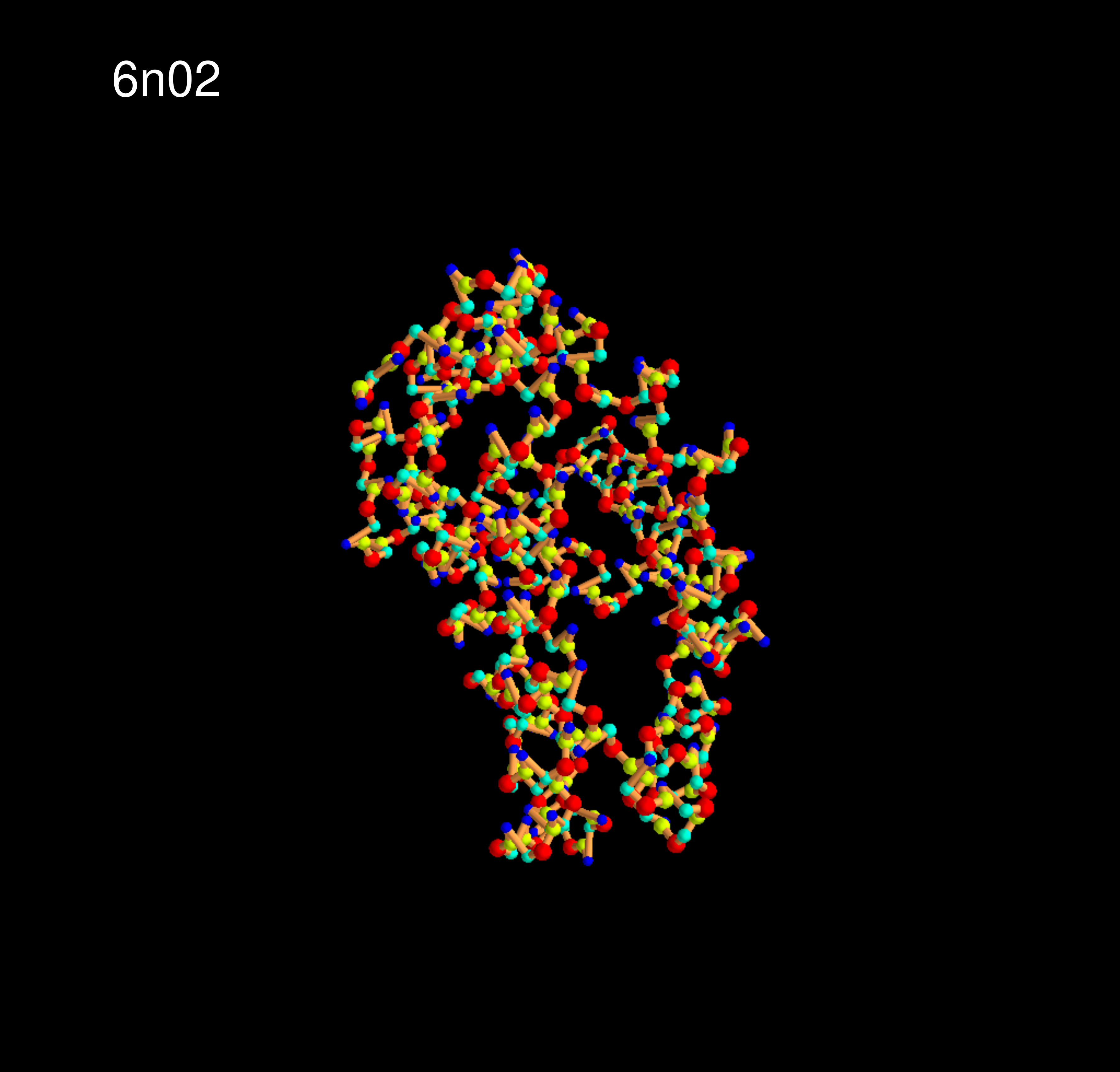}
        \captionsetup{labelformat=empty}
        \caption{}
        \label{supp_fig:pdb_6n02}
    \end{subfigure}
      \centering
      \begin{subfigure}{.24\textwidth} \centering
        \includegraphics[width=0.9\linewidth]{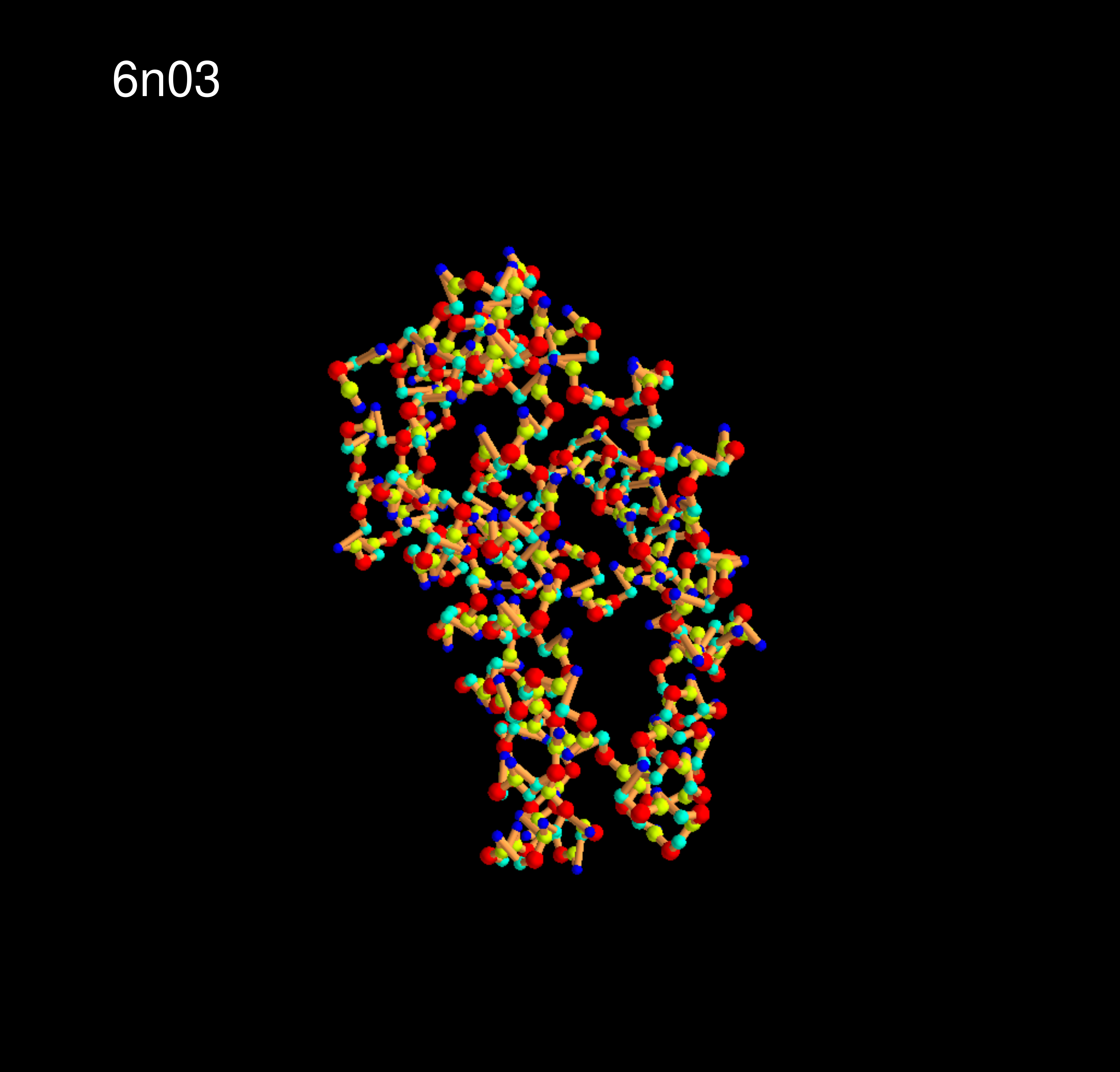}
        \captionsetup{labelformat=empty}
        \caption{}
        \label{supp_fig:pdb_6n03}
    \end{subfigure}
      \centering
      \begin{subfigure}{.24\textwidth} \centering
        \includegraphics[width=0.9\linewidth]{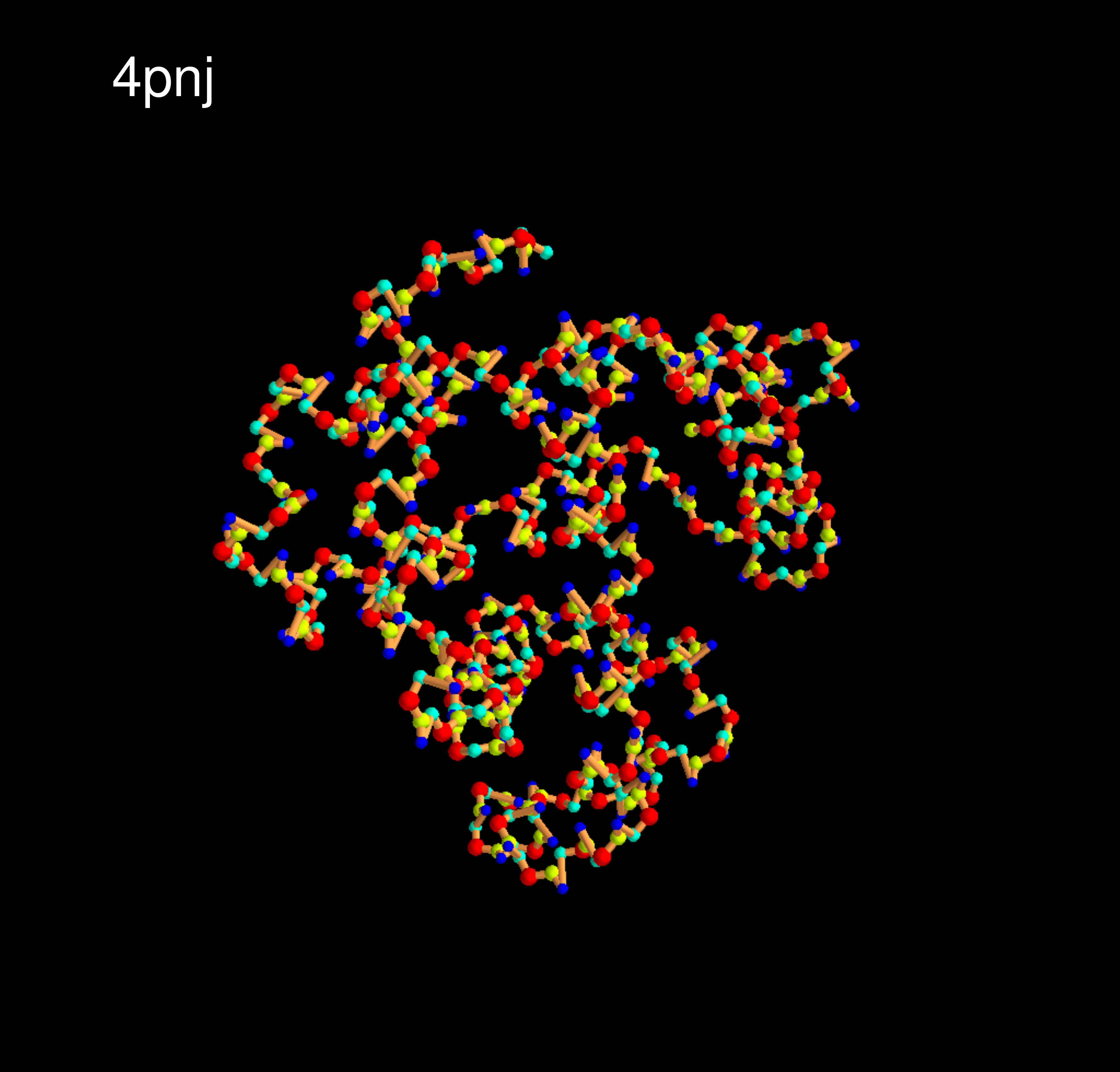}
        \captionsetup{labelformat=empty}
        \caption{}
        \label{supp_fig:pdb_4pnj}
    \end{subfigure}
      \centering
      \begin{subfigure}{.24\textwidth} \centering
        \includegraphics[width=0.9\linewidth]{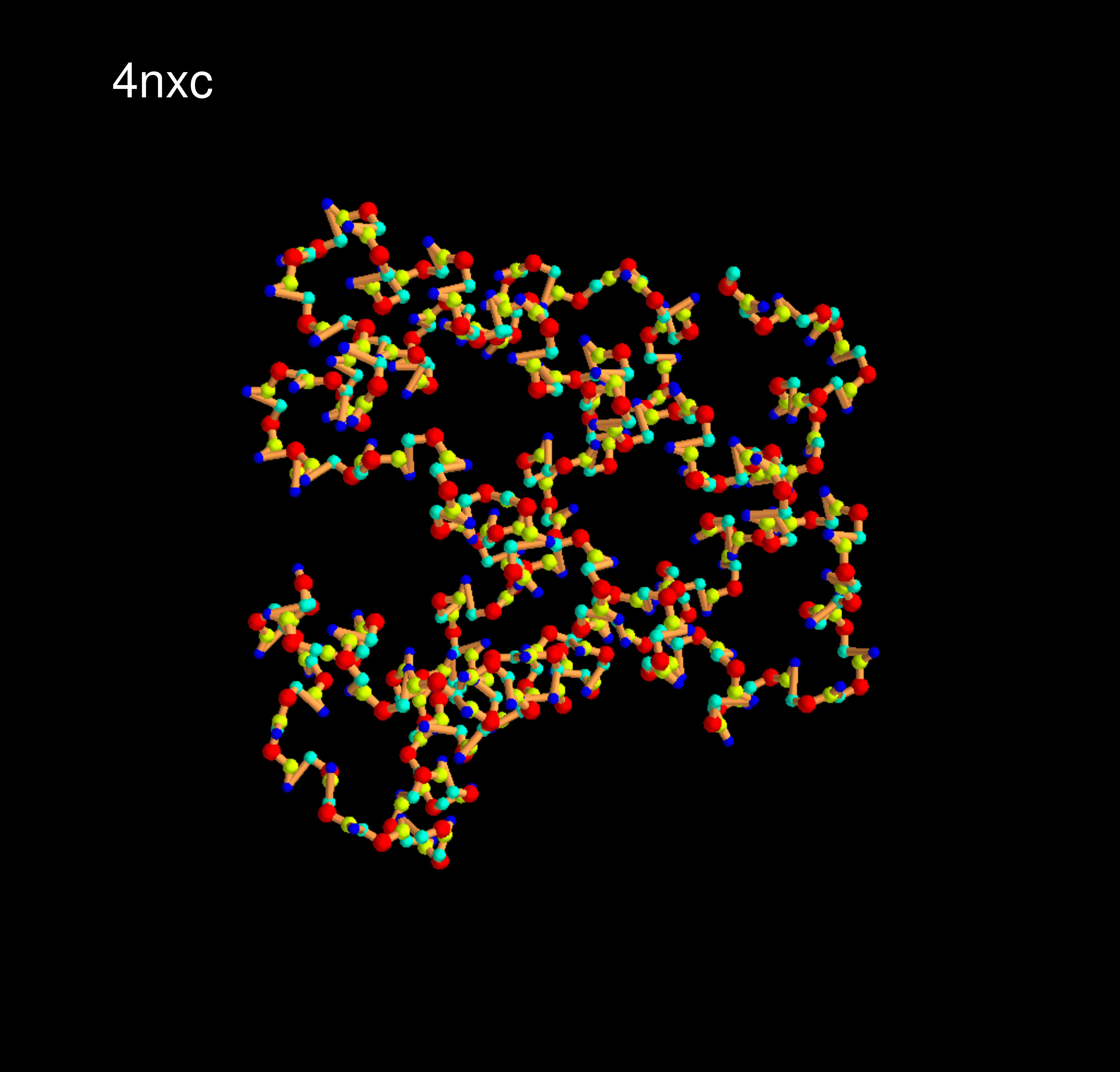}
        \captionsetup{labelformat=empty}
        \caption{}
        \label{supp_fig:pdb_4nxc}
    \end{subfigure}
      \centering
      \begin{subfigure}{.24\textwidth} \centering
        \includegraphics[width=0.9\linewidth]{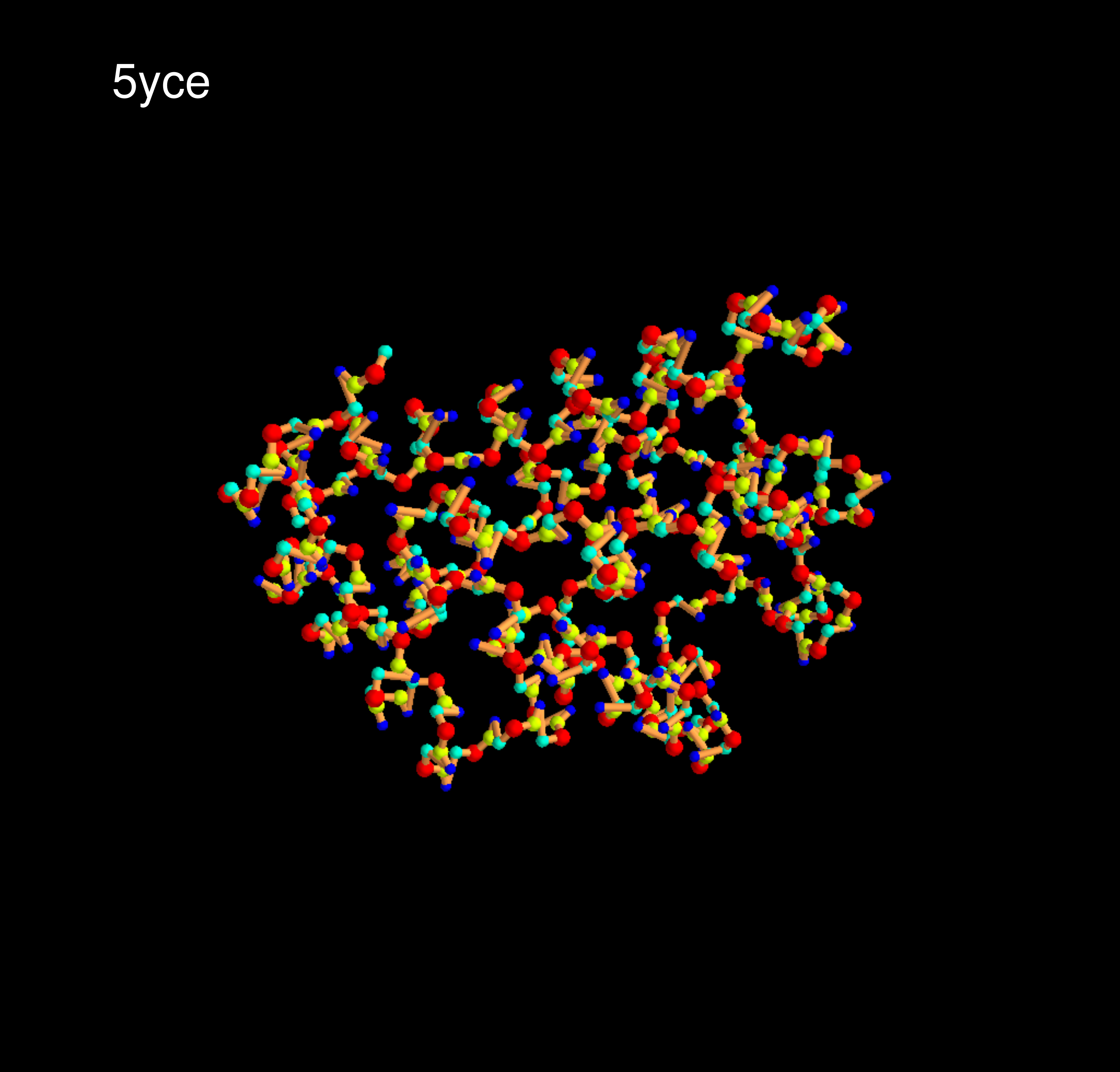}
        \captionsetup{labelformat=empty}
        \caption{}
        \label{supp_fig:pdb_5yce}
    \end{subfigure}
      \centering
      \begin{subfigure}{.24\textwidth} \centering
        \includegraphics[width=0.9\linewidth]{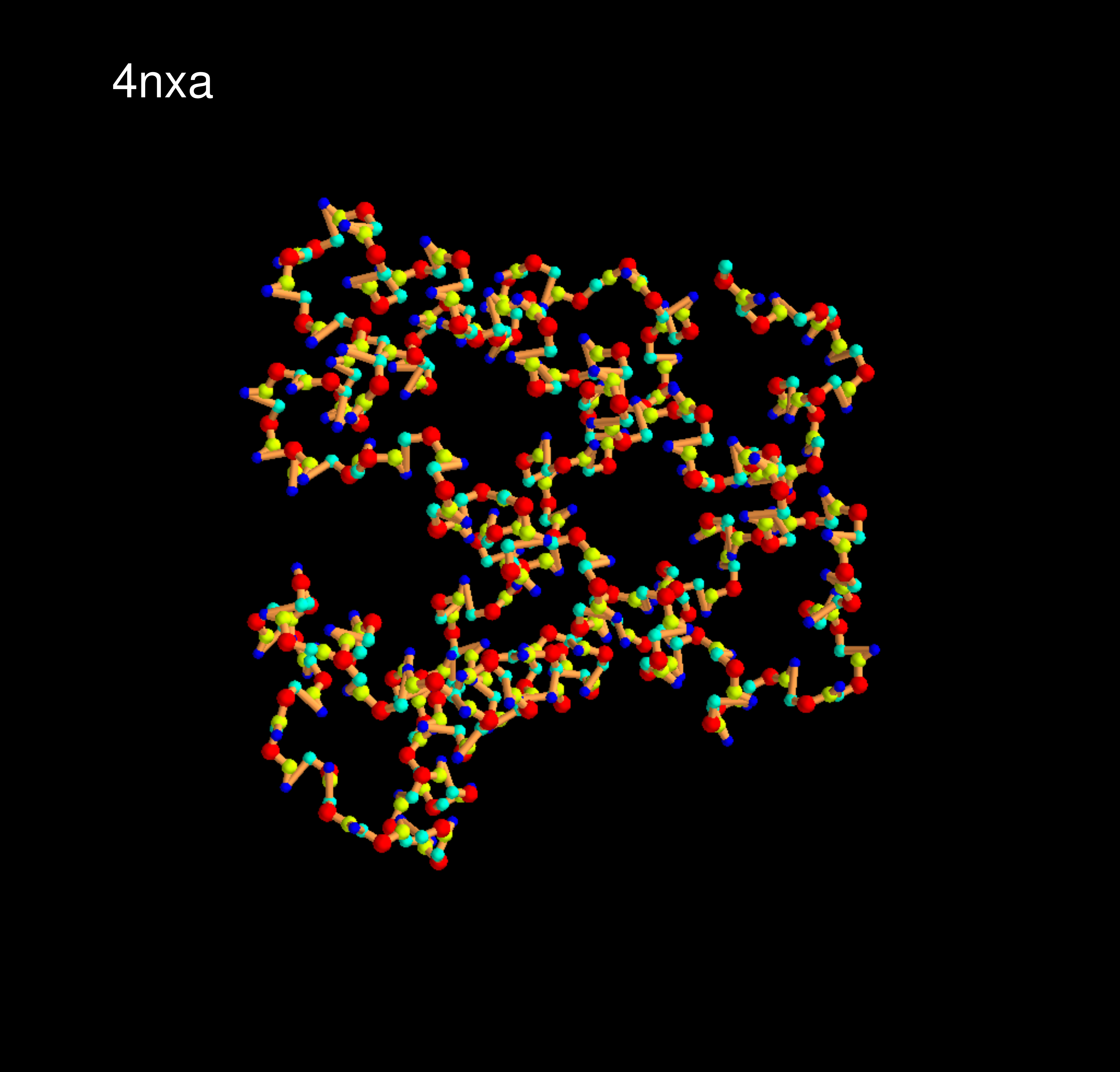}
        \captionsetup{labelformat=empty}
        \caption{}
        \label{supp_fig:pdb_4nxa}
    \end{subfigure}
      \centering
      \begin{subfigure}{.24\textwidth} \centering
        \includegraphics[width=0.9\linewidth]{figures/pdb_5yce.pdf}
        \captionsetup{labelformat=empty}
        \caption{}
        \label{supp_fig:pdb_5yce}
    \end{subfigure}
      \centering
      \begin{subfigure}{.24\textwidth} \centering
        \includegraphics[width=0.9\linewidth]{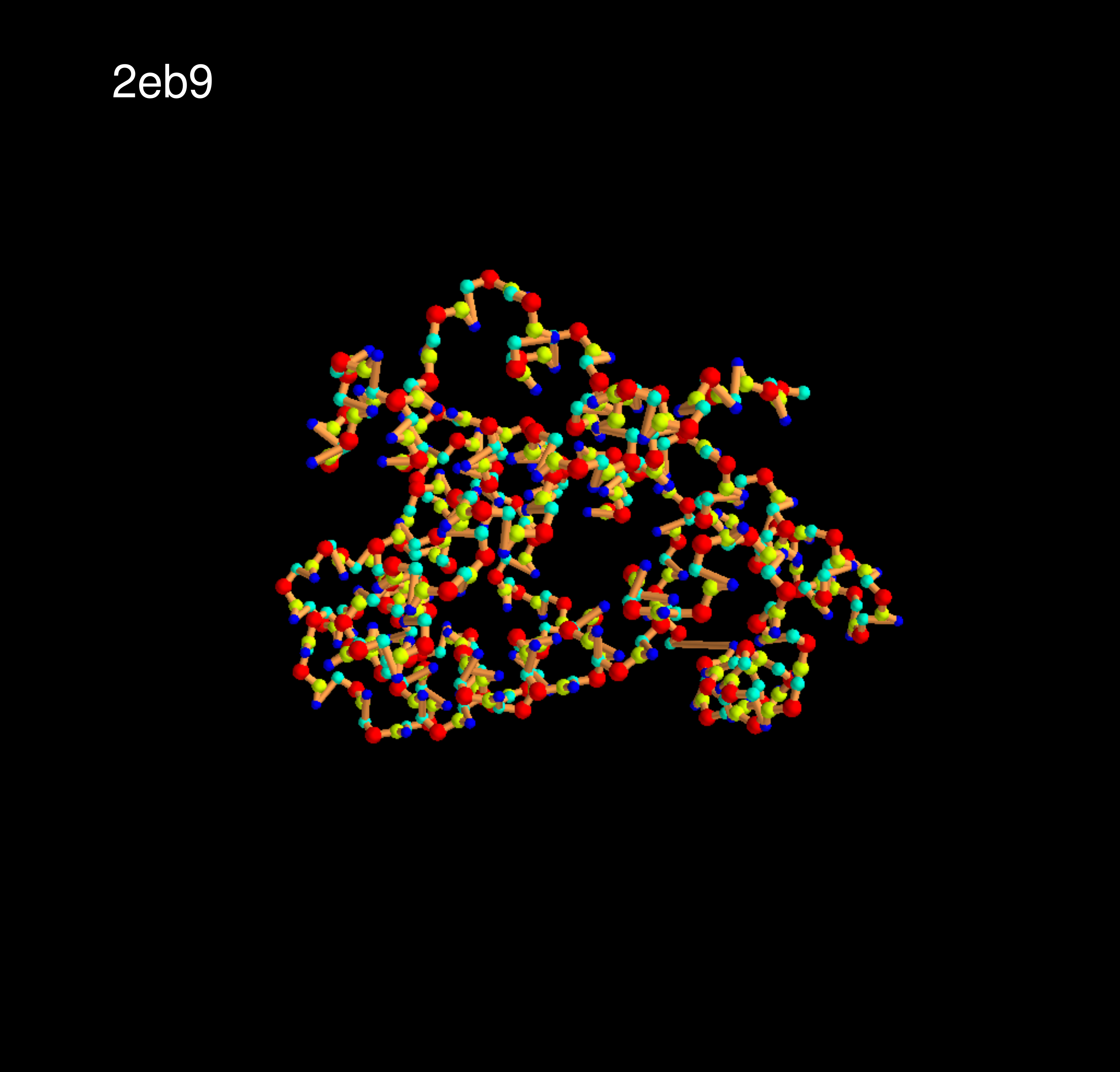}
        \captionsetup{labelformat=empty}
        \caption{}
        \label{supp_fig:pdb_2eb9}
    \end{subfigure}
      \caption{}
      \label{fig:pdb_8}
    \end{figure}

    \begin{figure}[!tbhp]
      \centering
      \begin{subfigure}{.33\textwidth} \centering
        \includegraphics[width=0.9\linewidth]{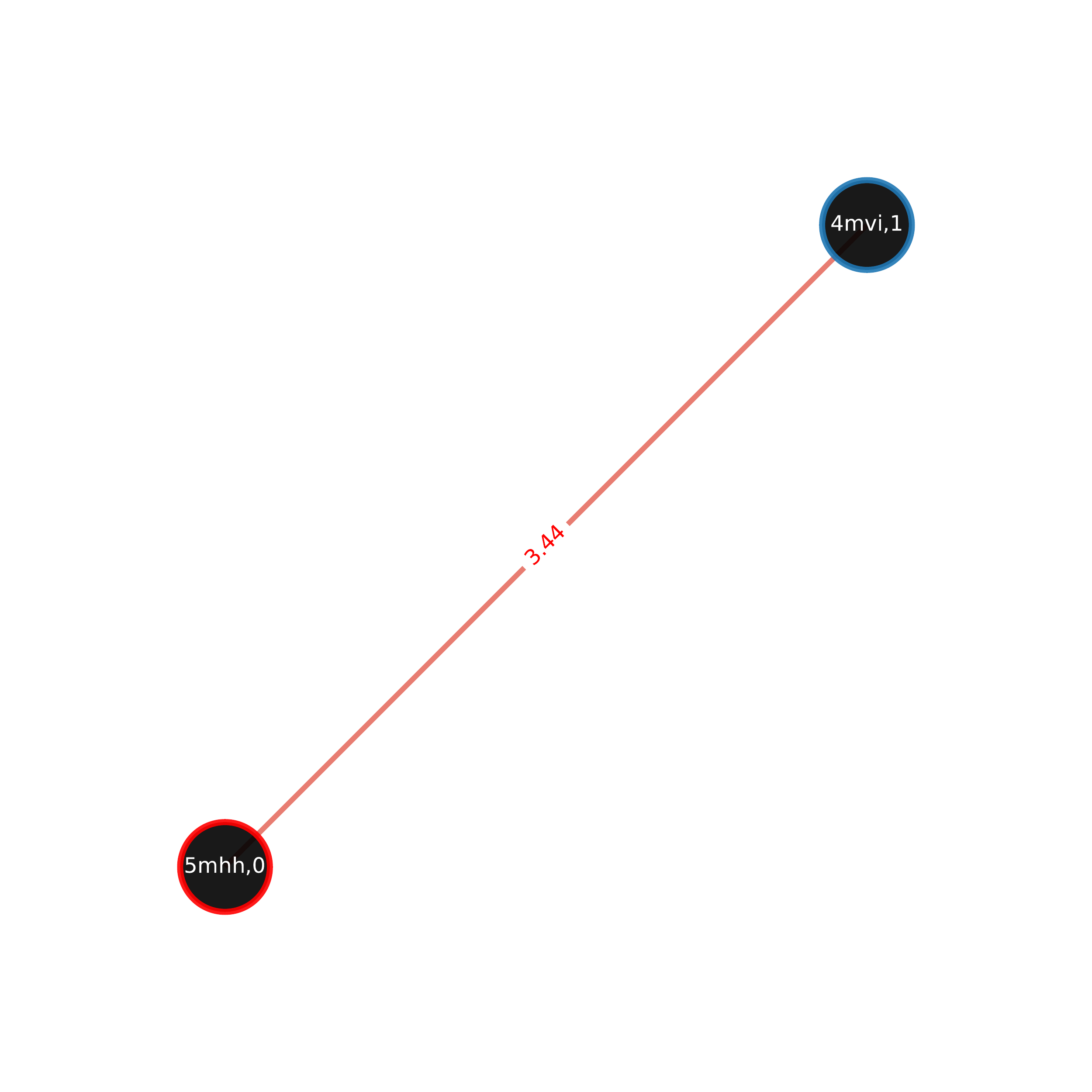}
        \captionsetup{labelformat=empty}
        \caption{}
        \label{supp_fig:pdb_hom_graph_9}
    \end{subfigure}
      \centering
      \begin{subfigure}{.33\textwidth} \centering
        \includegraphics[width=0.9\linewidth]{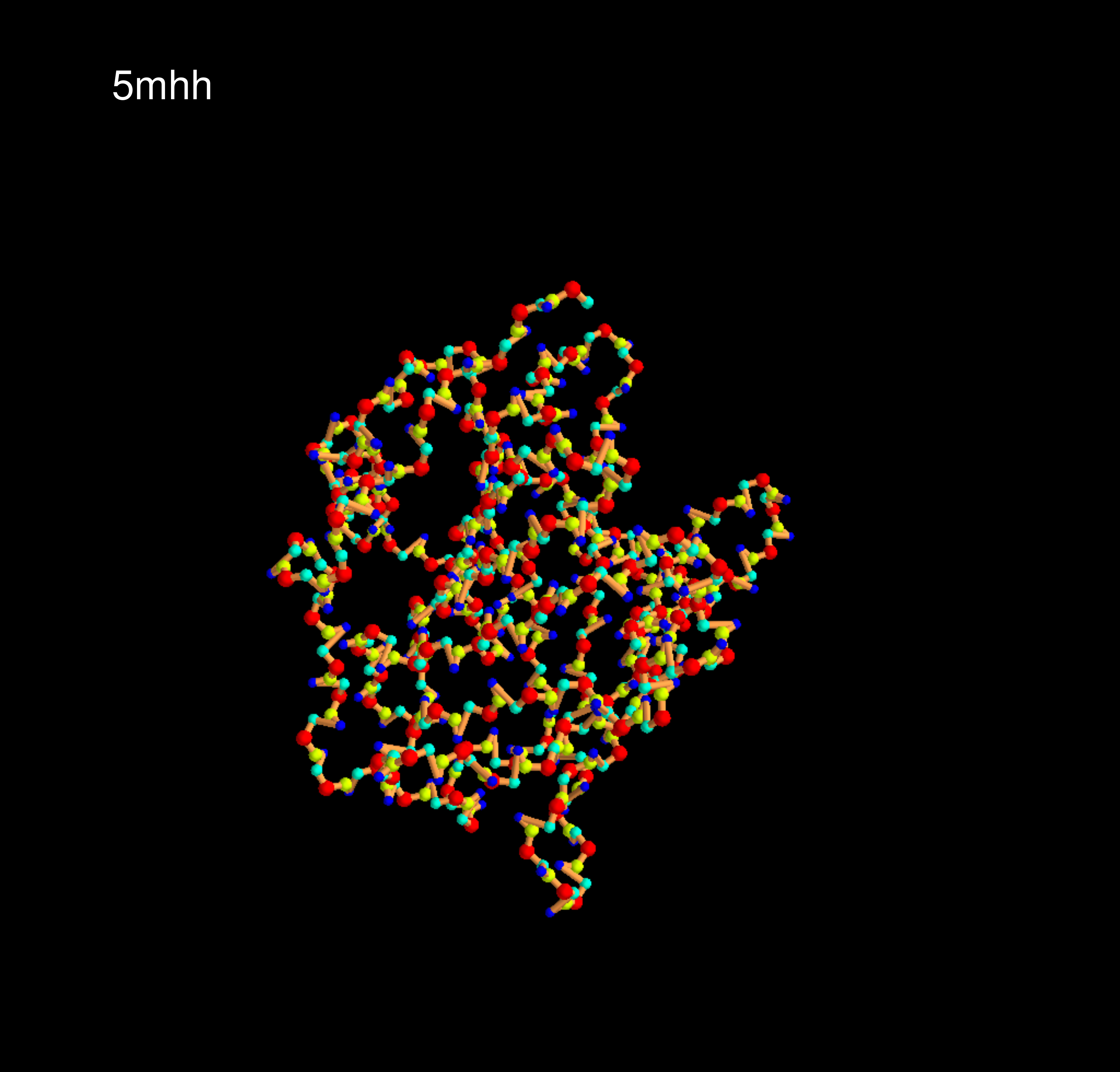}
        \captionsetup{labelformat=empty}
        \caption{}
        \label{supp_fig:pdb_5mhh}
    \end{subfigure}
      \centering
      \begin{subfigure}{.33\textwidth} \centering
        \includegraphics[width=0.9\linewidth]{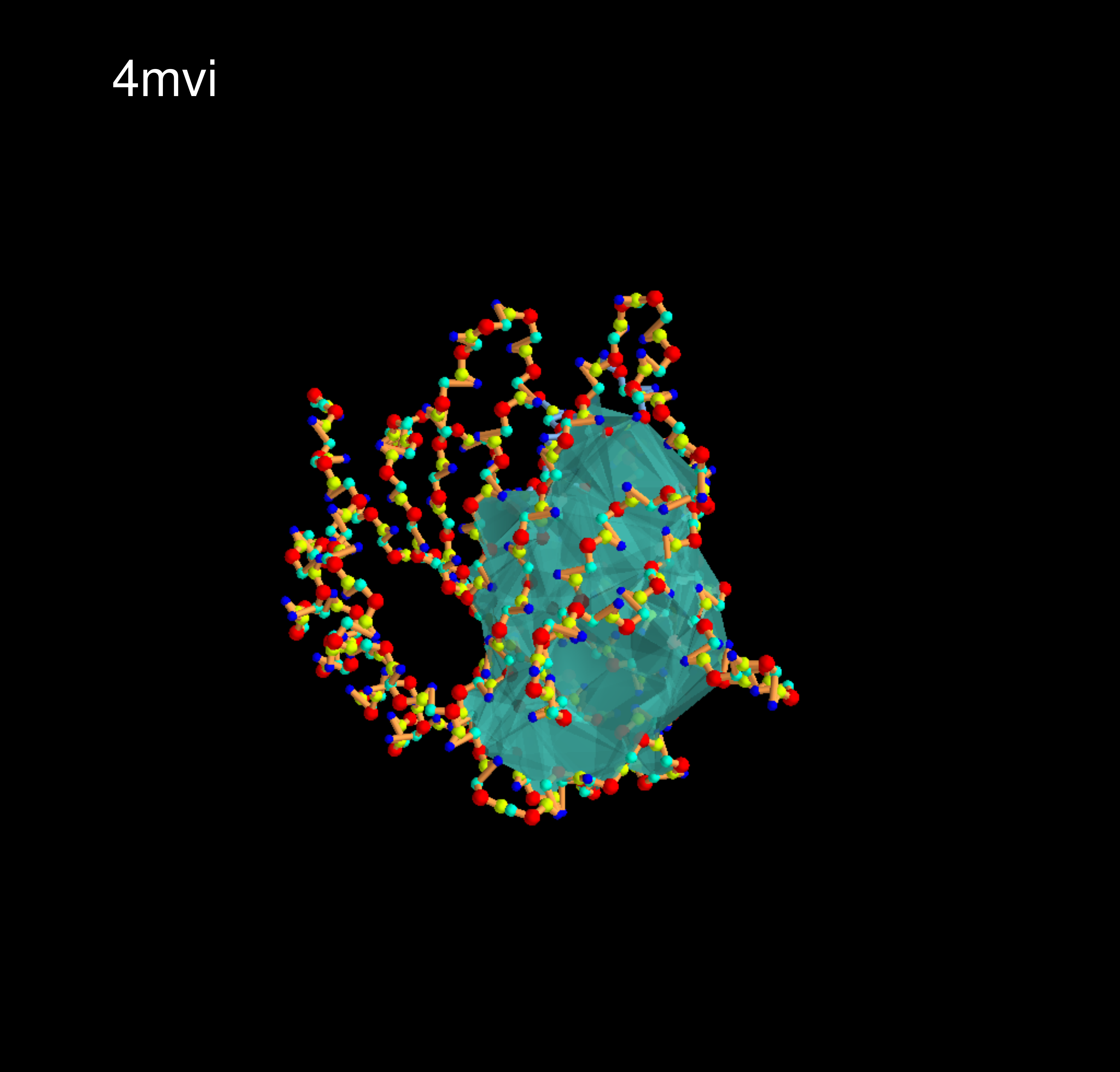}
        \captionsetup{labelformat=empty}
        \caption{}
        \label{supp_fig:pdb_4mvi}
    \end{subfigure}
      \caption{}
      \label{fig:pdb_9}
    \end{figure}

    \begin{figure}[!tbhp]
      \centering
      \begin{subfigure}{.48\textwidth} \centering
        \includegraphics[width=0.9\linewidth]{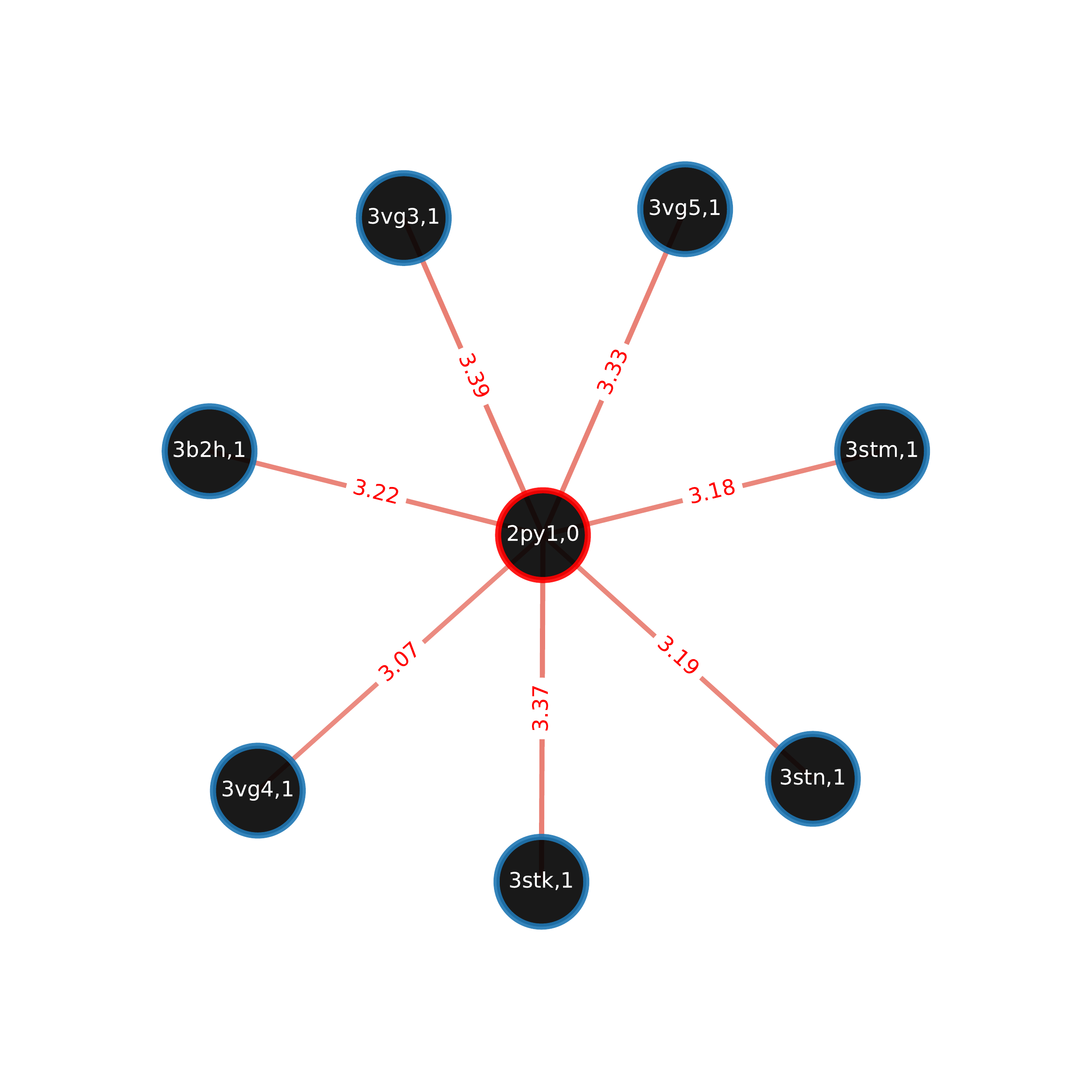}
        \captionsetup{labelformat=empty}
        \caption{}
        \label{supp_fig:pdb_hom_graph_10}
    \end{subfigure}
      \centering
      \begin{subfigure}{.24\textwidth} \centering
        \includegraphics[width=0.9\linewidth]{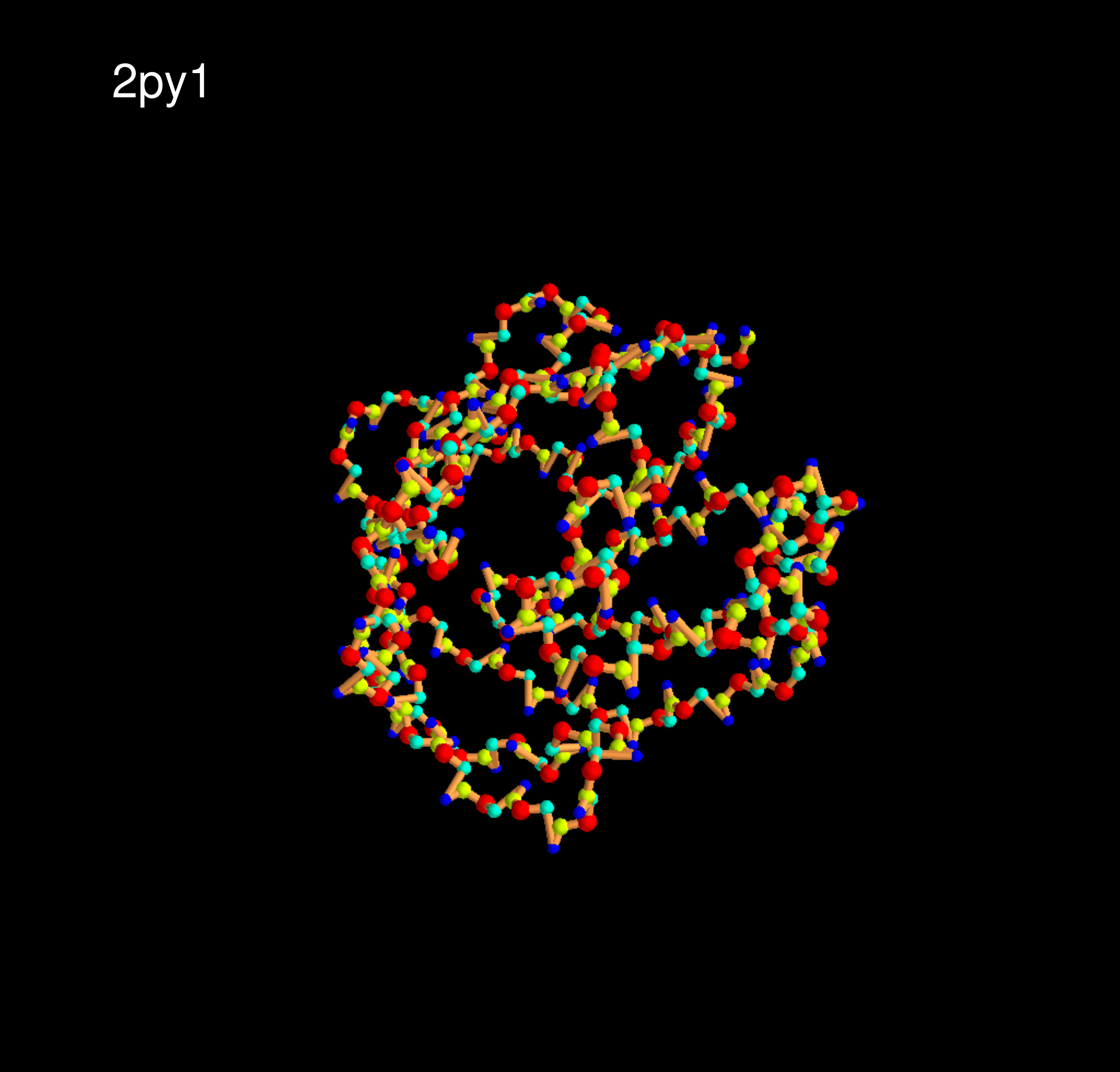}
        \captionsetup{labelformat=empty}
        \caption{}
        \label{supp_fig:pdb_2py1}
    \end{subfigure}
      \centering
      \begin{subfigure}{.24\textwidth} \centering
        \includegraphics[width=0.9\linewidth]{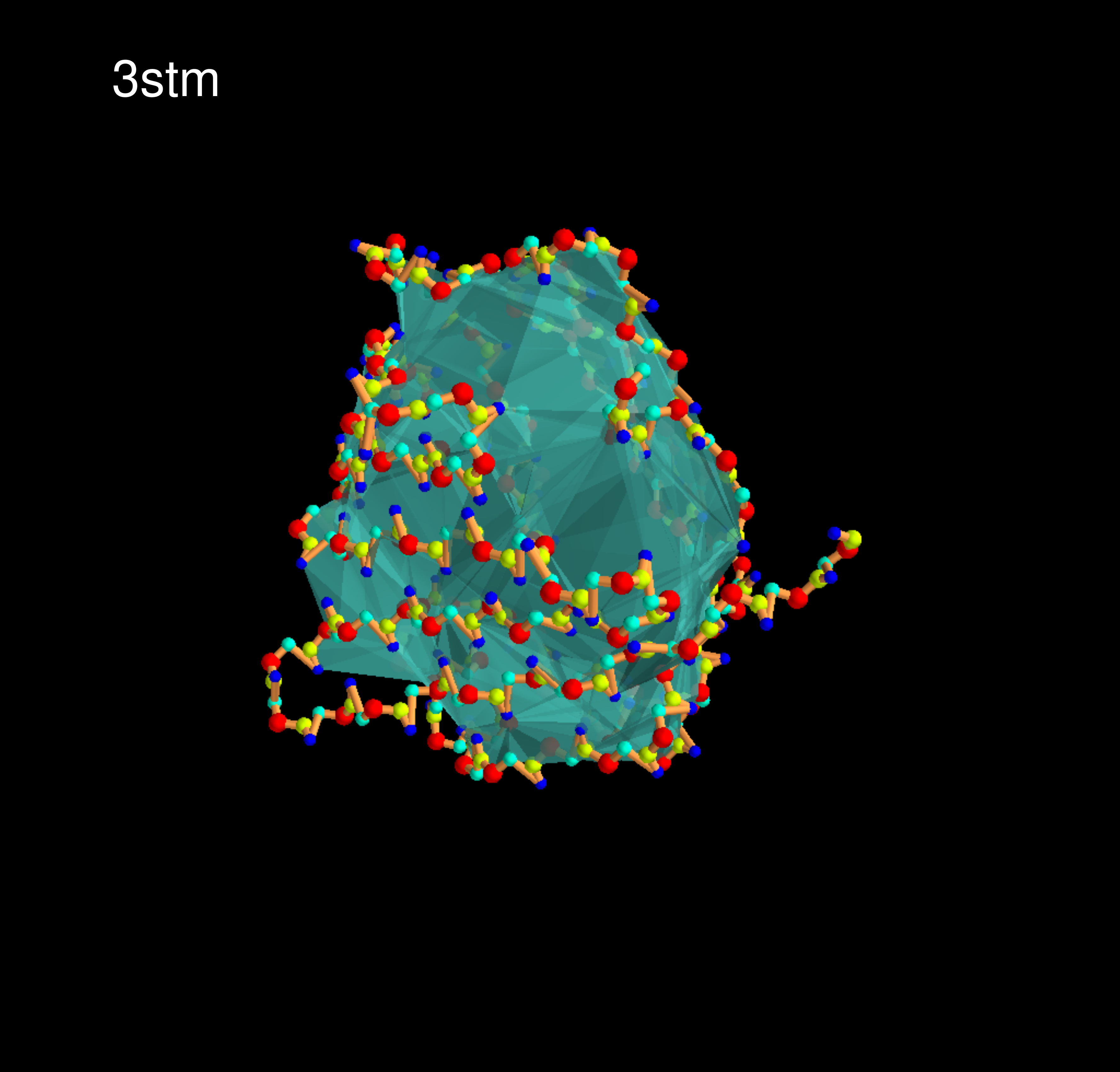}
        \captionsetup{labelformat=empty}
        \caption{}
        \label{supp_fig:pdb_3stm}
    \end{subfigure}
      \centering
      \begin{subfigure}{.24\textwidth} \centering
        \includegraphics[width=0.9\linewidth]{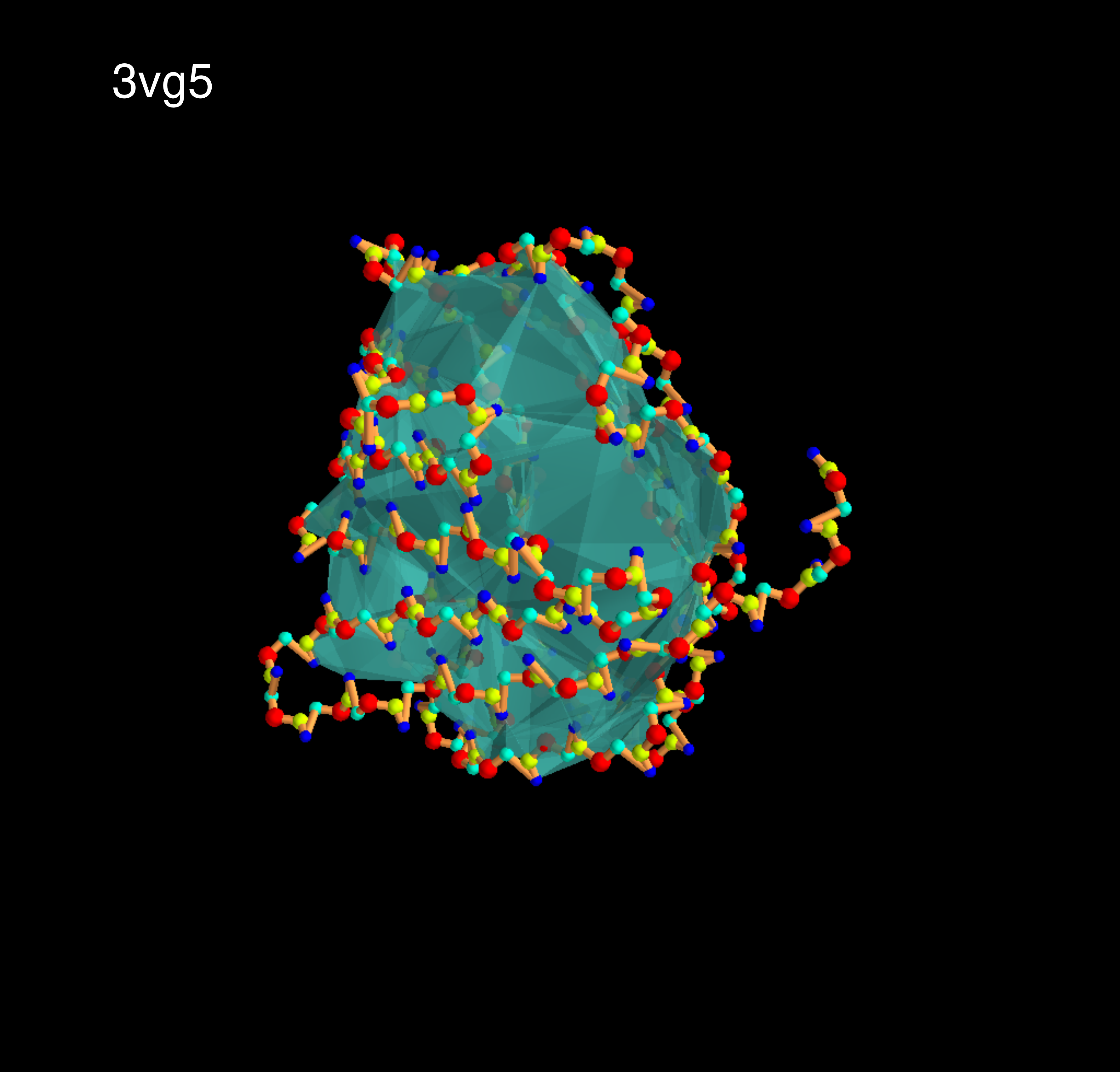}
        \captionsetup{labelformat=empty}
        \caption{}
        \label{supp_fig:pdb_3vg5}
    \end{subfigure}
      \centering
      \begin{subfigure}{.24\textwidth} \centering
        \includegraphics[width=0.9\linewidth]{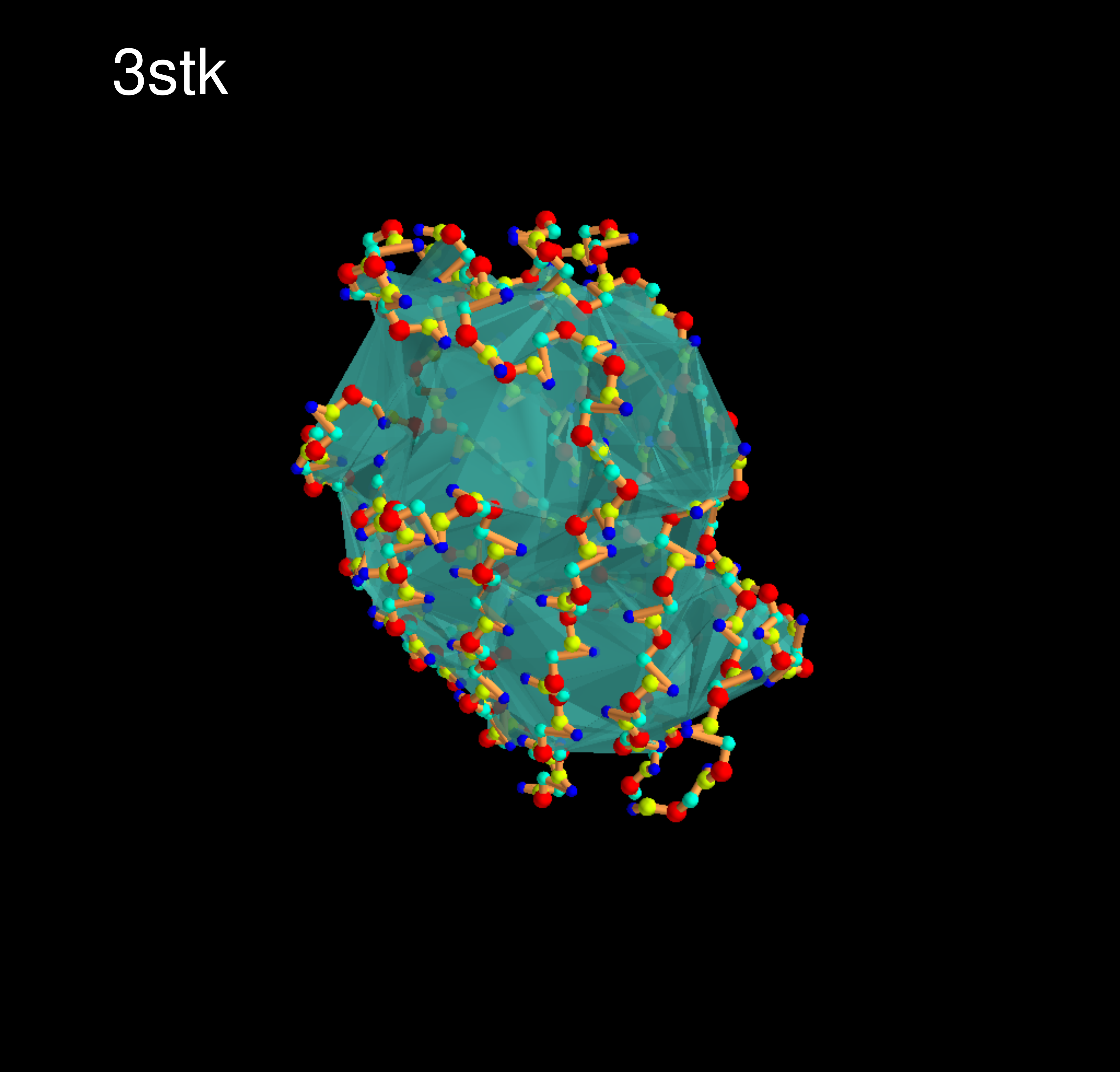}
        \captionsetup{labelformat=empty}
        \caption{}
        \label{supp_fig:pdb_3stk}
    \end{subfigure}
      \centering
      \begin{subfigure}{.24\textwidth} \centering
        \includegraphics[width=0.9\linewidth]{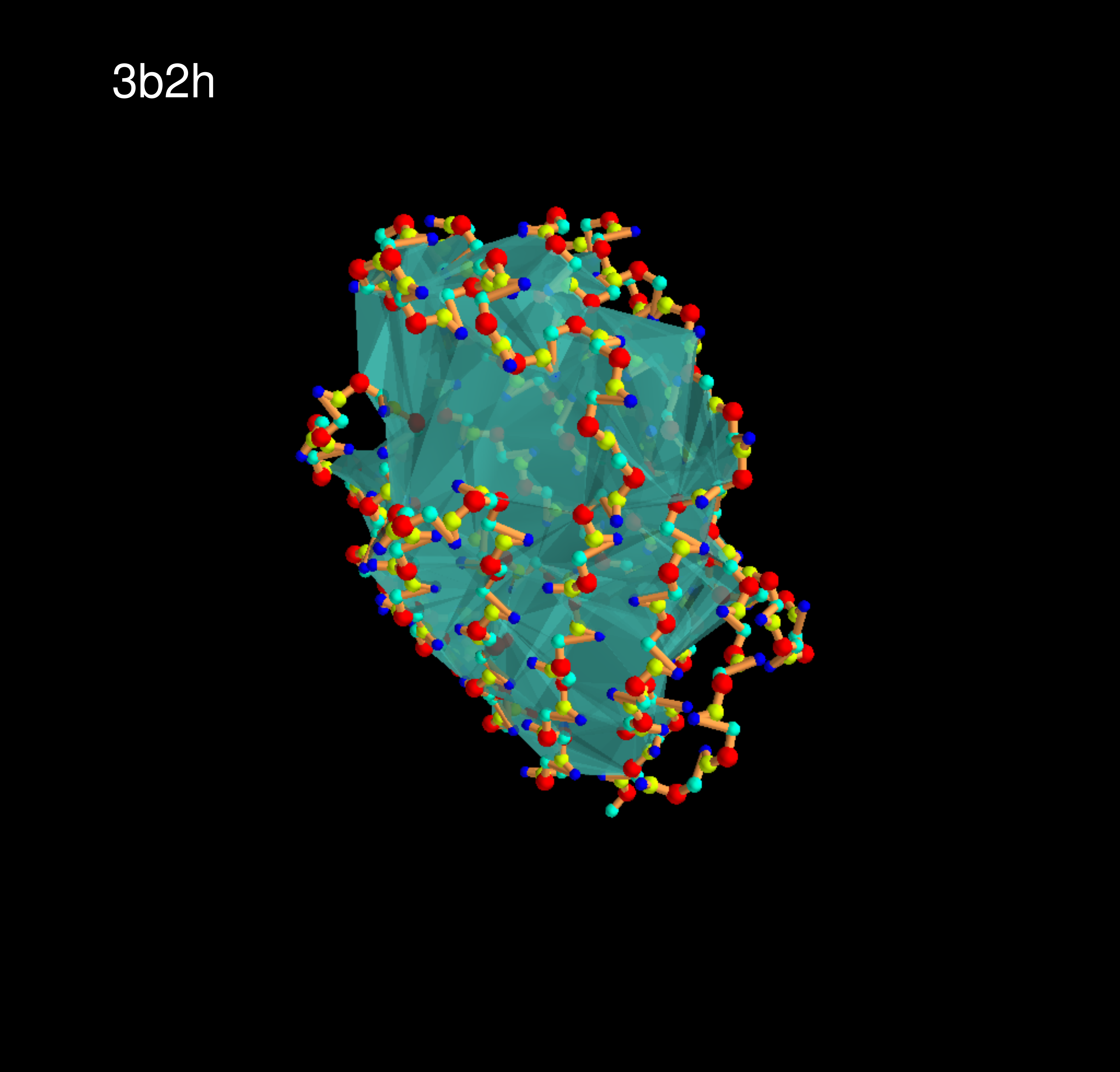}
        \captionsetup{labelformat=empty}
        \caption{}
        \label{supp_fig:pdb_3b2h}
    \end{subfigure}
      \centering
      \begin{subfigure}{.24\textwidth} \centering
        \includegraphics[width=0.9\linewidth]{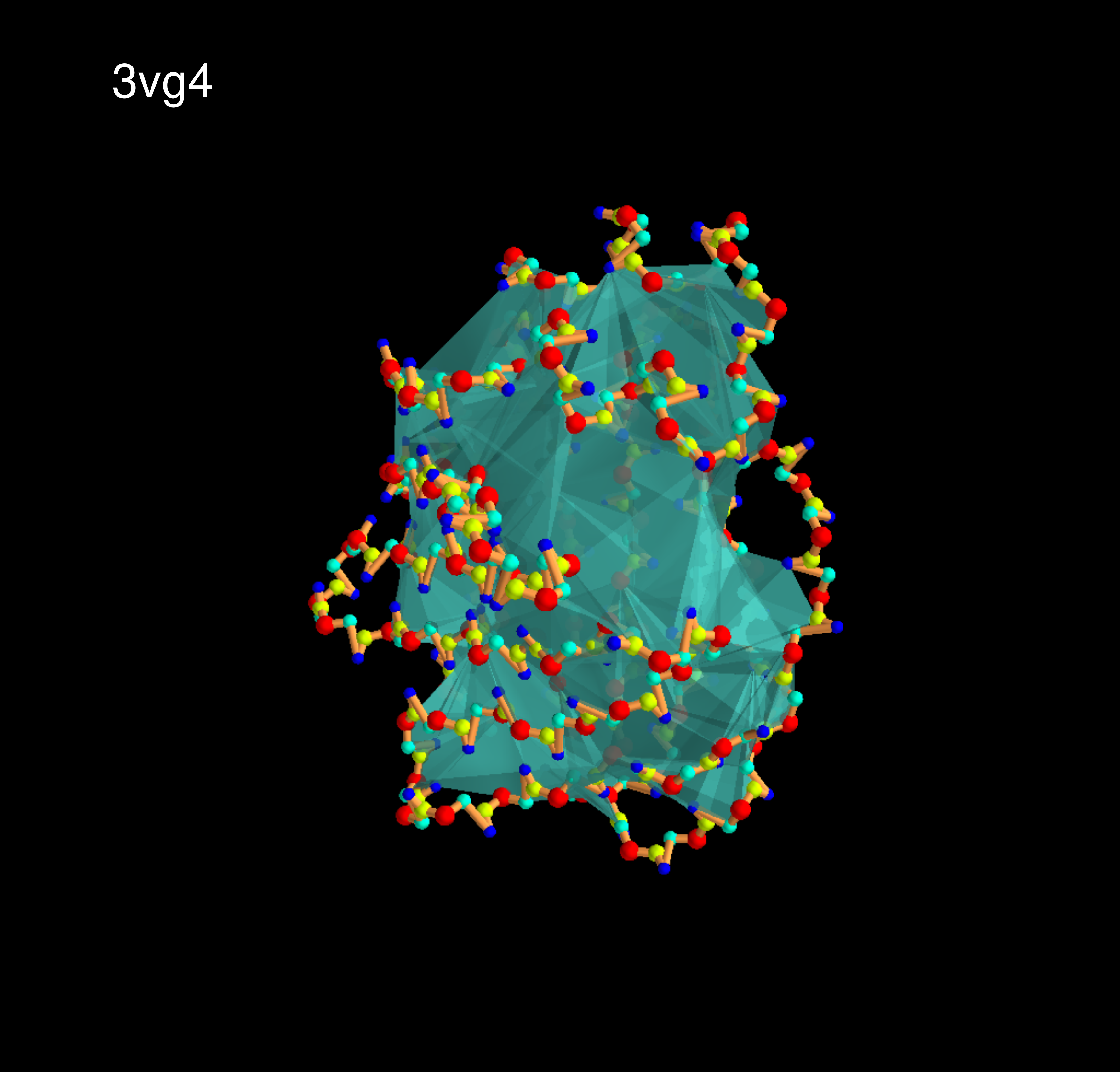}
        \captionsetup{labelformat=empty}
        \caption{}
        \label{supp_fig:pdb_3vg4}
    \end{subfigure}
      \centering
      \begin{subfigure}{.24\textwidth} \centering
        \includegraphics[width=0.9\linewidth]{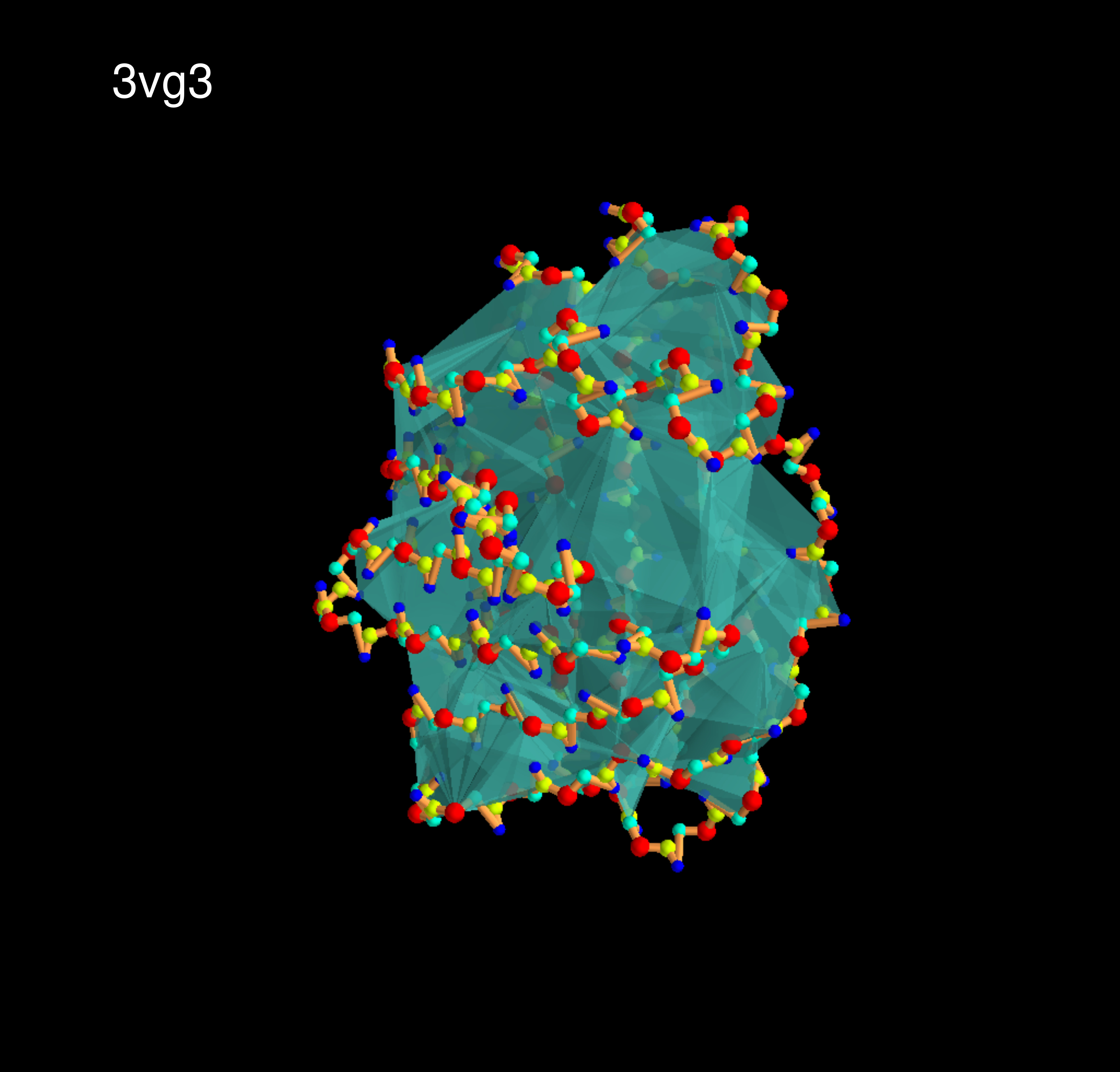}
        \captionsetup{labelformat=empty}
        \caption{}
        \label{supp_fig:pdb_3vg3}
    \end{subfigure}
      \centering
      \begin{subfigure}{.24\textwidth} \centering
        \includegraphics[width=0.9\linewidth]{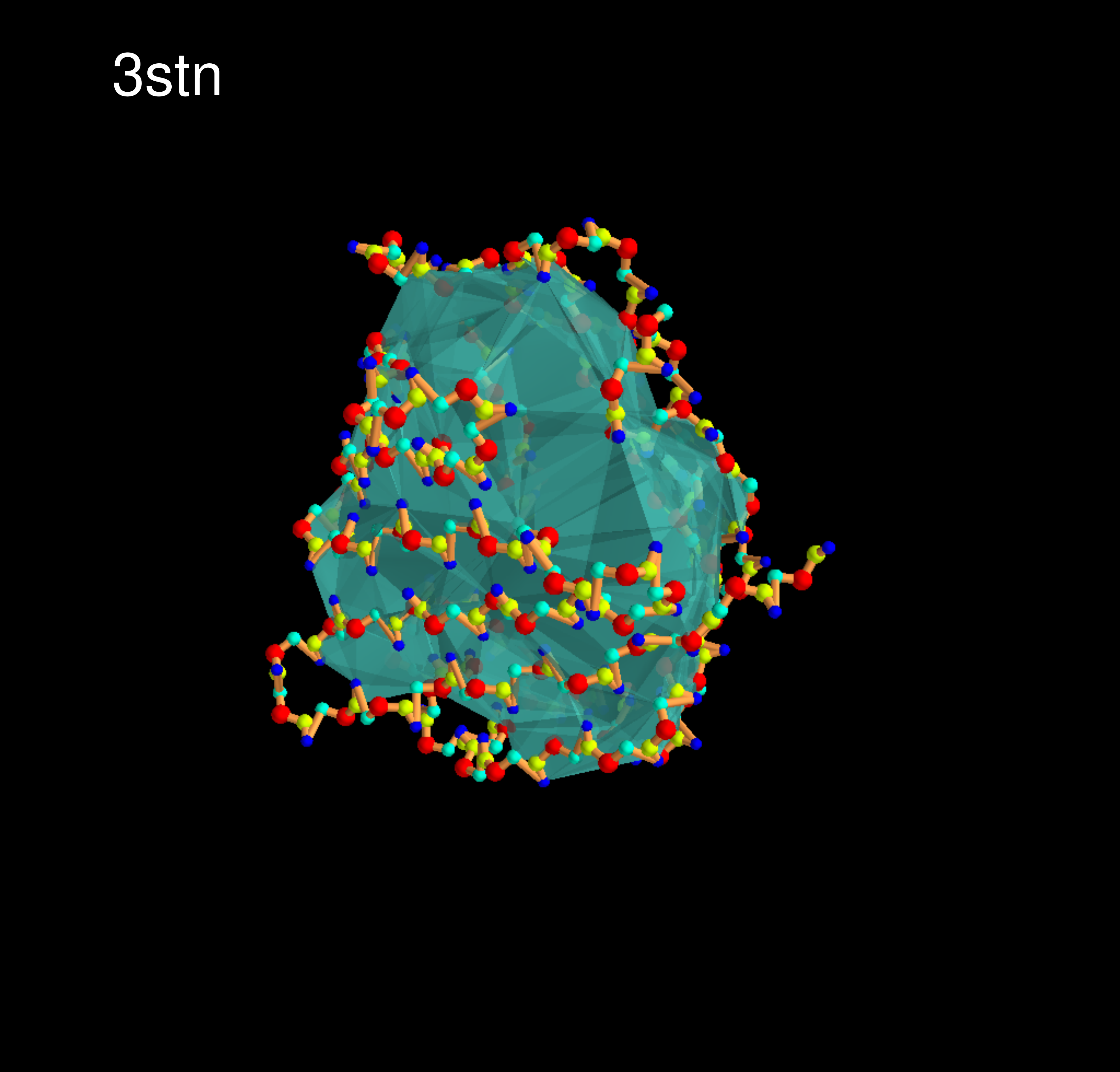}
        \captionsetup{labelformat=empty}
        \caption{}
        \label{supp_fig:pdb_3stn}
    \end{subfigure}
      \caption{}
      \label{fig:pdb_10}
    \end{figure}

    \begin{figure}[!tbhp]
      \centering
      \begin{subfigure}{.33\textwidth} \centering
        \includegraphics[width=0.9\linewidth]{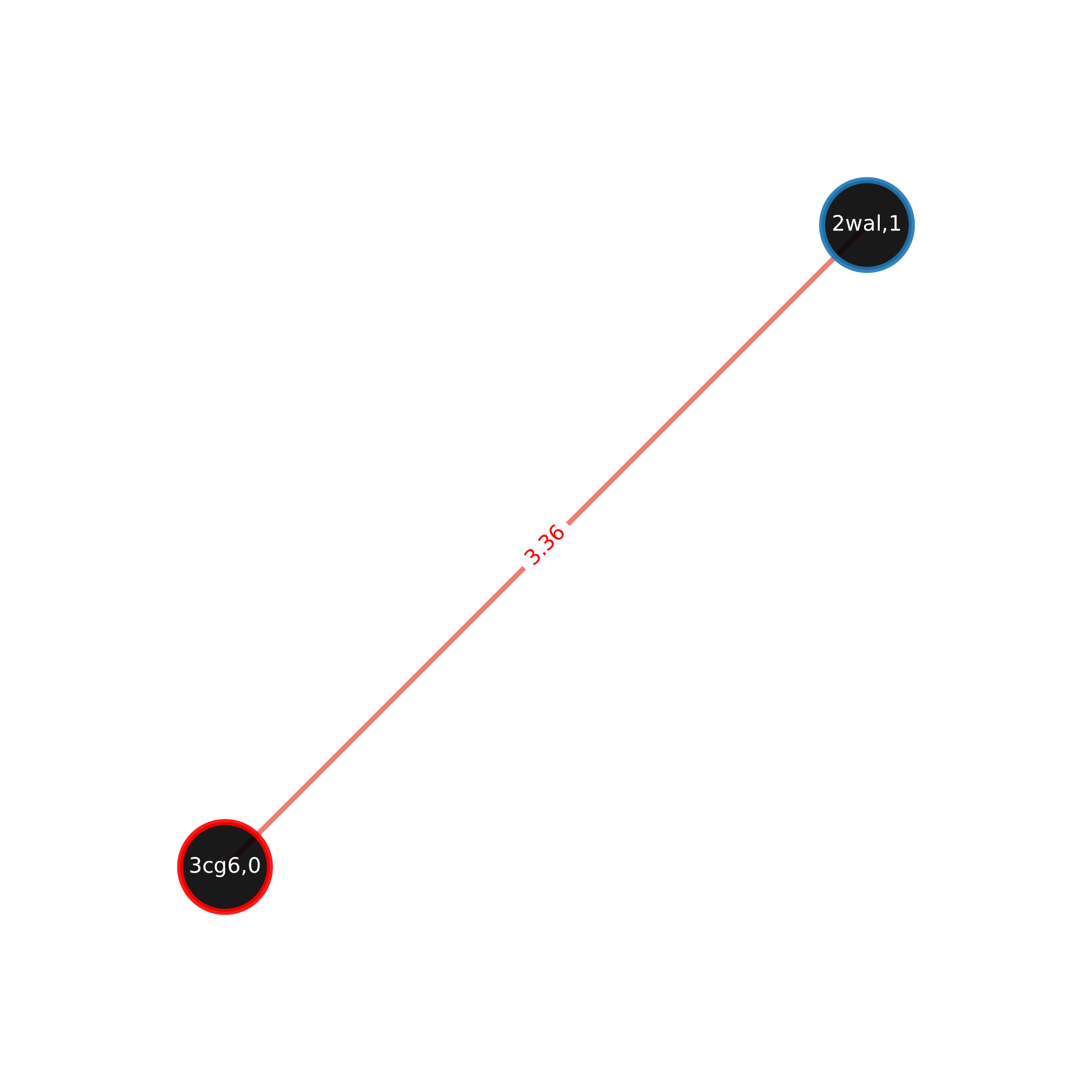}
        \captionsetup{labelformat=empty}
        \caption{}
        \label{supp_fig:pdb_hom_graph_11}
    \end{subfigure}
      \centering
      \begin{subfigure}{.33\textwidth} \centering
        \includegraphics[width=0.9\linewidth]{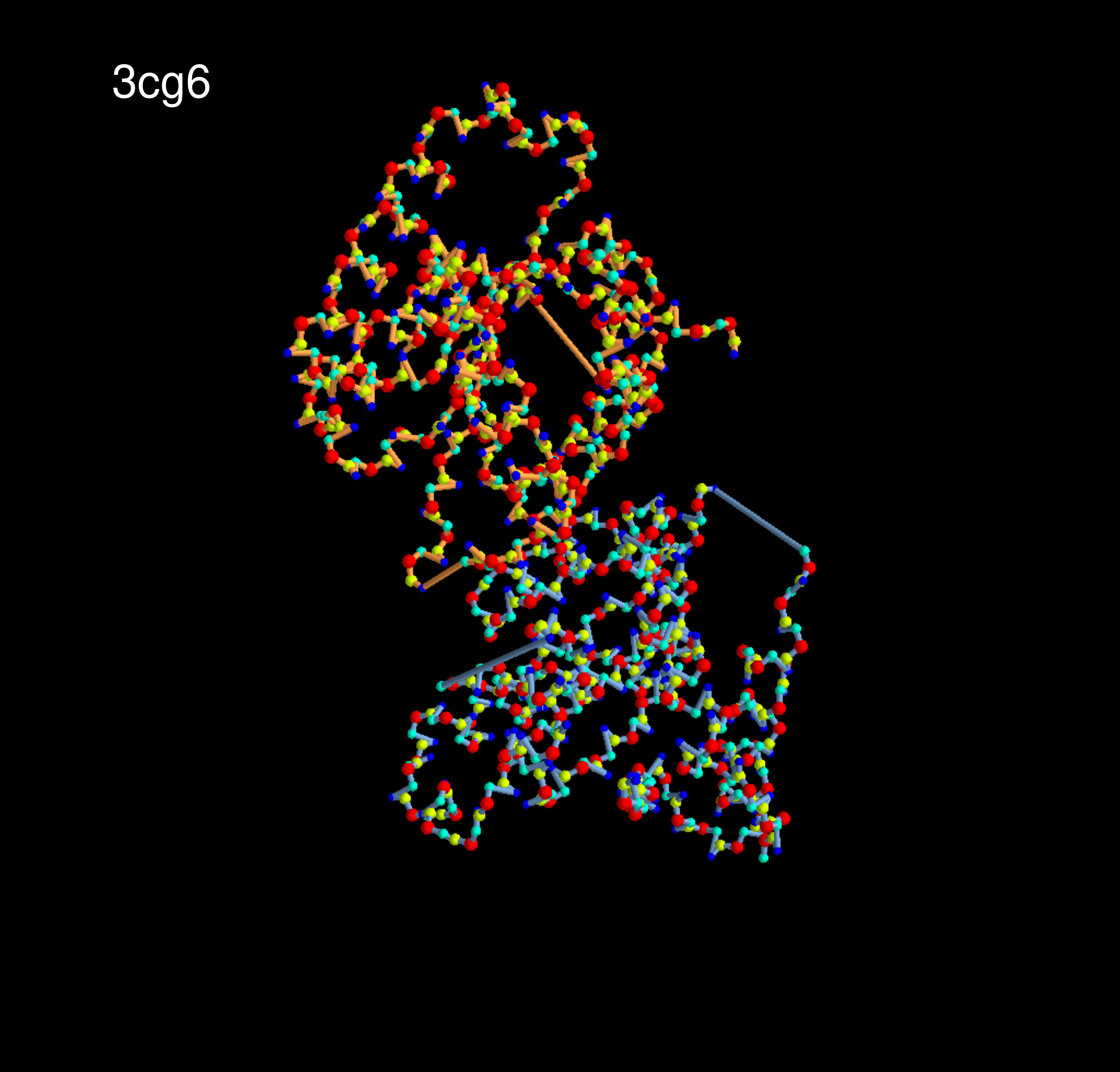}
        \captionsetup{labelformat=empty}
        \caption{}
        \label{supp_fig:pdb_2py1}
    \end{subfigure}
      \centering
      \begin{subfigure}{.33\textwidth} \centering
        \includegraphics[width=0.9\linewidth]{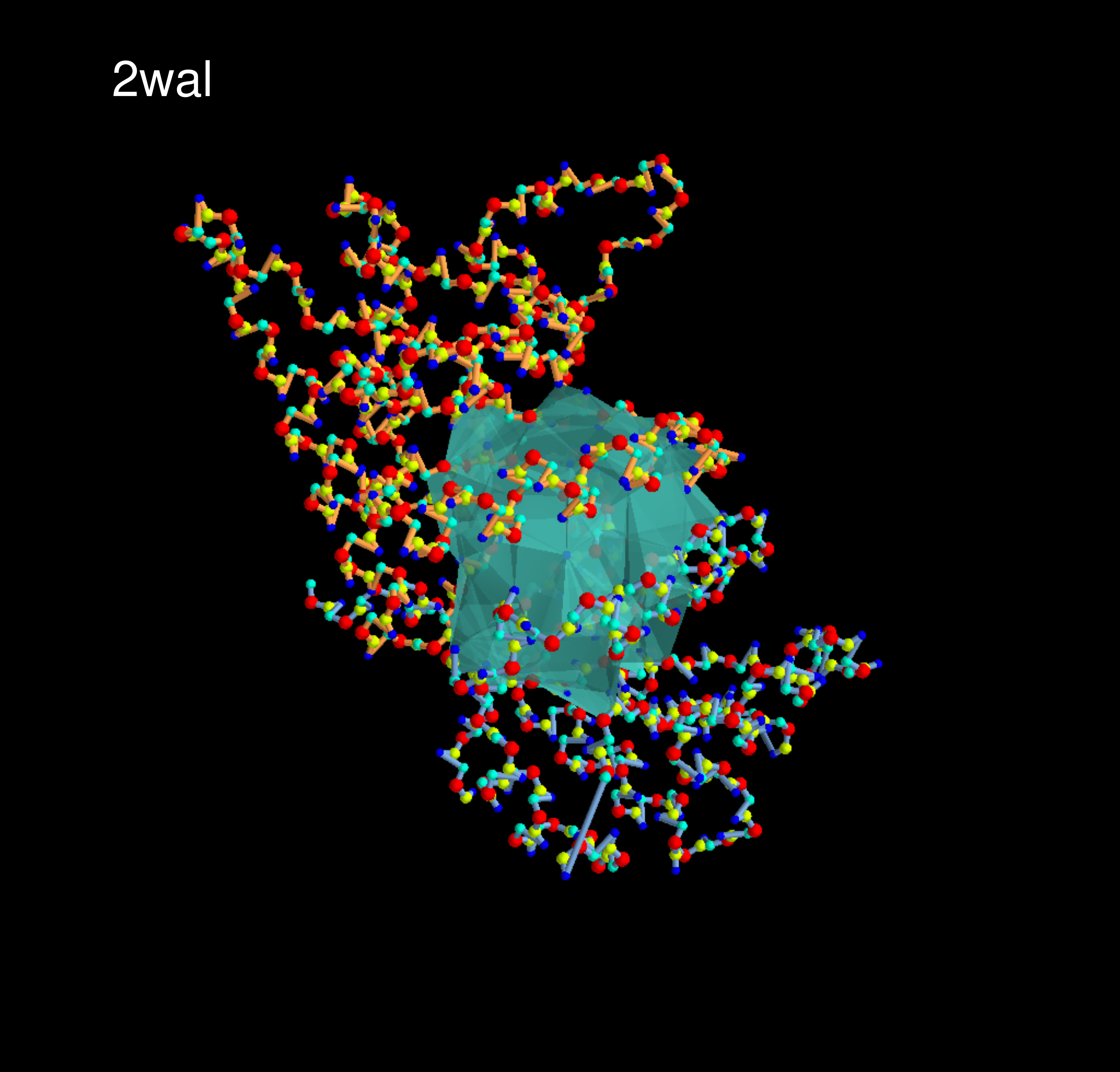}
        \captionsetup{labelformat=empty}
        \caption{}
        \label{supp_fig:pdb_3stm}
    \end{subfigure}
      \caption{}
      \label{fig:pdb_11}
    \end{figure}

    \begin{figure}[!tbhp]
      \centering
      \begin{subfigure}{.33\textwidth} \centering
        \includegraphics[width=0.9\linewidth]{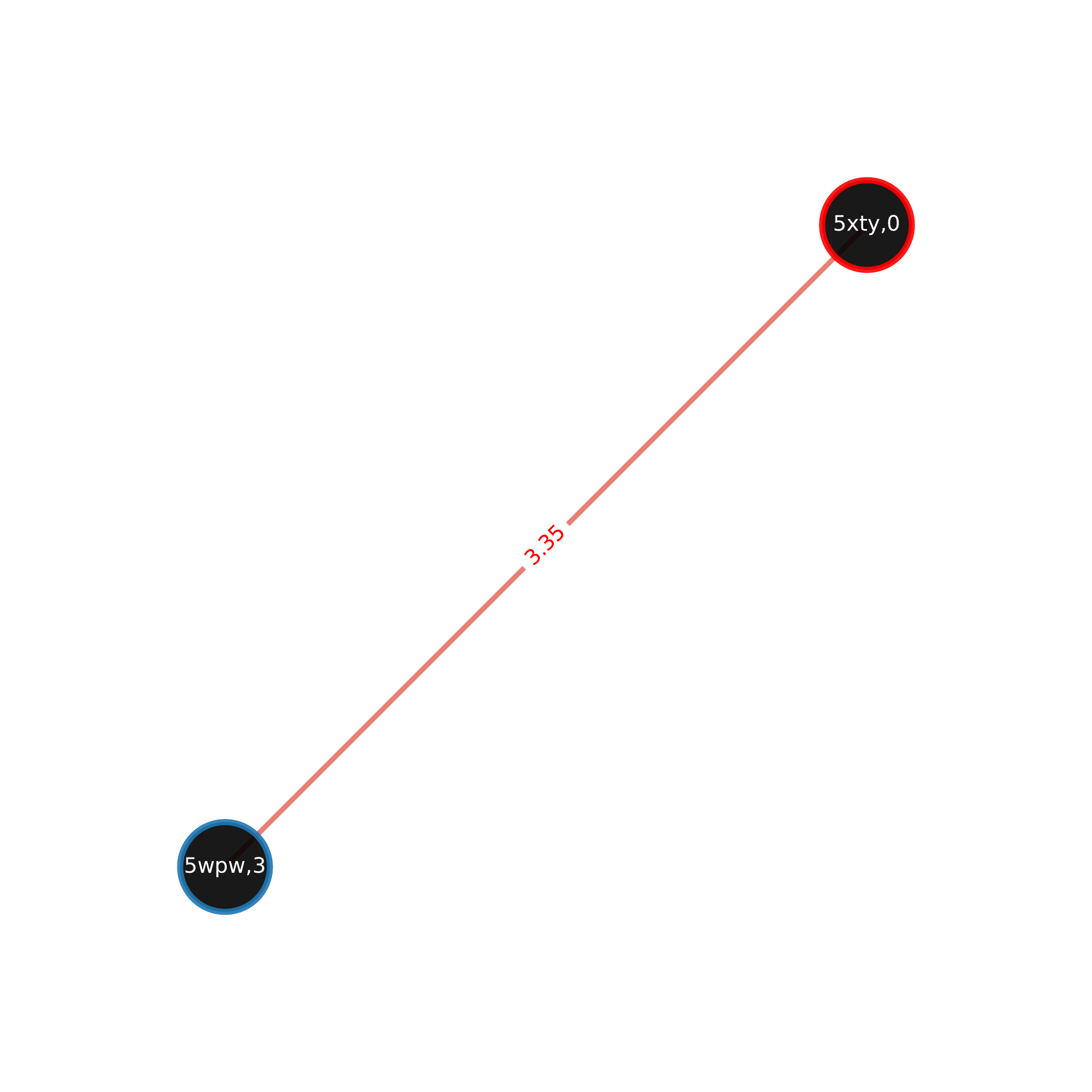}
        \captionsetup{labelformat=empty}
        \caption{}
        \label{supp_fig:pdb_hom_graph_12}
    \end{subfigure}
      \centering
      \begin{subfigure}{.33\textwidth} \centering
        \includegraphics[width=0.9\linewidth]{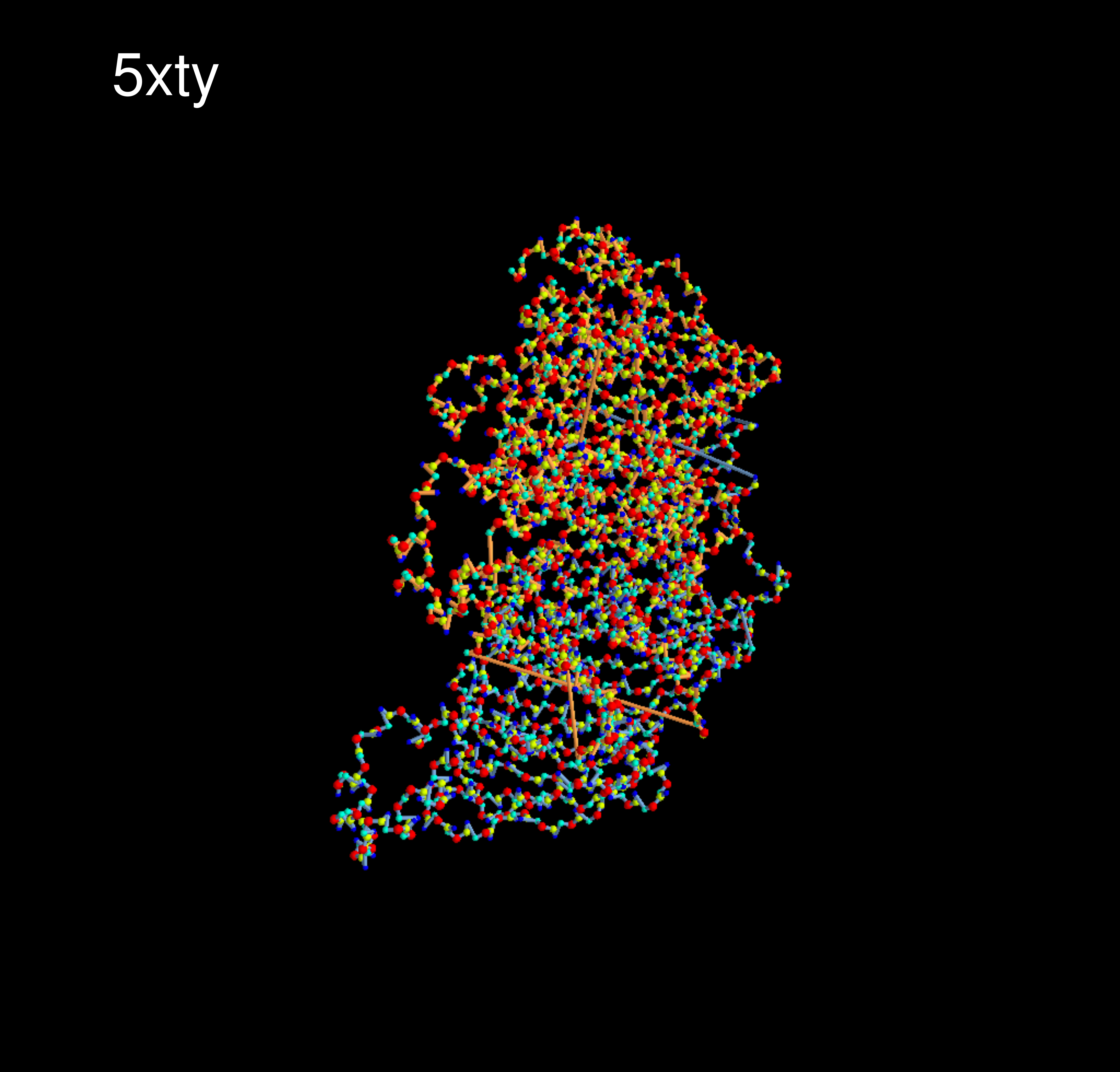}
        \captionsetup{labelformat=empty}
        \caption{}
        \label{supp_fig:pdb_5xty}
    \end{subfigure}
      \centering
      \begin{subfigure}{.33\textwidth} \centering
        \includegraphics[width=0.9\linewidth]{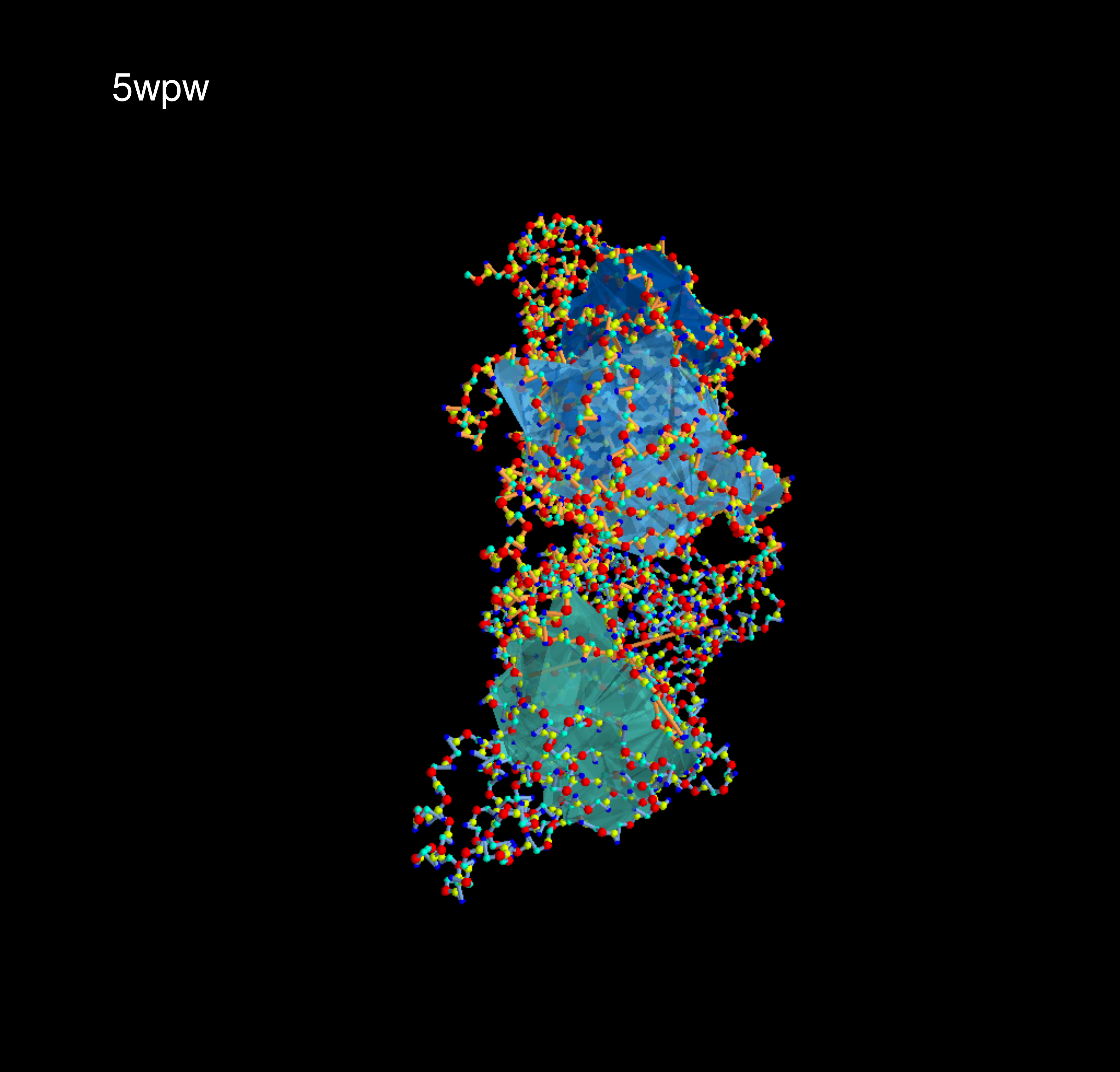}
        \captionsetup{labelformat=empty}
        \caption{}
        \label{supp_fig:pdb_5wpw}
    \end{subfigure}
      \caption{}
      \label{fig:pdb_12}
    \end{figure}

    \begin{figure}[!tbhp]
      \centering
      \begin{subfigure}{.33\textwidth} \centering
        \includegraphics[width=0.9\linewidth]{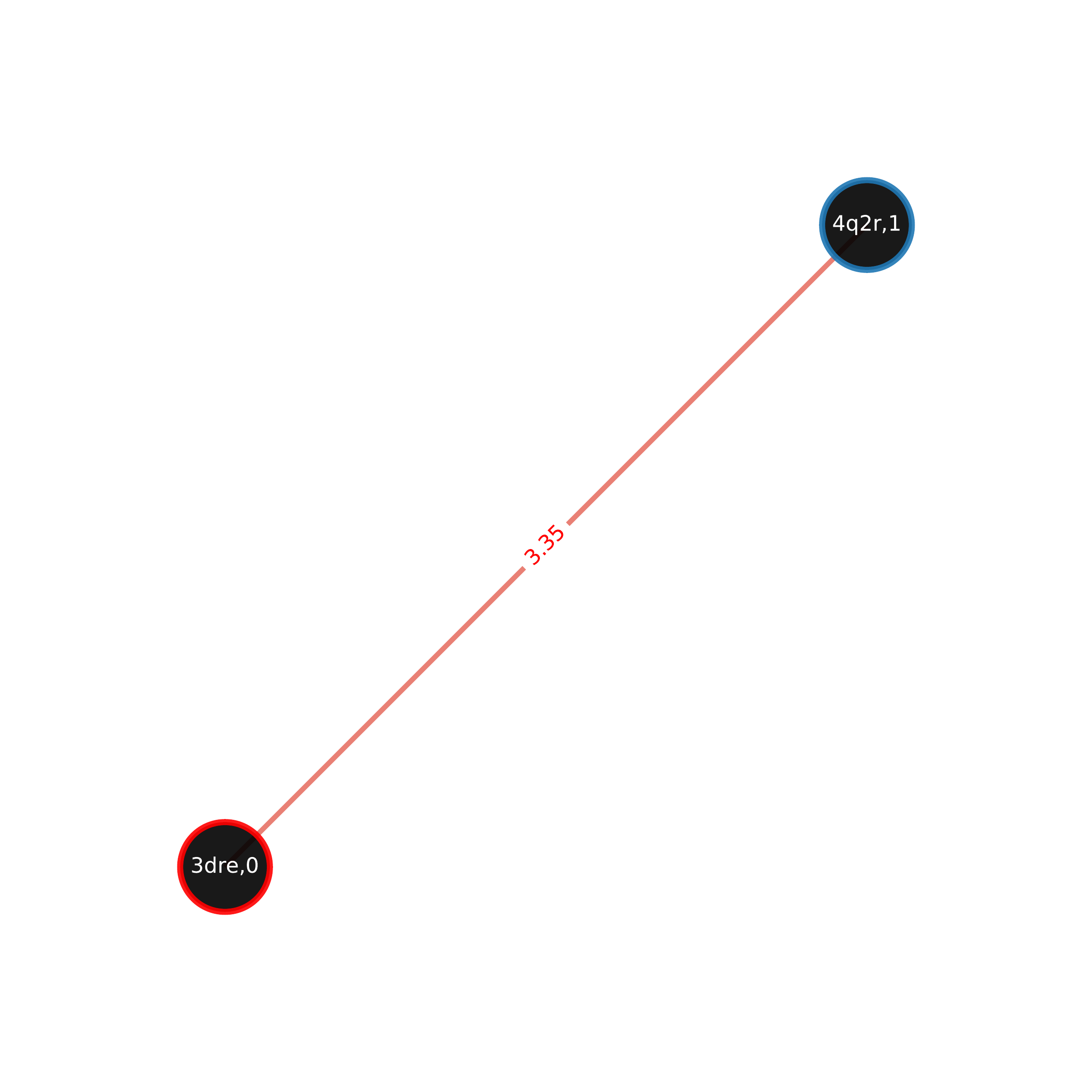}
        \captionsetup{labelformat=empty}
        \caption{}
        \label{supp_fig:pdb_hom_graph_13}
    \end{subfigure}
      \centering
      \begin{subfigure}{.33\textwidth} \centering
        \includegraphics[width=0.9\linewidth]{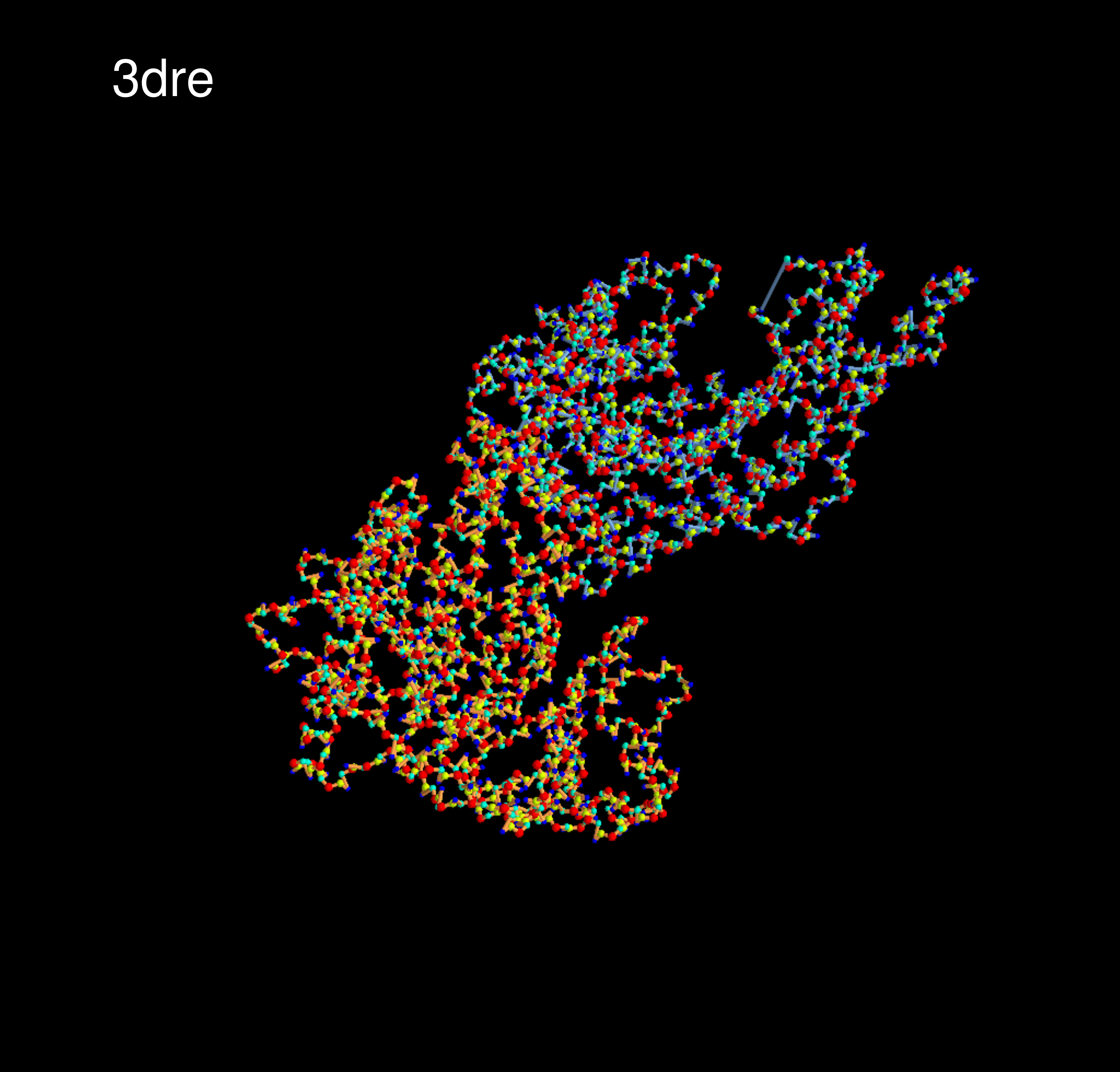}
        \captionsetup{labelformat=empty}
        \caption{}
        \label{supp_fig:pdb_3dre}
    \end{subfigure}
      \centering
      \begin{subfigure}{.33\textwidth} \centering
        \includegraphics[width=0.9\linewidth]{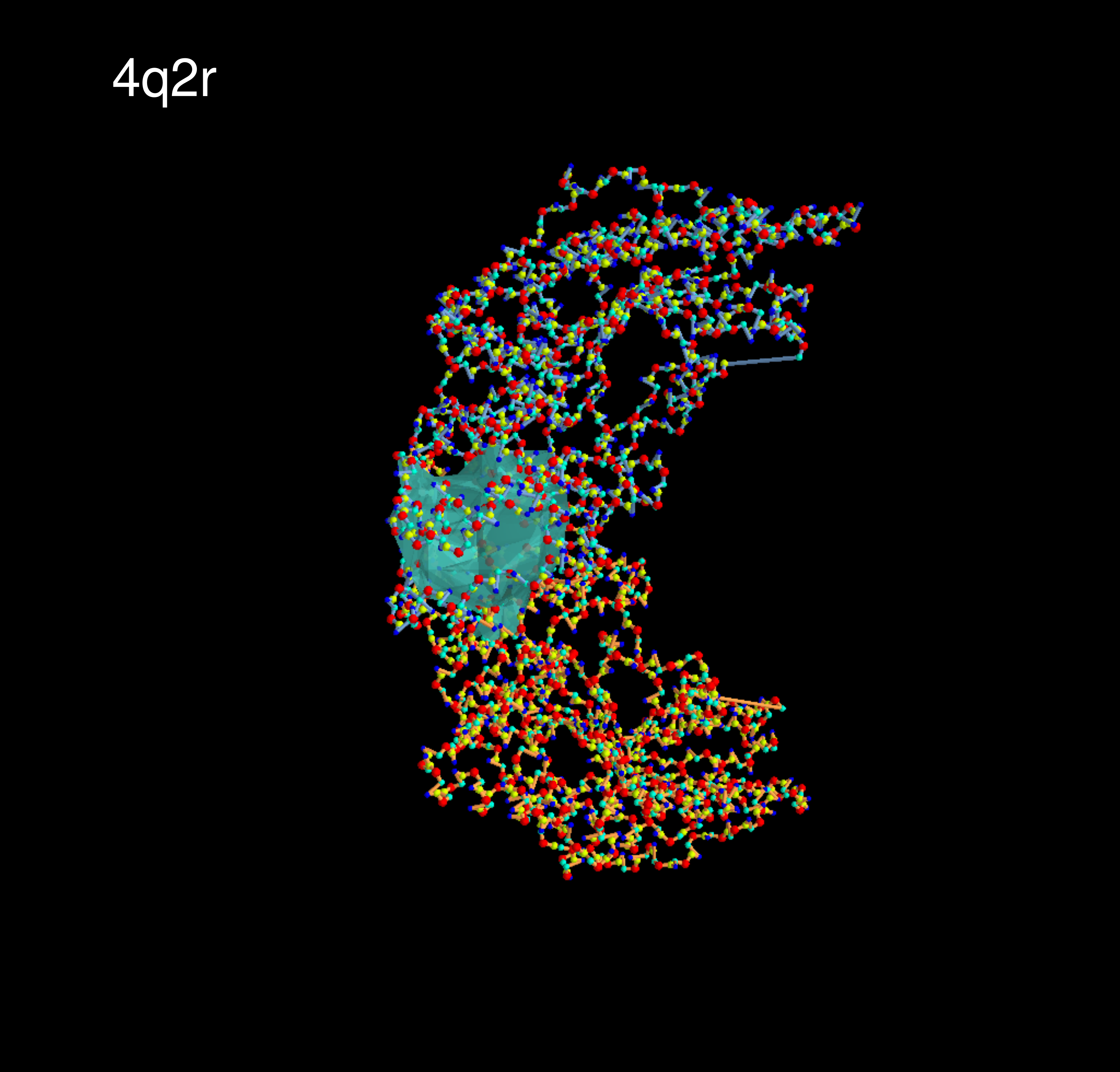}
        \captionsetup{labelformat=empty}
        \caption{}
        \label{supp_fig:pdb_4q2r}
    \end{subfigure}
      \caption{}
      \label{fig:pdb_13}
    \end{figure}

    \begin{figure}[!tbhp]
      \centering
      \begin{subfigure}{.33\textwidth} \centering
        \includegraphics[width=0.9\linewidth]{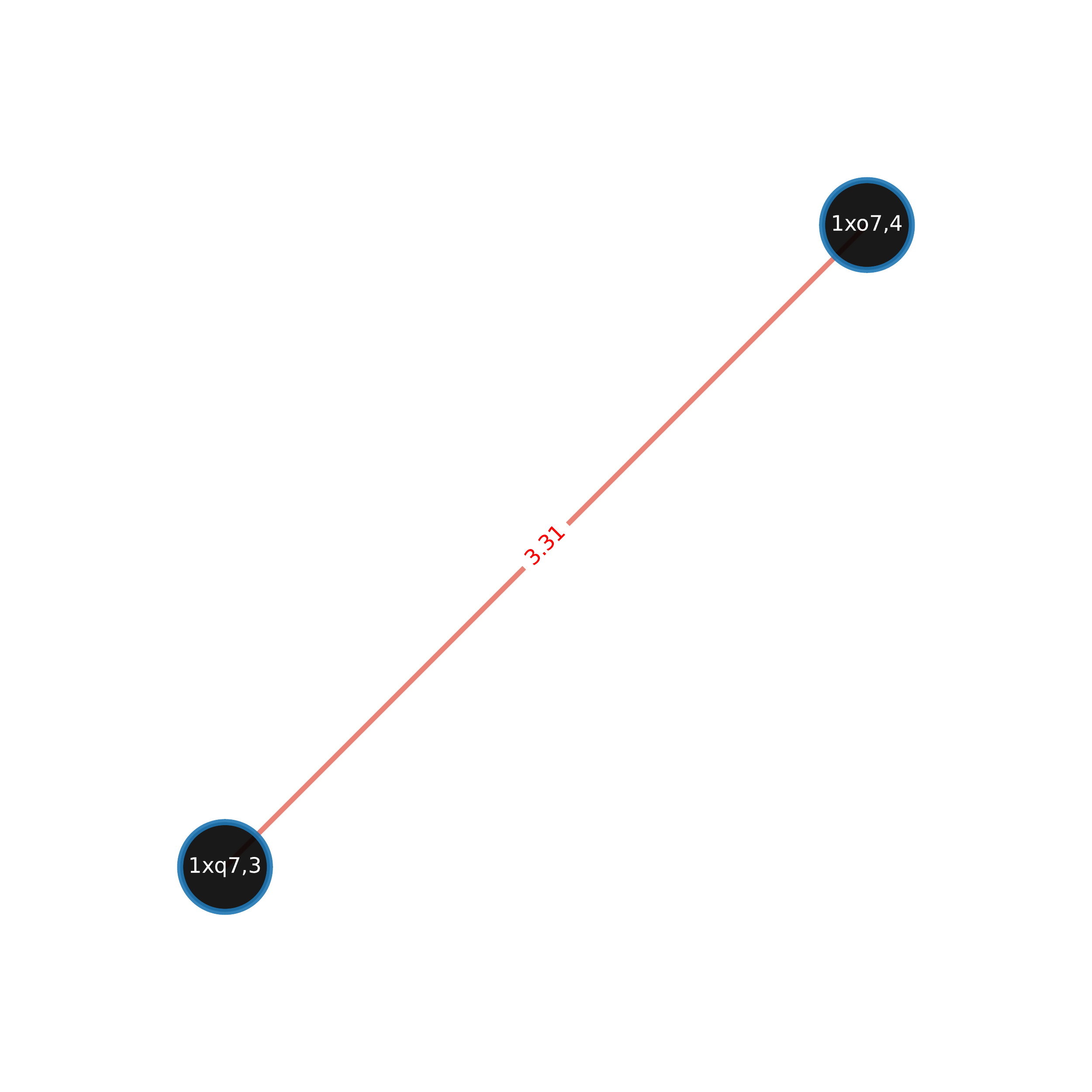}
        \captionsetup{labelformat=empty}
        \caption{}
        \label{supp_fig:pdb_hom_graph_14}
    \end{subfigure}
      \centering
      \begin{subfigure}{.33\textwidth} \centering
        \includegraphics[width=0.9\linewidth]{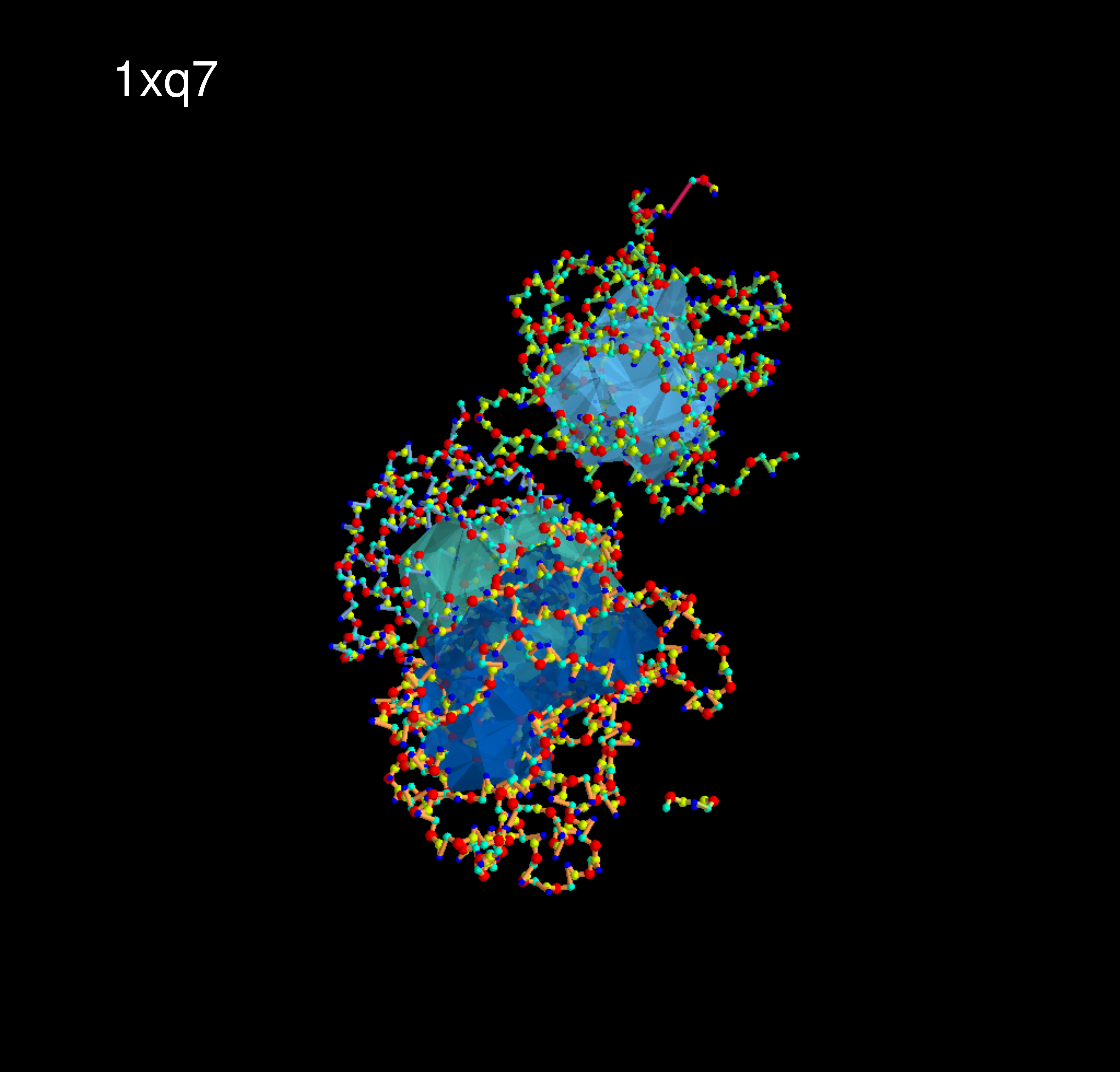}
        \captionsetup{labelformat=empty}
        \caption{}
        \label{supp_fig:pdb_1xq7}
    \end{subfigure}
      \centering
      \begin{subfigure}{.33\textwidth} \centering
        \includegraphics[width=0.9\linewidth]{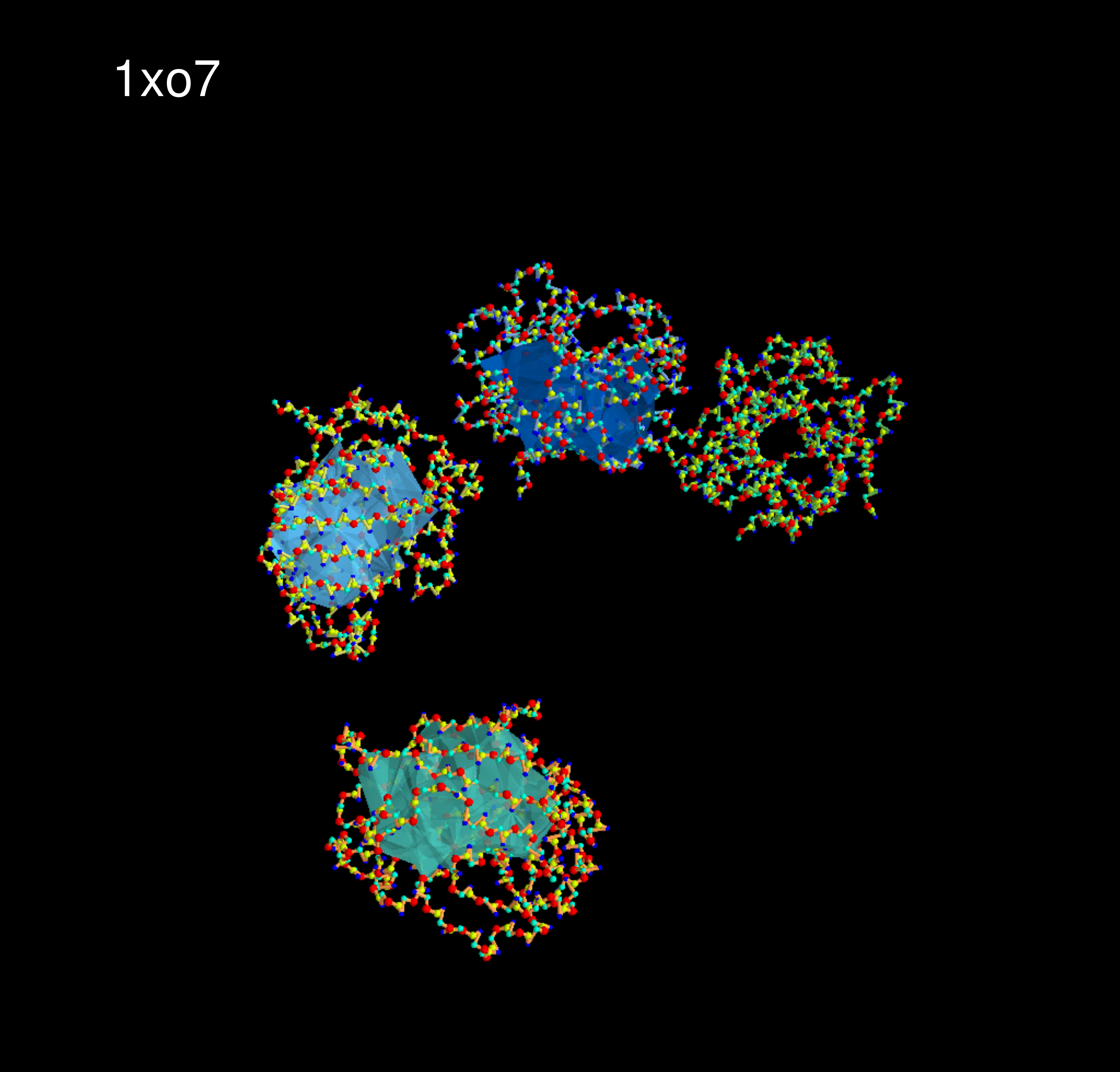}
        \captionsetup{labelformat=empty}
        \caption{}
        \label{supp_fig:pdb_1xo7}
    \end{subfigure}
      \caption{}
      \label{fig:pdb_14}
    \end{figure}

    \begin{figure}[!tbhp]
      \centering
      \begin{subfigure}{.48\textwidth} \centering
        \includegraphics[width=0.9\linewidth]{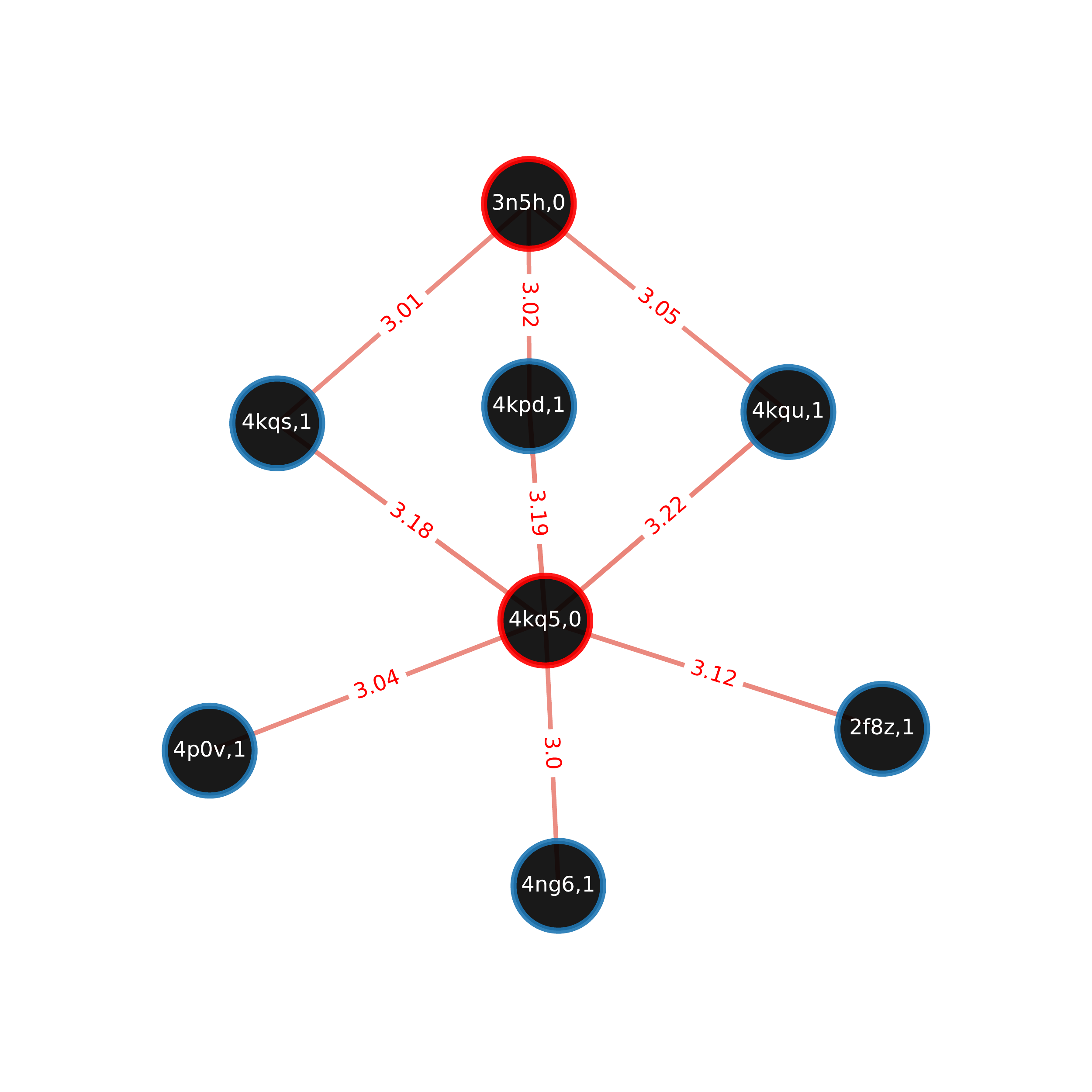}
        \captionsetup{labelformat=empty}
        \caption{}
        \label{supp_fig:pdb_hom_graph_15}
    \end{subfigure}
      \centering
      \begin{subfigure}{.24\textwidth} \centering
        \includegraphics[width=0.9\linewidth]{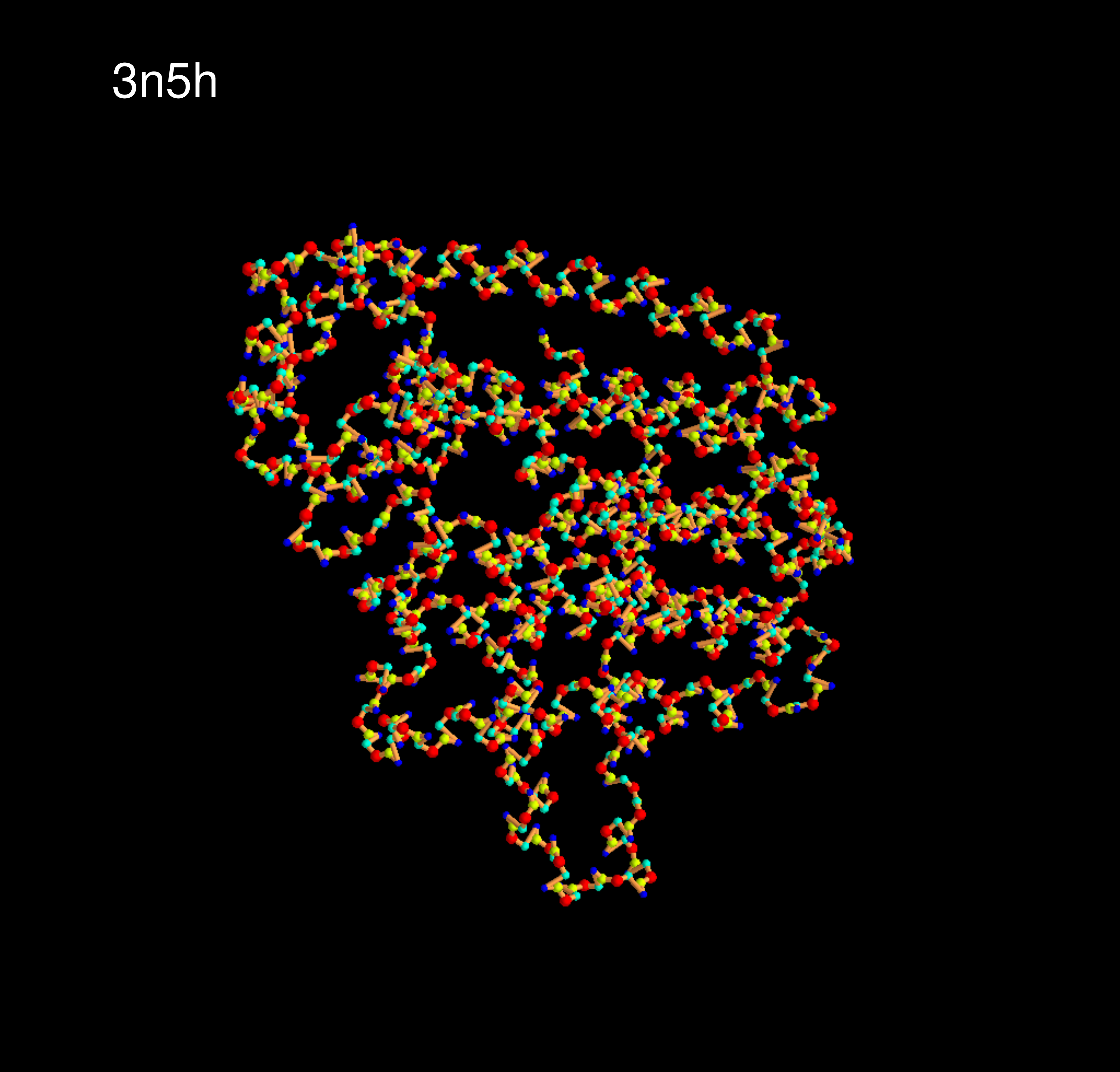}
        \captionsetup{labelformat=empty}
        \caption{}
        \label{supp_fig:pdb_3n5h}
    \end{subfigure}
      \centering
      \begin{subfigure}{.24\textwidth} \centering
        \includegraphics[width=0.9\linewidth]{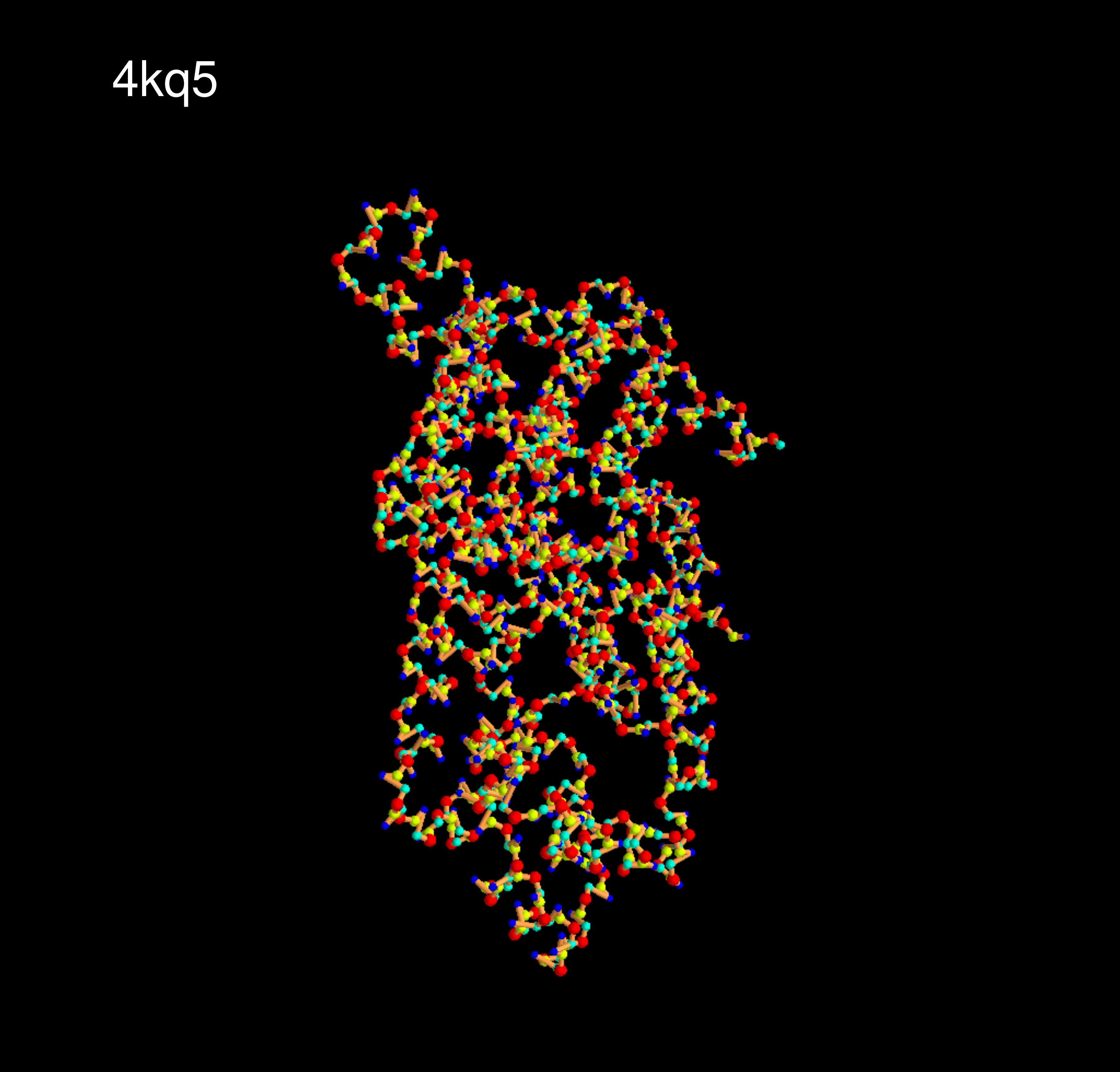}
        \captionsetup{labelformat=empty}
        \caption{}
        \label{supp_fig:pdb_4kq5}
    \end{subfigure}
      \centering
      \begin{subfigure}{.24\textwidth} \centering
        \includegraphics[width=0.9\linewidth]{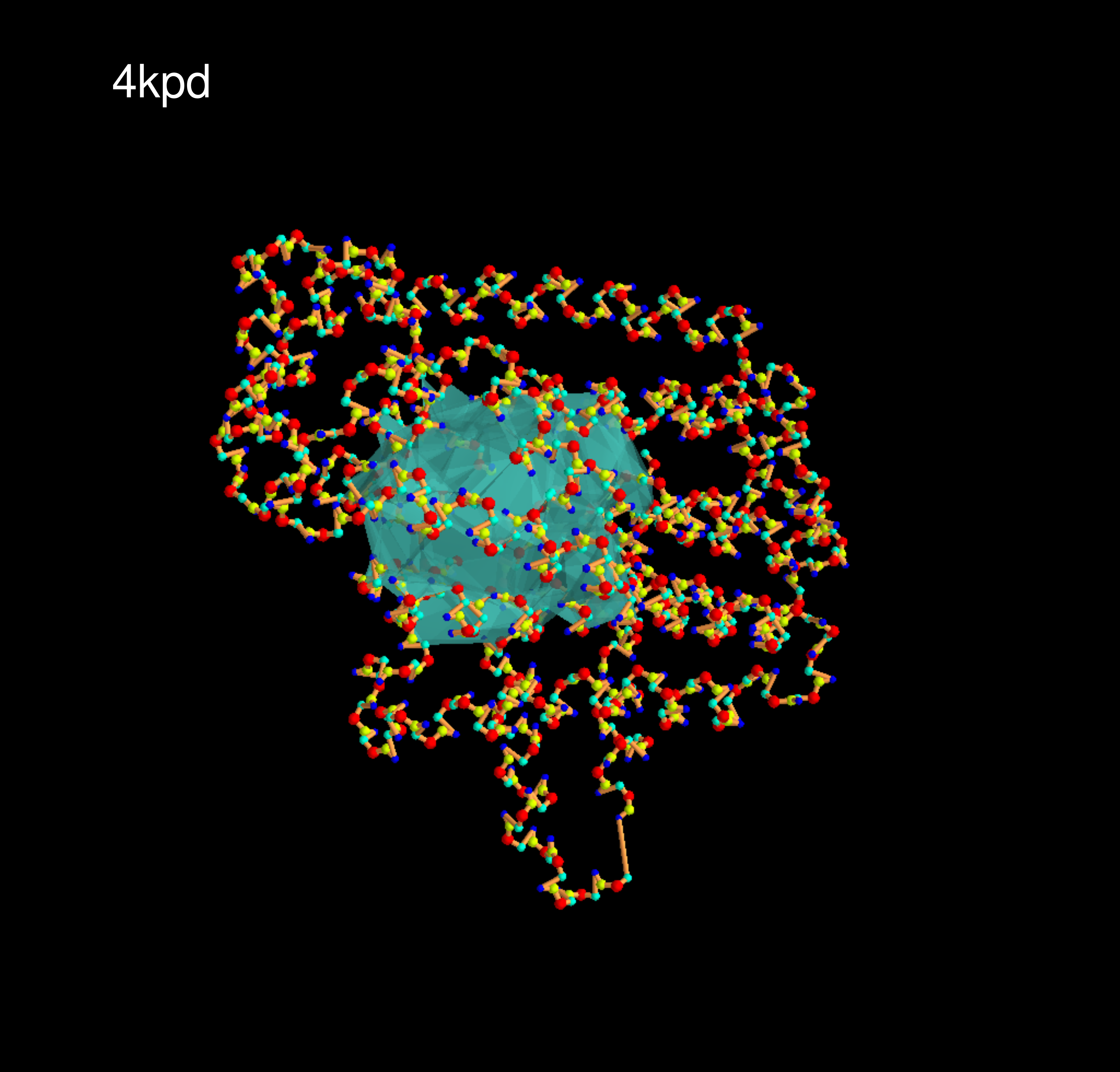}
        \captionsetup{labelformat=empty}
        \caption{}
        \label{supp_fig:pdb_4kpd}
    \end{subfigure}
      \centering
      \begin{subfigure}{.24\textwidth} \centering
        \includegraphics[width=0.9\linewidth]{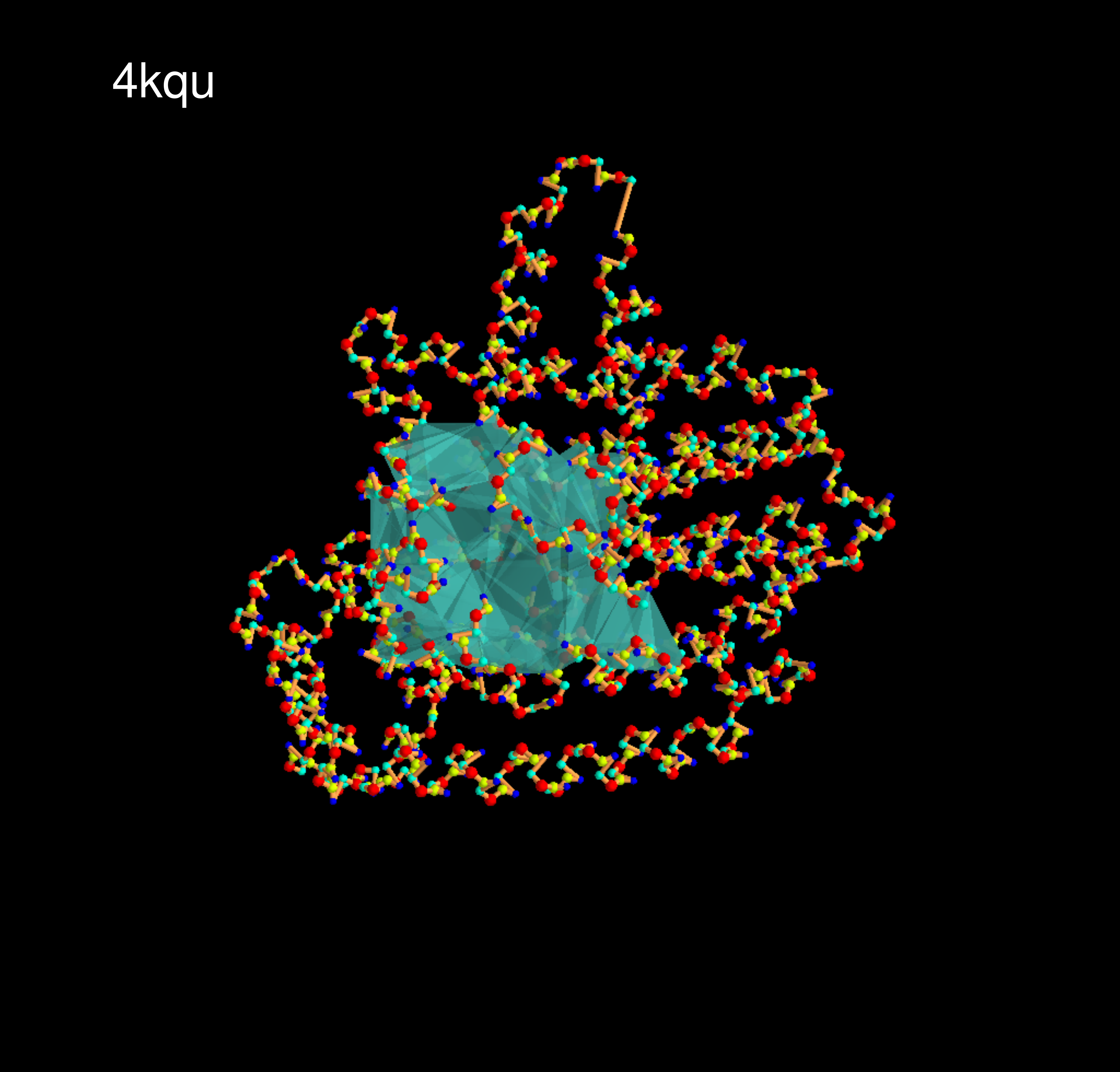}
        \captionsetup{labelformat=empty}
        \caption{}
        \label{supp_fig:pdb_4kqu}
    \end{subfigure}
      \centering
      \begin{subfigure}{.24\textwidth} \centering
        \includegraphics[width=0.9\linewidth]{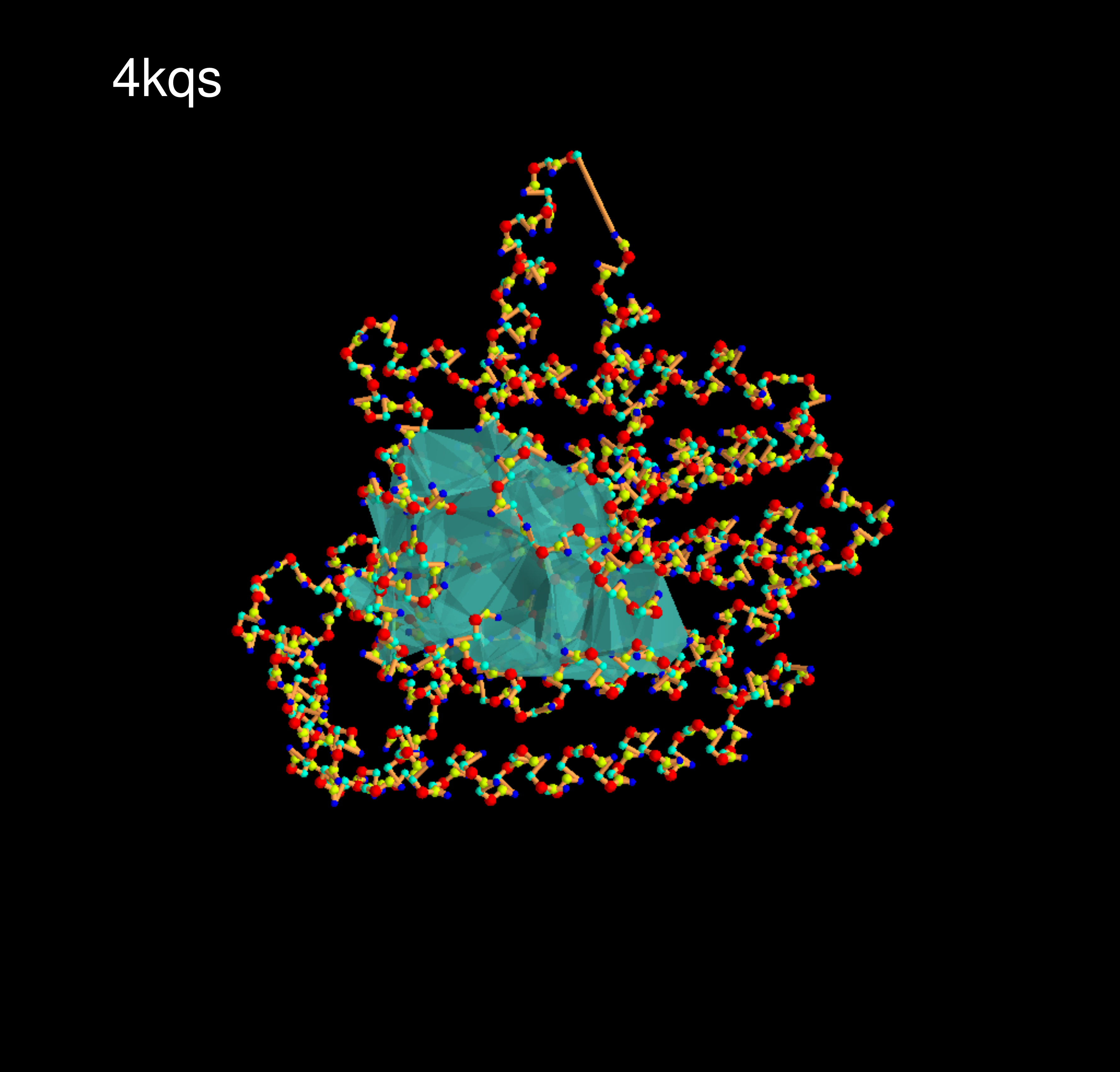}
        \captionsetup{labelformat=empty}
        \caption{}
        \label{supp_fig:pdb_4kqs}
    \end{subfigure}
      \centering
      \begin{subfigure}{.24\textwidth} \centering
        \includegraphics[width=0.9\linewidth]{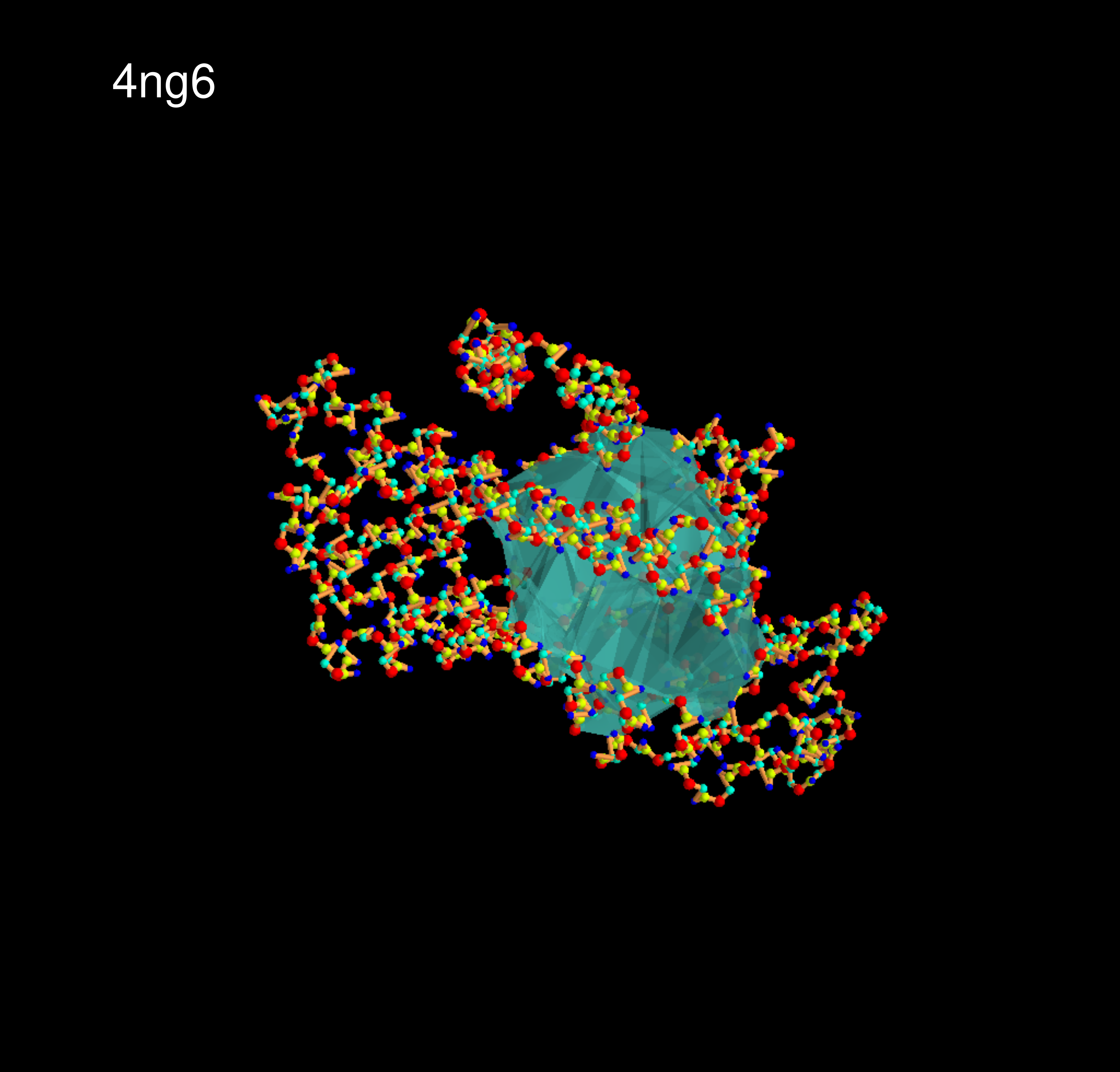}
        \captionsetup{labelformat=empty}
        \caption{}
        \label{supp_fig:pdb_4ng6}
    \end{subfigure}
      \centering
      \begin{subfigure}{.24\textwidth} \centering
        \includegraphics[width=0.9\linewidth]{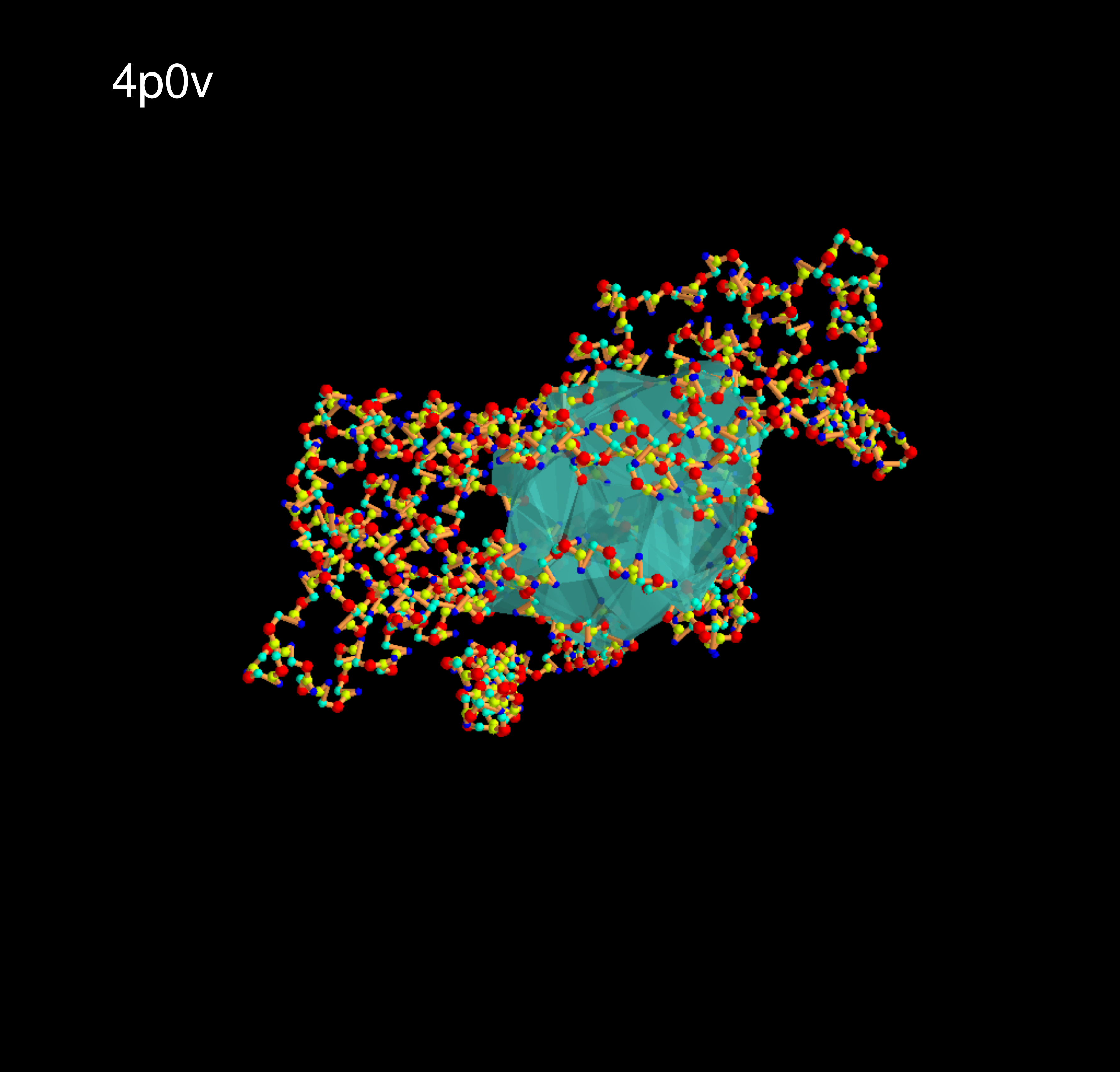}
        \captionsetup{labelformat=empty}
        \caption{}
        \label{supp_fig:pdb_4p0v}
    \end{subfigure}
      \centering
      \begin{subfigure}{.24\textwidth} \centering
        \includegraphics[width=0.9\linewidth]{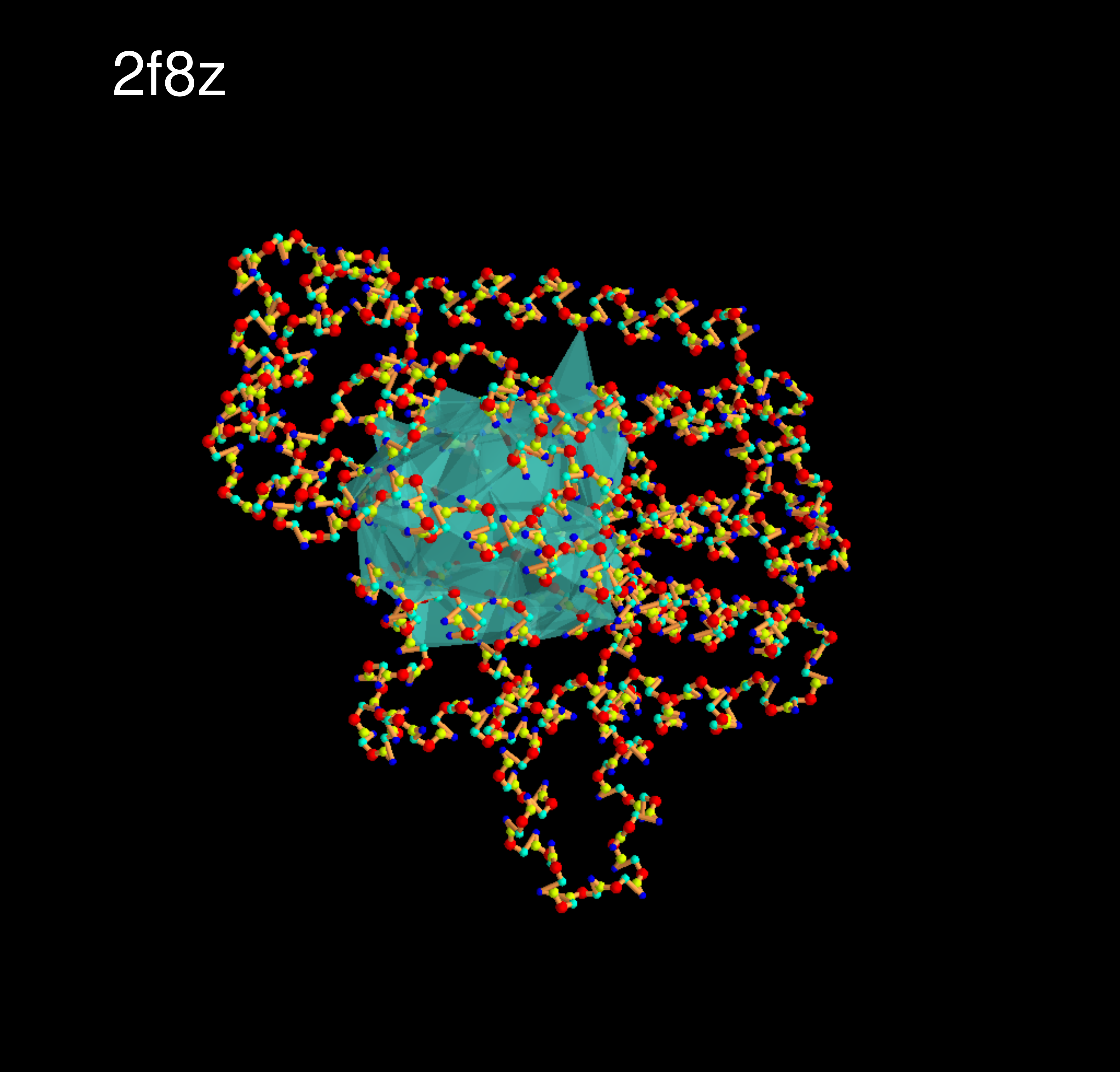}
        \captionsetup{labelformat=empty}
        \caption{}
        \label{supp_fig:pdb_2f8z}
    \end{subfigure}
      \caption{}
      \label{fig:pdb_15}
    \end{figure}

    \begin{figure}[!tbhp]
      \centering
      \begin{subfigure}{.48\textwidth} \centering
        \includegraphics[width=0.9\linewidth]{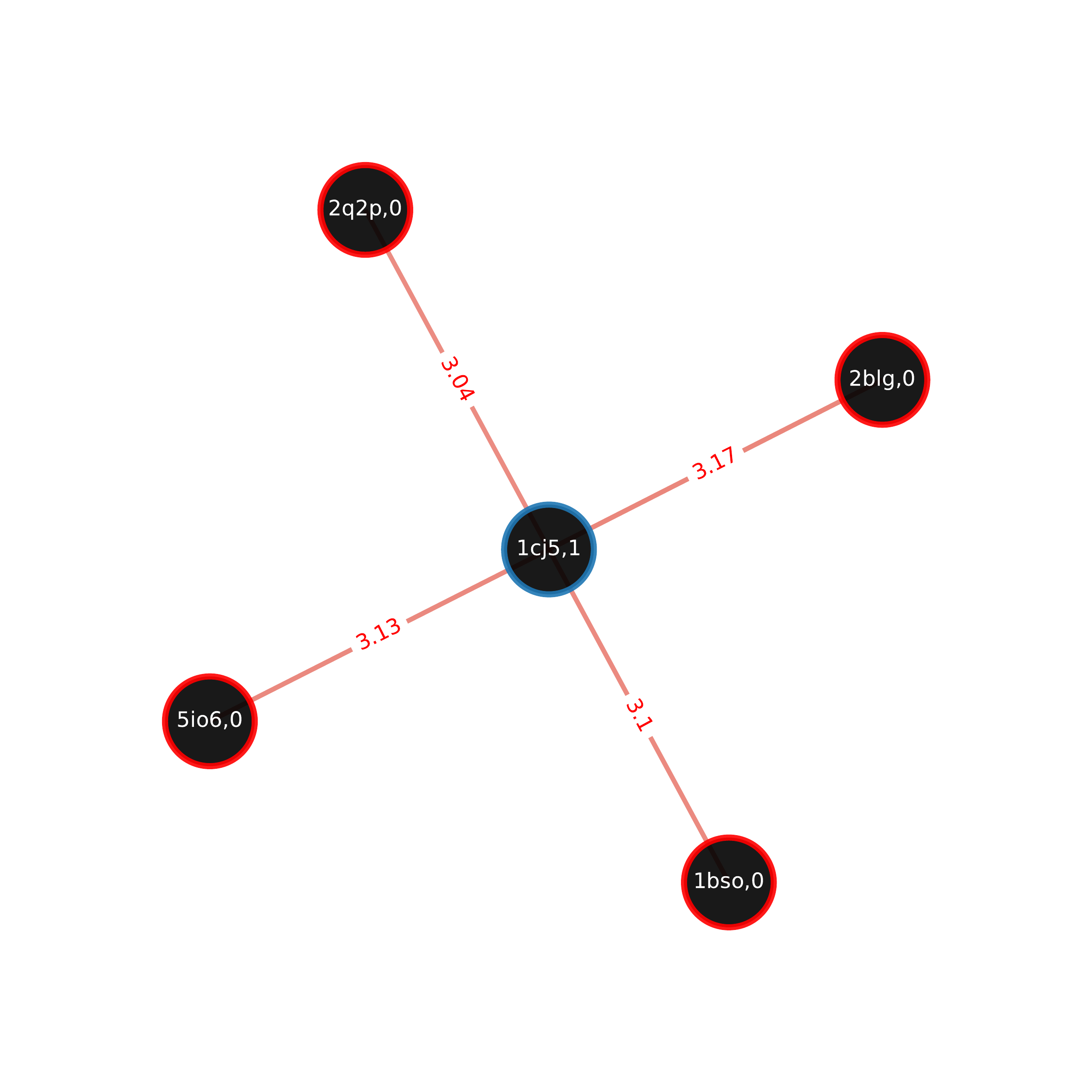}
        \captionsetup{labelformat=empty}
        \caption{}
        \label{supp_fig:pdb_hom_graph_16}
    \end{subfigure}
      \centering
      \begin{subfigure}{.24\textwidth} \centering
        \includegraphics[width=0.9\linewidth]{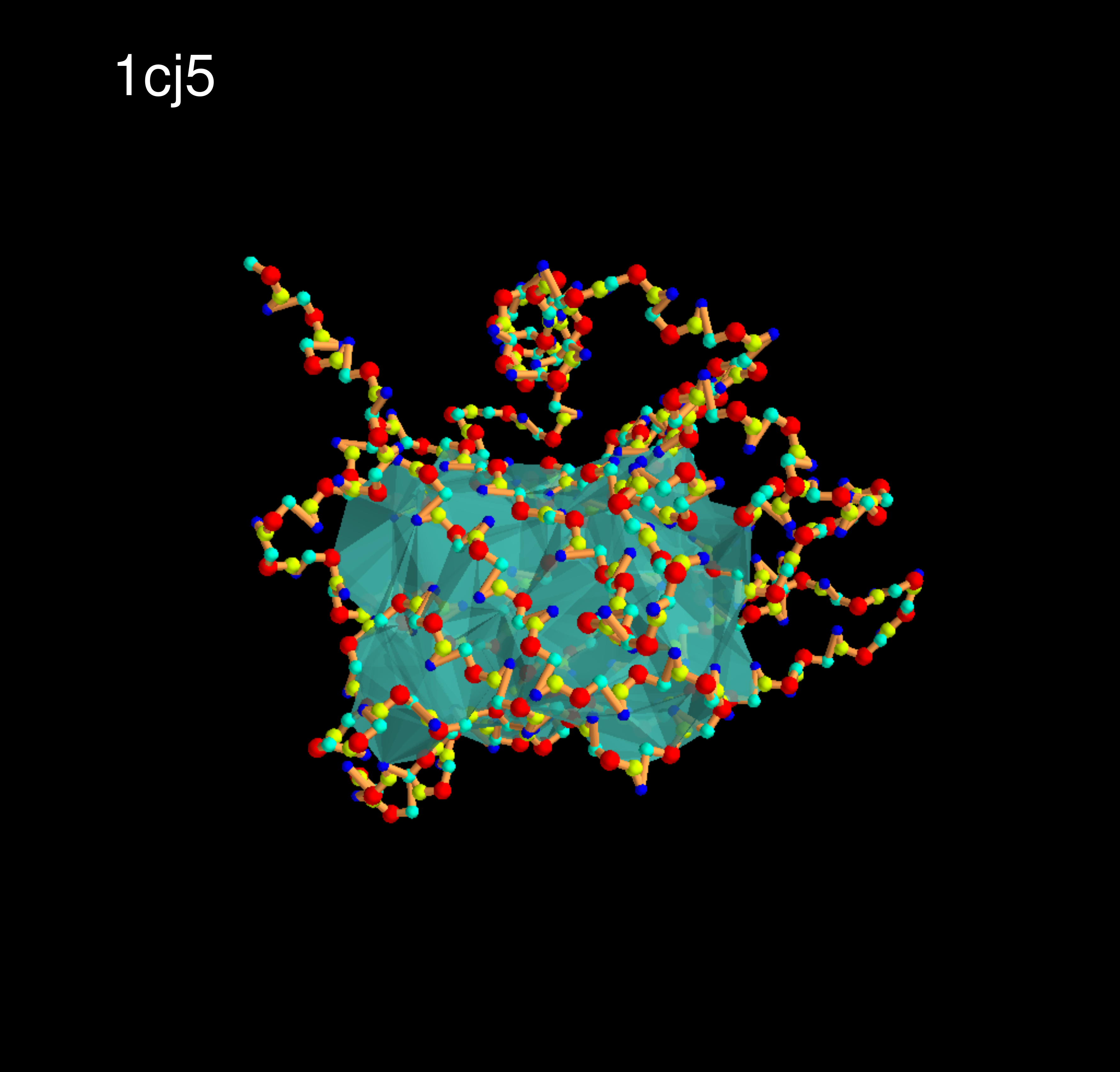}
        \captionsetup{labelformat=empty}
        \caption{}
        \label{supp_fig:pdb_1cj5}
    \end{subfigure}
      \centering
      \begin{subfigure}{.24\textwidth} \centering
        \includegraphics[width=0.9\linewidth]{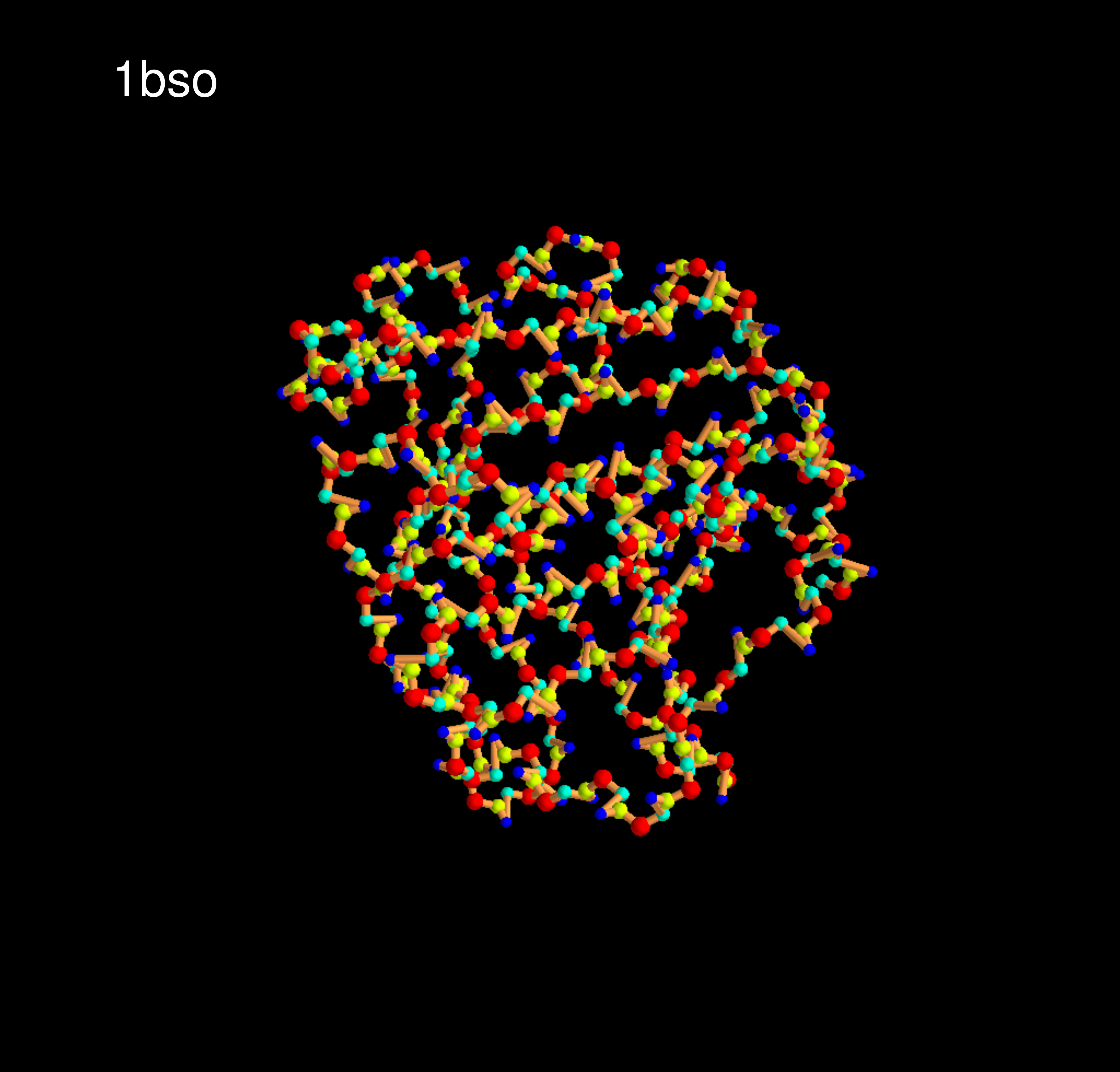}
        \captionsetup{labelformat=empty}
        \caption{}
        \label{supp_fig:pdb_1bso}
    \end{subfigure}
      \centering
      \begin{subfigure}{.24\textwidth} \centering
        \includegraphics[width=0.9\linewidth]{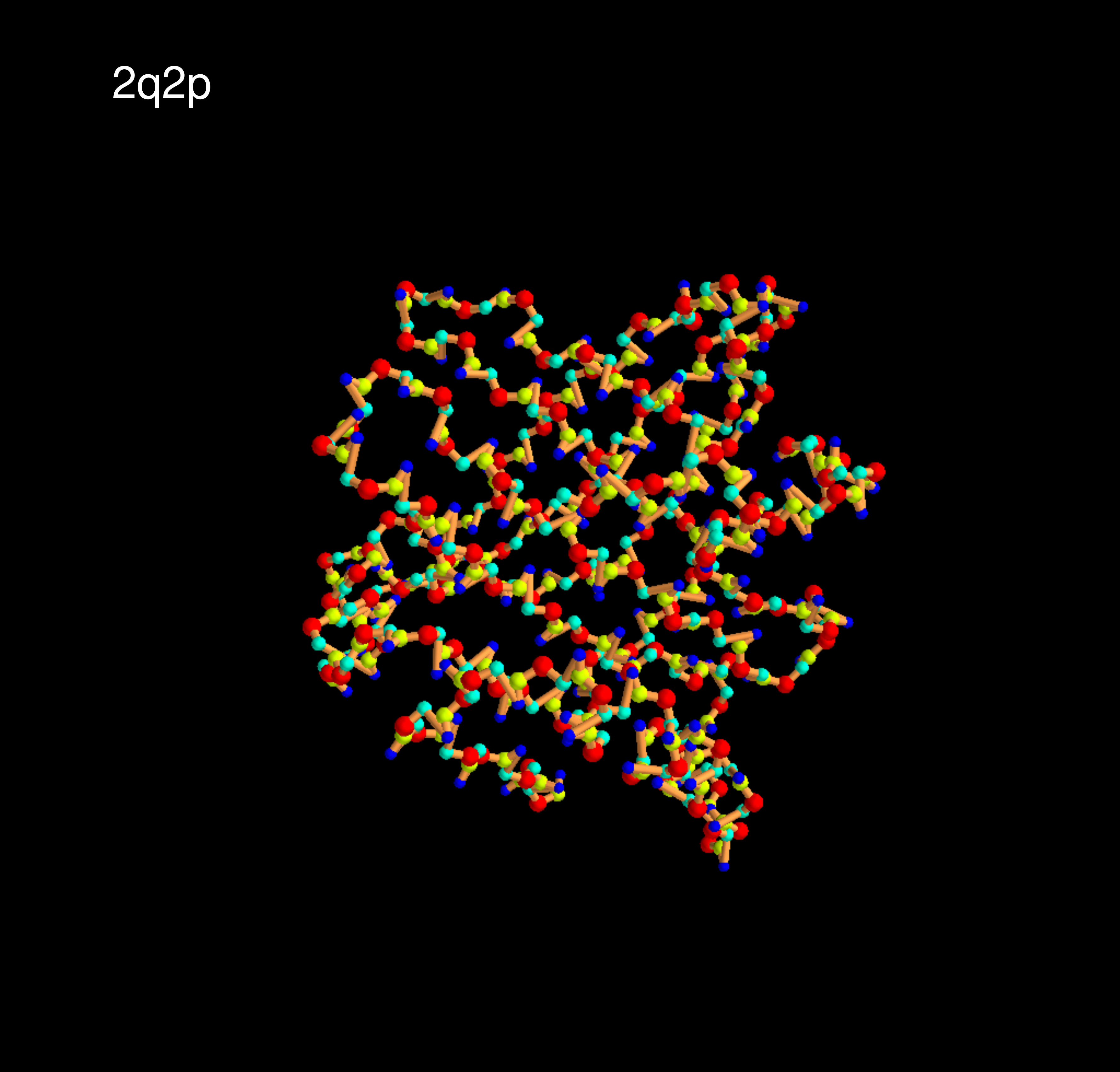}
        \captionsetup{labelformat=empty}
        \caption{}
        \label{supp_fig:pdb_2q2p}
    \end{subfigure}
      \centering
      \begin{subfigure}{.24\textwidth} \centering
        \includegraphics[width=0.9\linewidth]{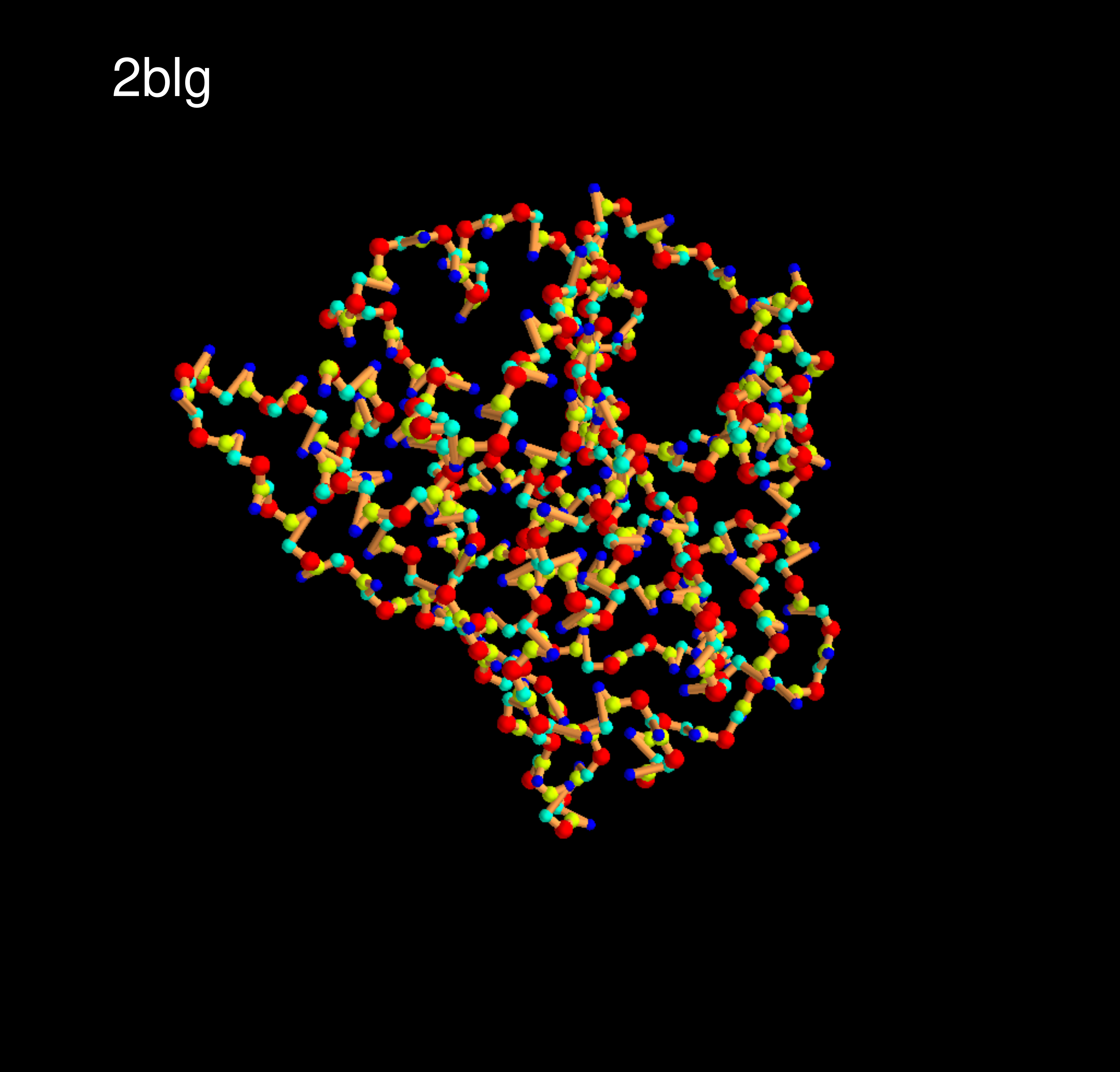}
        \captionsetup{labelformat=empty}
        \caption{}
        \label{supp_fig:pdb_2blg}
    \end{subfigure}
      \centering
      \begin{subfigure}{.24\textwidth} \centering
        \includegraphics[width=0.9\linewidth]{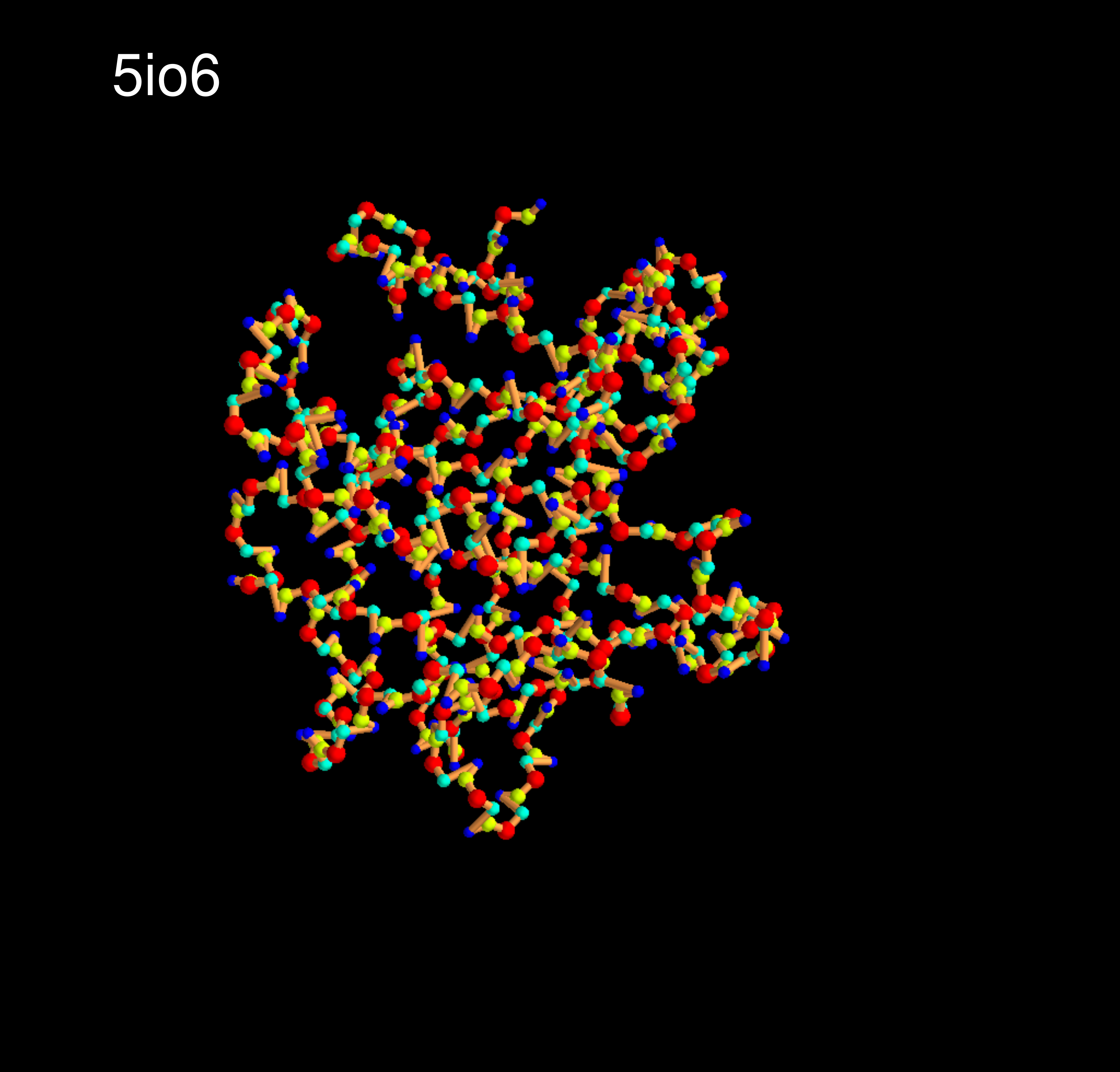}
        \captionsetup{labelformat=empty}
        \caption{}
        \label{supp_fig:pdb_5io6}
    \end{subfigure}
      \caption{}
      \label{fig:pdb_16}
    \end{figure}

    \begin{figure}[!tbhp]
      \centering
      \begin{subfigure}{.33\textwidth} \centering
        \includegraphics[width=0.9\linewidth]{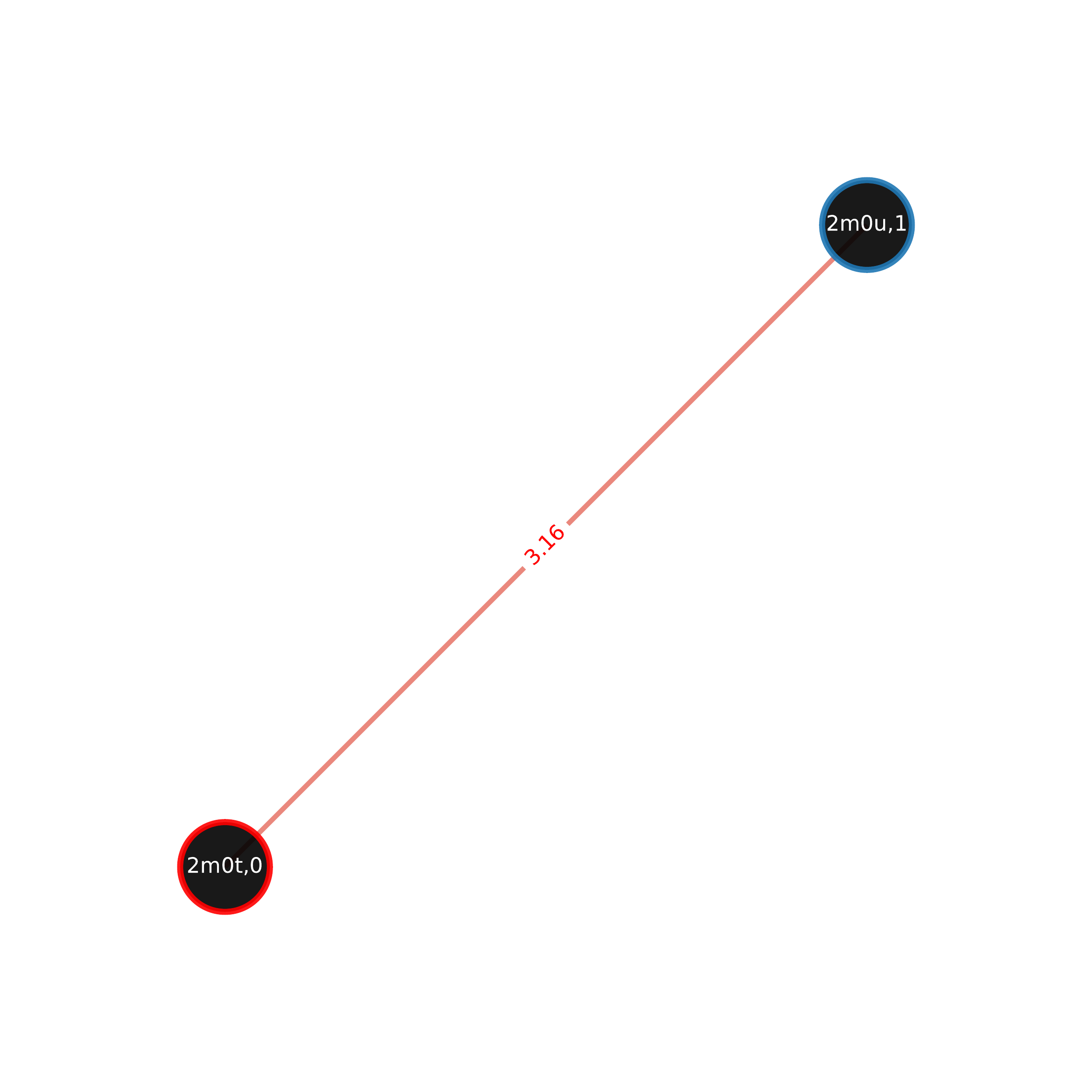}
        \captionsetup{labelformat=empty}
        \caption{}
        \label{supp_fig:pdb_hom_graph_17}
    \end{subfigure}
      \centering
      \begin{subfigure}{.33\textwidth} \centering
        \includegraphics[width=0.9\linewidth]{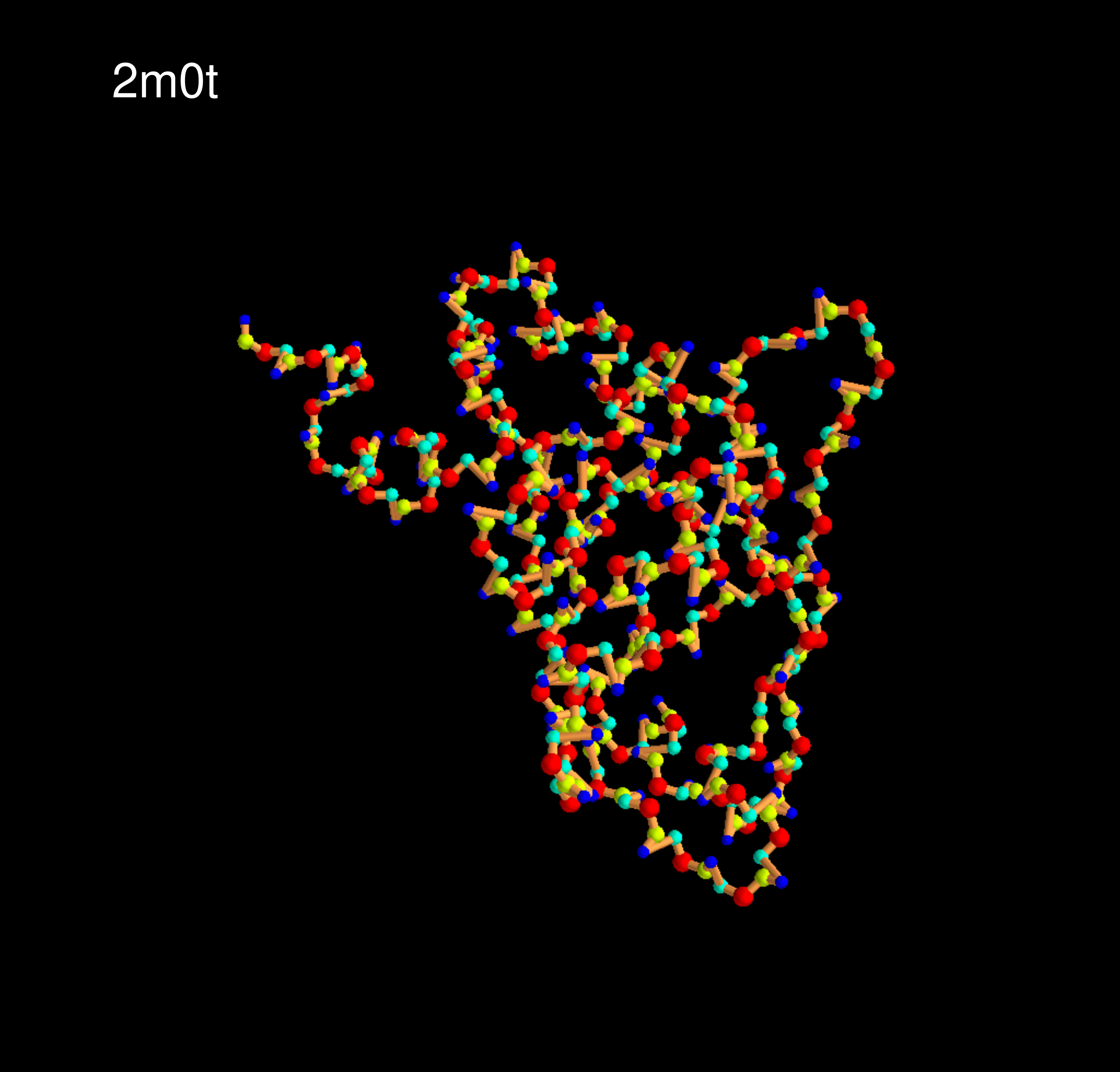}
        \captionsetup{labelformat=empty}
        \caption{}
        \label{supp_fig:pdb_2m0t}
    \end{subfigure}
      \centering
      \begin{subfigure}{.33\textwidth} \centering
        \includegraphics[width=0.9\linewidth]{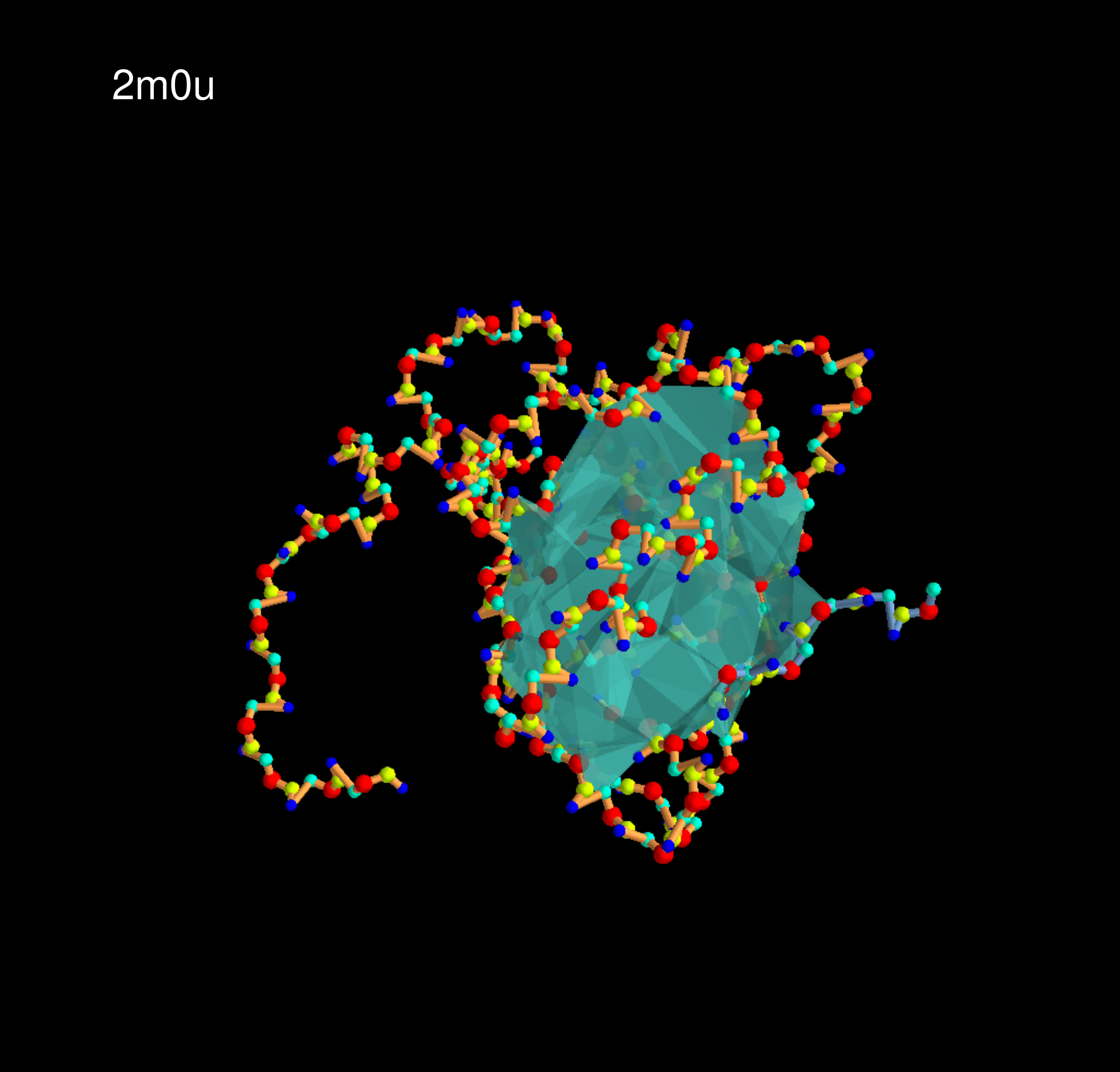}
        \captionsetup{labelformat=empty}
        \caption{}
        \label{supp_fig:pdb_2m0u}
    \end{subfigure}
      \caption{}
      \label{fig:pdb_17}
    \end{figure}

    \begin{figure}[!tbhp]
      \centering
      \begin{subfigure}{.33\textwidth} \centering
        \includegraphics[width=0.9\linewidth]{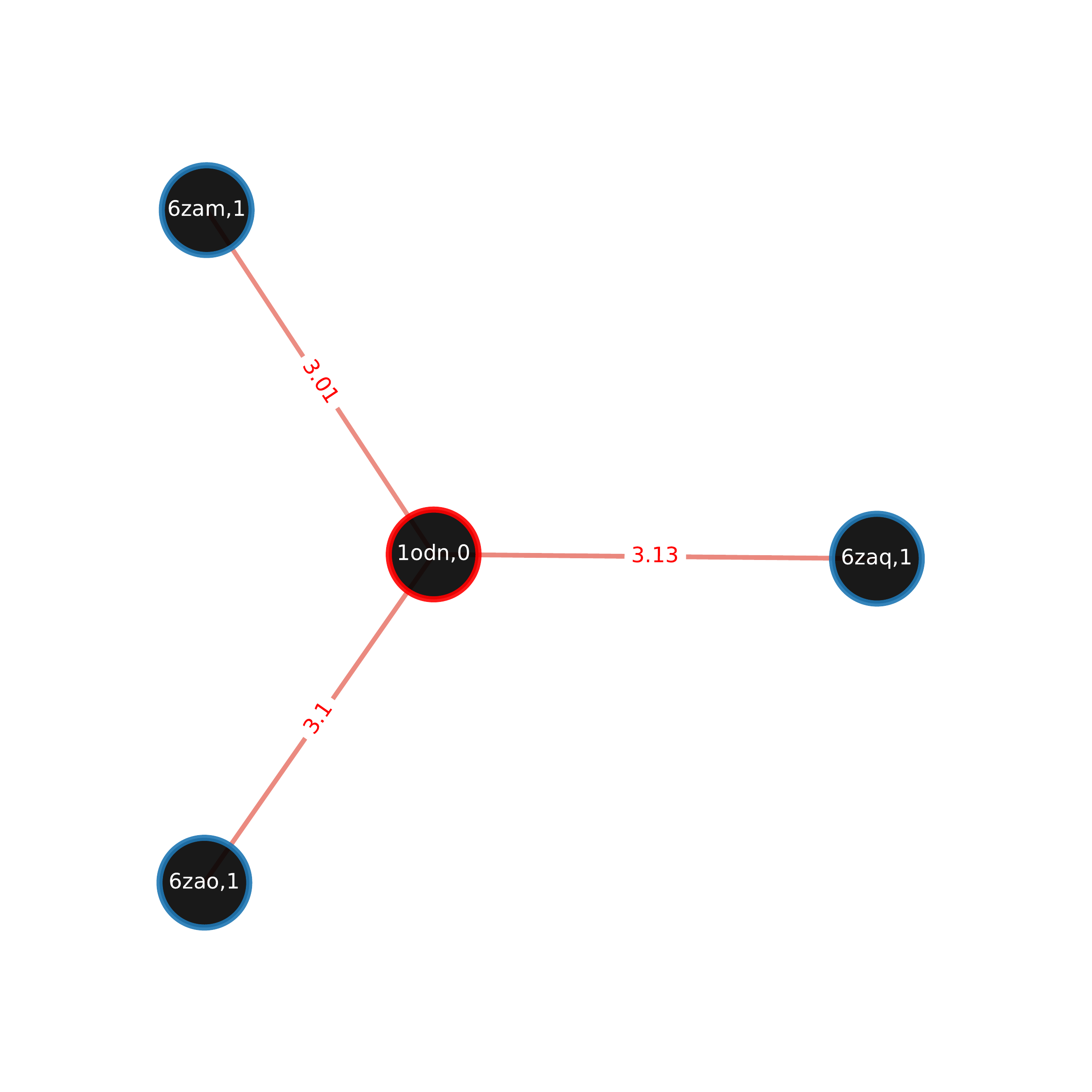}
        \captionsetup{labelformat=empty}
        \caption{}
        \label{supp_fig:pdb_hom_graph_18}
    \end{subfigure}
      \centering
      \begin{subfigure}{.33\textwidth} \centering
        \includegraphics[width=0.9\linewidth]{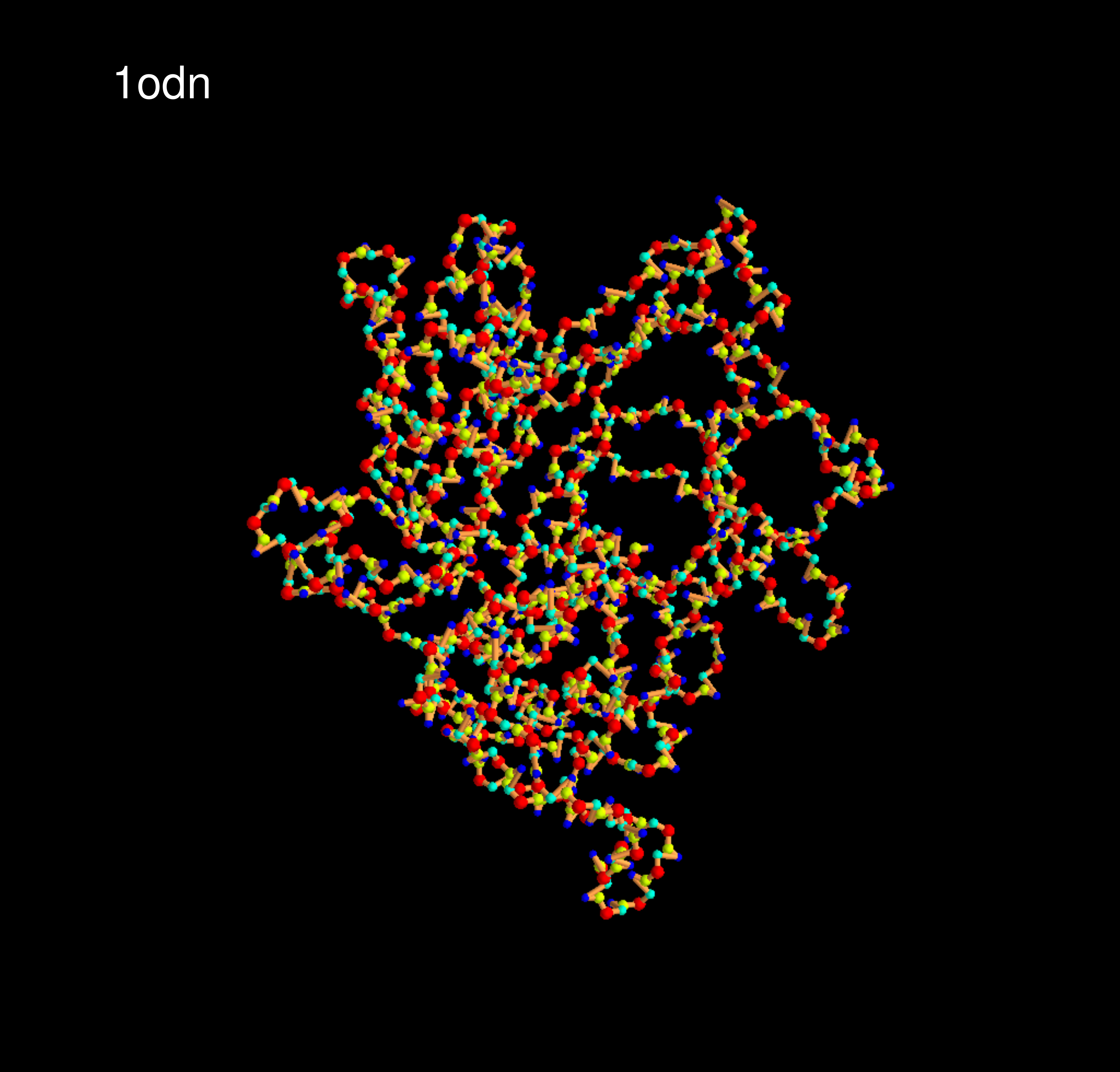}
        \captionsetup{labelformat=empty}
        \caption{}
        \label{supp_fig:pdb_1odn}
    \end{subfigure}
      \centering
      \begin{subfigure}{.33\textwidth} \centering
        \includegraphics[width=0.9\linewidth]{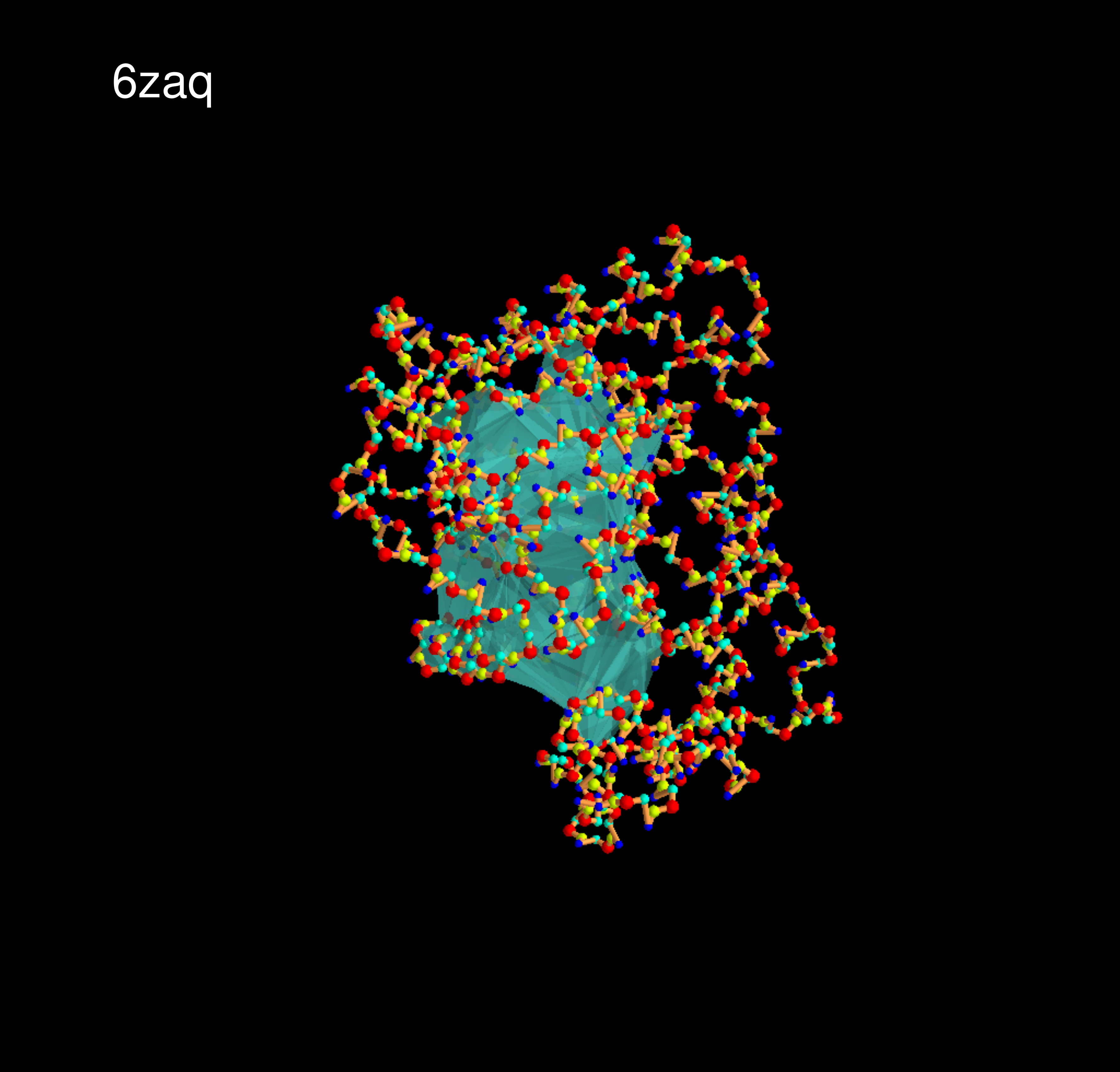}
        \captionsetup{labelformat=empty}
        \caption{}
        \label{supp_fig:pdb_6zaq}
    \end{subfigure}
      \centering
      \begin{subfigure}{.33\textwidth} \centering
        \includegraphics[width=0.9\linewidth]{figures/pdb_6zaq.pdf}
        \captionsetup{labelformat=empty}
        \caption{}
        \label{supp_fig:pdb_6zaq}
    \end{subfigure}
      \centering
      \begin{subfigure}{.33\textwidth} \centering
        \includegraphics[width=0.9\linewidth]{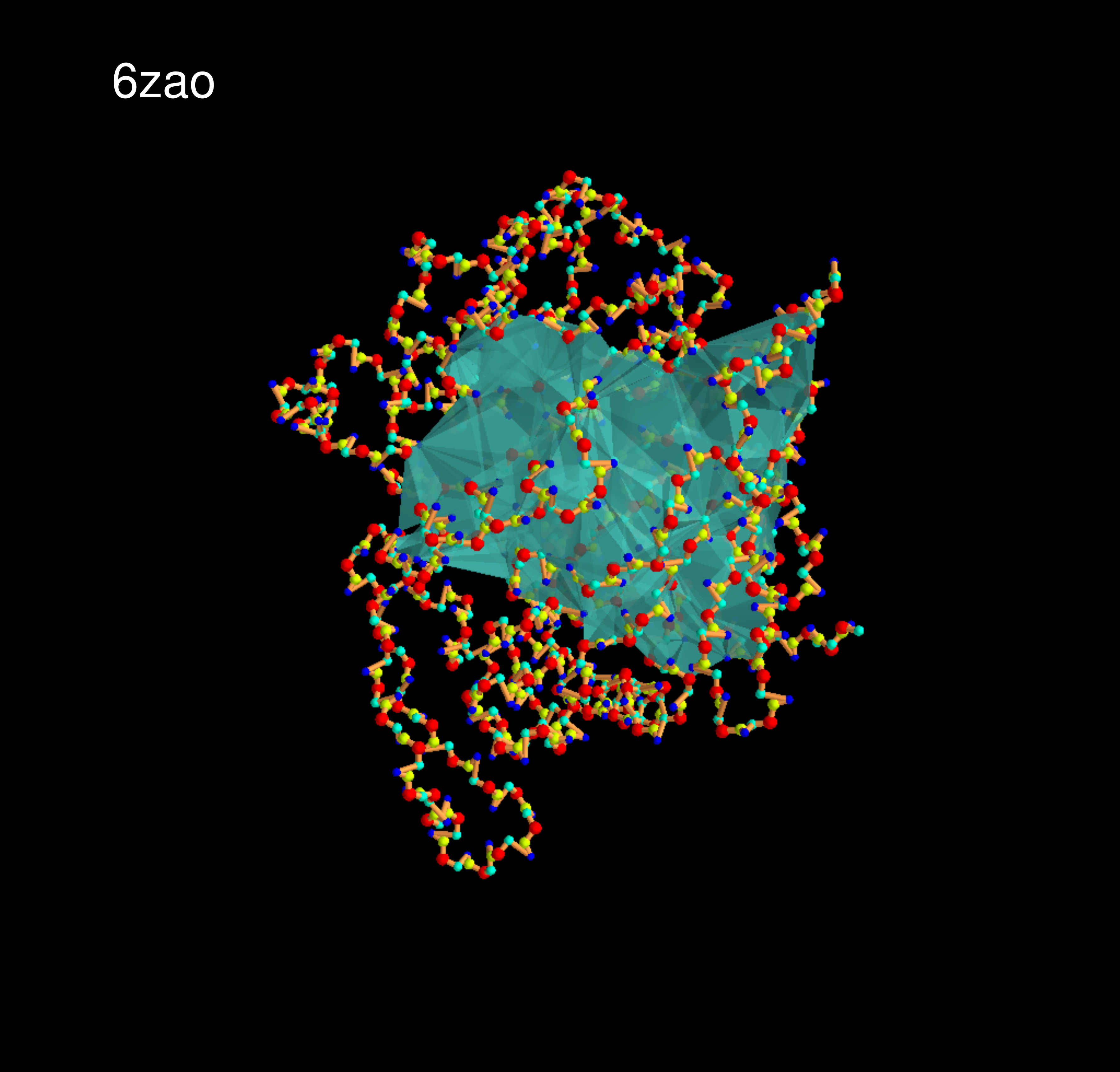}
        \captionsetup{labelformat=empty}
        \caption{}
        \label{supp_fig:pdb_6zao}
    \end{subfigure}
      \centering
      \begin{subfigure}{.33\textwidth} \centering
        \includegraphics[width=0.9\linewidth]{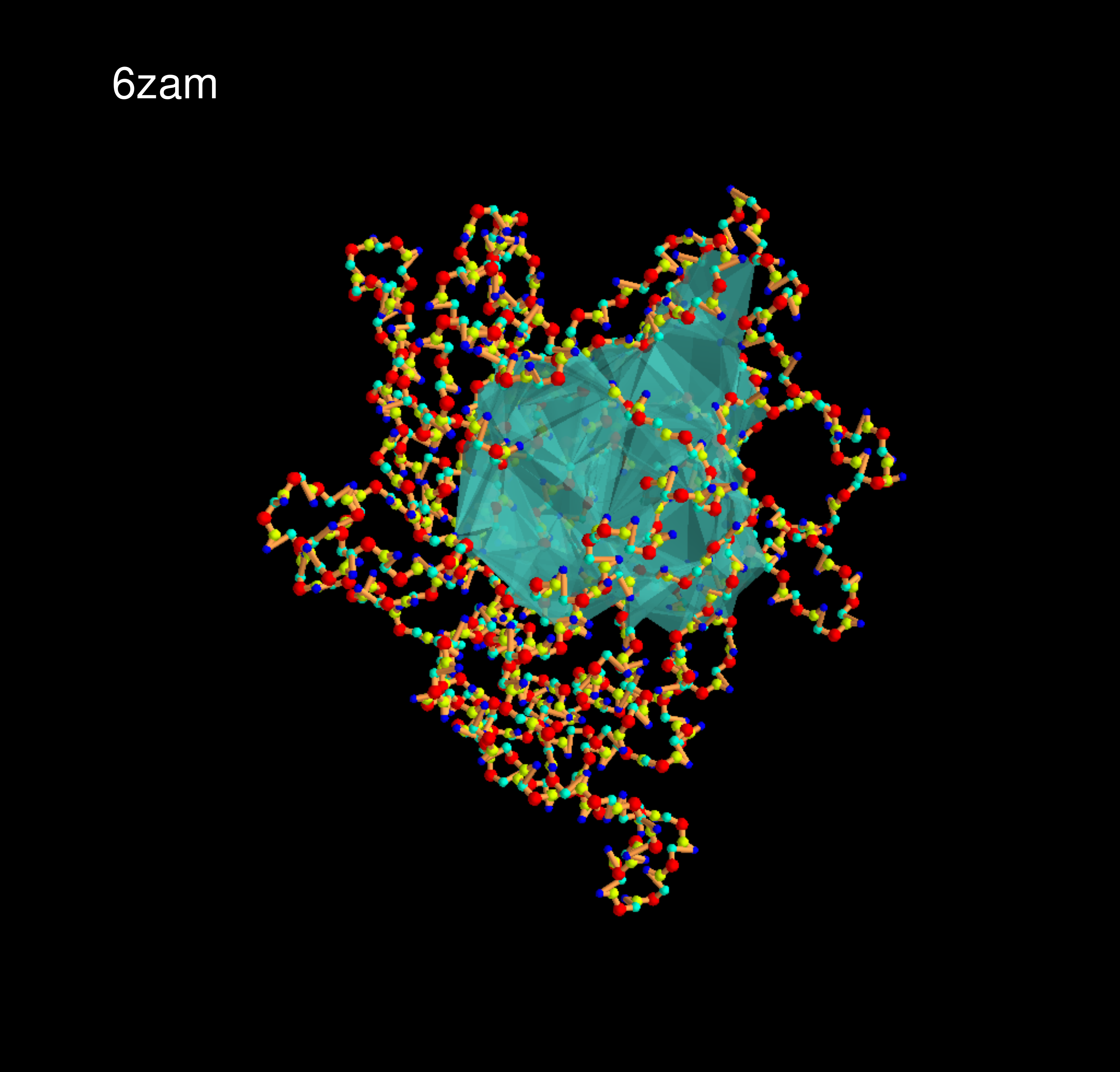}
        \captionsetup{labelformat=empty}
        \caption{}
        \label{supp_fig:pdb_6zam}
    \end{subfigure}
      \caption{}
      \label{fig:pdb_18}
    \end{figure}

    \begin{figure}[!tbhp]
      \centering
      \begin{subfigure}{.33\textwidth} \centering
        \includegraphics[width=0.9\linewidth]{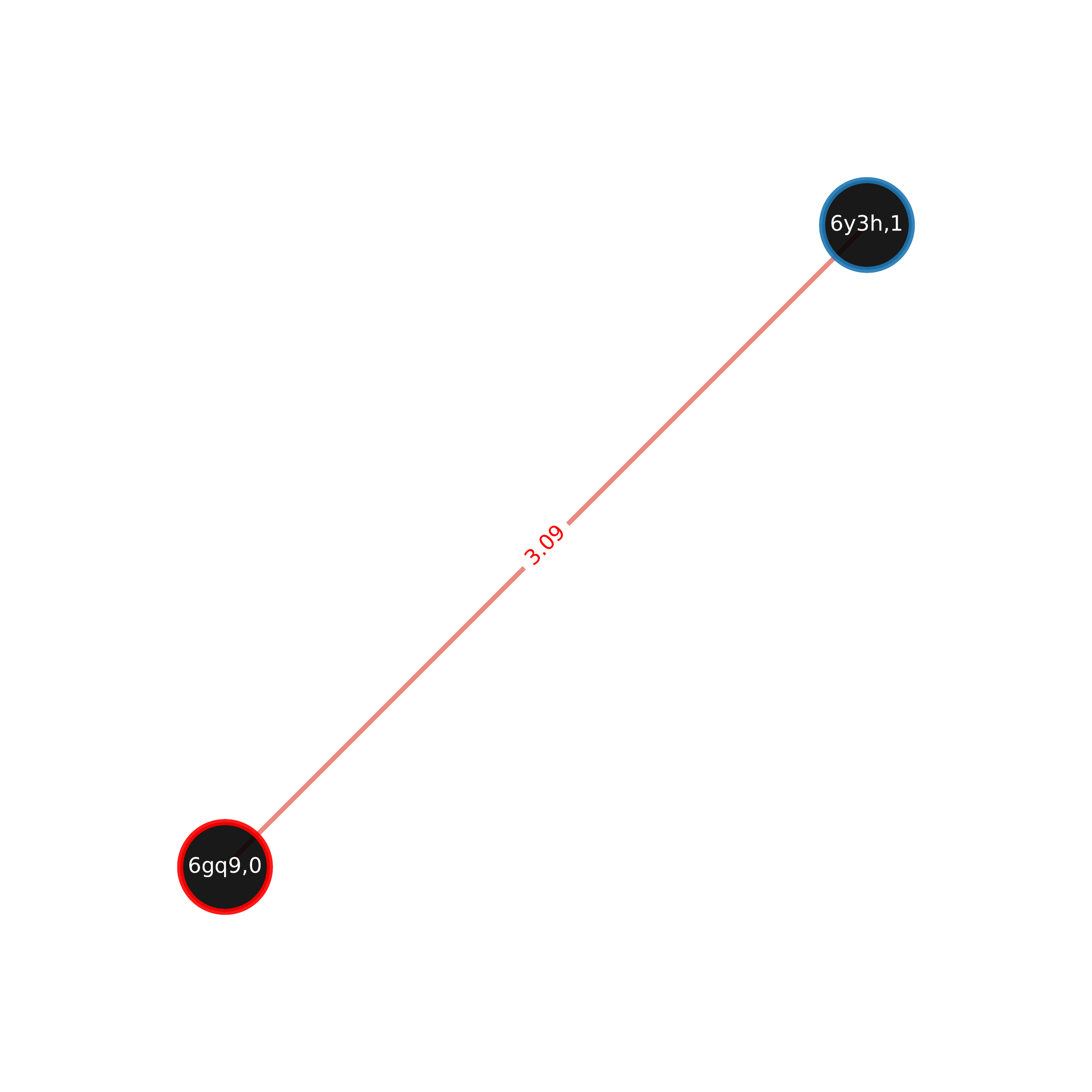}
        \captionsetup{labelformat=empty}
        \caption{}
        \label{supp_fig:pdb_hom_graph_19}
    \end{subfigure}
      \centering
      \begin{subfigure}{.33\textwidth} \centering
        \includegraphics[width=0.9\linewidth]{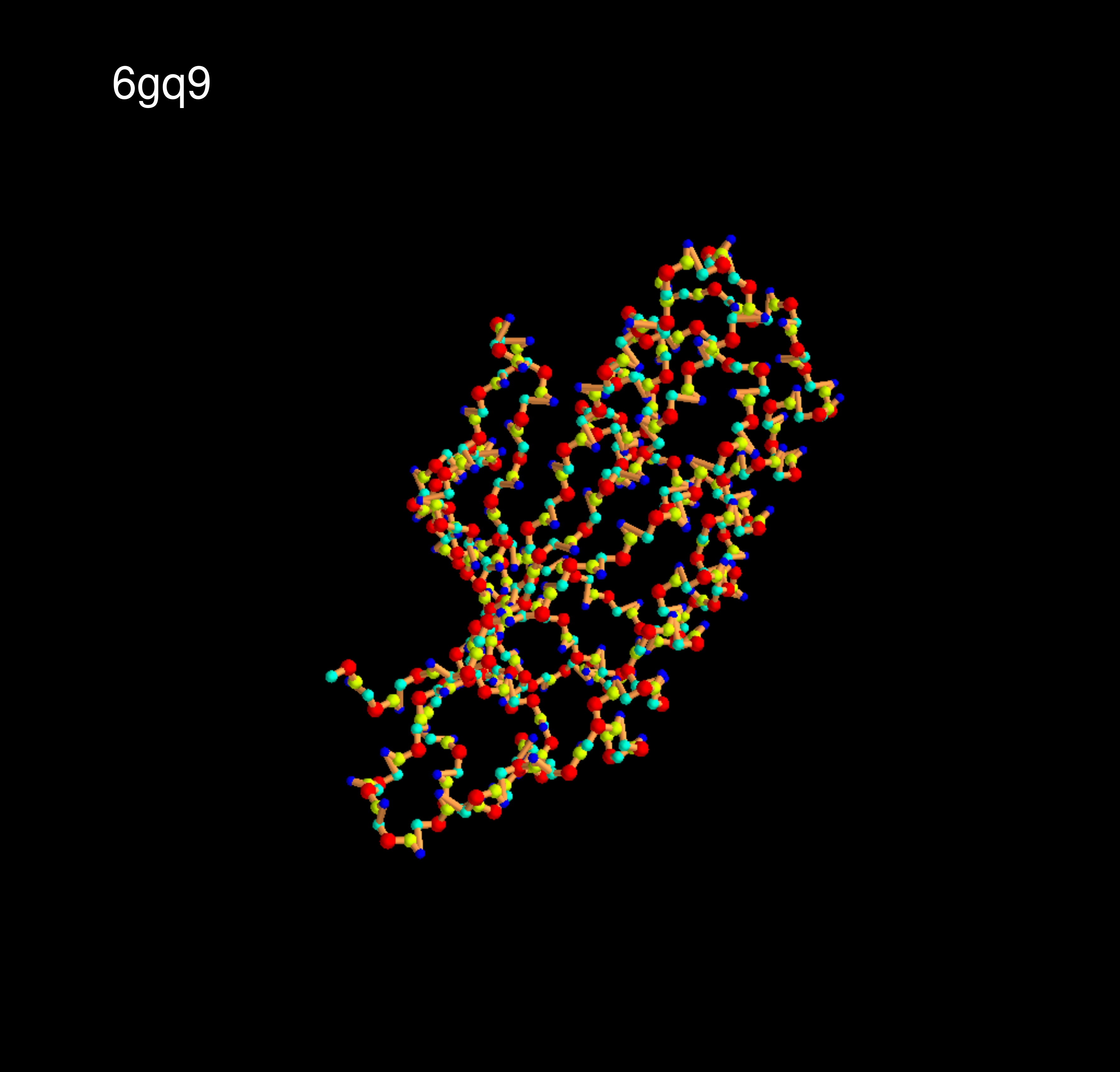}
        \captionsetup{labelformat=empty}
        \caption{}
        \label{supp_fig:pdb_6gq9}
    \end{subfigure}
      \centering
      \begin{subfigure}{.33\textwidth} \centering
        \includegraphics[width=0.9\linewidth]{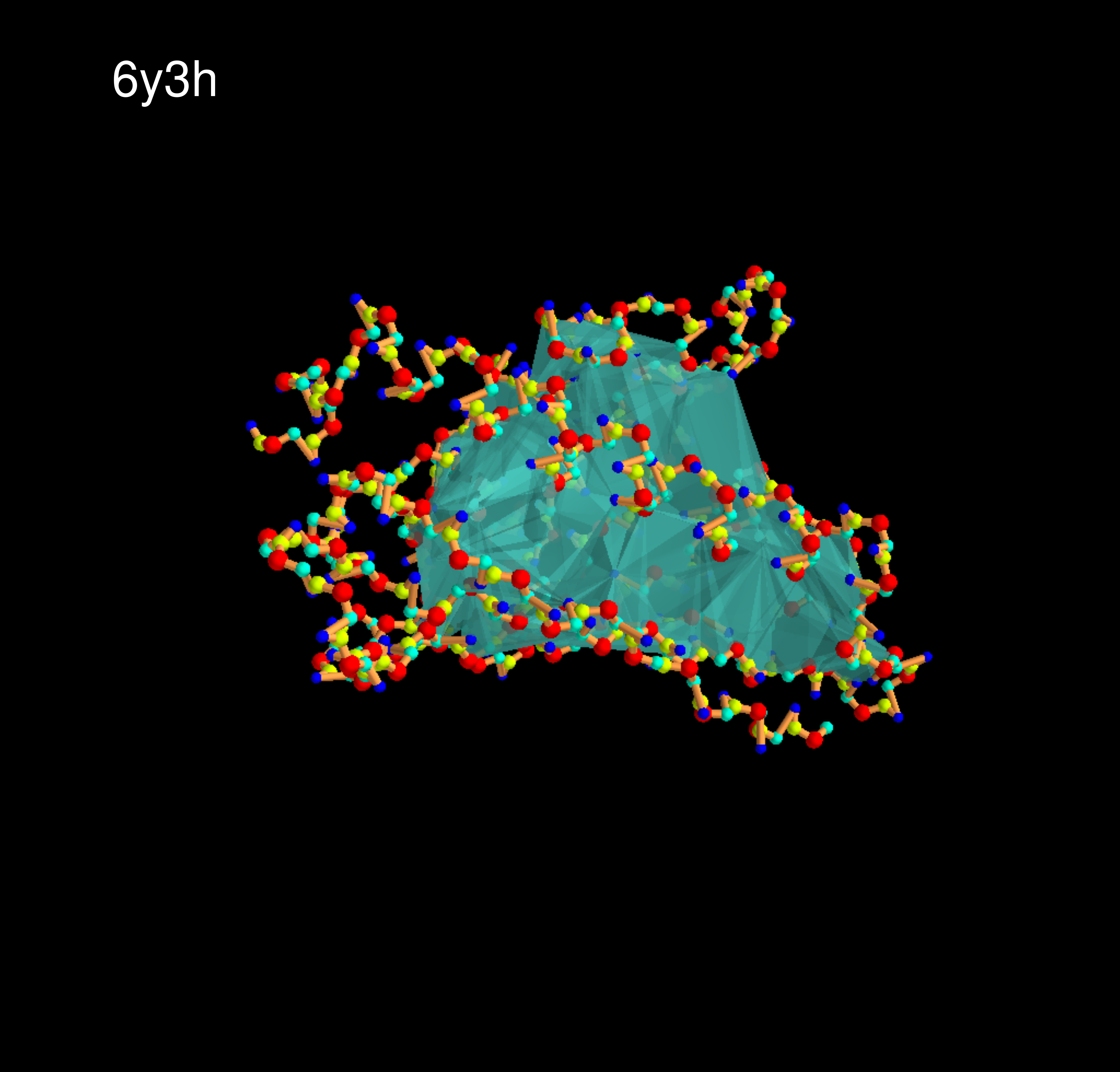}
        \captionsetup{labelformat=empty}
        \caption{}
        \label{supp_fig:pdb_6y3h}
    \end{subfigure}
      \caption{}
      \label{fig:pdb_19}
    \end{figure}

    \begin{figure}[!tbhp]
      \centering
      \begin{subfigure}{.48\textwidth} \centering
        \includegraphics[width=0.9\linewidth]{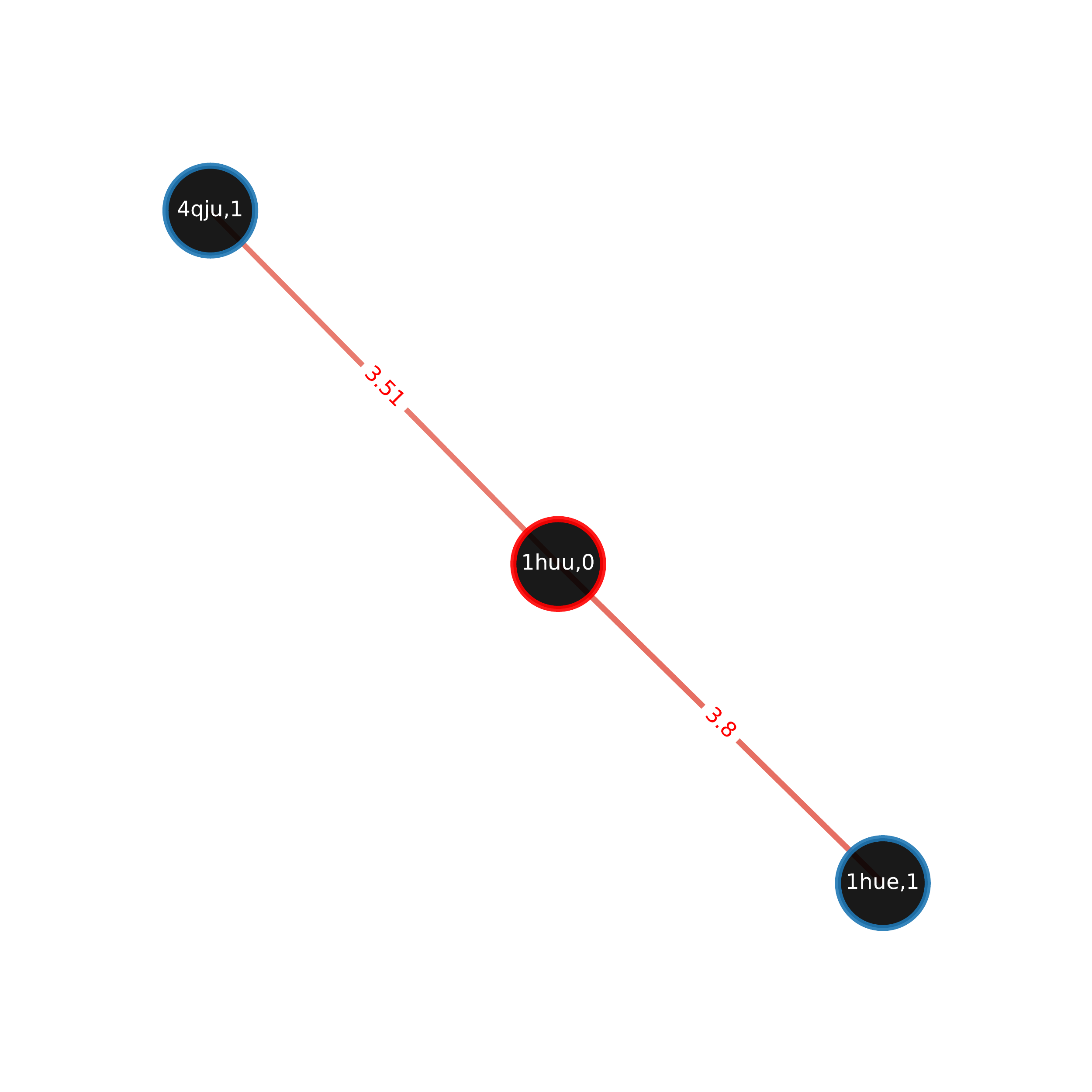}
        \captionsetup{labelformat=empty}
        \caption{}
        \label{supp_fig:pdb_hom_graph_20}
    \end{subfigure}
      \centering
      \begin{subfigure}{.48\textwidth} \centering
        \includegraphics[width=0.9\linewidth]{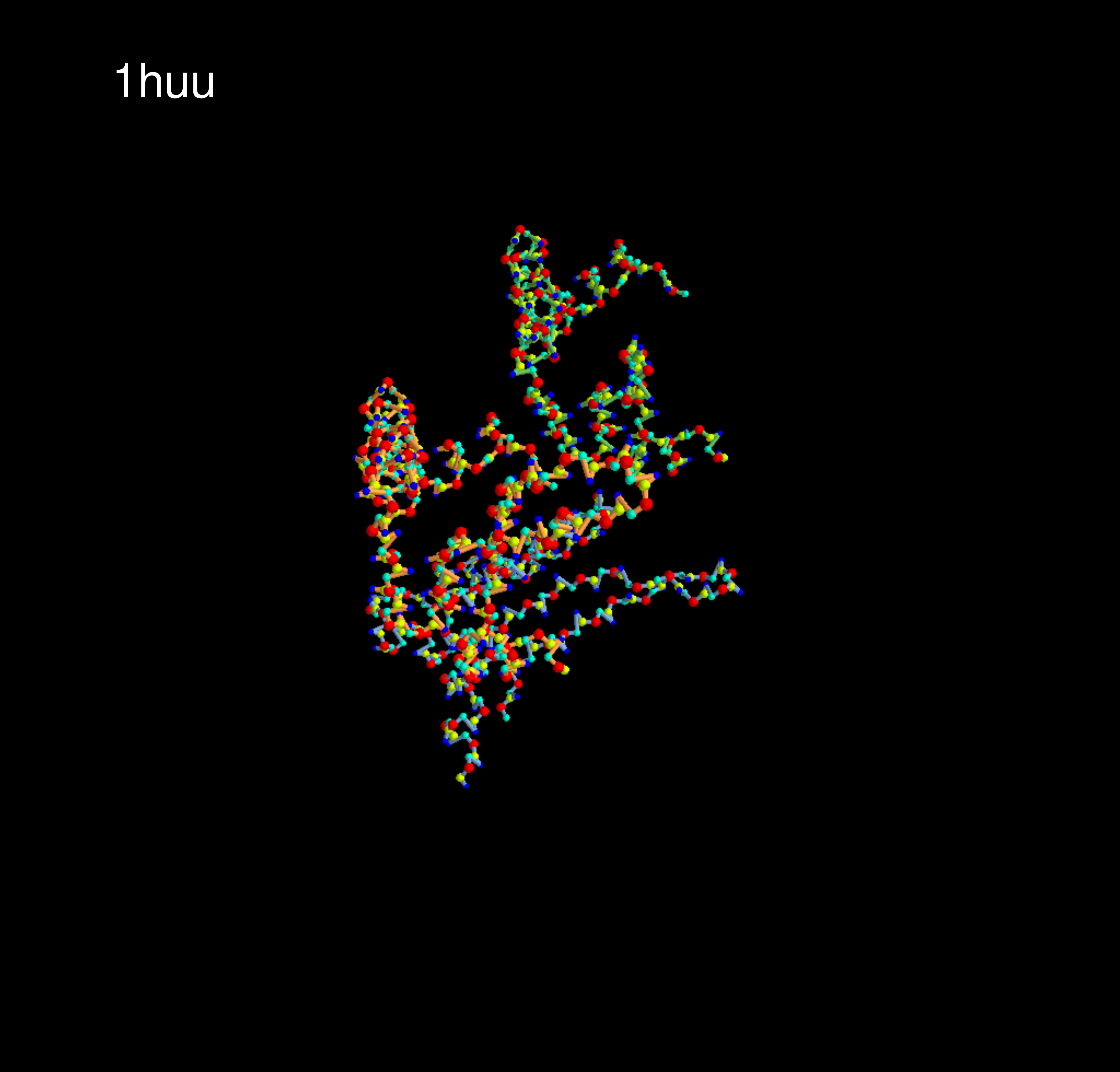}
        \captionsetup{labelformat=empty}
        \caption{}
        \label{supp_fig:pdb_1huu}
    \end{subfigure}
      \centering
      \begin{subfigure}{.48\textwidth} \centering
        \includegraphics[width=0.9\linewidth]{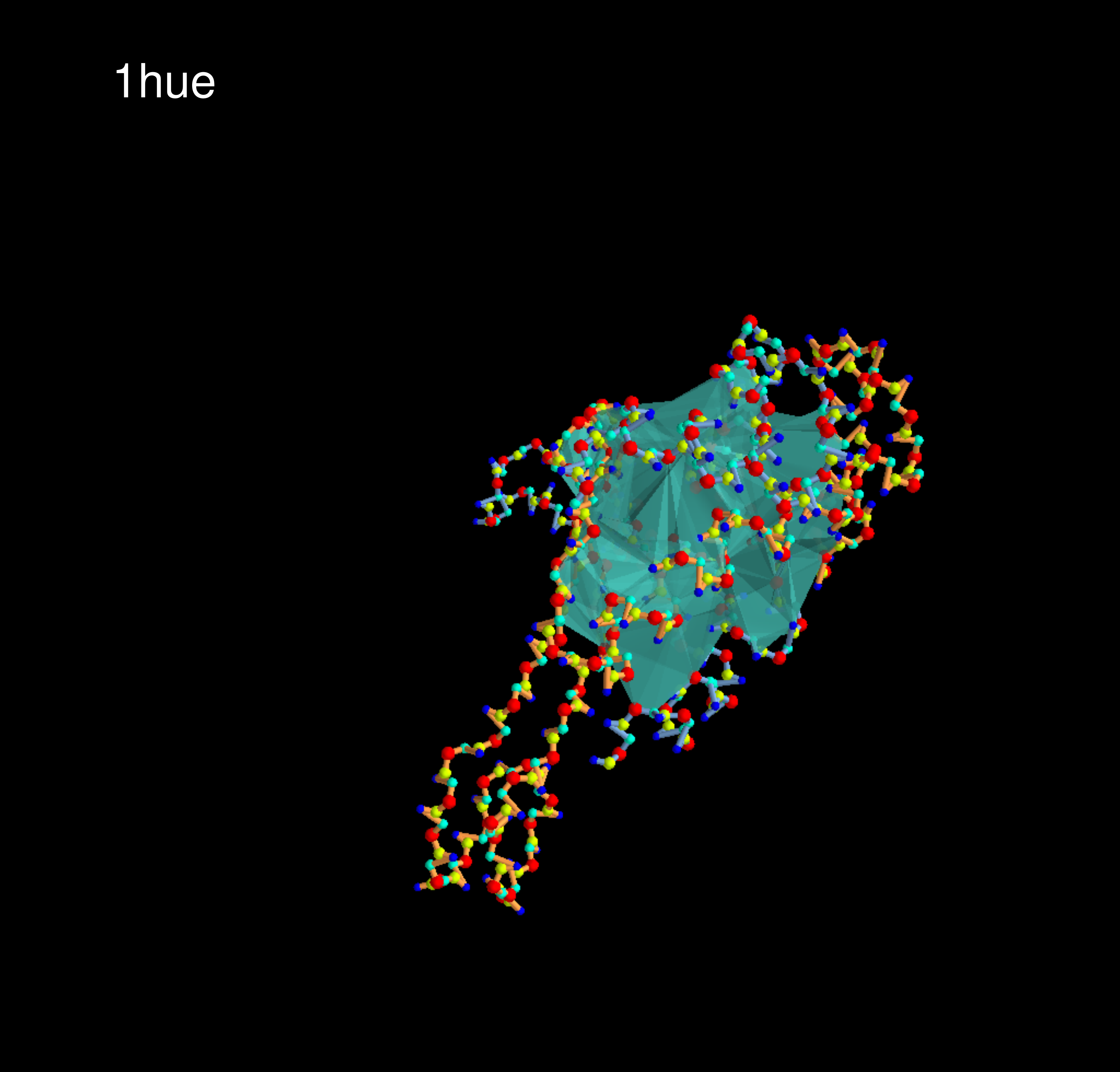}
        \captionsetup{labelformat=empty}
        \caption{}
        \label{supp_fig:pdb_1hue}
    \end{subfigure}
      \centering
      \begin{subfigure}{.48\textwidth} \centering
        \includegraphics[width=0.9\linewidth]{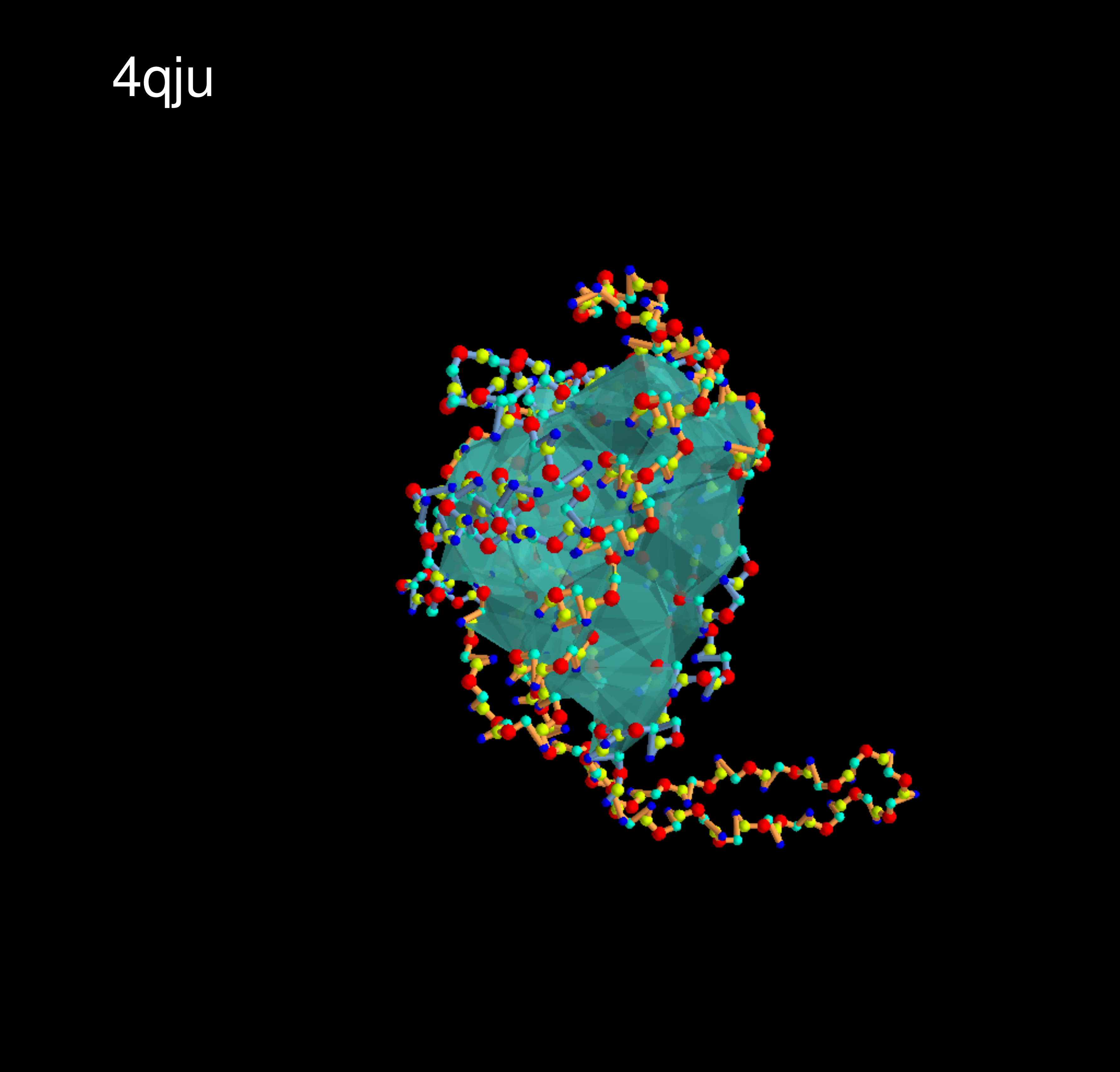}
        \captionsetup{labelformat=empty}
        \caption{}
        \label{supp_fig:pdb_4qju}
    \end{subfigure}
      \caption{}
      \label{fig:pdb_20}
    \end{figure}

    \begin{figure}[!tbhp]
      \centering
      \begin{subfigure}{.33\textwidth} \centering
        \includegraphics[width=0.9\linewidth]{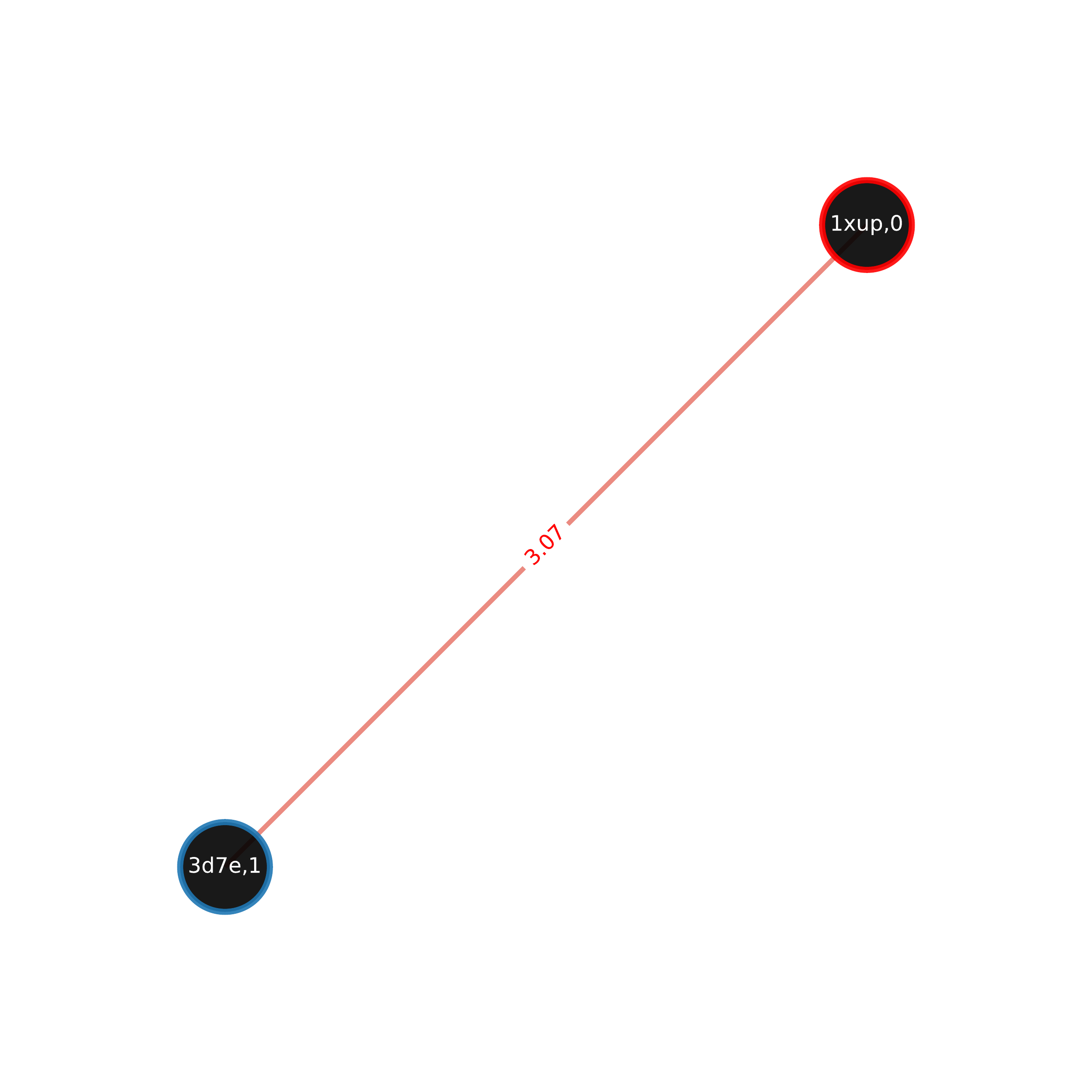}
        \captionsetup{labelformat=empty}
        \caption{}
        \label{supp_fig:pdb_hom_graph_21}
    \end{subfigure}
      \centering
      \begin{subfigure}{.33\textwidth} \centering
        \includegraphics[width=0.9\linewidth]{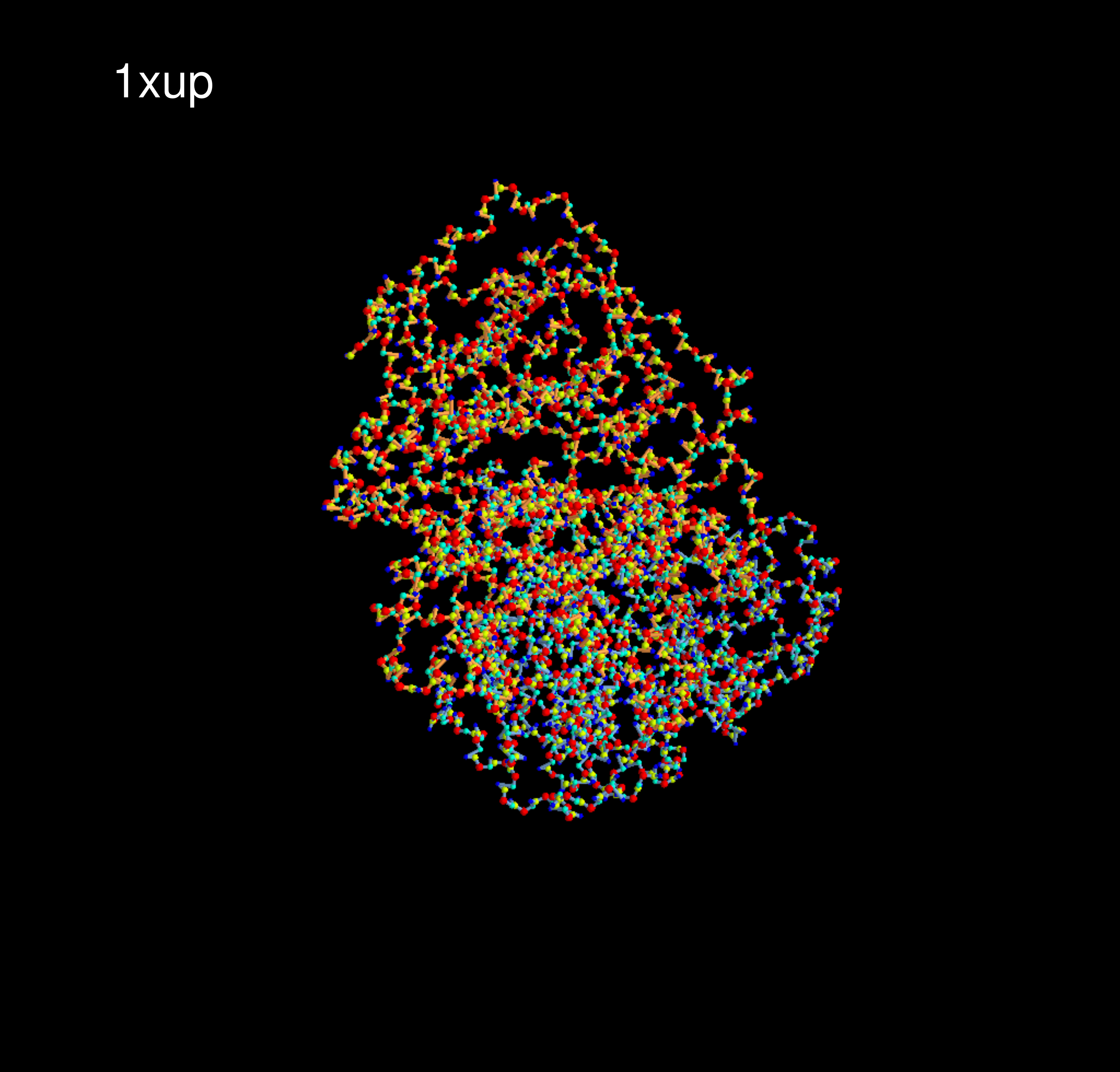}
        \captionsetup{labelformat=empty}
        \caption{}
        \label{supp_fig:pdb_1xup}
    \end{subfigure}
      \centering
      \begin{subfigure}{.33\textwidth} \centering
        \includegraphics[width=0.9\linewidth]{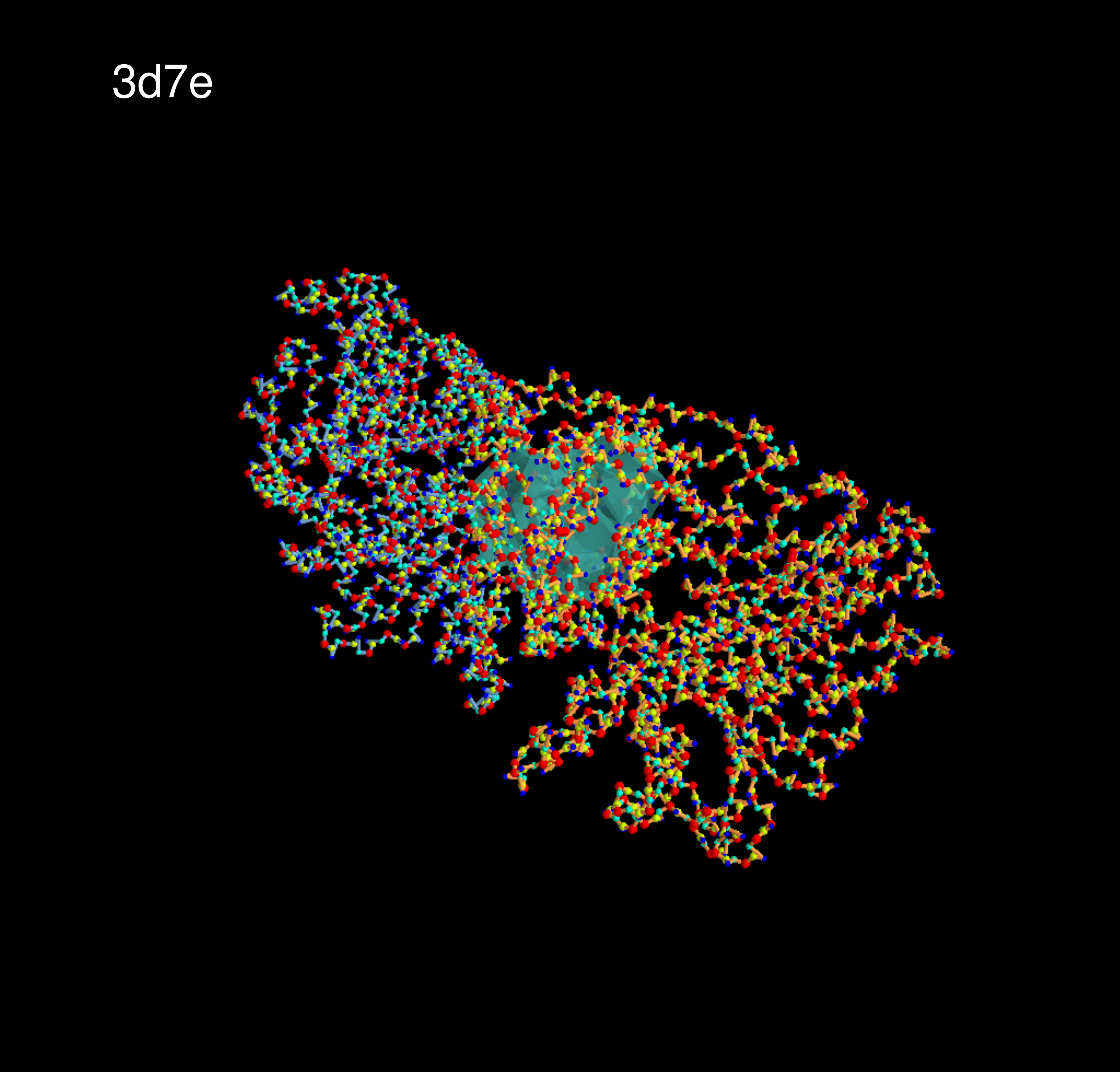}
        \captionsetup{labelformat=empty}
        \caption{}
        \label{supp_fig:pdb_3d7e}
    \end{subfigure}
      \caption{}
      \label{fig:pdb_21}
    \end{figure}

    \begin{figure}[!tbhp]
      \centering
      \begin{subfigure}{.33\textwidth} \centering
        \includegraphics[width=0.9\linewidth]{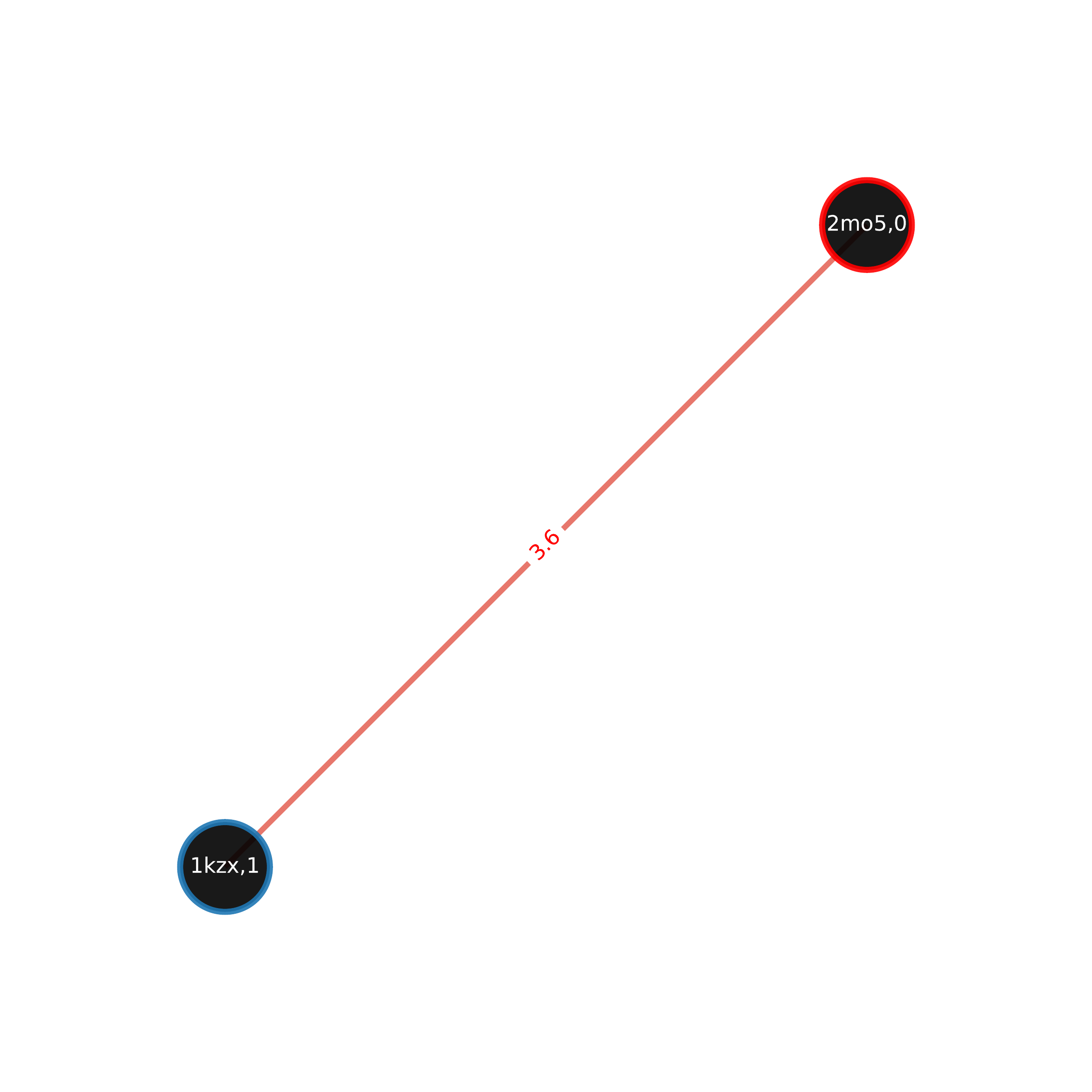}
        \captionsetup{labelformat=empty}
        \caption{}
        \label{supp_fig:pdb_hom_graph_22}
    \end{subfigure}
      \centering
      \begin{subfigure}{.33\textwidth} \centering
        \includegraphics[width=0.9\linewidth]{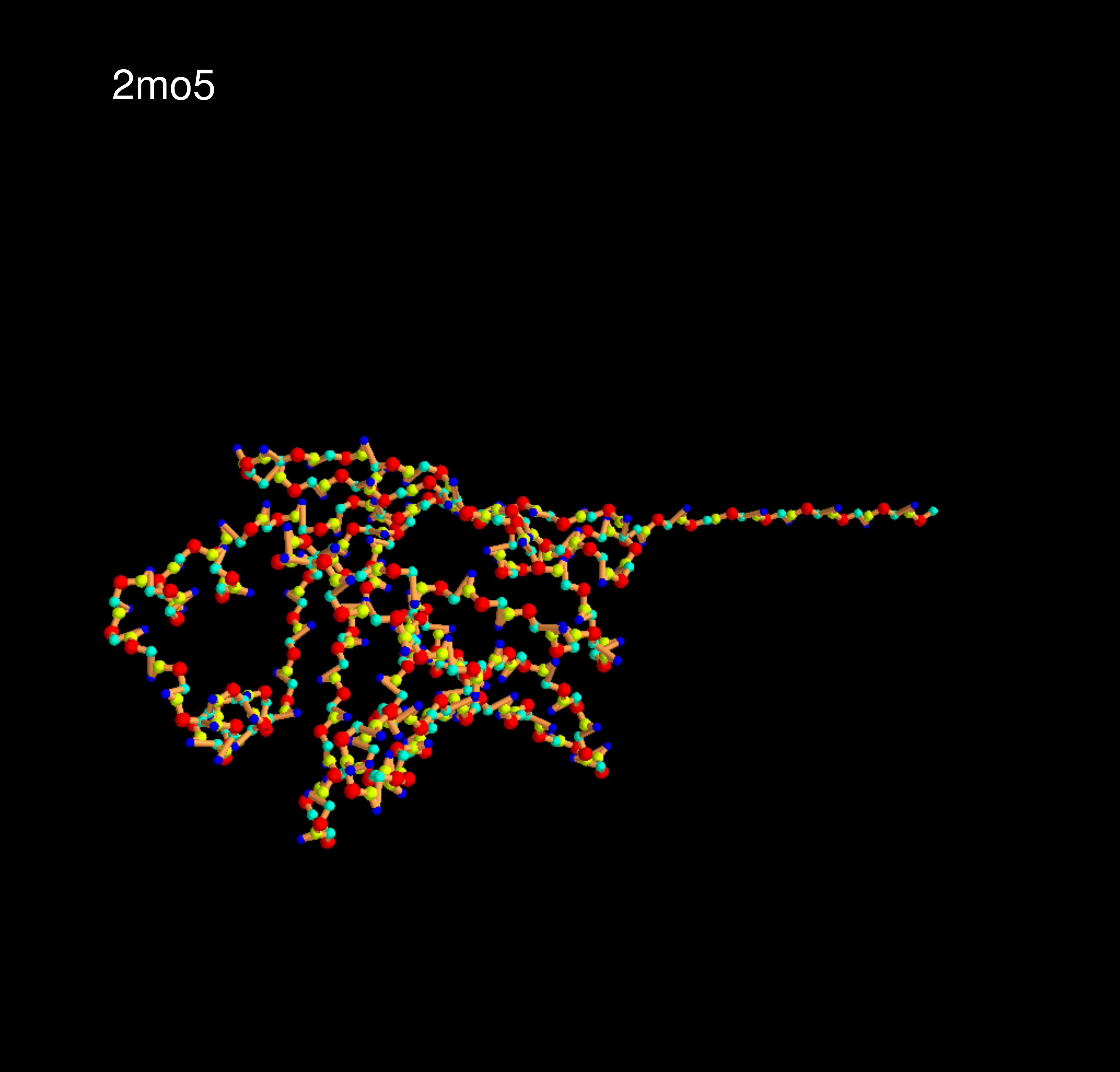}
        \captionsetup{labelformat=empty}
        \caption{}
        \label{supp_fig:pdb_2mo5}
    \end{subfigure}
      \centering
      \begin{subfigure}{.33\textwidth} \centering
        \includegraphics[width=0.9\linewidth]{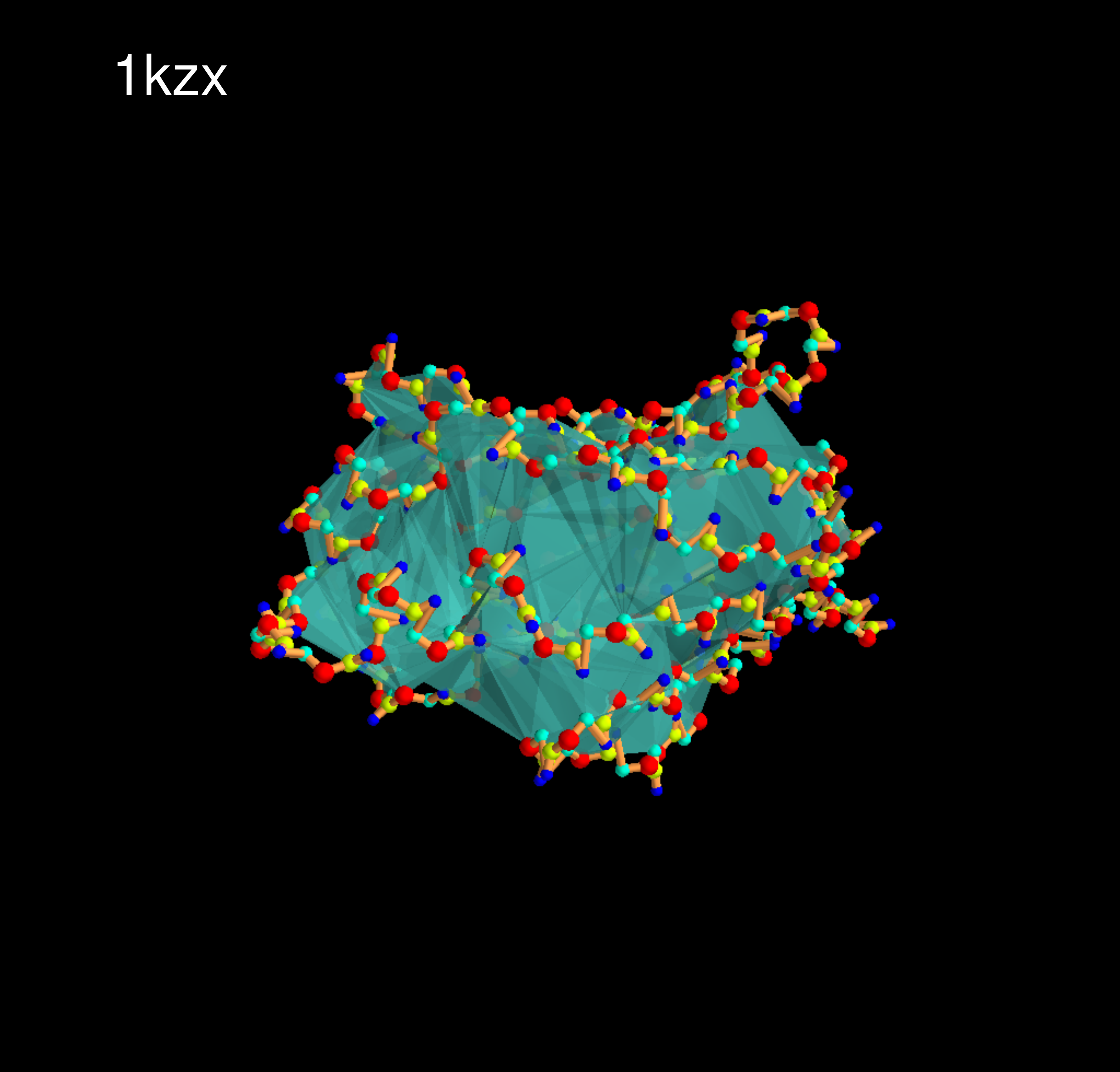}
        \captionsetup{labelformat=empty}
        \caption{}
        \label{supp_fig:pdb_1kzx}
    \end{subfigure}
      \caption{}
      \label{fig:pdb_22}
    \end{figure}

    \begin{figure}[!tbhp]
      \centering
      \begin{subfigure}{.48\textwidth} \centering
        \includegraphics[width=0.9\linewidth]{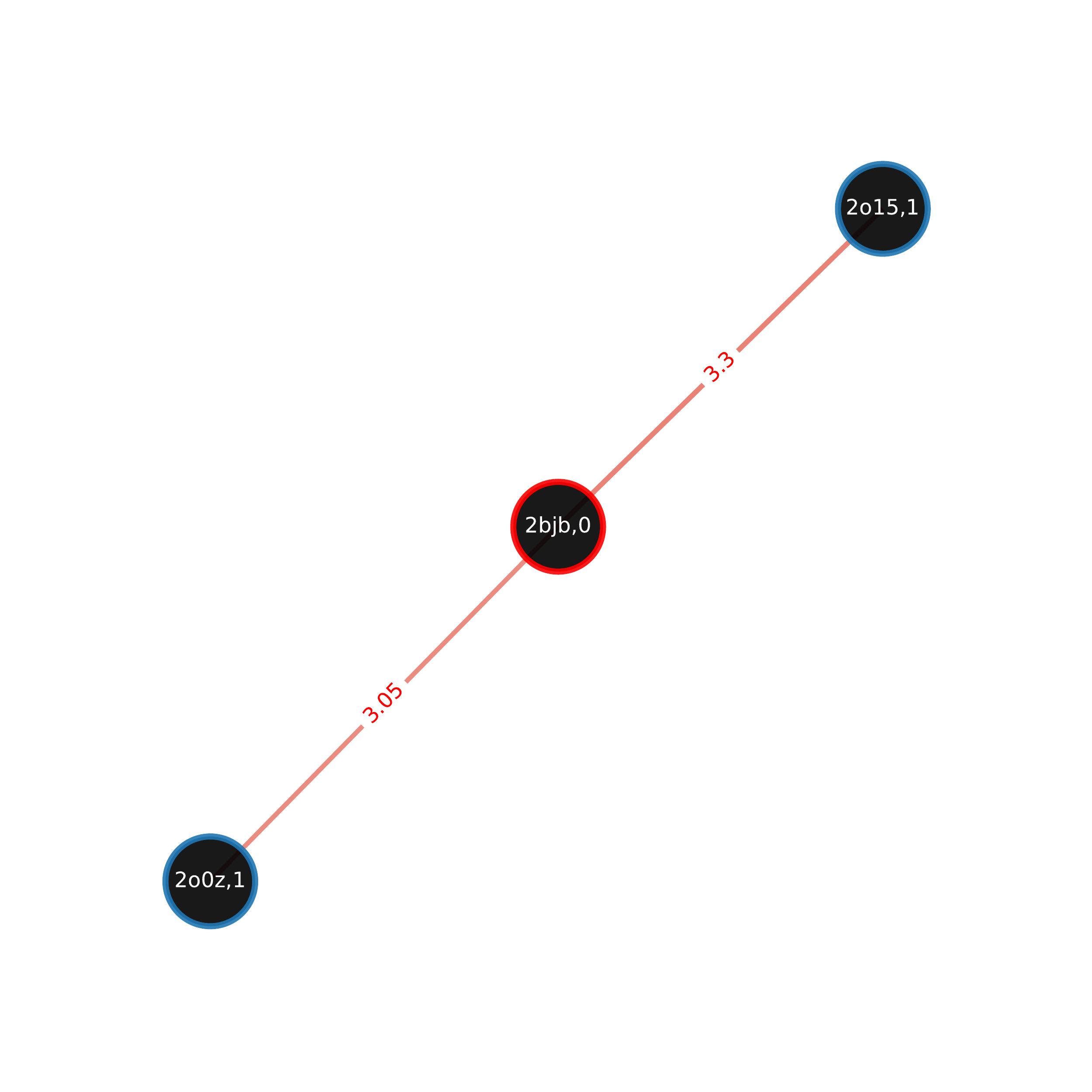}
        \captionsetup{labelformat=empty}
        \caption{}
        \label{supp_fig:pdb_hom_graph_23}
    \end{subfigure}
      \centering
      \begin{subfigure}{.48\textwidth} \centering
        \includegraphics[width=0.9\linewidth]{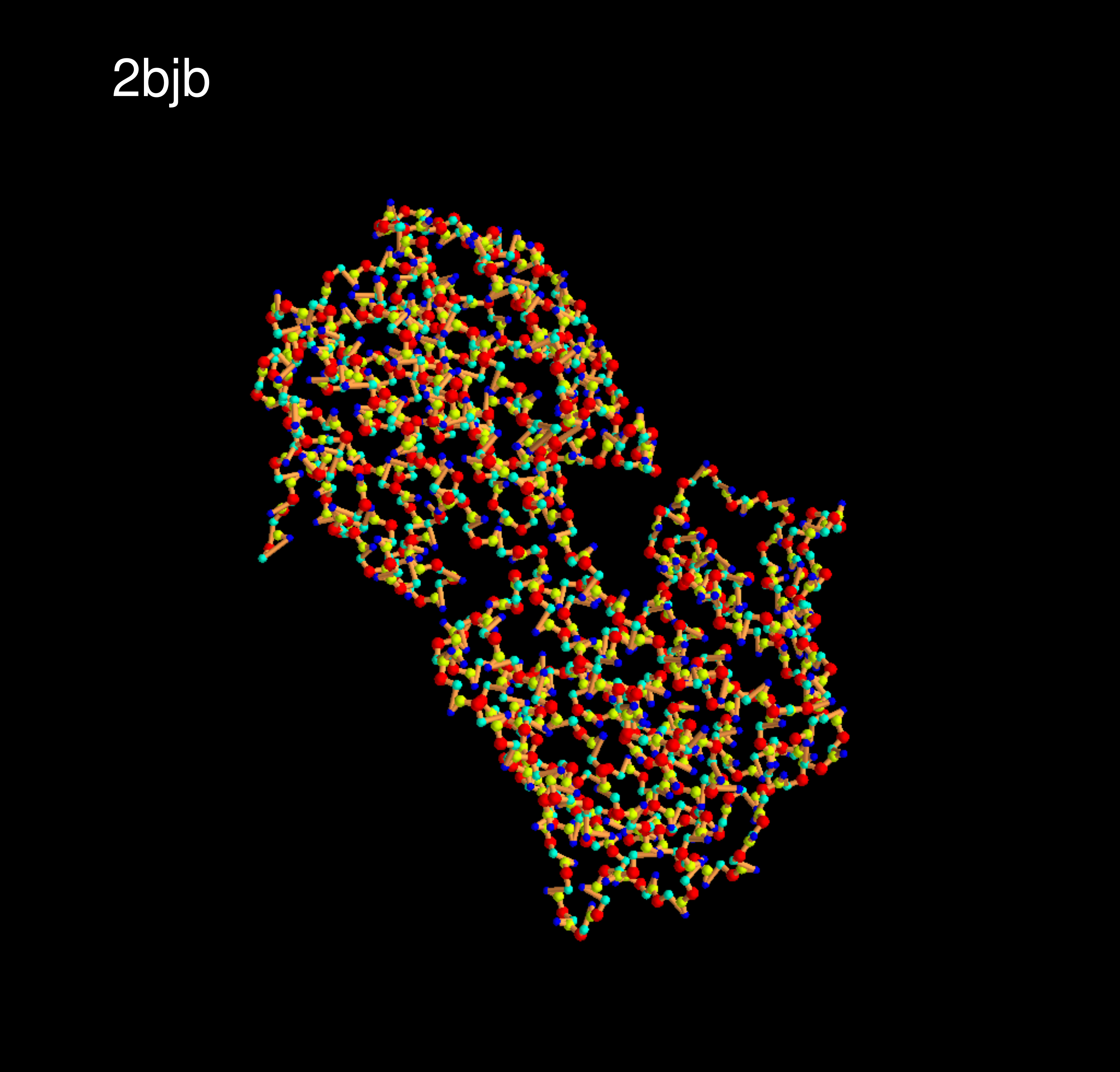}
        \captionsetup{labelformat=empty}
        \caption{}
        \label{supp_fig:pdb_2bjb}
    \end{subfigure}
      \centering
      \begin{subfigure}{.48\textwidth} \centering
        \includegraphics[width=0.9\linewidth]{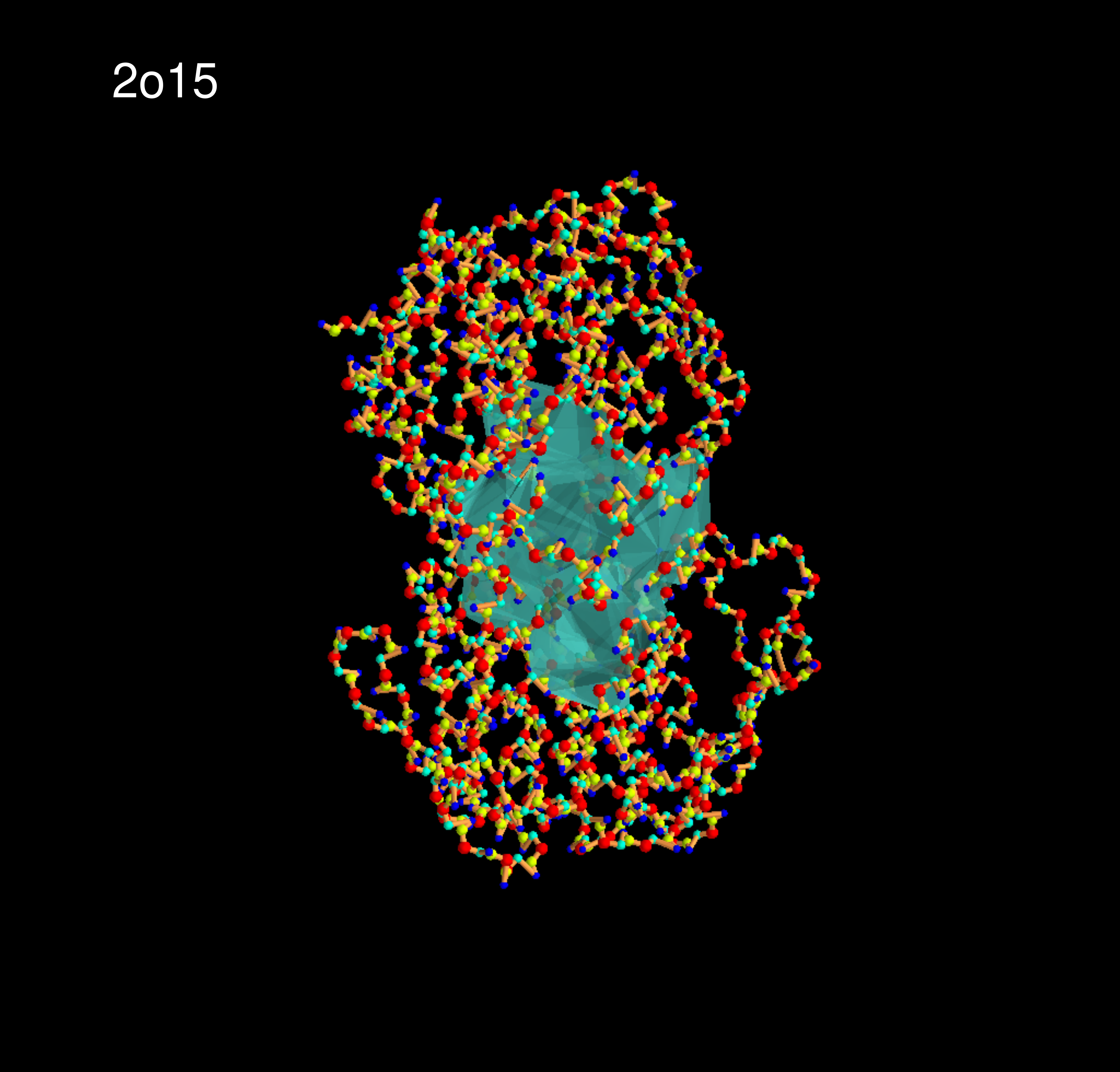}
        \captionsetup{labelformat=empty}
        \caption{}
        \label{supp_fig:pdb_2o15}
    \end{subfigure}
      \centering
      \begin{subfigure}{.48\textwidth} \centering
        \includegraphics[width=0.9\linewidth]{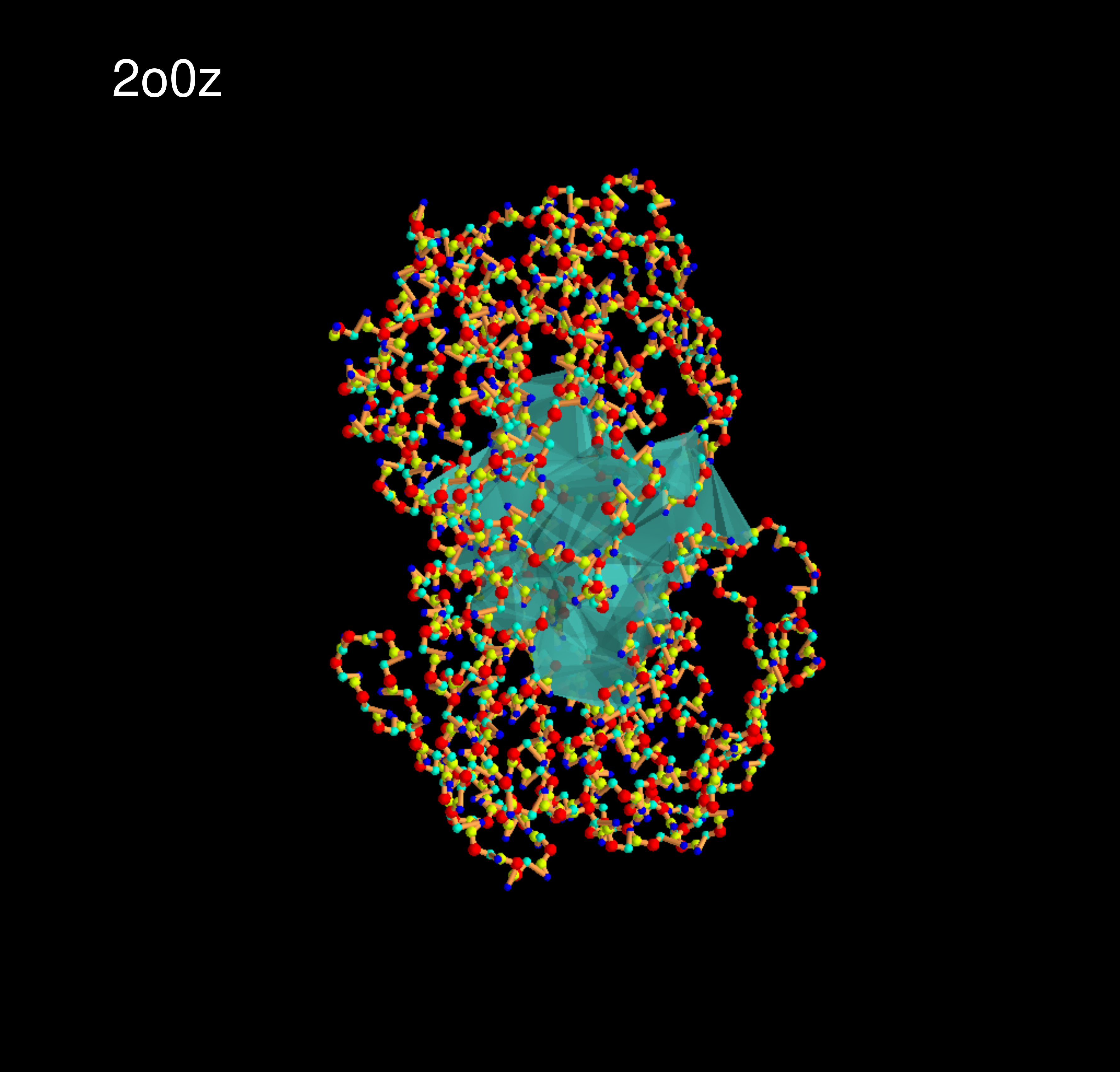}
        \captionsetup{labelformat=empty}
        \caption{}
        \label{supp_fig:pdb_2o0z}
    \end{subfigure}
      \caption{}
      \label{fig:pdb_23}
    \end{figure}

    \begin{figure}[!tbhp]
      \centering
      \begin{subfigure}{.33\textwidth} \centering
        \includegraphics[width=0.9\linewidth]{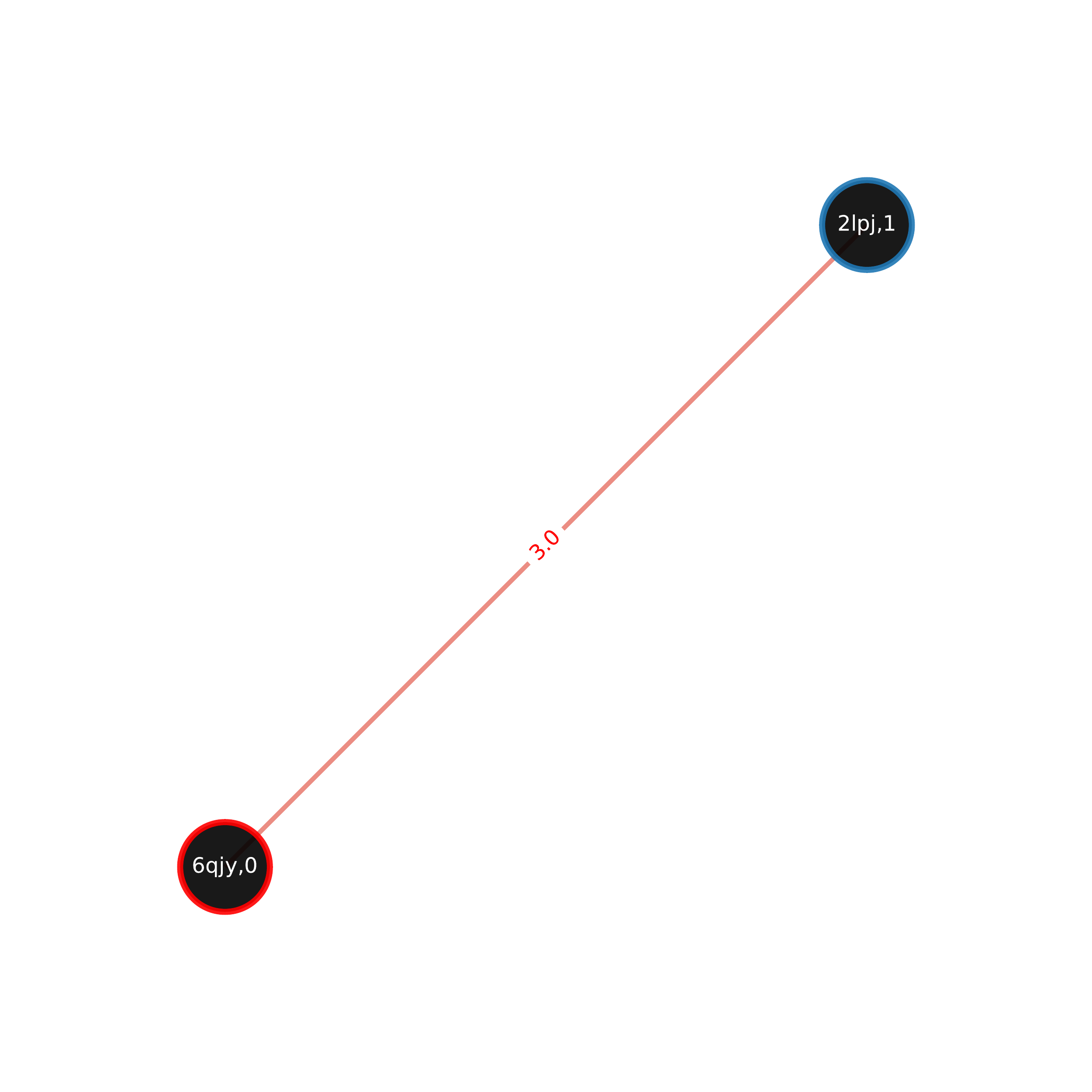}
        \captionsetup{labelformat=empty}
        \caption{}
        \label{supp_fig:pdb_hom_graph_24}
    \end{subfigure}
      \centering
      \begin{subfigure}{.33\textwidth} \centering
        \includegraphics[width=0.9\linewidth]{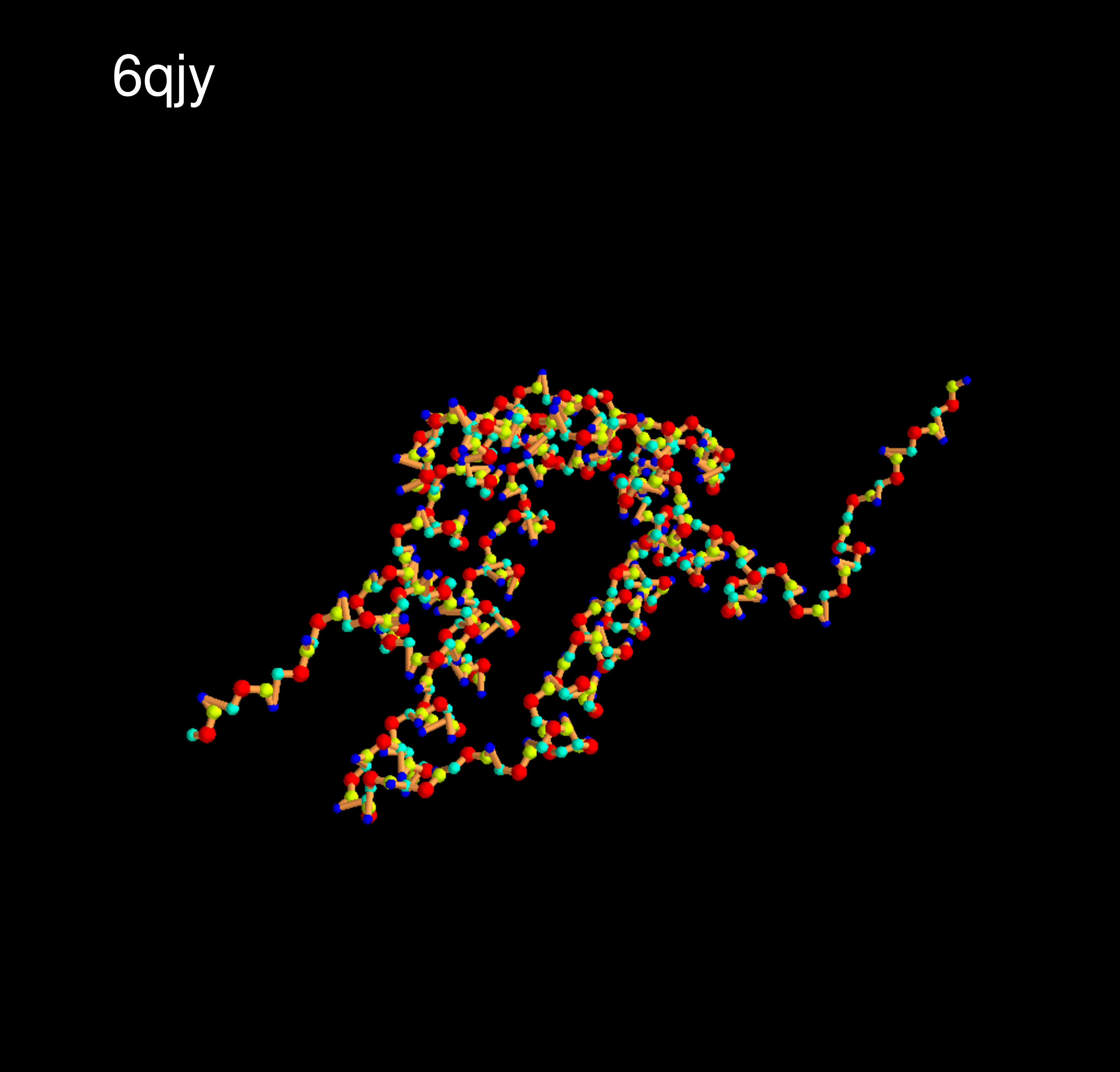}
        \captionsetup{labelformat=empty}
        \caption{}
        \label{supp_fig:pdb_6qjy}
    \end{subfigure}
      \centering
      \begin{subfigure}{.33\textwidth} \centering
        \includegraphics[width=0.9\linewidth]{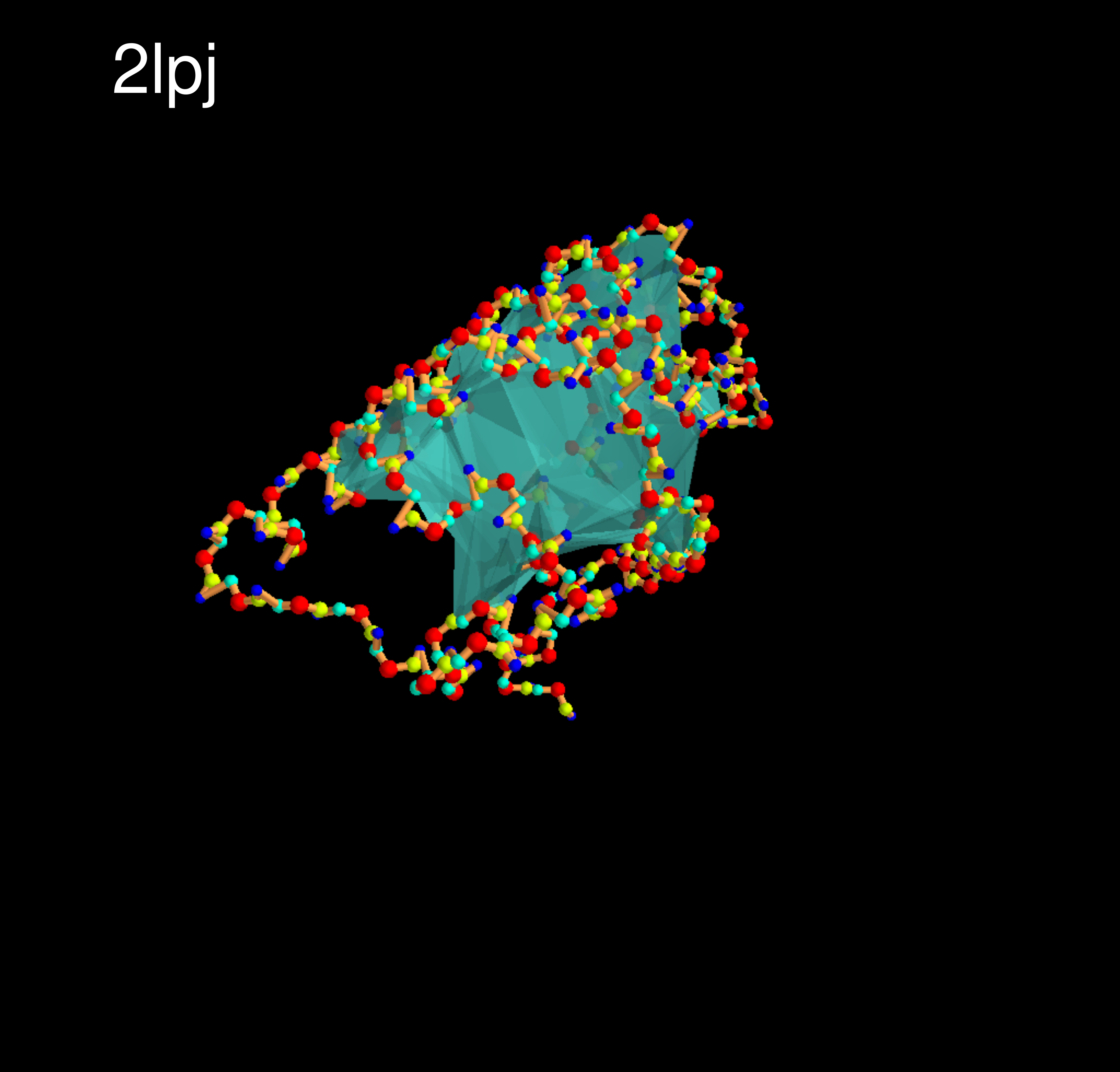}
        \captionsetup{labelformat=empty}
        \caption{}
        \label{supp_fig:pdb_2lpj}
    \end{subfigure}
      \caption{}
      \label{fig:pdb_24}
    \end{figure}

    \begin{figure}[!tbhp]
      \centering
      \begin{subfigure}{.24\textwidth} \centering
        \includegraphics[width=0.9\linewidth]{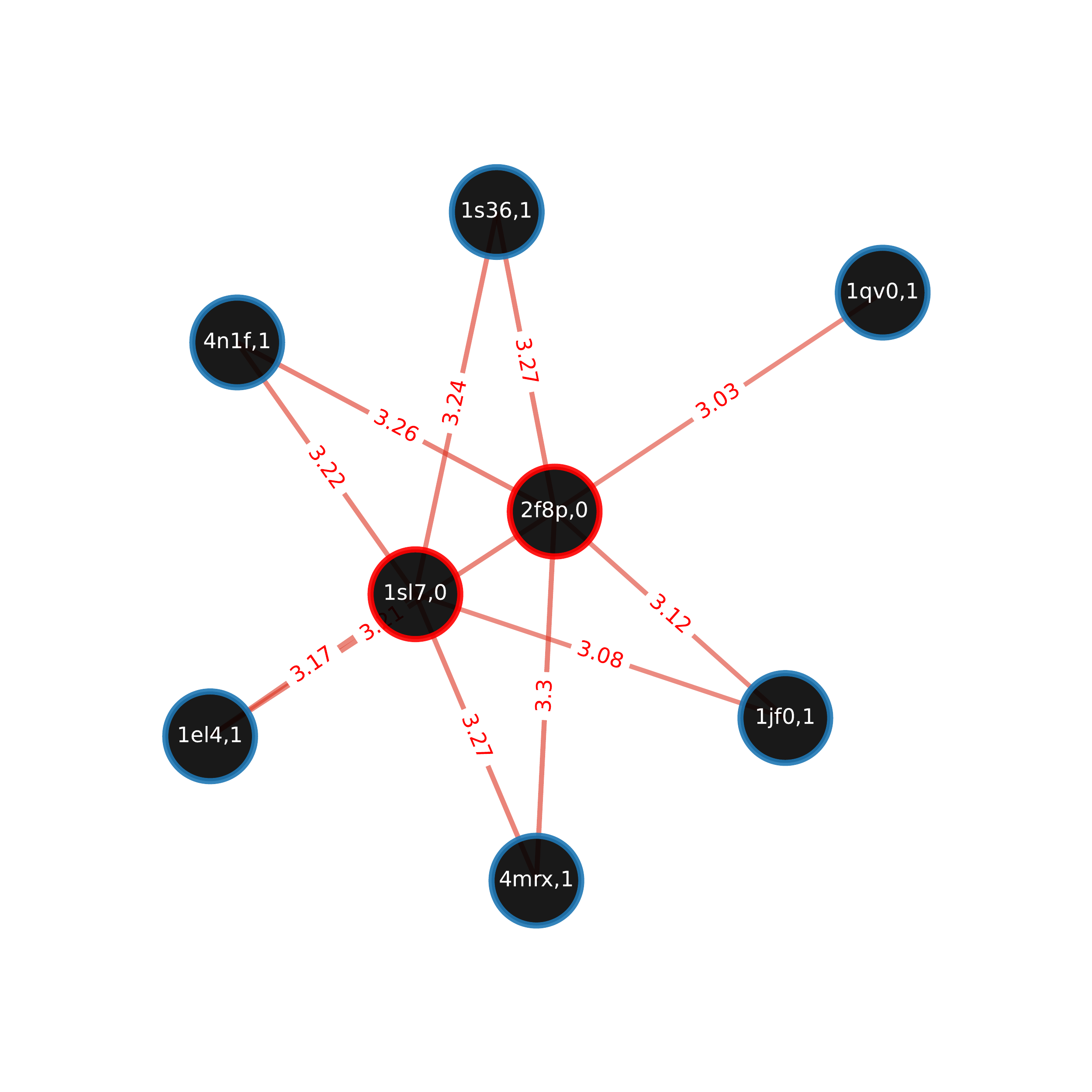}
        \captionsetup{labelformat=empty}
        \caption{}
        \label{supp_fig:pdb_hom_graph_25}
    \end{subfigure}
      \centering
      \begin{subfigure}{.24\textwidth} \centering
        \includegraphics[width=0.9\linewidth]{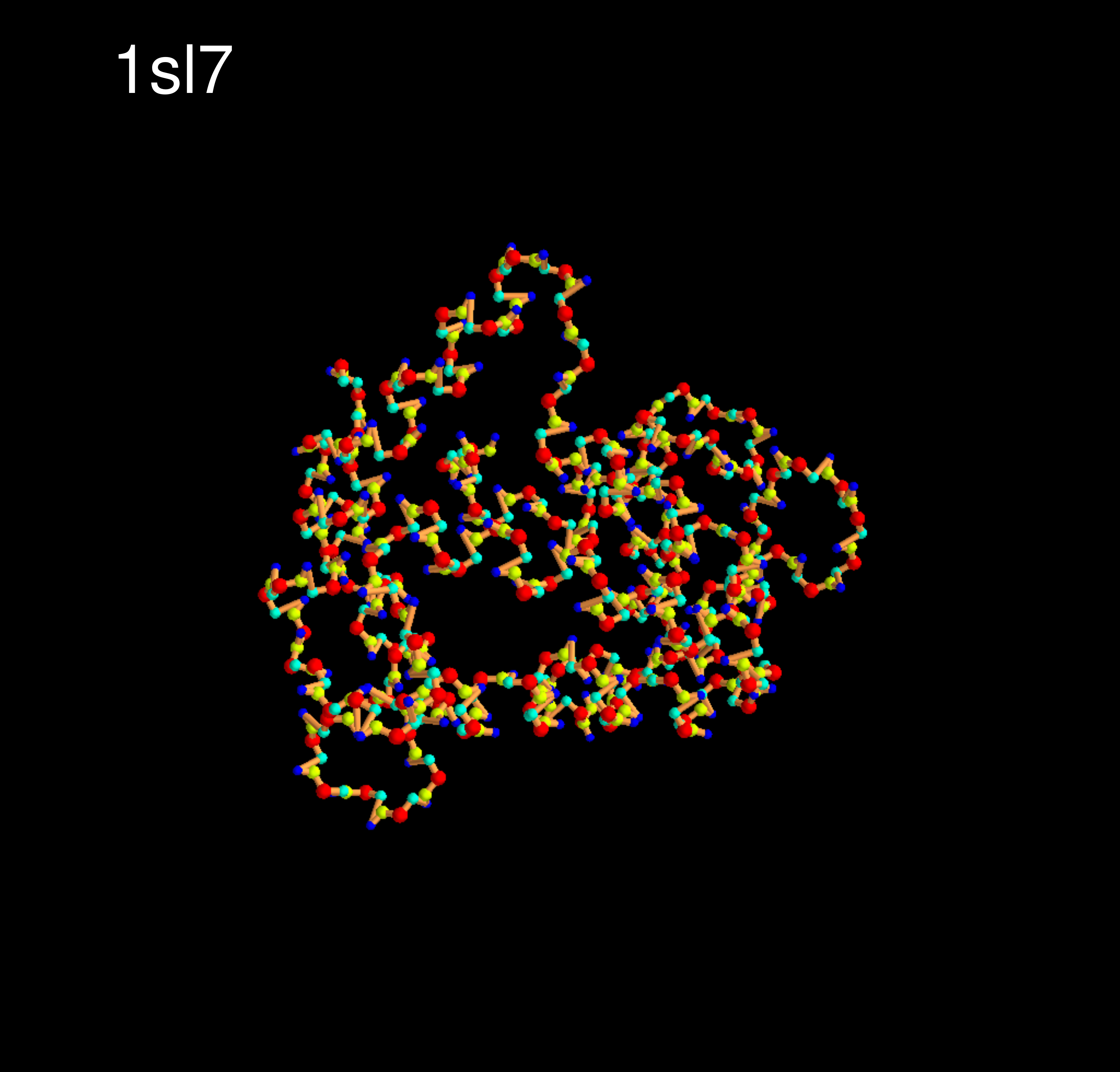}
        \captionsetup{labelformat=empty}
        \caption{}
        \label{supp_fig:pdb_1sl7}
    \end{subfigure}
      \centering
      \begin{subfigure}{.24\textwidth} \centering
        \includegraphics[width=0.9\linewidth]{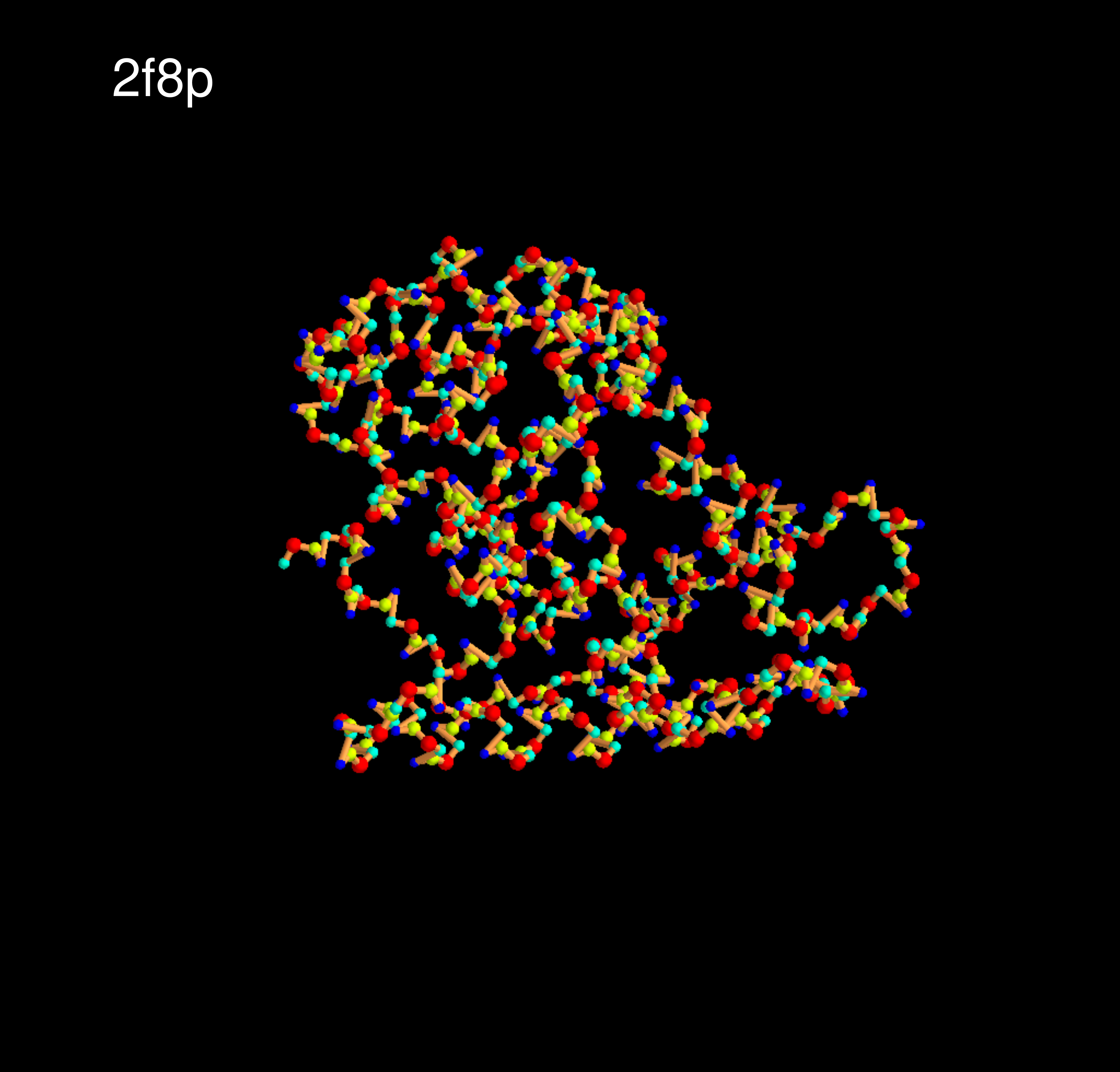}
        \captionsetup{labelformat=empty}
        \caption{}
        \label{supp_fig:pdb_2f8p}
    \end{subfigure}
      \centering
      \begin{subfigure}{.24\textwidth} \centering
        \includegraphics[width=0.9\linewidth]{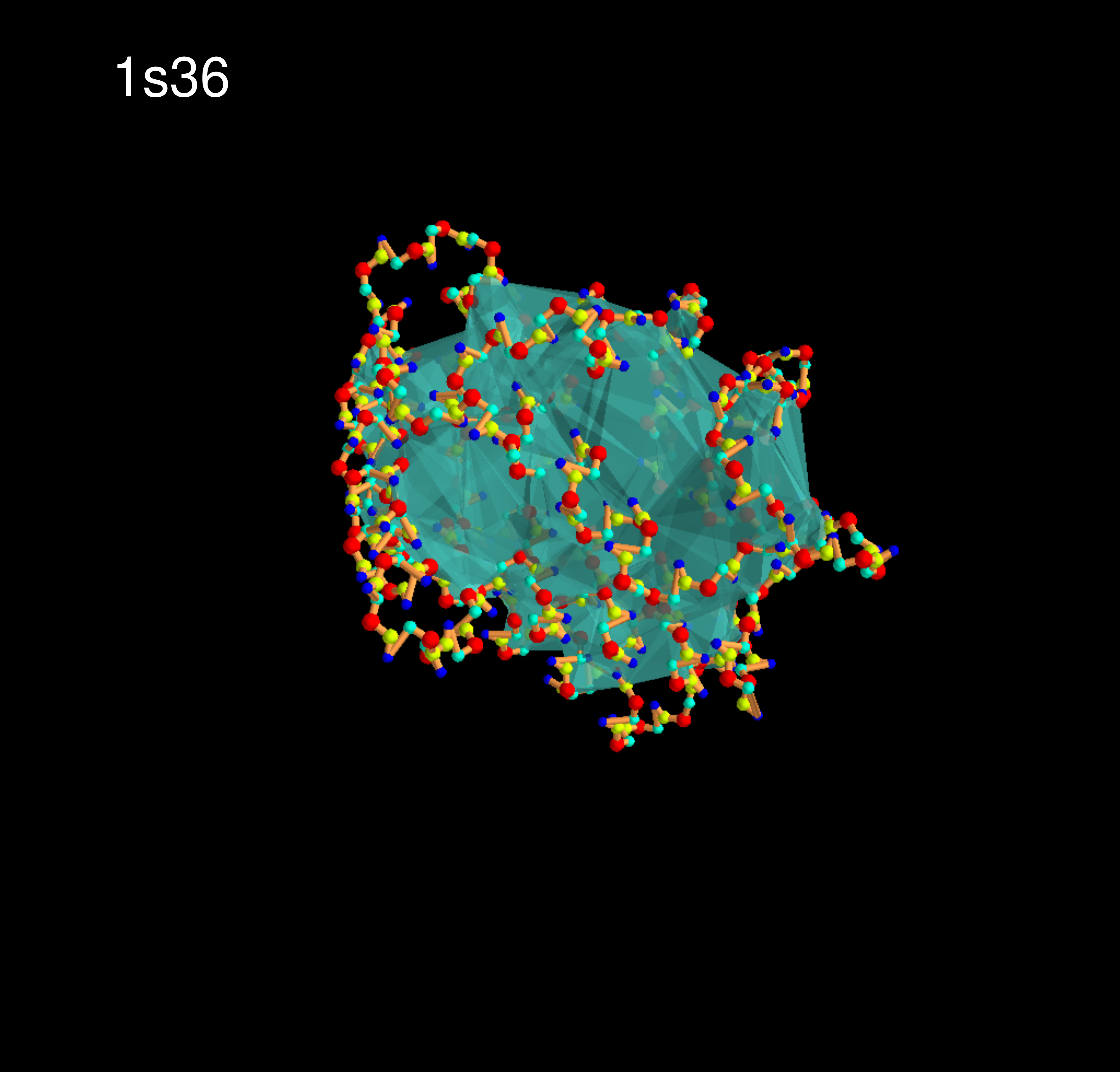}
        \captionsetup{labelformat=empty}
        \caption{}
        \label{supp_fig:pdb_1s36}
    \end{subfigure}
      \centering
      \begin{subfigure}{.24\textwidth} \centering
        \includegraphics[width=0.9\linewidth]{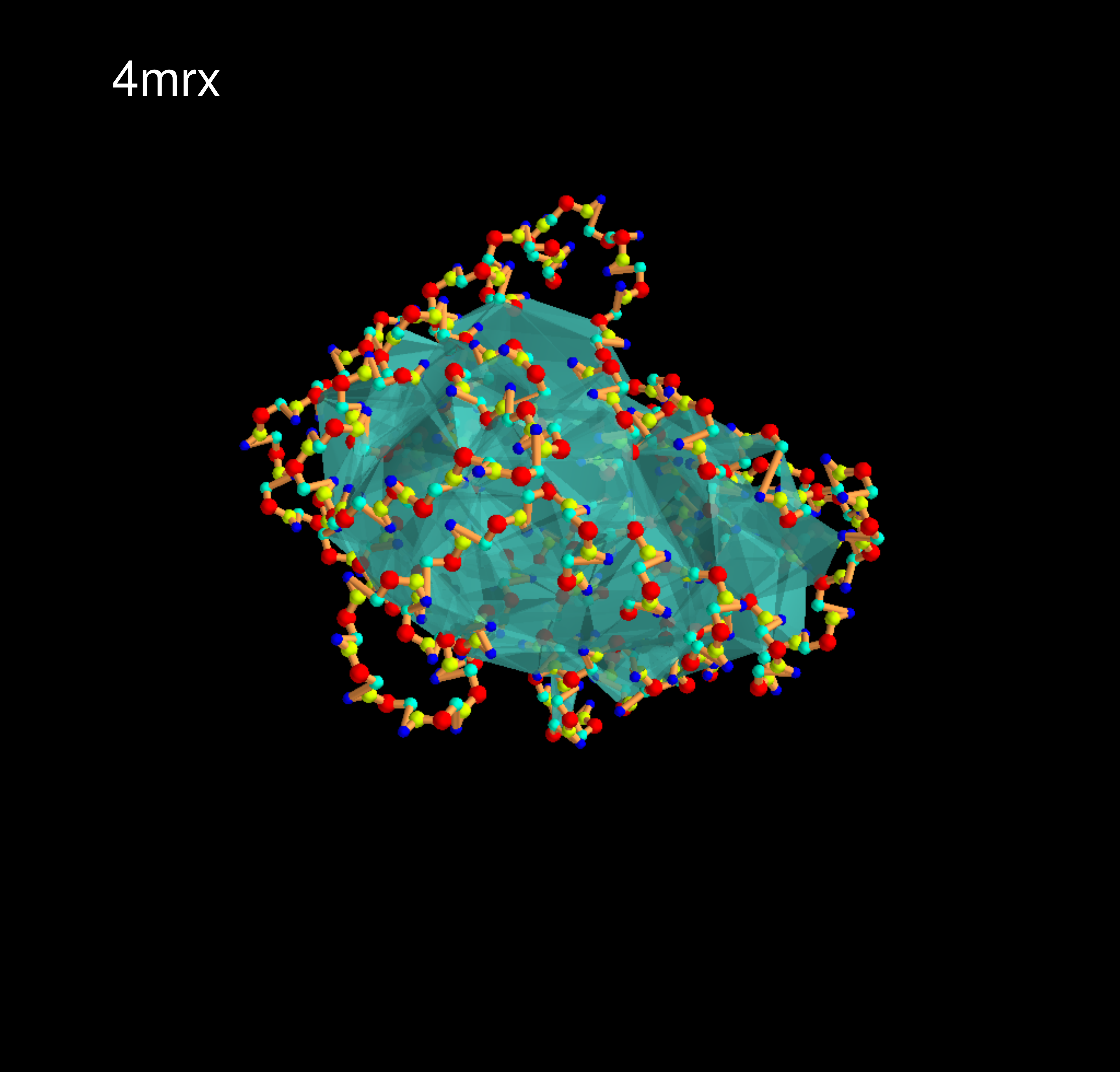}
        \captionsetup{labelformat=empty}
        \caption{}
        \label{supp_fig:pdb_4mrx}
    \end{subfigure}
      \centering
      \begin{subfigure}{.24\textwidth} \centering
        \includegraphics[width=0.9\linewidth]{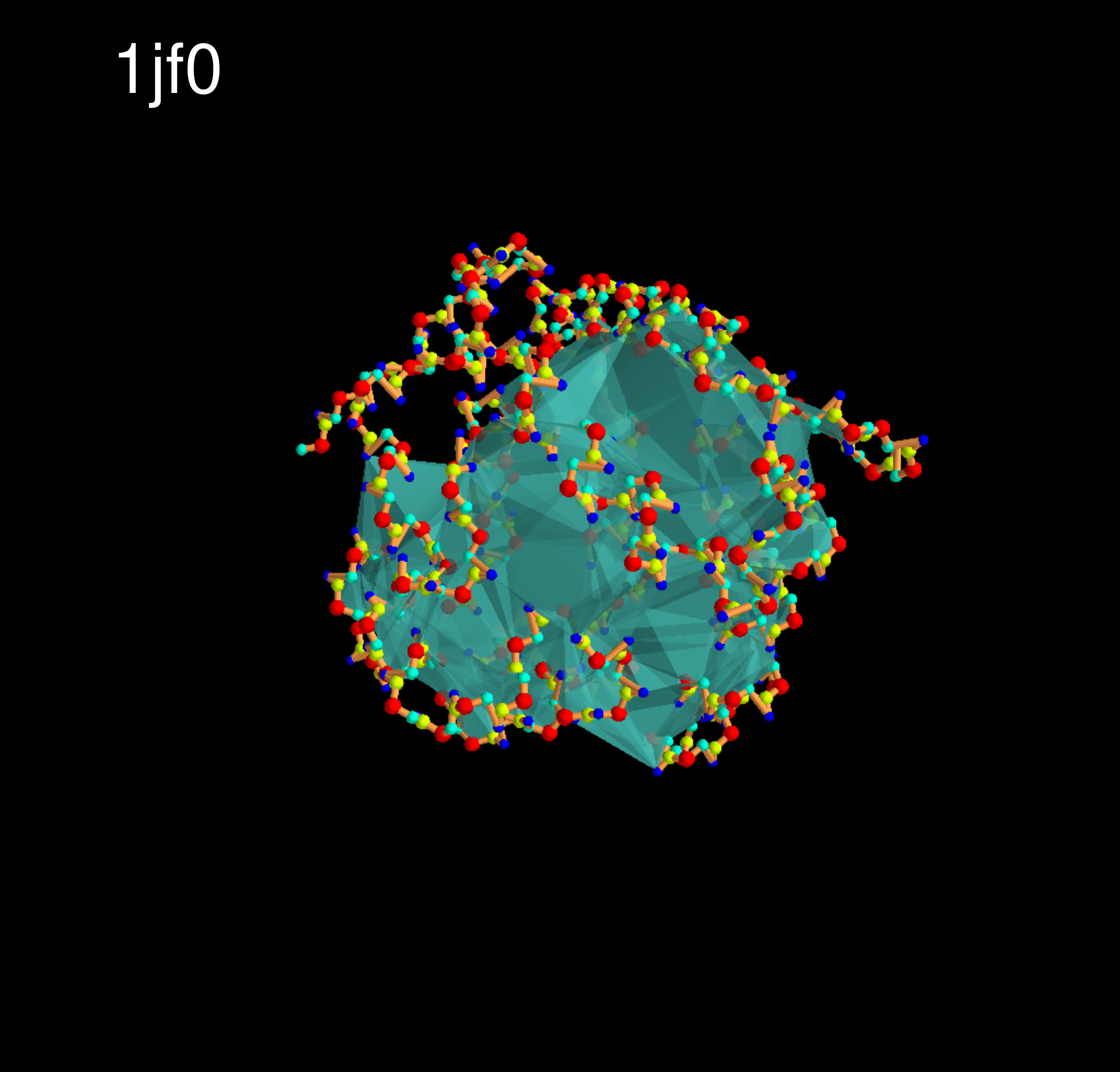}
        \captionsetup{labelformat=empty}
        \caption{}
        \label{supp_fig:pdb_1jf0}
    \end{subfigure}
      \centering
      \begin{subfigure}{.24\textwidth} \centering
        \includegraphics[width=0.9\linewidth]{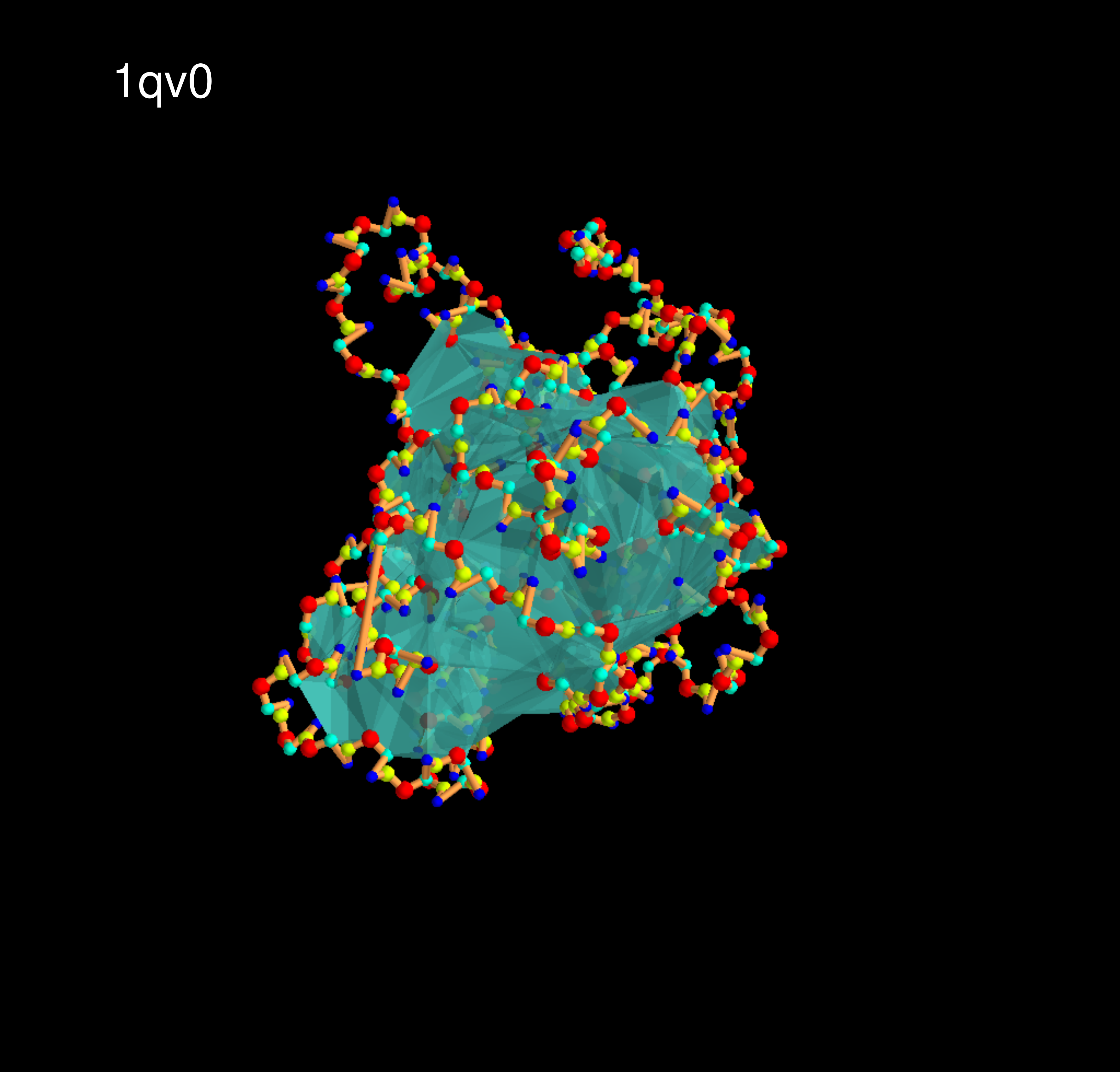}
        \captionsetup{labelformat=empty}
        \caption{}
        \label{supp_fig:pdb_1qv0}
    \end{subfigure}
      \centering
      \begin{subfigure}{.24\textwidth} \centering
        \includegraphics[width=0.9\linewidth]{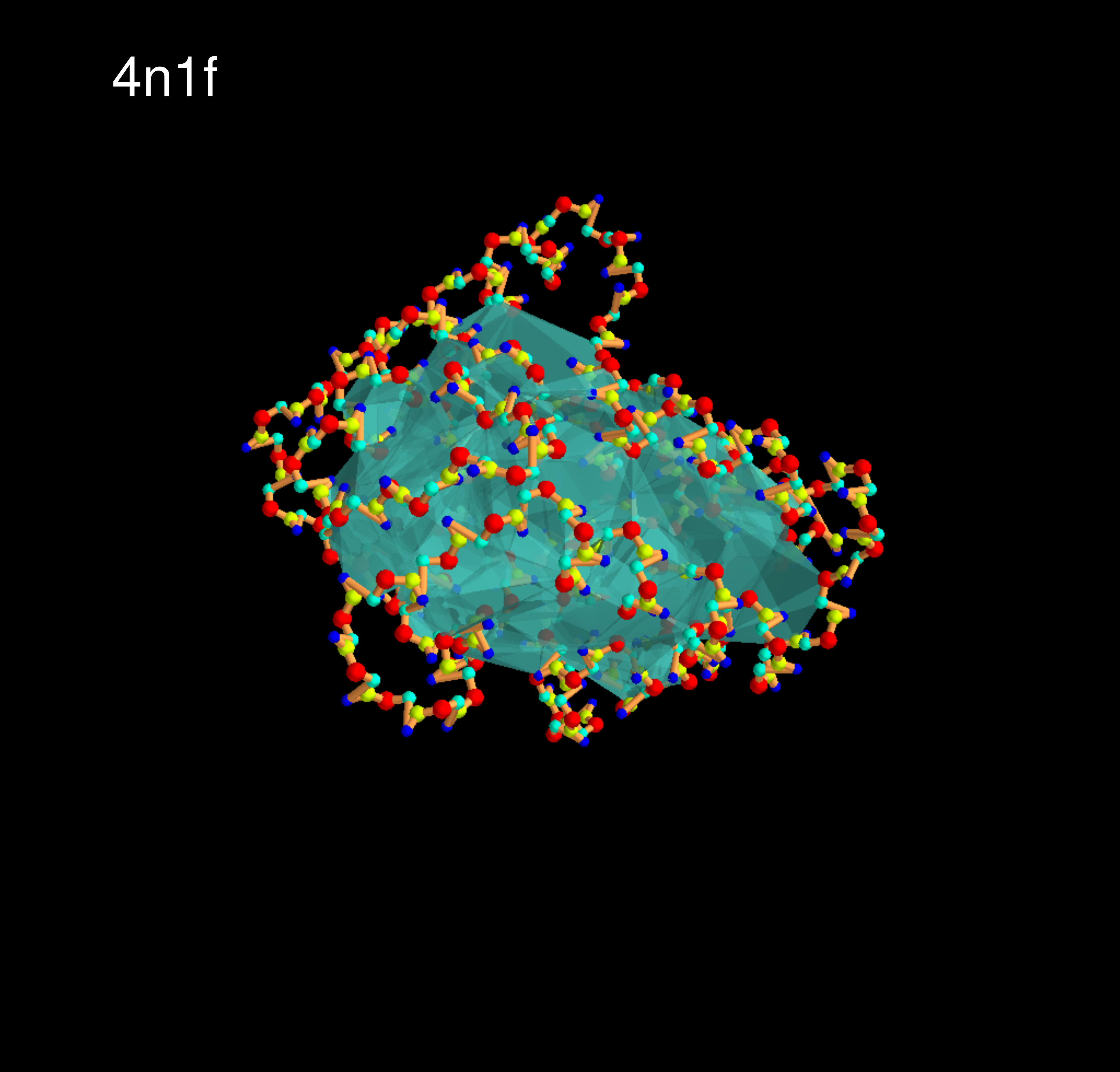}
        \captionsetup{labelformat=empty}
        \caption{}
        \label{supp_fig:pdb_4n1f}
    \end{subfigure}
      \centering
      \begin{subfigure}{.24\textwidth} \centering
        \includegraphics[width=0.9\linewidth]{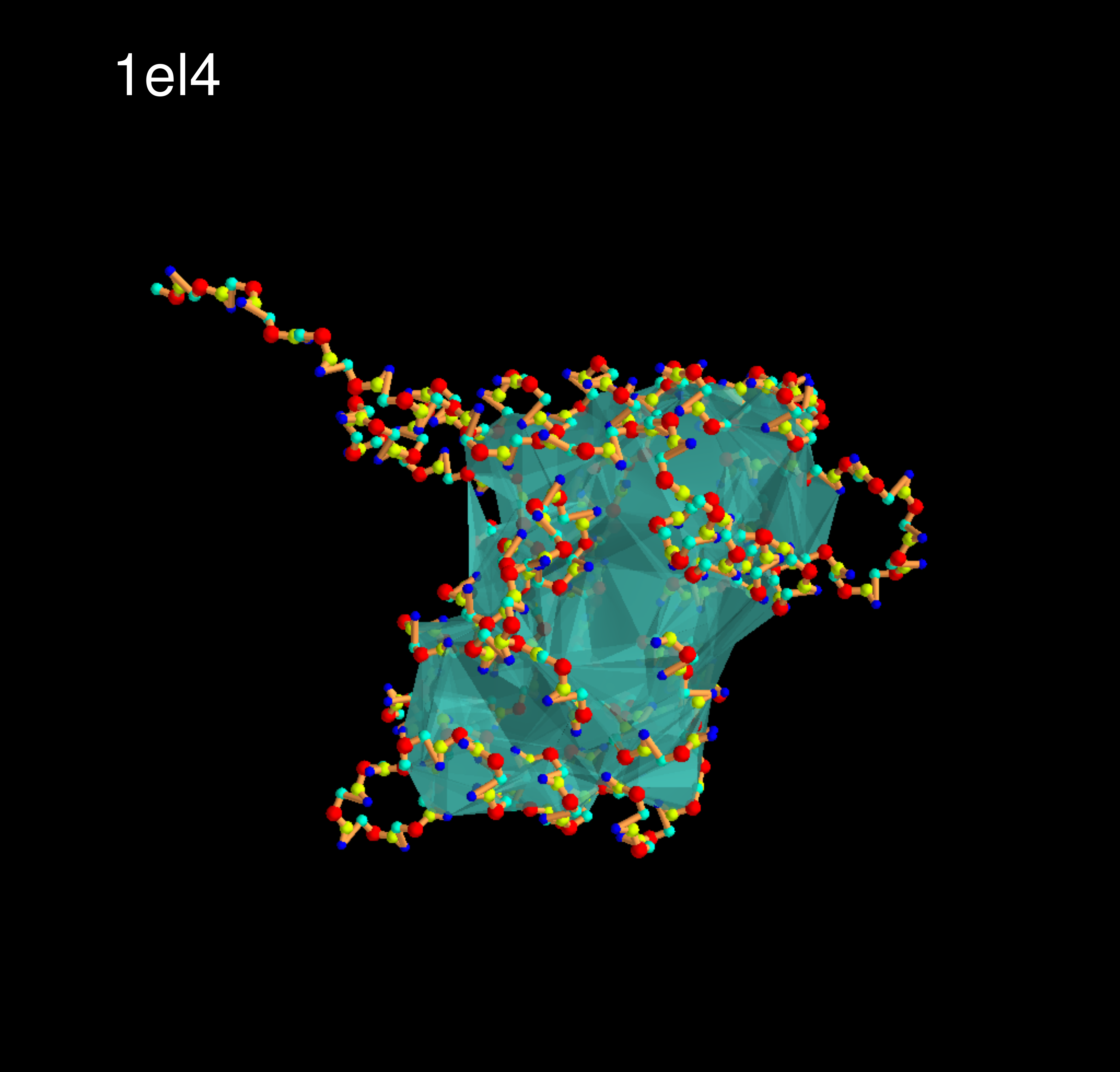}
        \captionsetup{labelformat=empty}
        \caption{}
        \label{supp_fig:pdb_1el4}
    \end{subfigure}
      \caption{}
      \label{fig:pdb_25}
    \end{figure}

\section*{Acknowledgement}

This research was supported by the Intramural Research Program of the 
NIH, the National Institute of Diabetes and Digestive and Kidney 
Diseases (NIDDK). This work utilized the computational resources of the NIH HPC Biowulf cluster
(http://hpc.nih.gov).

\newpage

\appendix

\begin{appendices}

\section{Serial-parallel homology computation}\label{app:serial_parallel}

\begin{figure}[tbhp]
  \centering
  \begin{subfigure}{0.49\linewidth} 
    \centering
  \includegraphics[width=0.95\textwidth]{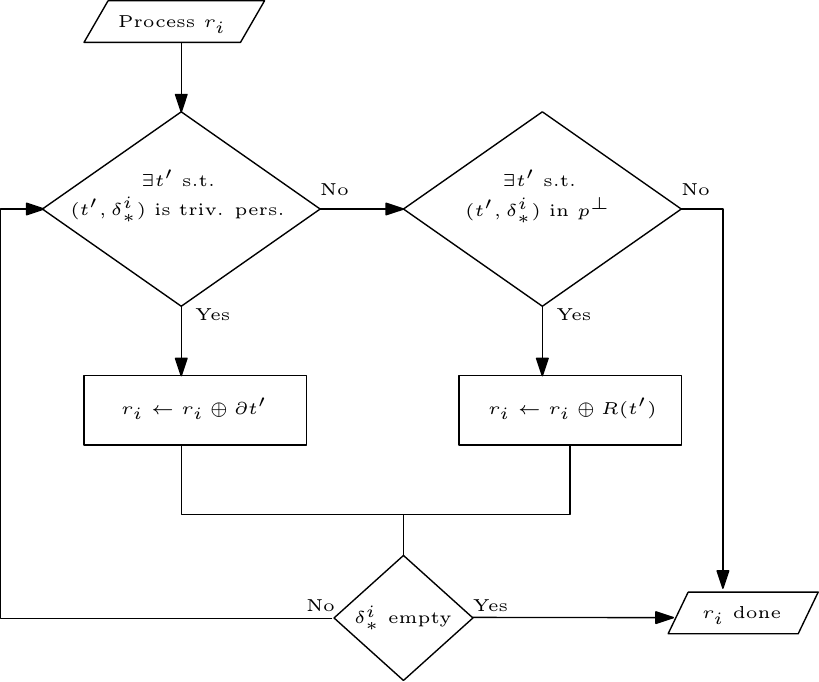}

  \caption{}

\label{fig:parallel_flowchart_hom}
\end{subfigure}
\centering
\begin{subfigure}{0.49\linewidth}
  \centering
  \includegraphics[width=0.95\textwidth]{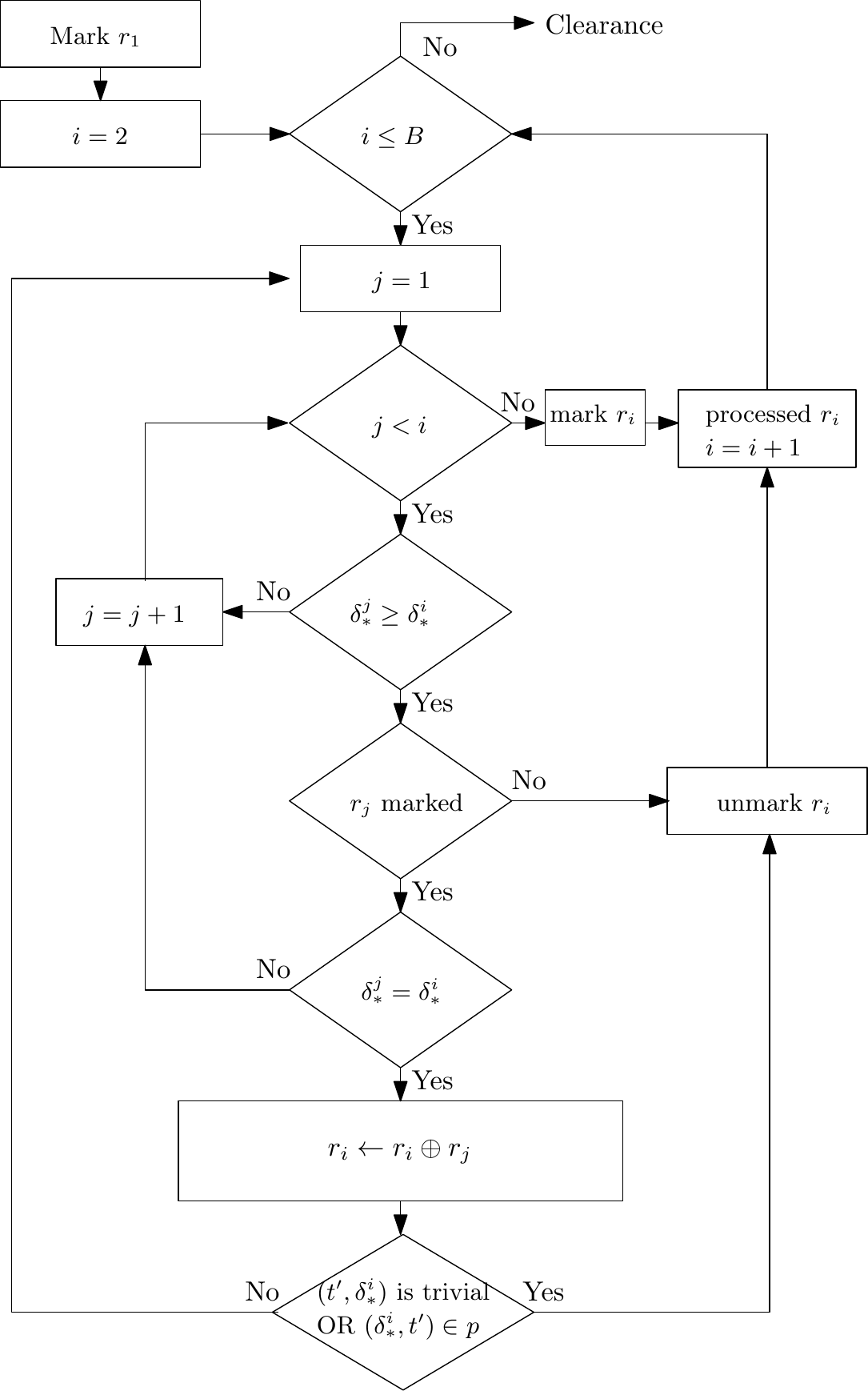} 

  \caption{}

\label{fig:serial_flowchart_hom}
\end{subfigure}

  \caption{Flowcharts for parallel and serial reductions that form the serial-parallel algorithm.
  (a) Parallel reduction: Each $r_i$ is reduced with $R$ independently, in parallel. (b) Serial
  reduction: After parallel reduction, columns in $\mathbf{r}$ are reduced with each other.}

  \label{fig:serial_parallel_flowcharts}
\end{figure}

\end{appendices}

\newpage
\bibliographystyle{unsrtnat}
\bibliography{PH_refs}

\end{document}